\documentclass[10pt,twocolumn,letterpaper]{article}

\usepackage[pagenumbers]{iccv} %

\usepackage{algorithm,amsfonts,bm,microtype,nicefrac,url,wrapfig,xcolor}
\usepackage{algorithmic}
\usepackage{mdframed}
\usepackage{tikz}
\usepackage{comment}
\usetikzlibrary{shapes.geometric}
\usetikzlibrary{shapes.misc, positioning}
\usetikzlibrary{decorations.pathreplacing,calligraphy}

\usepackage{tcolorbox}
\usepackage{pifont}

\usepackage[accsupp]{axessibility}
\newcommand{\tabfigure}[2]{\raisebox{-.5\height}{\includegraphics[#1]{#2}}}

\usepackage{array}
\newcolumntype{L}{>{\centering\arraybackslash}m{1.7cm}}

\definecolor{lightgraytable}{rgb}{0.86, 0.86, 0.86}
\definecolor{ForestGreen}{RGB}{58, 149, 102}
\definecolor{WrongChoiceRed}{RGB}{199, 54, 41}

\newcommand{\ziplora} {ZipLoRA }
\newcommand{\ours} {LoRA.rar }
\newcommand{\ourmethod} {LoRA.rar}
\newcommand{\ourmetric} {MARS$^2$}
\newcommand{\R}{\mathbb{R}}
\newcommand{\sm}{\textit{Supp.\ Mat}}

\definecolor{iccvblue}{rgb}{0.21,0.49,0.74}
\usepackage[pagebackref,breaklinks,colorlinks,allcolors=iccvblue]{hyperref}

\title{LoRA.rar: Learning to Merge LoRAs via Hypernetworks for\\ Subject-Style Conditioned Image Generation}

\author{Donald Shenaj$^{\diamondsuit,\spadesuit,}\thanks{Research completed during internship at Samsung R\&D Institute UK.\\ Project page: \url{https://donaldssh.github.io/LoRA.rar}.}$ ~~~~ Ondrej Bohdal$^\diamondsuit$ ~~~~ Mete Ozay$^\diamondsuit$ ~~~~ Pietro Zanuttigh$^\spadesuit$ ~~~~ Umberto Michieli$^\diamondsuit$ \\
$^\diamondsuit$Samsung R\&D Institute UK (SRUK)
~~~~
$^\spadesuit$University of Padova 
}

\begin{document}
\maketitle

\begin{abstract}
Recent advancements in image generation models have enabled personalized image creation with both user-defined subjects (content) and styles. Prior works achieved personalization by merging corresponding low-rank adapters (LoRAs) through optimization-based methods, which are computationally demanding and unsuitable for real-time use on resource-constrained devices like smartphones. To address this, we introduce \ourmethod, a method that not only improves image quality but also achieves a remarkable speedup of over $4000\times$ in the merging process. We collect a dataset of style and subject LoRAs and pre-train a hypernetwork on a diverse set of content-style LoRA pairs, learning an efficient merging strategy that generalizes to new, unseen content-style pairs, enabling fast, high-quality personalization. Moreover, we identify limitations in existing evaluation metrics for content-style quality and propose a new protocol using multimodal large language models (MLLMs) for more accurate assessment. Our method significantly outperforms the current state of the art in both content and style fidelity, as validated by MLLM assessments and human evaluations.

\end{abstract}

\section{Introduction}
\label{sec:intro}

The advent of text-to-image generation models based on denoising diffusion \cite{ho2020denoising} allowed for significant improvements in output quality. Furthermore, recently there has been growing interest in personalized image generation \cite{ruiz2023dreambooth,shah2024ziplora}, where users can generate images that depict particular subjects or styles by providing just a few reference images.

\begin{figure}[t]
    \centering
    \includegraphics[width=\linewidth]{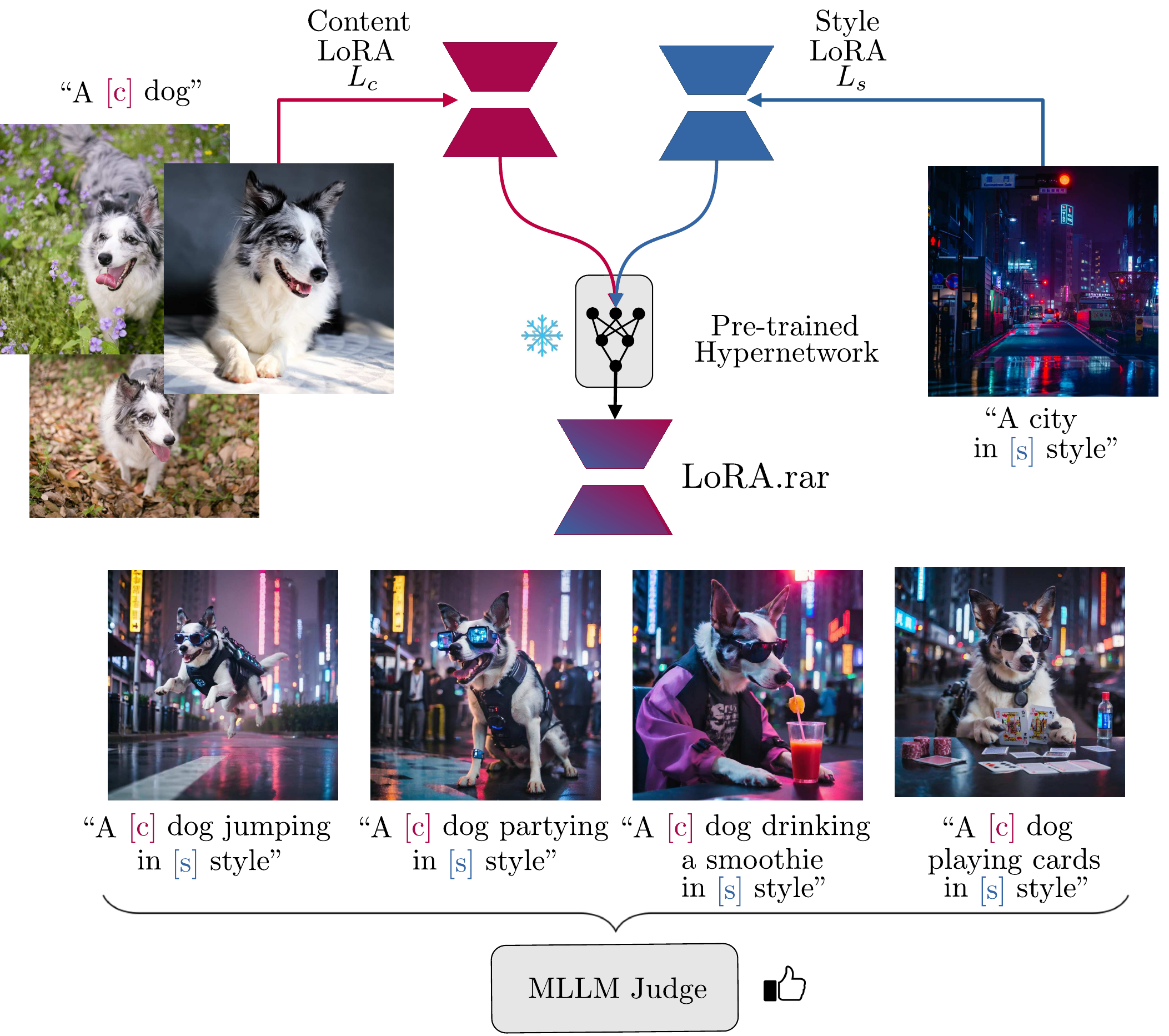}
    \caption{
    {
    We address the problem of joint content-style image generation by combining content and style LoRAs. Our method, \textbf{\ours}\!, uses a hypernetwork to dynamically predict the merging coefficients needed to combine content and style LoRAs. This enables high-quality, real-time LoRA merging. To evaluate the quality of the generated images, we propose a new MLLM protocol, which judges the fidelity of both content preservation and style transfer. The figure shows sample outputs generated by \textbf{\ours}\!.
    }
    }
    \label{fig:graphical_abstract}
\end{figure}

A key enabler of this personalization breakthrough is LoRA (Low-Rank Adapter), a parameter-efficient adaptation module introduced in \cite{hu2021lora}, which enables high-quality efficient personalization using only a small number of training samples.
This innovation has spurred extensive model sharing on open-source platforms like Civitai \cite{civitai} and Hugging Face \cite{huggingface}, making pre-trained LoRA parameters (LoRAs) readily available. 
The accessibility of these models has fueled interest in combining them to create images of personal subjects in various styles. For instance, users might apply a concept (\ie, subject) LoRA trained on a few photos of their pet, and combine it with a downloaded style LoRA to render their pet in an artistic style of their choice.

Averaging LoRAs can work acceptably when subject and style share significant visual characteristics, but fine-tuning of  merging coefficients (\ie, coefficients used to combine LoRAs) is typically required for more distinct subjects and styles. ZipLoRA \cite{shah2024ziplora} introduces an approach that directly optimizes merging coefficients through a customized objective function tailored to each subject-style LoRA combination. 
However, ZipLoRA's reliance on optimization for each new combination incurs a substantial computational cost, typically taking minutes to complete. This limitation restricts its practicality for real-time applications on resource-constrained devices like smartphones.
Achieving comparable or superior quality with respect to ZipLoRA while enabling real-time merging (\ie, in under a second) would make such technology far more accessible for deployment on resource-contraint devices.

In this paper, we introduce a method named \ourmethod\ for training a hypernetwork to learn merging coefficients for arbitrary subject and style LoRAs. Our hypernetwork is pre-trained on a curated dataset of LoRAs. During deployment, it generalizes to unseen subject-style combinations, generating merging coefficients instantly via a single forward pass and thus removing the need for retraining.
In cases where users prefer to input images rather than LoRAs, image-to-LoRA encoding methods like DiffLoRA \cite{wu2024difflora} can transform reference images into LoRAs. An overview of our approach and examples of generated images are shown in Fig.~\ref{fig:graphical_abstract}.

Our main contributions are as follows: 
\begin{enumerate}
  \setlength{\itemsep}{0em}
    \item We curate a dedicated dataset of LoRA weights, revealing their potential as a novel and valuable data modality for model merging.
    \item We propose \ourmethod, a novel method which pre-trains a small 0.5M-parameters hypernetwork to predict merging coefficients for arbitrary subject-style LoRAs.
    \item \ourmethod\ offers a fast and lightweight solution for merging subject-style LoRAs to generate images of any subject in any style. It generalizes seamlessly to unseen subject-style combinations at test time without requiring fine-tuning, unlike ZipLoRA. 
    \item We analyze the limitations of existing metrics (CLIP-I, CLIP-T, DINO) in assessing the fidelity of joint subject-style generation. To address this, we introduce \ourmetric, a new metric based on Multimodal Large Language Models (MLLMs), which aligns closely with user preferences and enables scalability 
    of quantitative studies.
    \item We show that \ourmethod\ consistently outperforms existing merging strategies 
    in subject-style personalization. 
\end{enumerate}

\begin{figure*}[t]
    \centering
    \includegraphics[width=0.9\linewidth, trim=0cm 0.15cm 0cm 0.2cm, clip]{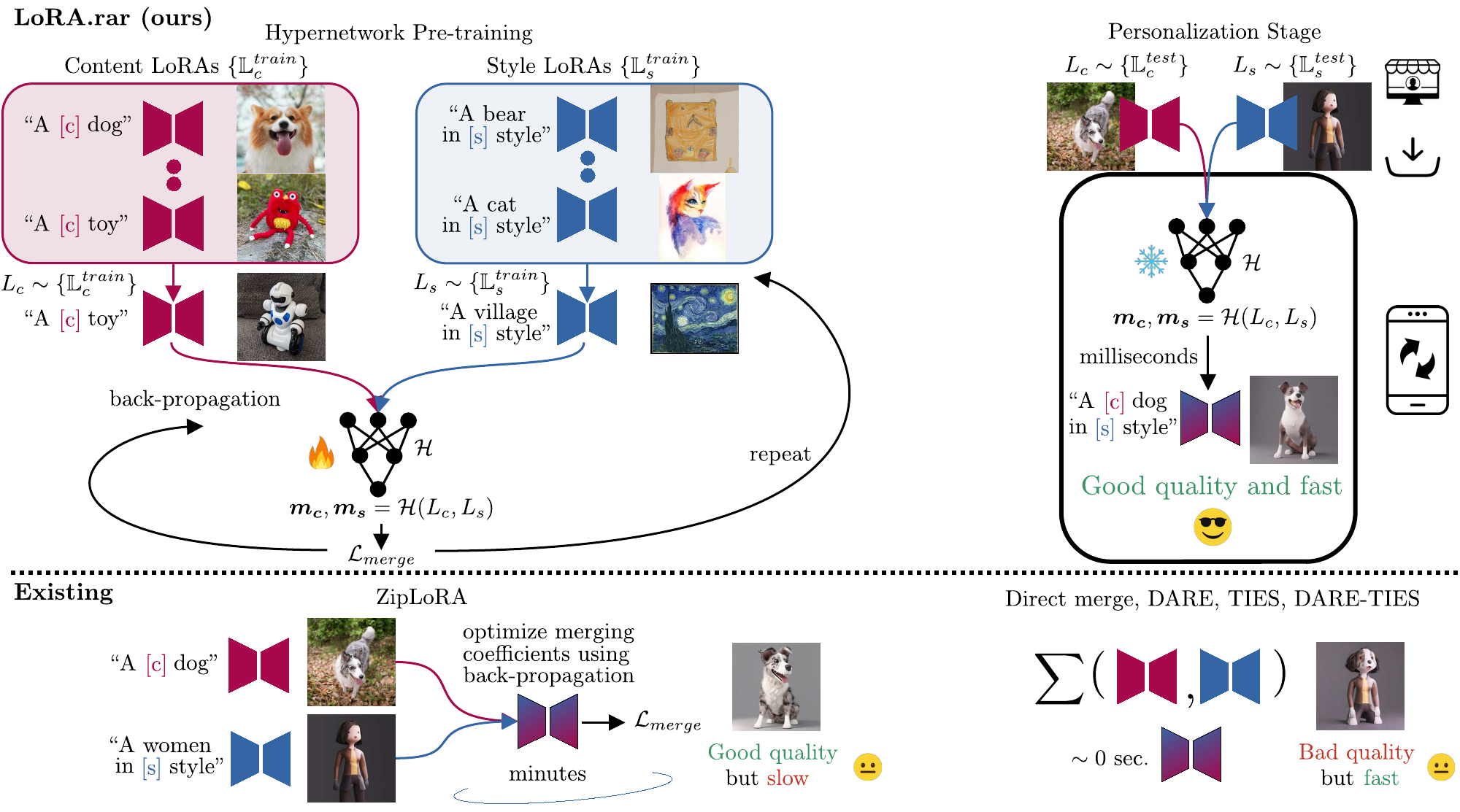}
    \caption{\textbf{Method Overview.} \ours pre-trains a hypernetwork that dynamically generates merging coefficients for new, unseen content-style LoRA pairs at deployment. In contrast, existing solutions are limited by either costly test-time training, as with ZipLoRA, or produce lower-quality outputs, as with conventional merging strategies.}
    \label{fig:overall_schematic}
\end{figure*}

\section{Related Work}

\noindent \textbf{Subject-Conditioned Image Generation} has been extensively explored in recent years.
DreamBooth \cite{ruiz2023dreambooth} fine-tunes the entire generative model on reference images for subject fidelity. 
Several techniques have been proposed to mitigate extensive training.
Textual Inversion \cite{gal2022image} optimizes token embeddings for subject encoding, with extensions such as \cite{voynov2023p+, arar2023domain, hao2023vico, yang2023controllable, tewel2023key} enhancing flexibility. %
However, these methods face limitations in scaling to multiple concepts.
Another line of work optimize specific network parts or employing specialized tuning, such as CustomDiffusion \cite{kumari2022customdiffusion}, LoRA \cite{hu2021lora,cloneofsimo,wu2024difflora}, SVDiff \cite{han2023svdiff}, and DreamArtist \cite{dong2022dreamartist}. In particular, LoRA has gained popularity for its training efficiency and adaptability to multiple concepts, making it a widely used approach for subject conditioning.

More recently, techniques for zero-shot personalization aim to avoid fine-tuning by either (i) using separate conditioning encoders (encoder-based approaches) \cite{Wei2023ELITEEV, gal2023encoder, ma2023subject, ye2023ip, patel2024lambda, li2023blip, pan2023kosmos, zhou2024customization, chen2023subject, zhou2024toffee}; or (ii) utilizing features from the generative model’s backbone to guide generation (encoder-free methods) \cite{zeng2024jedi, purushwalkam2024bootpig, aiello2025dreamcache}. However, these methods often require extensive additional storage, limiting their applicability in resource-constrained environments. %

\noindent \textbf{Subject- and Style-Conditioned Image Generation.} Beyond subject conditioning, many works tackle style-conditioned generation, such as StyleGAN \cite{karras2019style}, StyleDrop \cite{styledrop} and DreamArtist \cite{dong2022dreamartist}. However, these methods lack the ability to handle both subject and style conditioning jointly. 
Recent approaches addressing this challenge include CustomDiffusion \cite{kumari2022customdiffusion}, which learns multiple concepts through expensive joint training but struggles to disentangle style from subject, and HyperDreamBooth \cite{ruiz2024hyperdreambooth}, which generates personal subjects with good style editability via textual prompt. 
B-LoRA \cite{frenkel2025implicit} proposes a layer-wise LoRA tuning pipeline for either content or style.
Notably, ZipLoRA \cite{shah2024ziplora}, the closest reference for our work, merges pre-trained subject and style LoRAs via test-time optimization to discover the optimal merging coefficients. 
Concurrent work such as RB-Modulation \cite{rout2024rb} uses a style descriptor for modulation without LoRAs.
FreeTuner \cite{xu2024freetuner} and Break-for-Make \cite{xu2024break} further explore style disentanglement with separate content and style encoders or training subspaces. 

Our method focuses on efficient zero-shot merging of subject and style LoRAs aiming for high-quality subject preservation while preserving text editability.

\noindent \textbf{Model Merging} is an increasingly popular way to enhance the abilities of foundational models in both language \cite{hammoud2024model} and vision domains \cite{yang2024model}.
The simplest technique is direct arithmetic merge that averages the weights of multiple fine-tuned models \cite{wortsman2022model}. Despite its simplicity, it can improve performance and enable multi-tasking to at least a certain extent \cite{ilharco2023editing}. Following \cite{wortsman2022model}, diverse strategies have been proposed. %
TIES \cite{yadav2024ties} mitigates interference between parameters of different models due to different signs, while
DARE \cite{yu2024language} drops some weights and rescales the remaining ones to reduce redundancy and interference.
DARE-TIES \cite{goddard2024arcee} combines DARE and TIES, and demonstrates successful merging in complex scenarios.

\noindent \textbf{LoRA Merging for Image Generation} 
has recently gained attention. 
Mix-of-Show \cite{gu2023mixofshow} and LoRA-Composer \cite{yang2024lora} merge the concept of each LoRA in the output image for multi-concept generation (instead of subject-style).
Further, they require a custom version of LoRAs, hindering wide compatibility.
ZipLoRA \cite{shah2024ziplora} merges standard LoRAs, focusing on subject-style generation through parameter optimization.
However, this approach requires several minutes per merge at test time, limiting its usability in real-time scenarios. 
Our \ours builds upon ZipLoRA’s foundations, targeting both a more efficient solution and improved results.

\noindent \textbf{Hypernetworks,} or networks generating the weights of other networks \cite{ha2017hypernetworks, chauhan2024brief}, have found diverse use cases.
Hypernetworks are used to generate LoRA weights in \cite{ceritli-etal-2024-study, ruiz2024hyperdreambooth}, while \cite{bartholet2024non} uses them for model aggregation in federated learning. 
In contrast to previous approaches, we design a lightweight hypernetwork that takes any subject-style LoRA pair as input and predicts the merging coefficients for their combination, enabling efficient and high-quality joint subject-style personalization with no optimization overhead at test time.

\section{Method}
\label{sec:method}

Our objective is to design and train a hypernetwork that predicts weighting coefficients to merge content and style LoRAs. Using a set of LoRAs, we train this hypernetwork to produce suitable merging coefficients for unseen content and style LoRAs at the deployment stage.

We start by formulating the problem in Sec.~\ref{sec:problem_formulation}, detailing how LoRAs are applied to the base model and outlining the limitations of the current state-of-the-art approach. In Sec.~\ref{sec:method:dataset}, we describe the construction of the LoRA dataset used to train and evaluate our solution. Sec.~\ref{sec:method:structure} discusses the structural design of our hypernetwork, followed by an overview of the training procedure in Sec.~\ref{sec:method:training}.

\subsection{Problem Formulation}
\label{sec:problem_formulation}

We use a pre-trained image generation diffusion model $\mathcal{D}$ with weights $W_0$ and LoRA $L$ with weight update matrix $\Delta W$. For simplicity, we consider one layer at a time.
A model $\mathcal{D}$ that uses a LoRA $L$ is denoted as $\mathcal{D}_{L} = \mathcal{D} \oplus L$ with weights $W_0 + \Delta W$, where operation $\oplus$ means we apply LoRA $L$ to the base model $\mathcal{D}$. 
To specify content and style, we use LoRAs $L_c$ (content) and $L_s$ (style) with respective weight update matrix $\Delta W_c$ and $\Delta W_s$. 
Our objective is to merge $L_c$ and $L_s$ into $L_m$, producing a matrix $\Delta W_m$ that combines content and style coherently in generated images. The merging operation can vary, from simple averaging to advanced techniques like ZipLoRA's and ours.

\begin{figure}[t]
    \centering
    \includegraphics[width=\linewidth, trim=0cm 0.25cm 0cm 0.25cm, clip]{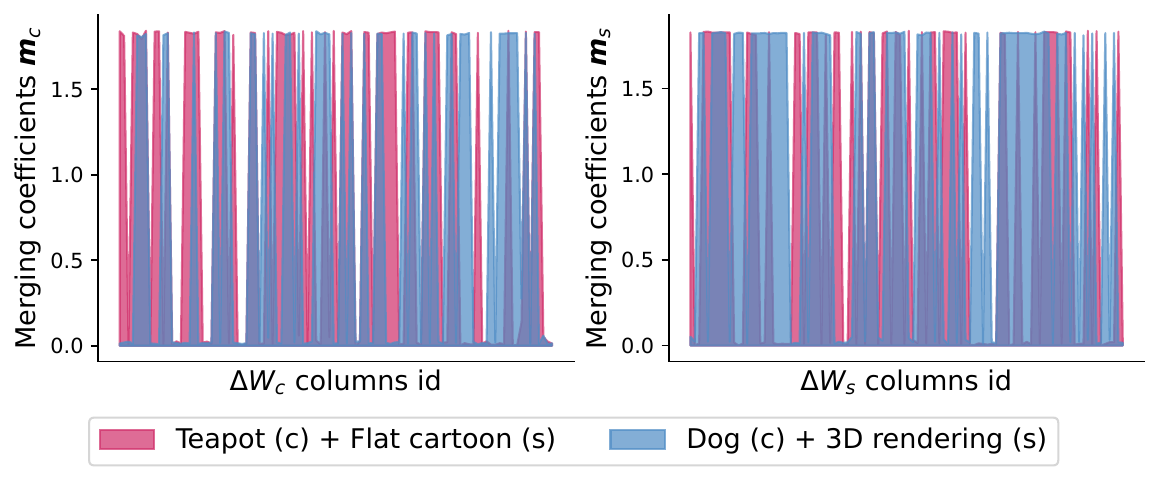}
    \caption{\textbf{ZipLoRA's Merging Coefficients} $\boldsymbol{m}_c, \boldsymbol{m}_s$ for randomly selected columns of the LoRA weight update matrices. The coefficients are visibly different for various combinations of content and style scenarios, showing the need for adaptive solutions.}
    \label{fig:ziplora_coefficients}
\end{figure}

ZipLoRA takes a gradient-based approach, learning column-wise merging coefficients $\boldsymbol{m}_c$ and $\boldsymbol{m}_s$ for $\Delta W_c$ and $\Delta W_s$, respectively, as:
$\Delta W_m =  \boldsymbol{m}_c \otimes \Delta W_c + \boldsymbol{m}_s\otimes \Delta W_s$, where $\otimes$ represents element-by-column multiplication. %
Although ZipLoRA achieves high-quality results, it requires training these coefficients from scratch for each content-style pair, with distinct coefficients for different combinations, as shown in Fig.~\ref{fig:ziplora_coefficients}. With ZipLoRA performing 100 gradient updates per pair, real-time performance is unfeasible, particularly on resource-constrained devices.

Our goal is to outperform ZipLoRA's image quality while accelerating merging coefficient generation time by orders of magnitude for unseen content-style pairs. To accomplish this, we pre-train a hypernetwork that predicts adaptive merging coefficients on the fly, enabling fast, high-quality merging in a single feed-forward pass. An overview is shown in Fig.~\ref{fig:overall_schematic}.

\subsection{LoRA Dataset Generation}
\label{sec:method:dataset}

To train our hypernetwork, we first build a dataset of LoRAs. Content LoRAs are trained on individual subjects from the DreamBooth dataset \cite{ruiz2023dreambooth}, and style LoRAs are trained on various styles from the StyleDrop / ZipLoRA datasets \cite{styledrop, shah2024ziplora}. Each LoRA is generated via the DreamBooth protocol.

We split the LoRA dataset into training $\{\mathbb{L}^{train}_c\}, \{\mathbb{L}^{train}_s\}$, validation $\{\mathbb{L}^{val}_c\}, \{\mathbb{L}^{val}_s\}$, and test $\{\mathbb{L}^{test}_c\}, \{\mathbb{L}^{test}_s\}$ sets. 
During training and evaluation, we sample content-style LoRA pairs. 
The hypernetwork is trained on the training sets, with hyperparameters and design choices tuned on the validation sets. The test sets are reserved to assess performance on novel content-style pairs.

\subsection{Hypernetwork Structure}
\label{sec:method:structure}

\begin{figure}
    \centering
    \resizebox{0.48\textwidth}{!}{%
\begin{tikzpicture}[font=\large]
\draw[line width=1pt] (0,0) rectangle (3,1.6);
\draw[fill=lightgray, line width=0.5pt] (0.6,0) rectangle (0.8,1.6);
\node[anchor=west] at (1.2,0.8) {$\Delta W_c$};
\node[anchor=west] at (-1.2,0.8) {Rows};
\draw[line width=1pt] (0,-0.4) rectangle (3, -2);
\draw[fill=lightgray, line width=0.5pt] (0.6, -0.4) rectangle (0.8, -2);
\node[anchor=west] at (0.5 ,-2.3) {$i$};

\node[anchor=west] at (1.2, -1.2) {$\Delta W_s$};
\node[anchor=west] at (-1.2, -1.2) {Rows};
\draw[line width=1pt] (4,-1.7) rectangle (5.6, 1.3);
\draw[fill=lightgray, line width=0.5pt] (4,0.7) rectangle (5.6, 0.5);

\node[anchor=west] at (4.5, 1) {$w_c^i$};

\node[anchor=west] at (4.3, -0.2) {$\Delta W_c$};
\draw[line width=1pt] (5.6,-1.7) rectangle (7.2, 1.3);
\draw[fill=lightgray, line width=0.5pt] (5.6,0.7) rectangle (7.2, 0.5);
\node[anchor=west] at (6, 1) {$w_s^i$};

\draw [line width=1.2pt, decorate,
	decoration = {calligraphic brace, mirror, amplitude=9pt}] (3.4, -2) --  (3.4,1.6);

\node[anchor=west] at (0.9 ,-2.5) {Columns};
\node[anchor=west] at (5.9, -0.2) {$\Delta W_s$};
\node[anchor=west] at (0.6, 2.3) {Content and Style LoRAs to merge};

\node at (11, 1) {Hypernetwork $\mathcal{H}$};
\node[rounded rectangle, draw, fill=teal!20] (input1)  at (9.5, 0.1)  {Input Layer 1};
\node[rounded rectangle, draw, fill=teal!20] (input2) at (12.5, 0.1) {Input Layer 2};
\node[rounded rectangle, draw, fill=teal!20] (output) at (11, -1) {Output Layer};

\node[anchor=west] at (4.6, -2.1) {Mini-batch};

\node (coeff) at (11, -2) {$\boldsymbol{m}_c, \boldsymbol{m}_s $ for each $i$};
\draw[->, line width=1pt] (input1) -- (output);
\draw[->, line width=1pt] (input2) -- (output);
\draw[->, line width=1pt] (output) -- (coeff);
\draw[line width=1pt] (7.2,0.2) rectangle (7.8,0.2);
\draw[line width=1pt] (7.8,0.2) rectangle (7.8,0.8);
\draw[line width=1pt] (7.8,0.8) rectangle (9.5,0.8);
\node[coordinate] (a) at (9.5,0.8) {};
\draw[->, line width=1pt] (a) -- (input1);
\end{tikzpicture}
}
    \vspace{-1.5em}
    \caption{\textbf{Hypernetwork Structure.} 
    The hypernetwork has multiple input layers, each matching the dimensionality of corresponding layers in the generative model. Here, an example layer with dimensions matching \textit{Input Layer 1} is shown. Content and style LoRAs are concatenated and input into the hypernetwork, which then predicts the columnwise merging coefficients for each specified layer.    
    }
    \label{fig:hypernet_overview}
\end{figure}
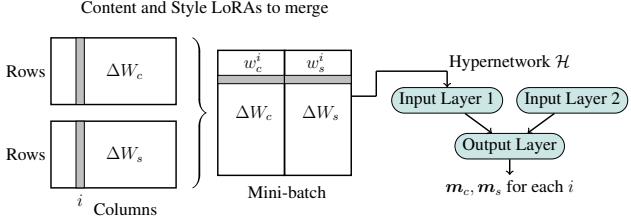

Our hypernetwork, $\mathcal{H}$, takes two LoRA update matrices as inputs: $\Delta W_c \in \R^{m \times n}$ for content and $\Delta W_s\in \R^{m \times n}$ for style, and predicts column-wise merging coefficients $\boldsymbol{m}_c\in \R^{n}$ and $\boldsymbol{m}_s\in \R^{n}$. 
Given the high dimensionality of each update matrix, flattening them directly as input would be impractical. To address this, we assume that the merging coefficient for each column can be predicted independently.

For each column $i$, we extract the respective content and style columns, $\boldsymbol{w}_c^i\!\!=\!\!\Delta W_c[:,i]$ and $\boldsymbol{w}_s^i\!\!=\!\!\Delta W_s[:,i]$, and concatenate them as $[\boldsymbol{w}_c^i, \boldsymbol{w}_s^i]\in \R^{2m}$ to form the input features for the hypernetwork. 
We treat different columns as a minibatch, allowing for efficient parallel processing. The full  hypernetwork input is thus ${\text{concat}(\Delta W_c^\intercal, \Delta W_s^\intercal, \text{dim}=1) \in \R^{n \times 2m}}$.

To accommodate the various LoRA matrix sizes within the diffusion model $\mathcal{D}$, we designed $\mathcal{H}$ with separate input layers tailored to each unique matrix size,
each mapped to a shared hidden dimension. In our case, the hypernetwork uses two input layers with ReLU non-linearities and a shared output layer to predict merging coefficients for each column.

Since different rows are treated as a mini-batch, overall the hypernetwork outputs $2n$ coefficients, one for each column of content and style LoRAs:
\begin{equation}
    \boldsymbol{m}_c, \boldsymbol{m}_s = \mathcal{H}(L_c, L_s).
\end{equation}
These coefficients are used to merge the LoRAs $L_c$ and $L_s$, resulting in the merged LoRA $L_m$ with update matrix $\Delta W_m$:
\begin{equation}
  \Delta W_m = \boldsymbol{m}_c \otimes \Delta W_c  + \boldsymbol{m}_s \otimes \Delta W_s.
  \label{eq:merge}
\end{equation}
Fig.~\ref{fig:hypernet_overview} provides an overview of how content and style LoRAs predict merging coefficients. Notably, we apply hypernetwork-guided merging for query and output LoRAs, while we use simple averaging for key and value LoRAs. This configuration empirically outperformed other tested options, as detailed in the \sm.

\subsection{Hypernetwork Training}
\label{sec:method:training}

We train the hypernetwork $\mathcal{H}$ by sampling content-style LoRA pairs from the training set $\{\mathbb{L}^{train}_c\}, \{\mathbb{L}^{train}_s\}$. The hypernetwork generates merging coefficients, which are then used to compute a merging loss $\mathcal{L}_{merge}$ that updates the weights of $\mathcal{H}$. 
We discover that the merging loss $\mathcal{L}_{merge}$ of \cite{shah2024ziplora}, which was originally proposed to
optimize the merging coefficients for
a specific subject-style LoRA pair at test time, could be 
repurposed more effectively to optimize the weights of the hypernetwork $\mathcal{H}$ instead. 
This novel application 
produces better merging coefficients and promotes generalization to any new subject-style LoRA pair. The merging loss includes terms that ensure both content and style fidelity, while also encouraging orthogonality between content and style merging coefficients. Specifically, $\mathcal{L}_{merge}$ is defined as:
\begin{align}
\mathcal{L}_{merge} =& \| (\mathcal{D} \oplus L_m)(\boldsymbol{x}_c, \boldsymbol{p}_c) - (\mathcal{D} \oplus L_c)(\boldsymbol{x}_c, \boldsymbol{p}_c)\|_2  \nonumber \\ 
    +& \| (\mathcal{D} \oplus L_m)(\boldsymbol{x}_s, \boldsymbol{p}_s) - (\mathcal{D} \oplus L_s)(\boldsymbol{x}_s, \boldsymbol{p}_s)\|_2 \nonumber \\
    + &\lambda |\boldsymbol{m}_c \cdot \boldsymbol{m}_s|,
\label{eq:optfn}
\end{align}
where $\boldsymbol{x}_c, \boldsymbol{x}_s$ are the noisy latents, and $\boldsymbol{p}_c, \boldsymbol{p}_s$ are the text prompts for content and style reference images respectively \cite{shah2024ziplora}. The term $\lambda$ controls the strength of the orthogonality-promoting regularization term.

The training process is formalized in Algorithm~\ref{alg:hypertraining}.
Architecture choices for the hypernetwork, as well as hyperparameters, are optimized on the validation set. The final evaluation is conducted on the test set (which includes unknown subjects and styles substantially different from the training ones), when the hypernetwork $\mathcal{H}$ simply predicts the merging coefficients for new content and style LoRAs $L_c \sim \{\mathbb{L}^{test}_c\}, L_s \sim \{\mathbb{L}^{test}_s\}$.

\begin{algorithm}[t]
\small
\linespread{1.2}\selectfont
\caption{Hypernetwork training.} %
\label{alg:hypertraining}
\vspace{.25em}
\begin{algorithmic}[1]
    \REQUIRE{\# training steps $T$, learning rate $\eta$, base model $\mathcal{D}$, training dataset of content and style LoRAs $\{\mathbb{L}^{train}_c\}, \{\mathbb{L}^{train}_s\}$}
    \vspace{-1.4em} %
    \STATE \textbf{Initialize} hypernetwork $\mathcal{H}$
    \FOR{$t=1,\ldots,T$}
        \STATE Sample content and style LoRAs from the training set: ${L_c \sim \{\mathbb{L}^{train}_c\}, L_s \sim \{\mathbb{L}^{train}_s\}}$
        \STATE Predict merging coefficients $\boldsymbol{m}_c, \boldsymbol{m}_s = \mathcal{H}(L_c, L_s)$
        \STATE 
        Obtain merged LoRA $L_m$ with weight update matrix $\Delta W_m$ computed via Eq.~\eqref{eq:merge} %
        \STATE Compute $\mathcal{L}_{merge}$ using Eq.~\eqref{eq:optfn}
        \STATE Update $\mathcal{H}\leftarrow \mathcal{H} -\eta\nabla_{\mathcal{H}} \mathcal{L}_{merge}$
    \ENDFOR
\end{algorithmic}
\end{algorithm}

\section{Joint Subject-Style Evaluation Metrics}
\label{sec:method:metrics}

In this section we discuss how to evaluate personalized image generation methods across diverse subjects and styles.

\noindent \textbf{Limitations of Existing Metrics.}
Developing reliable metrics that align with user preferences is crucial for scaling text-to-image models, especially when direct feedback is unavailable. Metrics CLIP-I, CLIP-T, and DINO \cite{ruiz2023dreambooth} are widely used for single-concept personalization (\ie, personalizing to either a style or subject) in benchmarks such as DreamBooth~\cite{ruiz2023dreambooth}, DreamBench++~\cite{peng2024dreambench}, and ImagenHub~\cite{ku2024imagenhub}.

\begin{figure}[]
    \centering
    \footnotesize
    \begin{tabular}{c c c}
        Reference & Direct merge & \ours (Ours) \\
        \tabfigure{width=0.12\textwidth}{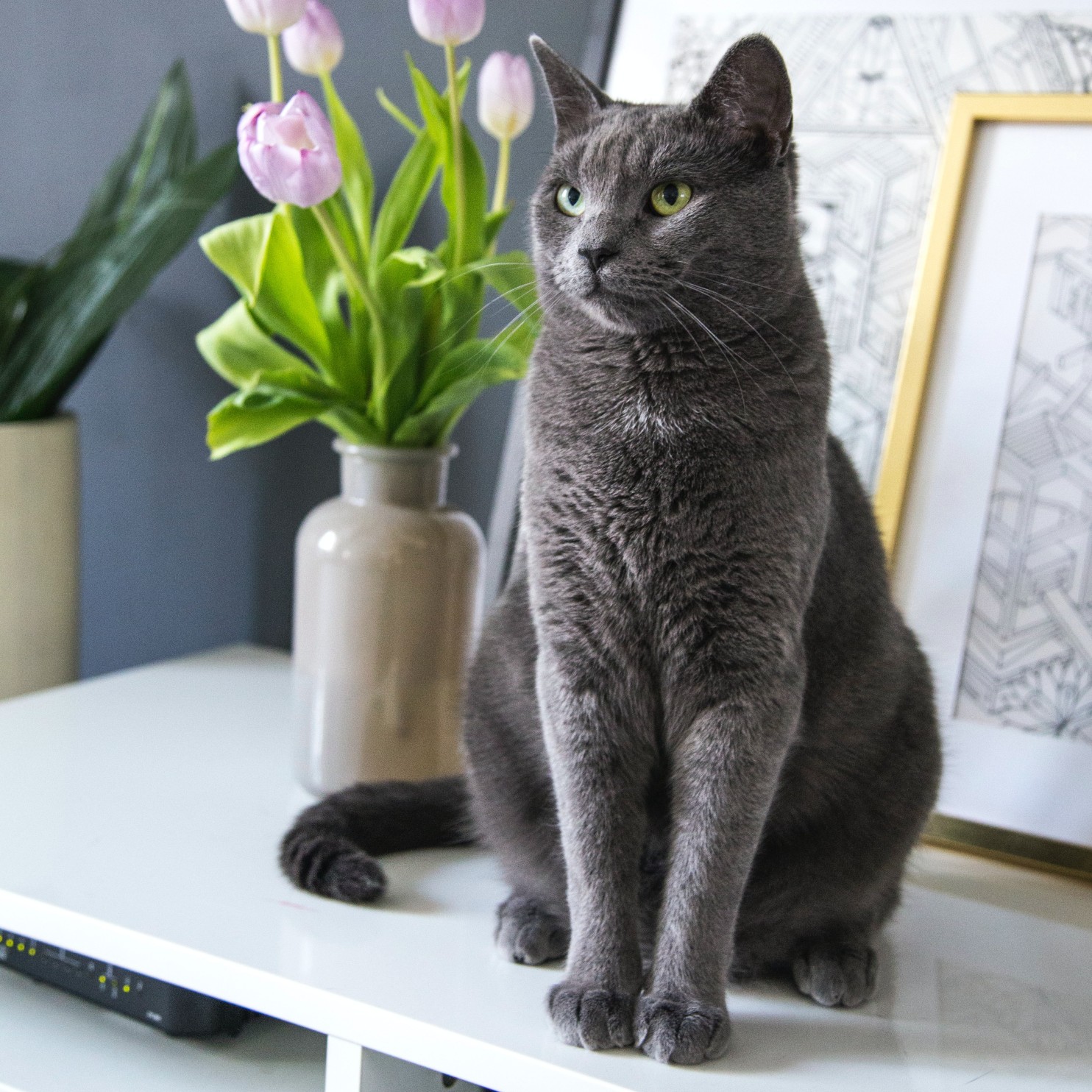}%
        \llap{\raisebox{0.02\textwidth}{%
    \includegraphics[height=0.05\textwidth]{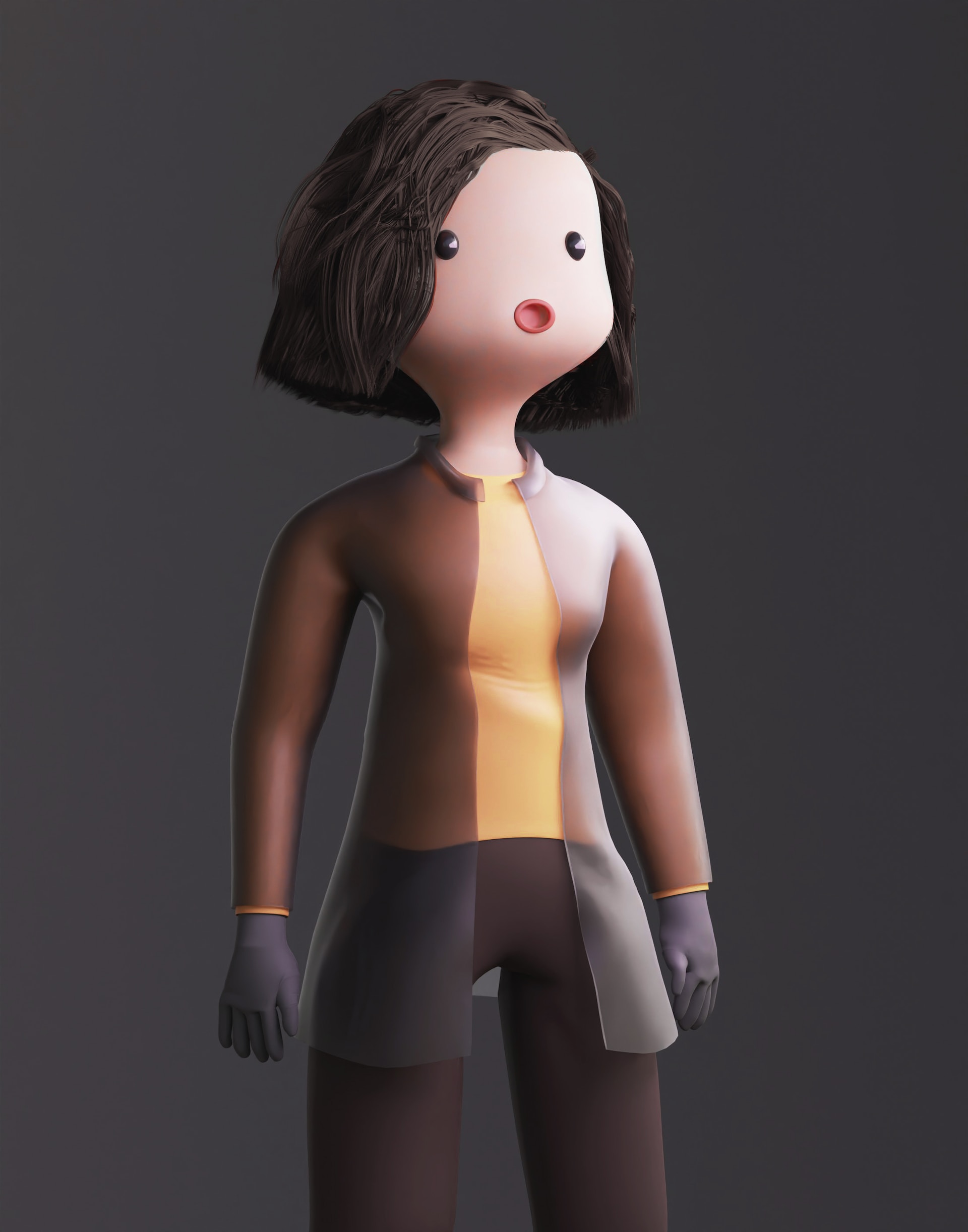}%
            }}
         & \fcolorbox{WrongChoiceRed}{white}{\tabfigure{width=0.12\textwidth}{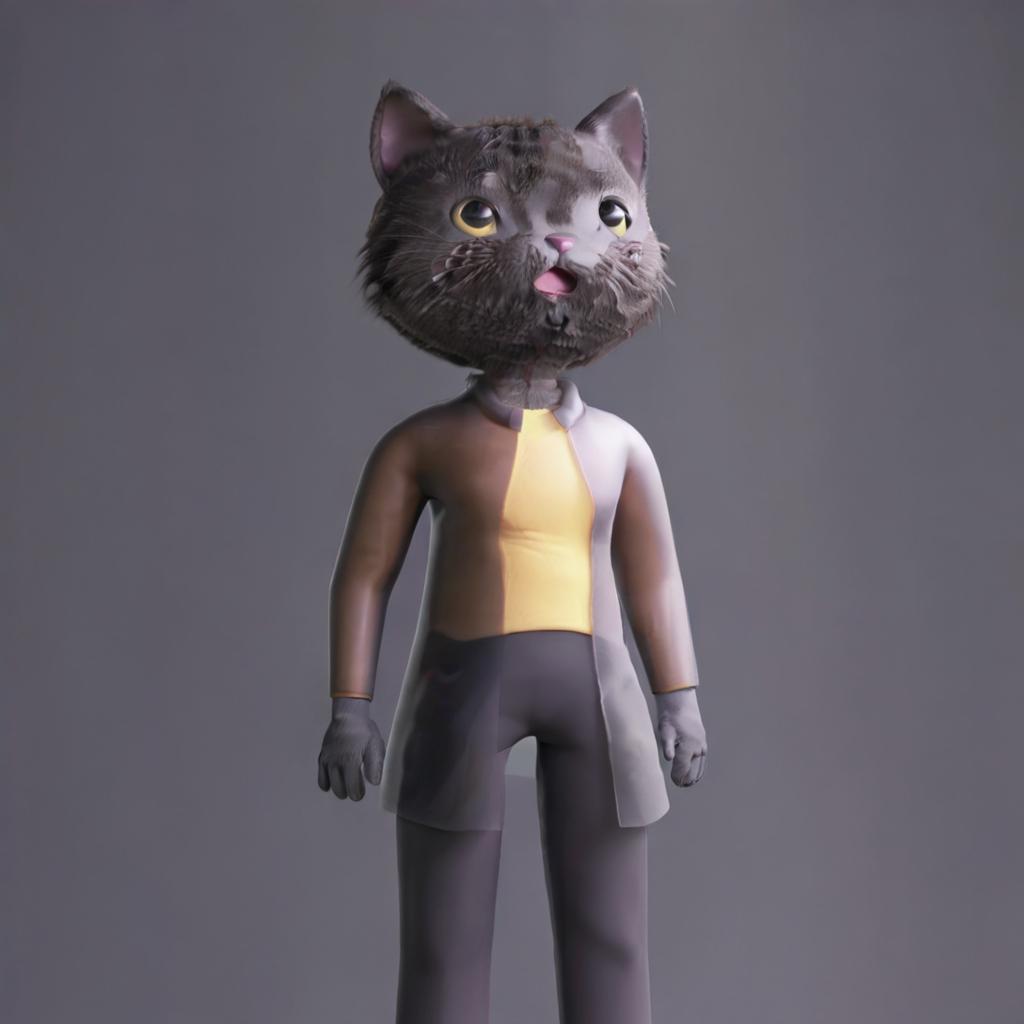}} &  \fcolorbox{ForestGreen}{white}{\tabfigure{width=0.12\textwidth}{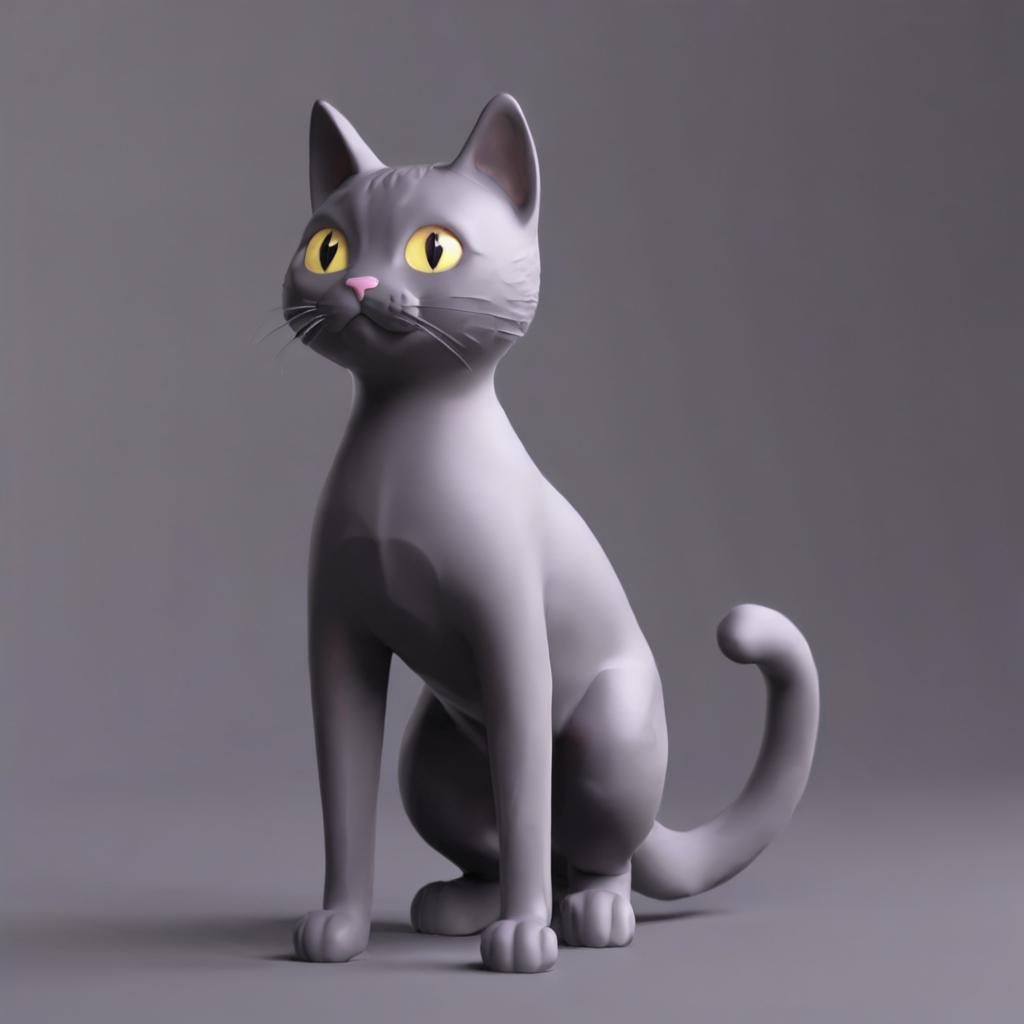}} \\
        &  {\scriptsize \textbf{CLIP-I: 0.817}}  & {\scriptsize \textbf{CLIP-I: 0.695}}  \\
        &  {\scriptsize DINO: 0.180} & {\scriptsize DINO: 0.293} \\
        &  {\scriptsize CLIP-T: 0.319}  & {\scriptsize CLIP-T: 0.334} \\

        \tabfigure{width=0.12\textwidth}{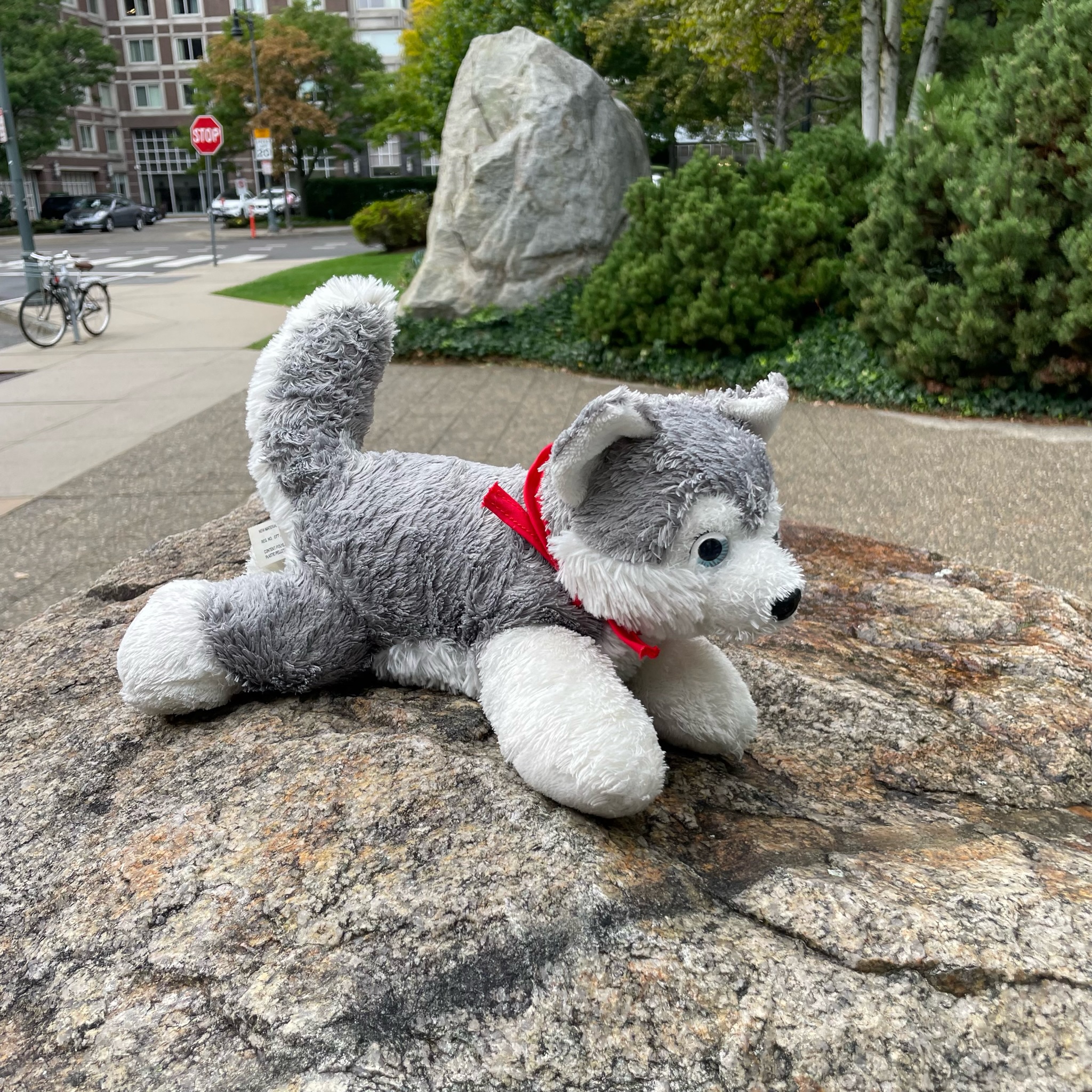}%
        \llap{\raisebox{0.02\textwidth}{%
              \includegraphics[height=0.05\textwidth]{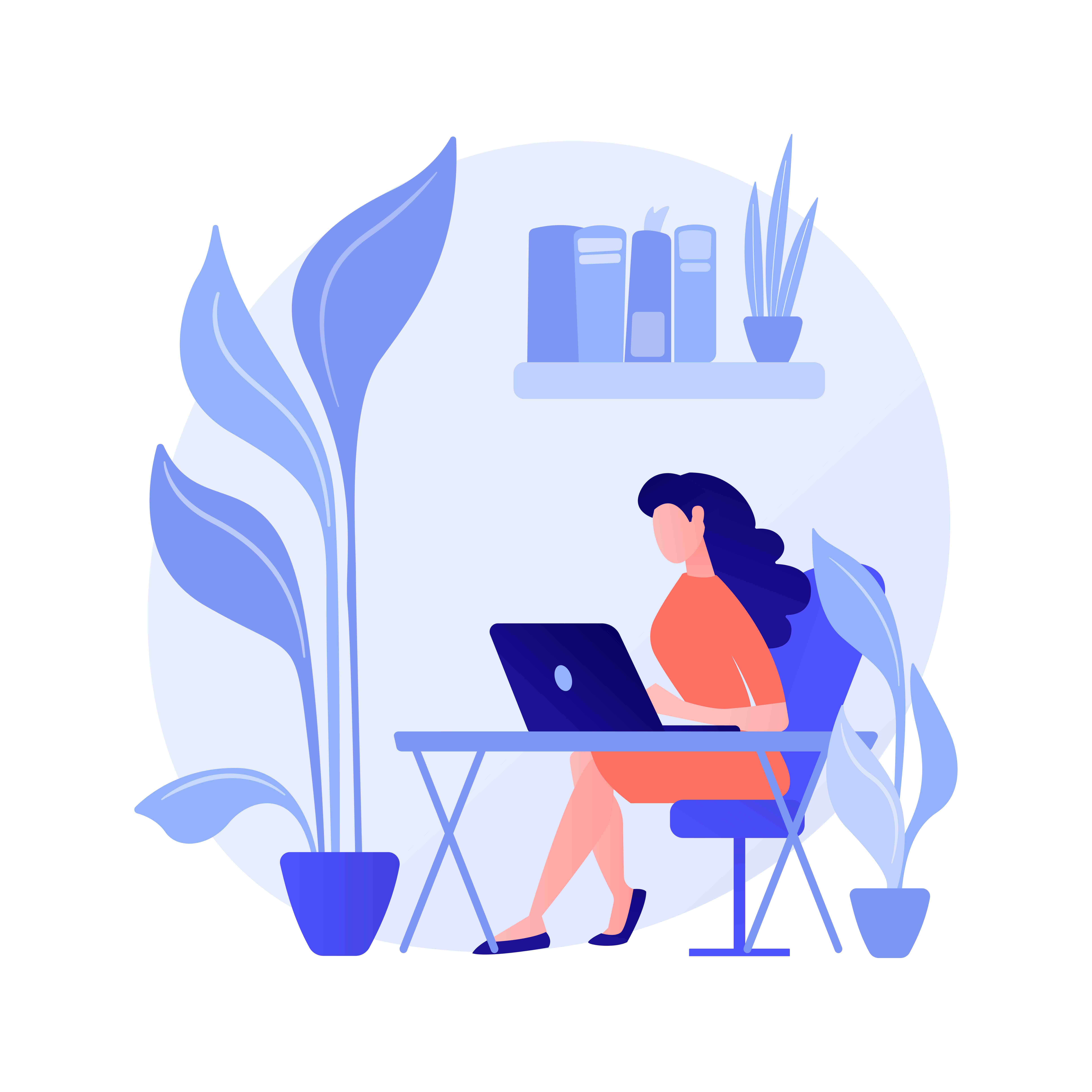}%
            }}
         & \fcolorbox{WrongChoiceRed}{white}{\tabfigure{width=0.12\textwidth}{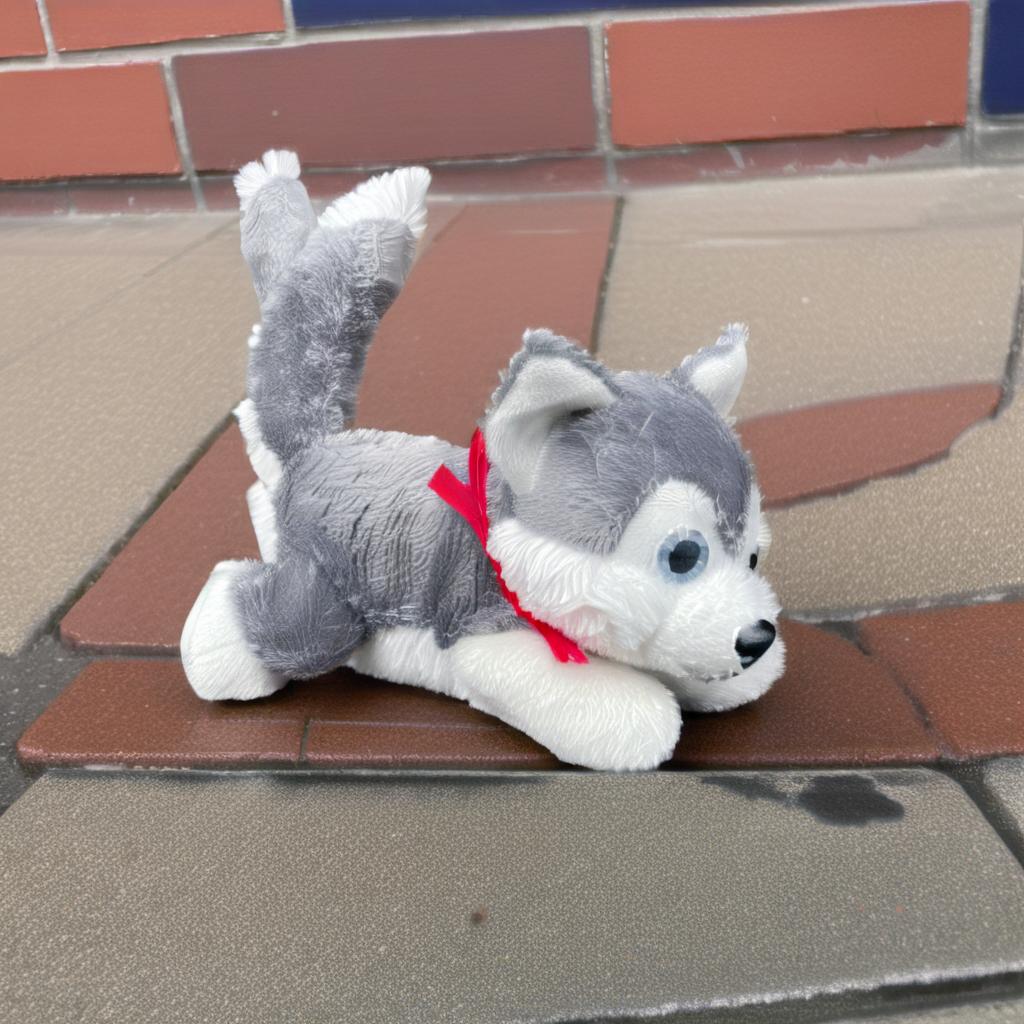}} 
         &  \fcolorbox{ForestGreen}{white}{\tabfigure{width=0.12\textwidth}{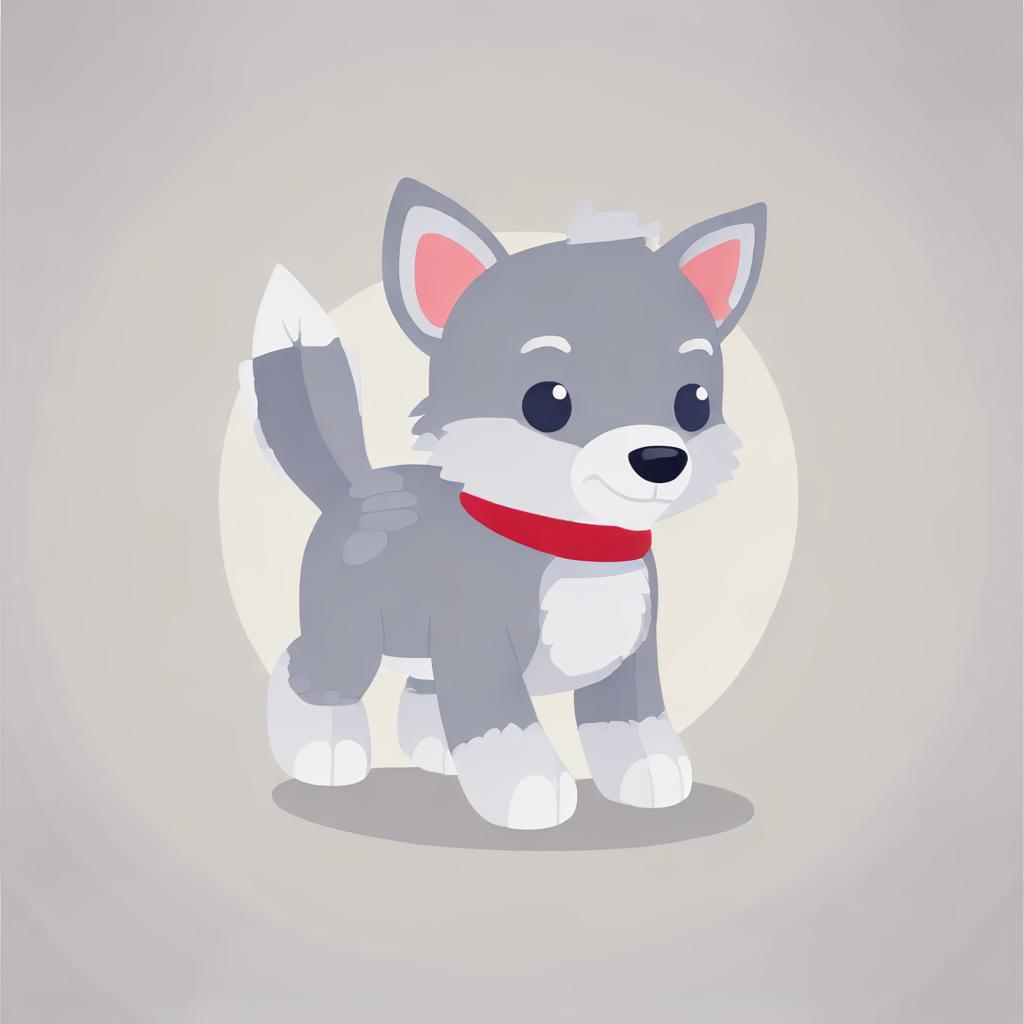}} \\
        &  {\scriptsize CLIP-I: 0.445} & {\scriptsize CLIP-I: 0.671} \\
        &  {\scriptsize \textbf{DINO: 0.801} } & {\scriptsize \textbf{DINO: 0.411}} \\
        &  {\scriptsize CLIP-T: 0.259 }& {\scriptsize CLIP-T: 0.298 }\\
    \end{tabular}
    \caption{\textbf{Limitation of Existing Metrics.} Top: CLIP-I is maximized when the style image (shown in the small upper right thumbnail) content is replicated. Bottom: DINO is maximized when the generated image has no style transfer.}
    \label{tab:metrics_failure}
\end{figure}

However, these metrics may not reliably evaluate joint subject-style personalization, as illustrated in Fig.~\ref{tab:metrics_failure}. 
Specifically, the CLIP-I score tends to favor style fidelity, often overlooking accurate representation of the subject (top of Fig.~\ref{tab:metrics_failure}), while the DINO score prioritizes the original subject replication overlooking stylistics integration (bottom of Fig.~\ref{tab:metrics_failure}).
CLIP-T, typically used for text alignment, supports subject recontextualization but is less suited to style-content prompts like ``\verb|A [c] <class name> in [s] style|''.
Here \verb|[c]| is a unique rare token identifier for content, \verb|<class name>| is the class name following \cite{ruiz2023dreambooth}, and \verb|[s]| is a short description of the style as in StyleDrop \cite{styledrop}.

\noindent \textbf{Evaluation via Multimodal Large Language Models (MLLMs).} 
To overcome the limitations of conventional metrics, we propose leveraging MLLMs for evaluation. LLMs have shown high effectiveness in evaluating text-based outputs \cite{zheng2023judging}, and their application has recently been extended to multimodal tasks involving both text and images \cite{chen2024mllm}. For example, recent works \cite{zeng2024can, zong2024vlicl} have successfully utilized MLLMs to determine whether generated images meet specified criteria, such as color or object presence.

\begin{figure}[t]
    \centering
    \includegraphics[width=\linewidth]{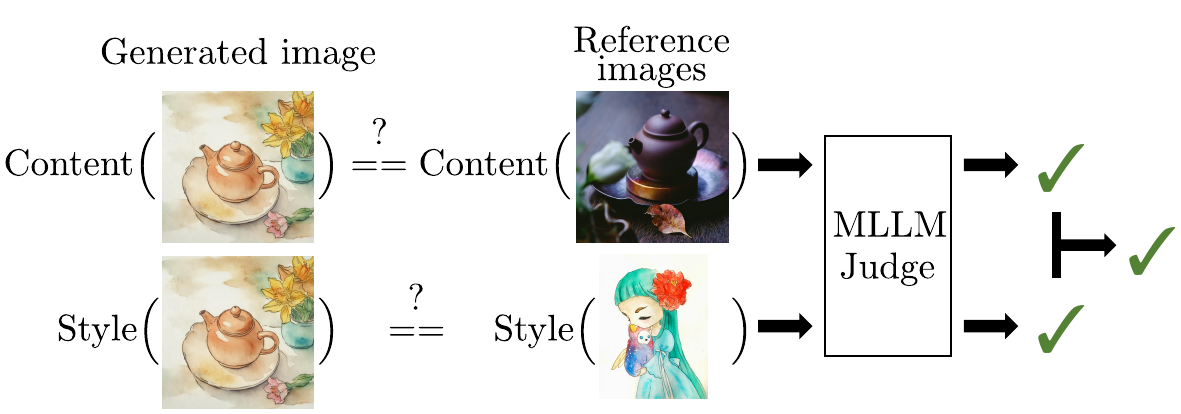}
    \caption{\textbf{Evaluation via MLLM Judge}.
    Generated images are checked separately for content and style. We mark the image as correct if both are approved.}
    \label{fig:mllm_eval}
\end{figure}

Among specialized MLLM judge models, LLaVA-Critic \cite{xiong2024llavacritic} stands out for its accuracy in assessing output quality in multimodal contexts. In this work, we use LLaVA-Critic to evaluate whether generated images accurately represent the intended subject (content) and style. 
Our protocol is as follows: the MLLM judge first assesses if each generated image meets the specified style and content independently. For clarity, we provide reference images for both style and content along with detailed evaluation prompts. Binary ratings are used for both style and content evaluations, with an image deemed correct, \ie, final score of 1, only if it fulfills both criteria, and 0 otherwise. When there are multiple reference images, we consider a generated image accurate if the MLLM model identifies it as correct for more than half of the reference images. 
We call the new metric \ourmetric: Multimodal Assistant Rating Subject\&Style.
The process is illustrated in Fig.~\ref{fig:mllm_eval}, with more details in the \sm.

For each content-style pair, we generate multiple images and evaluate both the average and best sample quality according to the MLLM judge. This dual evaluation not only facilitates a fair comparison with existing literature but also provides flexibility for users in downstream applications, allowing them to select the most preferred sample.

\noindent \textbf{Human Evaluation.} 
To complement automated metrics, we also conduct human evaluations on a subset of generated images, comparing our results with those from ZipLoRA, the primary competitor. 
We consider two cases: 1) \textit{randomly} select one generated sample from each approach for every test content-style LoRA pair; 2) take a \textit{best} sample, \ie accepted by the MLLM model (if there are multiple samples with correct style and content, randomly choose one of them).
For unbiased feedback, we anonymize method names, asking evaluators to rate whether our solution produces images that are better, similar, or worse than the baseline.
This human evaluation offers insights into real-world user preferences and serves as qualitative validation of our approach.

\section{Experiments}

\noindent \textbf{Baselines.} 
We compare our approach to several established methods, including: joint training of both content and style via Dreambooth \cite{ruiz2023dreambooth}; direct merging of LoRA weights \cite{wortsman2022model}; general model merging techniques such as DARE \cite{yu2024language}, TIES \cite{yadav2024ties}, and DARE-TIES \cite{goddard2024arcee}; and ZipLoRA \cite{shah2024ziplora}, which is specifically designed for merging subject and style LoRAs.
ZipLoRA has also been compared in \cite{shah2024ziplora} with strategies such as StyleDrop \cite{styledrop}, Custom Diffusion \cite{kumari2022customdiffusion} and Mix-of-show \cite{gu2023mixofshow}. These methods, however, have been shown to perform less effectively while being computationally costly, so we exclude them from further comparison in this work.

\noindent \textbf{Implementation Details.}  
All experiments use the SDXL v1 
\cite{Podell2023SDXLIL} unless specified, following the setup in \cite{shah2024ziplora}. For subject LoRAs, we adopt rare unique token identifiers as in \cite{ruiz2023dreambooth}. In contrast, style LoRAs are fine-tuned using text description identifiers, following \cite{styledrop}, where these were found more effective for style representation.  More details 
in \sm.

\noindent \textbf{Datasets.} Our hypernetwork is trained on a set of LoRAs rather than images. 
The datasets used to train the LoRAs include 30 subjects (each with 4–5 images) and 26 styles (each represented by a single image).
For training, validation, and testing, we split the subjects into 20-5-5 and styles into 18-3-5 (see the \sm.\ for details), yielding a total of 360 subject-style LoRA combinations for hypernetwork training, a quantity shown to be sufficient for robust performance. %
Our hypernetwork operates on each column of the LoRA weight update matrix independently, so each combination of subject and style LoRAs represents thousands of training examples. Therefore, we can train the hypernetwork to converge with just a few hundreds of subject-style combinations.

\noindent \textbf{Evaluation Details.} 
For our MLLM-based metric, \ourmetric, we use the LLaVA-Critic 7b model \cite{xiong2024llavacritic}.
The prompts used for the MLLM model are detailed in the \sm.
For human evaluation, evaluators are presented with content and style reference images alongside randomly ordered outputs from each method. Each of our 25 evaluators assesses 25 pairs of images comparing two approaches. An example task precedes the evaluation to clarify the assessment criteria (see \sm.). Evaluators choose the option that best reflects target content and style, with choices between \textit{Option 1}, \textit{Comparable}, and \textit{Option 2}.

\begin{table}[t]
    \centering
        \begin{tabular}{l c c}
            \toprule
           & Average case & Best case \\
            \toprule
                 Joint Training  \cite{ruiz2023dreambooth} & 0.53 &0.84 \\
                Direct Merge \cite{wortsman2022model}  &0.40 & 0.76 \\
                DARE \cite{yu2024language} &0.34& 0.72\\
                TIES \cite{yadav2024ties} & 0.43&0.80\\
                DARE-TIES \cite{goddard2024arcee} & 0.30 & 0.60\\
                \ziplora \cite{shah2024ziplora} & 0.58 &\textbf{1.00} \\
                \textbf{\ours (ours)} &\textbf{0.71} &\textbf{1.00} \\
                         \bottomrule
        \end{tabular}
    
    \caption{\textbf{MLLM Evaluation.} Ratio of generated images with the correct content and style on the combinations of test subjects and styles according to our new metric \ourmetric. Our solution leads to better images compared to  existing approaches.}
    \vspace{-0.2cm}
    \label{tab:mllm_evaluation}
\end{table}

\subsection{Quantitative Analysis}

We quantitatively evaluate our contributions in four main ways: (1) performance of \ours through our MLLM-based \ourmetric\ metric as described in Sec.~\ref{sec:method:metrics}; (2) studying the alignment of MLLM judge with human preference; (3) via a human evaluation study on a subset of generated samples; (4) by studying the produced merging coefficients.

\noindent\textbf{1) MLLM Evaluation Results} are presented in Table~\ref{tab:mllm_evaluation}. Our solution consistently outperforms all methods, including ZipLoRA, in both content and style accuracy. 
For the best sample (selected by MLLM from 10 generated images as one with correct style and content if available), both our solution and ZipLoRA achieve perfect accuracy, indicating that users can reliably choose preferred outputs when multiple samples are available. 
Across all generated images, on average our solution performs better than ZipLoRA, likely benefitting from its capacity to leverage knowledge learned from diverse content-style LoRA combinations.

\noindent\textbf{2) Metrics Alignment with Human Preference.}
We computed the correlation between CLIP-I, DINO, and our \ourmetric\ metric against the human evaluation score for the direct merge approach (which has the highest CLIP-I and DINO, see \textit{Supp.\ Mat} for complete evaluation on these metrics), and obtained $[0.08, -0.01, 0.76]$ respectively. 
This result shows the \ourmetric\ metric is well-aligned with human preference, while CLIP-I and DINO are not appropriate for evaluating joint subject-style generation, as discussed in Sec.~\ref{sec:method:metrics}.

\begin{figure}[t]
    \centering
    \includegraphics[width=\linewidth, trim=0cm 0.25cm 0cm 0.25cm, clip]{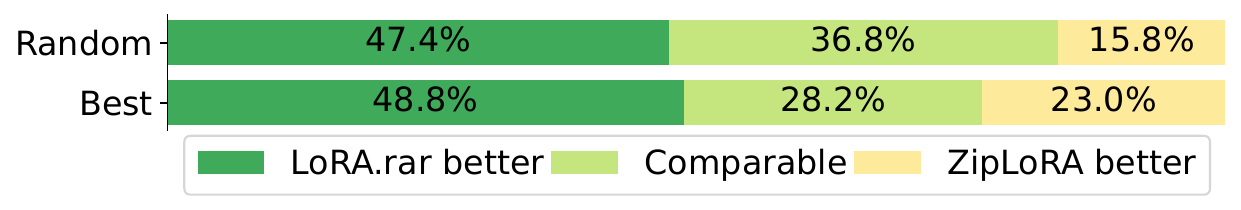}
    
    \caption{\textbf{Human Evaluation} for generated images sampled randomly or according to \ourmetric. More than 75\% respondents consider \ours comparable or better than ZipLoRA.}
        \vspace{-0.1cm}
        \label{fig:human_eval}
\end{figure}

\noindent\textbf{3) Human Evaluation} results are reported in Fig.~\ref{fig:human_eval}. This evaluation was conducted on a subset of generated samples as described earlier and focused on ZipLoRA as the primary comparison baseline, given the time constraints of manual assessment.
The results indicate that our solution compares favorably with ZipLoRA, confirming that our generated images are typically either better or comparable in quality. Furthermore, our solution can operate in real-time for new subject-style combinations, unlike ZipLoRA.

\noindent\textbf{4) Analysis of Merging Coefficients}
\begin{figure}[t]
    \centering
    \includegraphics[width=0.99\linewidth]{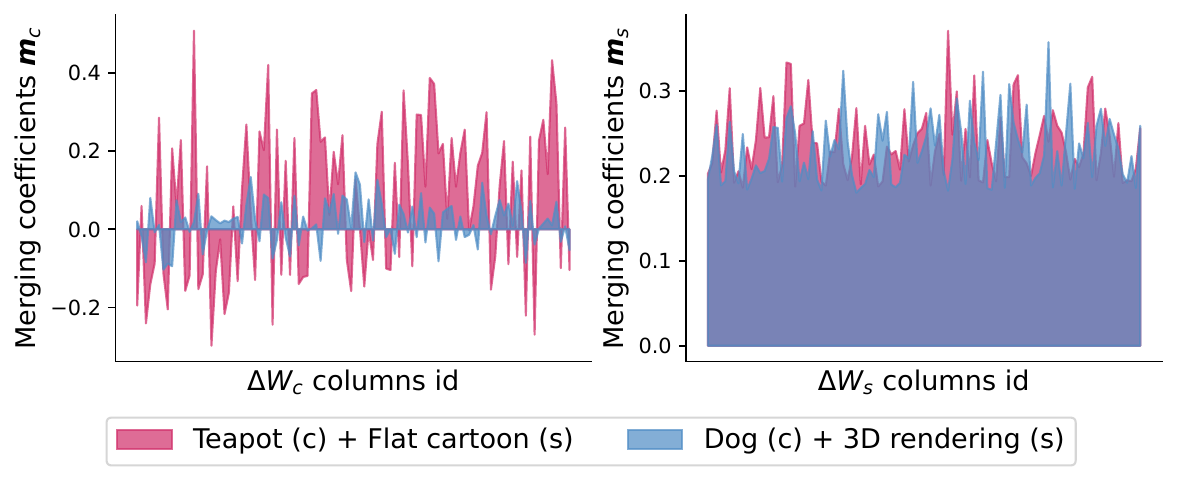}
    
    \vspace{-0.2cm}
    \caption{\textbf{LoRA.rar's Merging Coefficients} $\boldsymbol{m}_c, \boldsymbol{m}_s$ for randomly selected columns of the LoRA weight update matrices. \ours learns a non-trivial strategy with superior performance.}
    \vspace{-0.2cm}
        \label{fig:lorarar_coefficients}
\end{figure}
learned by \ours is shown in Fig.~\ref{fig:lorarar_coefficients}. 
\ours learns a non-trivial adaptive merging strategy, with diverse coefficients.
This adaptability allows \ours to flexibly combine content and style representations, likely contributing to its superior performance. 
ZipLoRA, instead, mostly converges to 
a binary selection of either subject or style for each weight (see Fig.~\ref{fig:ziplora_coefficients}). 
This leads to overfitting to one of the two aspects (\eg, subject rendered too realistic and/or without style acquisition, style not applied consistently), limiting its capacity to finely integrate details across styles and subjects. 
LoRA.rar, thanks to pre-training on diverse LoRA pairs, finds better merging coefficients to integrate subject and style without overfitting.

\subsection{Qualitative Analysis}
We conduct a qualitative analysis of \ours by: (1) comparing the images generated by \ours with those produced by competing methods, and (2) analyzing the diversity of images generated by \ours across contents and styles. 

\begin{figure*}[!t]
    \centering
    \resizebox{0.46\textwidth}{!}{%
    \begin{tabular}[c]{L L L L L L L}
     \small A \textit{[C] teapot} in \textit{[S]} style & 3D Rendering & Oil Painting & Watercolor Painting & Flat Cartoon Illustration & Glowing \\
        \tabfigure{width=0.12\textwidth}{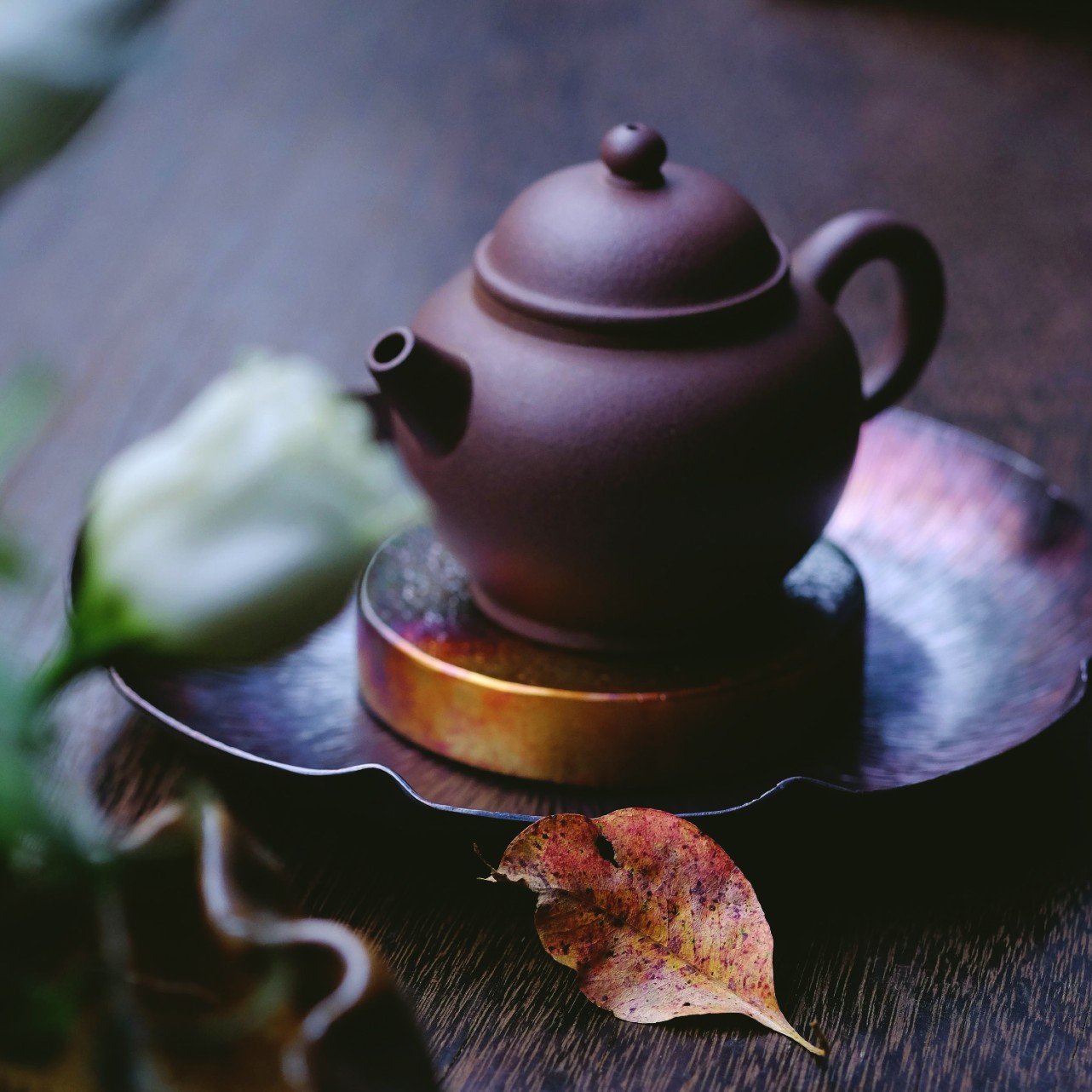}  & \tabfigure{width=0.12\textwidth}{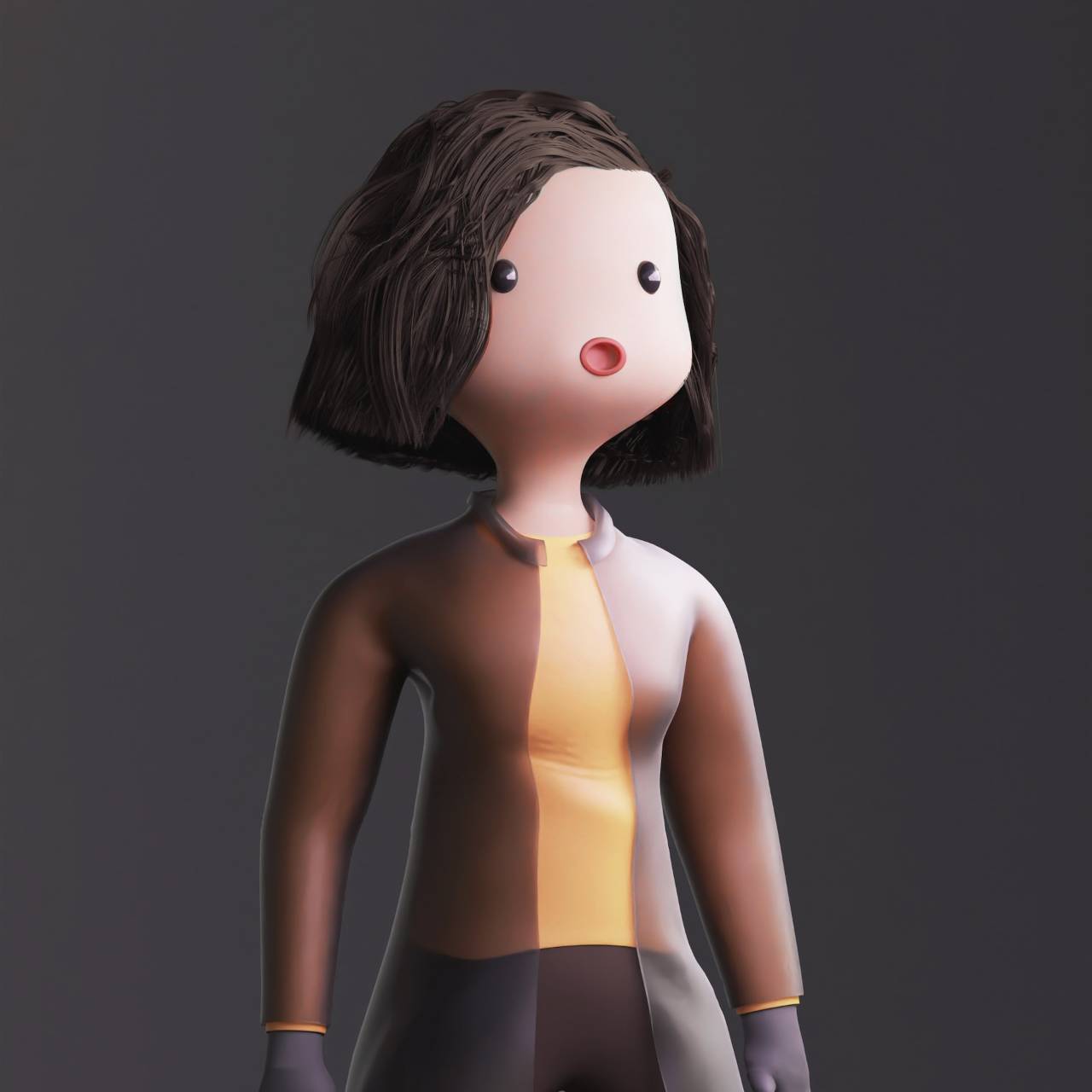} 
        & \tabfigure{width=0.12\textwidth}{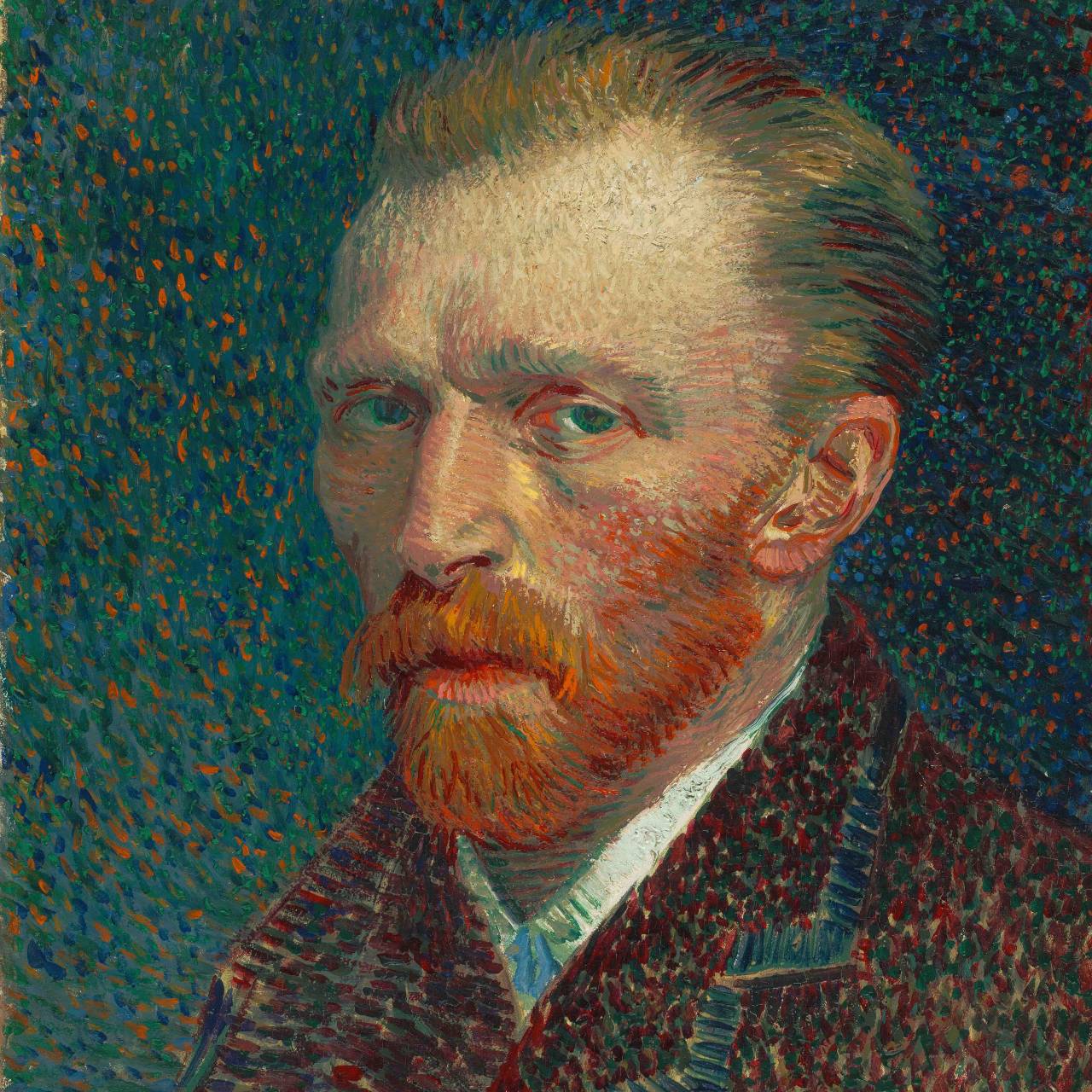}
        & \tabfigure{width=0.12\textwidth}{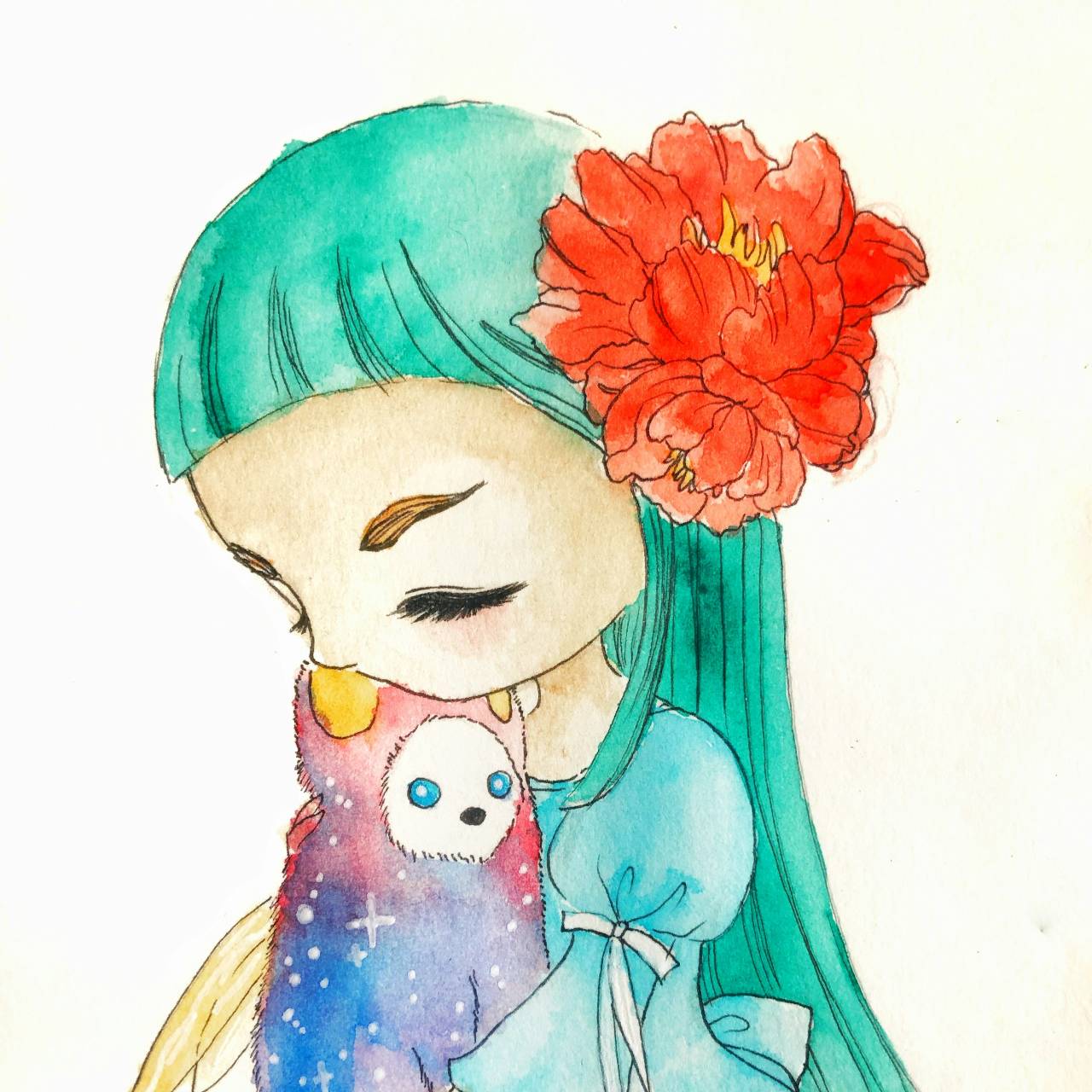}
        & \tabfigure{width=0.12\textwidth}{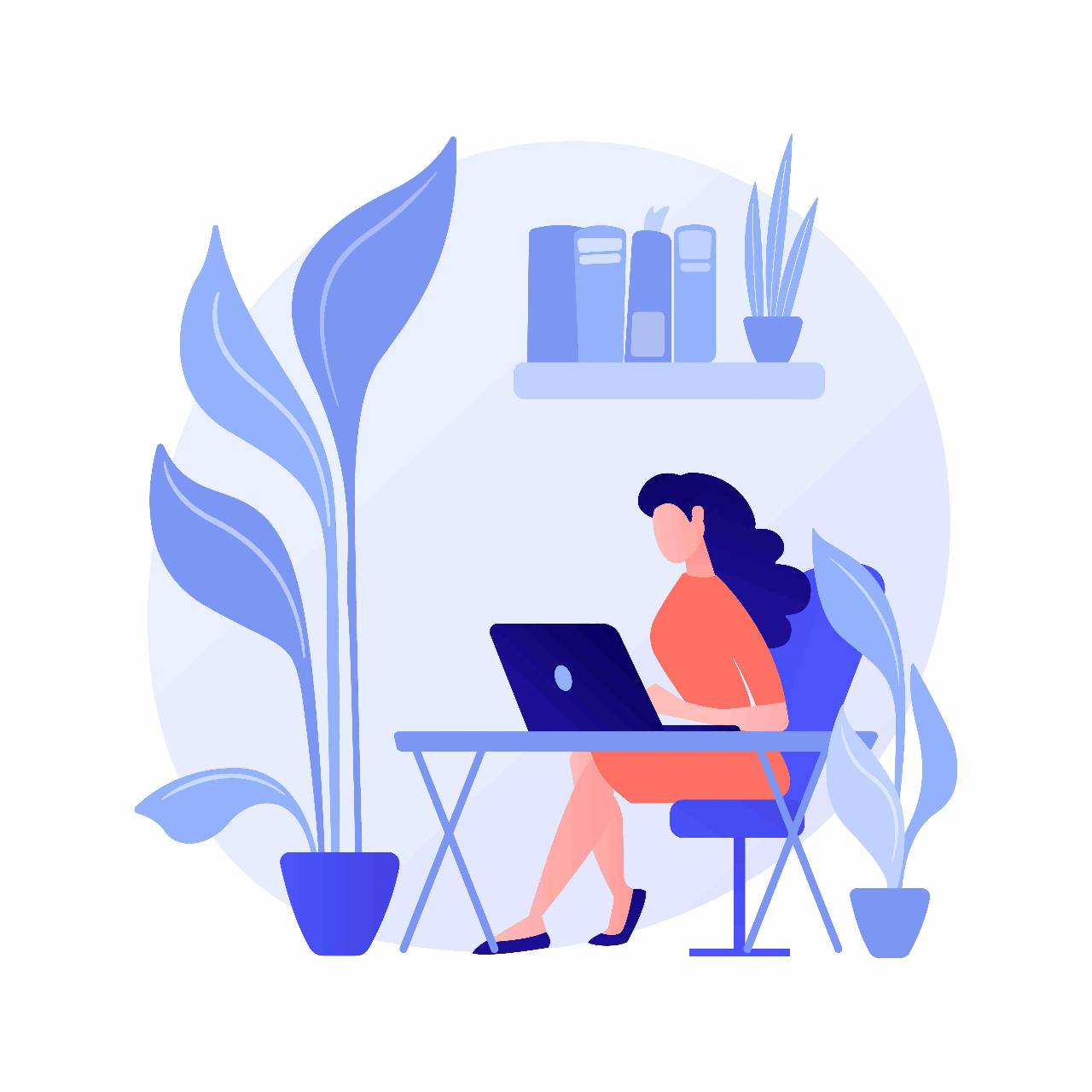}
        & \tabfigure{width=0.12\textwidth}{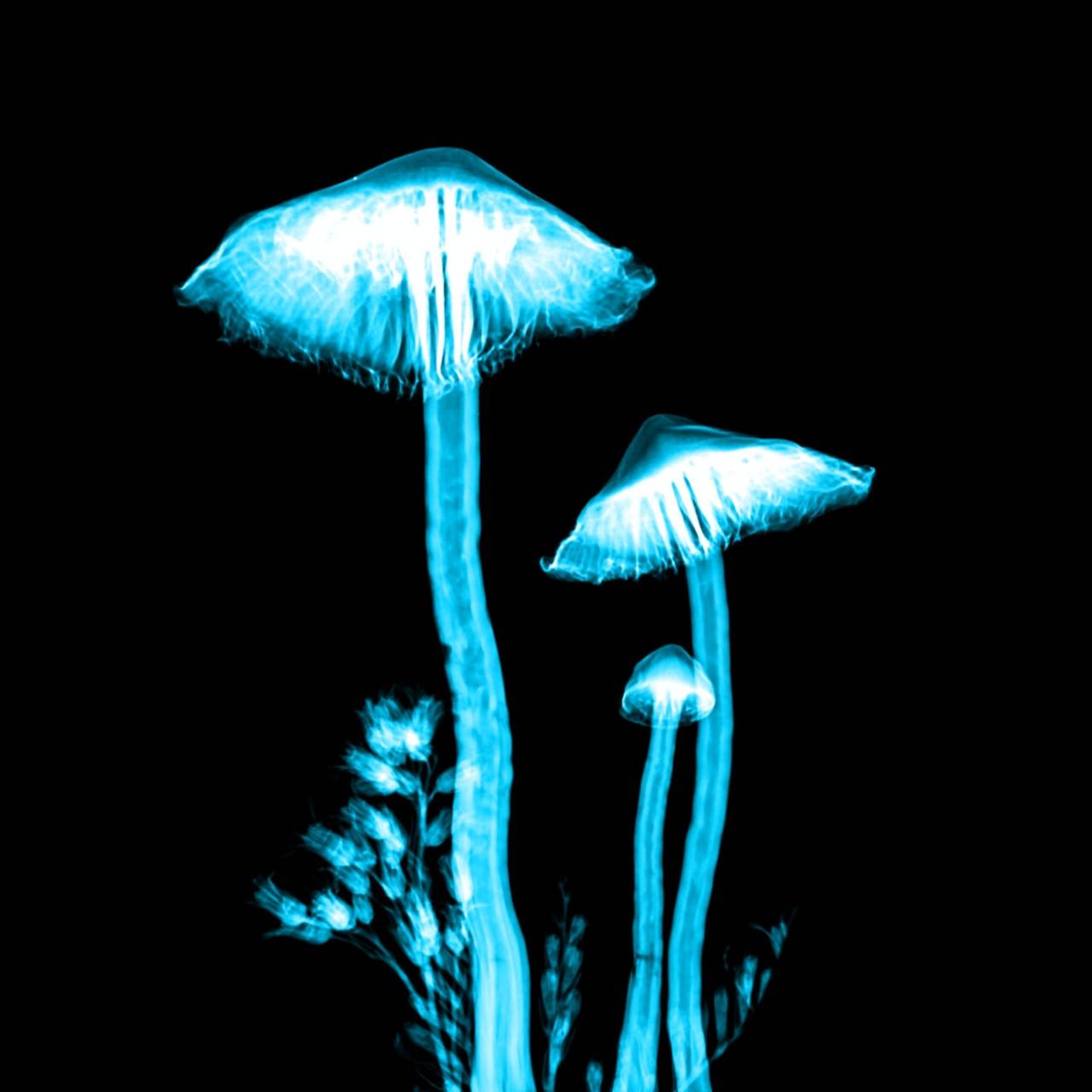} \\

    \begin{minipage}[c][1.7cm][c]{\linewidth} %
        \centering Joint Training 
    \end{minipage}
    &
        \tabfigure{width=0.12\textwidth}{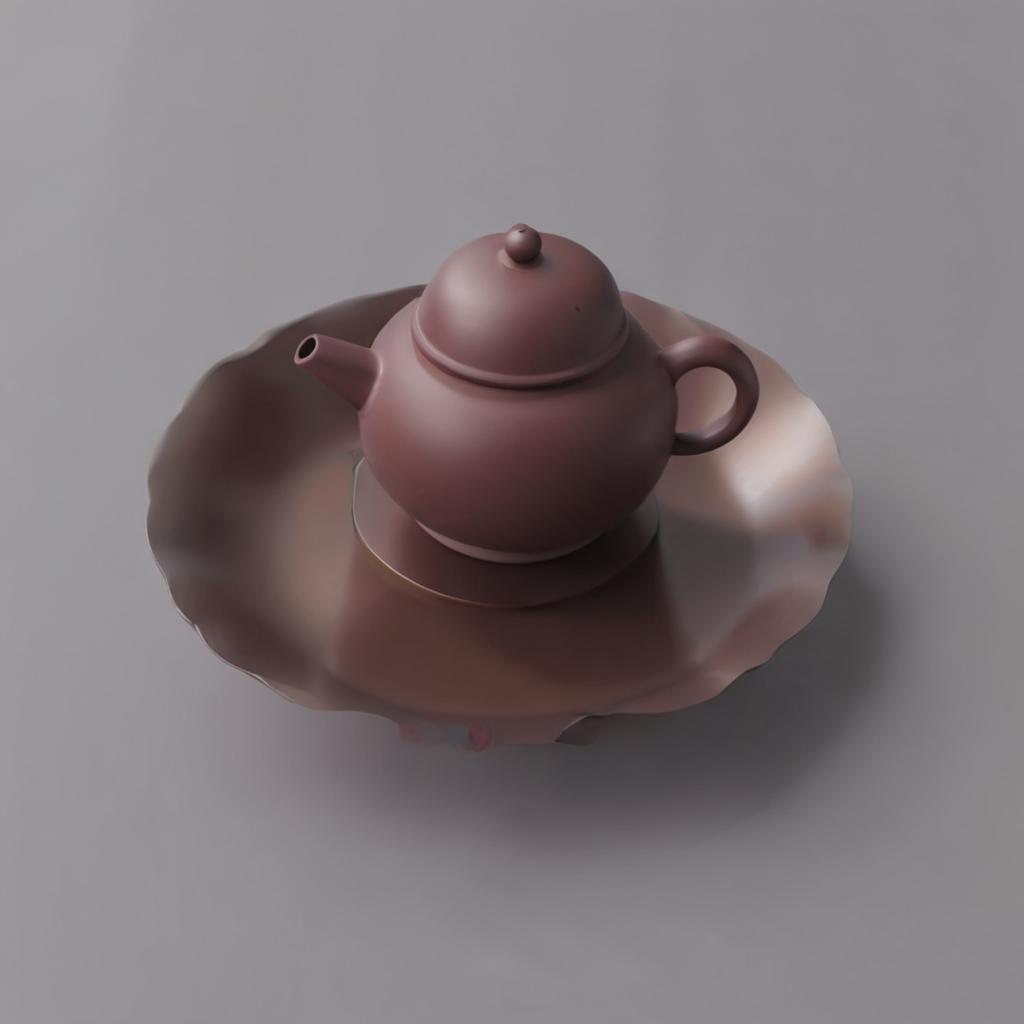} &
        \tabfigure{width=0.12\textwidth}{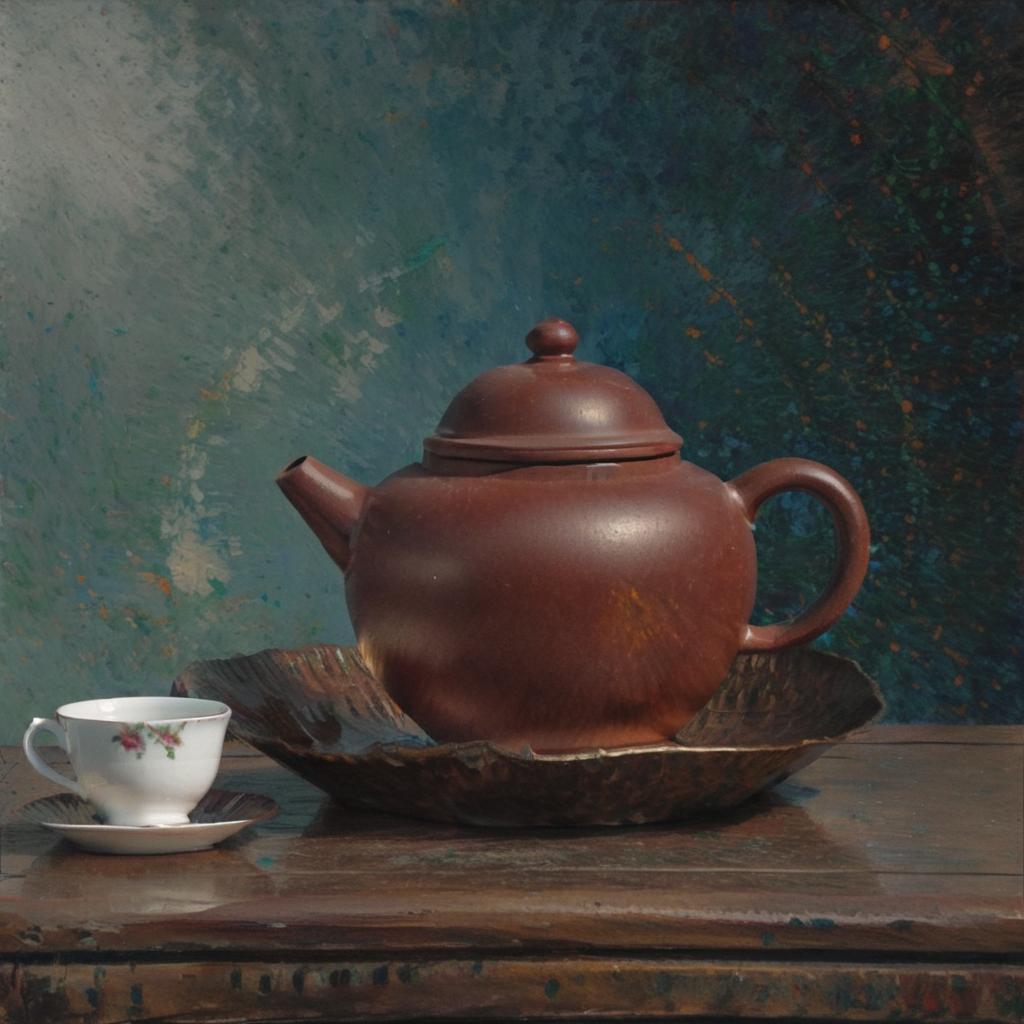} &
        \tabfigure{width=0.12\textwidth}{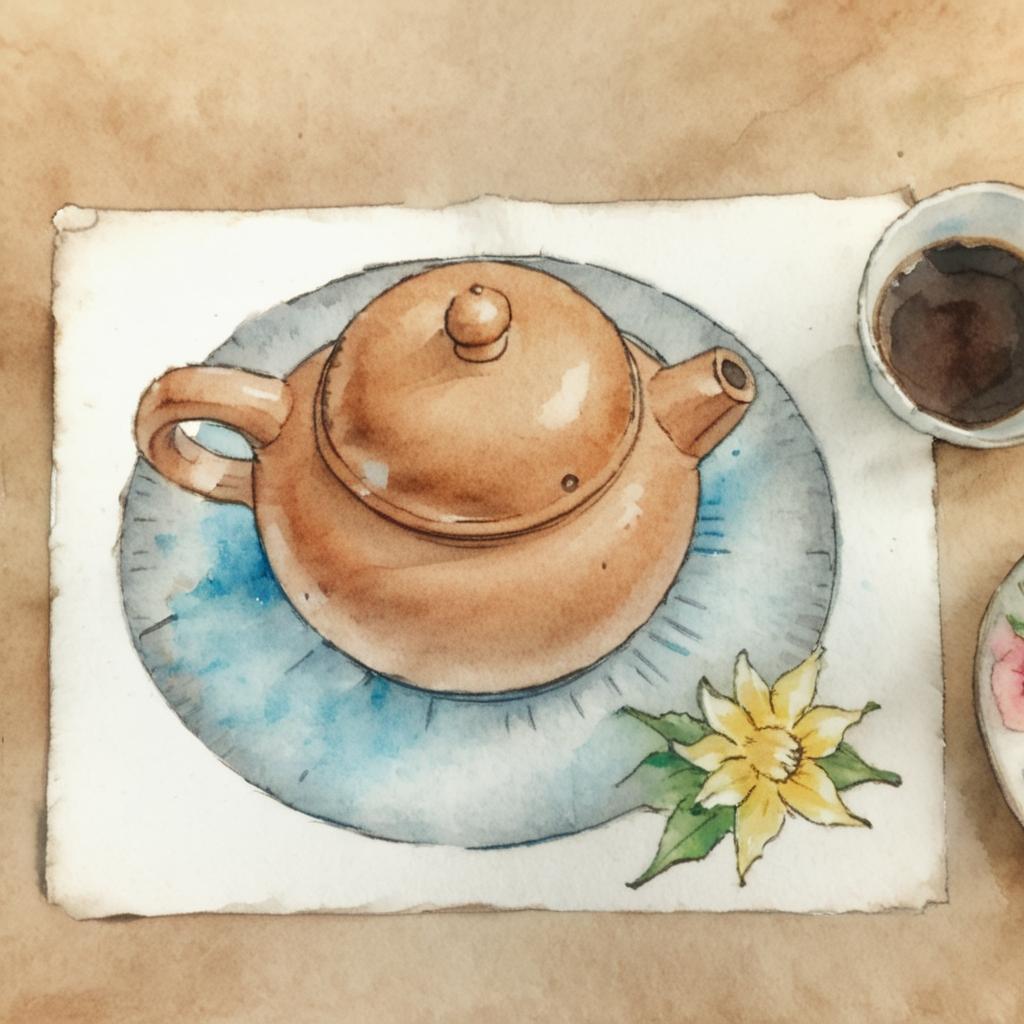} &
        \tabfigure{width=0.12\textwidth}{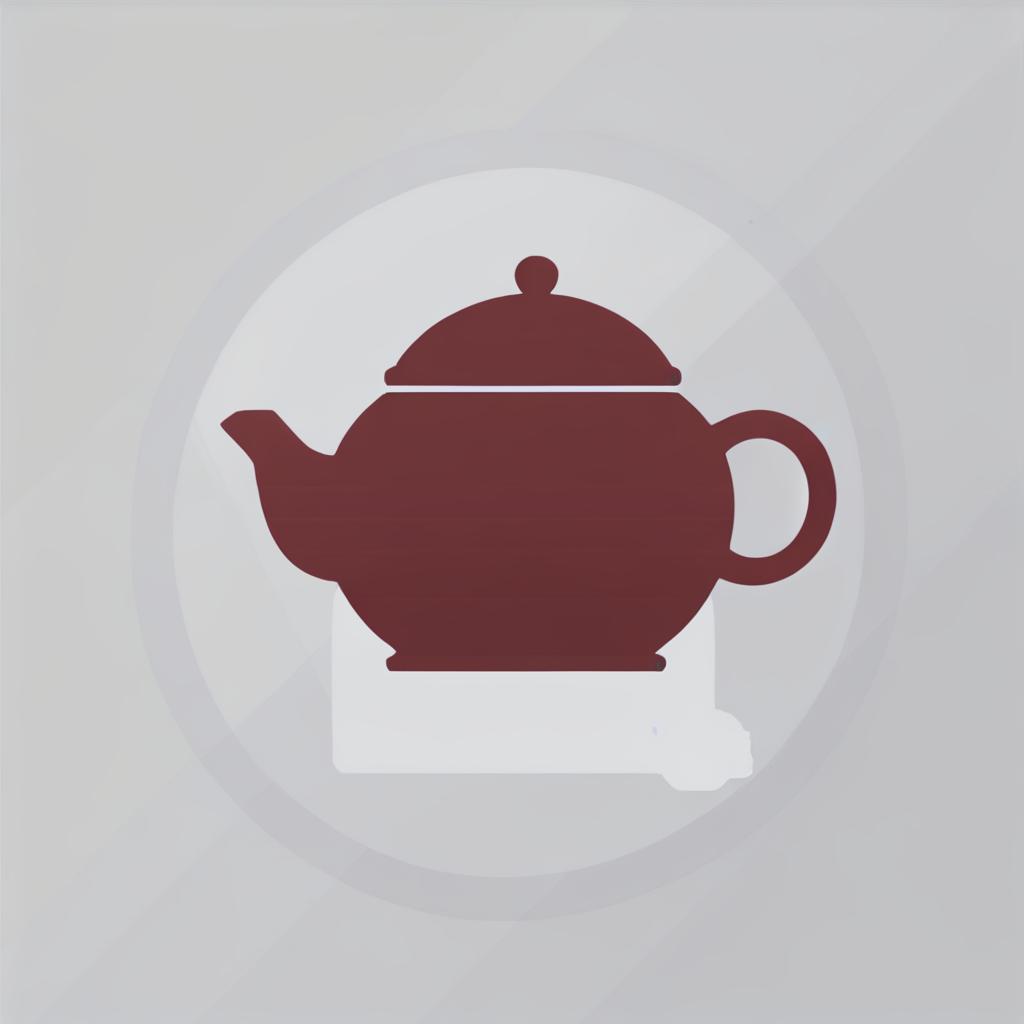} &
        \tabfigure{width=0.12\textwidth}{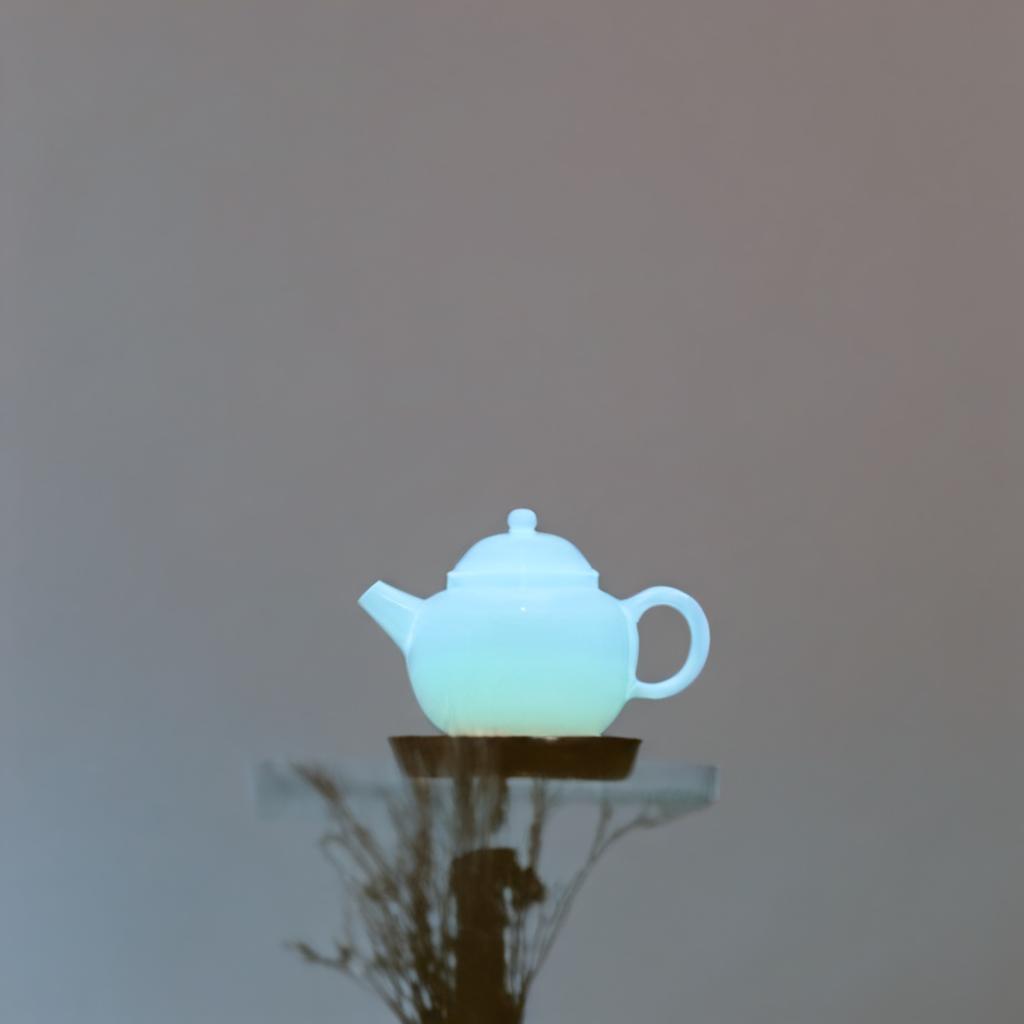} \\
     \begin{minipage}[c][1.7cm][c]{\linewidth} %
        \centering Direct Merge
    \end{minipage}
    &
        \tabfigure{width=0.12\textwidth}{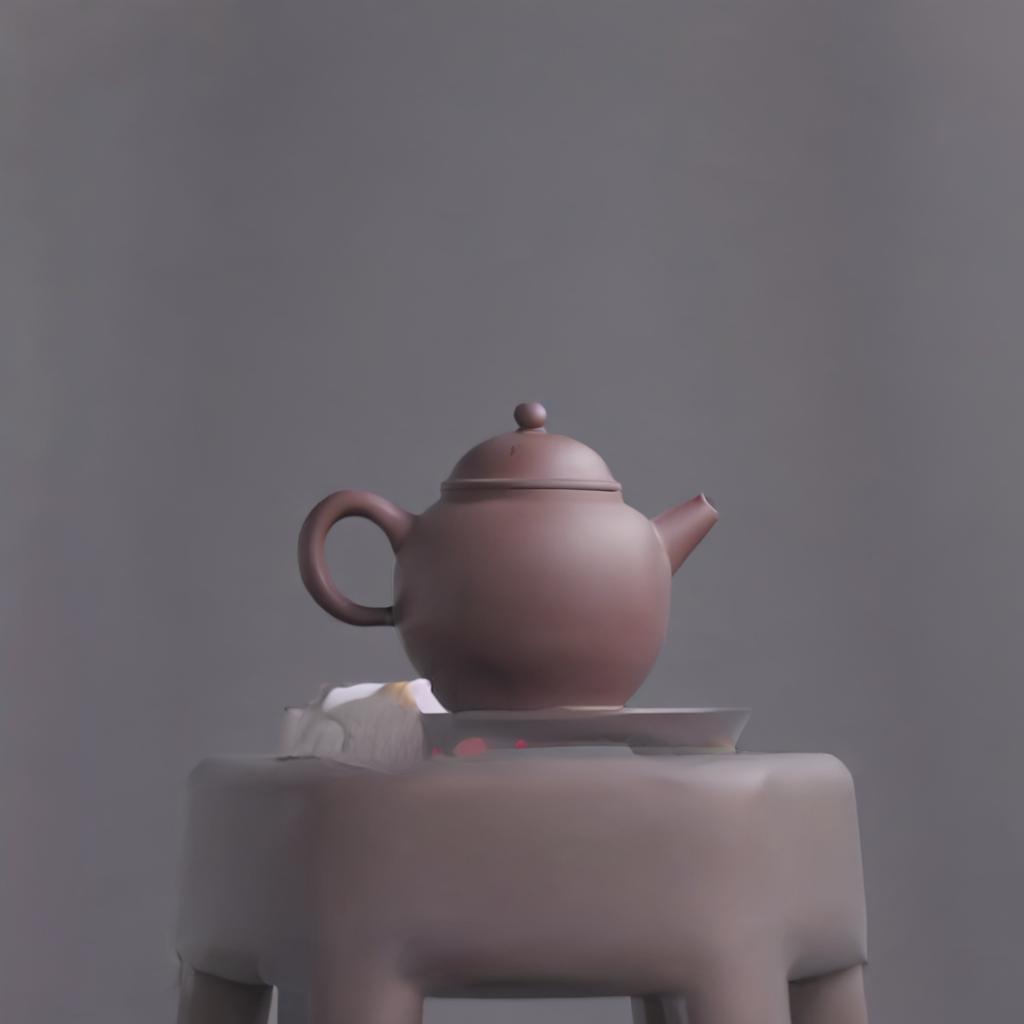} &
        \tabfigure{width=0.12\textwidth}{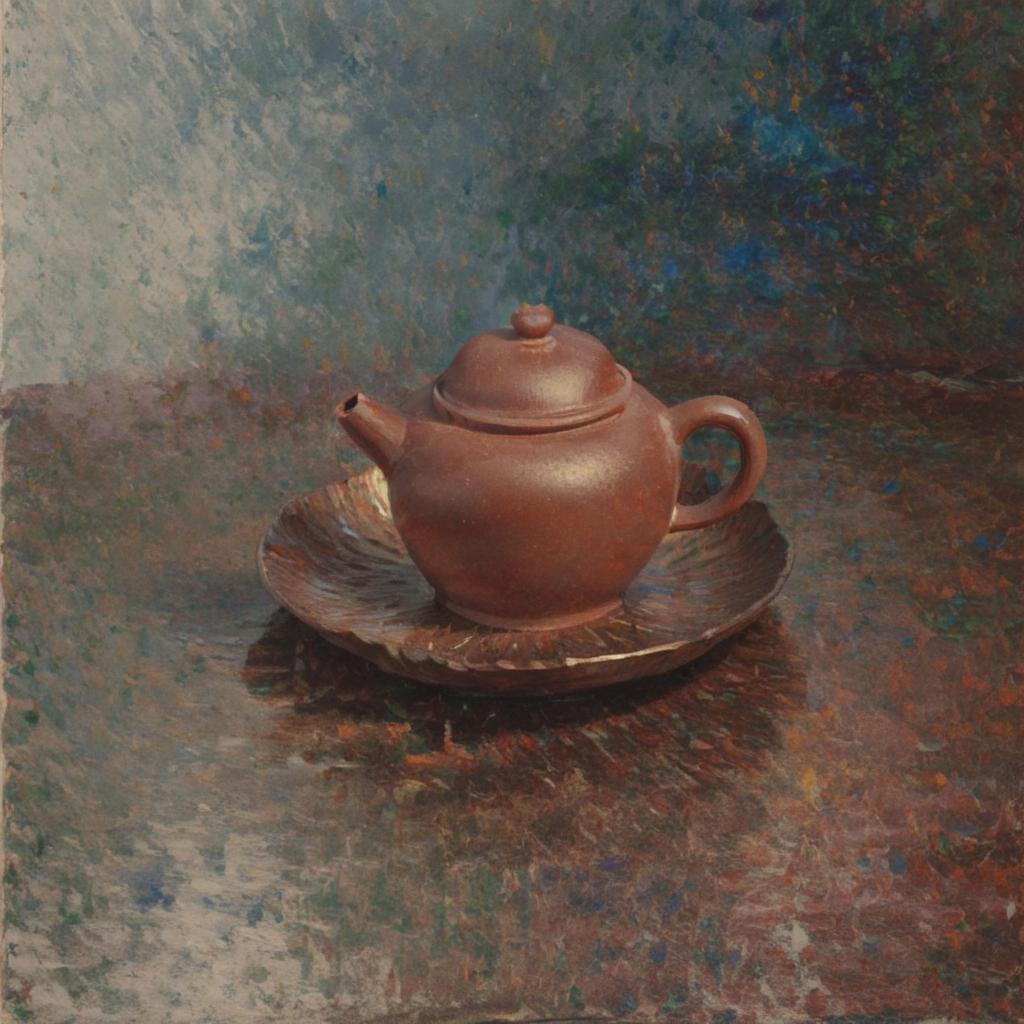} &
        \tabfigure{width=0.12\textwidth}{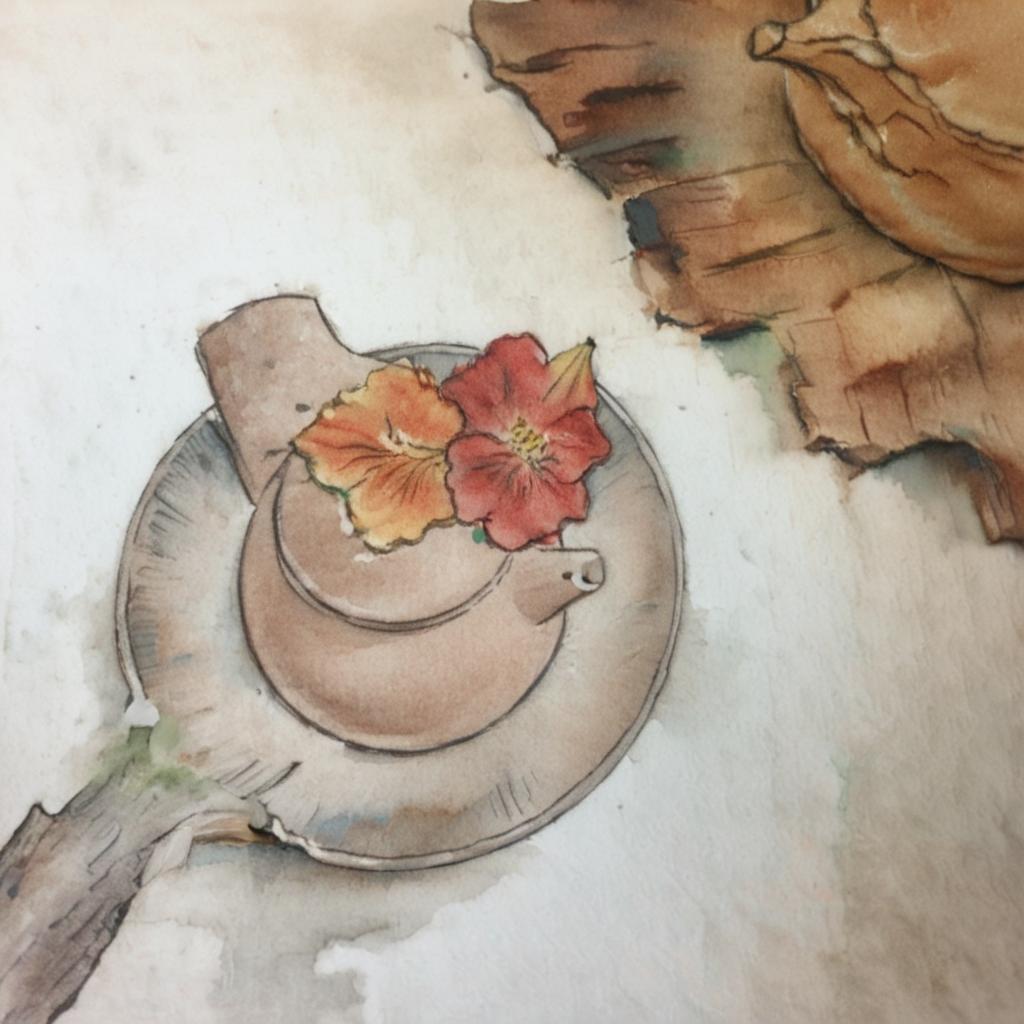} &
        \tabfigure{width=0.12\textwidth}{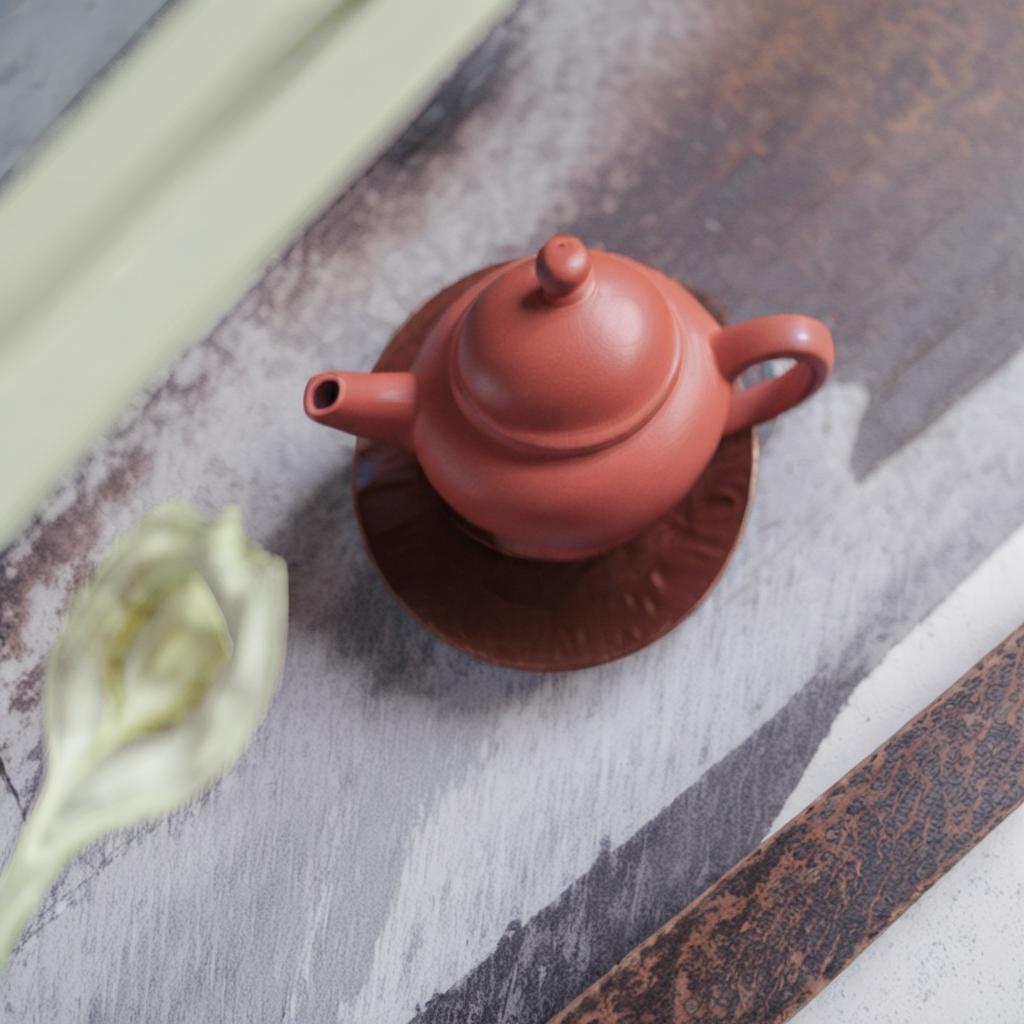} &
        \tabfigure{width=0.12\textwidth}{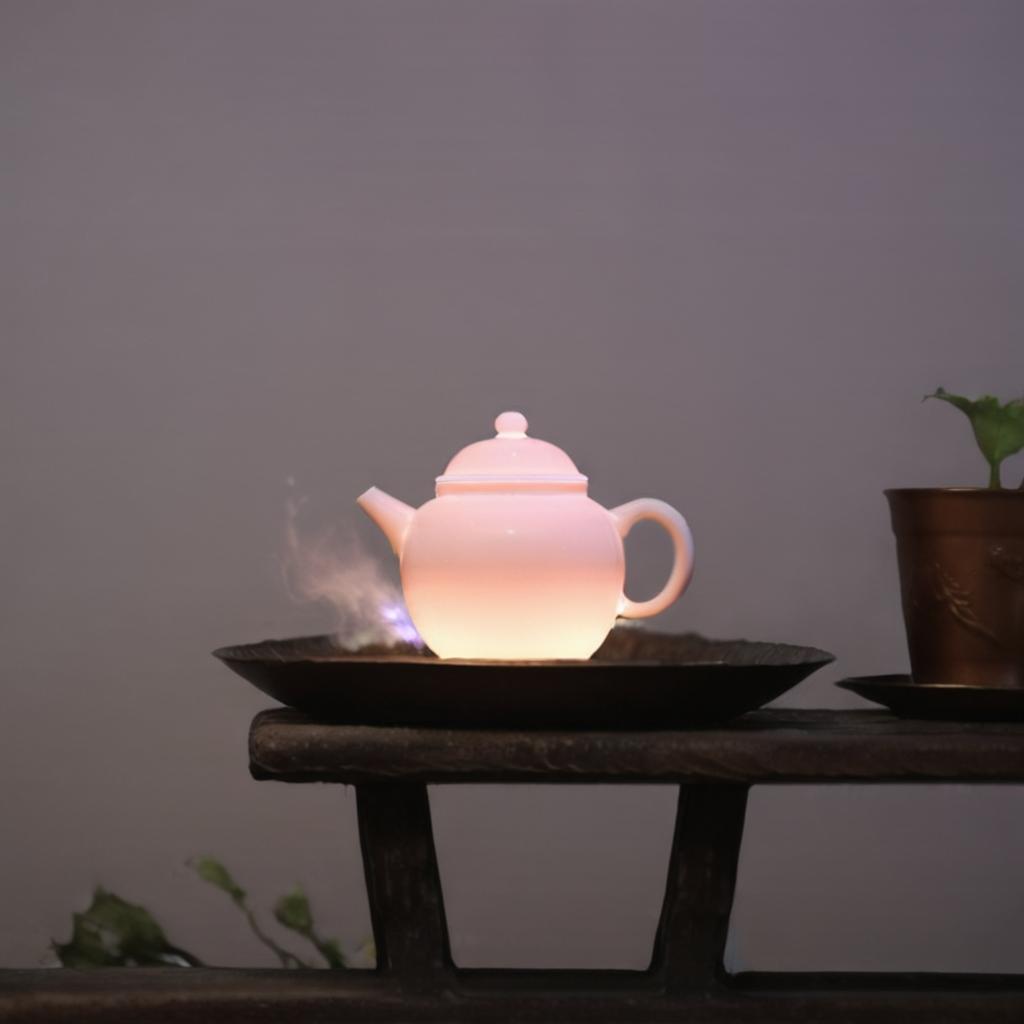} \\
     \begin{minipage}[c][1.7cm][c]{\linewidth} %
        \centering DARE
    \end{minipage}
    &
        \tabfigure{width=0.12\textwidth}{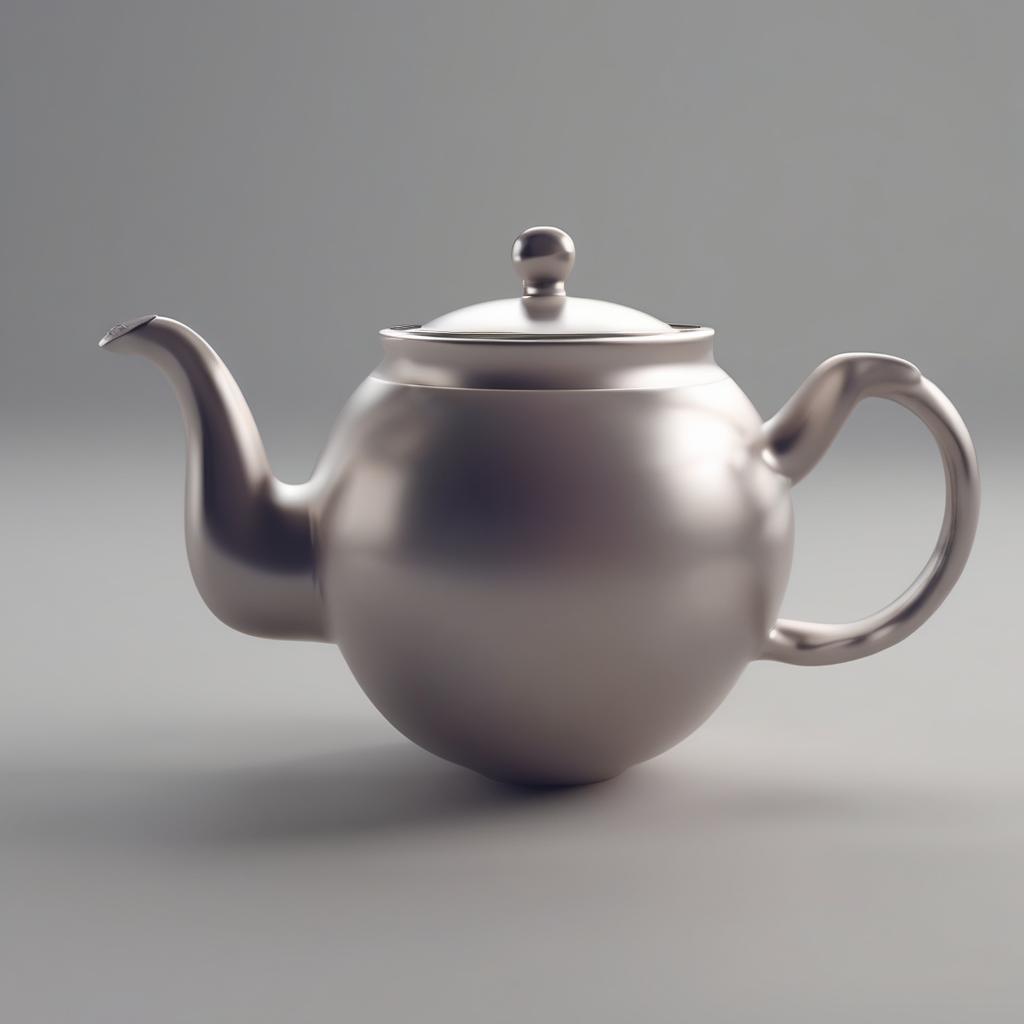} &
        \tabfigure{width=0.12\textwidth}{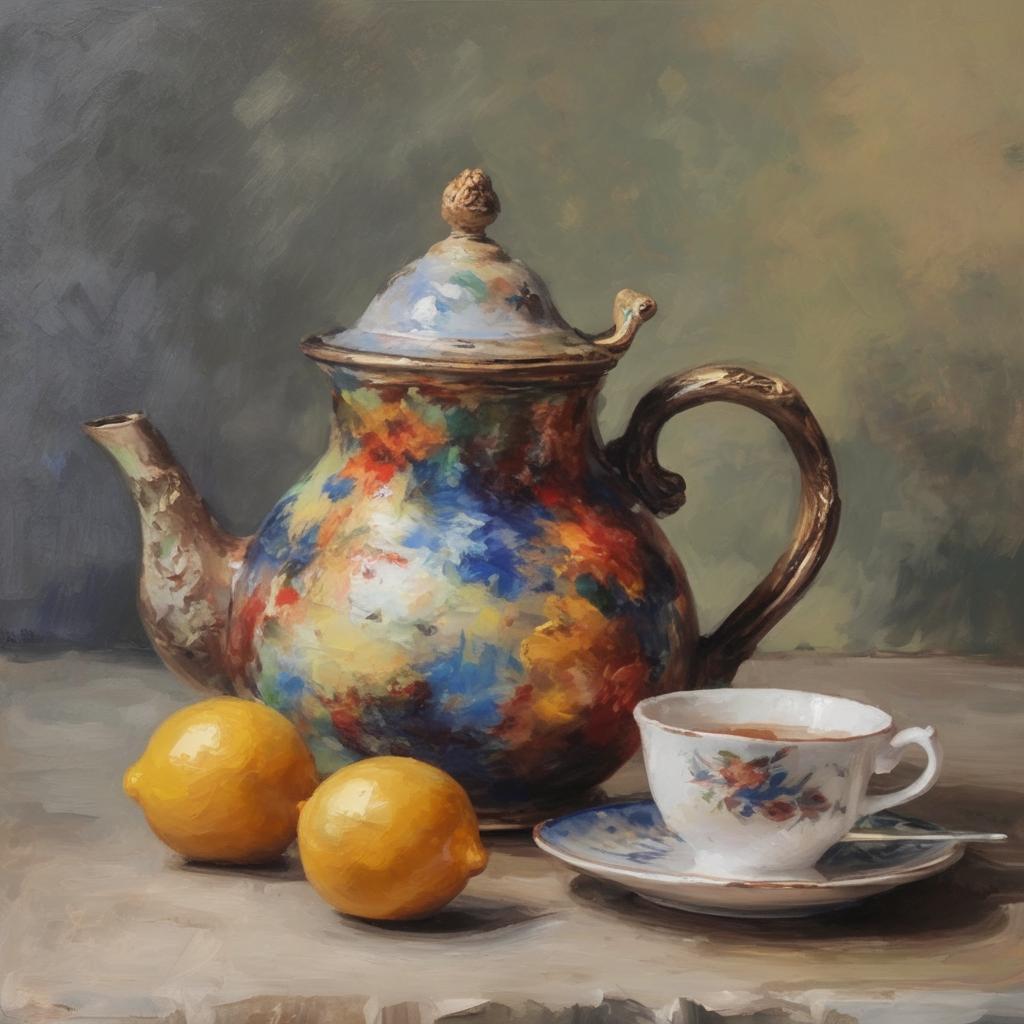} &
        \tabfigure{width=0.12\textwidth}{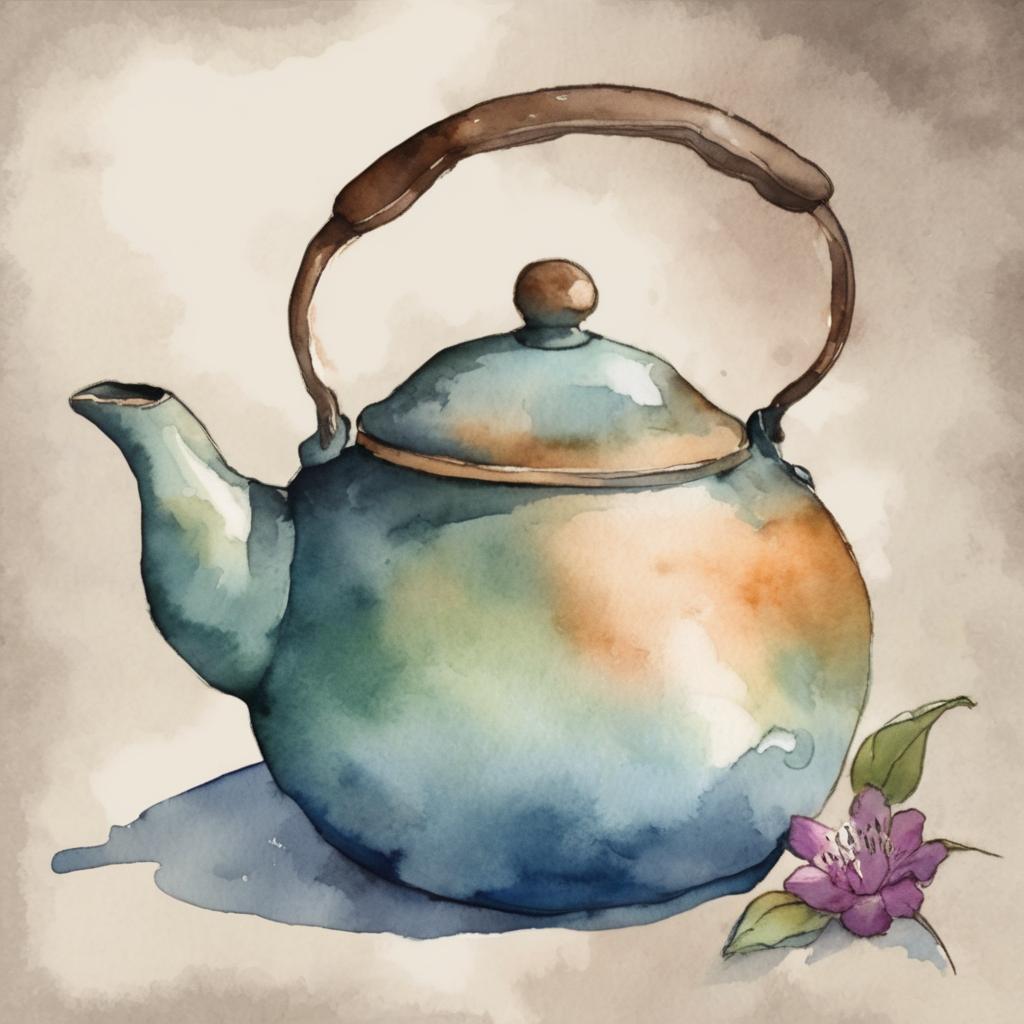} &
        \tabfigure{width=0.12\textwidth}{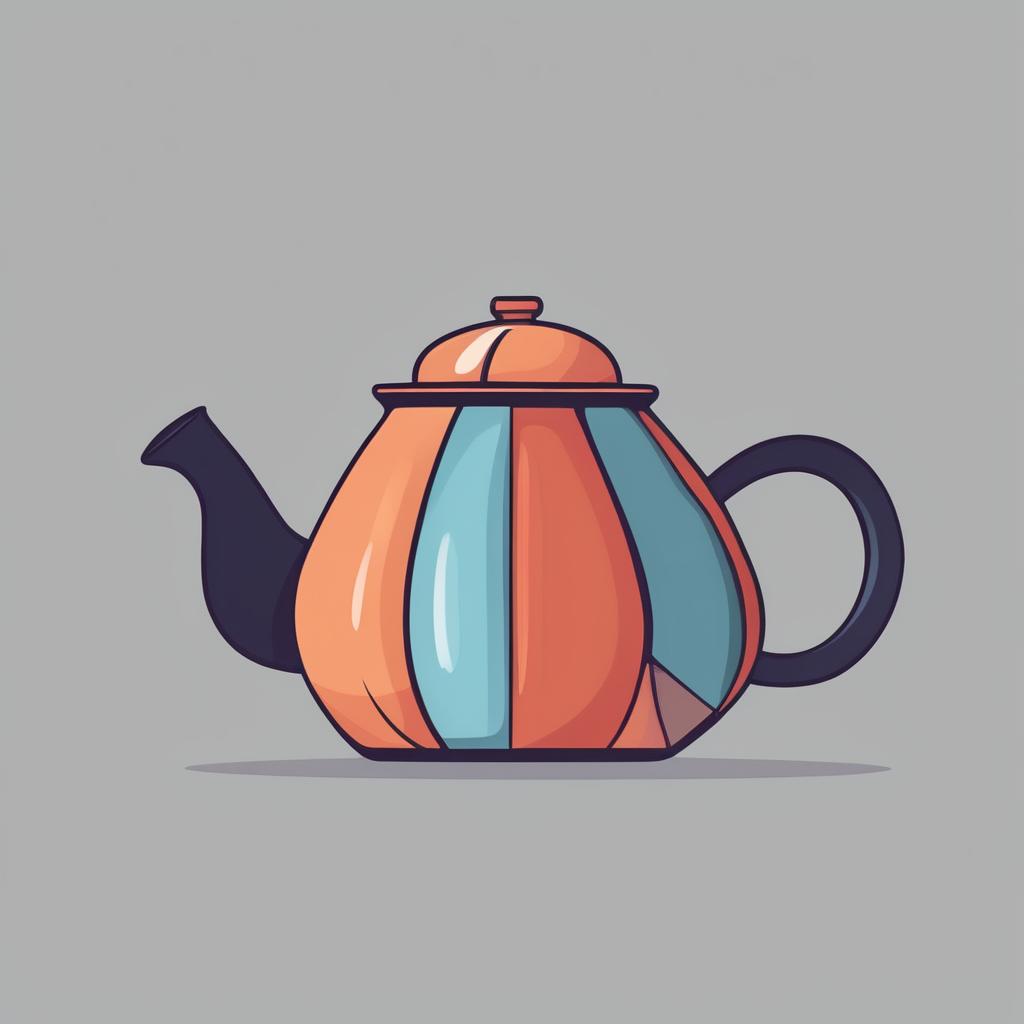} &
        \tabfigure{width=0.12\textwidth}{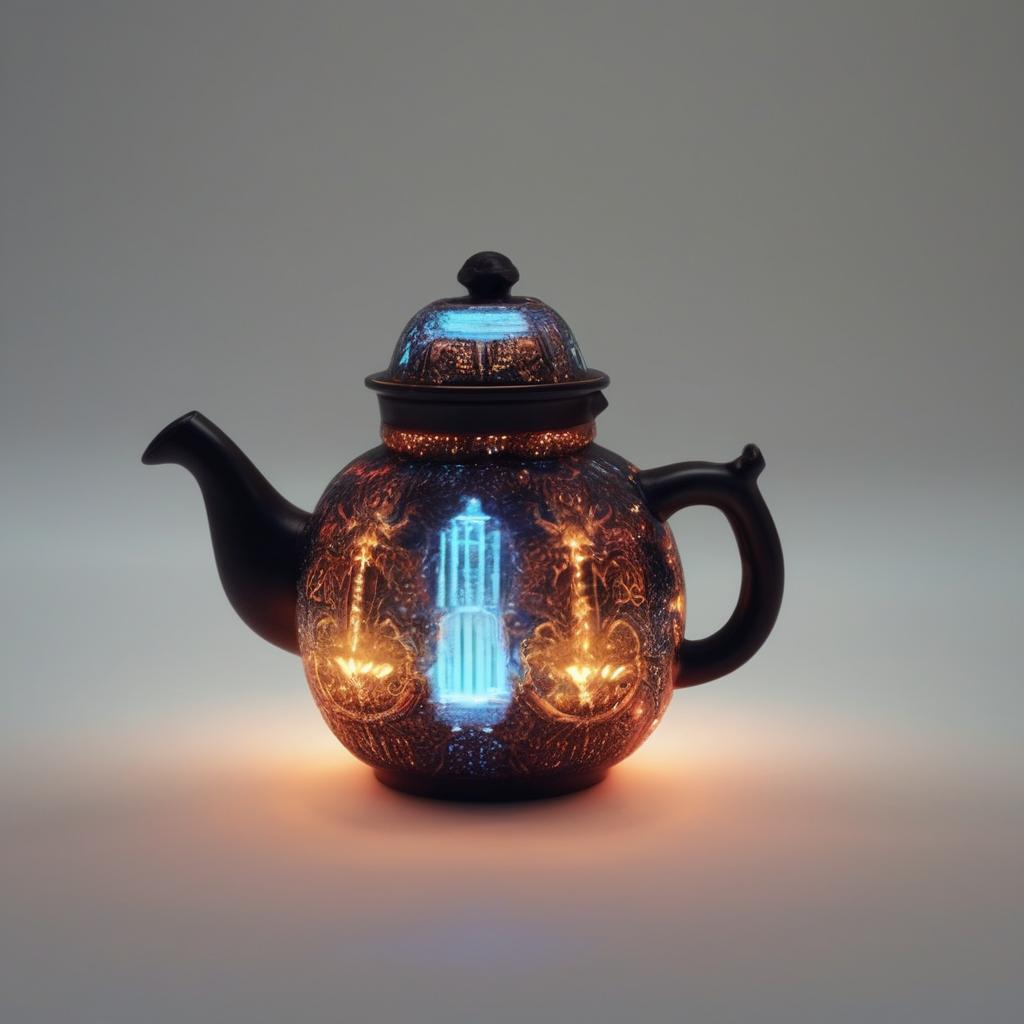} \\
    \begin{minipage}[c][1.7cm][c]{\linewidth} %
        \centering TIES
    \end{minipage}
    &
        \tabfigure{width=0.12\textwidth}{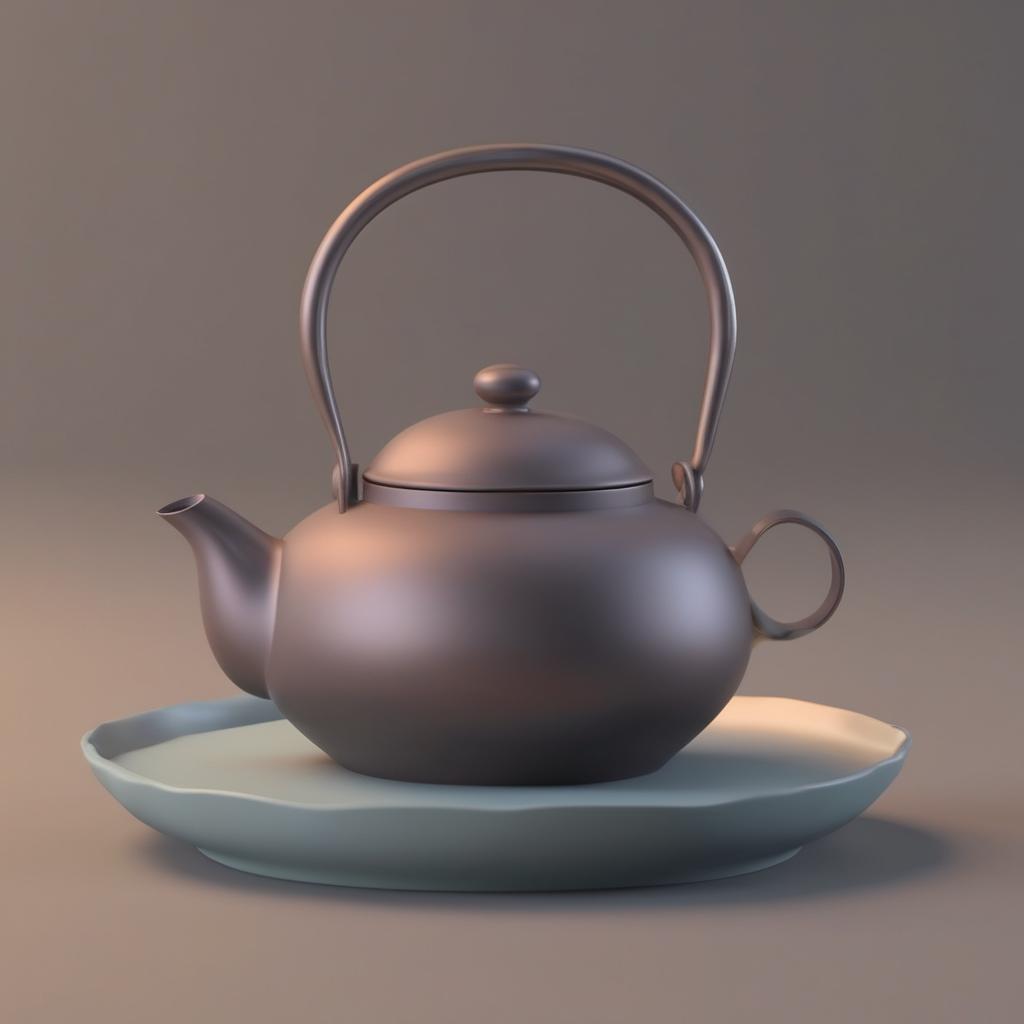} &
        \tabfigure{width=0.12\textwidth}{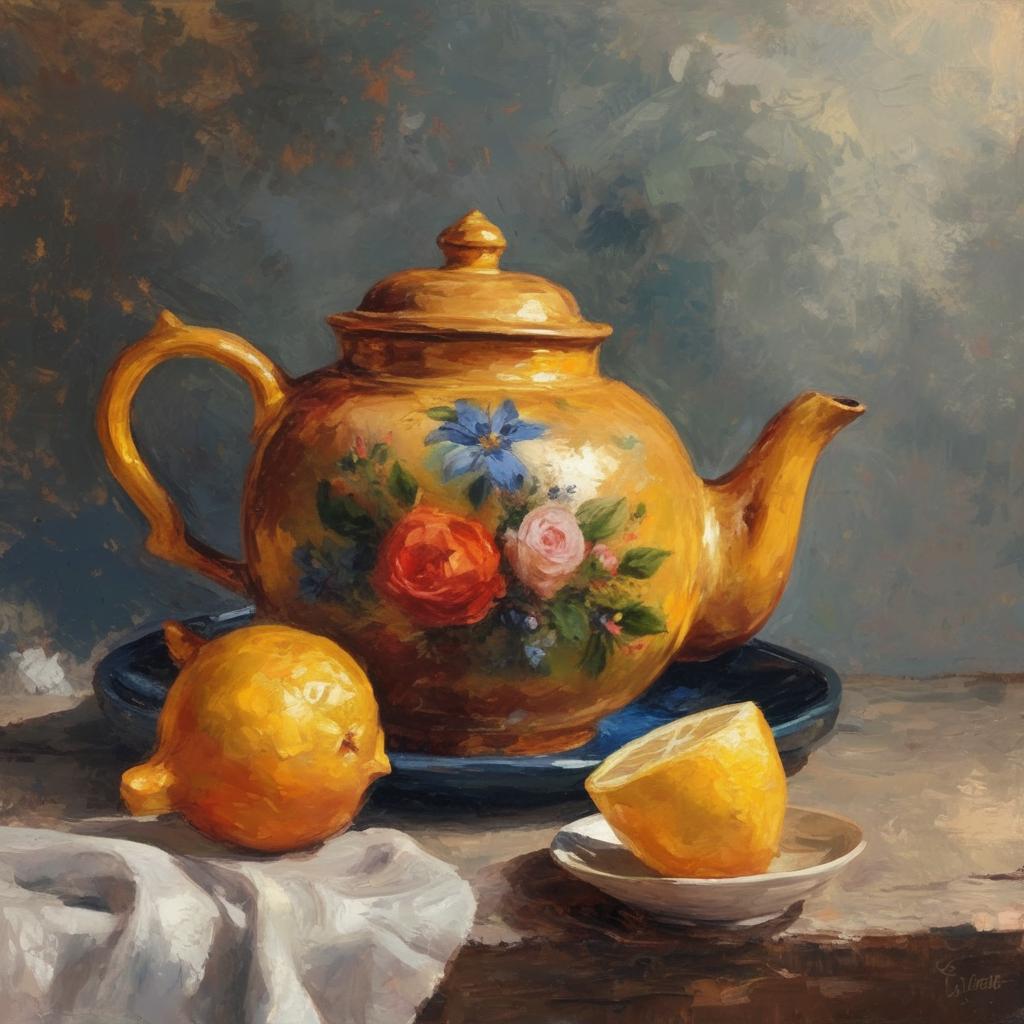} &
        \tabfigure{width=0.12\textwidth}{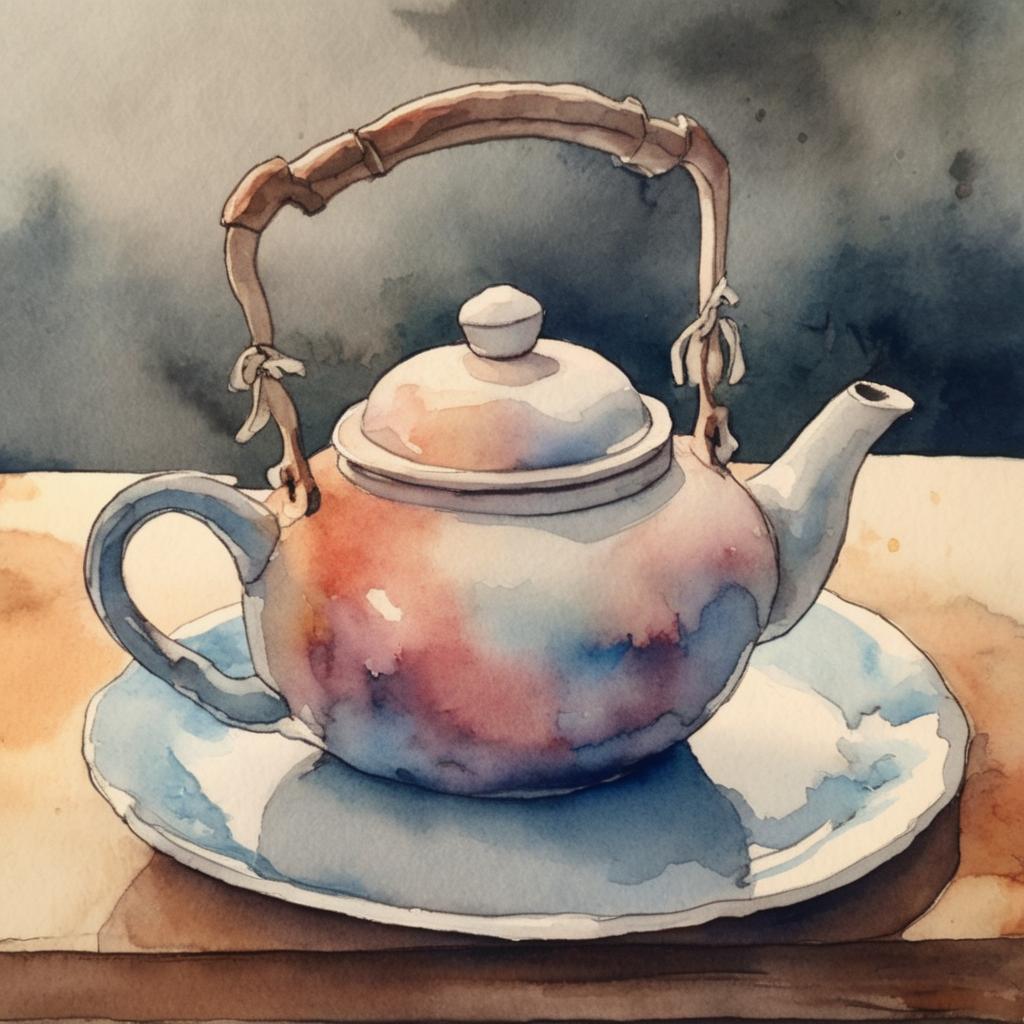} &
        \tabfigure{width=0.12\textwidth}{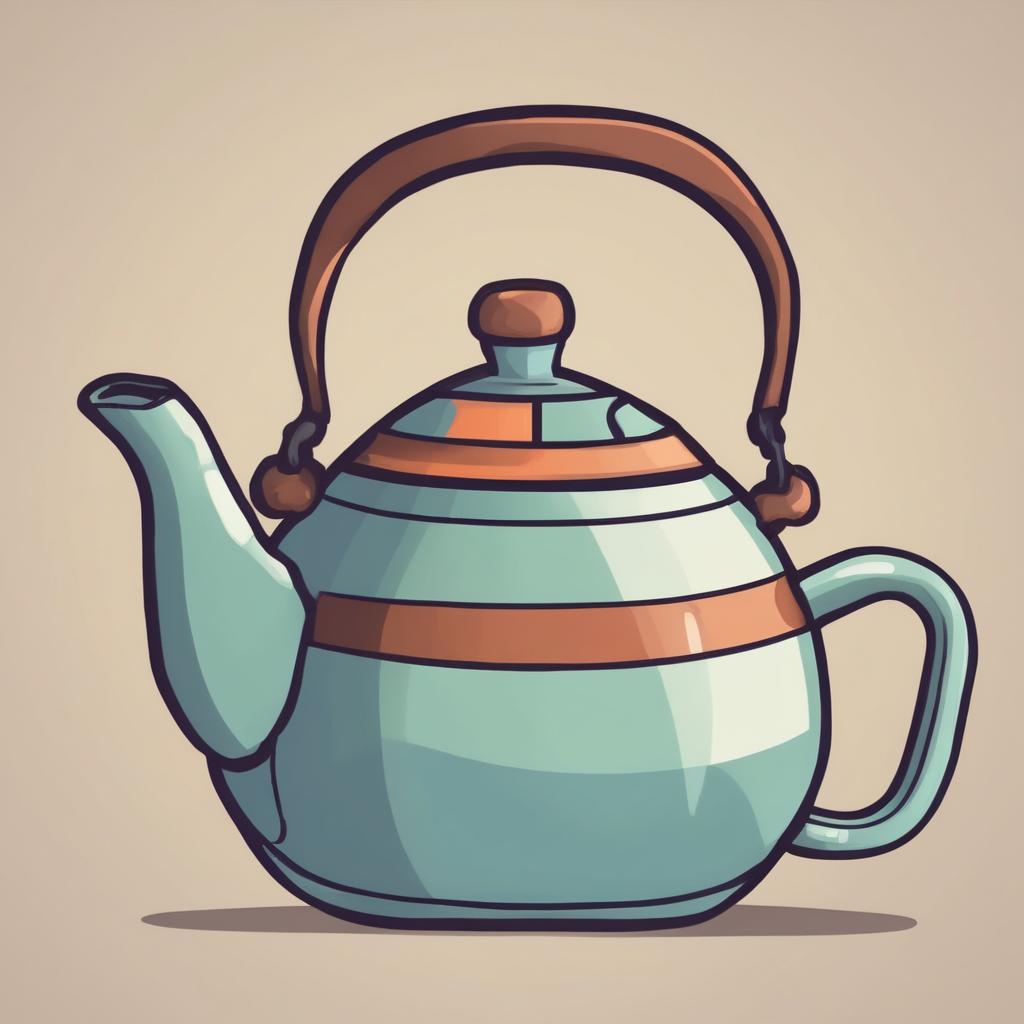} &
        \tabfigure{width=0.12\textwidth}{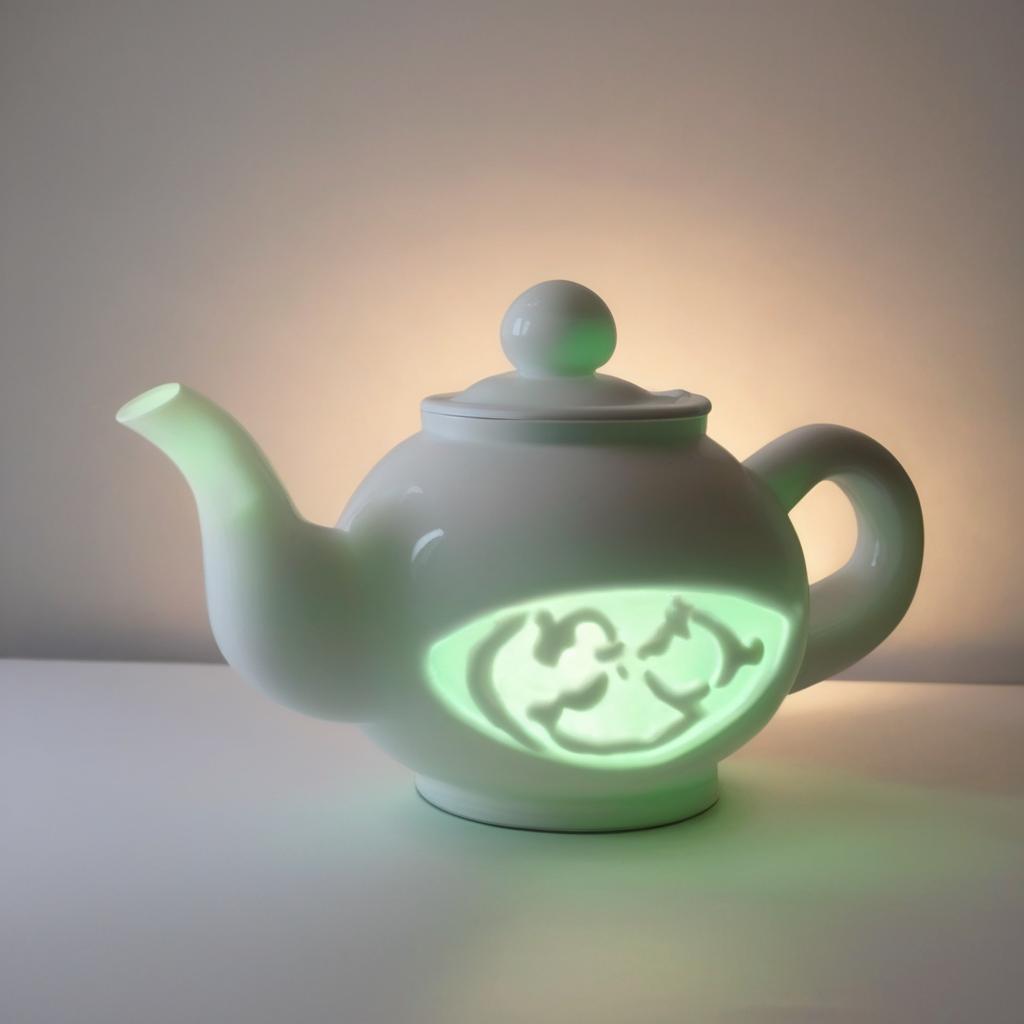} \\
    \begin{minipage}[c][1.7cm][c]{\linewidth} %
        \centering DARE-TIES
    \end{minipage}
    &
        \tabfigure{width=0.12\textwidth}{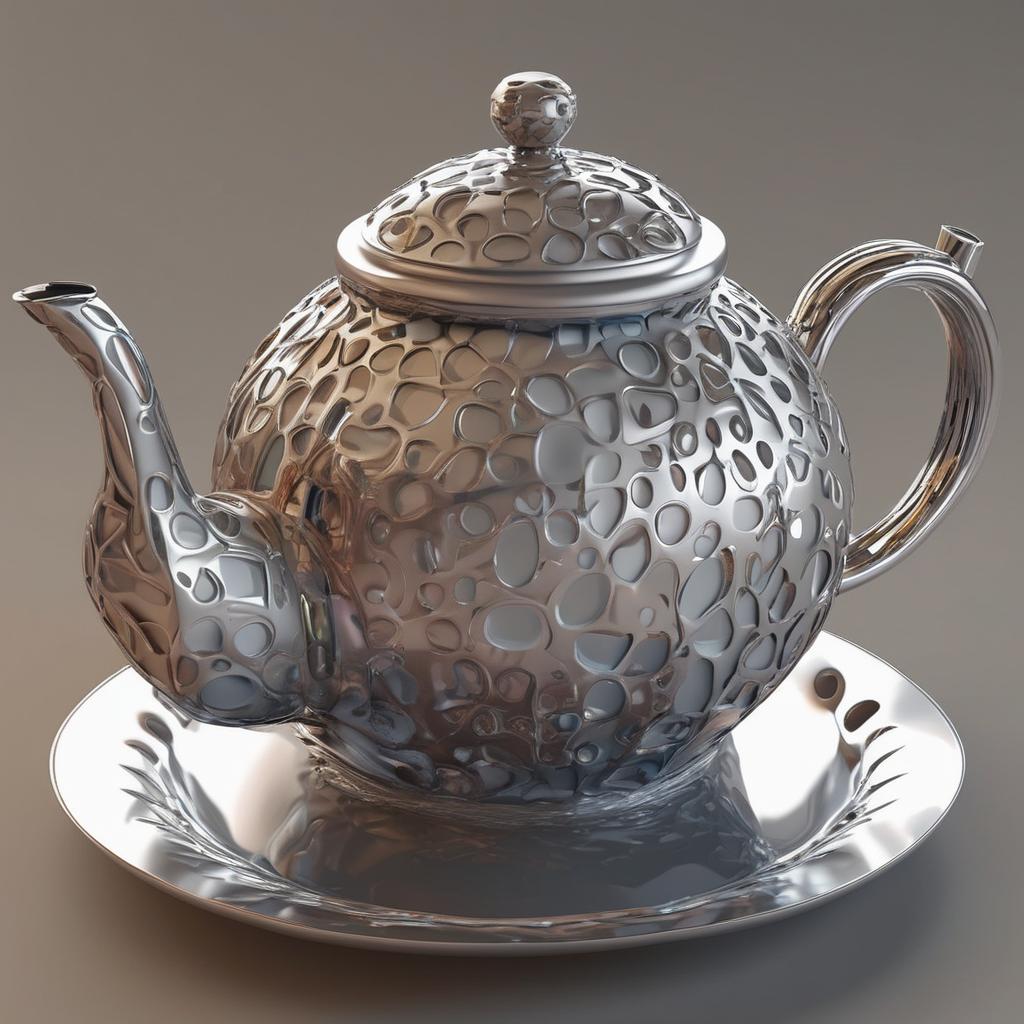} &
        \tabfigure{width=0.12\textwidth}{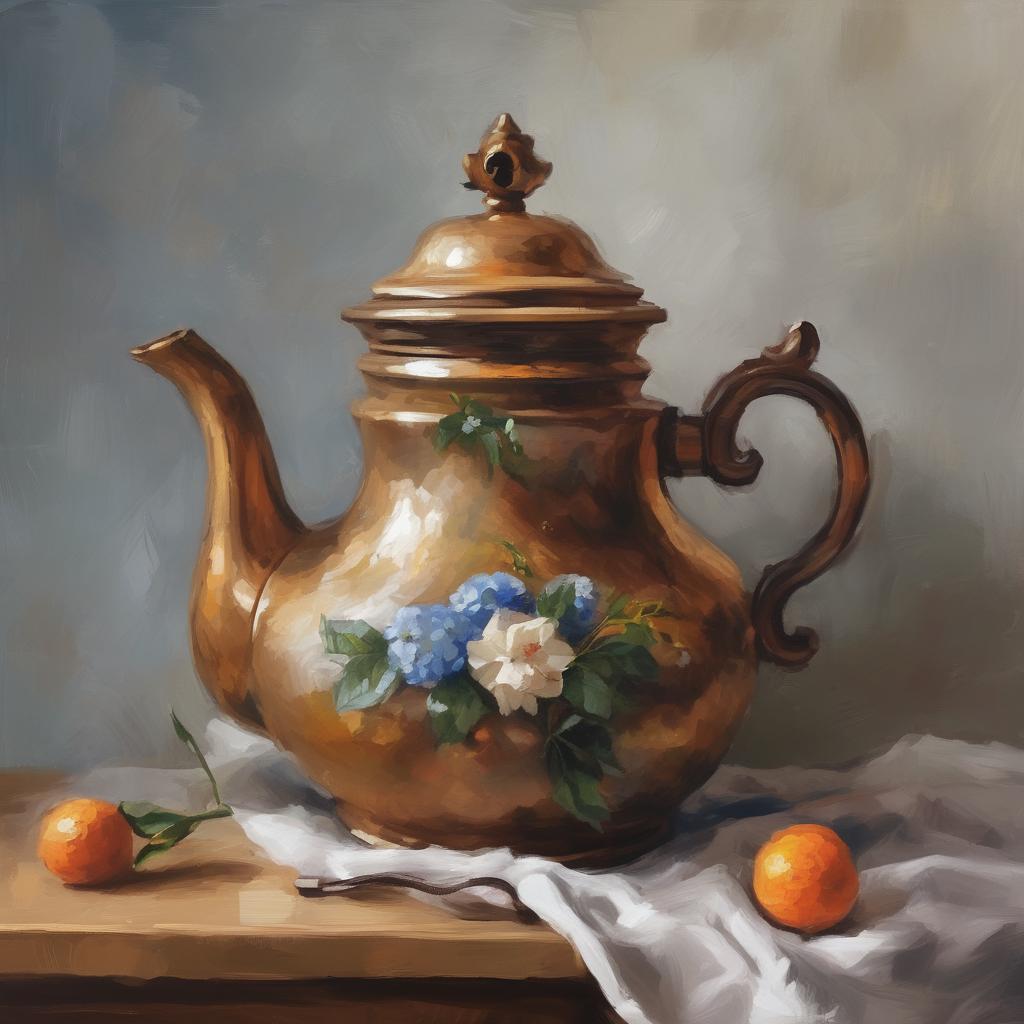} &
        \tabfigure{width=0.12\textwidth}{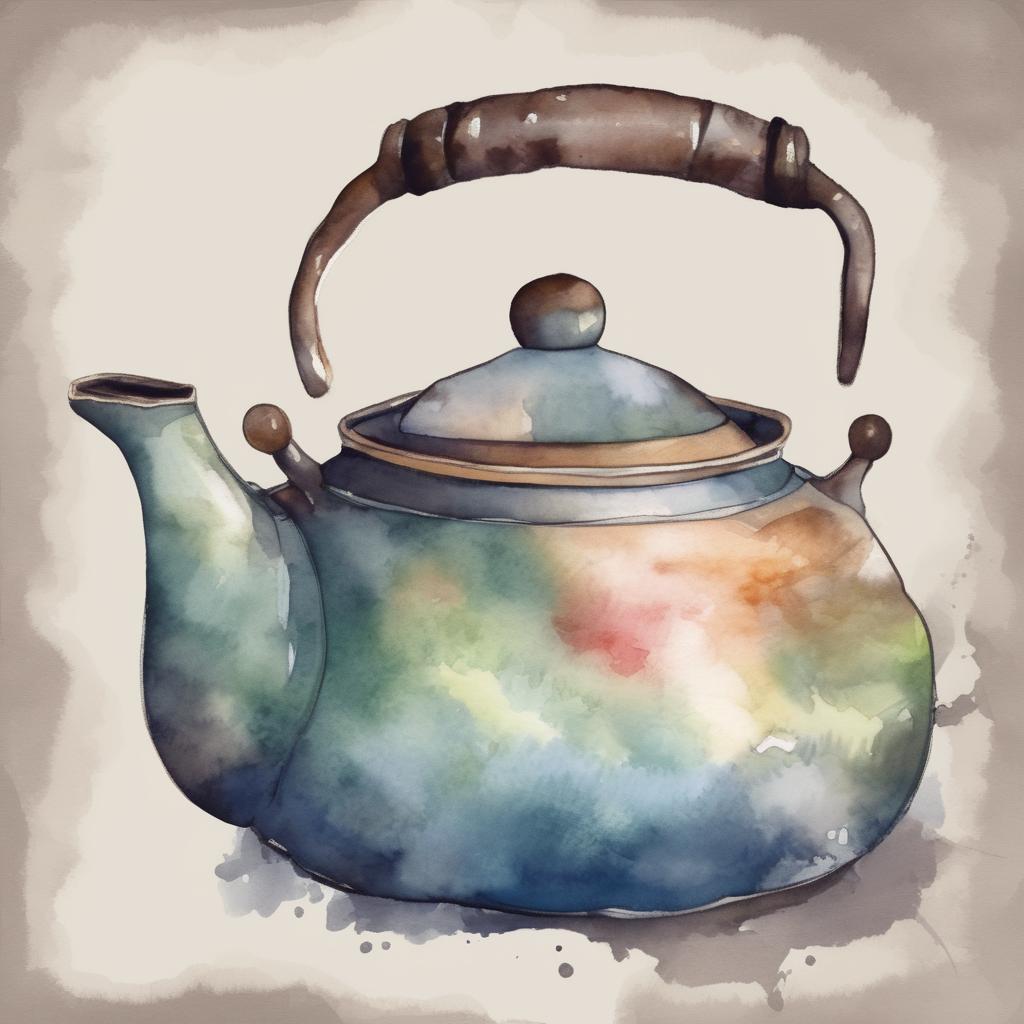} &
        \tabfigure{width=0.12\textwidth}{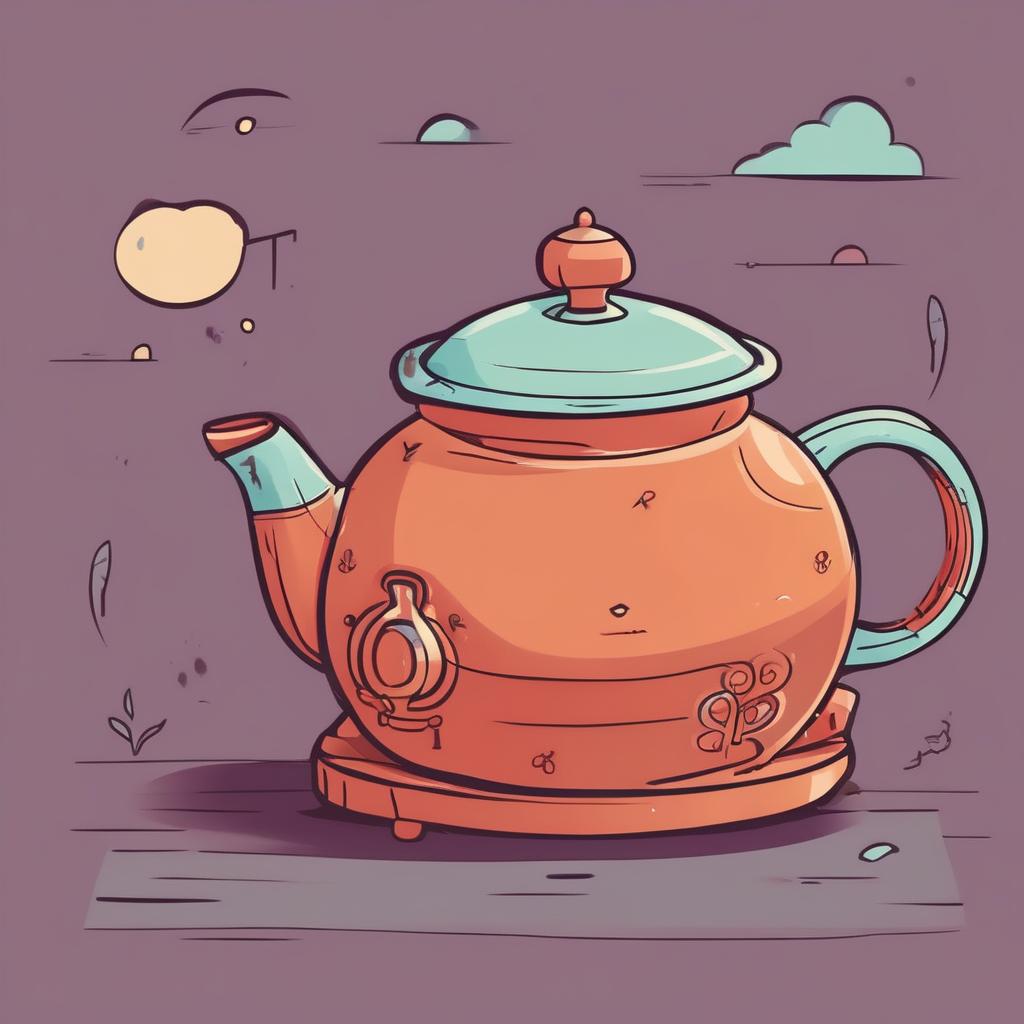} &
        \tabfigure{width=0.12\textwidth}{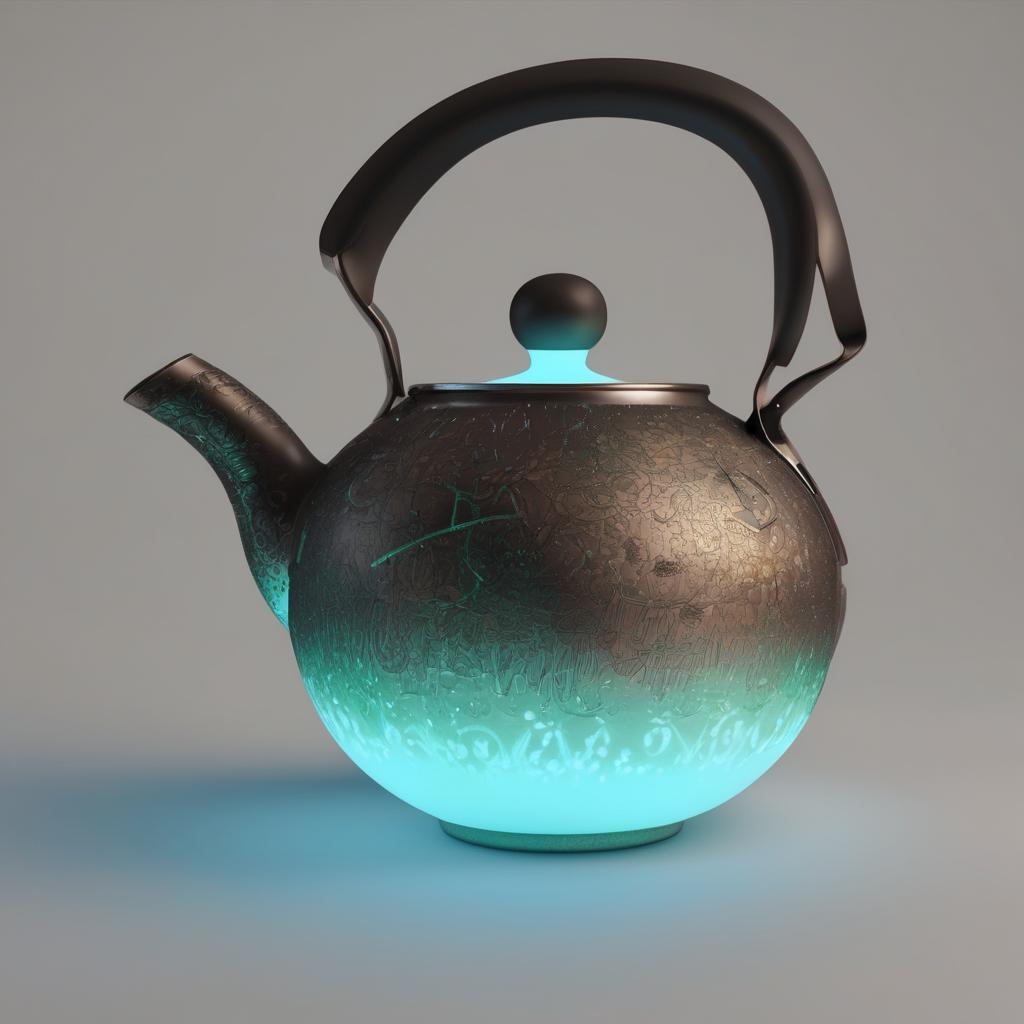} \\
    \begin{minipage}[c][1.7cm][c]{\linewidth} %
        \centering \ziplora
    \end{minipage}
    &
        \tabfigure{width=0.12\textwidth}{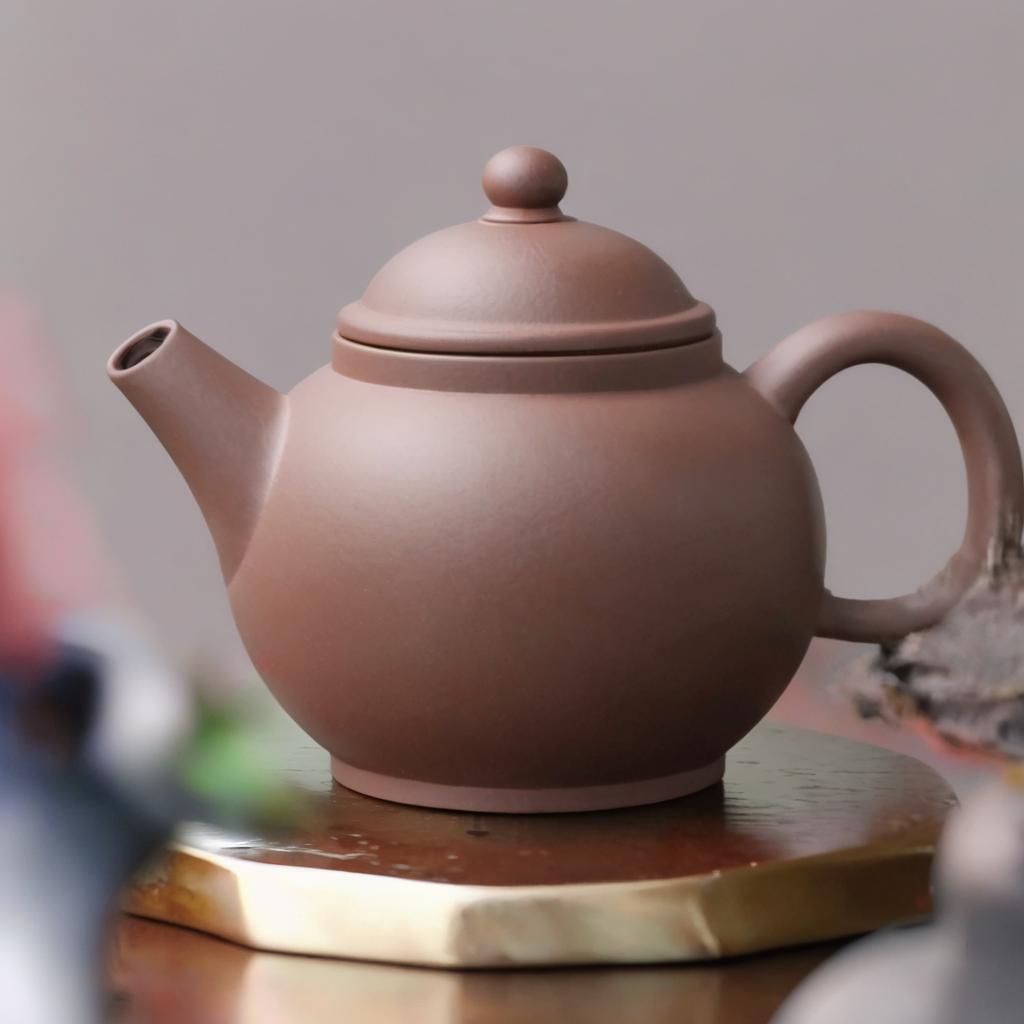} &
        \tabfigure{width=0.12\textwidth}{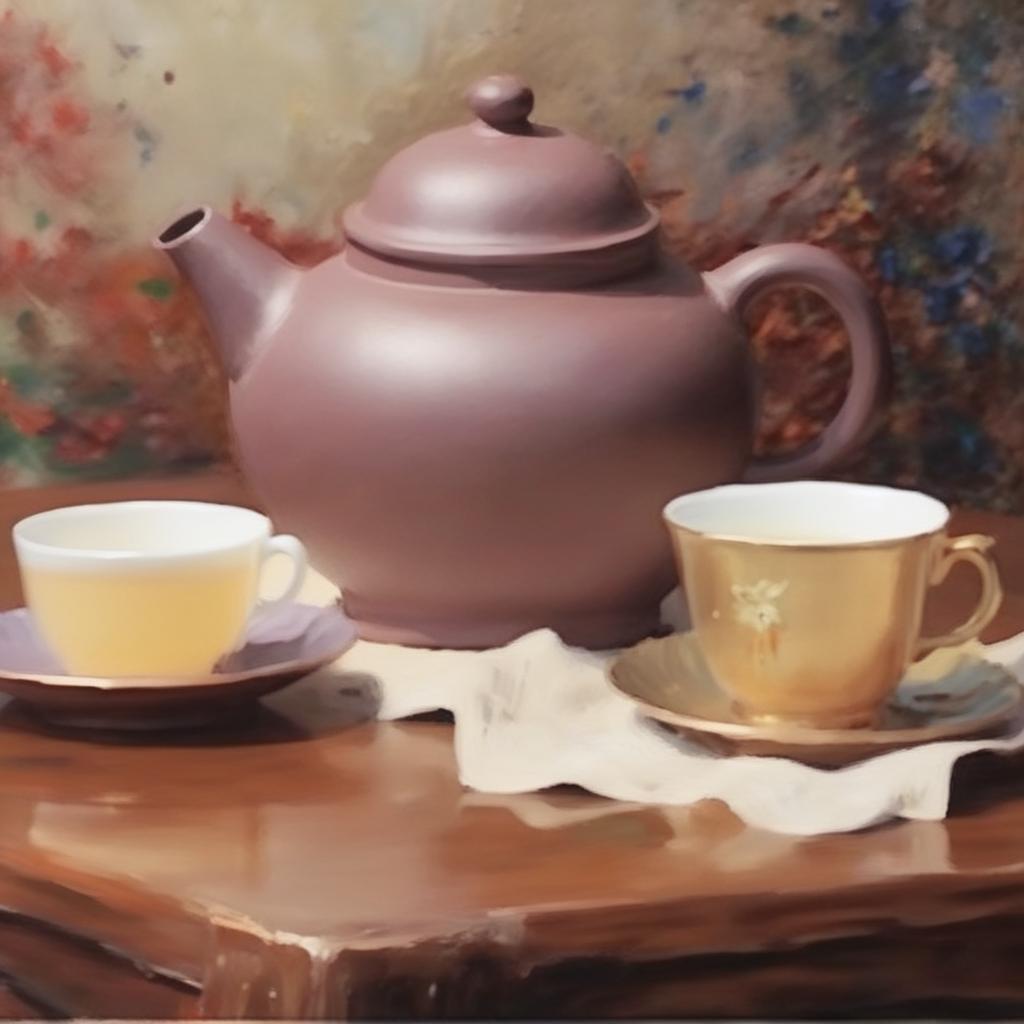} &
        \tabfigure{width=0.12\textwidth}{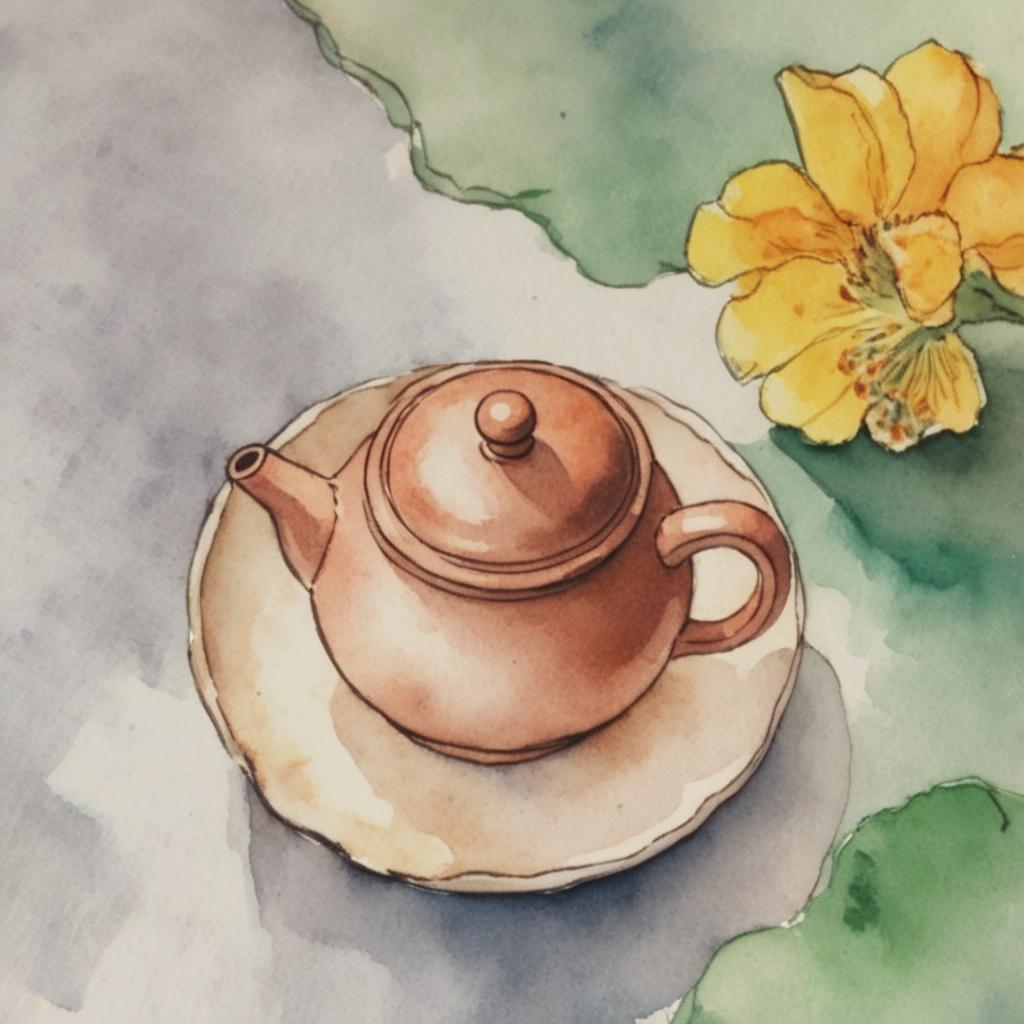} &
        \tabfigure{width=0.12\textwidth}{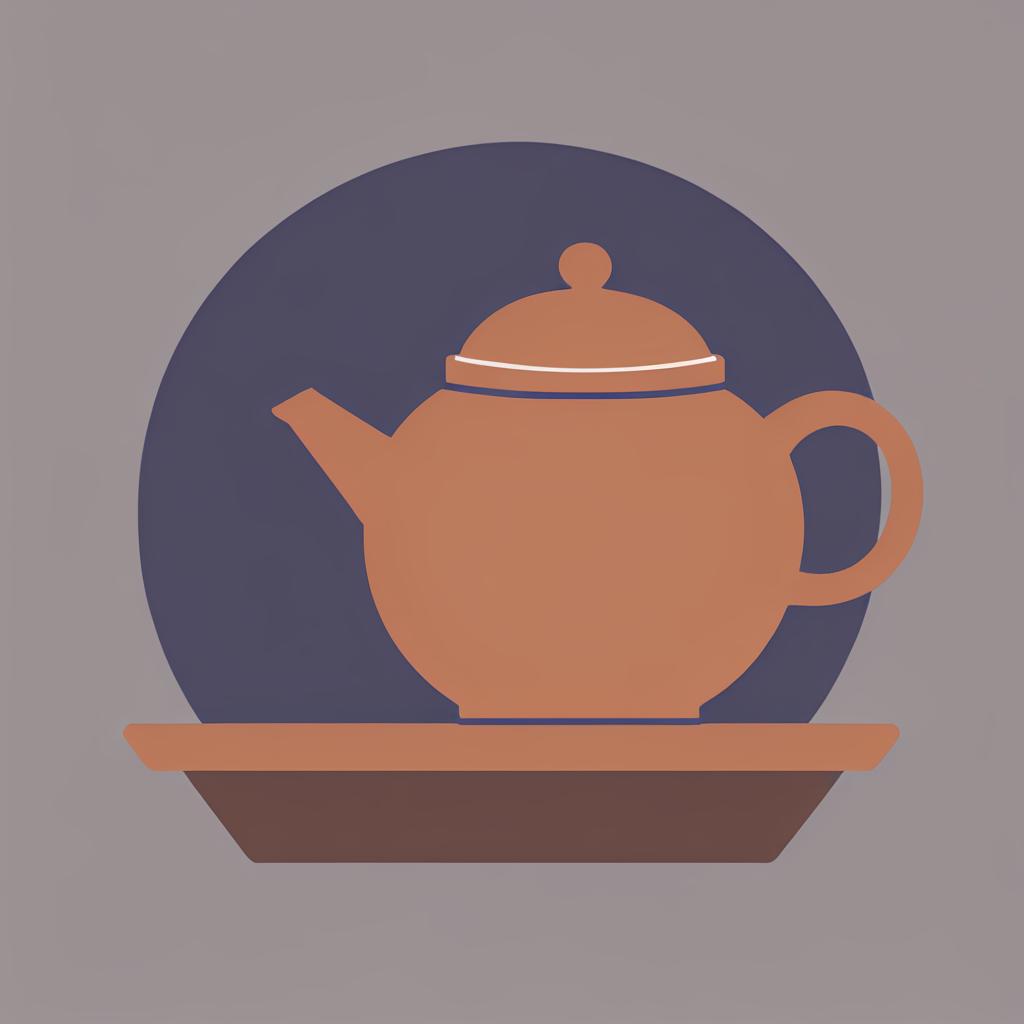} &
        \tabfigure{width=0.12\textwidth}{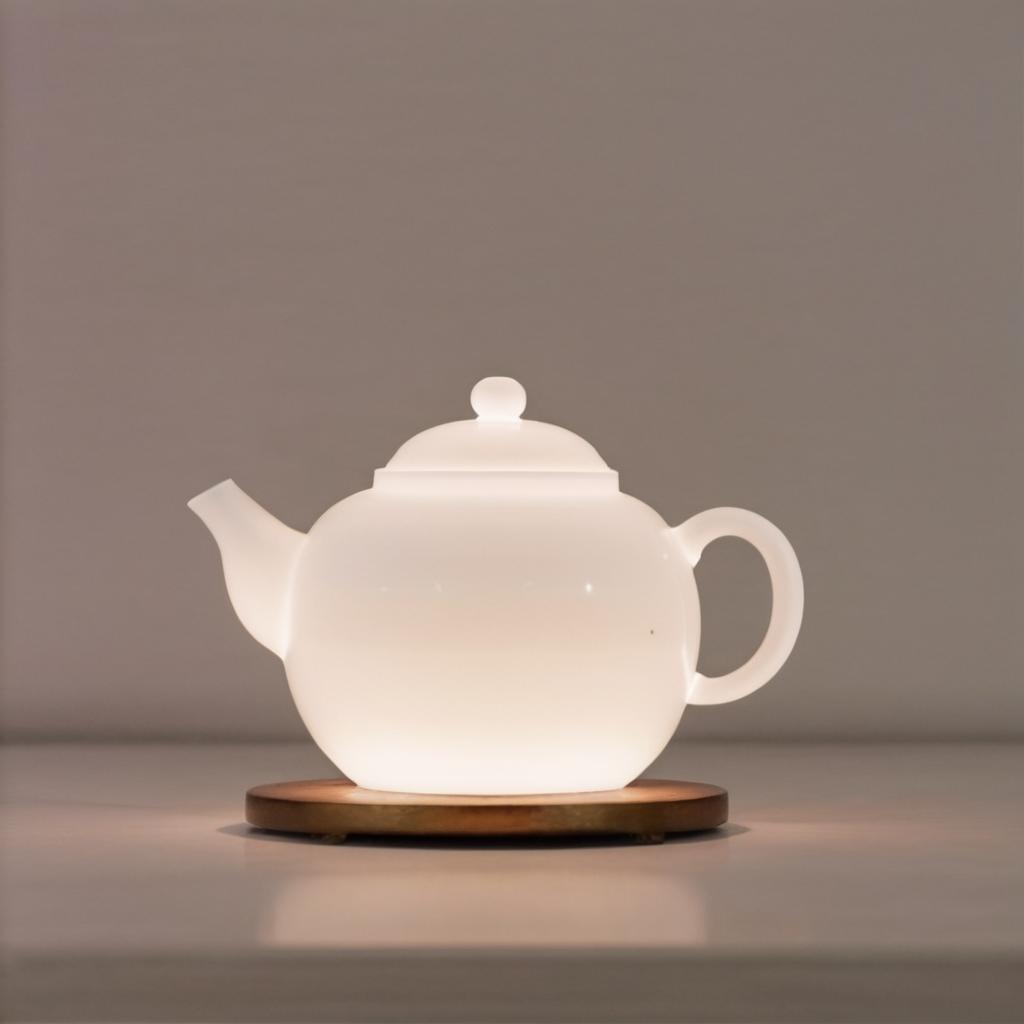} \\
    \begin{minipage}[c][1.7cm][c]{\linewidth} %
        \centering \textbf{\ours (ours)}
    \end{minipage}
        &
        \tabfigure{width=0.12\textwidth}{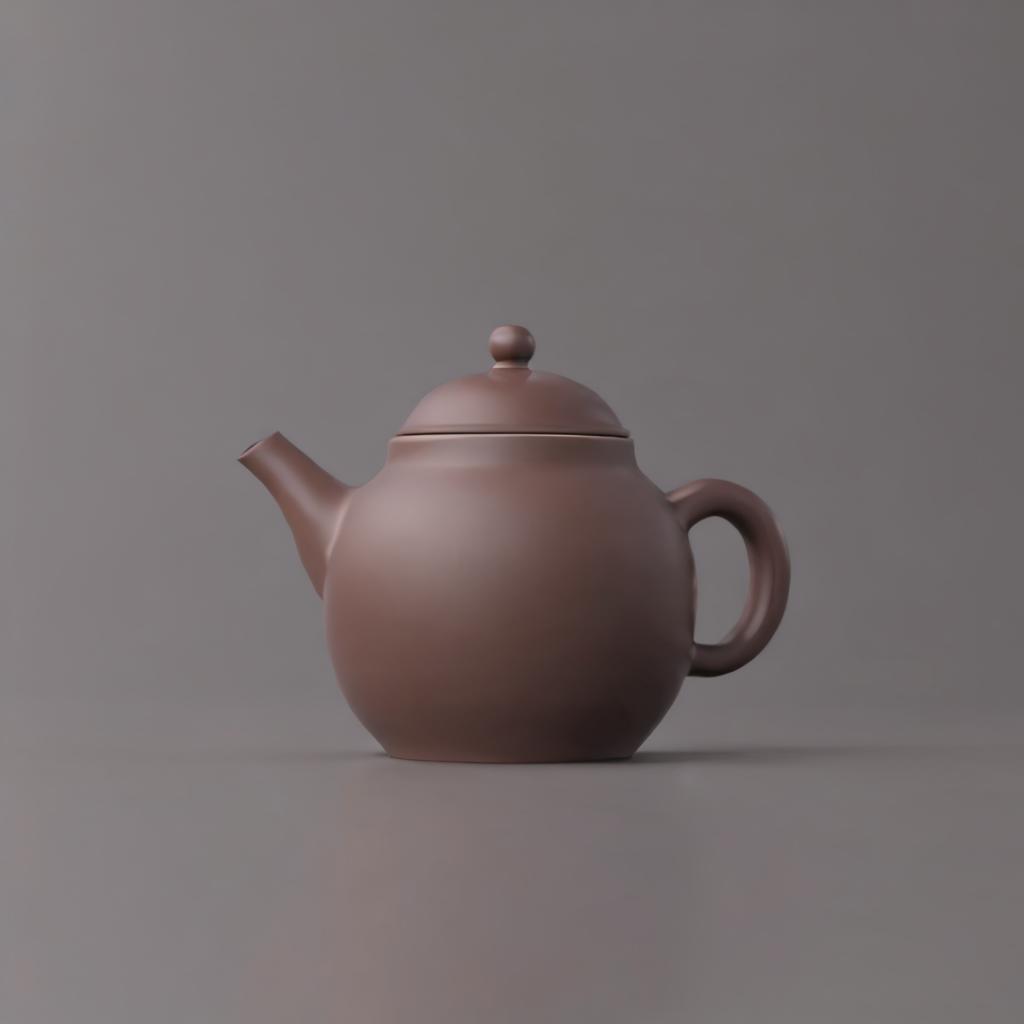} &
        \tabfigure{width=0.12\textwidth}{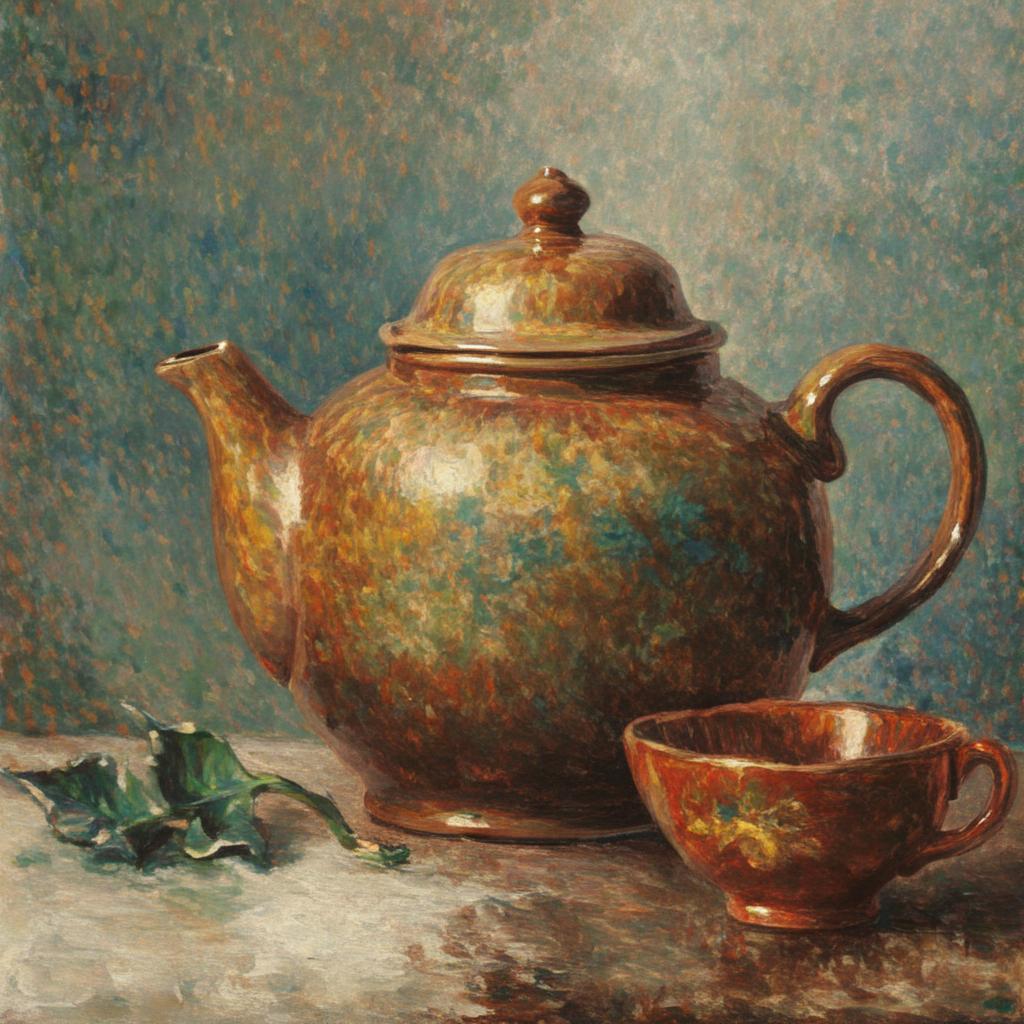} &
        \tabfigure{width=0.12\textwidth}{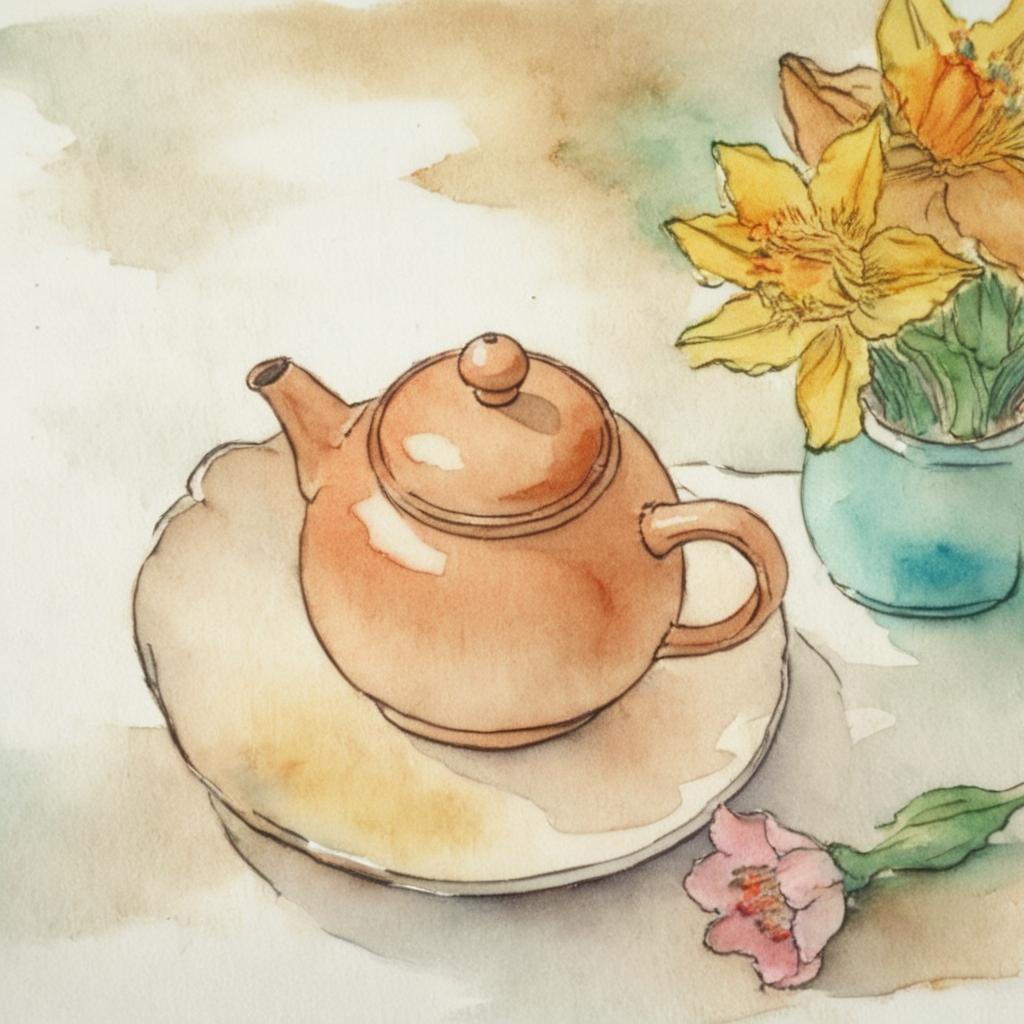} &
        \tabfigure{width=0.12\textwidth}{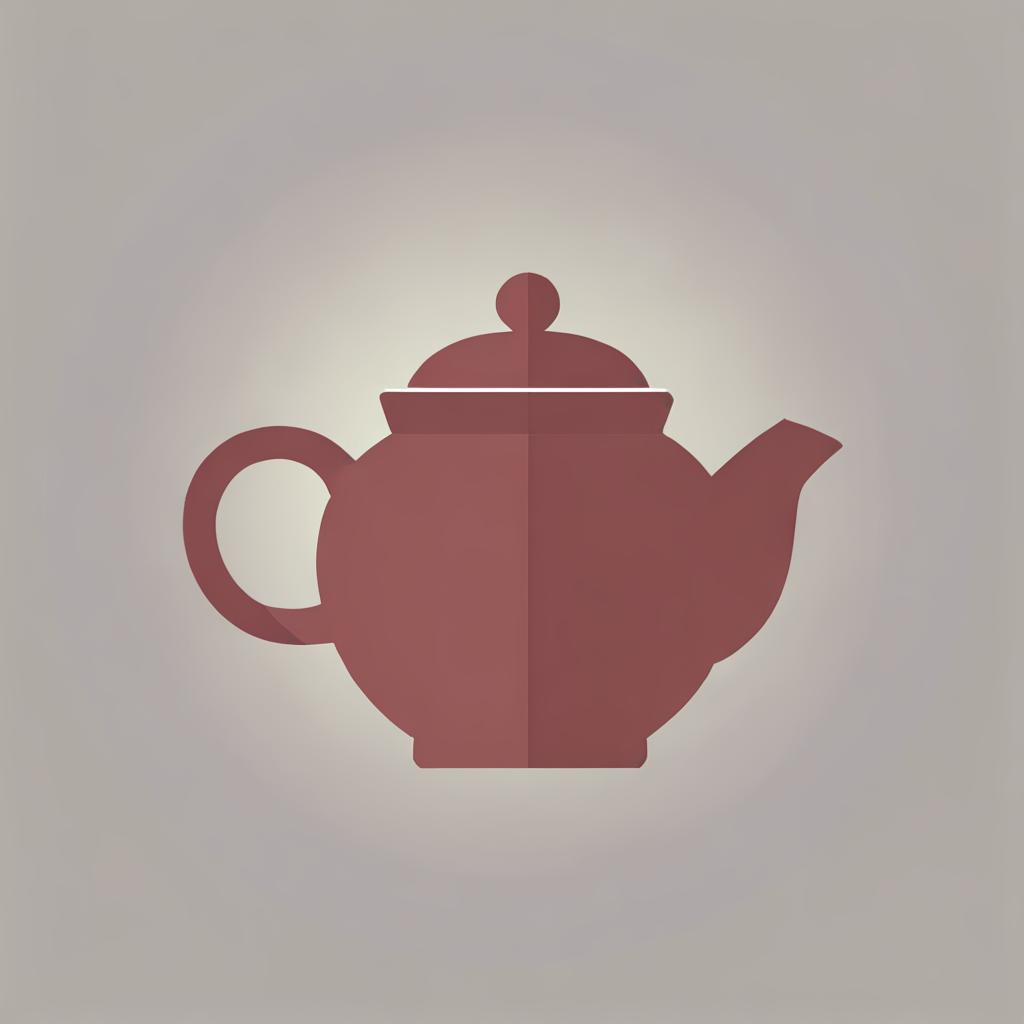} &
        \tabfigure{width=0.12\textwidth}{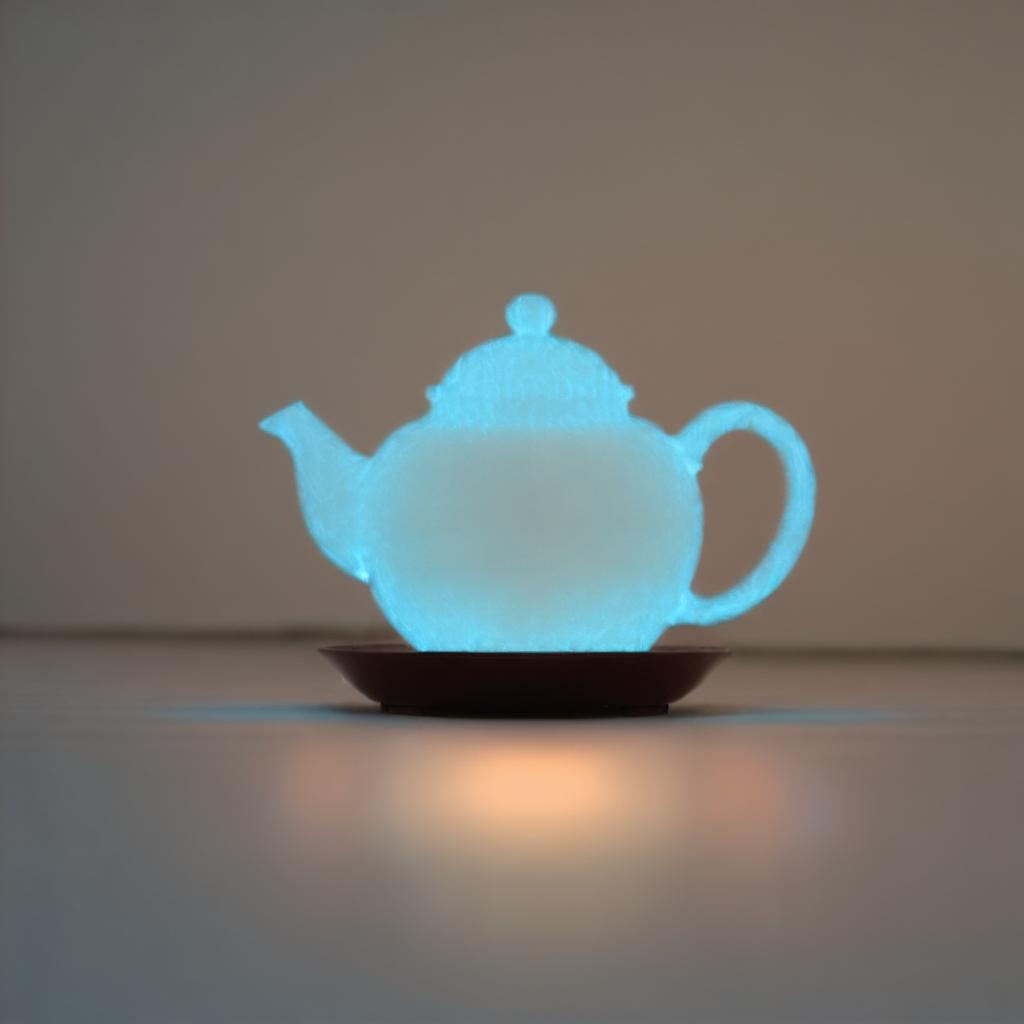} \\
    \end{tabular}
    }
    \resizebox{0.46\textwidth}{!}{%
    \begin{tabular}[c]{L L L L L L L}
     \small A \textit{[C] stuffed animal} in \textit{[S]} style & 3D Rendering & Oil Painting & Watercolor Painting & Flat Cartoon Illustration & Glowing \\
        \tabfigure{width=0.12\textwidth}{figures/dataset/test/wolf_plushie/00.jpg}  & \tabfigure{width=0.12\textwidth}{figures/qualitative/3d_rendering.jpg} 
        & \tabfigure{width=0.12\textwidth}{figures/qualitative/oil.jpg}
        & \tabfigure{width=0.12\textwidth}{figures/qualitative/watercolor.jpg}
        & \tabfigure{width=0.12\textwidth}{figures/qualitative/flat.jpg}
        & \tabfigure{width=0.12\textwidth}{figures/qualitative/glowing.jpg} \\

    \begin{minipage}[c][1.7cm][c]{\linewidth} %
        \centering Joint Training 
    \end{minipage}
    &
        \tabfigure{width=0.12\textwidth}{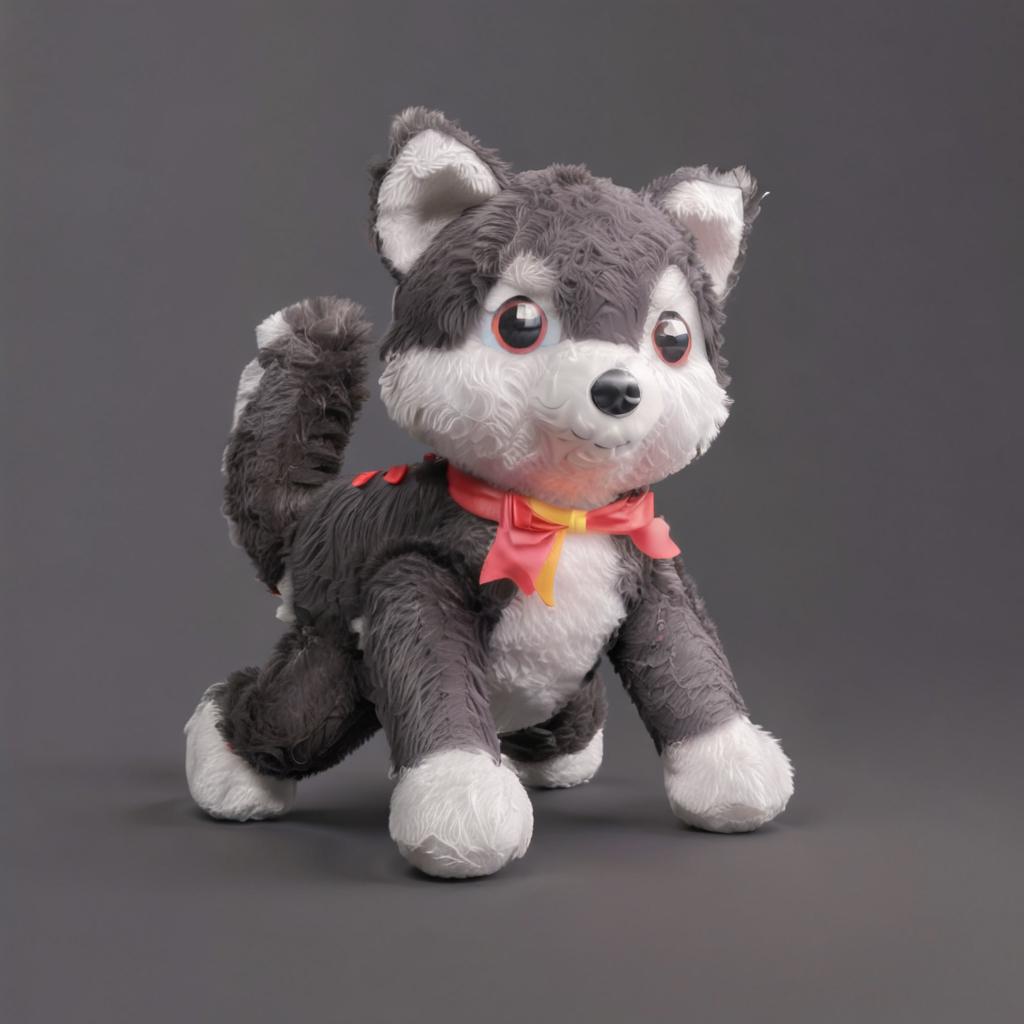} &
        \tabfigure{width=0.12\textwidth}{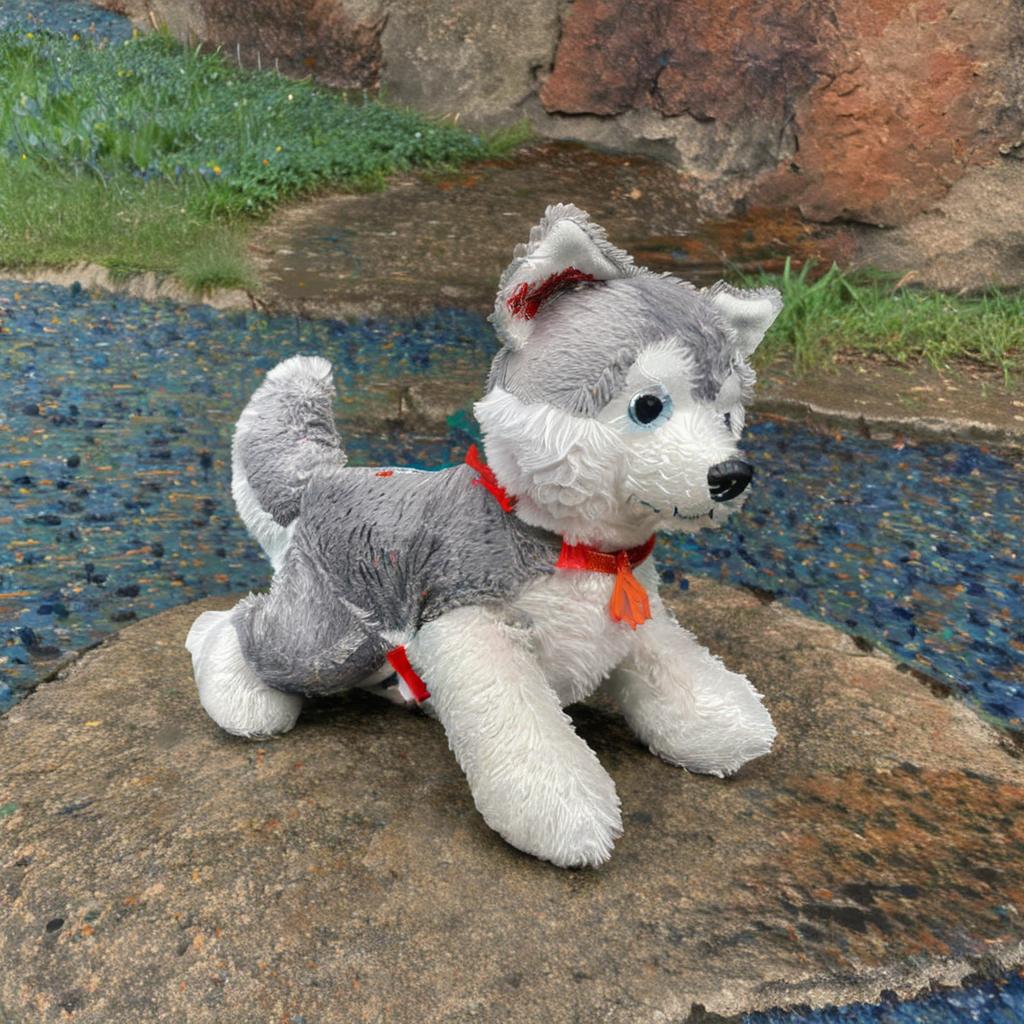} &
        \tabfigure{width=0.12\textwidth}{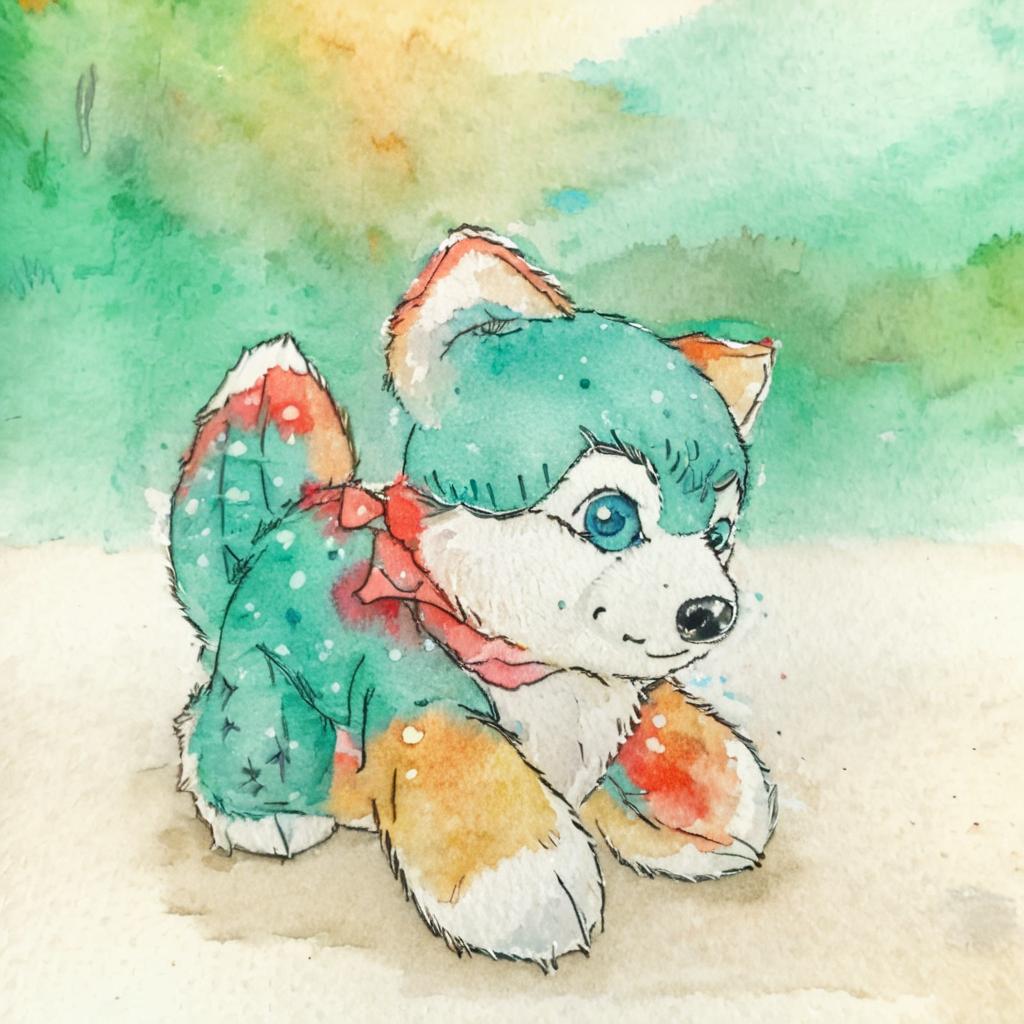} &
        \tabfigure{width=0.12\textwidth}{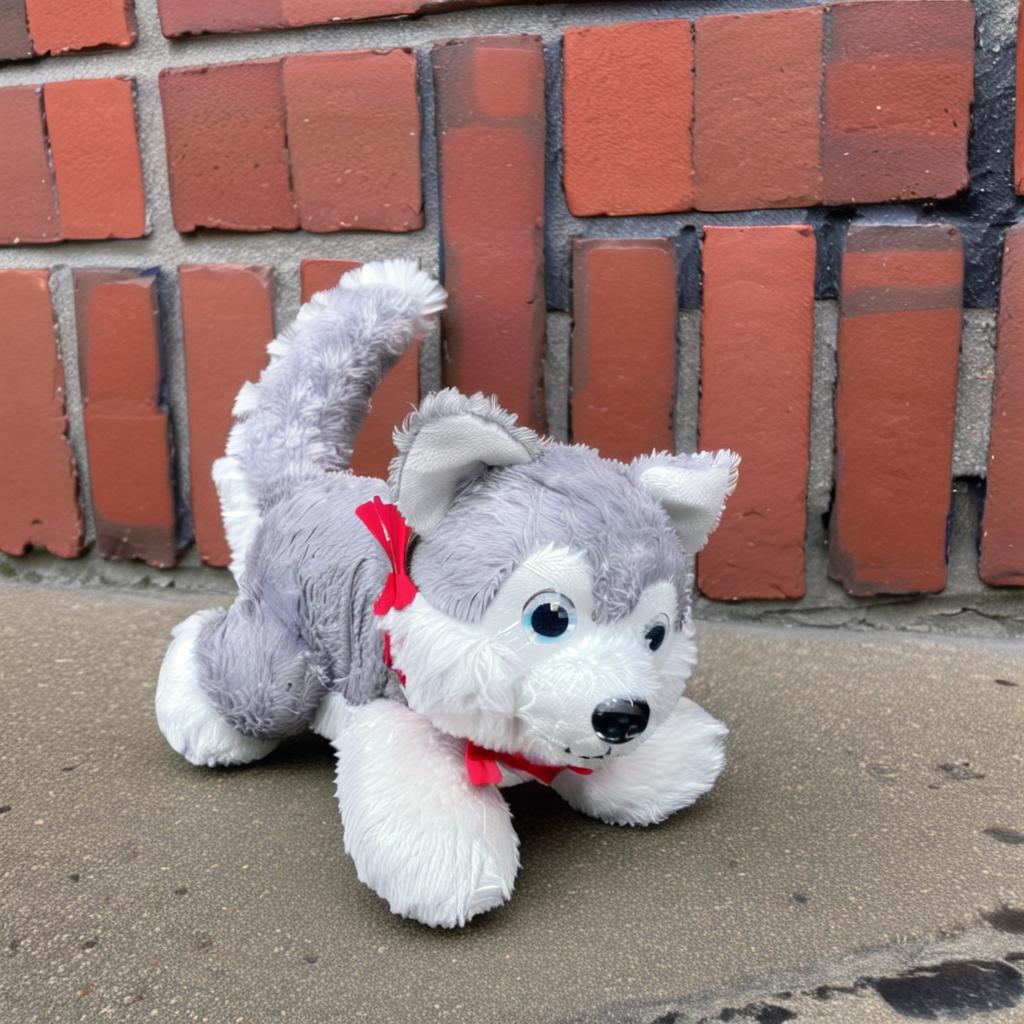} &
        \tabfigure{width=0.12\textwidth}{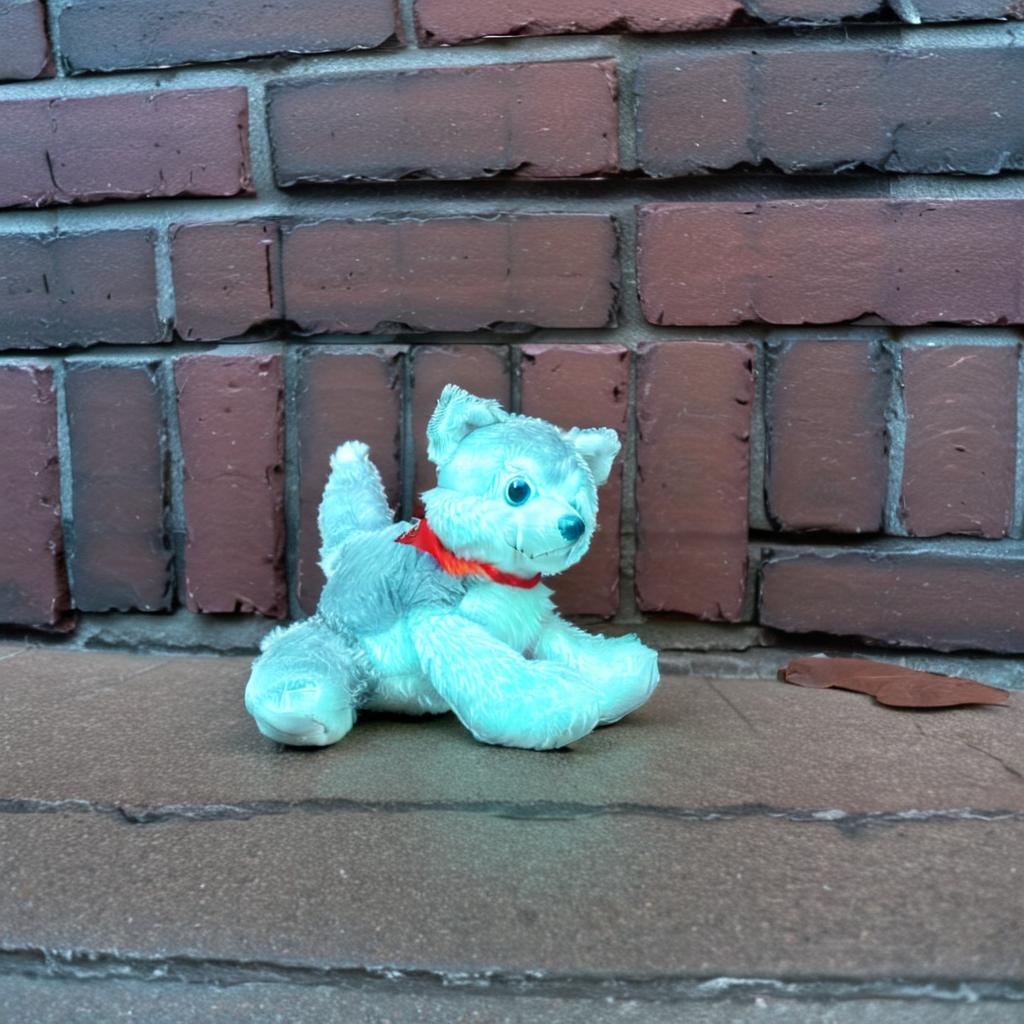} \\
     \begin{minipage}[c][1.7cm][c]{\linewidth} %
        \centering Direct Merge
    \end{minipage}
    &
        \tabfigure{width=0.12\textwidth}{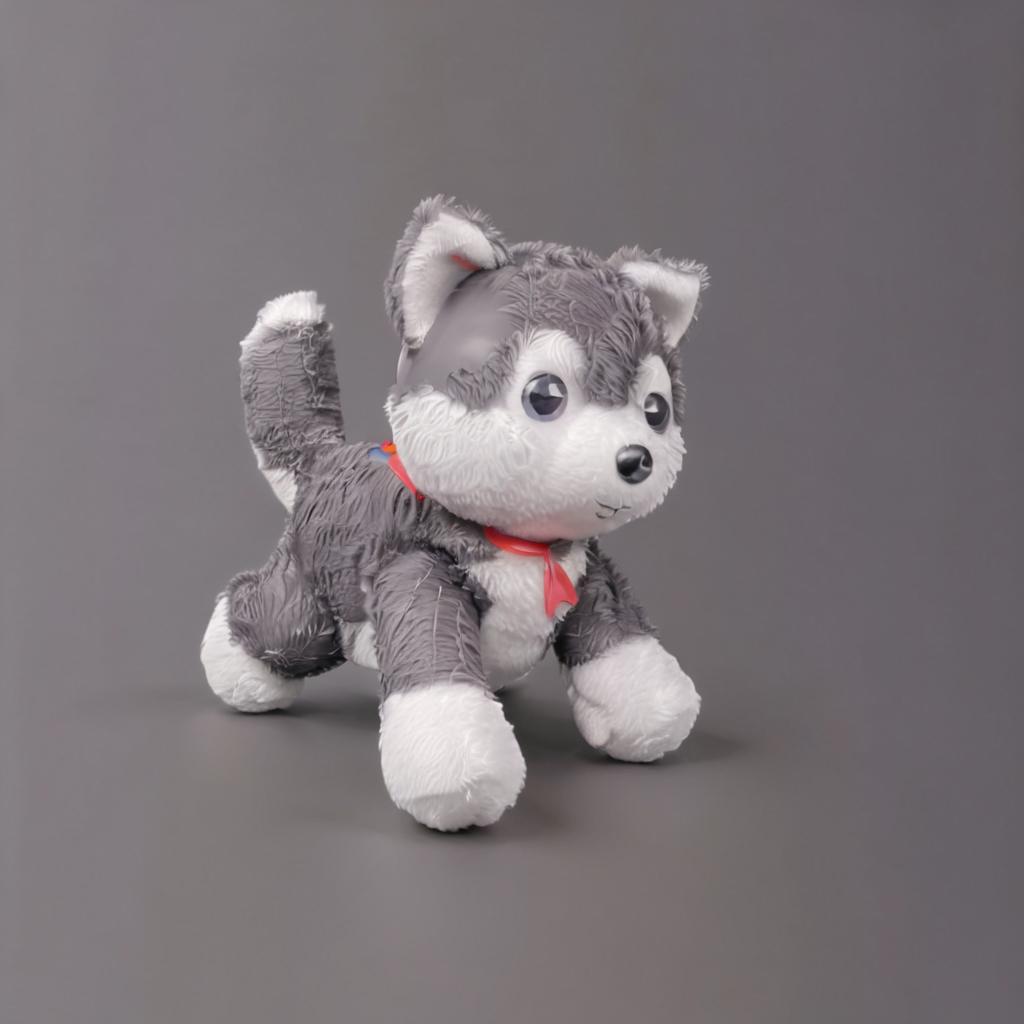} &
        \tabfigure{width=0.12\textwidth}{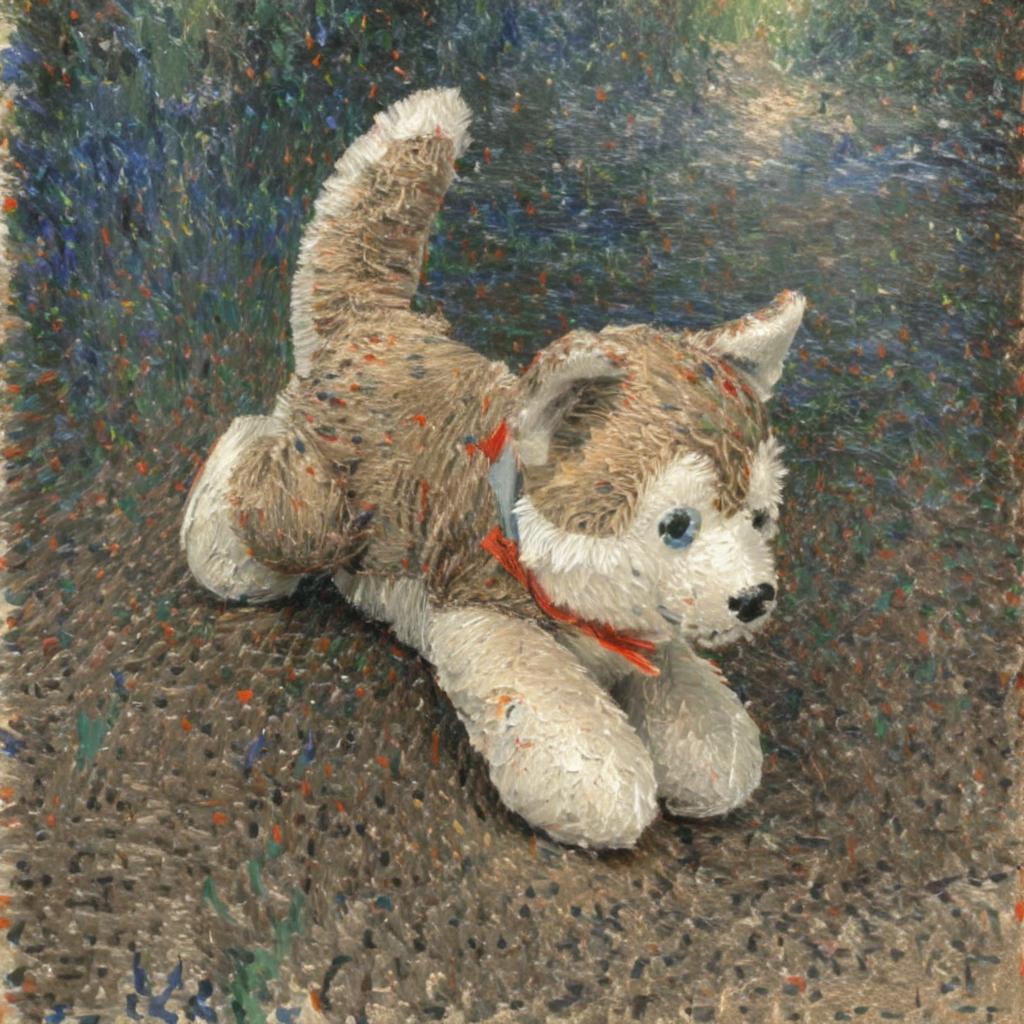} &
        \tabfigure{width=0.12\textwidth}{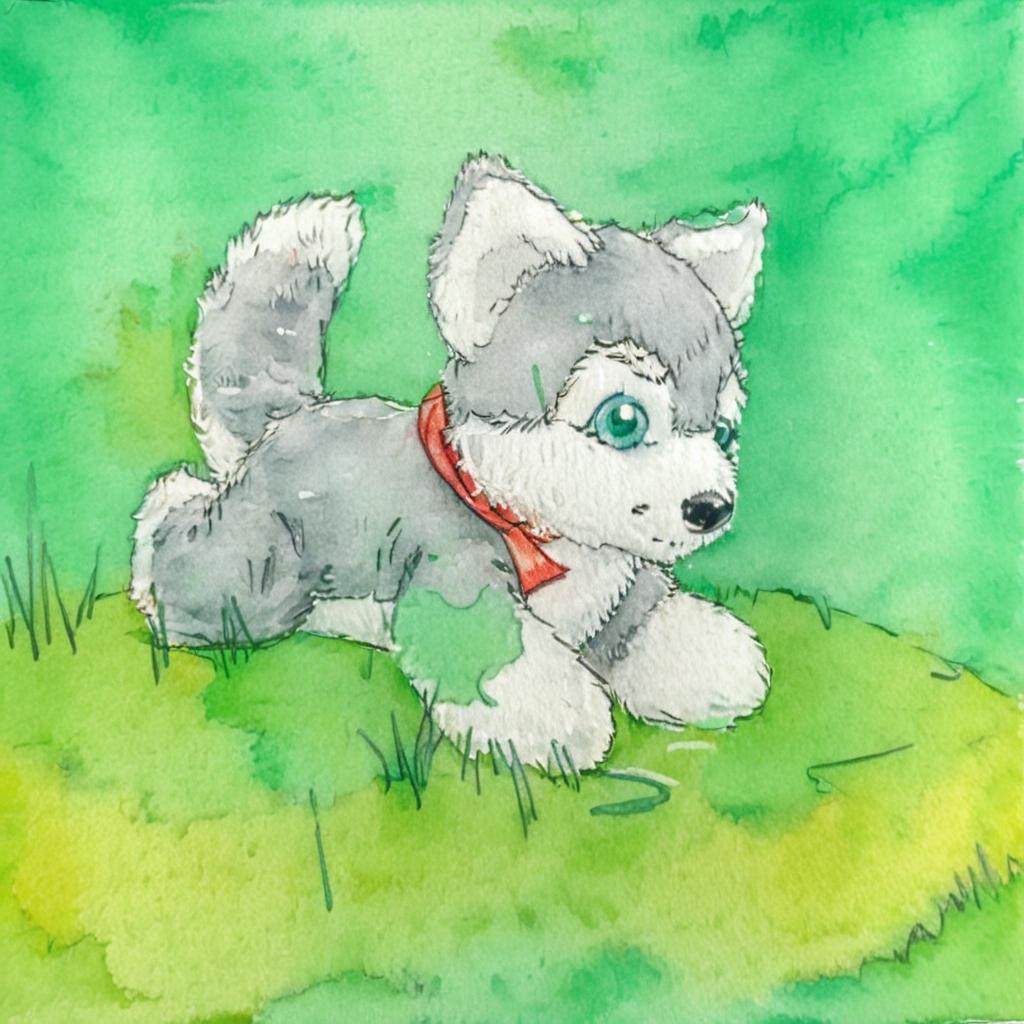} &
        \tabfigure{width=0.12\textwidth}{figures/qualitative/wolf/direct_merge/flat_cartoon.jpg} &
        \tabfigure{width=0.12\textwidth}{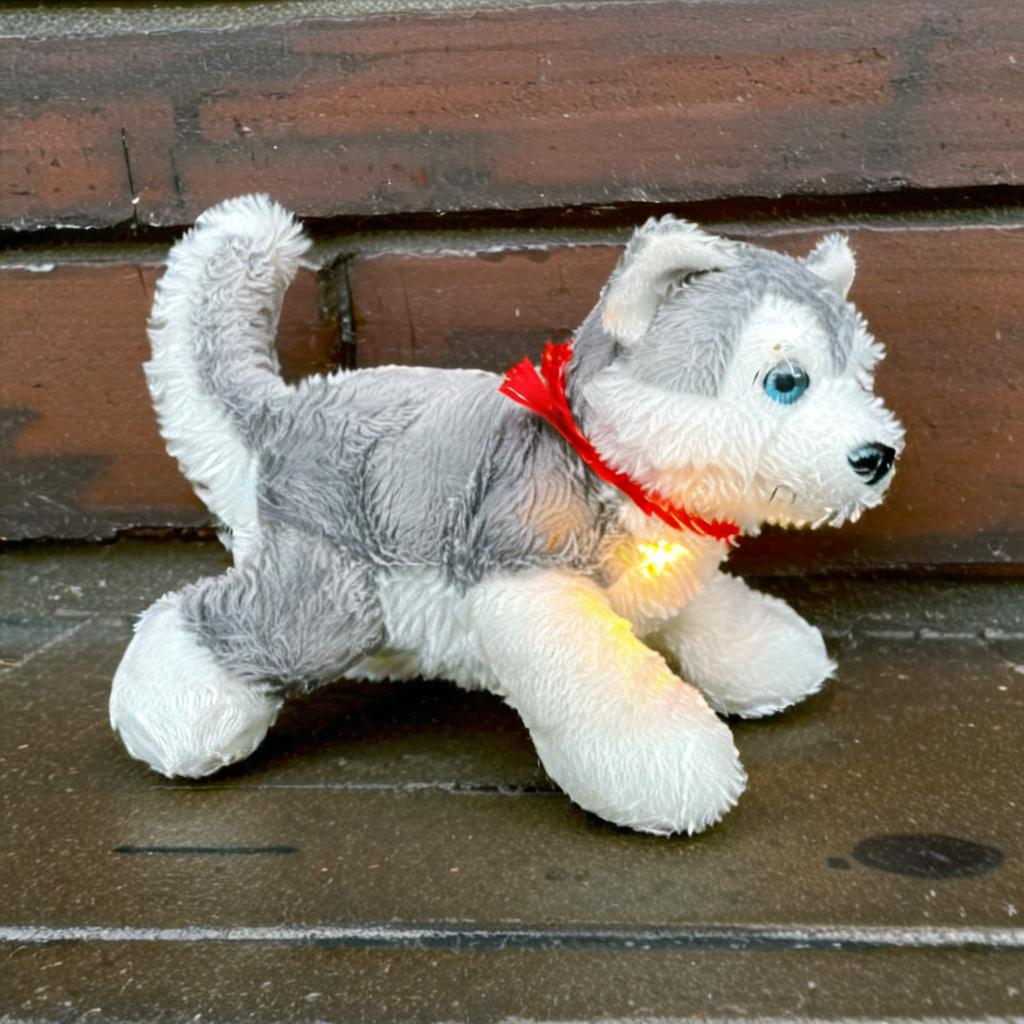} \\
     \begin{minipage}[c][1.7cm][c]{\linewidth} %
        \centering DARE
    \end{minipage}
    &
        \tabfigure{width=0.12\textwidth}{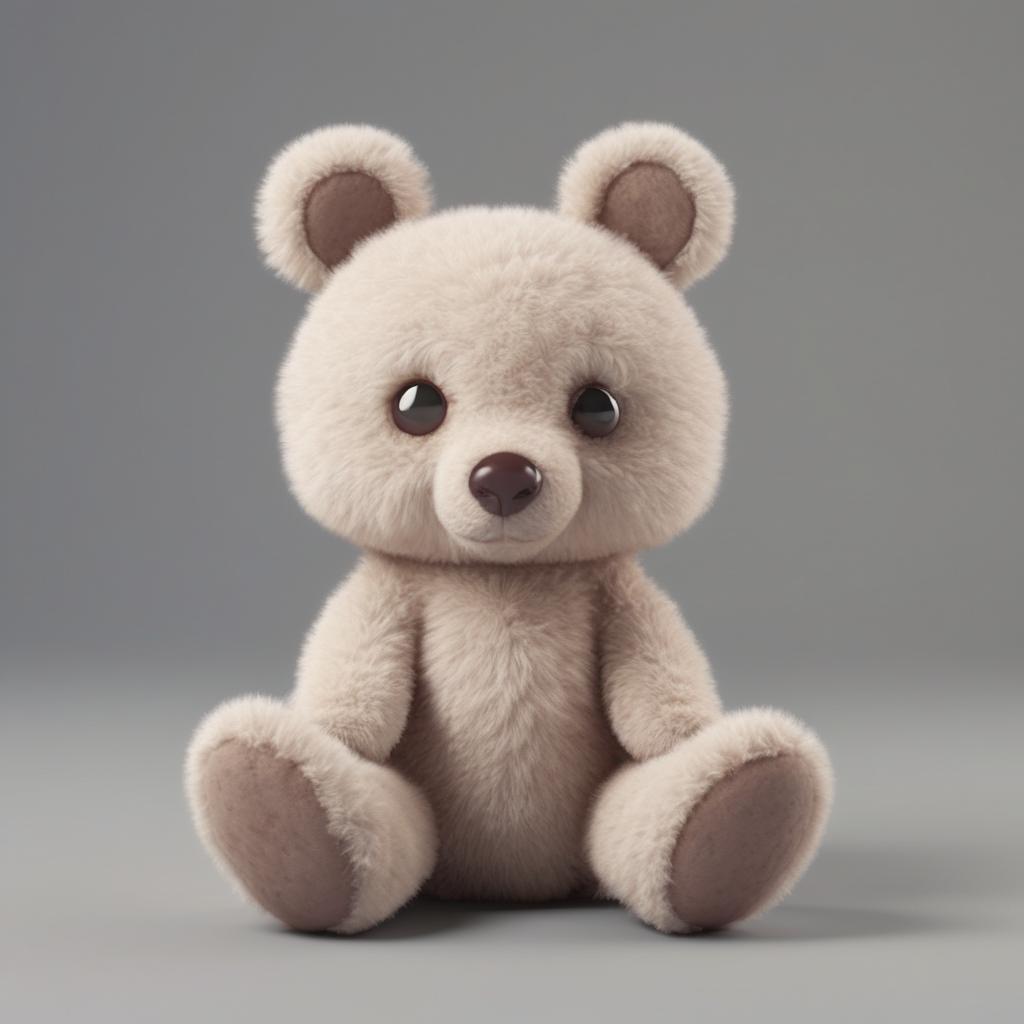} &
        \tabfigure{width=0.12\textwidth}{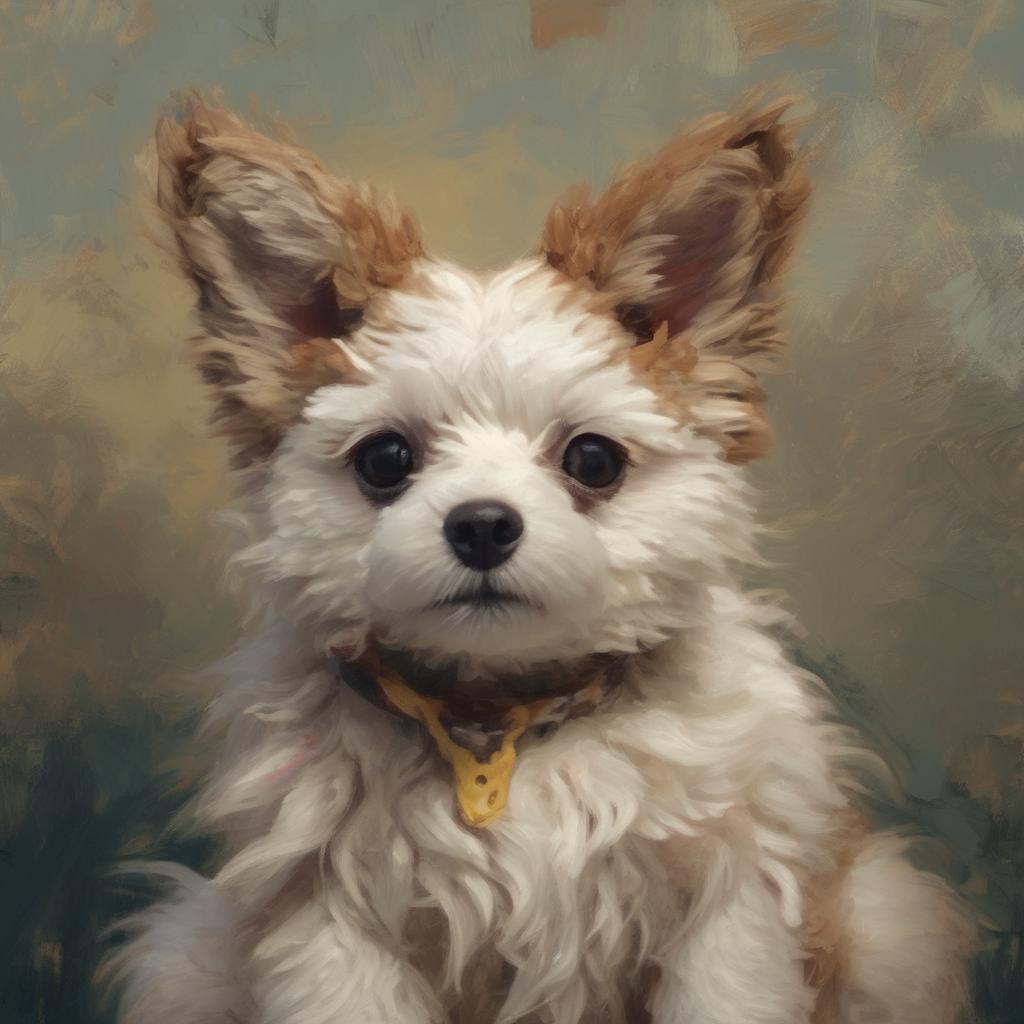} &
        \tabfigure{width=0.12\textwidth}{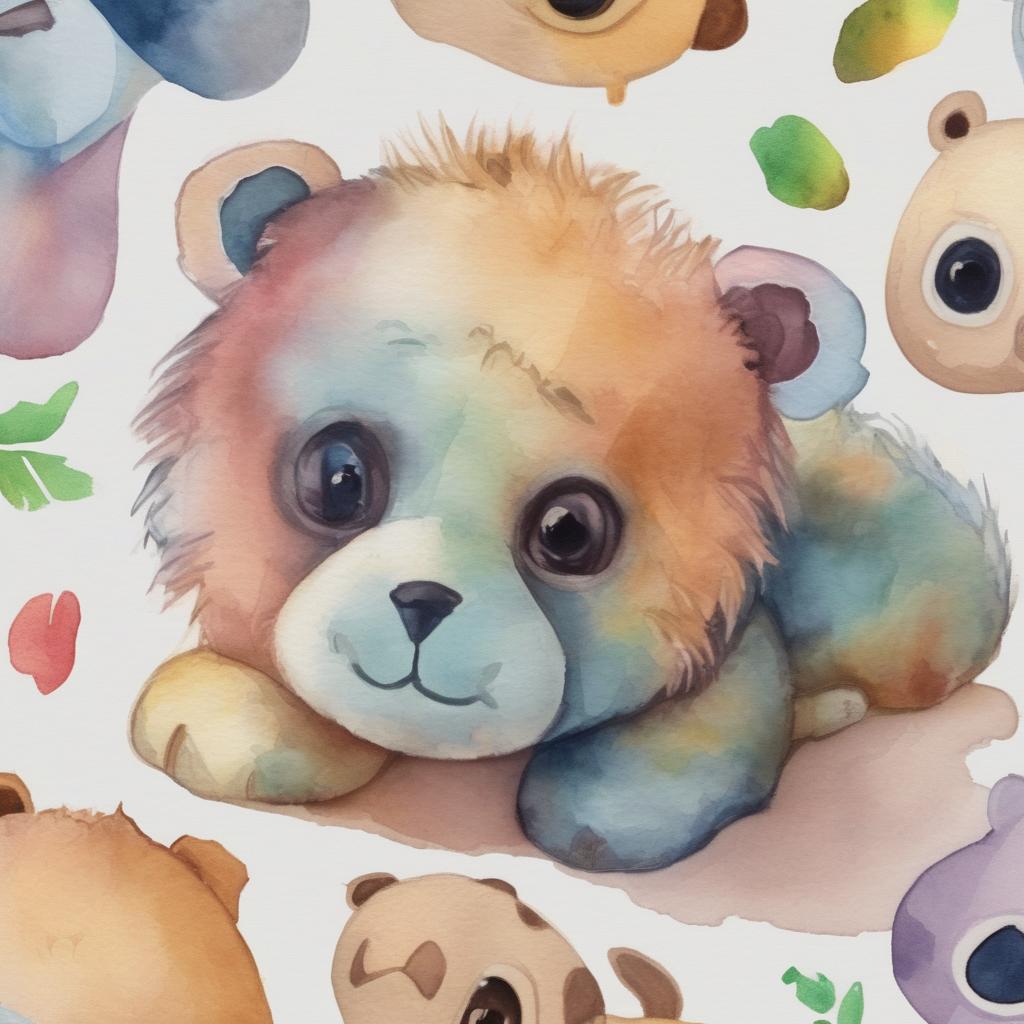} &
        \tabfigure{width=0.12\textwidth}{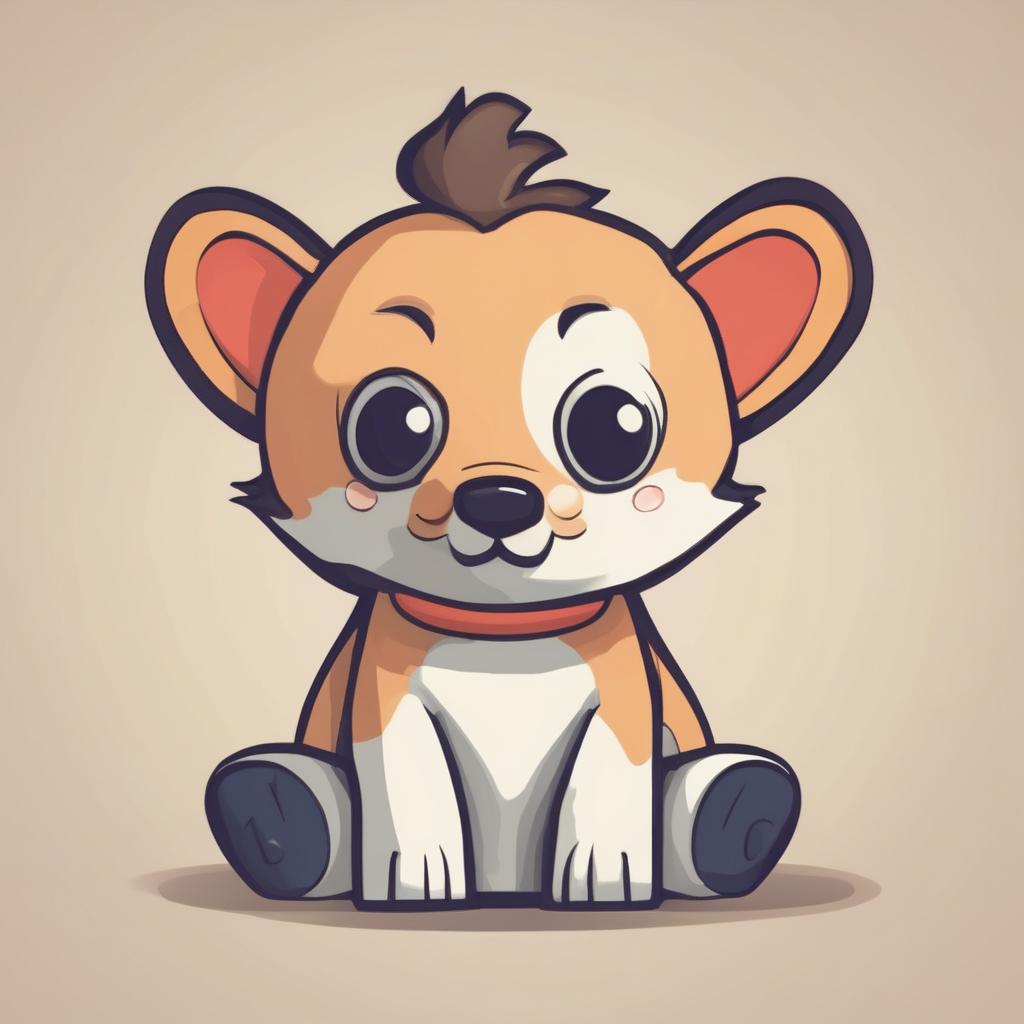} &
        \tabfigure{width=0.12\textwidth}{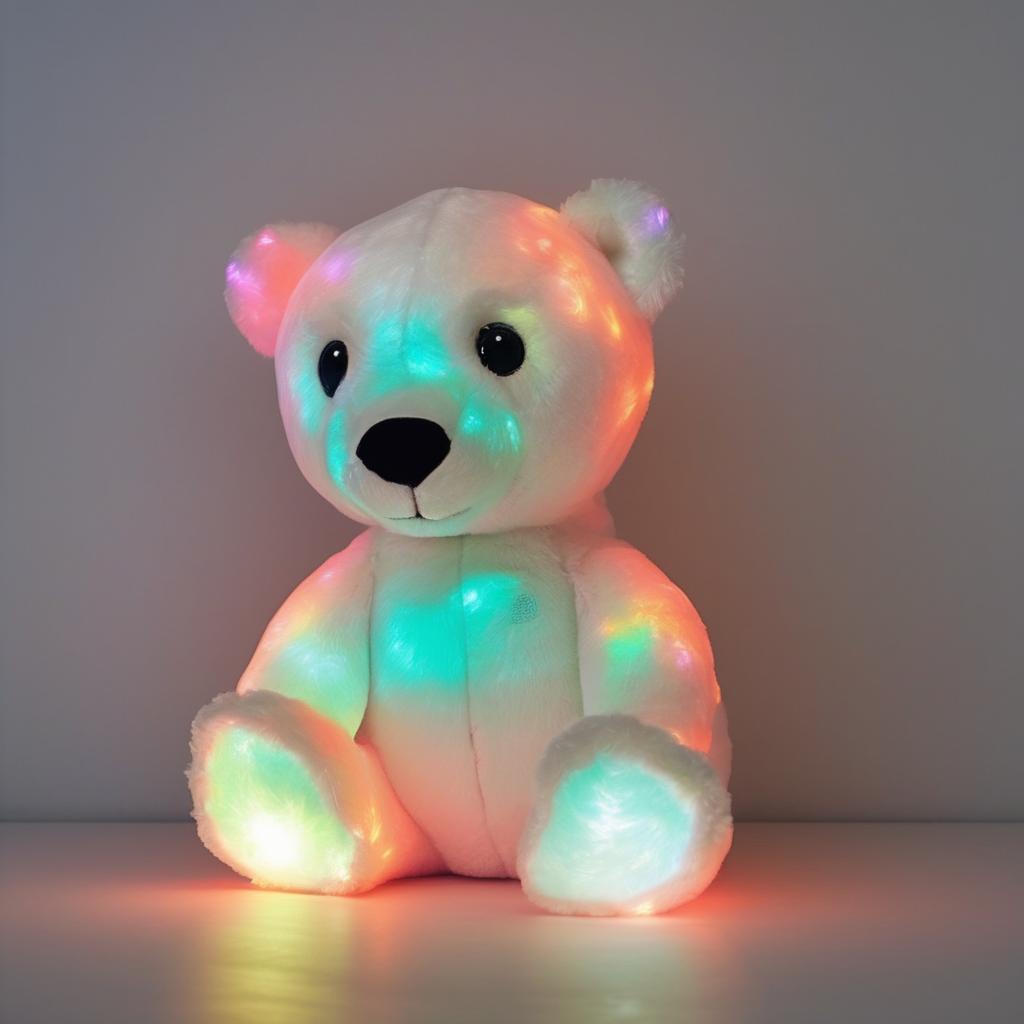} \\
    \begin{minipage}[c][1.7cm][c]{\linewidth} %
        \centering TIES
    \end{minipage}
    &
        \tabfigure{width=0.12\textwidth}{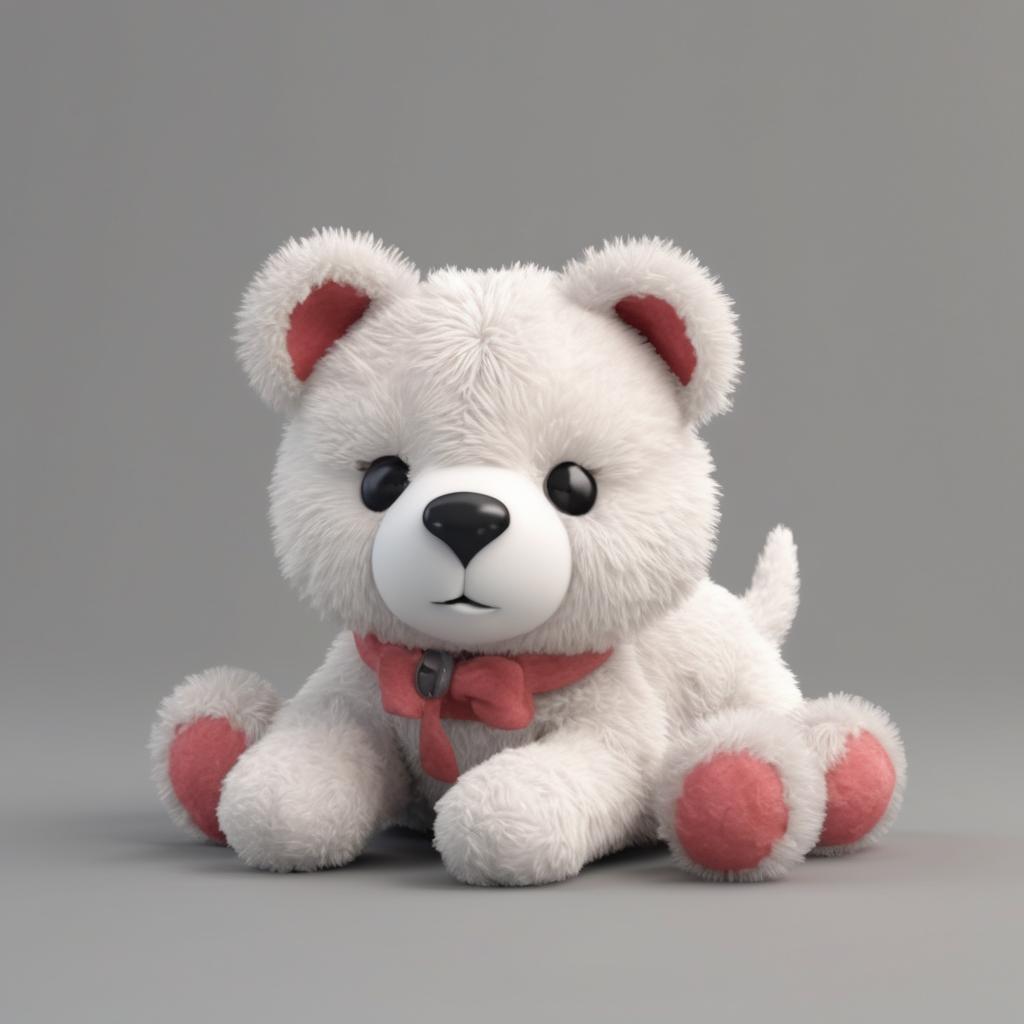} &
        \tabfigure{width=0.12\textwidth}{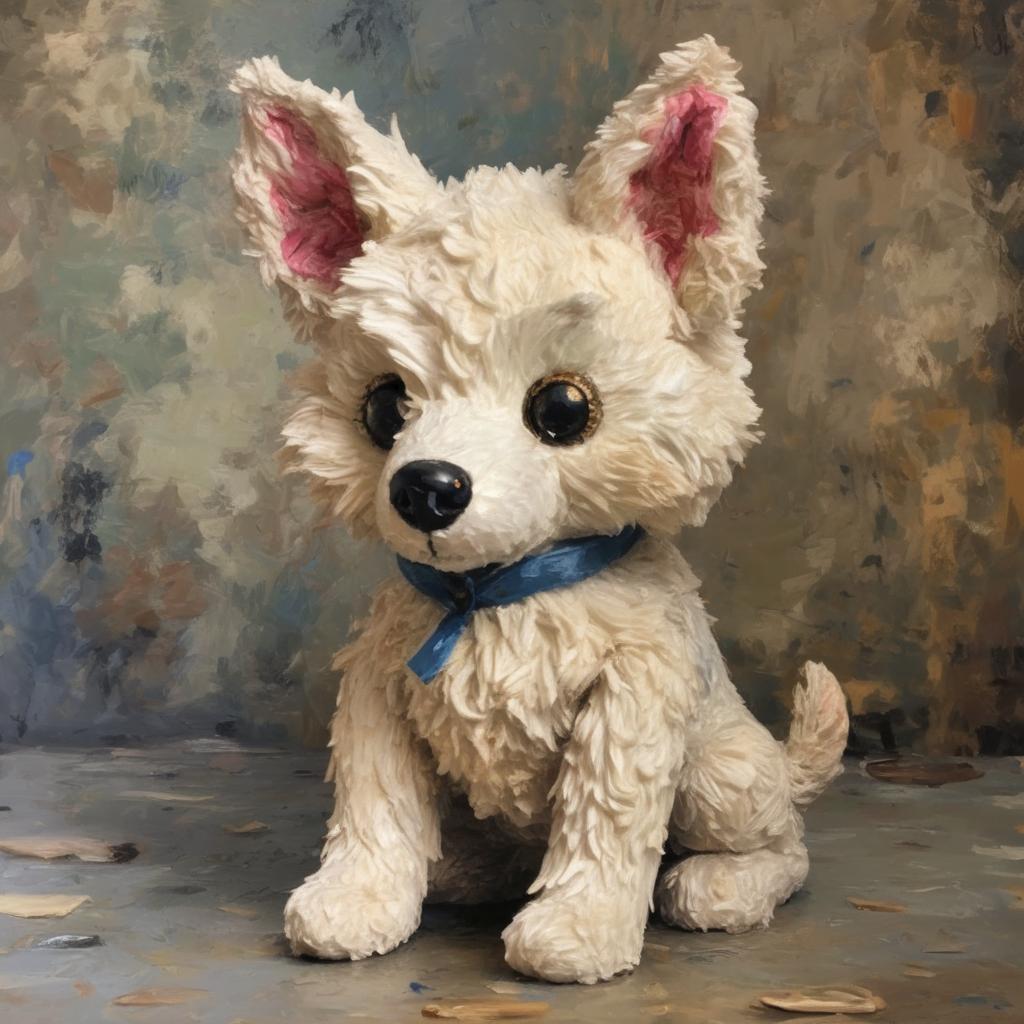} &
        \tabfigure{width=0.12\textwidth}{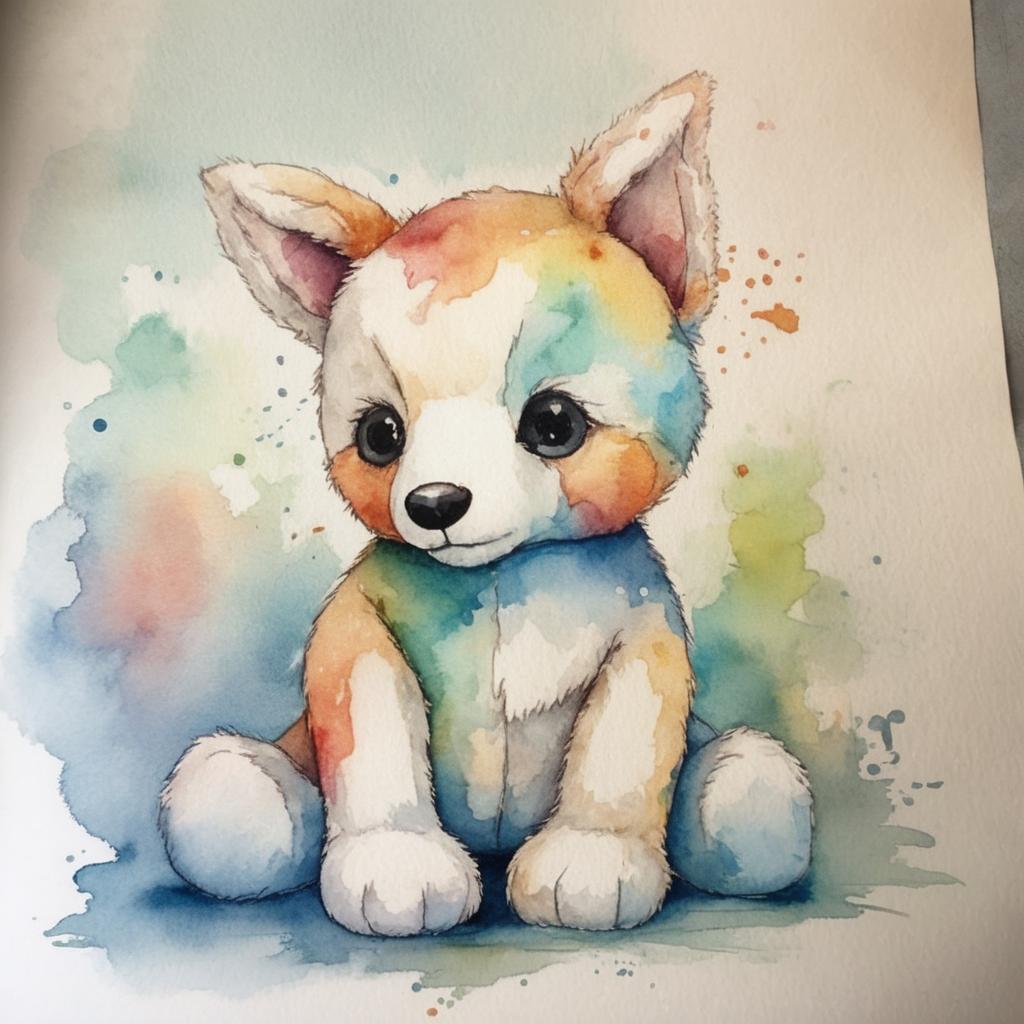} &
        \tabfigure{width=0.12\textwidth}{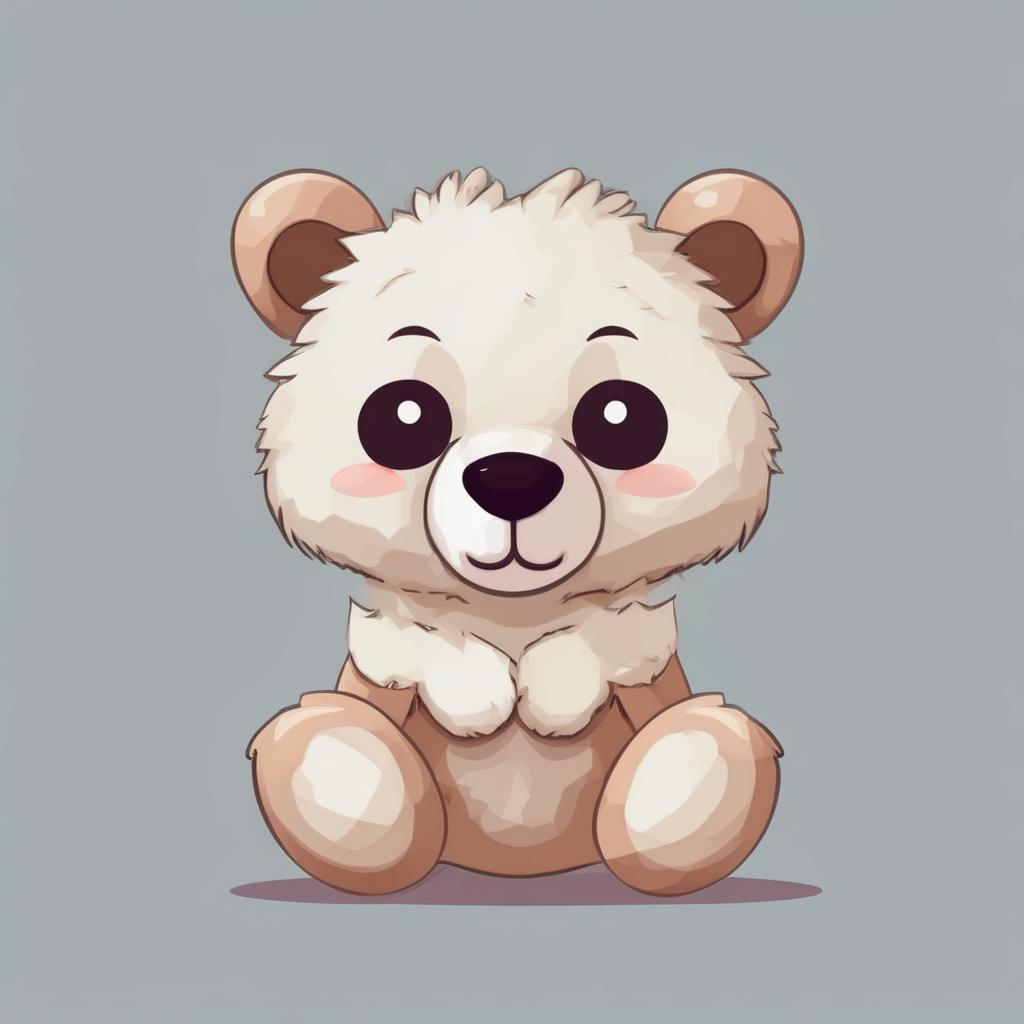} &
        \tabfigure{width=0.12\textwidth}{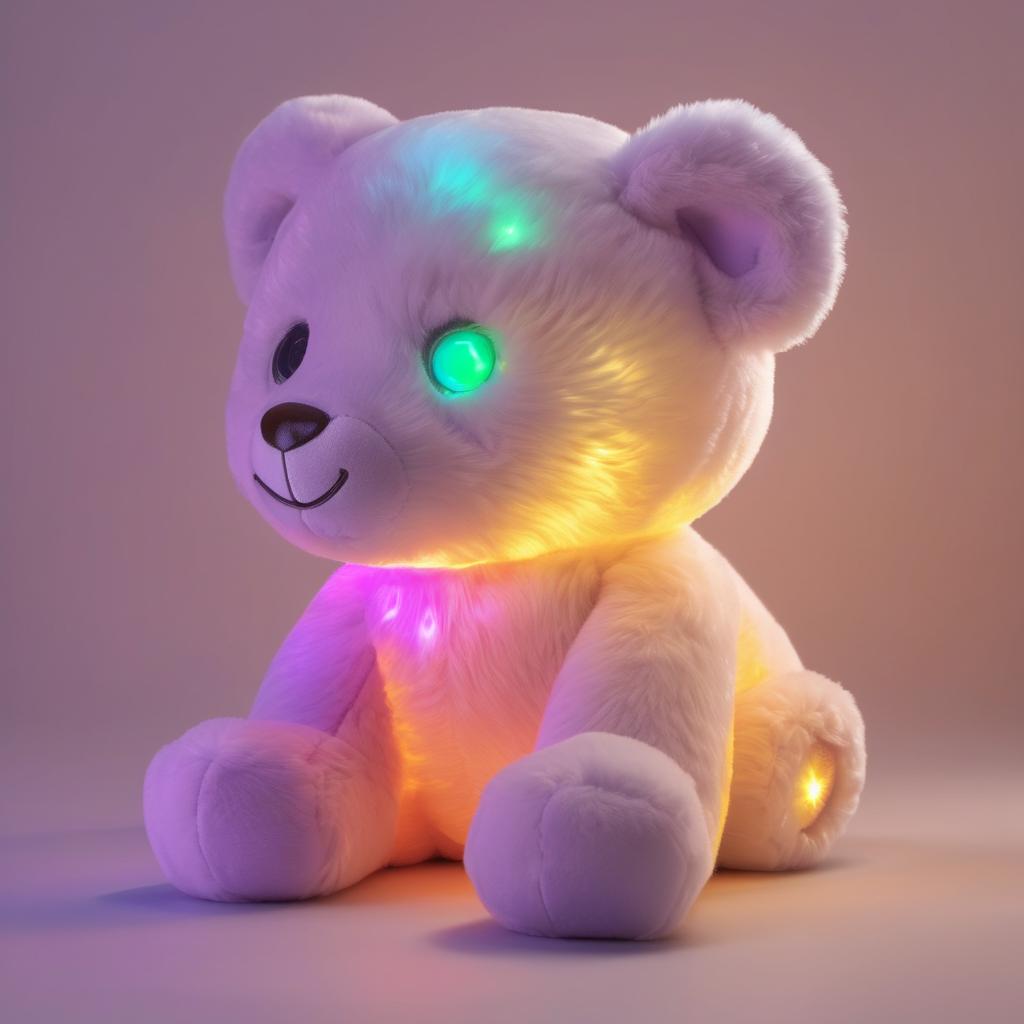} \\
    \begin{minipage}[c][1.7cm][c]{\linewidth} %
        \centering DARE-TIES
    \end{minipage}
    &
        \tabfigure{width=0.12\textwidth}{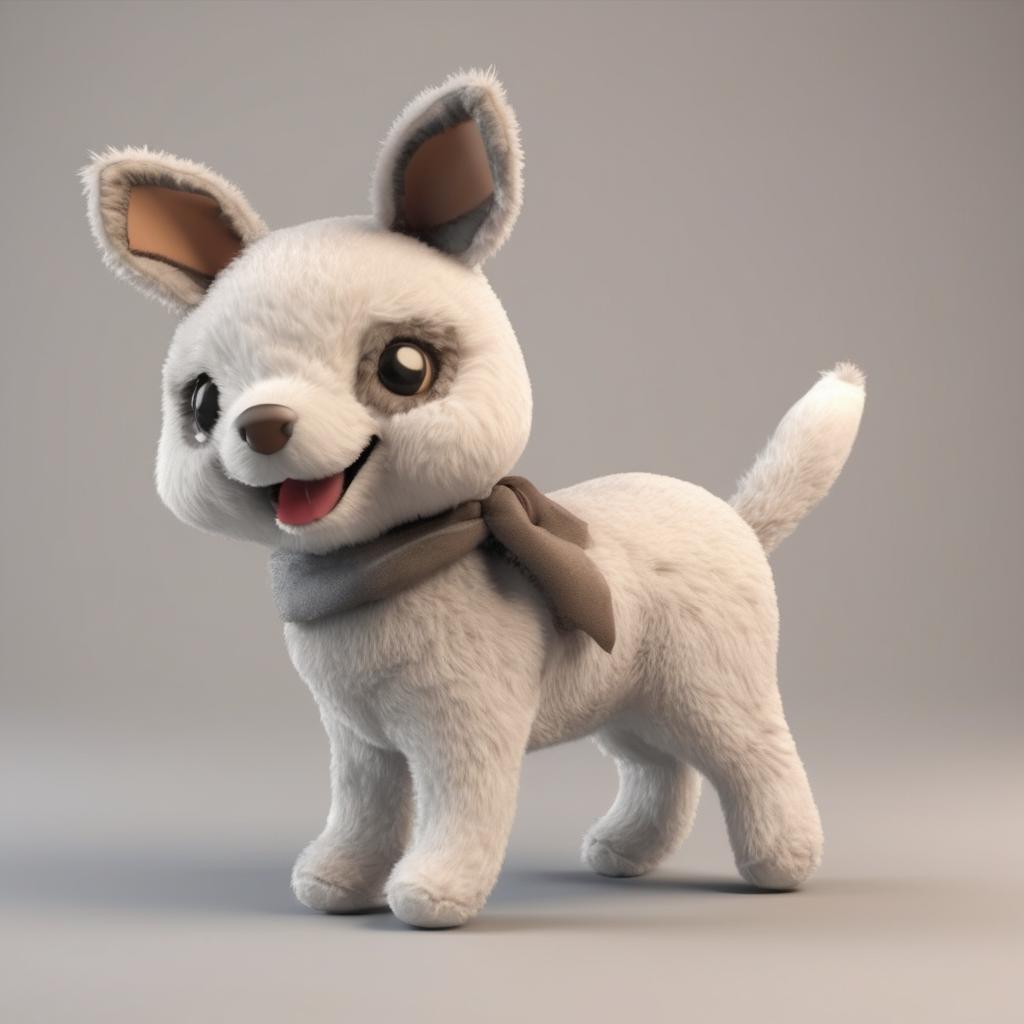} &
        \tabfigure{width=0.12\textwidth}{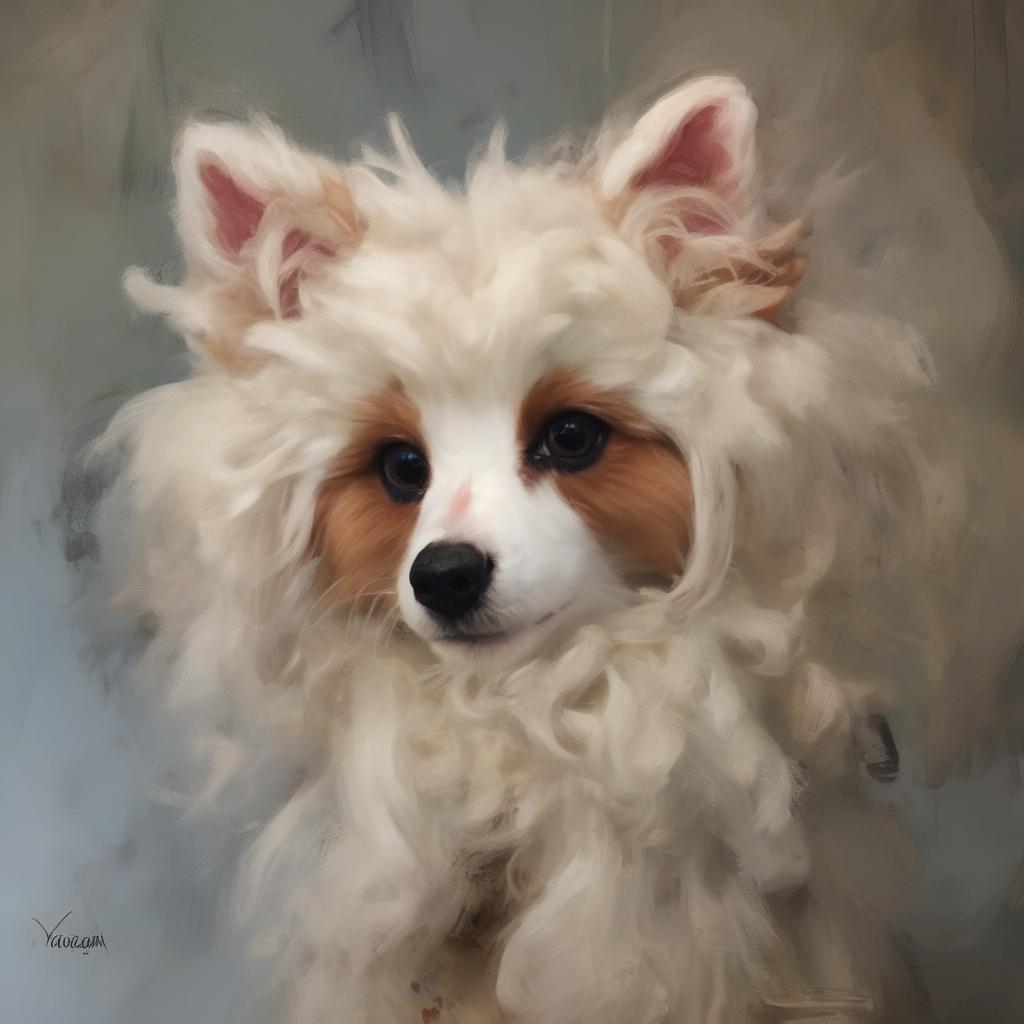} &
        \tabfigure{width=0.12\textwidth}{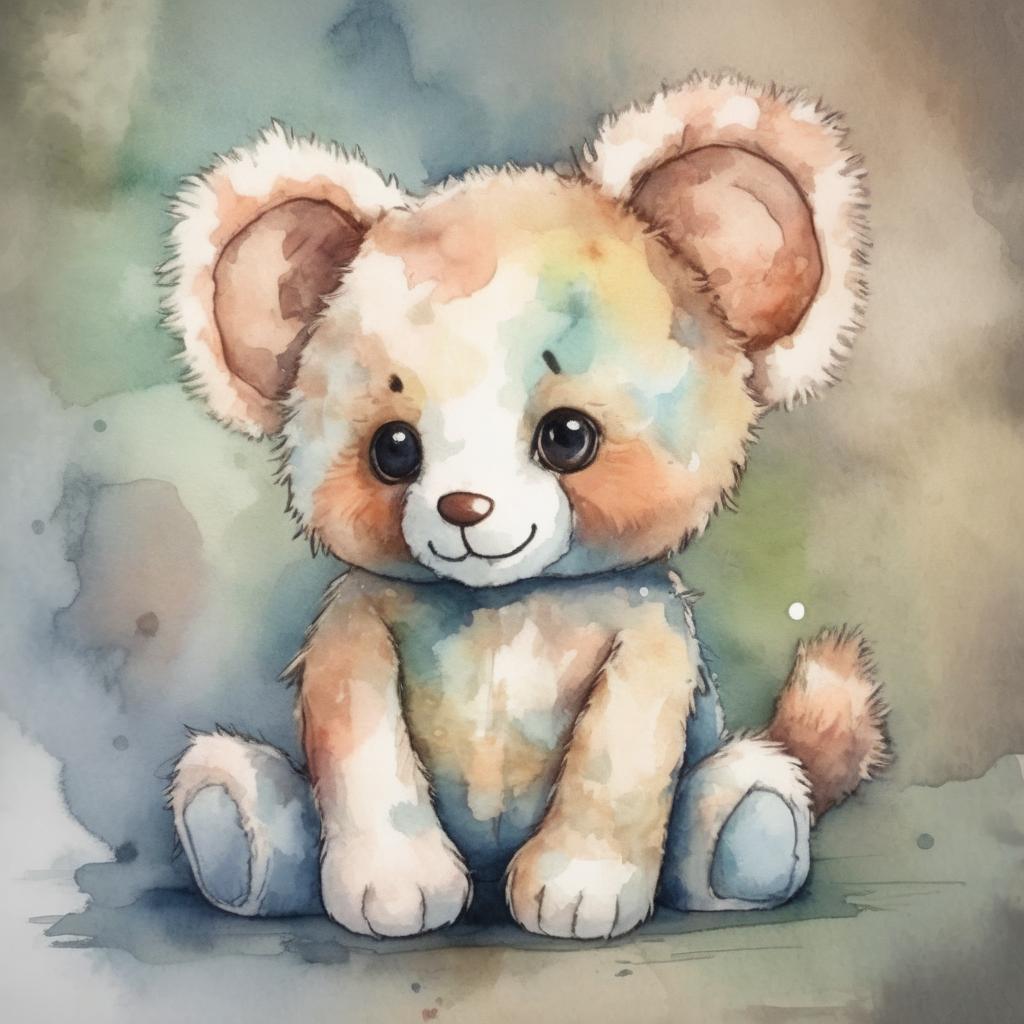} &
        \tabfigure{width=0.12\textwidth}{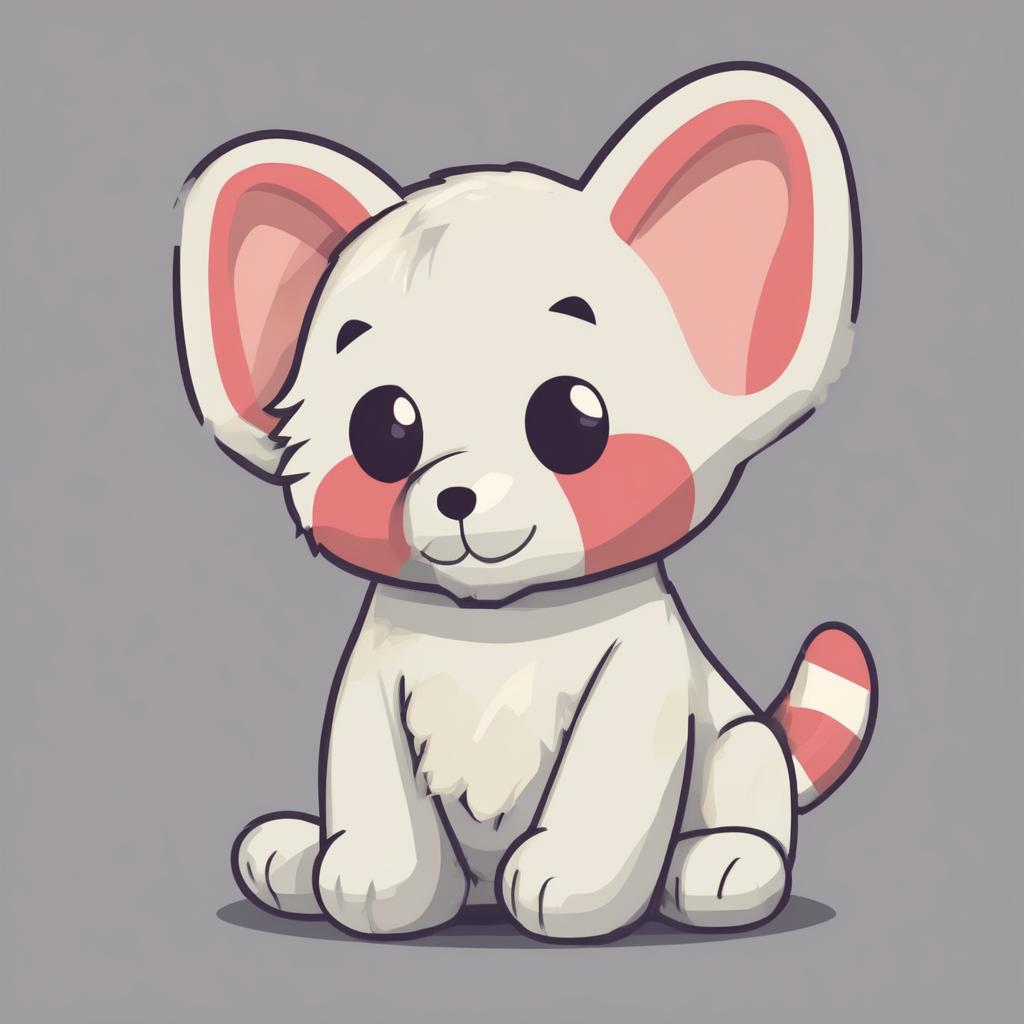} &
        \tabfigure{width=0.12\textwidth}{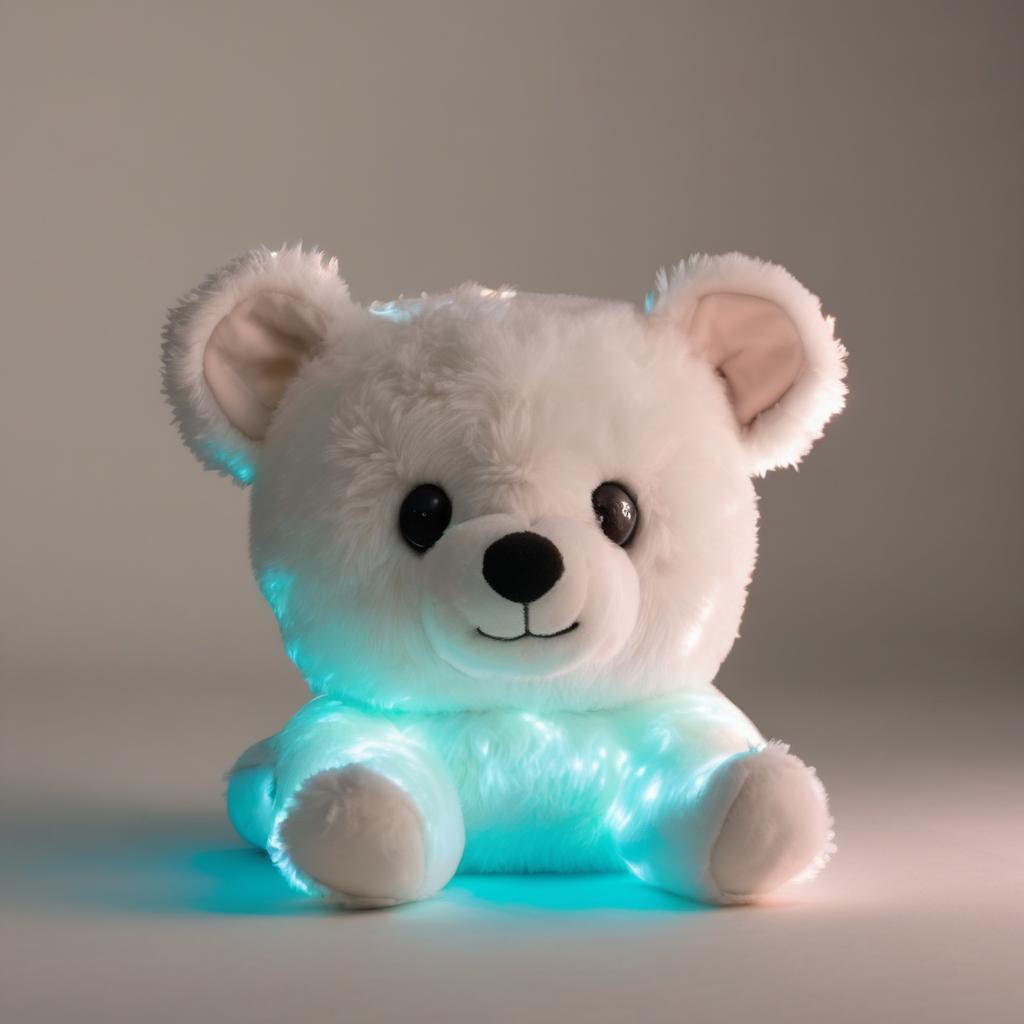} \\
    \begin{minipage}[c][1.7cm][c]{\linewidth} %
        \centering \ziplora
    \end{minipage}
    &
        \tabfigure{width=0.12\textwidth}{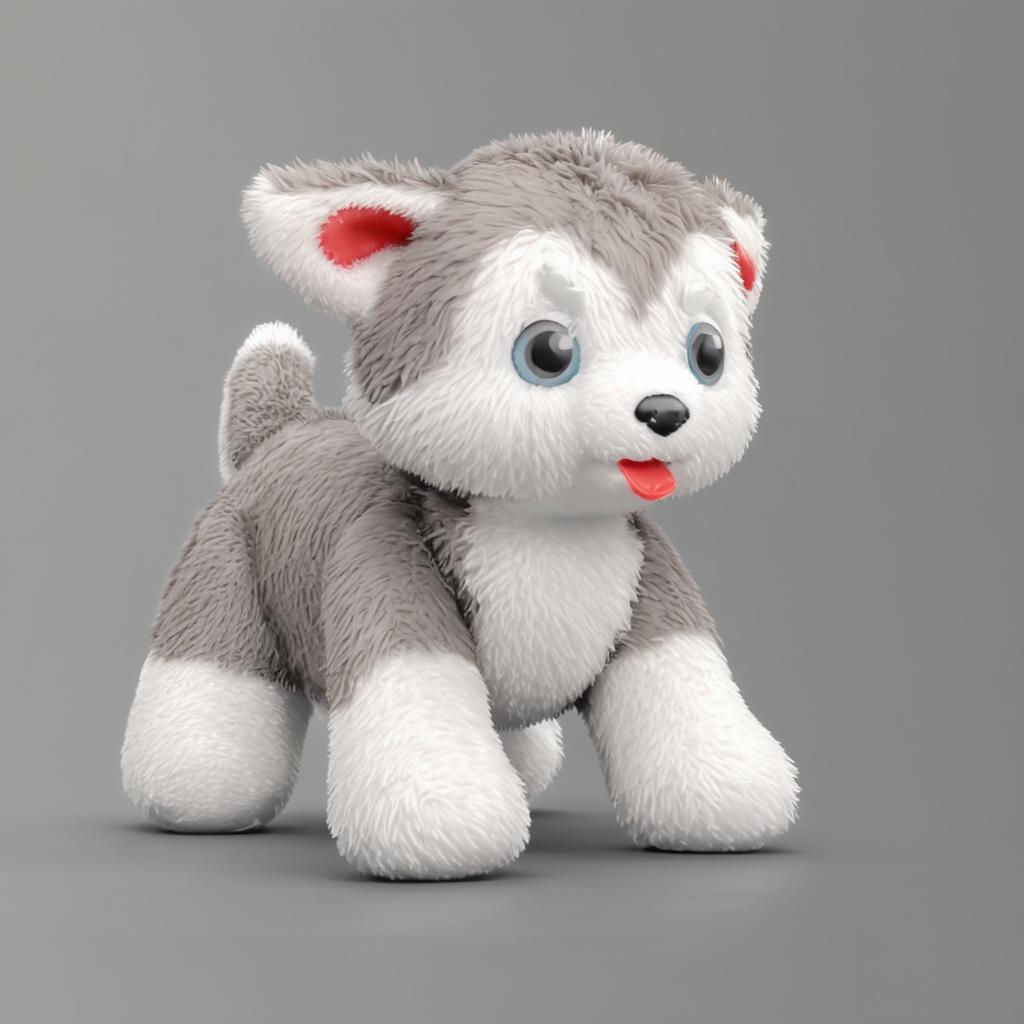} &
        \tabfigure{width=0.12\textwidth}{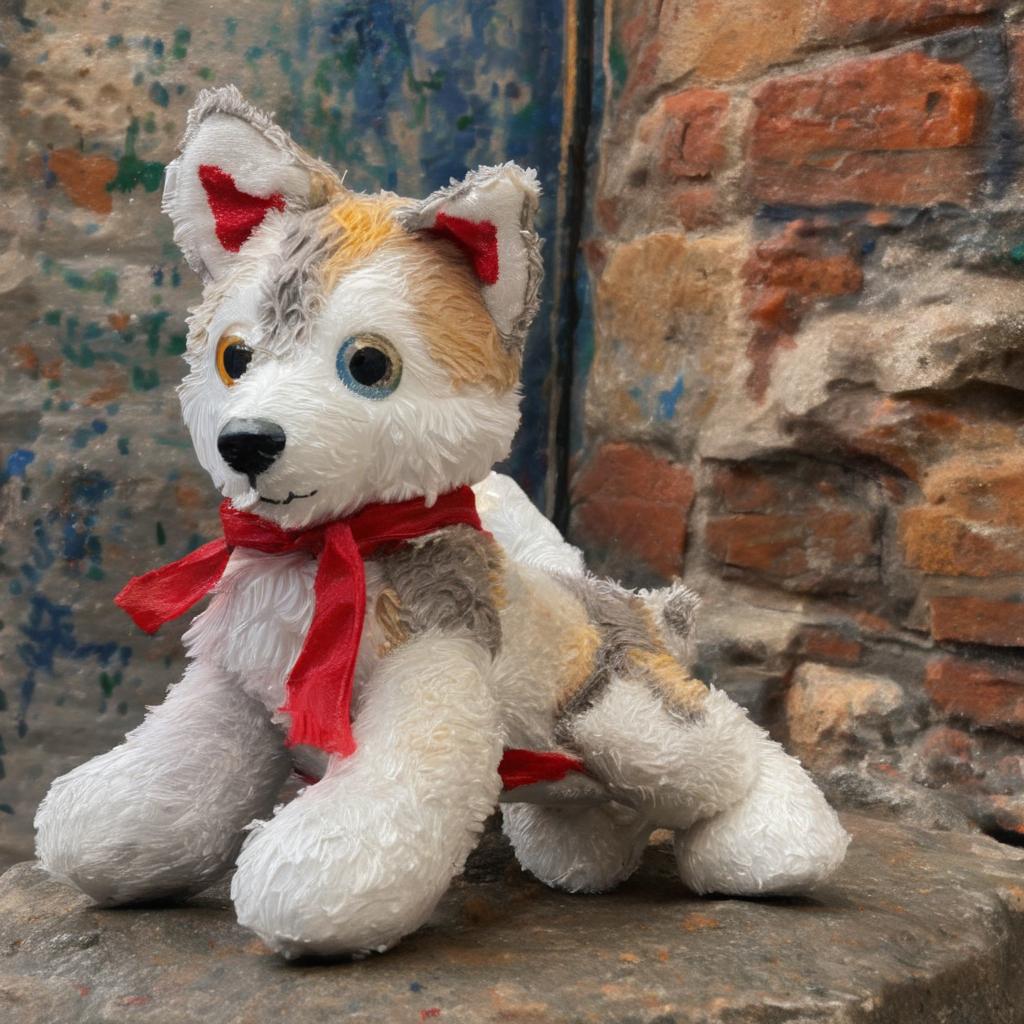} &
        \tabfigure{width=0.12\textwidth}{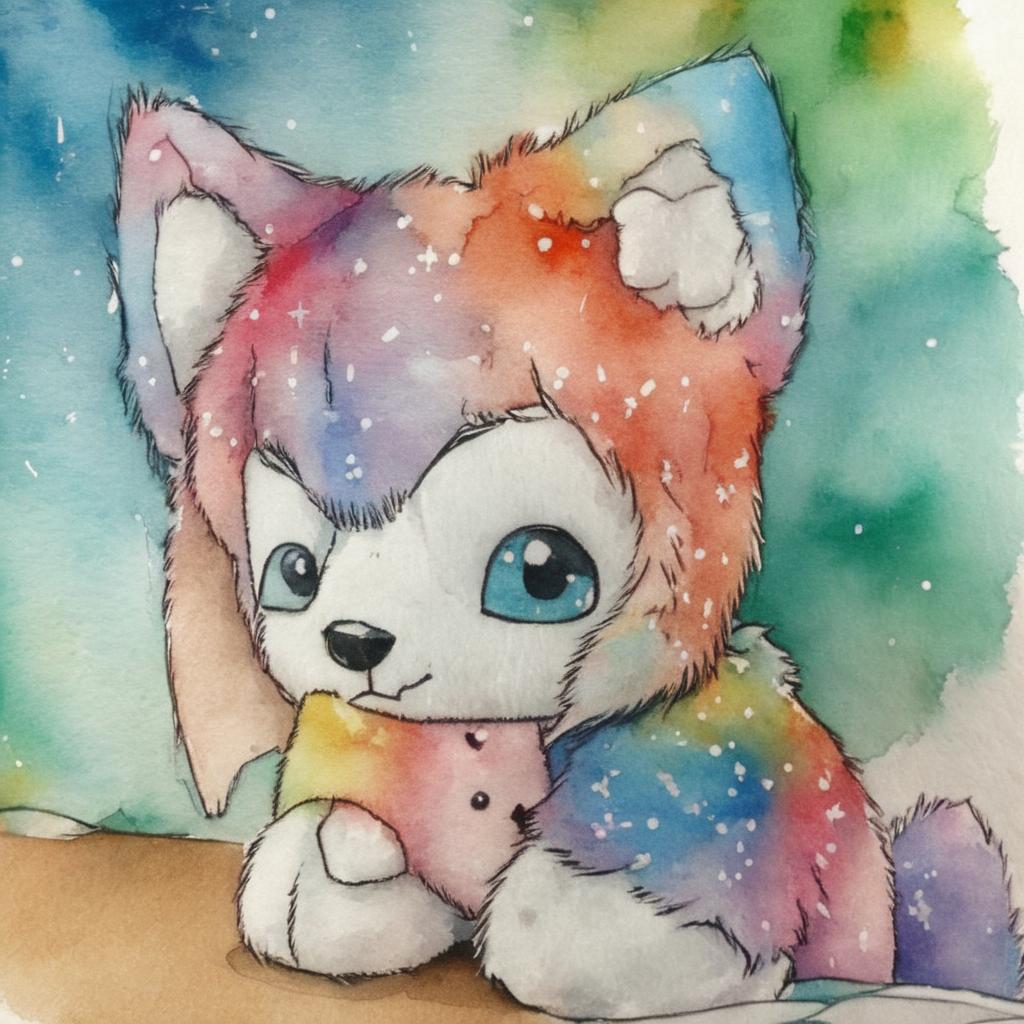} &
        \tabfigure{width=0.12\textwidth}{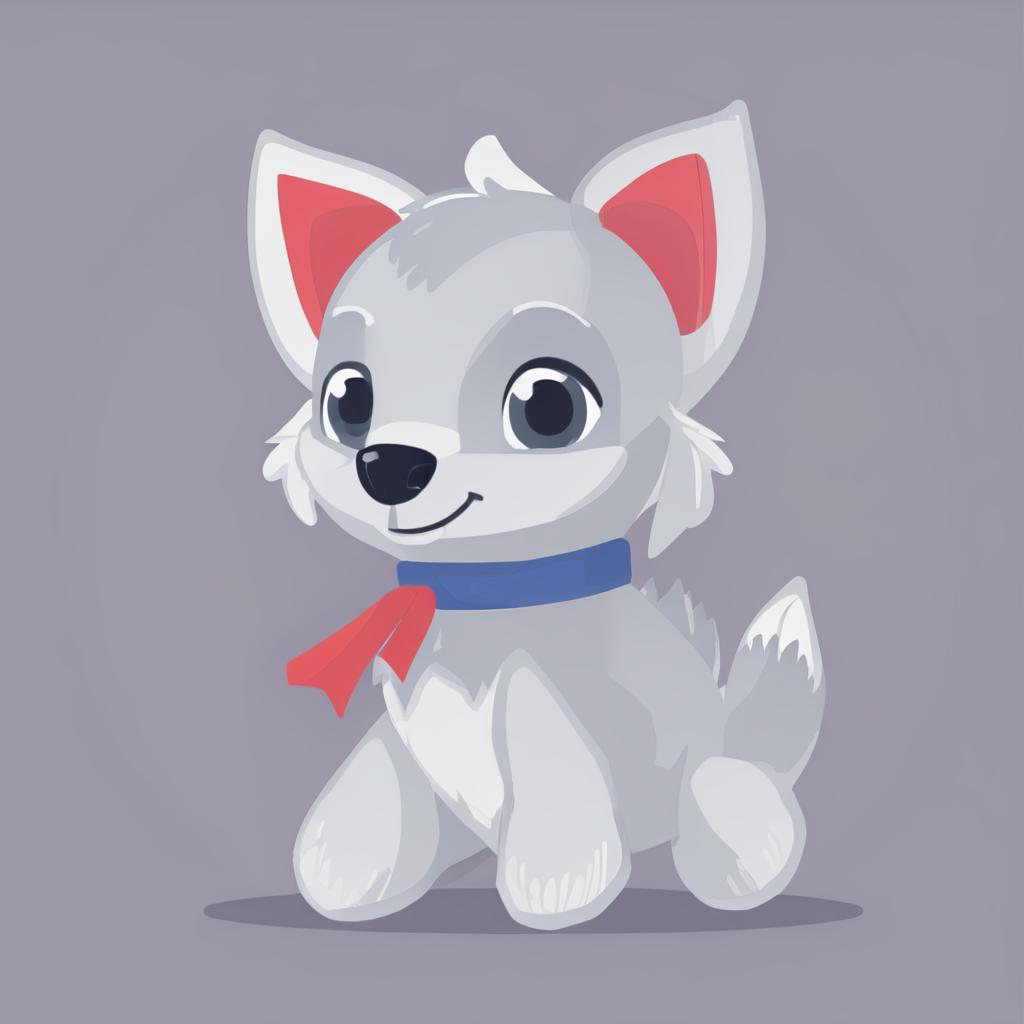} &
        \tabfigure{width=0.12\textwidth}{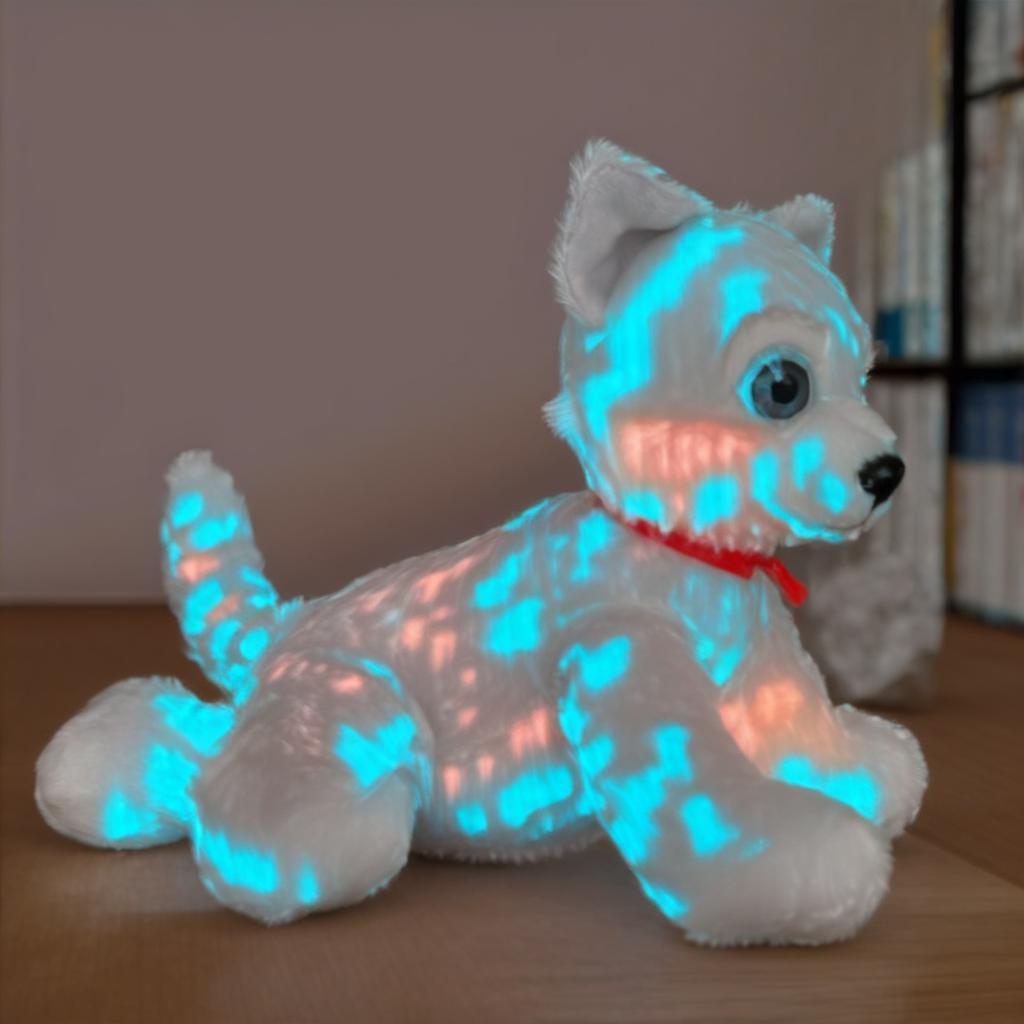} \\
    \begin{minipage}[c][1.7cm][c]{\linewidth} %
        \centering \textbf{\ours (ours)}
    \end{minipage}
        &
        \tabfigure{width=0.12\textwidth}{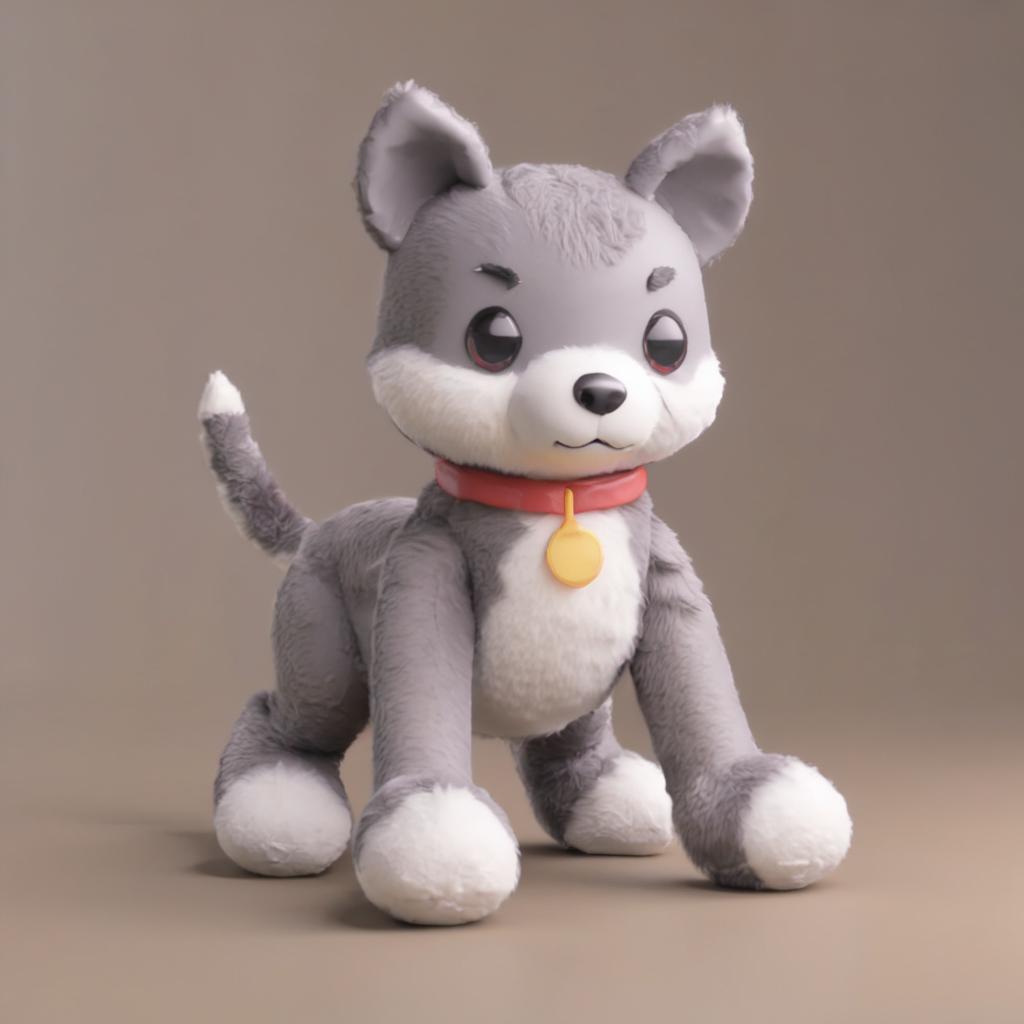} &
        \tabfigure{width=0.12\textwidth}{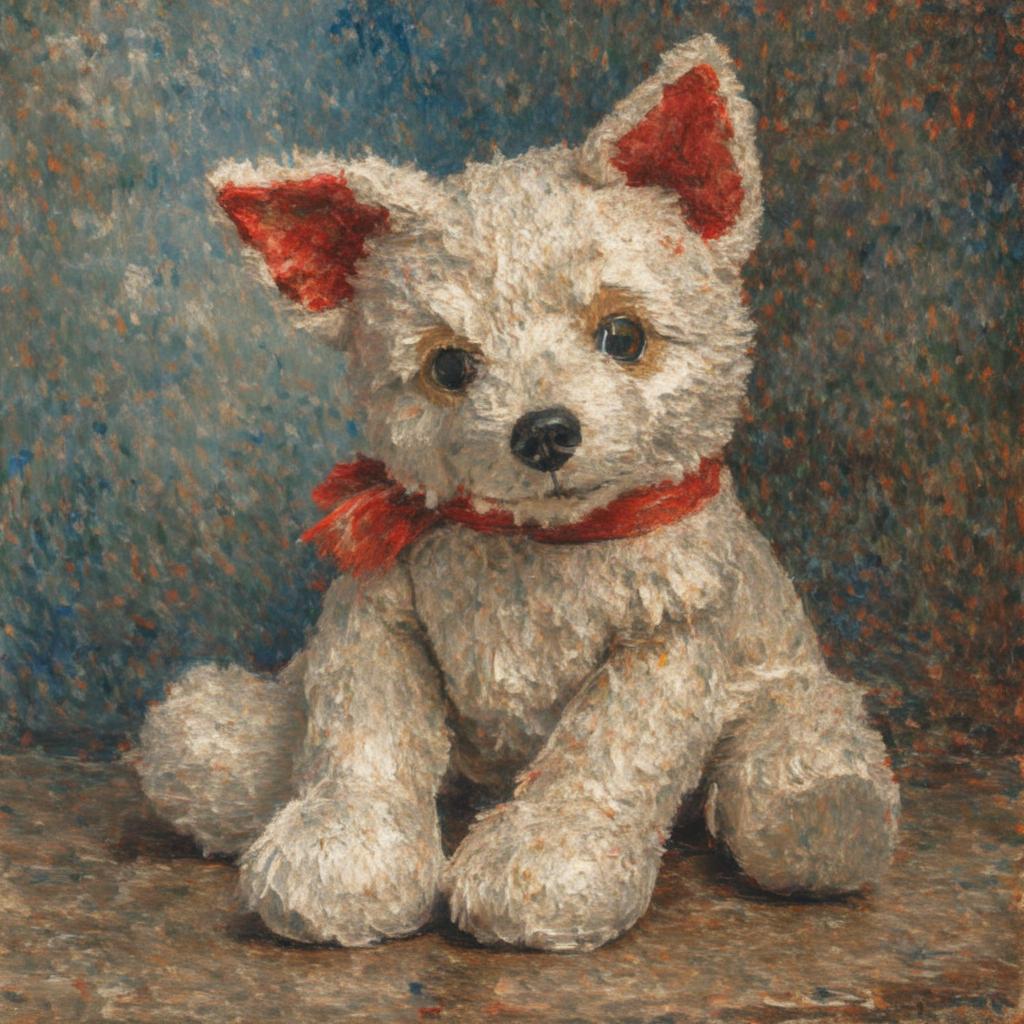} &
        \tabfigure{width=0.12\textwidth}{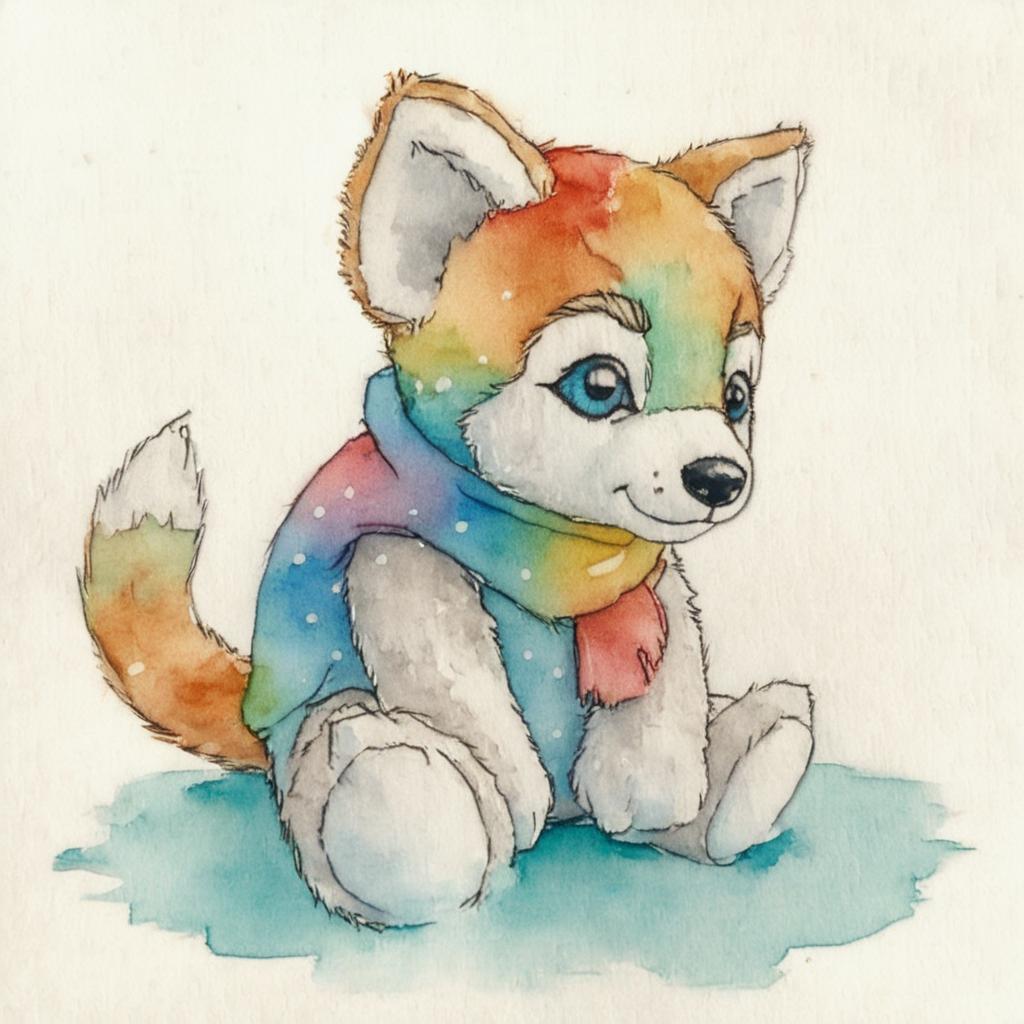} &
        \tabfigure{width=0.12\textwidth}{figures/qualitative/wolf/ours/flat_cartoon.jpg} &
        \tabfigure{width=0.12\textwidth}{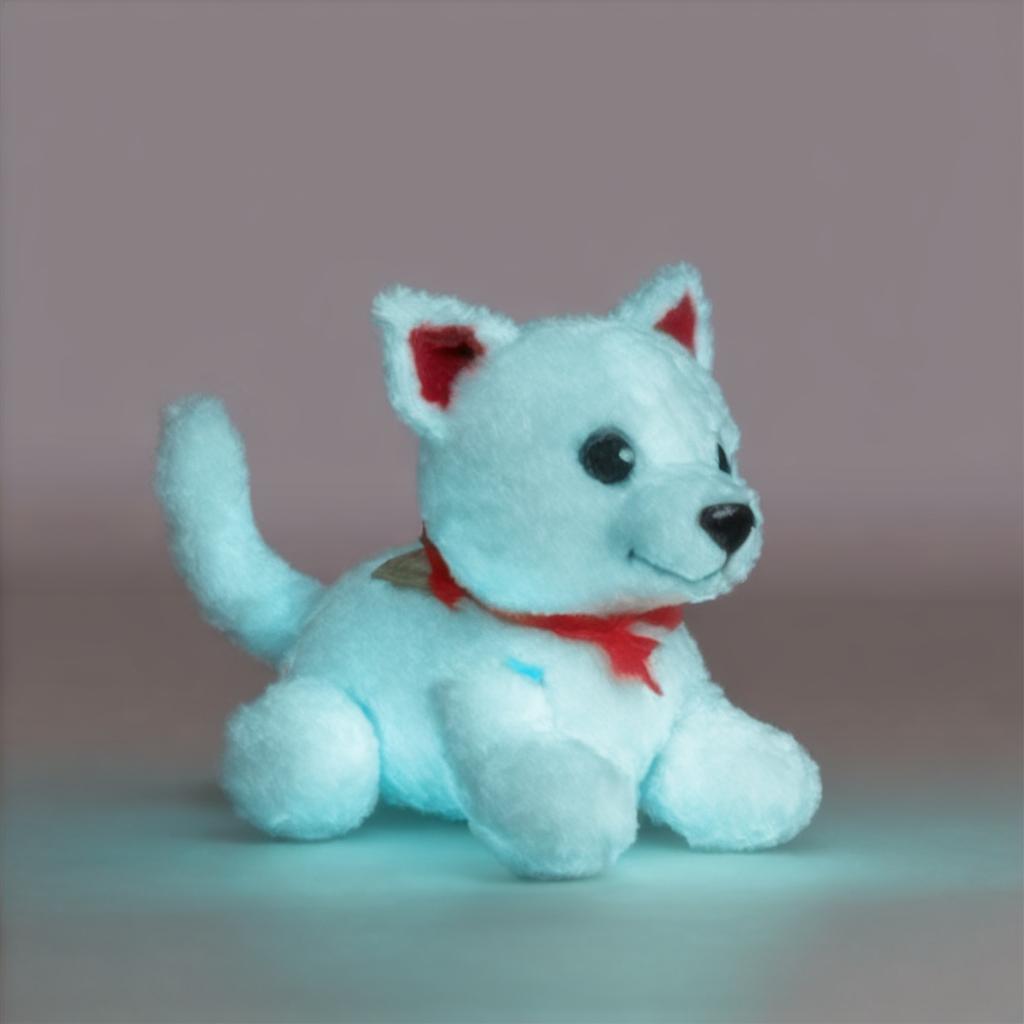} \\
    \end{tabular}
    }
    \caption{\textbf{Qualitative Comparison.} \ours generates better images than other merging strategies, including ZipLoRA.}
    \vspace{-0.4cm}
    \label{fig:qualitative_comparison}
\end{figure*}

\noindent\textbf{1) Comparison against state of the art} is shown in Fig.~\ref{fig:qualitative_comparison}.
The results demonstrate that \ours excels in capturing fine details across various styles, consistently producing high-quality images. 
While \ziplora also generates high-quality images, \ours outperforms it in terms of overall fidelity to both content and style.
A limitation of ZipLoRA is in too realistic generation, \eg, the teapot in 3D rendering style is immersed in a photorealistic scene, and the wolf plushie in oil painting does not resemble a painting.
Other approaches show less consistent results, \ie,
direct merge is able to produce a teapot or a stuffed animal in 3D rendering style (with minor inaccuracies), but fails at generating flat cartoon illustrations, where there is no one-to-one mapping of content and style. DARE, TIES, DARE-TIES do not produce satisfactory results: either the style or the content are incorrect, or both. Joint training, presents improved results compared to direct merge, but has the same limitations.

\begin{figure}[!t]
    \centering
    \resizebox{0.46\textwidth}{!}{%
    \begin{tabular}[c]{L L L L L L}
\raisebox{-7mm}{\begin{tikzpicture}
    \draw[->] (0, -1.5) -- (1.5, -1.5) node[below, midway] {Styles};
    \draw[->] (0, -1.5) -- (0, -3) node[right, midway] {Contents};
\end{tikzpicture}}
     &
    \tabfigure{width=0.12\textwidth}{figures/qualitative/3d_rendering.jpg} &
    \tabfigure{width=0.12\textwidth}{figures/qualitative/oil.jpg} &
    \tabfigure{width=0.12\textwidth}{figures/qualitative/watercolor.jpg} & \tabfigure{width=0.12\textwidth}{figures/qualitative/flat.jpg}  & \tabfigure{width=0.12\textwidth}{figures/qualitative/glowing.jpg} \\
    
    \tabfigure{width=0.12\textwidth}{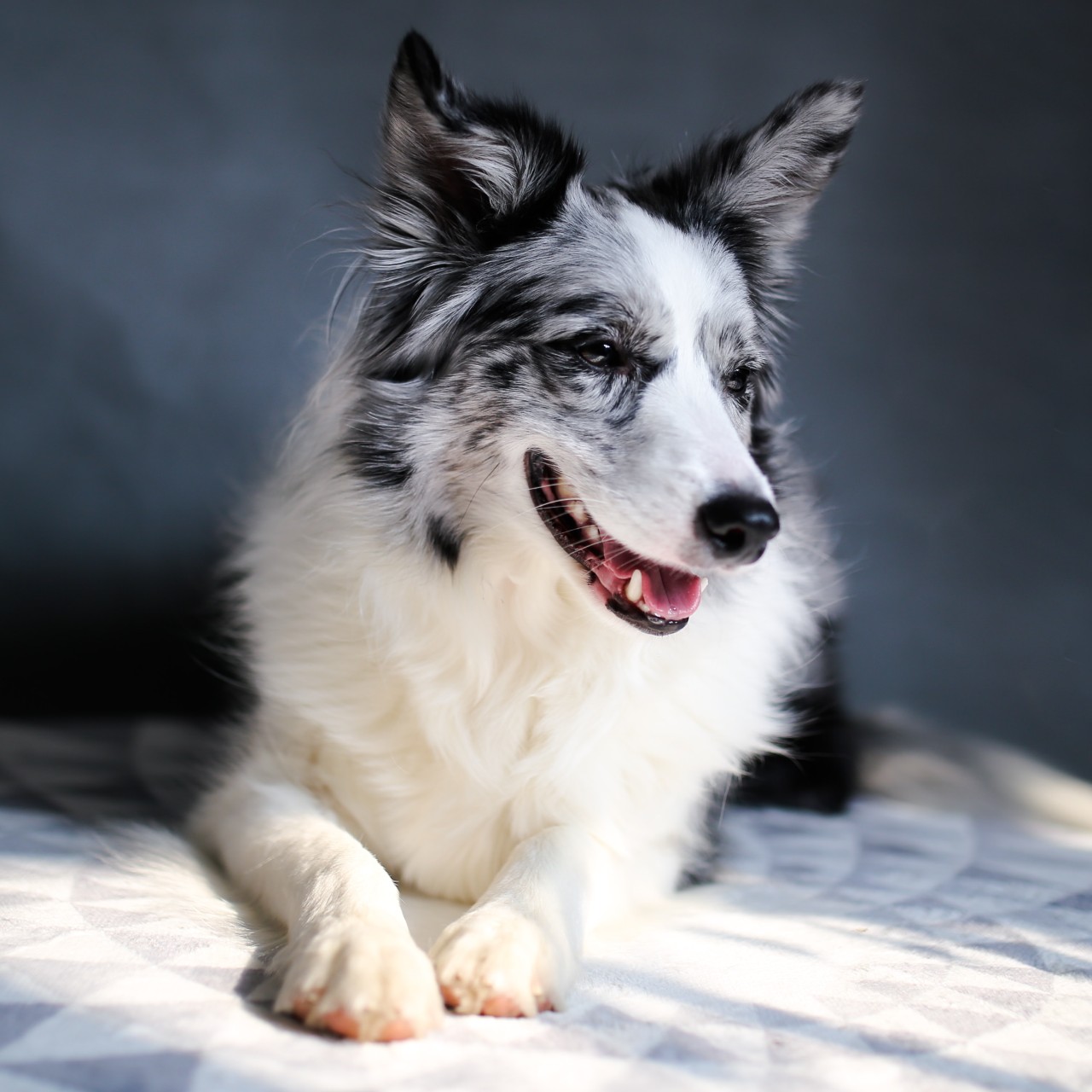} &
    \tabfigure{width=0.12\textwidth}{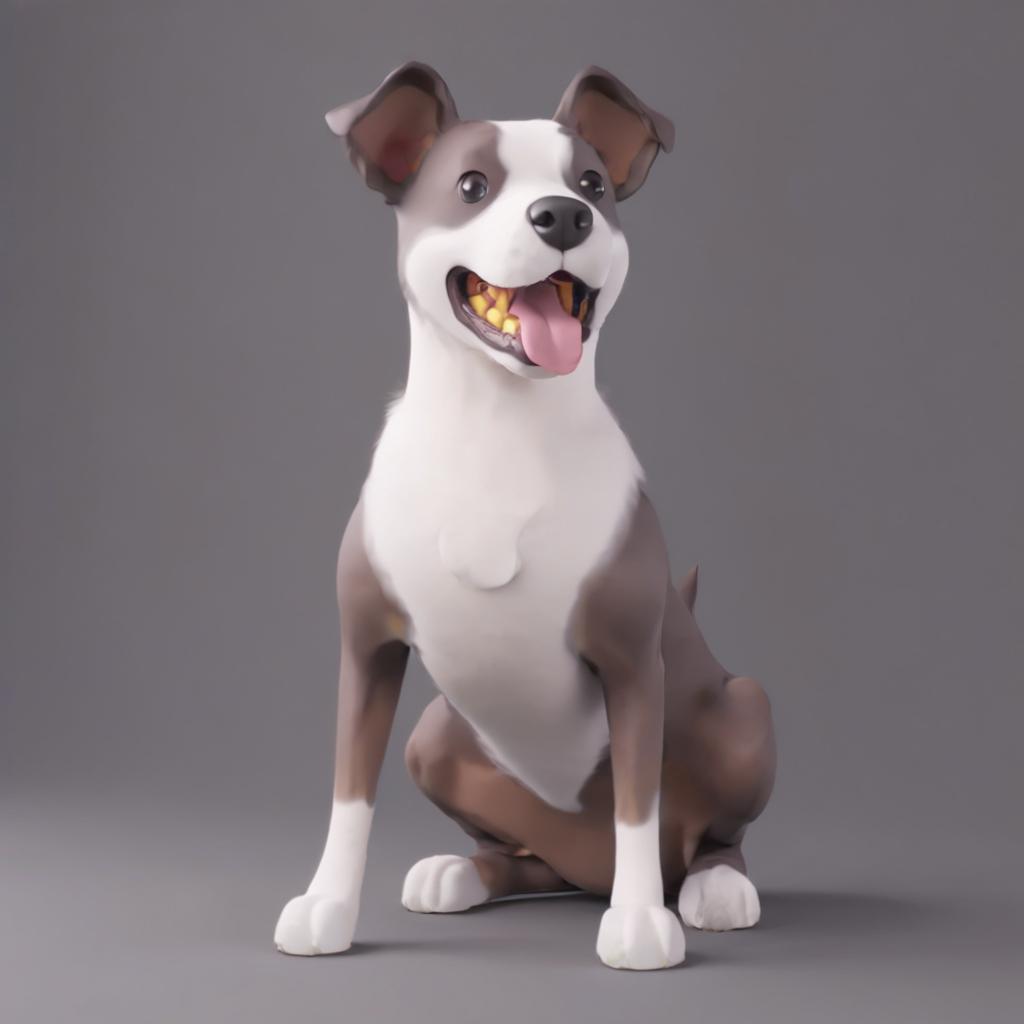} &
    \tabfigure{width=0.12\textwidth}{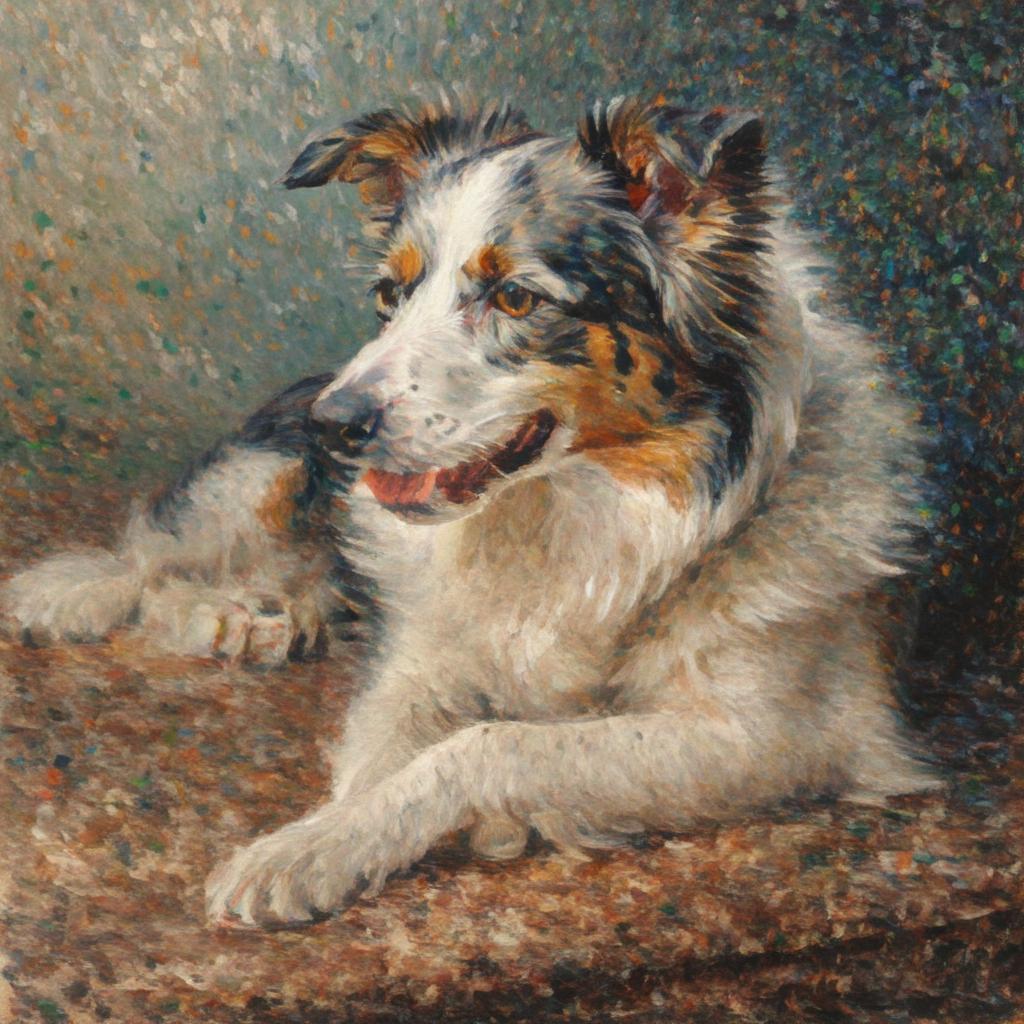} &
    \tabfigure{width=0.12\textwidth}{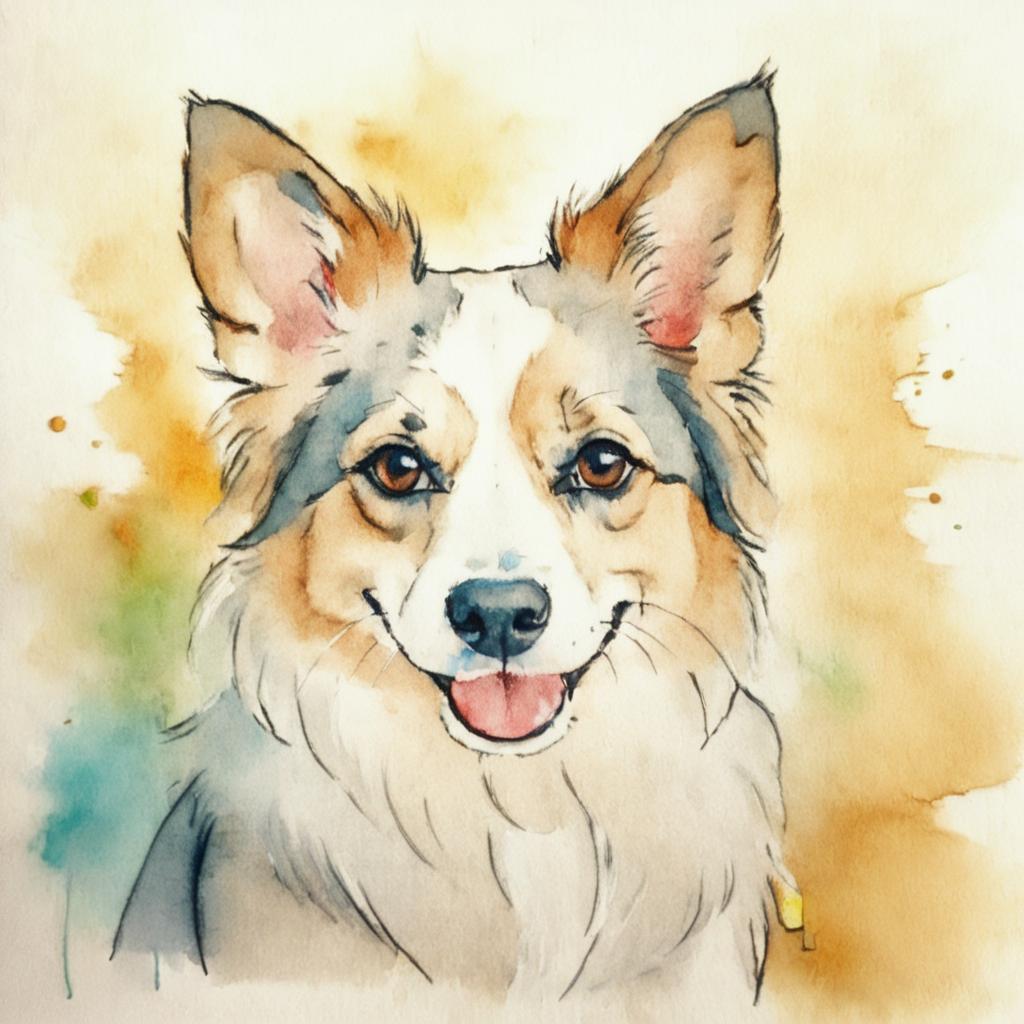} &
    \tabfigure{width=0.12\textwidth}{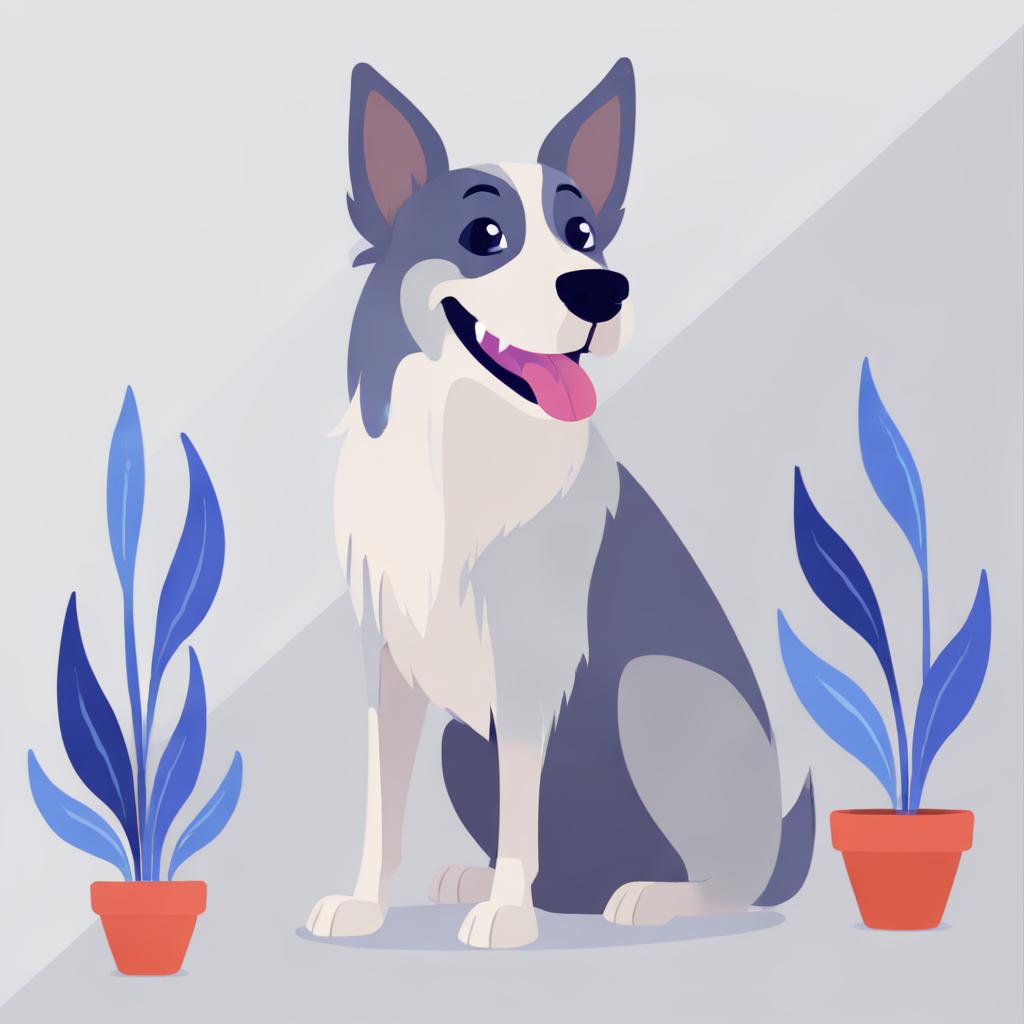} &
    \tabfigure{width=0.12\textwidth}{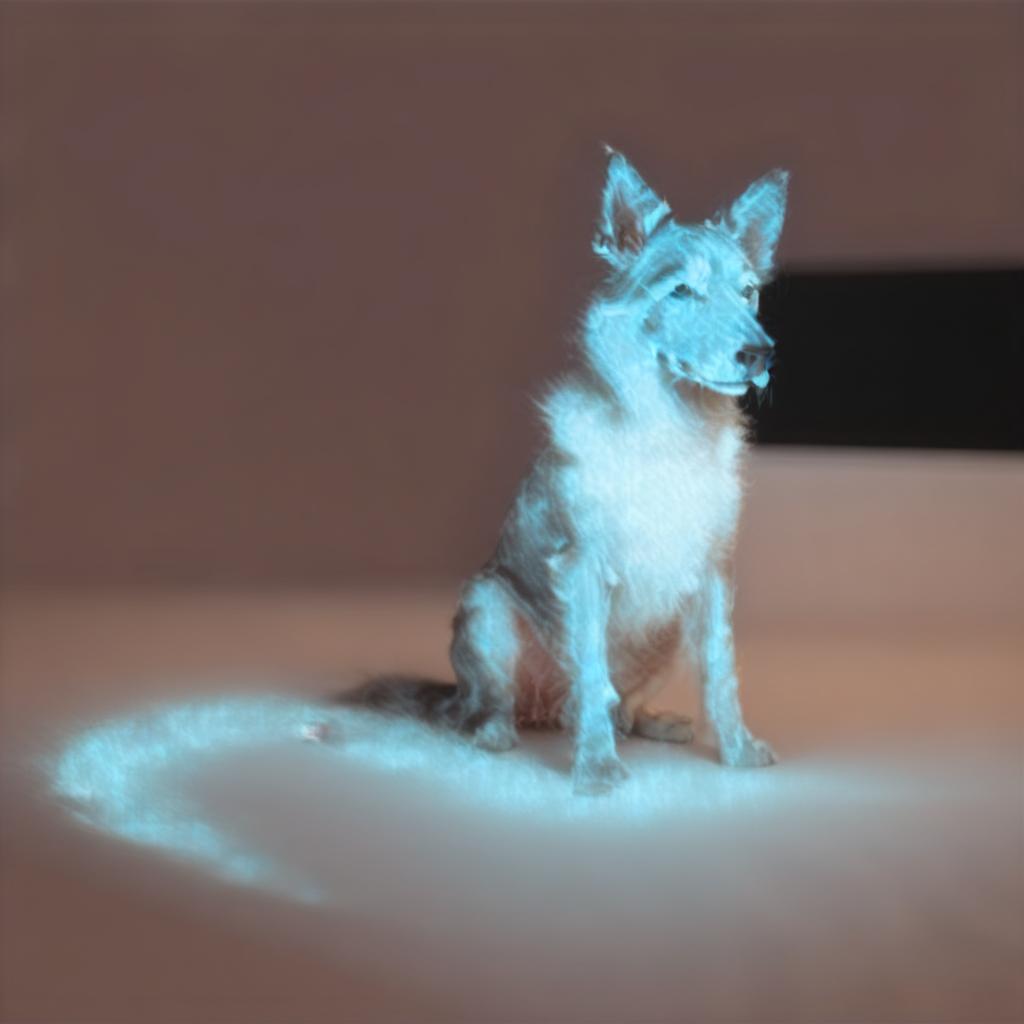} \\

    \tabfigure{width=0.12\textwidth}{figures/dataset/test/cat2/03.jpg} &
    \tabfigure{width=0.12\textwidth}{figures/qualitative/cat2/ours/3d_rendering.jpg} &
    \tabfigure{width=0.12\textwidth}{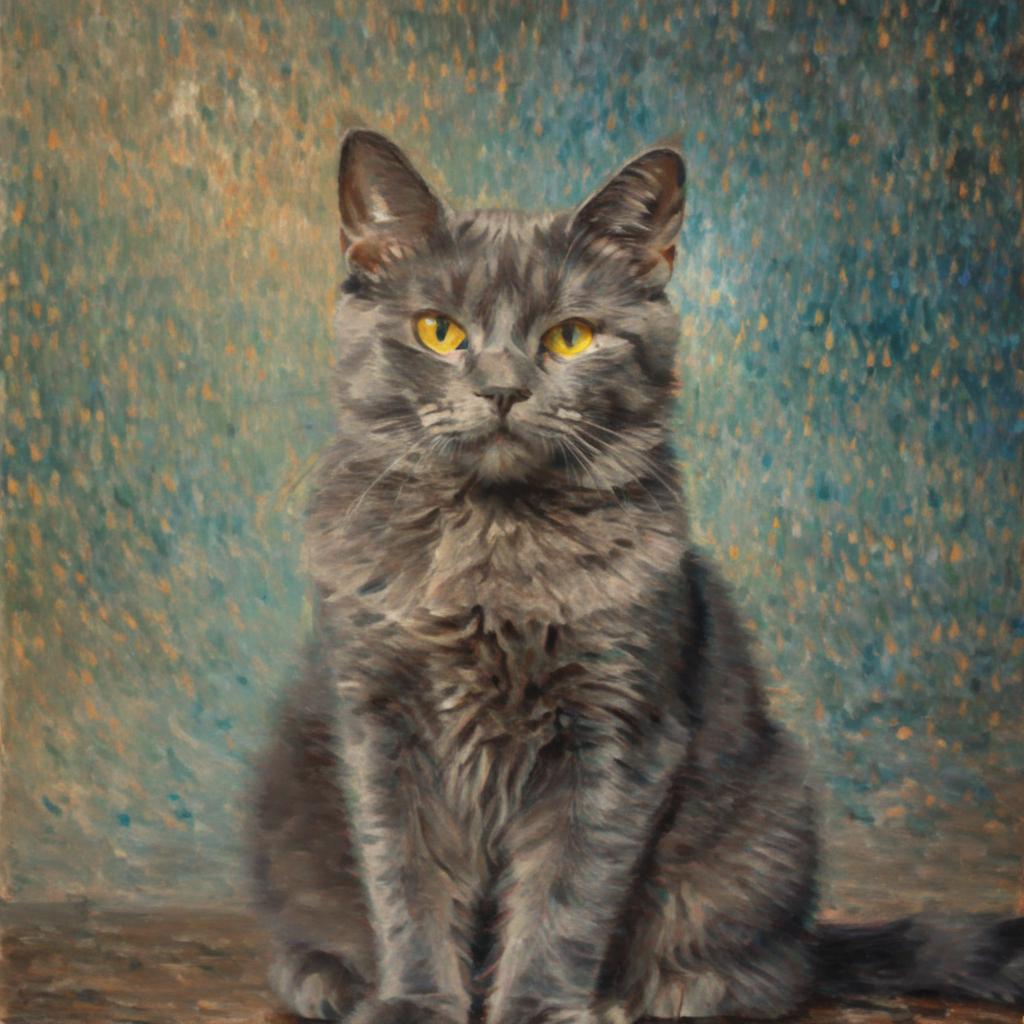} &
    \tabfigure{width=0.12\textwidth}{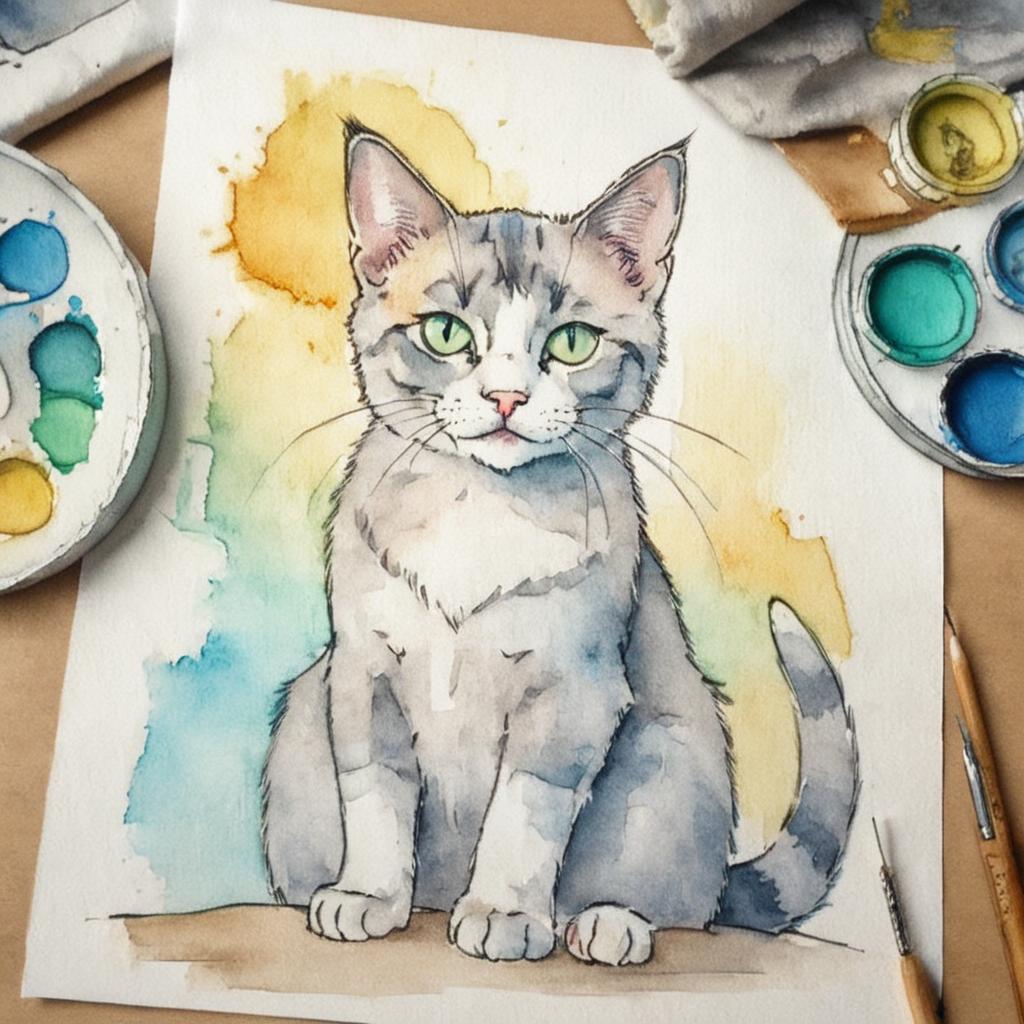} &
    \tabfigure{width=0.12\textwidth}{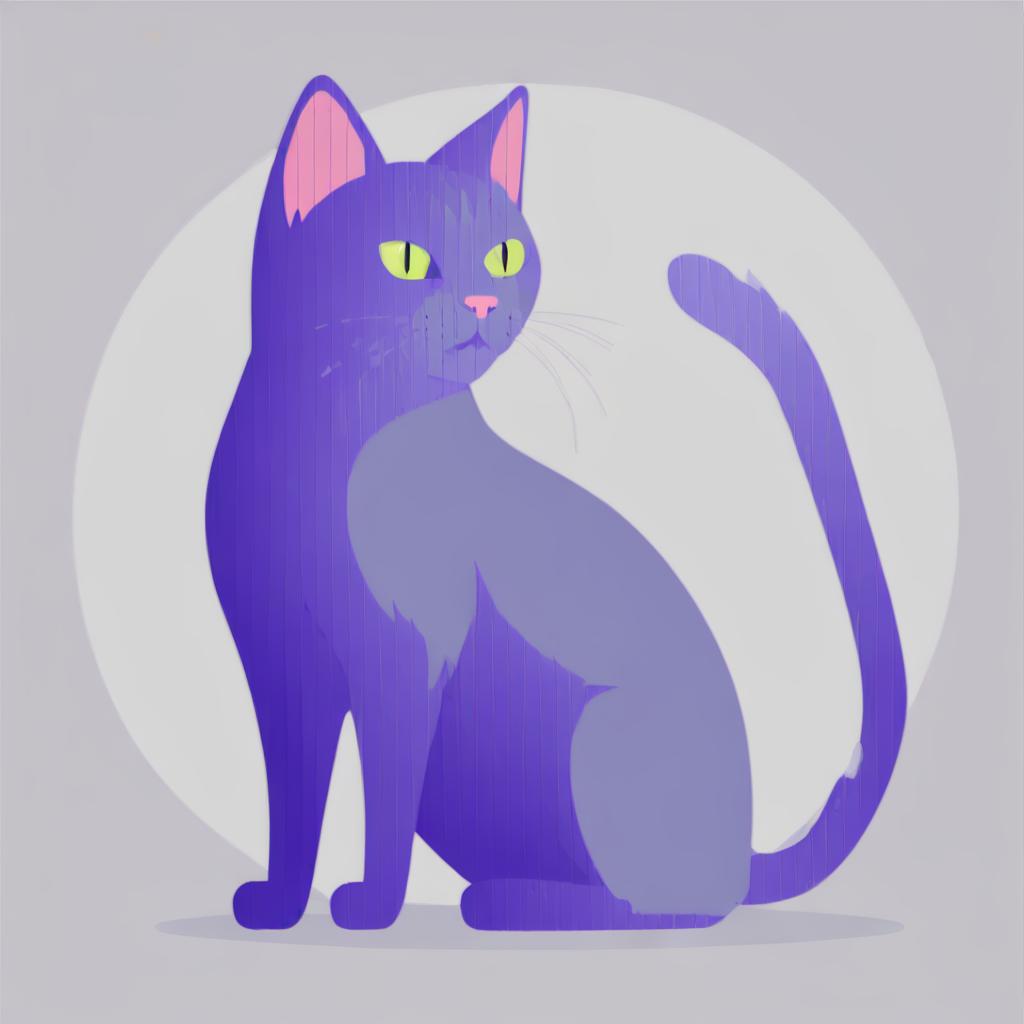} &
    \tabfigure{width=0.12\textwidth}{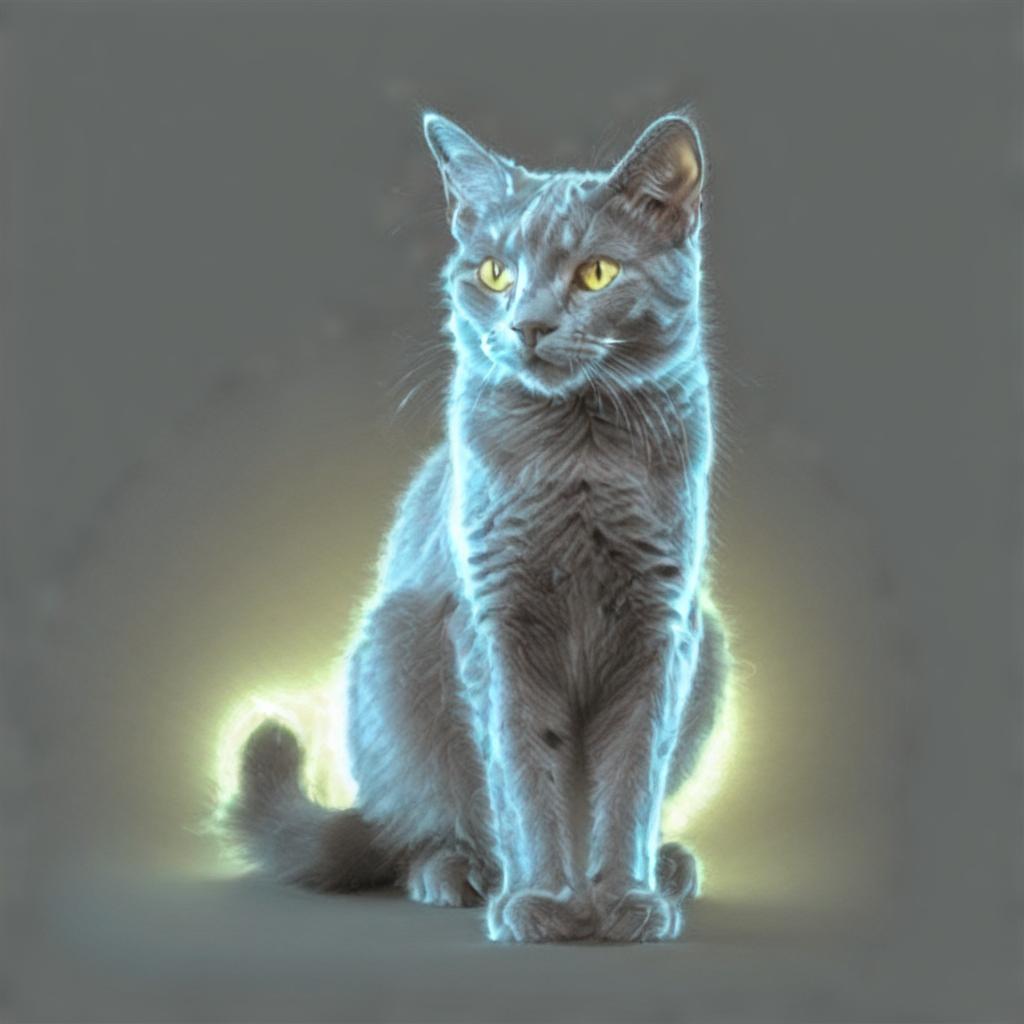} \\

    \tabfigure{width=0.12\textwidth}{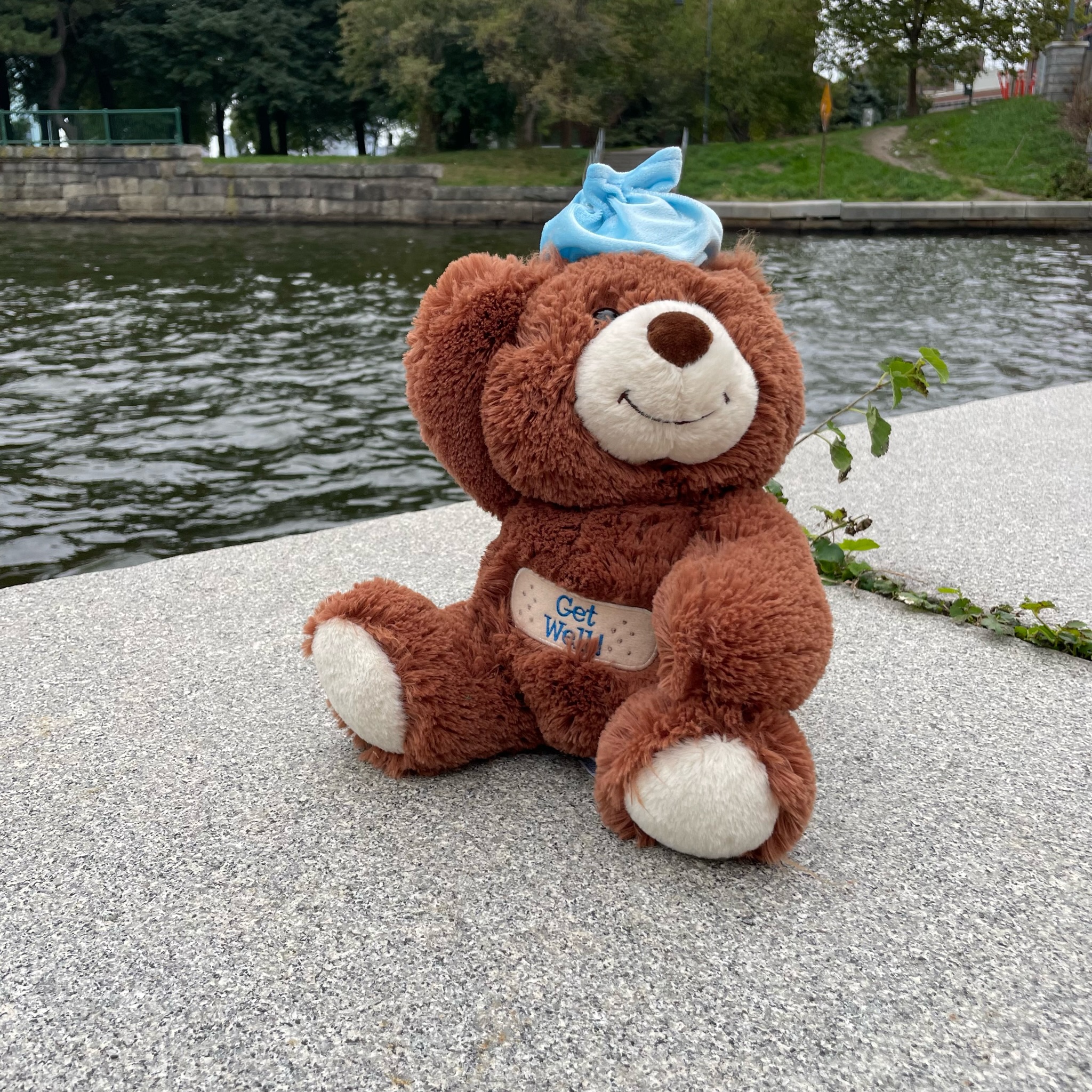} &
    \tabfigure{width=0.12\textwidth}{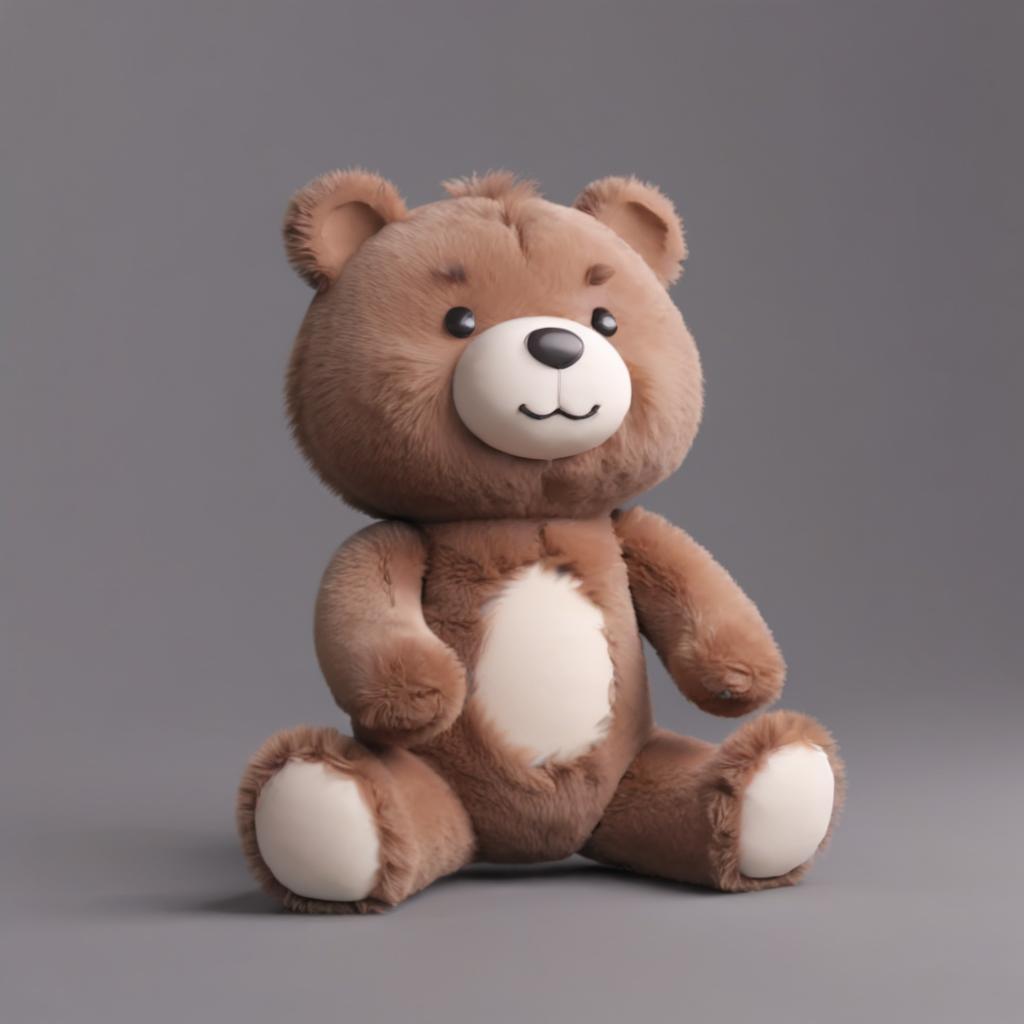} &
    \tabfigure{width=0.12\textwidth}{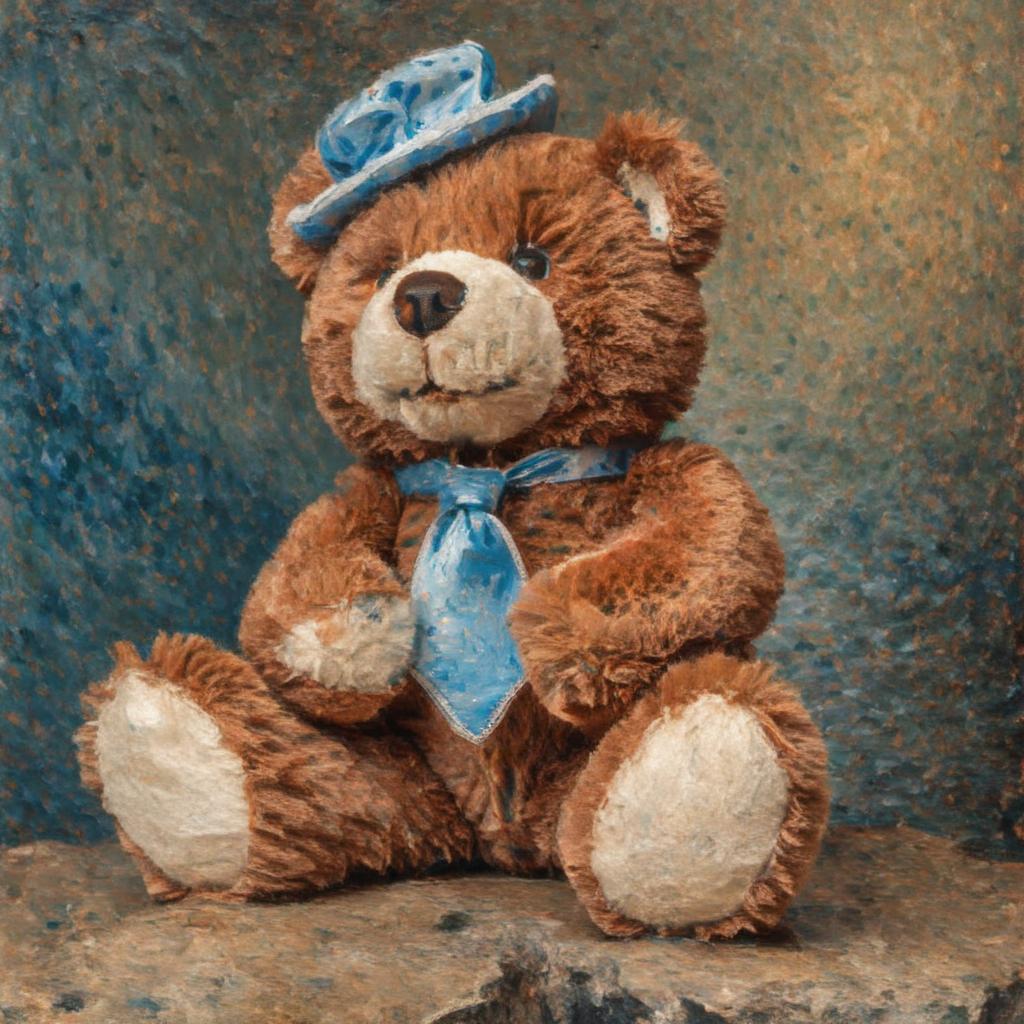} &
    \tabfigure{width=0.12\textwidth}{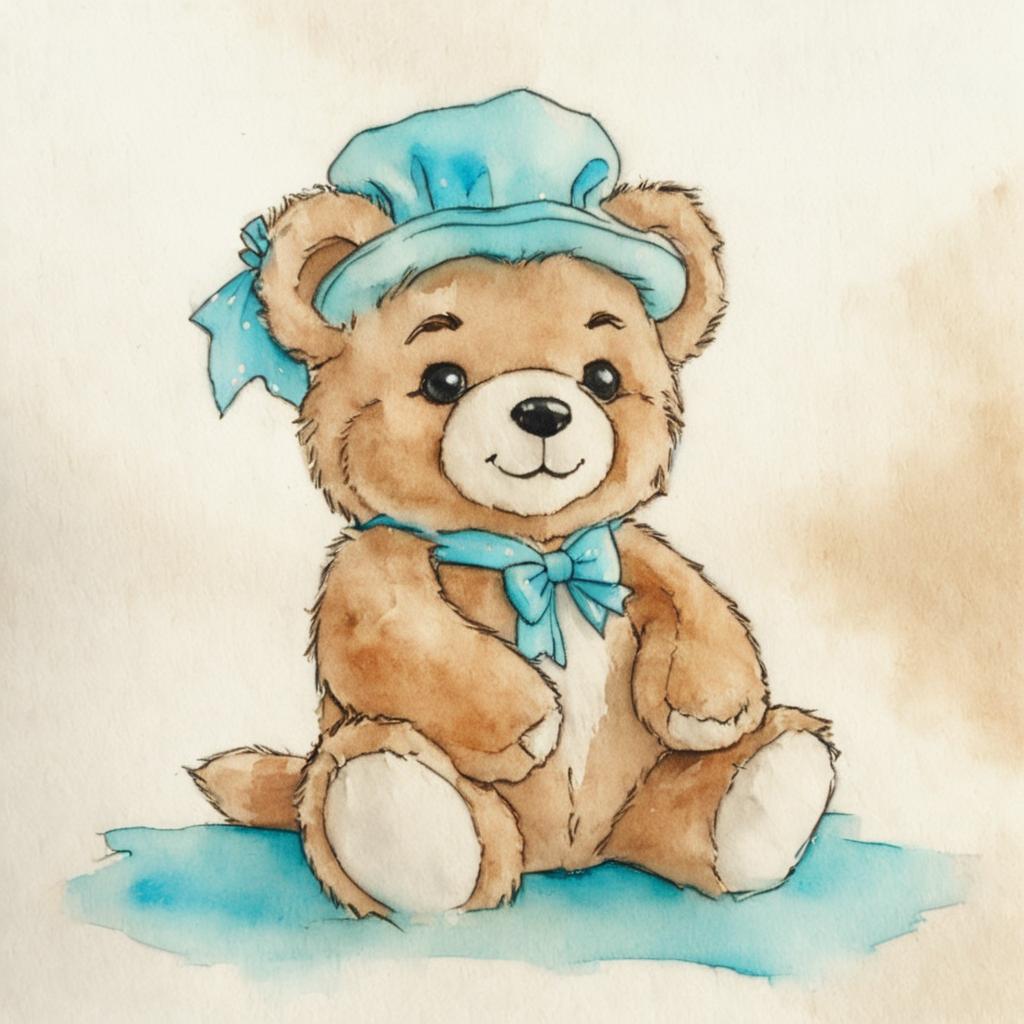} &
    \tabfigure{width=0.12\textwidth}{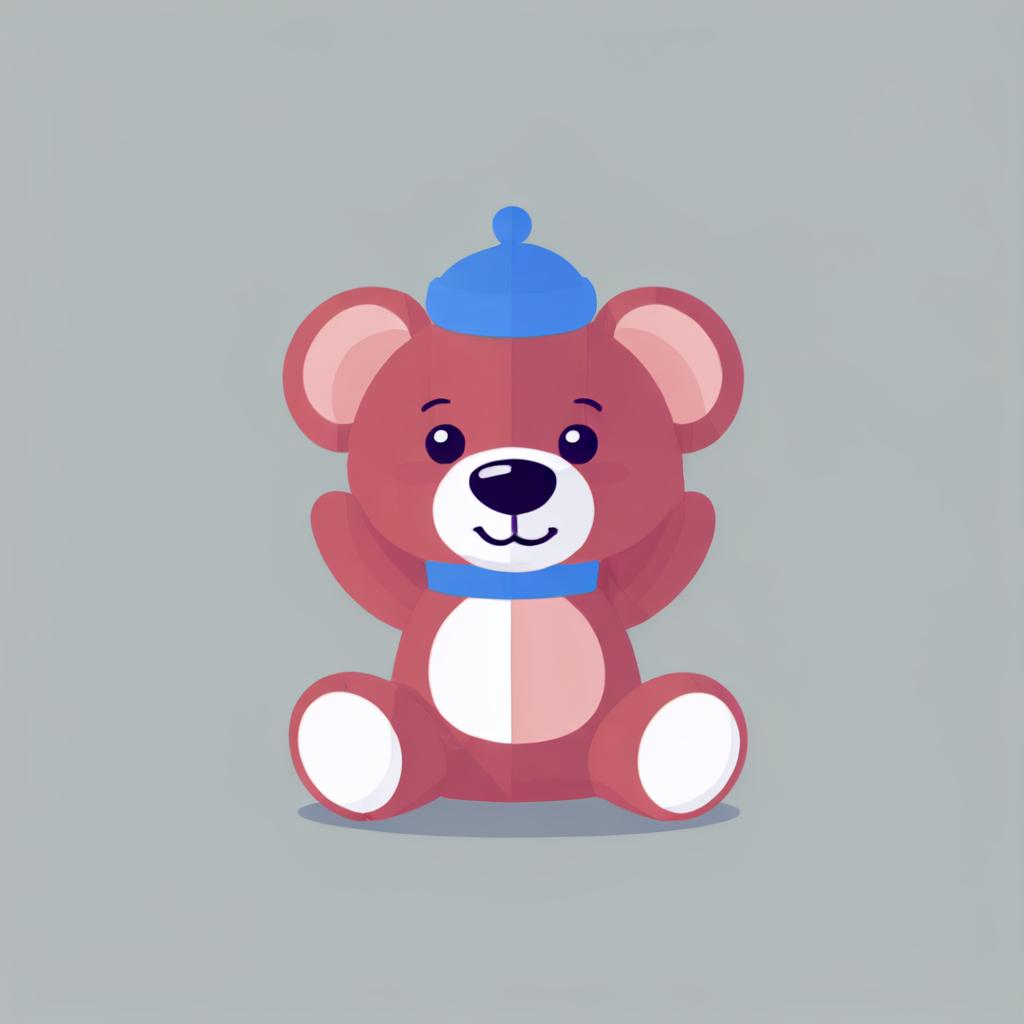} &
    \tabfigure{width=0.12\textwidth}{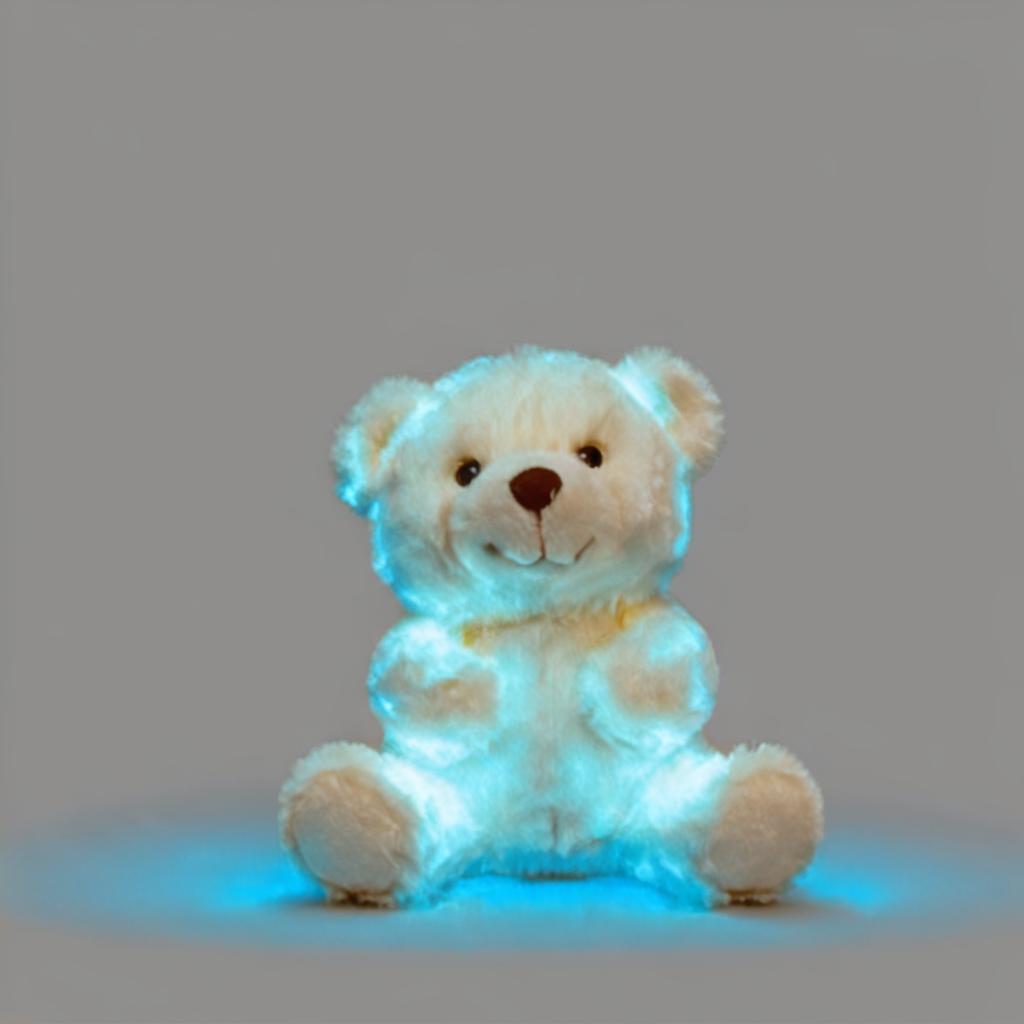} \\

    \tabfigure{width=0.12\textwidth}{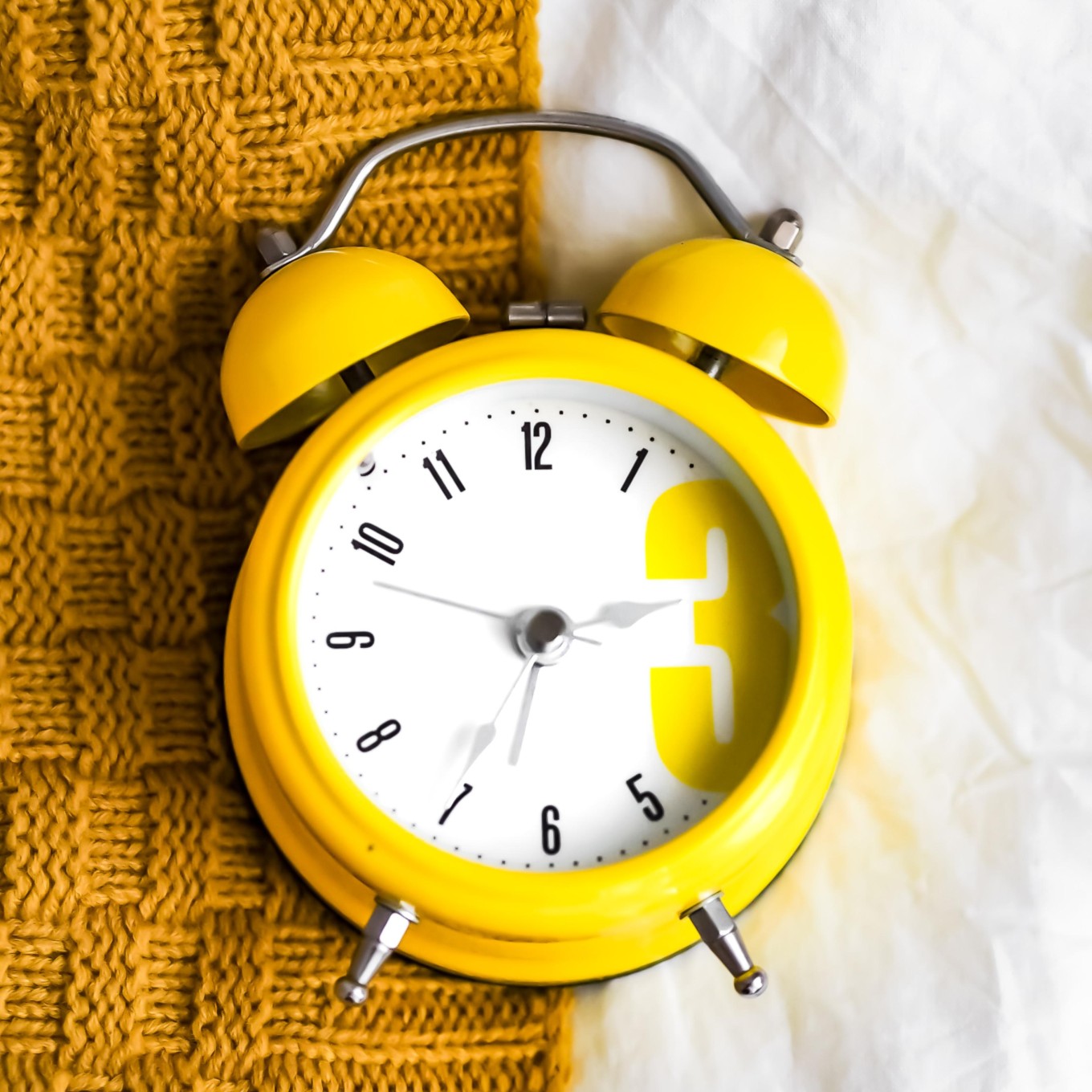} &
    \tabfigure{width=0.12\textwidth}{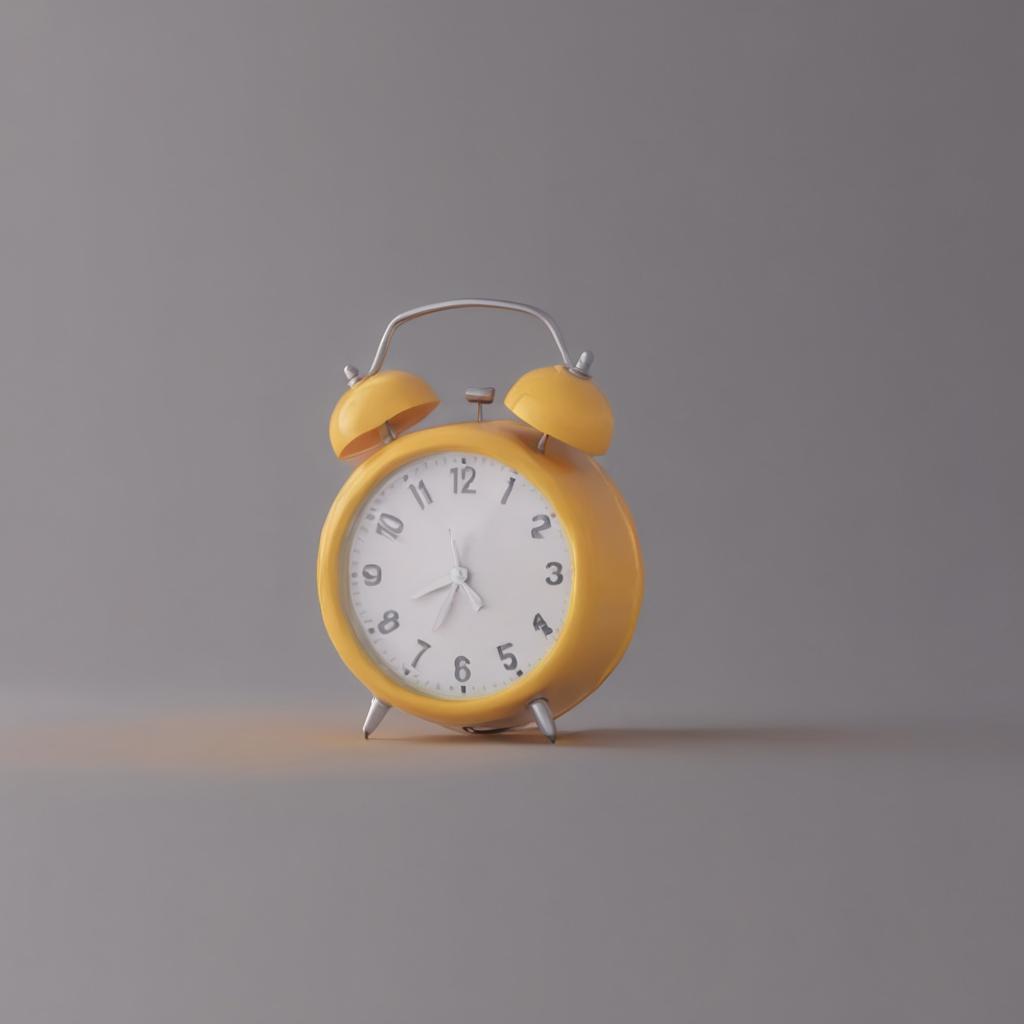} &
    \tabfigure{width=0.12\textwidth}{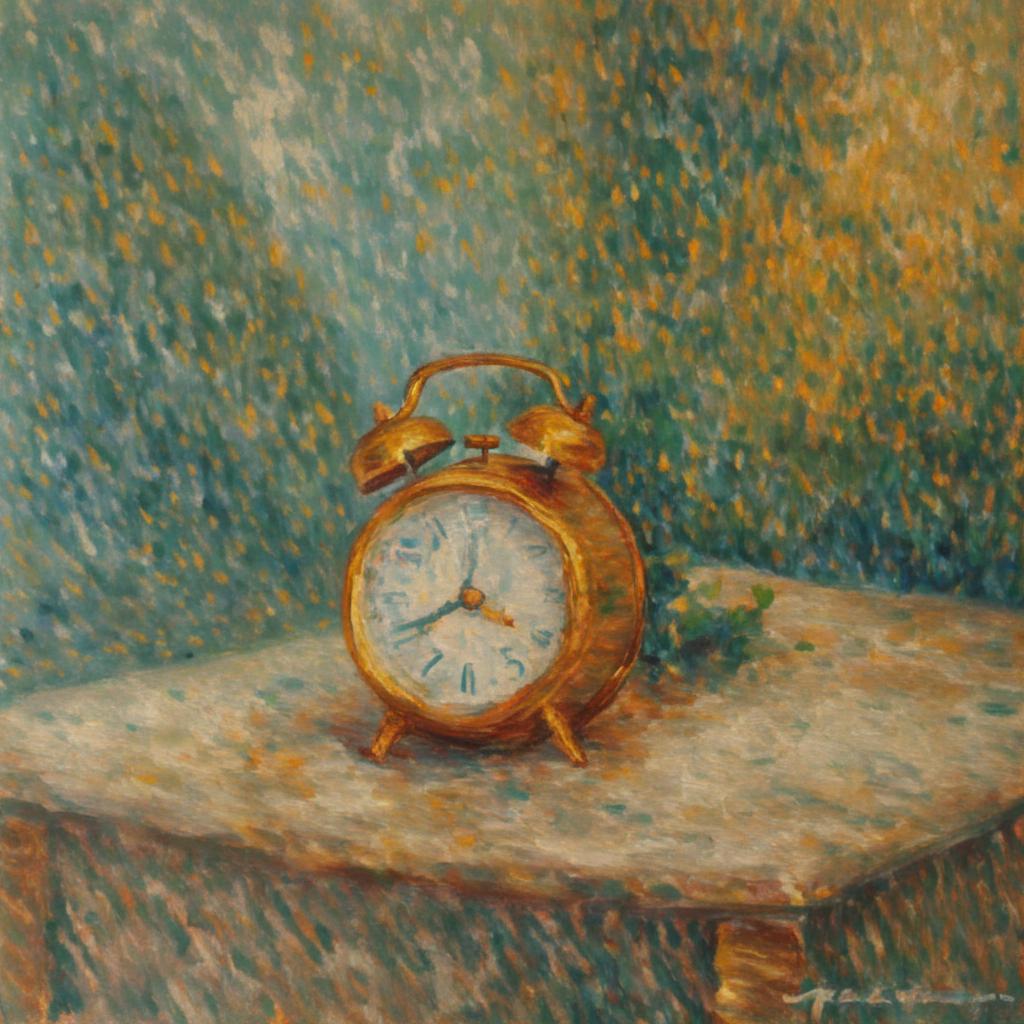} &
    \tabfigure{width=0.12\textwidth}{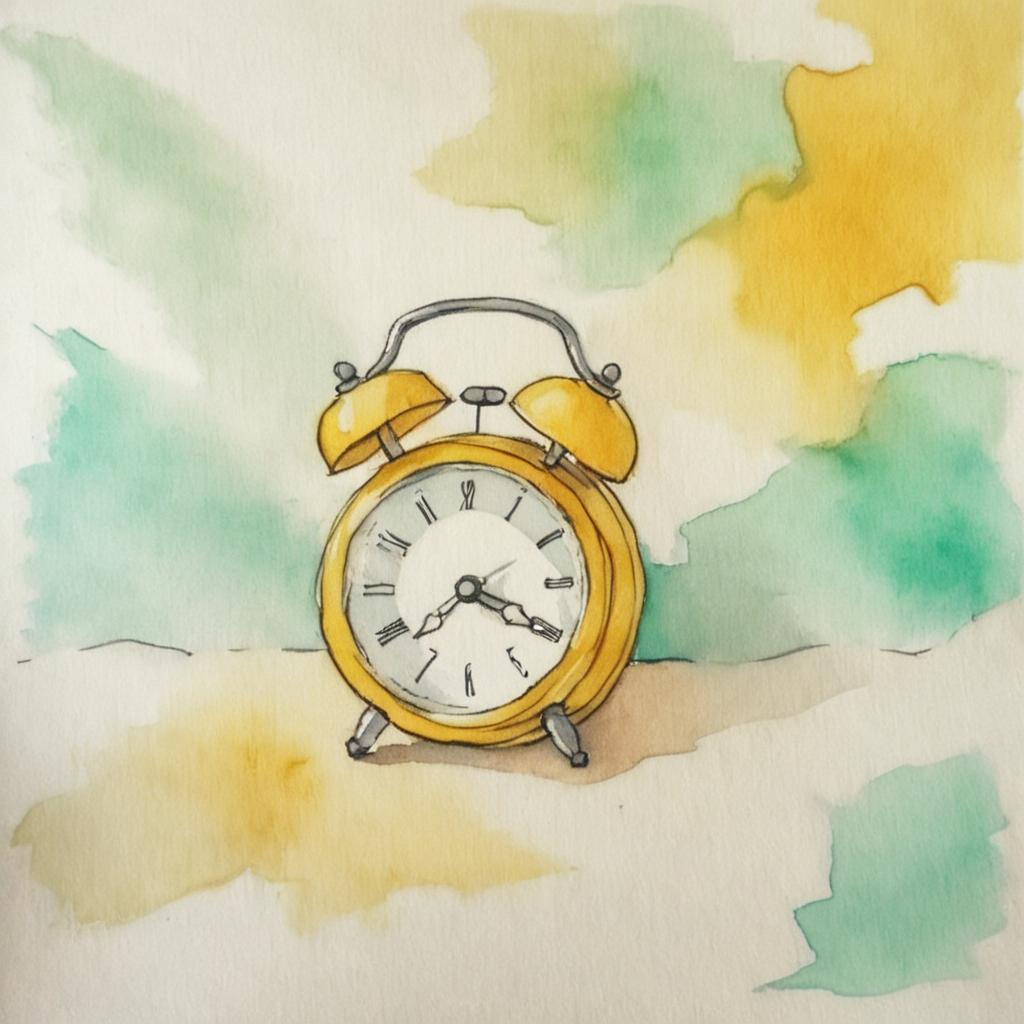} &
    \tabfigure{width=0.12\textwidth}{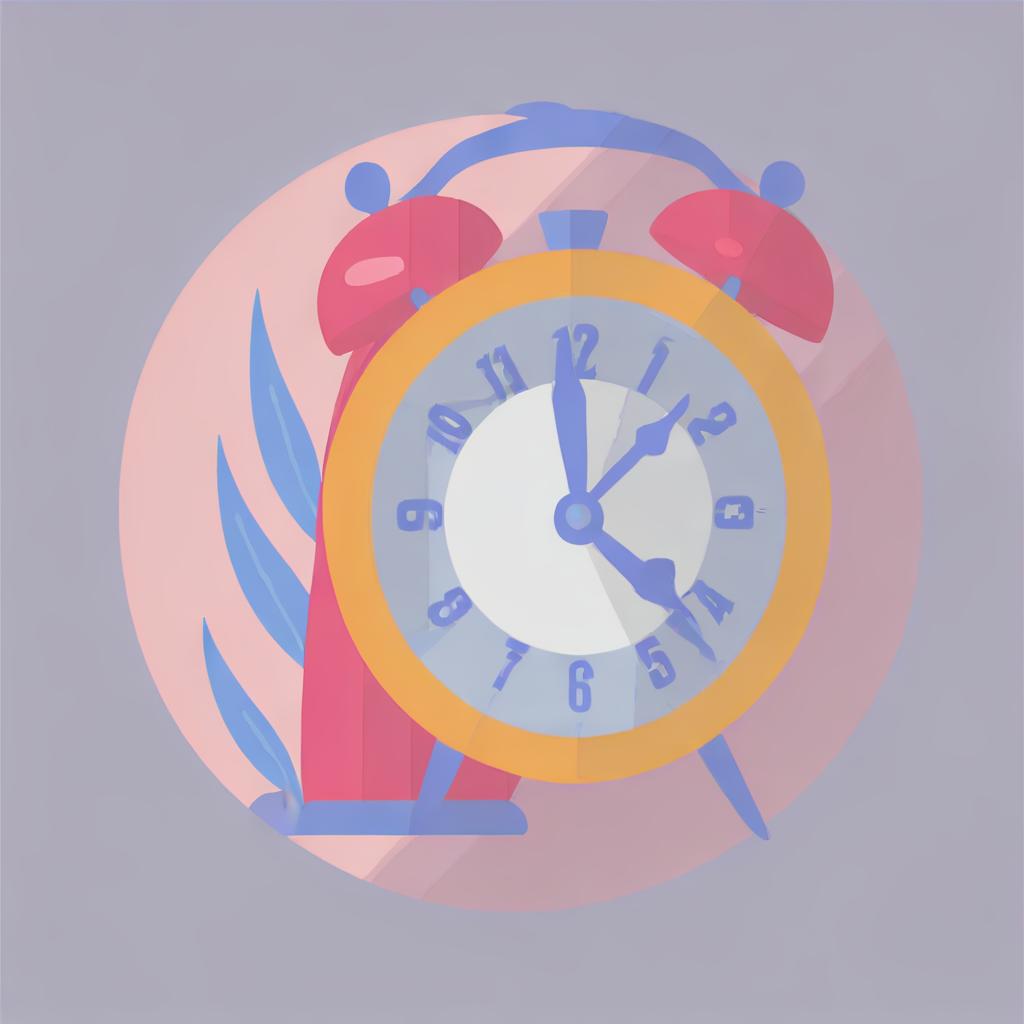} &
    \tabfigure{width=0.12\textwidth}{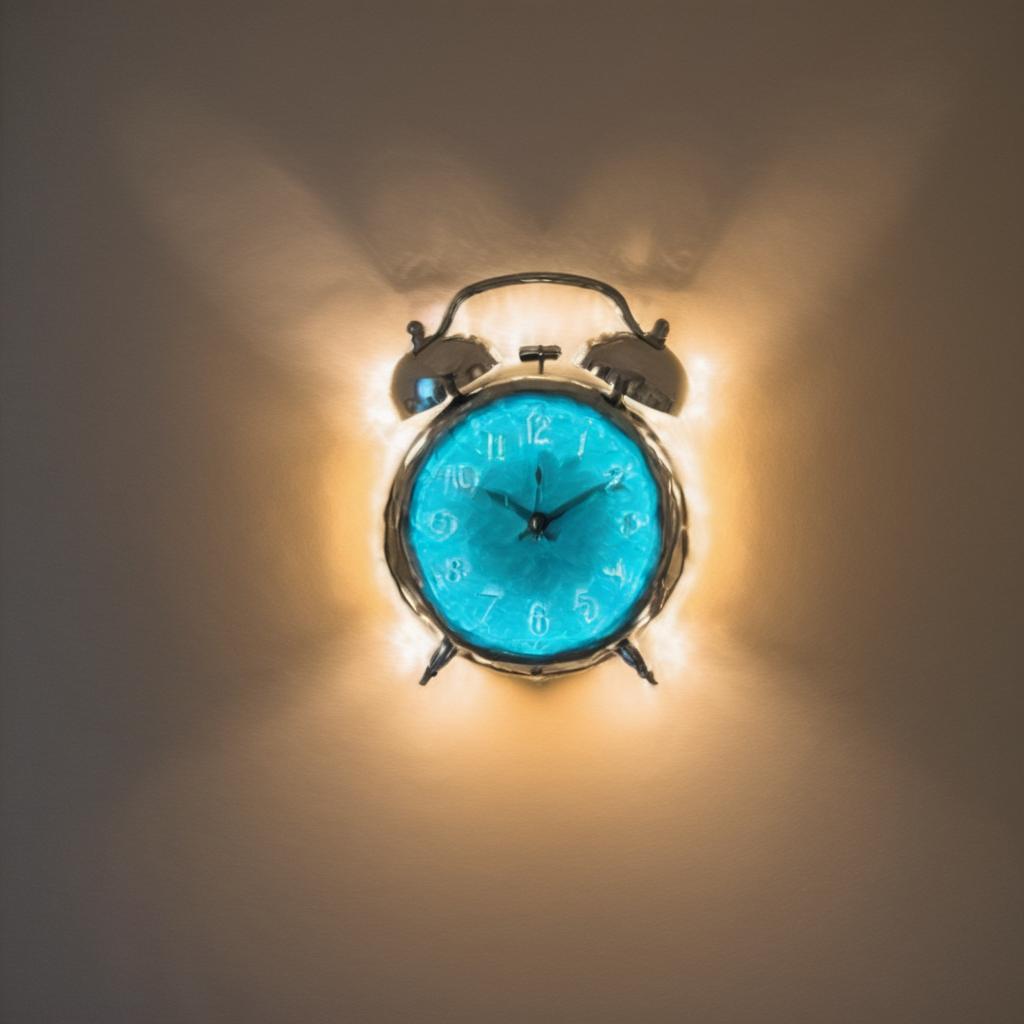} \\
    \end{tabular}
    }
    \caption{\textbf{\ours Evaluation} across different subject-style combinations. Our solution consistently produces good results.}
    \label{fig:evaluation_across_combinations_latex}
\end{figure}

\noindent\textbf{2) Reliability across different subject-style pairs} is shown in Fig.~\ref{fig:evaluation_across_combinations_latex}, where \ours consistently works well across diverse combinations of contents and styles, highlighting its versatility and effectiveness (more examples in \sm.)

\subsection{Additional Analyses}

\begin{table}[t]
\setlength{\tabcolsep}{2.5pt}
\centering
\resizebox{1.0\columnwidth}{!}{
        \begin{tabular}{lcc}
            \toprule
            & ZipLoRA & \textbf{\ours (ours)} \\
            \midrule
             Time to predict merging coeffs & 158s & \textbf{0.037s} \\
             \# Parameters & 1.5M$^\dag$ & \textbf{0.49M} \\
             \# Attempts to \textit{good*} image & 2.55 & \textbf{2.28} \\
             Extra memory at test time & 4GB & \textbf{0GB} \\
             \bottomrule
        \end{tabular}
        }
    \caption{\textbf{Footprint Analysis.} Our \ours is more than $4000\times$ faster and uses $3\times$ fewer parameters than \ziplora, despite using a hypernetwork. *: a \textit{good} image is accepted by \ourmetric. $^\dag$: value for one subject-style pair only.
    }
    \vspace{-0.1cm}
    \label{tab:runtimes}
\end{table}
\begin{table}[t]
    \centering
        \begin{tabular}{l c c}
            \toprule
           & Average case & Best case \\
            \toprule
                \ziplora \cite{shah2024ziplora} & 0.51 & 0.84 \\
                \textbf{\ours (ours)} & \textbf{0.56} & \textbf{0.92} \\
                         \bottomrule
        \end{tabular}
    \caption{\textbf{MLLM Evaluation on KOALA 700m.}
    }
    \label{tab:mllm_evaluation_koala}
\end{table}
We provide a detailed analysis of resource usage in Table~\ref{tab:runtimes}. Our findings highlight the efficiency and scalability of \ours in comparison to ZipLoRA:
(1) \textbf{Runtime Efficiency:} our solution generates the merging coefficients over $4,000$ times faster than ZipLoRA on an NVIDIA 4090, achieving real-time performance. 
While ZipLoRA requires 100 training steps for each content-style pair, \ours generates merging coefficients in a single forward pass (per layer) using a pre-trained hypernetwork.
(2) \textbf{Parameter Storage:} ZipLoRA needs to store the learned coefficients for every combination of content and style for later use. \ours only needs to store the hypernetwork, which has $3$ times fewer parameters than a single ZipLoRA combination. %
(3) \textbf{Sample Efficiency:} on average, \ours requires fewer attempts than ZipLoRA to produce a high-quality image that aligns with both content and style—2.28 attempts for \ours versus 2.55 for ZipLoRA. This improvement reflects LoRA.rar's enhanced accuracy in generating visually coherent outputs without extensive retries, further optimizing resource usage and user experience.
(4) \textbf{Memory Consumption at Test Time:} \ours is efficient in terms of memory, which is dominated by the generative model ($\sim$15GB), with negligible overhead for our approach, while ZipLoRA requires additional 4GB ($\sim$19GB totally).
(5) \textbf{Performance on lightweight diffusion model} is shown in \cref{tab:mllm_evaluation_koala} where \ours  robustly outperforms ZipLoRA also on KOALA 700m \cite{lee2024koala}. See the \sm\ for the qualitative results.

\section{Conclusion}

In this work, we introduced LoRA.rar, a novel method for joint subject-style personalized image generation.
\ours leverages a hypernetwork to generate coefficients for merging content and style LoRAs. 
By training on diverse content-style LoRA pairs, our method can generalize to new, unseen pairs.
Our experiments show that \ours consistently outperforms existing methods in image quality, as assessed by both human evaluators and an MLLM-based judge specifically designed to address the challenges of joint content-style personalization. 
Crucially, \ours generates the merging coefficients in real time, bypassing the need for test-time optimization used by state-of-the-art methods.

\clearpage

\section*{Acknowledgment}
This work was partially supported by the European Union under the Italian National Recovery and Resilience Plan (NRRP) Mission 4, Component 2, Investment 1.3, CUP C93C22005250001, partnership on “Telecommunications of the Future” (PE00000001 - program “RESTART”).

{
    \small
    \bibliographystyle{ieeenat_fullname}
    \bibliography{main}
}

\clearpage

\renewcommand{\thefigure}{A\arabic{figure}}
\renewcommand{\theHfigure}{A\arabic{figure}}
\renewcommand{\thesection}{A\arabic{section}}
\renewcommand{\theHsection}{A\arabic{section}}
\renewcommand{\theequation}{A\arabic{equation}}
\renewcommand{\theHequation}{A\arabic{equation}}
\renewcommand{\thetable}{A\arabic{table}}
\renewcommand{\theHtable}{A\arabic{table}}

\setcounter{equation}{0}
\setcounter{figure}{0}
\setcounter{table}{0}
\setcounter{section}{0}

\maketitlesupplementary

This document includes additional material that was not possible to include in the main paper.
Sec.~\ref{sec:supp:evaluation} presents additional details regarding both MLLM-based and human evaluation, further information on image generation prompts, and it also includes dataset attribution and partitioning details.
Sec.~\ref{sec:supp:results} shows additional results: performance via standard metrics, a thorough ablation study on the hypernetwork design, results on a lightweight diffusion model, generalization to new concepts, new splits, and recontextualization output generations.
Sec.~\ref{sec:supp:discussion} outlines limitations of our approach and discusses its societal impact.

\begin{figure}[b]
    \centering
\begin{tcolorbox}[colback=blue!5!white, colframe=blue!75!black, title=\textbf{Subject Assessment Prompt}]

\textbf{System Prompt}

You are a helpful assistant.\\

\textbf{User Prompt}

Your task is to identify if the test image shows the same subject as the support image.\\

Support image:

\begin{verbatim}
{Image}
\end{verbatim}

Test image:

\begin{verbatim}
{Image}
\end{verbatim}

Pay attention to the details of the subject, it should for example have the same color. However, the general style of the image may be different. 

Does the test image show the same subject as the support image? 

Answer with \textbf{Yes} or \textbf{No} only.\\

\end{tcolorbox}
    \caption{\textbf{Subject Assessment Prompt.} Prompt used to evaluate the subject fidelity on generated images via our MLLM-based metric \ourmetric.}    \label{fig:suppl:subject-prompt}
\end{figure}

\section{Additional Details}
\label{sec:supp:evaluation}

\subsection{MLLM-based Evaluation}

\textbf{Evaluation Prompts.} 
We show the prompts we have used with our MLLM-based \ourmetric~metric using the LLaVA-Critic-7b model. 
The subject assessment prompt is shown in Fig.~\ref{fig:suppl:subject-prompt}, while the style assessment prompt is in Fig.~\ref{fig:suppl:style-prompt}.

\begin{figure}[t]
    \centering
    \includegraphics[width=\linewidth]{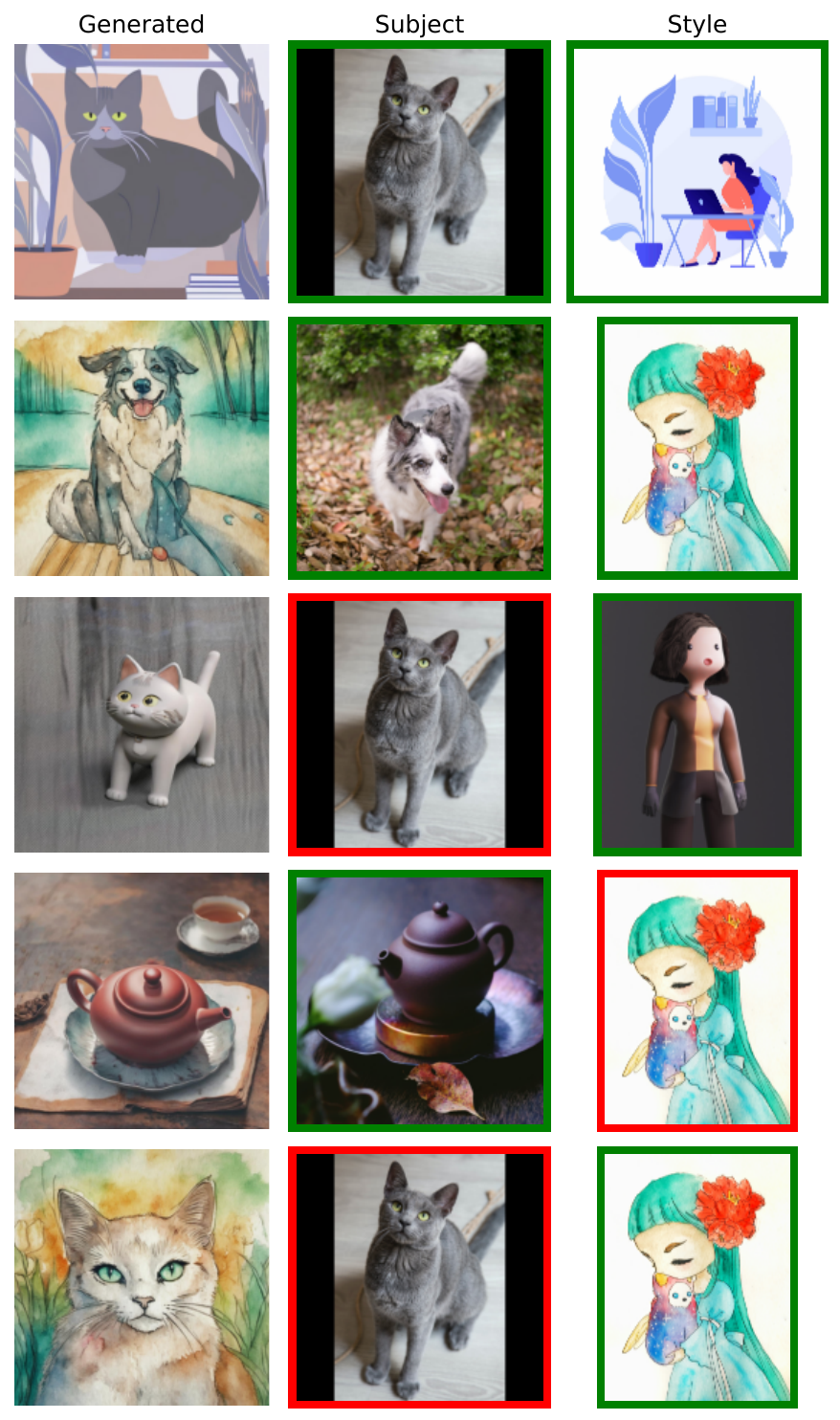}
    \caption{\textbf{MLLM Judge Assessment Samples.} This figure illustrates how the MLLM judge evaluates generated images for subject and style alignment. First column: examples of generated images. Second and third columns: reference subject and style, respectively. Green boxes indicate that the MLLM judge confirms the generated image aligns with the reference subject or style, whereas red boxes denote a mismatch.}
    \label{fig:mllm_eval_examples}
\end{figure}

We test separately for correctness of the generated subject and style as we have found such approach to be more robust. 
We have also manually checked how accurate the MLLM model is in assessing the correctness of the subject and style, taken singularly, and found the quality to be suitable for the task.
We show examples of how the MLLM judge assesses various generated images in terms of the subject or style in Fig.~\ref{fig:mllm_eval_examples}.
In the first and second row, the generated images reproduce the reference subject in the reference style and, therefore, are correctly accepted by the MLLM judge.
Images in third and fifth rows reproduce a generic cat (\eg, white rather than gray) in the correct style, hence the MLLM judge accepts the style but not the subject preservation.
The teapot in the fourth row is preserved in the generated image, but the style is incorrect (\eg, more similar to an oil painting rather than watercolor painting).

\begin{figure}[t]
    \centering
\begin{tcolorbox}[colback=blue!5!white, colframe=blue!75!black, title=\textbf{Style Assessment Prompt}]

\textbf{System Prompt}

You are a helpful assistant.\\

\textbf{User Prompt}

Your task is to identify if the test image shows the subject in \{style\} style. An example image in the \{style\} style is provided.\\

Example image in the \{style\} style:

\begin{verbatim}
{Image}
\end{verbatim}

Test image:

\begin{verbatim}
{Image}
\end{verbatim}

The example image shows an illustration of the \{style\} style and the details of the subject are expected to be different.

Do not check similarity with the subject.

Is the test image in the \{style\} style?

Answer with \textbf{Yes} or \textbf{No} only.\\

\end{tcolorbox}
    \caption{\textbf{Style Assessment Prompt.} Prompt used to evaluate the style on generated images via our MLLM-based metric \ourmetric.}
    \label{fig:suppl:style-prompt}
\end{figure}

\subsection{Human Evaluation Study}

As part of the human evaluation study, we asked 25 participants to compare two generated images at a time, given reference subject and style images. The images are generated by either our approach or ZipLoRA, and they are randomly ordered in each pair. We test 25 subject-style combinations with one pair of generated images for each.
The combinations are also randomly ordered. We consider two scenarios, one where we use randomly generated images and one where we take the ``best'' images as judged by the MLLM judge. In the ``best'' scenario, we gathered all the images that satisfied both subject and style according to the MLLM judge and then selected one randomly among those--there was always at least one such example for each approach.

\begin{figure}[t]
    \centering
    \includegraphics[width=\linewidth]{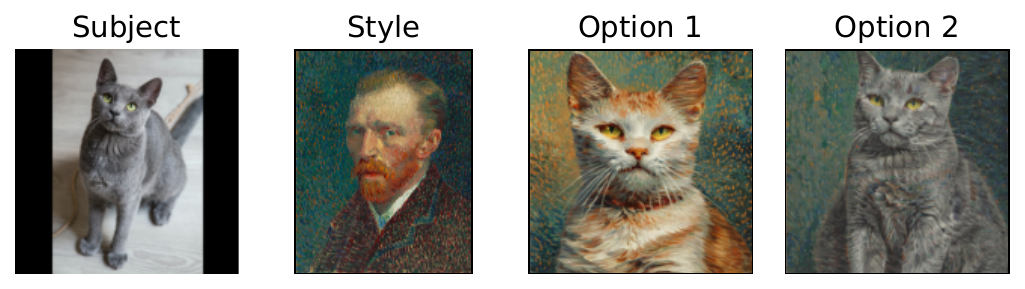}
    \caption{\textbf{Example Case for Evaluators.} Example used to teach human evaluators how to evaluate the generated images. In this example, the participant should select \textit{Option 2} as better, because the generated image in Option 2 represents the target subject in the target style. Option 1 follows the style, but generates a random cat instead.}
    \label{fig:human_eval_example}
\end{figure}

We introduced and explained the task to the evaluators via the example shown in Fig.~\ref{fig:human_eval_example} and the following textual instruction: \textit{``Your task is to evaluate which of two generated images better represents the given subject and style -- or if they are similarly good. You are provided with an image showing the subject (e.g. black cat) and an image showing the image style (e.g. van Gogh style painting), and two generated images such as in the example below. In this example you would select option 2 as better because it shows a cat that looks like the one in the subject image, and both images follow the style.''}.

The evaluation was done via a web app that shows the images and lets the participant click on a button saying which option is better among: \textit{``Option 1'', ``Similar'', ``Option 2''}.

\subsection{Additional Experimental Details}
\label{sec:supp:experimental}

\noindent \textbf{Prompts Used for Image Generation.} 
The prompts used to generate the images for the main paper qualitative and quantitative results are of the form: ``\verb|A [c] <class name> in [s] style|''. 
For ``\verb|[c]|'' we used the rare token used to train the content LoRAs and for ``\verb|<class name>|'' we used the same name as DreamBooth \cite{ruiz2023dreambooth}. Finally, for ``\verb|[s]|'' we used the short text description as in StyleDrop, in particular it corresponds to the style name that we assigned (after removing the number, if present). The full list of names is detailed in \cref{sec:dataset_partitioning}.

\begin{table*}[!h]
    \centering
    \begin{tabular}{p{0.08\textwidth} p{0.42\textwidth} p{0.42\textwidth} }
    \toprule
         & Contents & Styles \\
         \toprule
       Train  &  backpack, backpack dog, berry bowl, candle, cat \#1, colorful sneaker, dog \#1, dog \#5, dog \#6, dog \#7, duck toy, fancy boot, grey sloth plushie, monster toy, pink sunglasses, poop emoji, rc car, red cartoon, robot toy, shiny sneaker, vase &  3D rendering \#1, 3D rendering \#3, abstract rainbow, black statue, cartoon line drawing, flat cartoon illustration \#1, glowing 3D rendering, kid crayon drawing, line drawing, melting golden rendering, oil painting \#3, sticker, watercolor painting \#2, watercolor painting \#4, watercolor painting \#5, watercolor painting \#6, watercolor painting \#7, wooden sculpture\\
       \midrule
       Validation &  dog \#2, dog \#3, clock, bear plushie & 3D rendering \#2, oil painting \#1, watercolor painting \#1 \\
       \midrule
       Test &  dog \#8, cat \#2, wolf plushie, teapot, can & 3D rendering \#4, oil painting \#2, watercolor painting \#3, flat cartoon illustration \#2, glowing\\
       \bottomrule
    \end{tabular}
    \caption{\textbf{Dataset partitioning.} Contents and styles LoRAs train/validation/test splits.}
    \label{tab:dataset_split}
\end{table*}

\noindent \textbf{Additional Implementation Details.}  
Base LoRAs are trained as in \cite{shah2024ziplora}, for $1000$ fine-tuning steps, with batch size 1, a learning rate of $5\times 10^{-5}$ and a rank of $64$. The text encoder remains frozen during training. The hypernetwork used is a two-layer MLP with two separate input layers of size 1280 and 2560, followed by a ReLU activation function, a shared hidden layer of size 128, and two outputs.
We train our hypernetwork for 100 different $\{L_c, L_s\}$ combinations (totalling 5000 steps), with  $\lambda\!=\!0.01$, learning rate $0.01$ and the AdamW optimizer. 
For ZipLoRA, we use a training setup of 100 steps with the same $\lambda$ and learning rate. 
The DARE, TIES, and DARE-TIES baselines are evaluated with uniform weights and a density of 0.5.
For joint training, we used a multi-concept variant of Dreambooth LoRA  as in \cite{shah2024ziplora}.
In all experiments, 50 diffusion inference steps are used.

\subsection{Additional Dataset Details}
\label{sec:supp:data_attribution}

\begin{figure}[!h]
    \centering
    \resizebox{0.5\textwidth}{!}{
    \begin{tabular}[c]{c c c c c c}
        Contents [c] & \tabfigure{width=0.12\textwidth}{figures/dataset/test/dog8/04.jpg} 
        & \tabfigure{width=0.12\textwidth}{figures/dataset/test/cat2/03.jpg}
        & \tabfigure{width=0.12\textwidth}{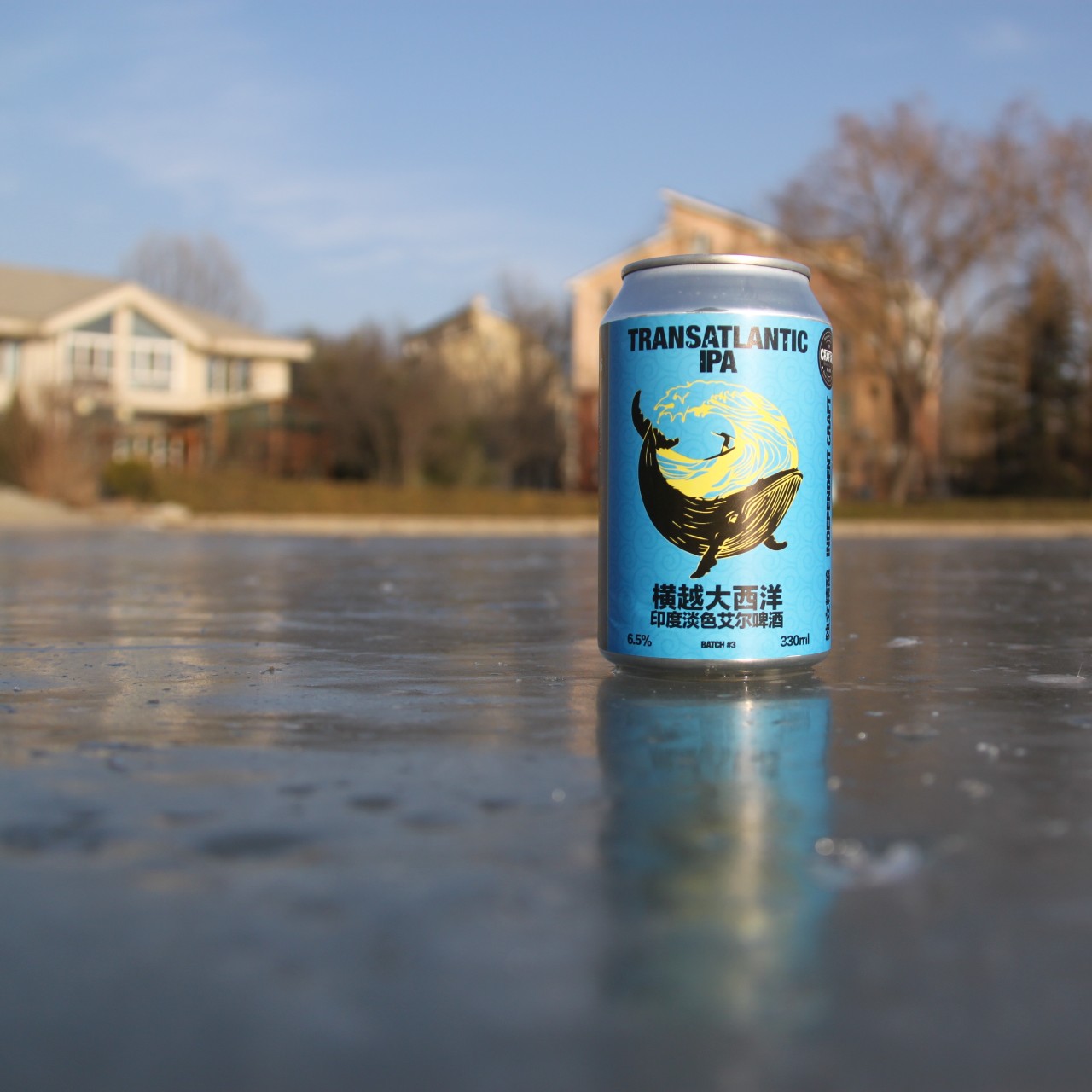}
        & \tabfigure{width=0.12\textwidth}{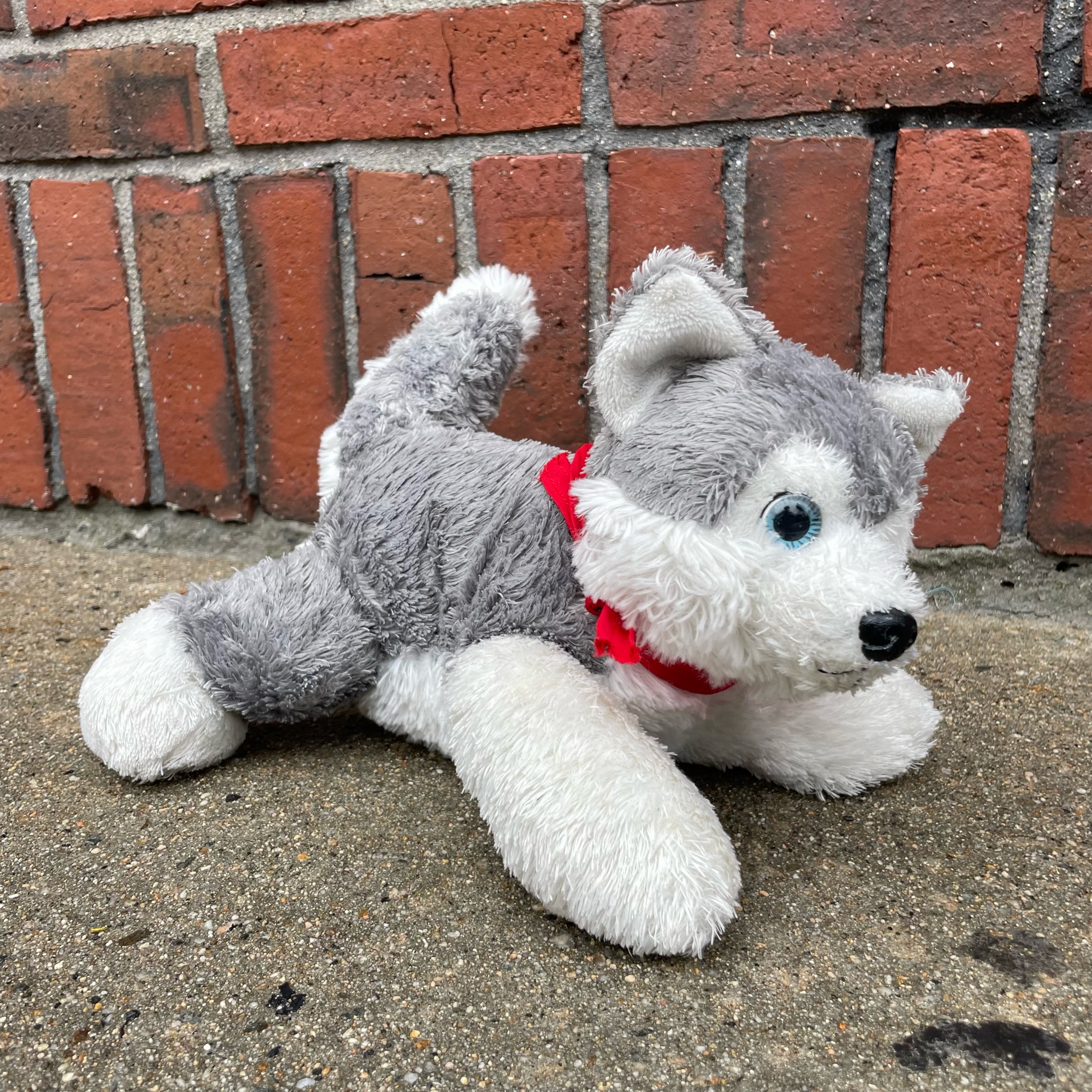}
        & \tabfigure{width=0.12\textwidth}{figures/dataset/test/teapot/00.jpg} \\
        Styles [s] & \tabfigure{width=0.12\textwidth}{figures/dataset/test/3d_rendering.jpg} 
        & \tabfigure{width=0.12\textwidth}{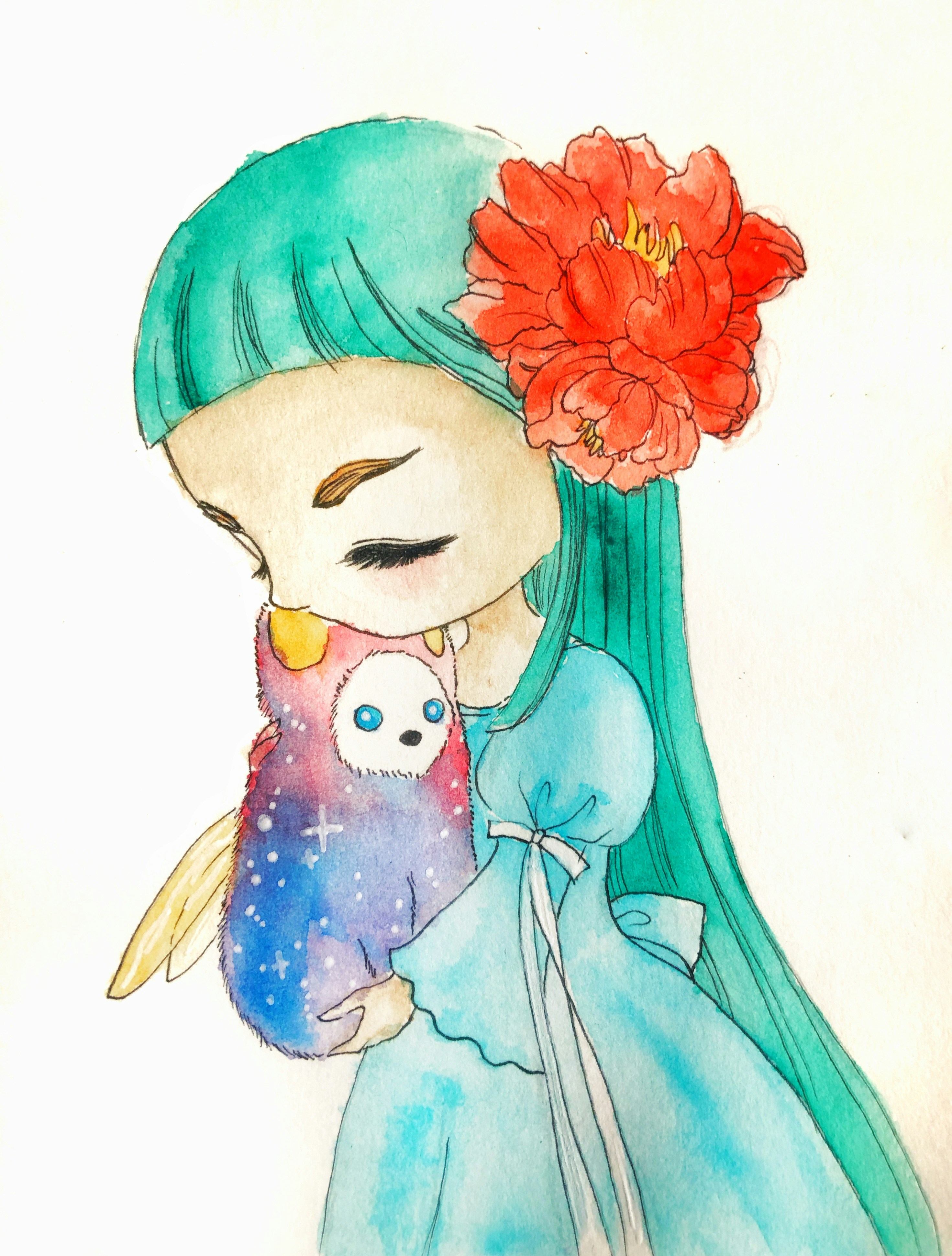}
        & \tabfigure{width=0.12\textwidth}{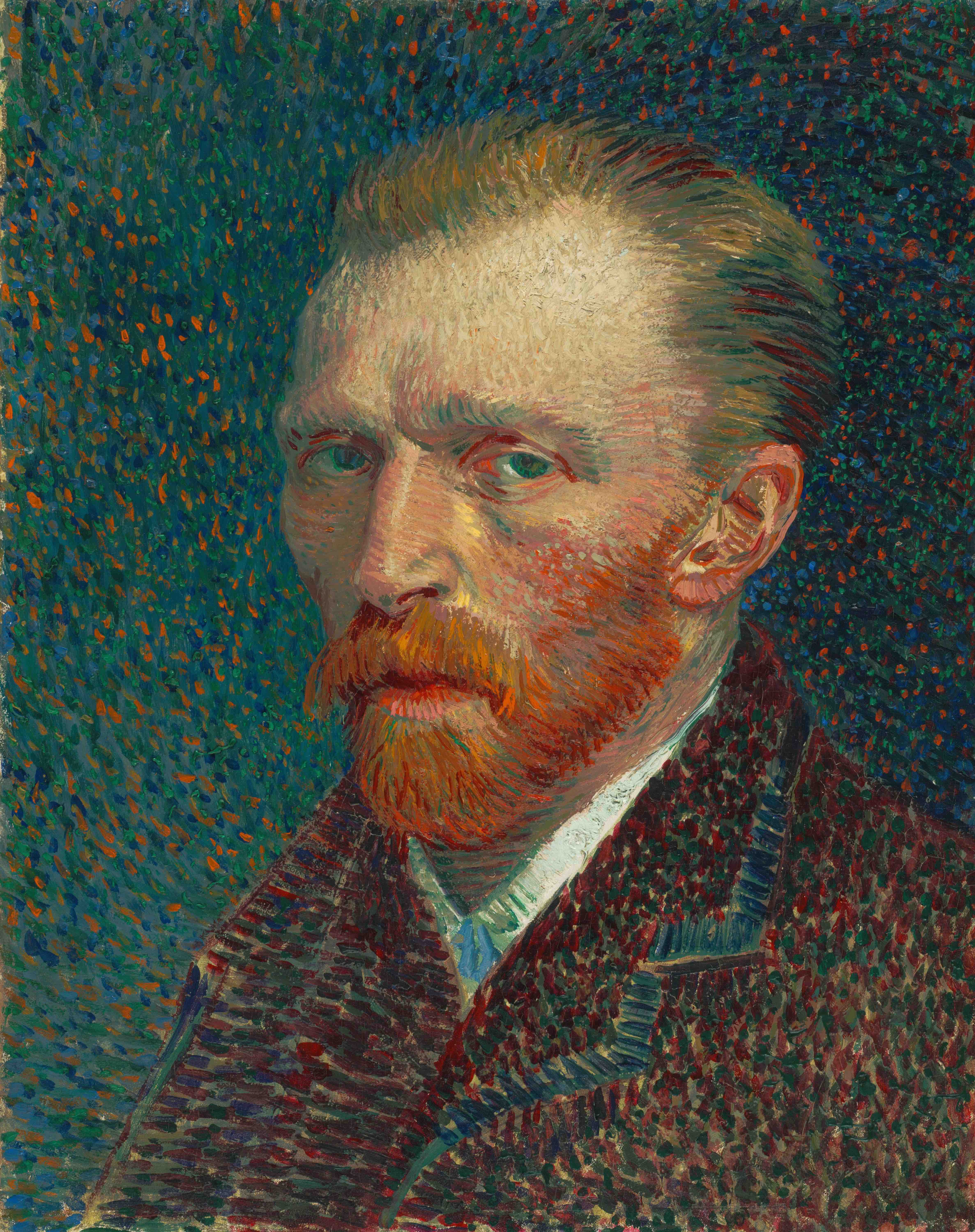} 
        & \tabfigure{width=0.12\textwidth}{figures/dataset/test/flat.jpg}
        & \tabfigure{width=0.12\textwidth}{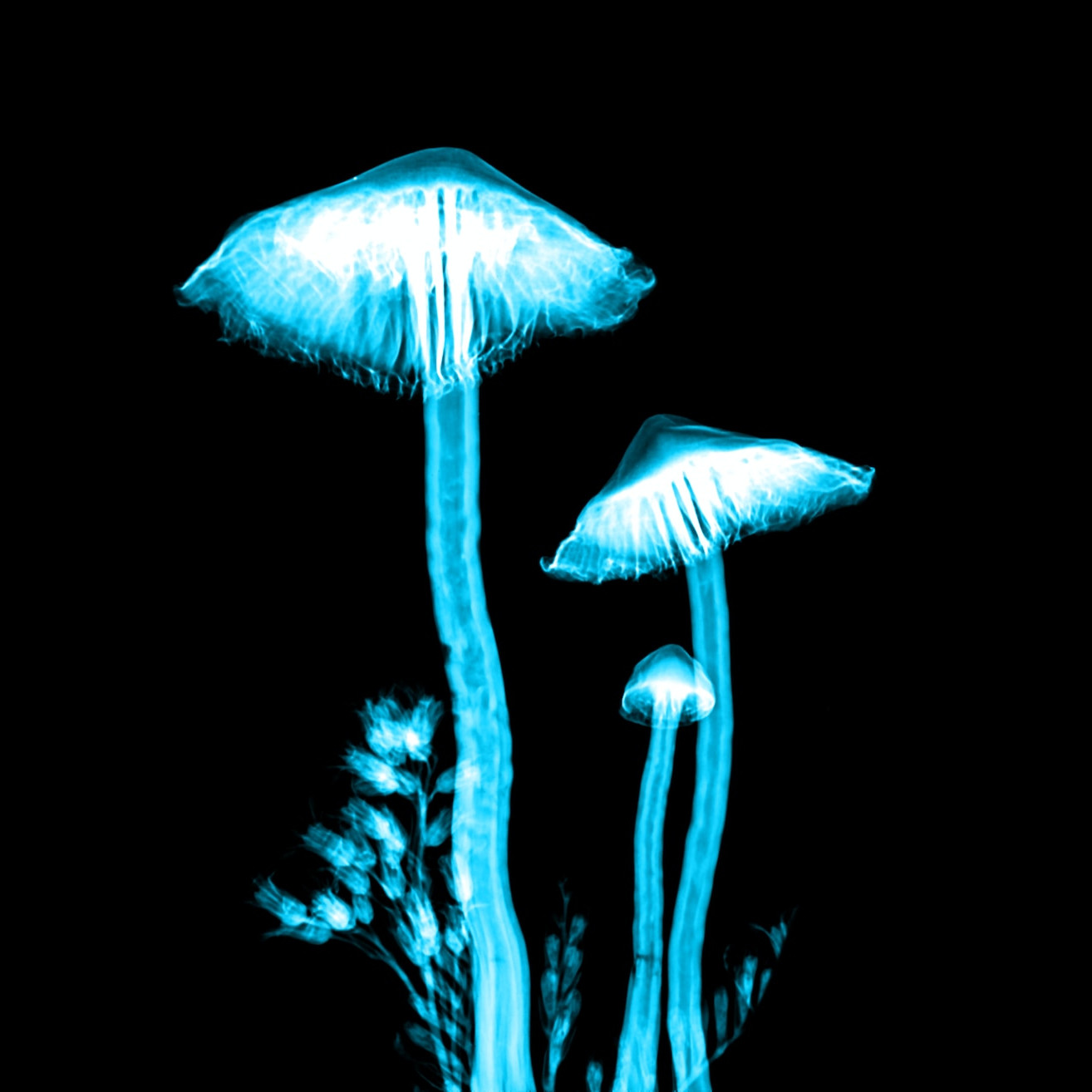} 
    \end{tabular}
    }
    \caption{\textbf{Test Set Samples.} Subject and styles of the test set in our data partitioning.}
    \label{tab:test_dataset}
\end{figure}
\begin{figure}[!h]
    \centering
    \resizebox{0.5\textwidth}{!}{
    \begin{tabular}[c]{c c c c c}
        Contents [c] & \tabfigure{width=0.12\textwidth}{figures/dataset/validation/dog2} 
        & \tabfigure{width=0.12\textwidth}{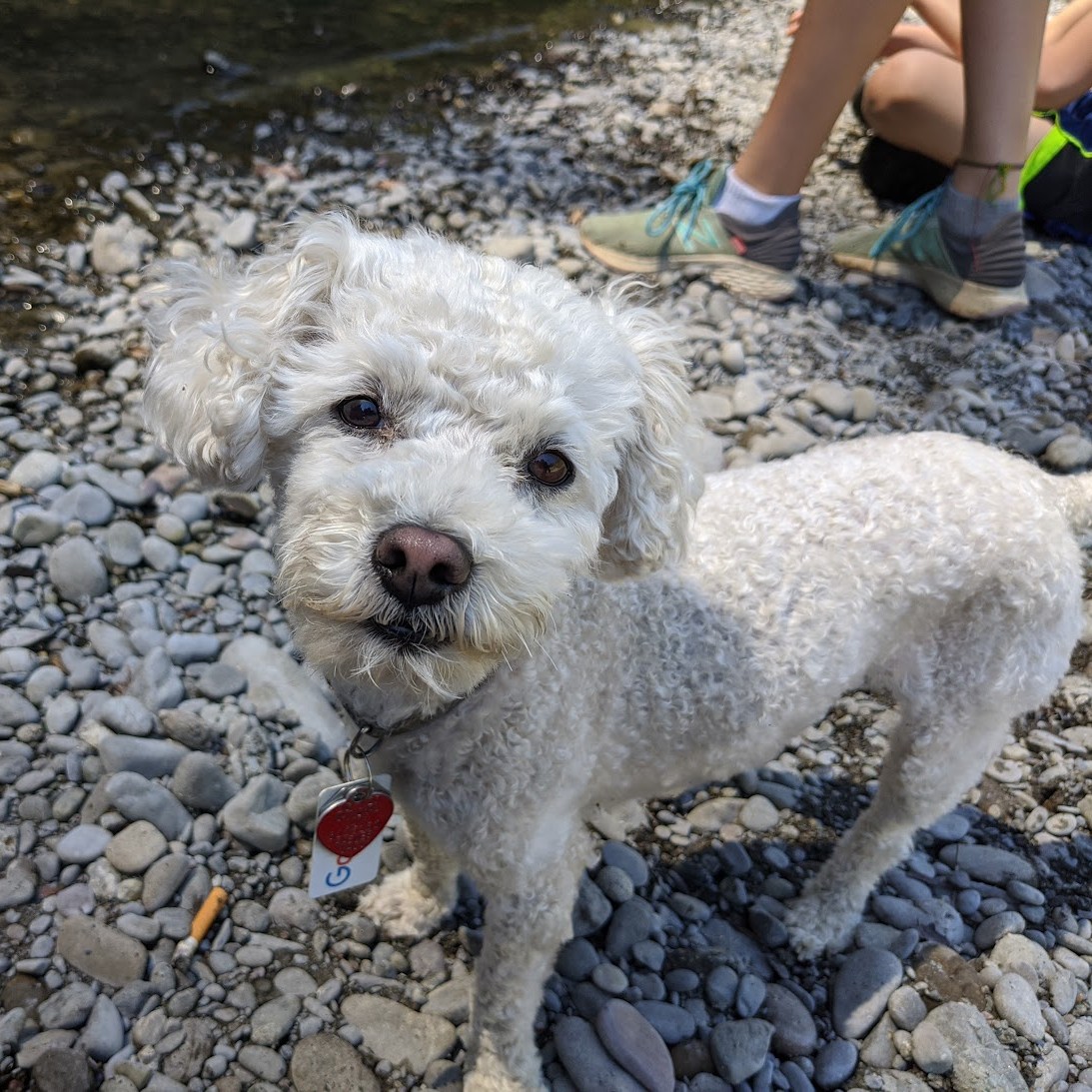} 
        \vspace{0.05cm}
        & \tabfigure{width=0.12\textwidth}{figures/dataset/validation/clock}
        & \tabfigure{width=0.12\textwidth}{figures/dataset/validation/bear_plushie}\\
        Styles [s] & \tabfigure{width=0.12\textwidth}{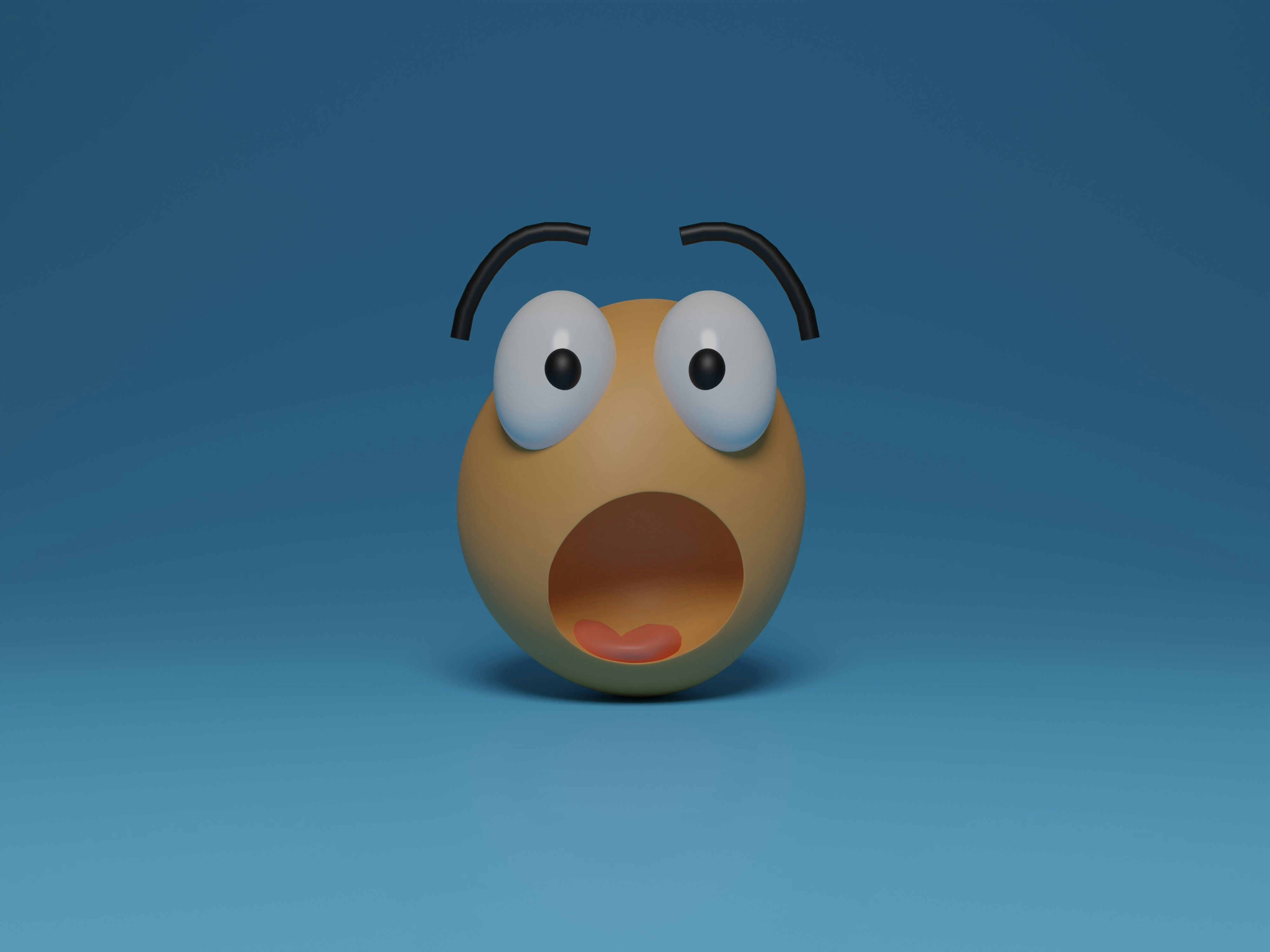} 
        & 
        \tabfigure{width=0.12\textwidth}{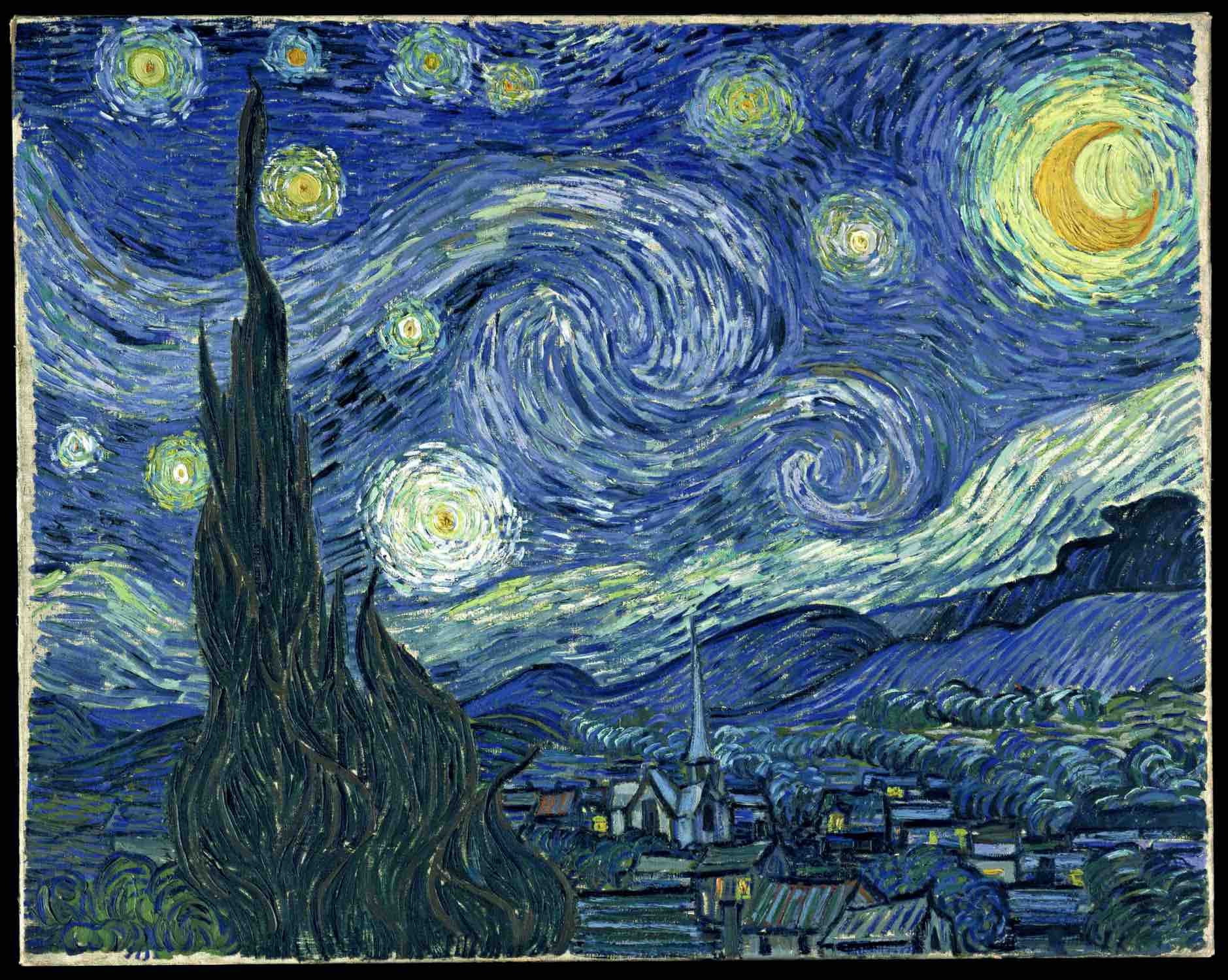}
        & \tabfigure{width=0.12\textwidth}{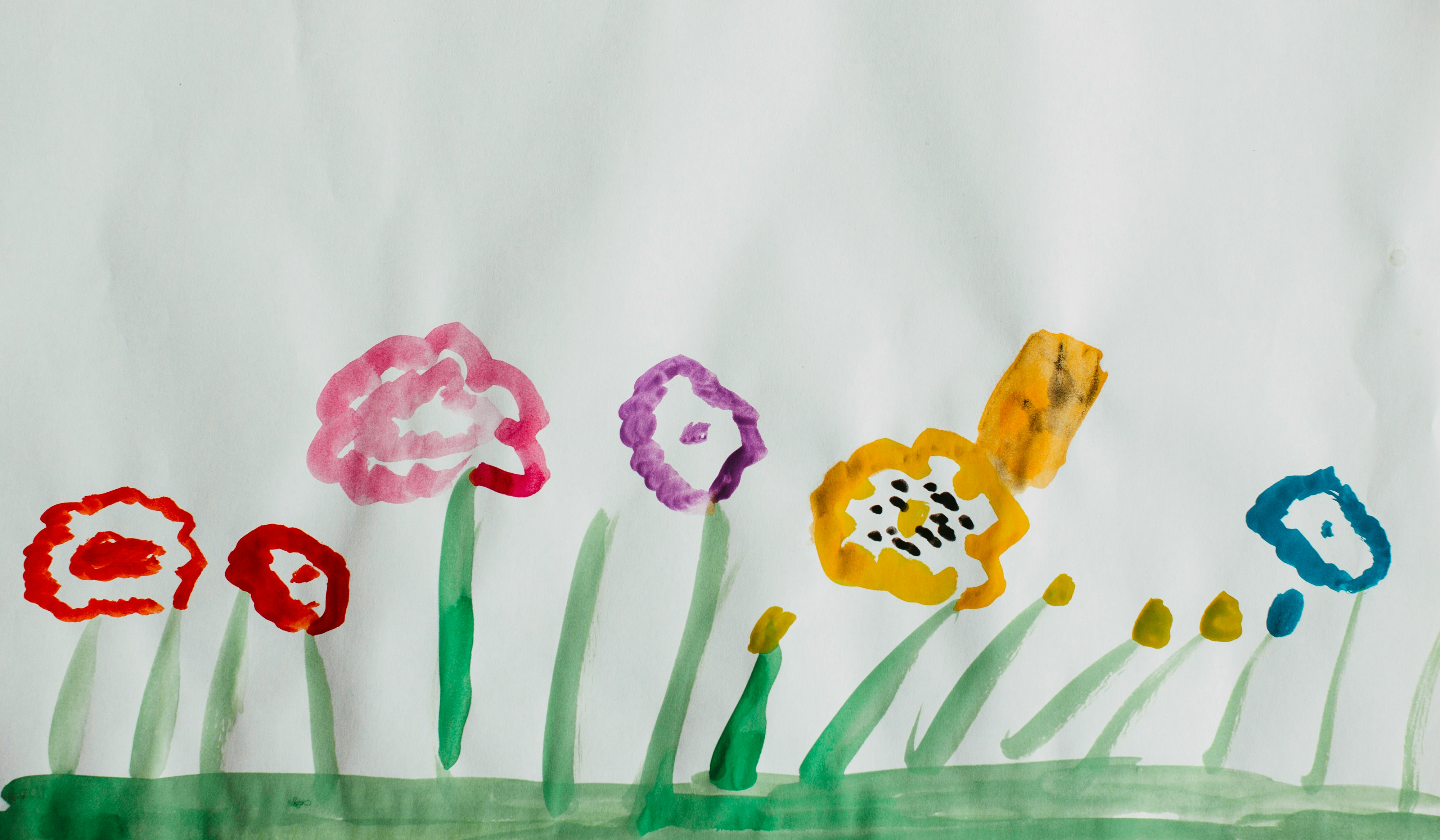} 
        & \\
    \end{tabular}
    }
    \caption{\textbf{Validation Set Samples.} Subject and styles of the validation set in our data partitioning.}
    \label{tab:val_dataset}
\end{figure}
We use the style images from the datasets collected by StyleDrop / ZipLoRA \cite{styledrop, shah2024ziplora}, while the subject images are taken from the DreamBooth \cite{ruiz2023dreambooth} dataset. 
Note that these datasets do not contain any human subjects data or personally identifiable information. We provide image attributions below for each image that we used in our experiments. 
We refer readers to manuscripts and project websites of StyleDrop, ZipLoRA and DreamBooth for more detailed information about the usage policy and licensing of these images.

\noindent \textbf{Attribution for Style Reference Images}
\label{sec:supp:data_attribution:style}
StyleDrop project webpage provides the image attribution information \href{https://github.com/styledrop/styledrop.github.io/blob/main/images/assets/data.md}{here}.
In particular, we used the following 20 styles: 
\href{https://unsplash.com/photos/0e6nHU8GRUY}{S1 (3D rendering \#1)}, 
\href{https://unsplash.com/photos/pink-yellow-and-green-flower-decors-6dY9cFY-qTo}{S2 (watercolor painting \#1)},  
\href{https://www.freepik.com/free-psd/three-dimensional-real-estate-icon-mock-up_32453229.htm}{S3 (3D rendering \#3)}, 
\href{https://it.freepik.com/vettori-gratuito/adesivo-albero-di-pino-su-sfondo-bianco_20710341.htm}{S4 (sticker)}, 
\href{https://www.freepik.com/free-vector/young-woman-walking-dog-leash-girl-leading-pet-park-flat-illustration_11236131.htm}{S5 (flat cartoon illustration \#2)}, 
\href{https://unsplash.com/photos/0pJPixfGfVo}{S6 (watercolor painting \#5)}, 
\href{https://img.freepik.com/free-vector/biophilic-design-workspace-abstract-concept_335657-3081.jpg}{S7 (flat cartoon illustration \#1)}, 
\href{https://unsplash.com/photos/a-golden-flower-with-drops-of-liquid-on-it-Prx96KdmWj0}{S8 (melting golden rendering)}, 
\href{https://github.com/styledrop/styledrop.github.io/blob/main/images/assets/image_6487327_crayon_02.jpg}{S9 (kid crawyon drawing)}, 
\href{https://unsplash.com/photos/a-wooden-carving-of-a-man-with-a-beard-CuWq_99U0xs}{S10 (wooden sculpture)},
\href{https://upload.wikimedia.org/wikipedia/commons/thumb/a/aa/Vincent_van_Gogh_-_Self-portrait_with_grey_felt_hat_-_Google_Art_Project.jpg/1024px-Vincent_van_Gogh_-_Self-portrait_with_grey_felt_hat_-_Google_Art_Project.jpg}{S11 (oil painting \#3)},  
\href{https://images.unsplash.com/photo-1578927107994-75410e4dcd51}{S12 (watercolor painting \#7)},  
\href{https://images.unsplash.com/photo-1612760721786-a42eb89aba02}{S13 (watercolor painting \#6)},  
\href{https://upload.wikimedia.org/wikipedia/commons/6/66/VanGogh-starry_night_ballance1.jpg}{S14 (oil painting \#1)},  
\href{https://upload.wikimedia.org/wikipedia/commons/d/de/Van_Gogh_Starry_Night_Drawing.jpg}{S15 (line drawing)},  
\href{https://upload.wikimedia.org/wikipedia/commons/thumb/4/4c/Vincent_van_Gogh_-_Self-Portrait_-_Google_Art_Project_%28454045%29.jpg/1024px-Vincent_van_Gogh_-_Self-Portrait_-_Google_Art_Project_%28454045%29.jpg}{S16 (oil painting \#2)},
\href{https://img.freepik.com/free-psd/abstract-background-design_1297-124.jpg}{S17 (abstract rainbow colored flowing smoke wave design)},
\href{https://images.unsplash.com/photo-1538836026403-e143e8a59f04}{S18 (glowing)},  
\href{https://images.unsplash.com/photo-1644664477908-f8c4b1d215c4}{S19 (glowing 3D rendering)},  
\href{https://images.unsplash.com/photo-1634926878768-2a5b3c42f139}{S20 (3D rendering \#4)}.
Additionally, we also used 6 styles from ZipLoRA (linked as hyperlinks): 
\href{https://unsplash.com/photos/t0Bv0OBQuTg}{S21 (3D rendering \#2)}, 
\href{https://unsplash.com/photos/H9g_HE6ZgGA}{S22 (watercolor painting \#2)},
\href{https://unsplash.com/photos/jI3Lp0FYEz0}{S23 (watercolor painting \#3)}, 
\href{https://unsplash.com/photos/kHuCUkkExbc}{S24 (watercolor painting \#4)}, 
\href{https://www.instagram.com/p/CqwU1bavm0T/}{S25 (cartoon line drawing)}, 
\href{https://unsplash.com/photos/gargoyle-statue-gZzUo--BTZ4}{S26 (black statue)}. 

\noindent \textbf{Attribution for Subject Reference Images}
The DreamBooth project webpage provides the image attribution information \href{https://github.com/google/dreambooth/blob/main/dataset/references_and_licenses.txt}{here}.
Specifically, the sources of the content images that we used in our experiments
are as follows (linked as hyperlinks):
\href{https://github.com/google/dreambooth/tree/main/dataset/backpack}{C1},
\href{https://github.com/google/dreambooth/tree/main/dataset/backpack_dog}{C2},
\href{https://github.com/google/dreambooth/tree/main/dataset/bear_plushie}{C3},
\href{https://github.com/google/dreambooth/tree/main/dataset/berry_bowl}{C4},
\href{https://github.com/google/dreambooth/tree/main/dataset/can}{C5},
\href{https://github.com/google/dreambooth/tree/main/dataset/candle}{C6},
\href{https://github.com/google/dreambooth/tree/main/dataset/cat}{C7},
\href{https://github.com/google/dreambooth/tree/main/dataset/cat2}{C8},
\href{https://github.com/google/dreambooth/tree/main/dataset/clock}{C9},
\href{https://github.com/google/dreambooth/tree/main/dataset/colorful_sneaker}{C10},
\href{https://github.com/google/dreambooth/tree/main/dataset/dog}{C11},
\href{https://github.com/google/dreambooth/tree/main/dataset/dog2}{C12},
\href{https://github.com/google/dreambooth/tree/main/dataset/dog3}{C13},
\href{https://github.com/google/dreambooth/tree/main/dataset/dog5}{C14},
\href{https://github.com/google/dreambooth/tree/main/dataset/dog6}{C15},
\href{https://github.com/google/dreambooth/tree/main/dataset/dog7}{C16},
\href{https://github.com/google/dreambooth/tree/main/dataset/dog8}{C17},
\href{https://github.com/google/dreambooth/tree/main/dataset/duck_toy}{C18},
\href{https://github.com/google/dreambooth/tree/main/dataset/fancy_boot}{C19},
\href{https://github.com/google/dreambooth/tree/main/dataset/rey_sloth_plushie}{C20},
\href{https://github.com/google/dreambooth/tree/main/dataset/monster_toy}{C21},
\href{https://github.com/google/dreambooth/tree/main/dataset/pink_sunglasses}{C22},
\href{https://github.com/google/dreambooth/tree/main/dataset/poop_emoji}{C23},
\href{https://github.com/google/dreambooth/tree/main/dataset/rc_car}{C24},
\href{https://github.com/google/dreambooth/tree/main/dataset/red_cartoon}{C25},
\href{https://github.com/google/dreambooth/tree/main/dataset/robot_toy}{C26},
\href{https://github.com/google/dreambooth/tree/main/dataset/shiny_sneaker}{C27},
\href{https://github.com/google/dreambooth/tree/main/dataset/teapot}{C28},
\href{https://github.com/google/dreambooth/tree/main/dataset/vase}{C29},
\href{https://github.com/google/dreambooth/tree/main/dataset/wolf_plushie}{C30}.

\noindent \textbf{Dataset Partitioning}
\label{sec:dataset_partitioning}
There are 30 subjects and 26 styles overall. We split the subjects and styles randomly, but with the constraint that there is a good representation of different subjects and styles in each split as some subjects and styles are similar to each other. For example we aimed at avoiding only testing on different dogs or only on painting styles.

We split the subjects and styles into training, validation and test splits as shown in \cref{tab:dataset_split}. In \cref{tab:test_dataset} and \cref{tab:val_dataset} we show images taken from the test and validation sets respectively (used to train the test and validation LoRAs).

\section{Additional Results}
\label{sec:supp:results}

\subsection{Performance via Standard Metrics}
Standard metrics evaluations (DINO, CLIP-I, CLIP-T) are reported in Table~\ref{tab:metrics_main}. We include this analysis for informational purposes only.
As explained in Sec.~4 of the main paper,
 these metrics are not optimal for the joint subject-style personalization task. 
Specifically, DINO (CLIP-I) is maximized when the subject (style) reference images are copied without meaningful integration, so more attention should be given to MLLM and human evaluation results.
\begin{table}[!h]
    \centering
        \begin{tabular}{l c c c}
            \toprule
            &CLIP-I & DINO & CLIP-T\\
            \toprule
                 Joint Training \cite{ruiz2023dreambooth}  &0.623&0.764&    0.329\\
                 Direct Merge \cite{wortsman2022model} &0.657&0.747&  0.305\\  
                DARE \cite{yu2024language} &0.630&0.576&  0.360\\  
                TIES \cite{yadav2024ties} &0.620&0.592&  0.358\\  
                
                DARE-TIES \cite{goddard2024arcee} &0.618&0.559&  0.355\\  \ziplora \cite{shah2024ziplora} &0.643&0.741&  0.334\\ \textbf{\ours (ours)} &0.656&0.643& 0.344\\ 
             \bottomrule
        \end{tabular}
        \caption{\textbf{Standard Metrics.} \ours attains similar results, but these metrics are inadequate for
    joint subject-style changes.}
    \label{tab:metrics_main}
\end{table}

\begin{figure}[!h]
    \centering
    \includegraphics[width=0.99\columnwidth]{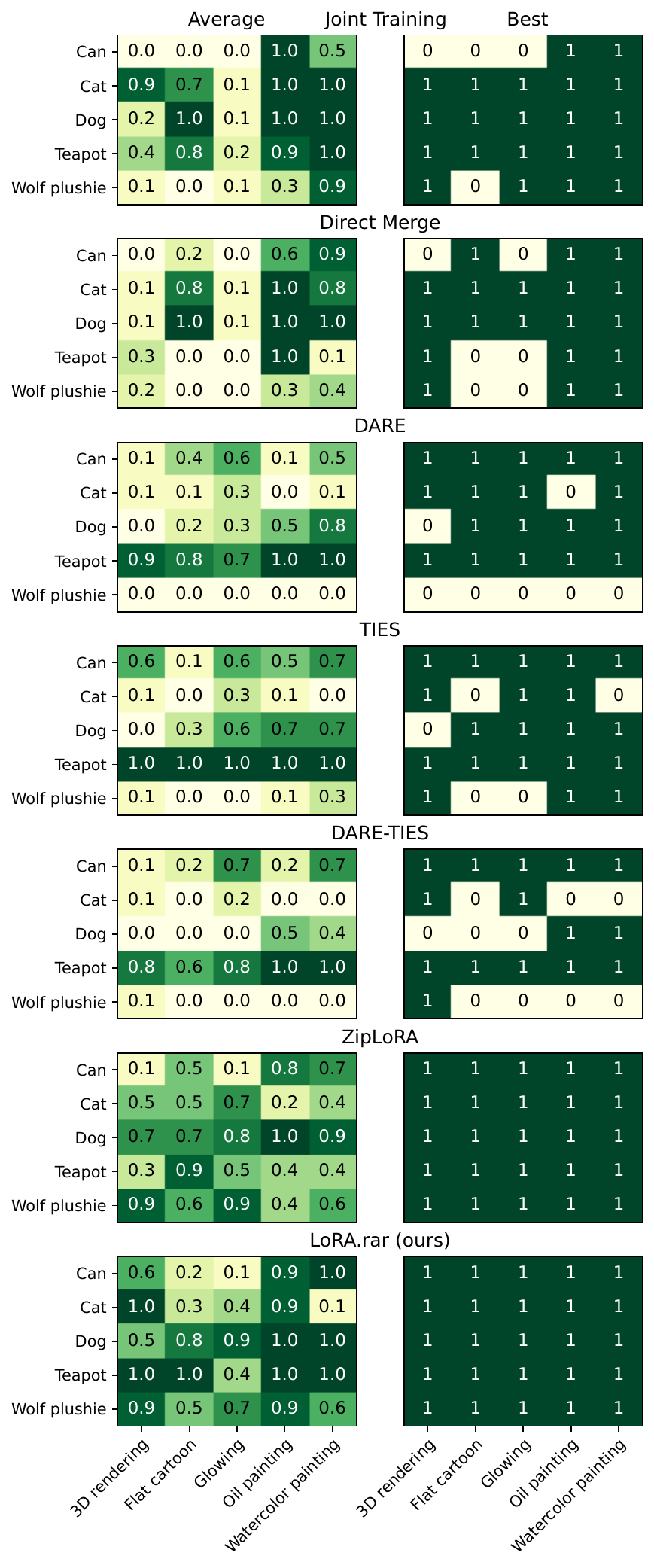}
    \vspace{-0.2cm}
    \caption{\textbf{MLLM Evaluation per Test Subject and Style.} Ratio of generated images with the correct content and style according to our metric \ourmetric. Our solution leads to better images compared to  existing approaches.}
    \vspace{-0.2cm}
        \label{fig:supp:mllm_results_per_subject_per_style}
\end{figure}

\subsection{MLLM Results per Subject and Style}
We provide results of MLLM evaluation for each test subject and style in Fig.~\ref{fig:supp:mllm_results_per_subject_per_style}. We report the results for both the average case as well as the best case. The results indicate that there are certain subjects and styles that are more challenging than others, for example the \textit{can} subject or the \textit{glowing} style. We also see that \ourmethod~and ZipLoRA are in general significantly more successful than the other approaches, and they can be successful also in cases where other approaches typically fail, for example in the case of the \textit{wolf plushie} subject.

\subsection{Ablation Study on Hypernetwork}
We conducted an ablation study on the hypernetwork design by exhaustively exploring all possible configurations to determine which components should have their merging coefficients predicted by the hypernetwork
We used the validation set and MLLM judge for this investigation, and we report the results in Table~\ref{tab:hypernet_ablation}. We observe that the best results are obtained by \textit{Query, Output} case that we have used; however, a few other combinations also achieve good results such as \textit{Query, Key, Output}; \textit{Query, Value} and \textit{Value}.

\begin{table}[t]
    \centering
        \begin{tabular}{l c c}
            \toprule
           & \textbf{Average case} & \textbf{Best case} \\
            \toprule
                Key & 0.28 & 0.75 \\
                Value & 0.43 & 0.83 \\
                Query & 0.28 & 0.75 \\
                Output & 0.39 & 0.83  \\
                \midrule
                Key, Value & 0.40 & 0.92 \\
                Key, Query & 0.31 & 0.75  \\
                Key, Output & 0.44 & 0.75\\
                Query, Value & 0.42 & 0.83 \\
                \textbf{Query, Output} & 0.48 & 0.92 \\
                Value, Output & 0.29 & 0.58 \\
                \midrule
                Query, Key, Value &  0.41 & 0.83 \\
                Query, Value, Output & 0.23 & 0.33 \\
                Query, Key, Output & 0.49 & 0.83 \\
                Key, Value, Output &  0.29 & 0.50 \\
                \midrule
                Query, Key, Value, Output & 0.23 & 0.50 \\    
             \bottomrule
        \end{tabular}
    \caption{\textbf{Ablation Study via MLLM Evaluation.} Ratio of generated images with the correct content and style on the combinations of validation subjects and styles according to our \ourmetric metric.}
    \label{tab:hypernet_ablation}
\end{table}

\begin{figure*}[!t]
    \centering
    \resizebox{0.46\textwidth}{!}{%
    \begin{tabular}[c]{L L L L L L L}
     \small A \textit{[C] cat} in \textit{[S]} style & 3D Rendering & Oil Painting & Watercolor Painting & Flat Cartoon Illustration & Glowing \\
        \tabfigure{width=0.12\textwidth}{figures/dataset/test/cat2/03.jpg}  & \tabfigure{width=0.12\textwidth}{figures/qualitative/3d_rendering.jpg} 
        & \tabfigure{width=0.12\textwidth}{figures/qualitative/oil.jpg}
        & \tabfigure{width=0.12\textwidth}{figures/qualitative/watercolor.jpg}
        & \tabfigure{width=0.12\textwidth}{figures/qualitative/flat.jpg}
        & \tabfigure{width=0.12\textwidth}{figures/qualitative/glowing.jpg} \\
    \begin{minipage}[c][1.7cm][c]{\linewidth} %
        \centering \ziplora
    \end{minipage}
    &
        \tabfigure{width=0.12\textwidth}{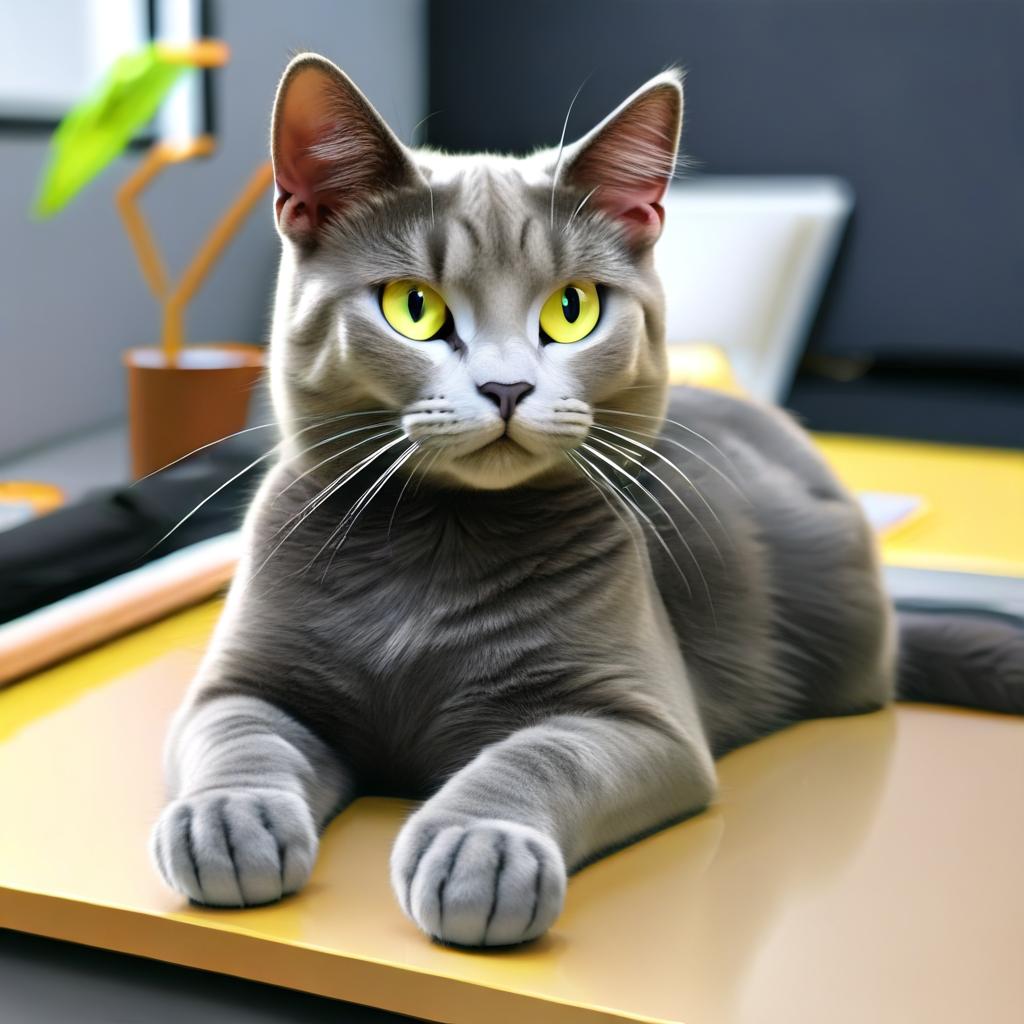} &
        \tabfigure{width=0.12\textwidth}{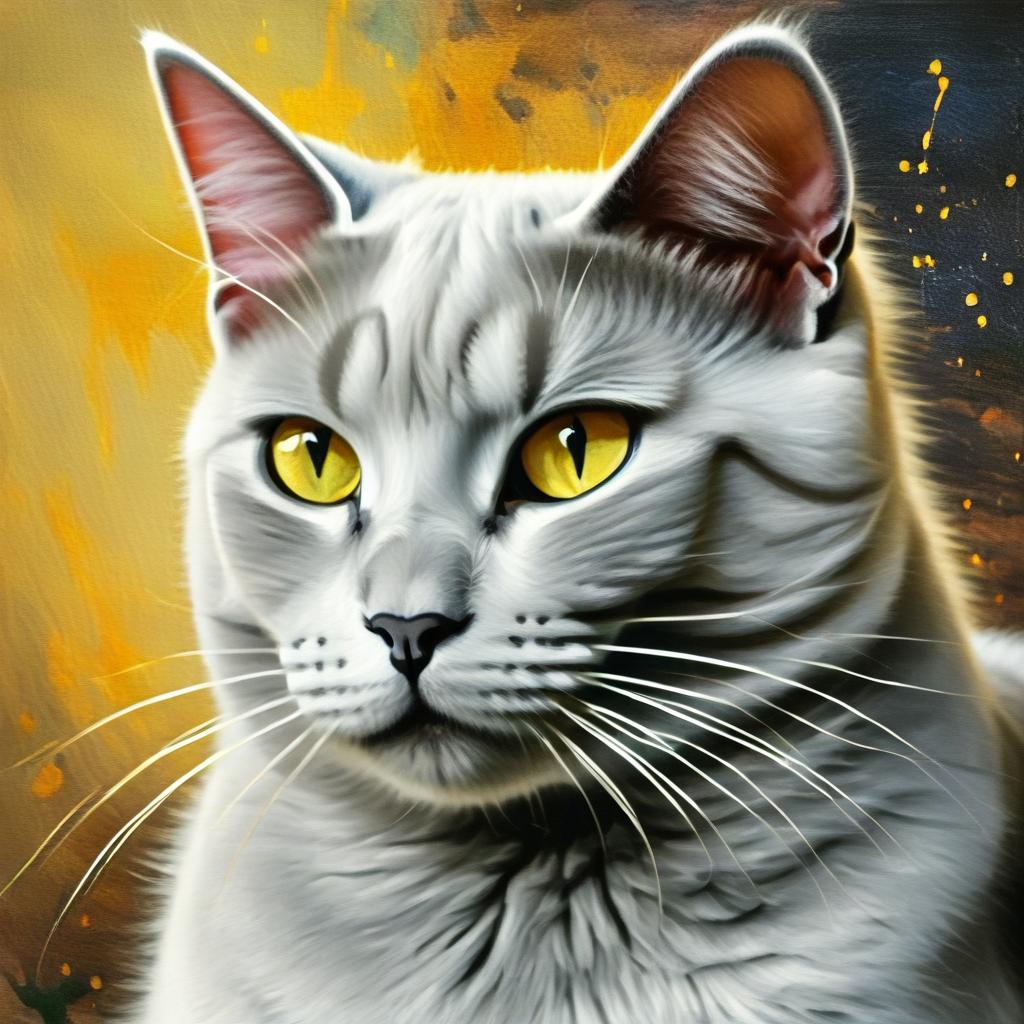} &
        \tabfigure{width=0.12\textwidth}{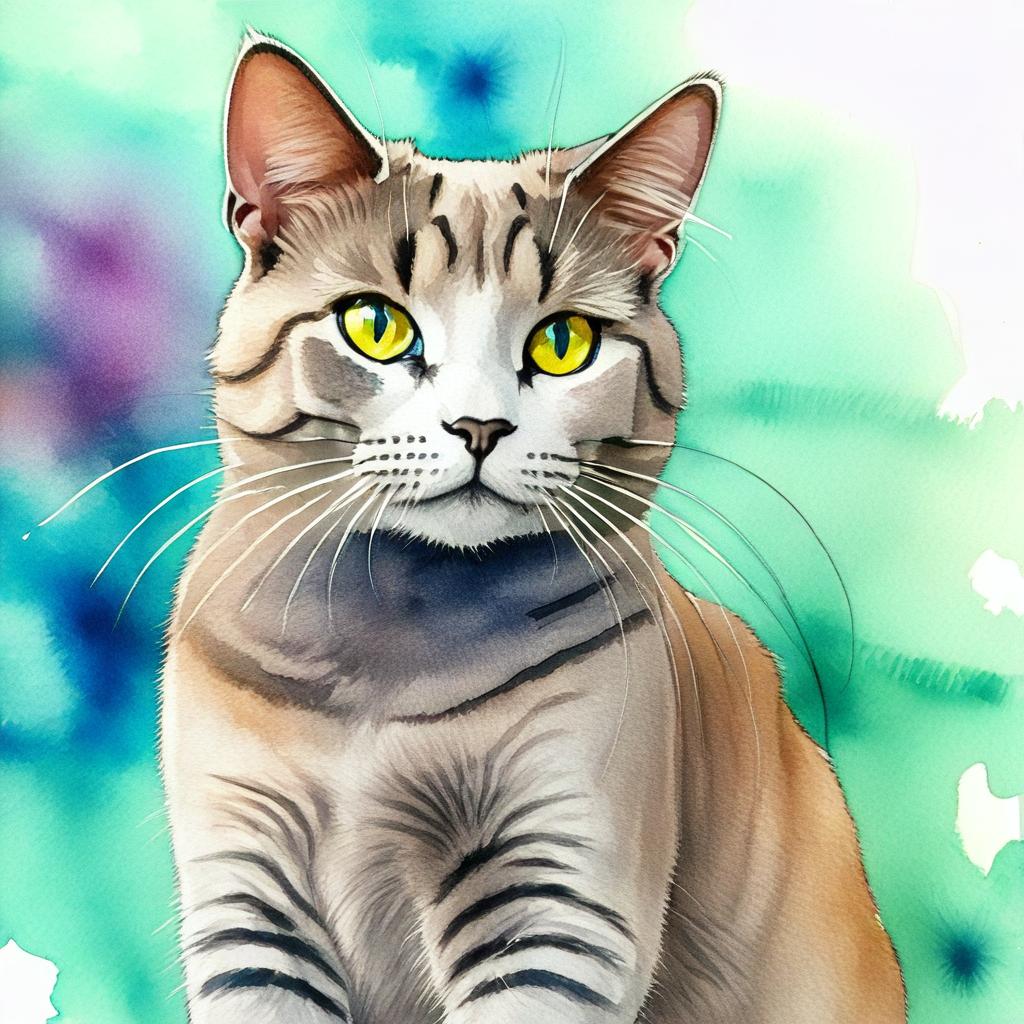} &
        \tabfigure{width=0.12\textwidth}{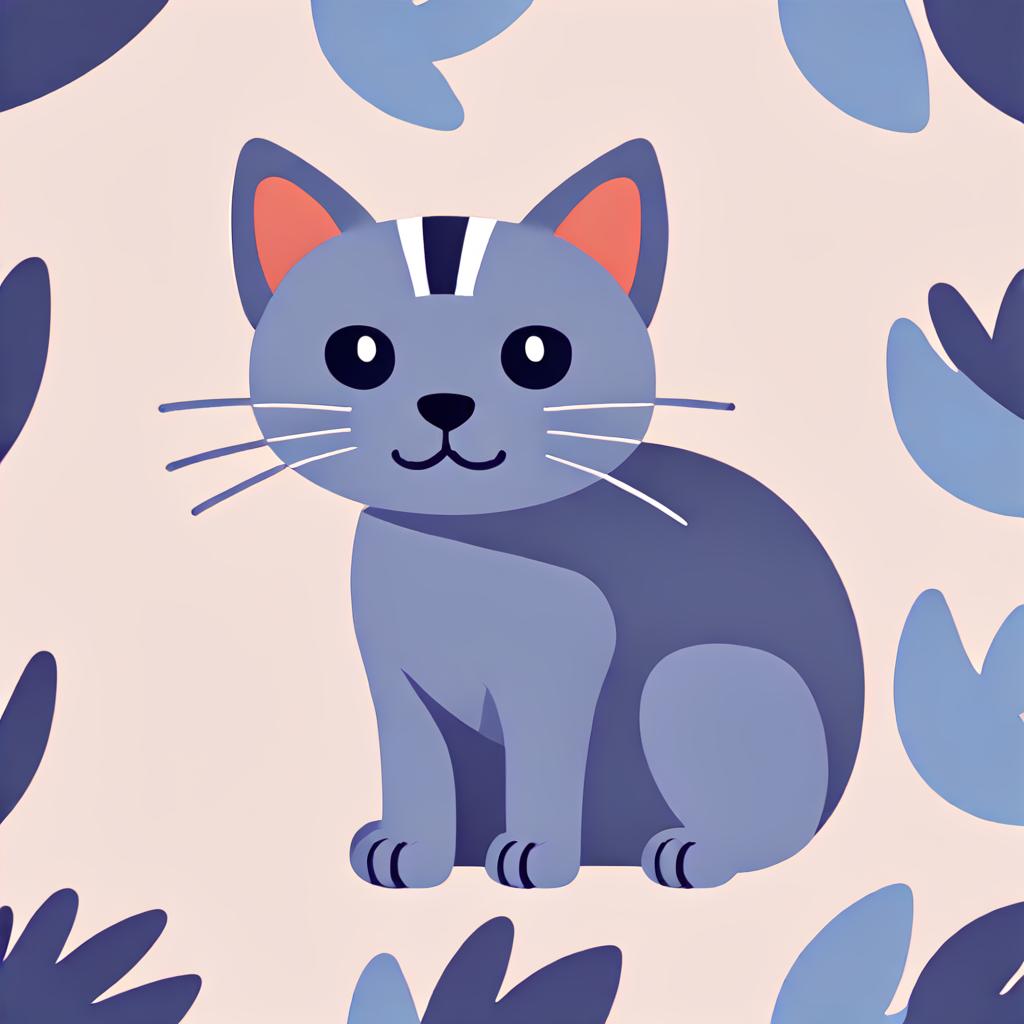} &
        \tabfigure{width=0.12\textwidth}{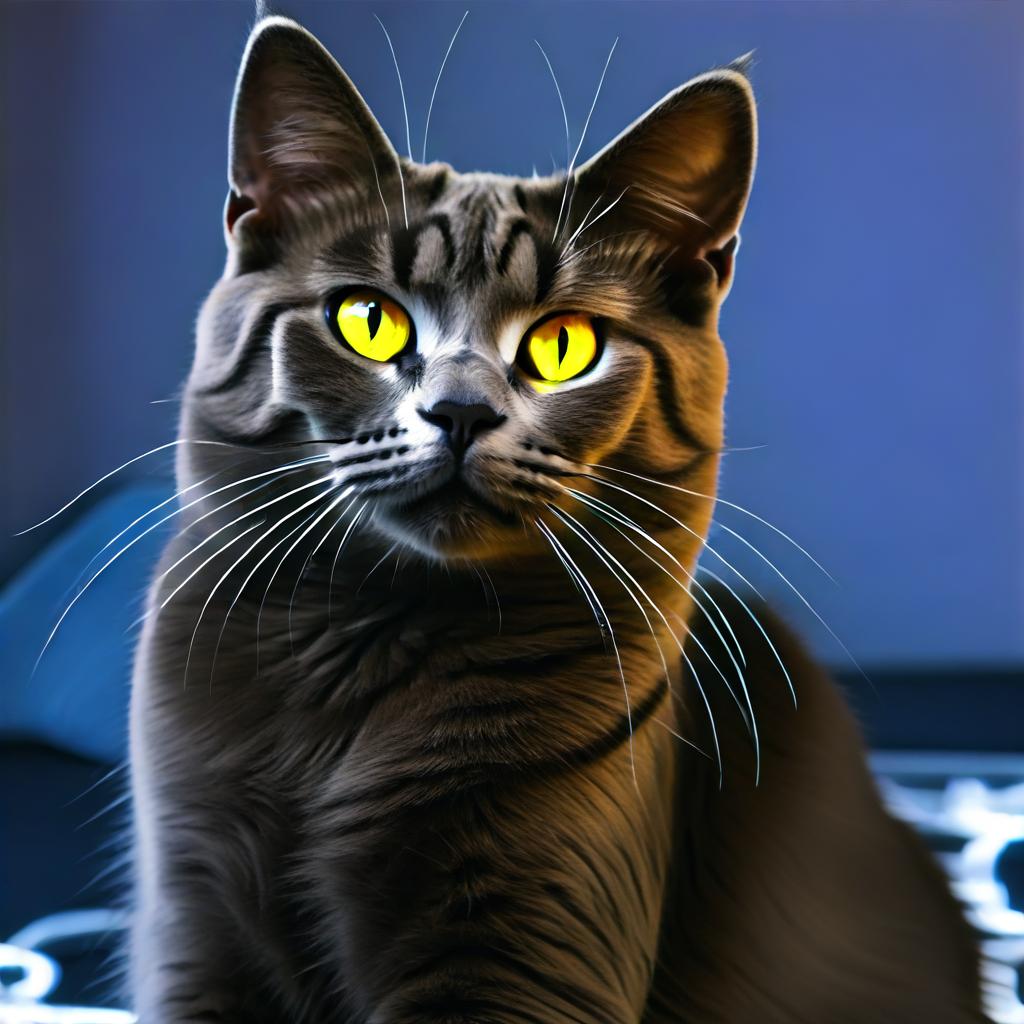} & \\
    \begin{minipage}[c][1.7cm][c]{\linewidth} %
        \centering \textbf{\ours (ours)}
    \end{minipage}
        &
        \tabfigure{width=0.12\textwidth}{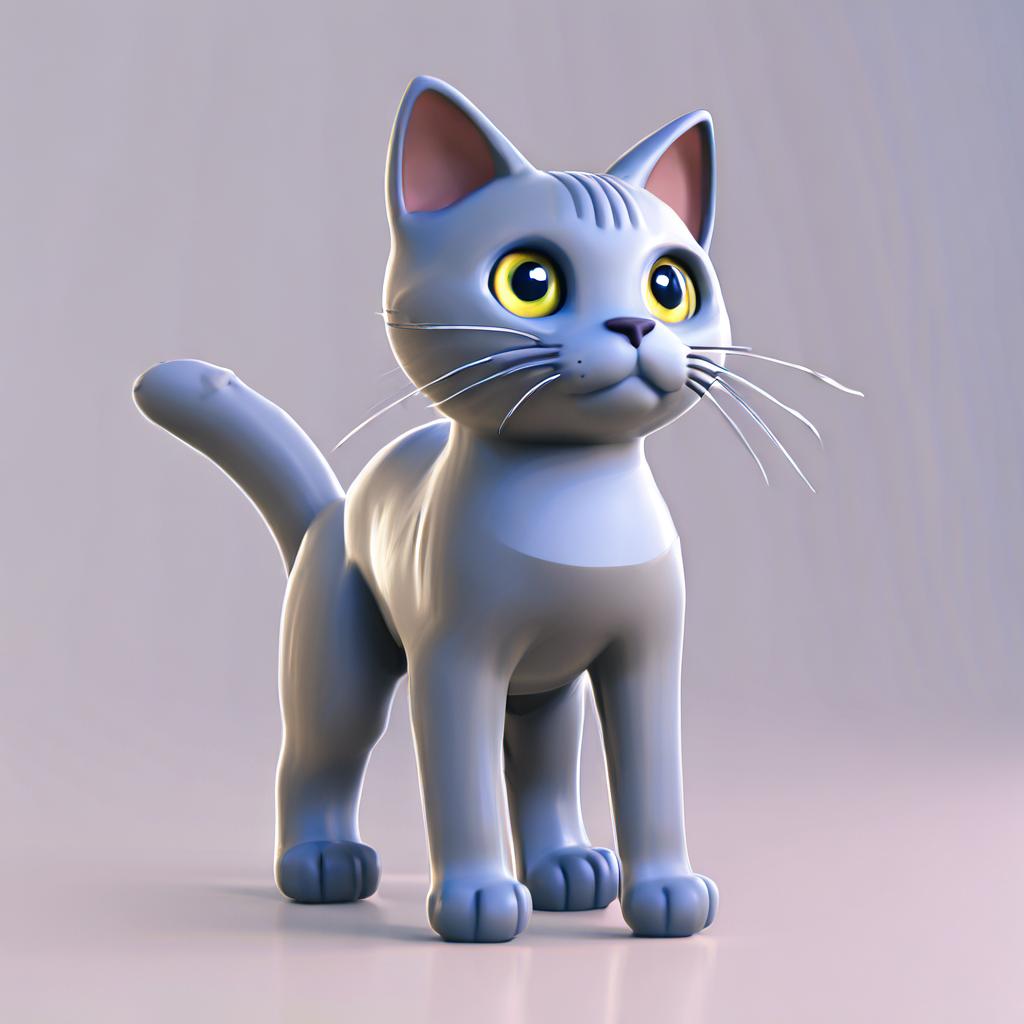} &
        \tabfigure{width=0.12\textwidth}{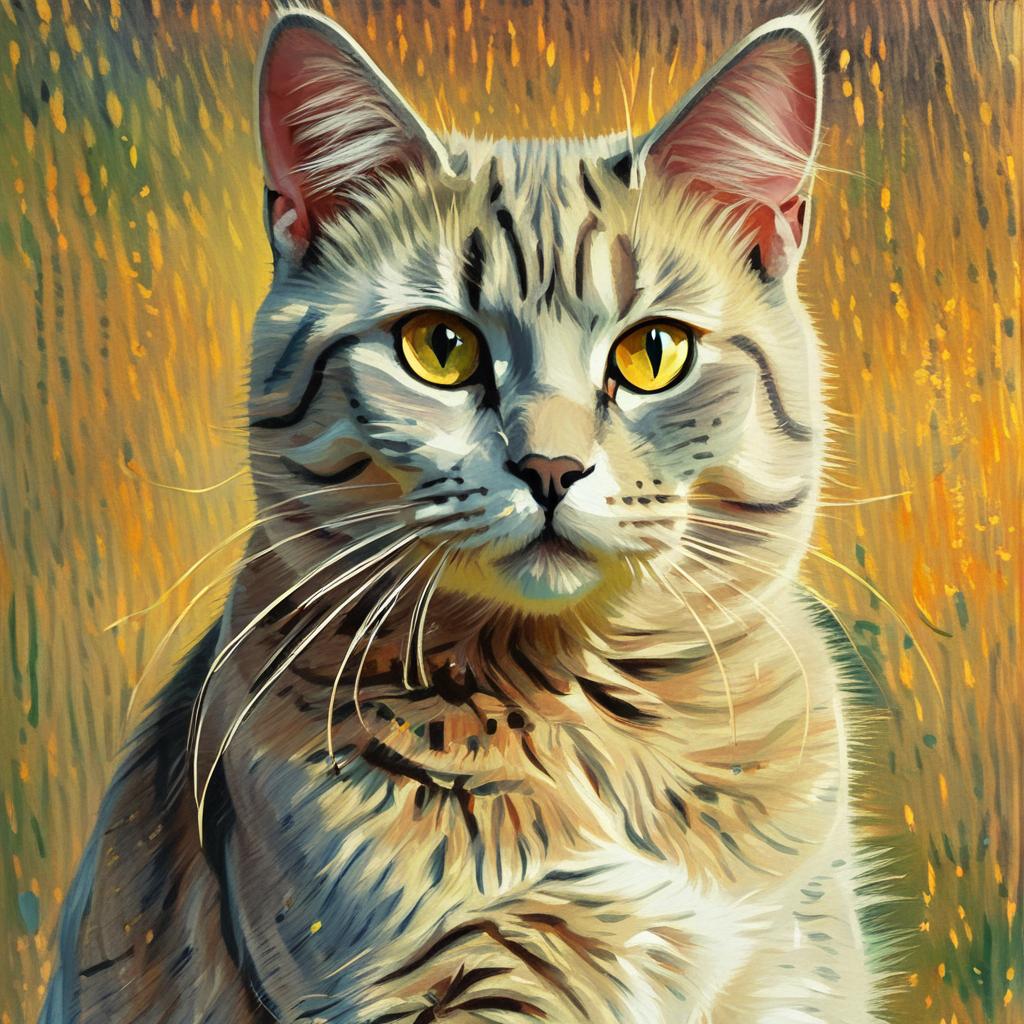} &
        \tabfigure{width=0.12\textwidth}{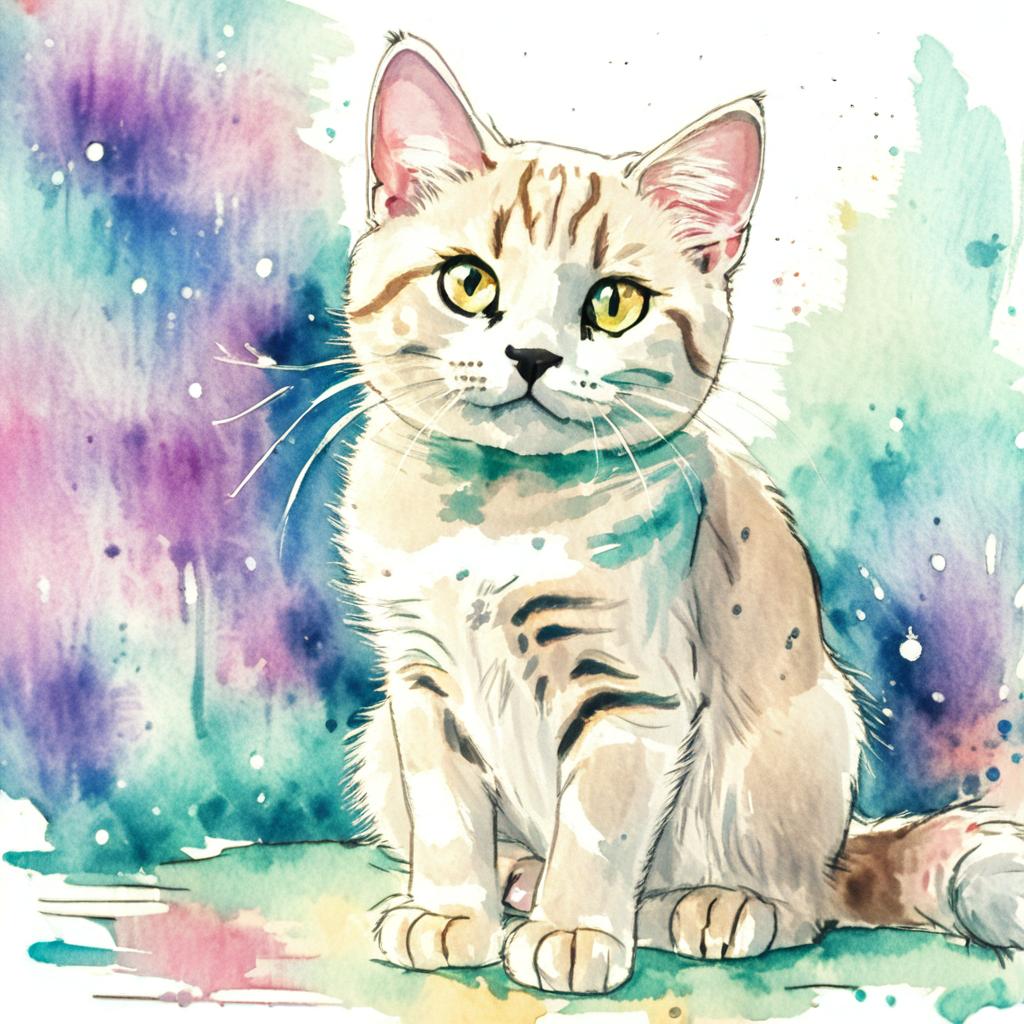} &
        \tabfigure{width=0.12\textwidth}{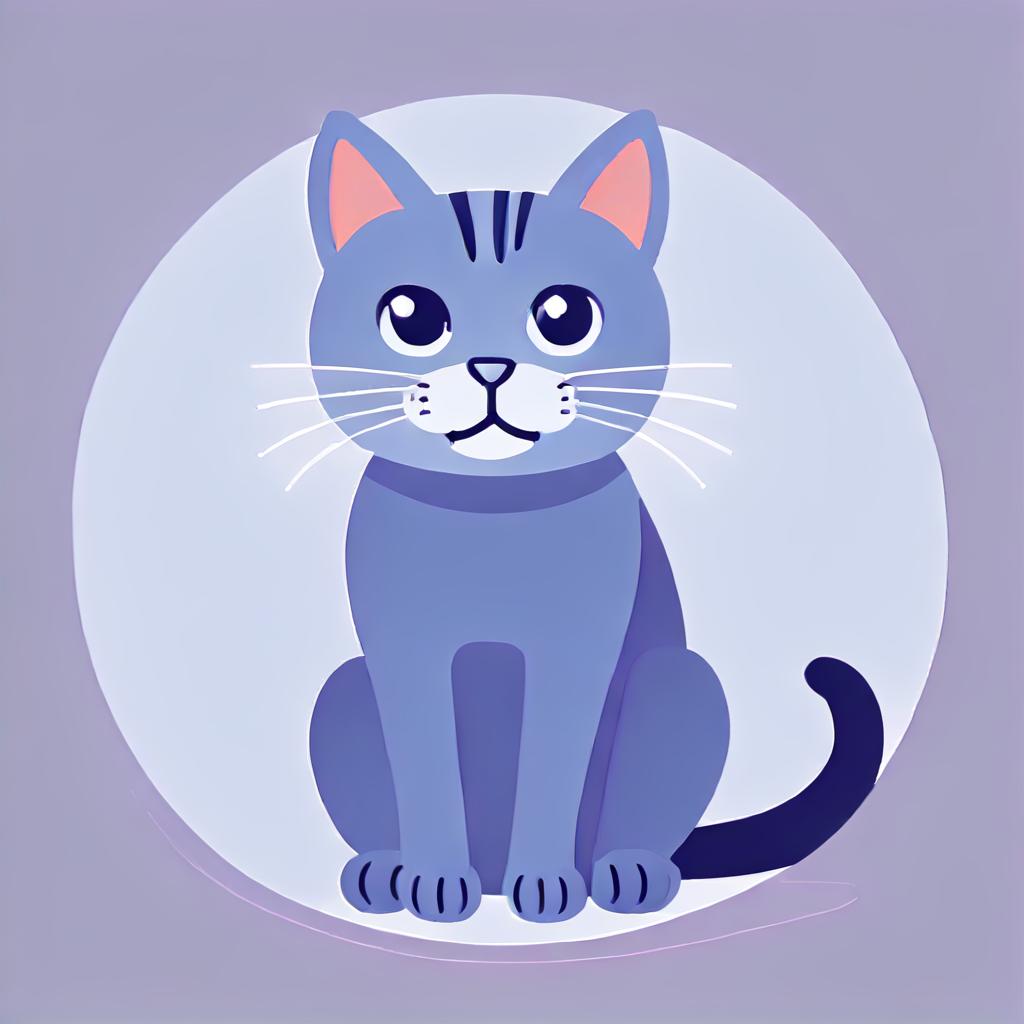} &
        \tabfigure{width=0.12\textwidth}{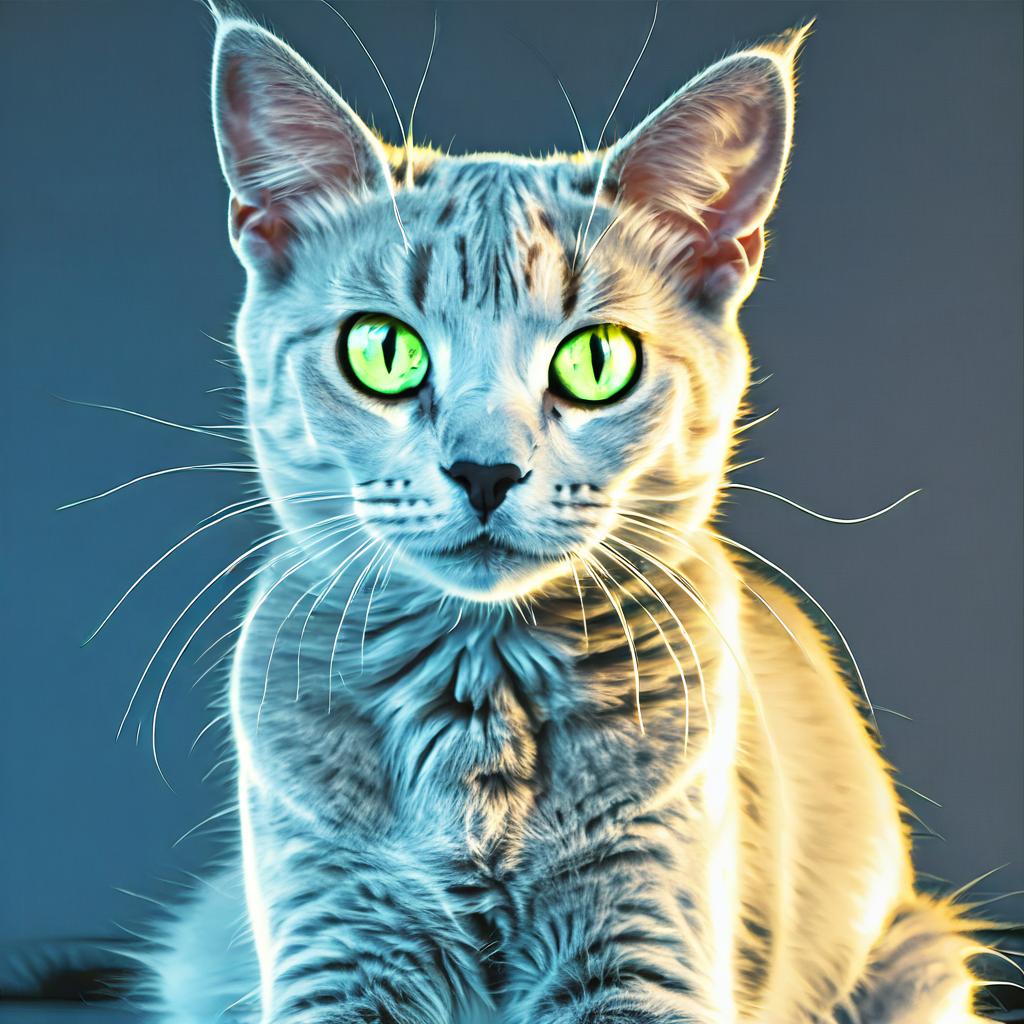} & \\
    \end{tabular}
    }
    \resizebox{0.46\textwidth}{!}{%
    \begin{tabular}[c]{L L L L L L L}
     \small A \textit{[C] dog} in \textit{[S]} style & 3D Rendering & Oil Painting & Watercolor Painting & Flat Cartoon Illustration & Glowing \\
        \tabfigure{width=0.12\textwidth}{figures/dataset/test/dog8/04.jpg}  & \tabfigure{width=0.12\textwidth}{figures/qualitative/3d_rendering.jpg} 
        & \tabfigure{width=0.12\textwidth}{figures/qualitative/oil.jpg}
        & \tabfigure{width=0.12\textwidth}{figures/qualitative/watercolor.jpg}
        & \tabfigure{width=0.12\textwidth}{figures/qualitative/flat.jpg}
        & \tabfigure{width=0.12\textwidth}{figures/qualitative/glowing.jpg} \\
    \begin{minipage}[c][1.7cm][c]{\linewidth} %
        \centering \ziplora
    \end{minipage}
    &
        \tabfigure{width=0.12\textwidth}{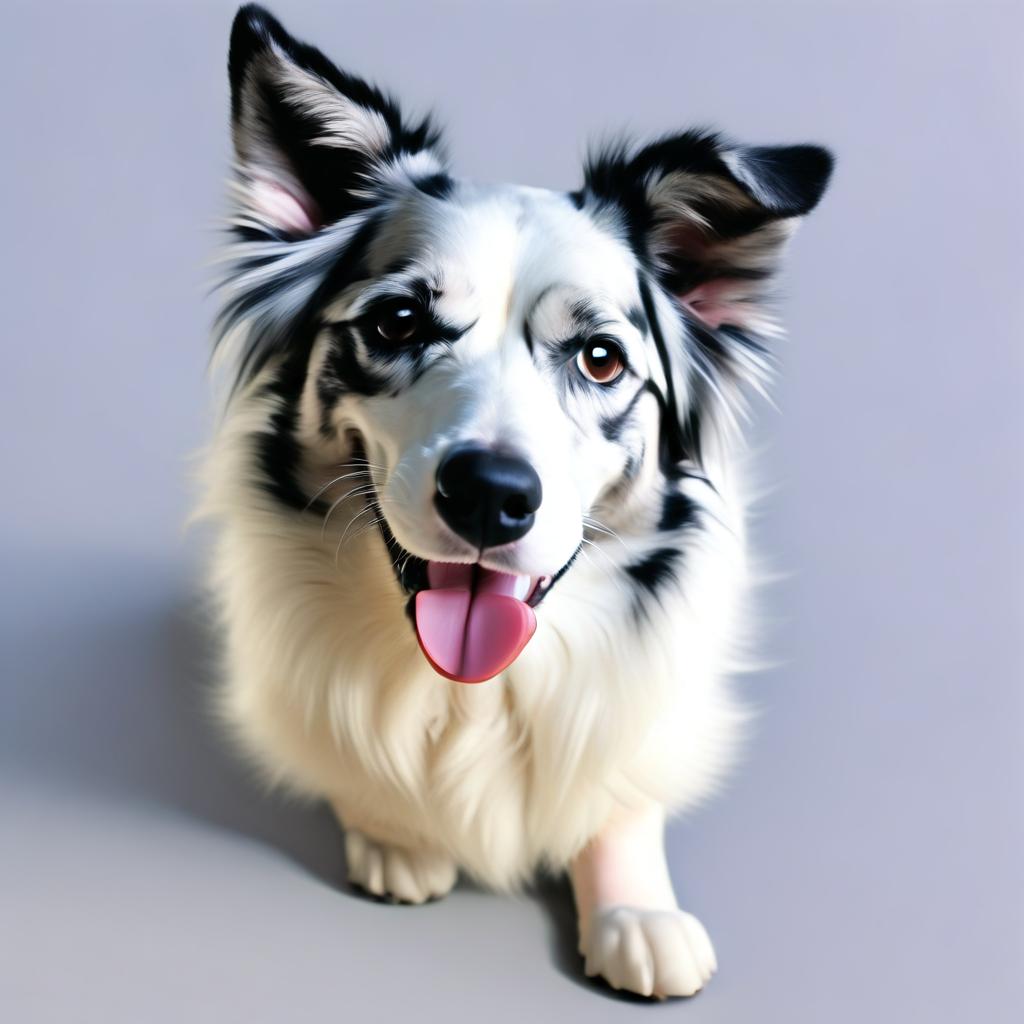} &
        \tabfigure{width=0.12\textwidth}{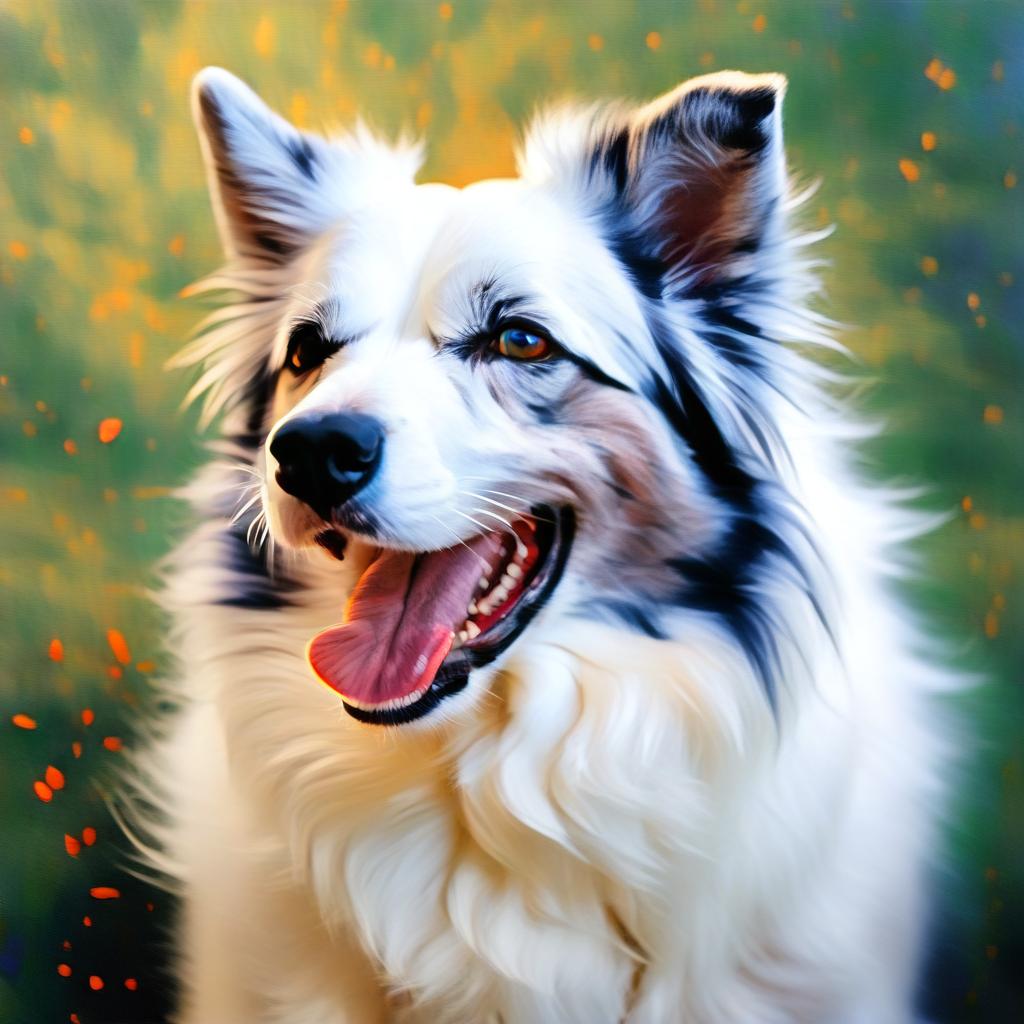} &
        \tabfigure{width=0.12\textwidth}{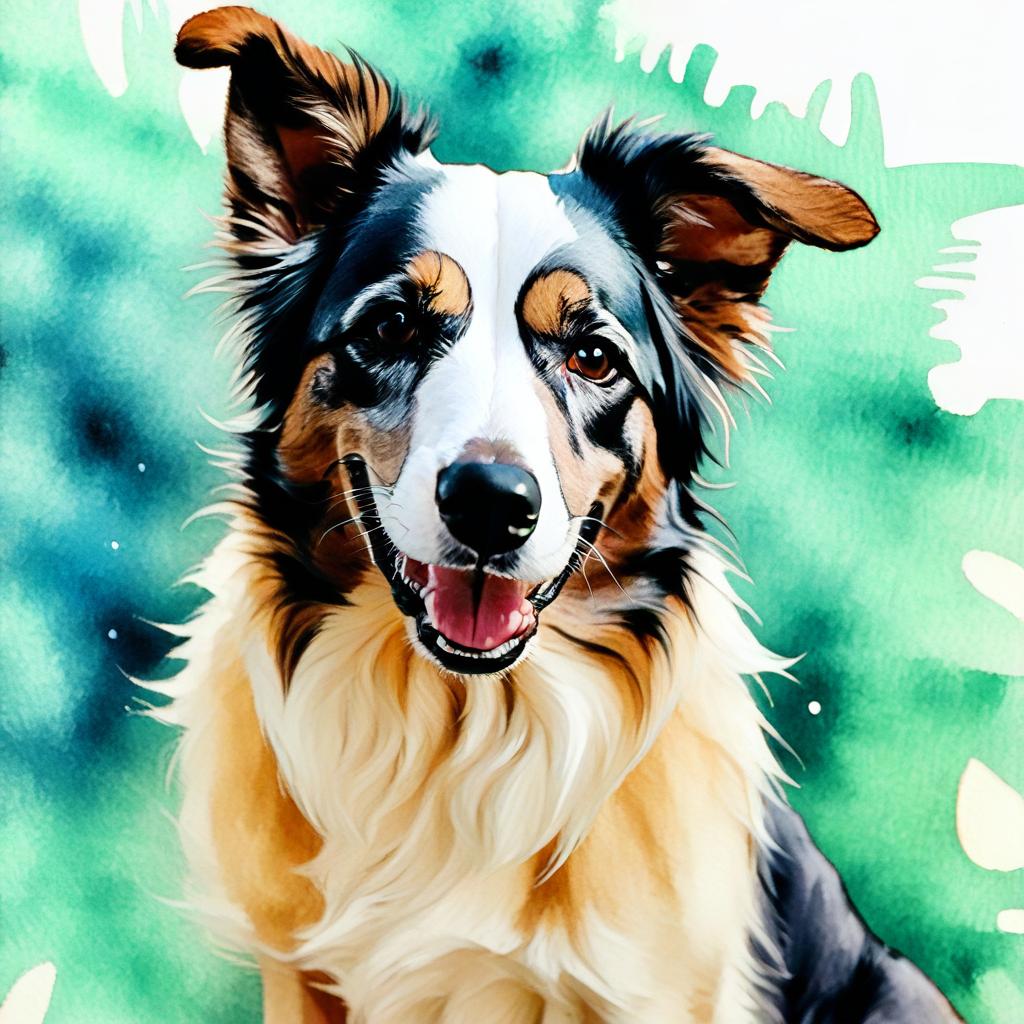} &
        \tabfigure{width=0.12\textwidth}{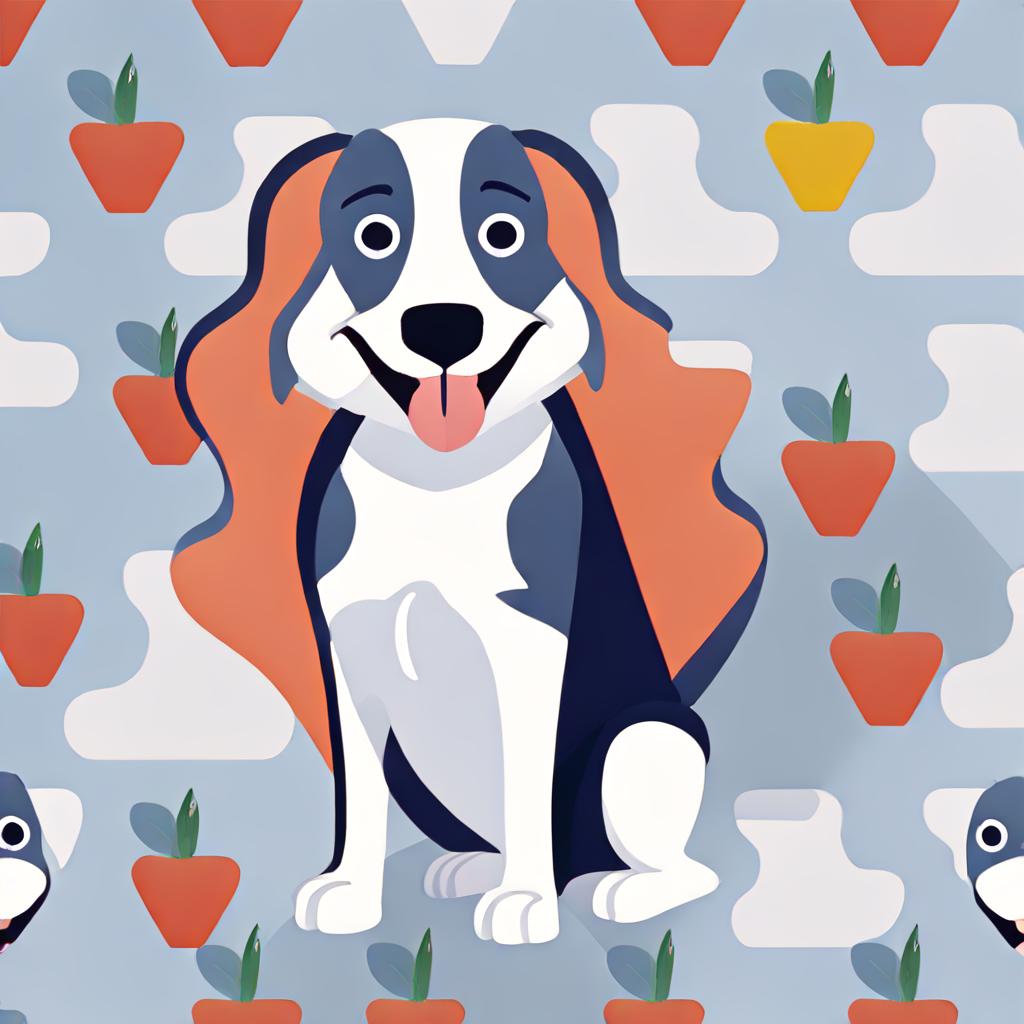} &
        \tabfigure{width=0.12\textwidth}{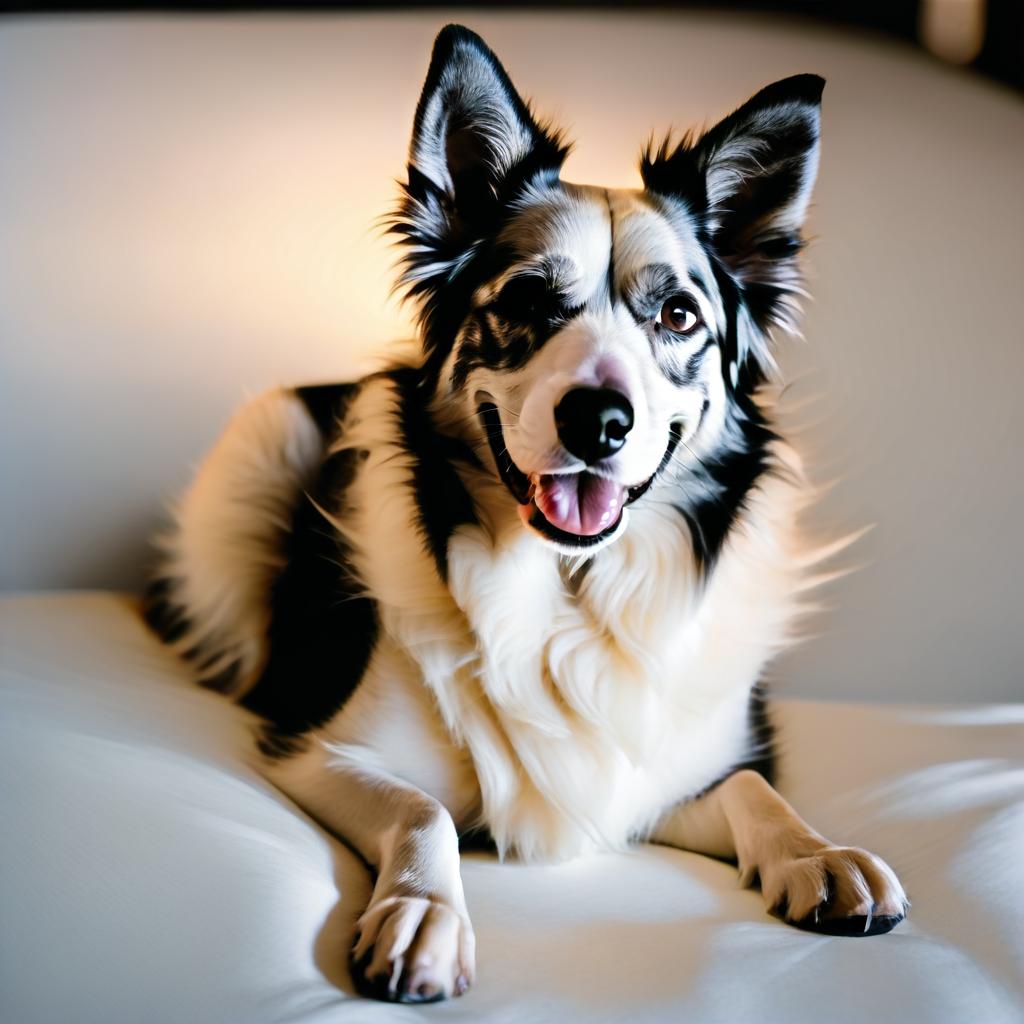} & \\
    \begin{minipage}[c][1.7cm][c]{\linewidth} %
        \centering \textbf{\ours (ours)}
    \end{minipage}
        &
        \tabfigure{width=0.12\textwidth}{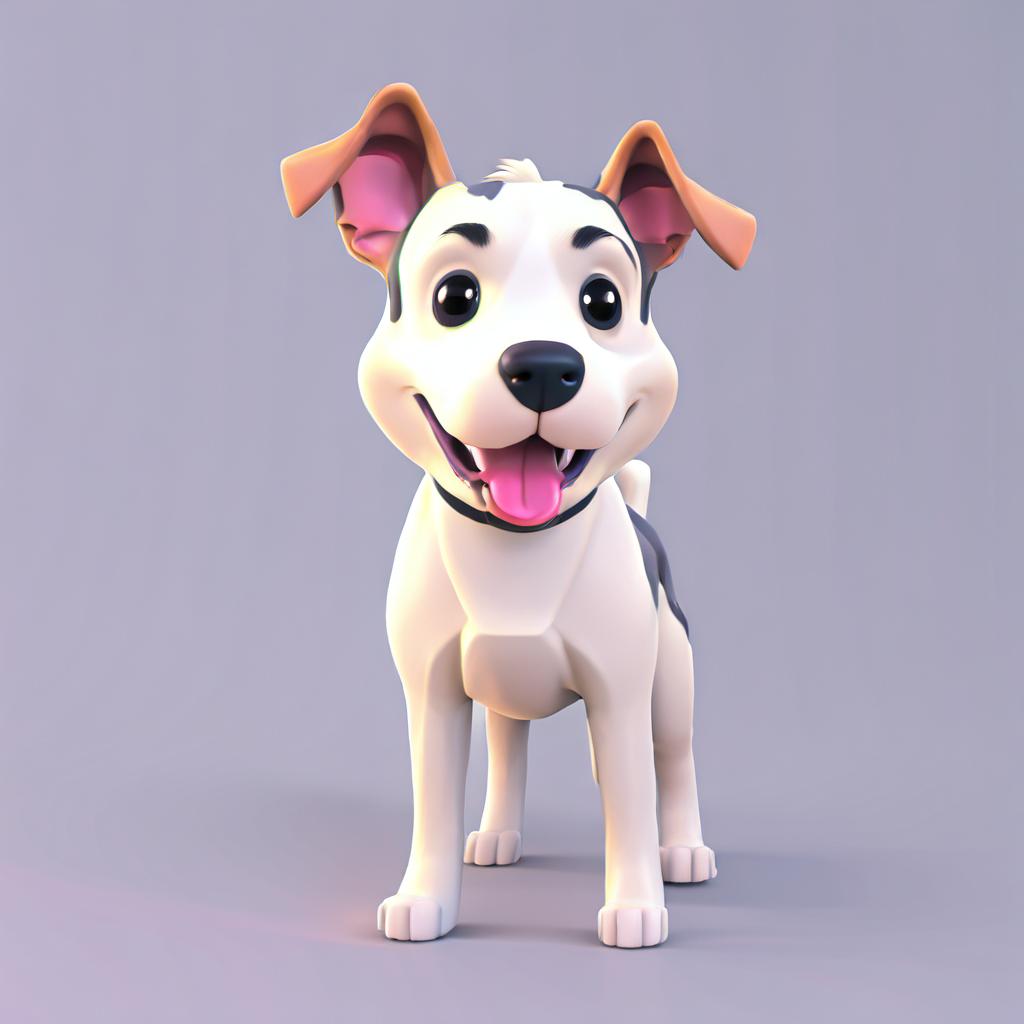} &
        \tabfigure{width=0.12\textwidth}{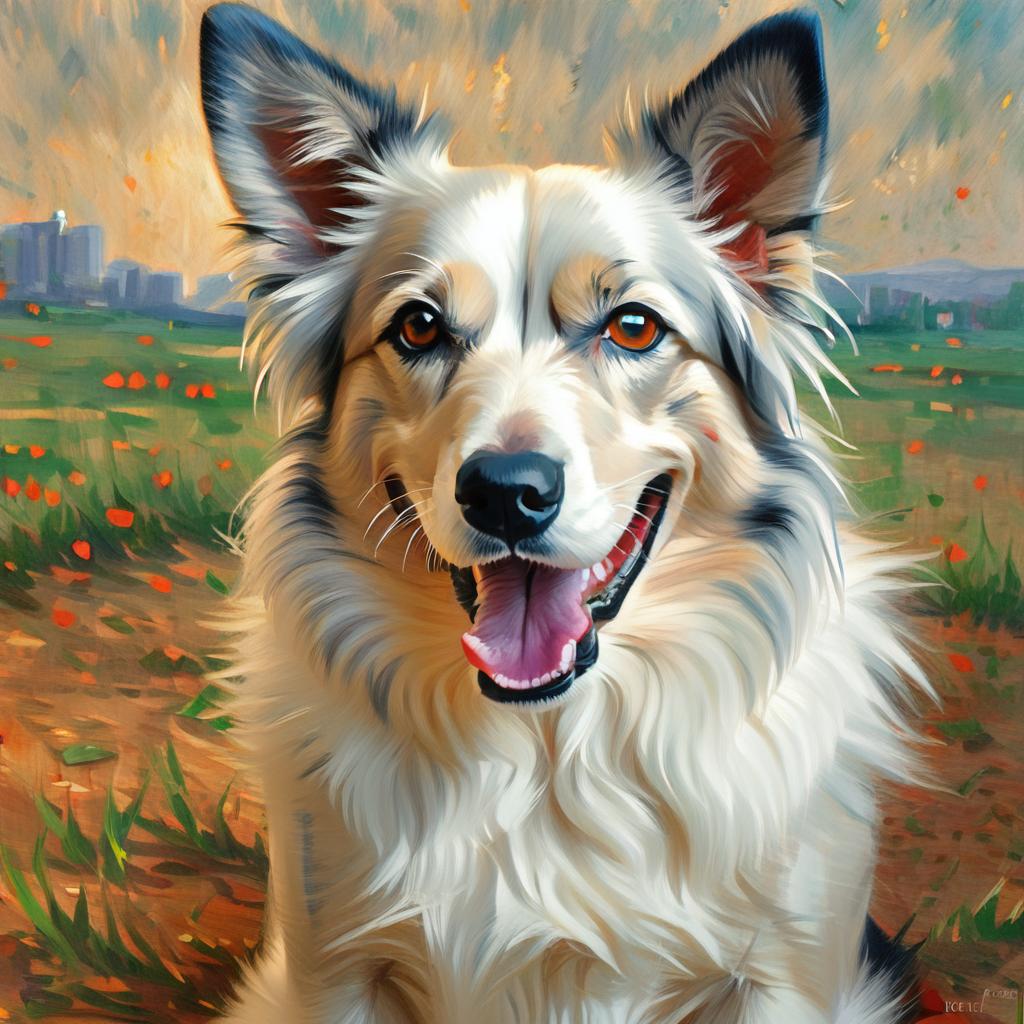} &
        \tabfigure{width=0.12\textwidth}{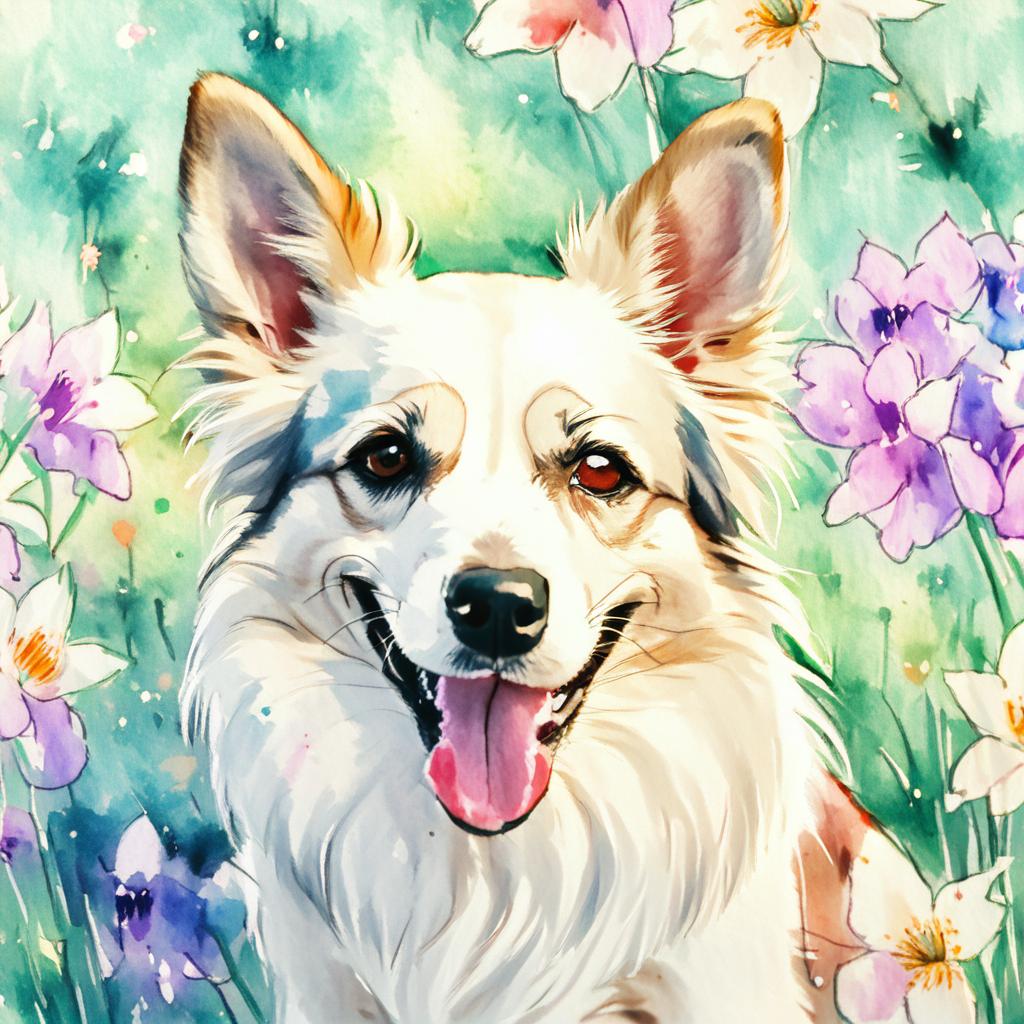} &
        \tabfigure{width=0.12\textwidth}{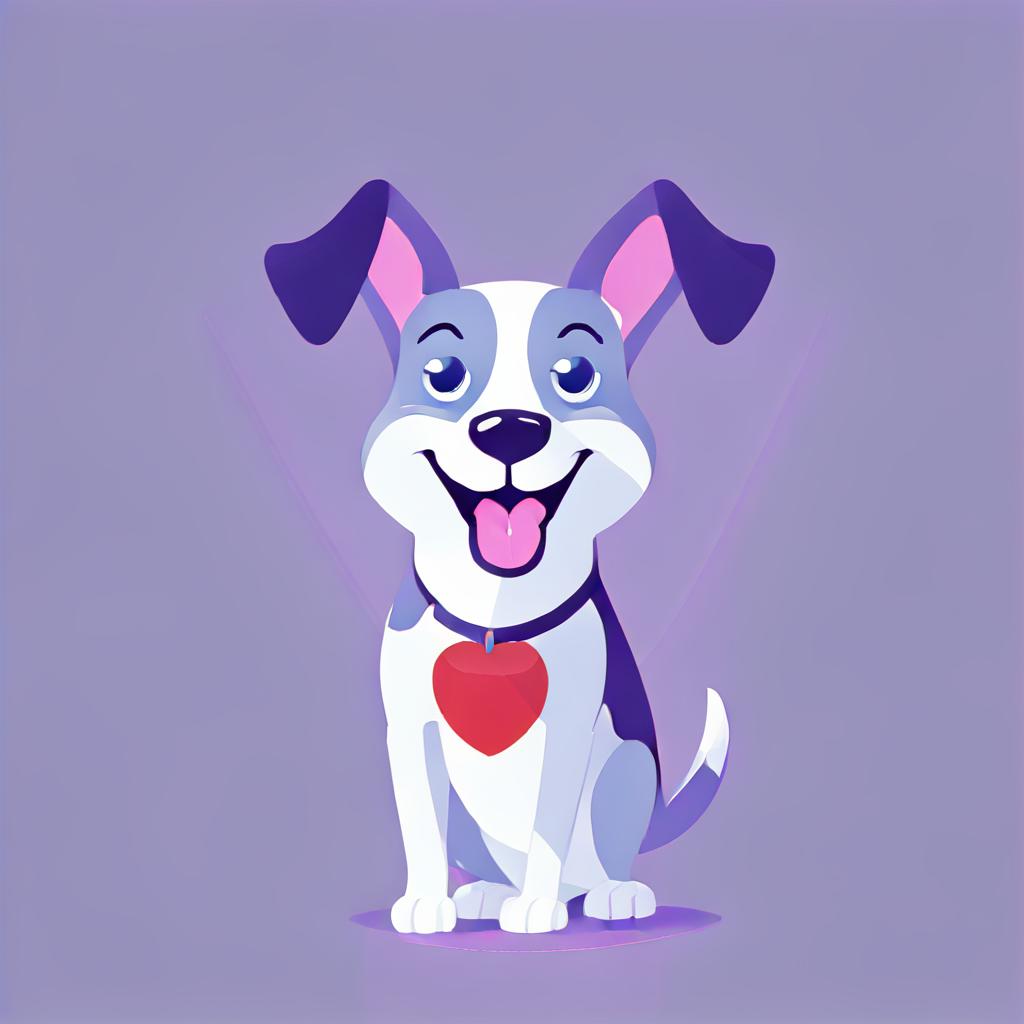} &
        \tabfigure{width=0.12\textwidth}{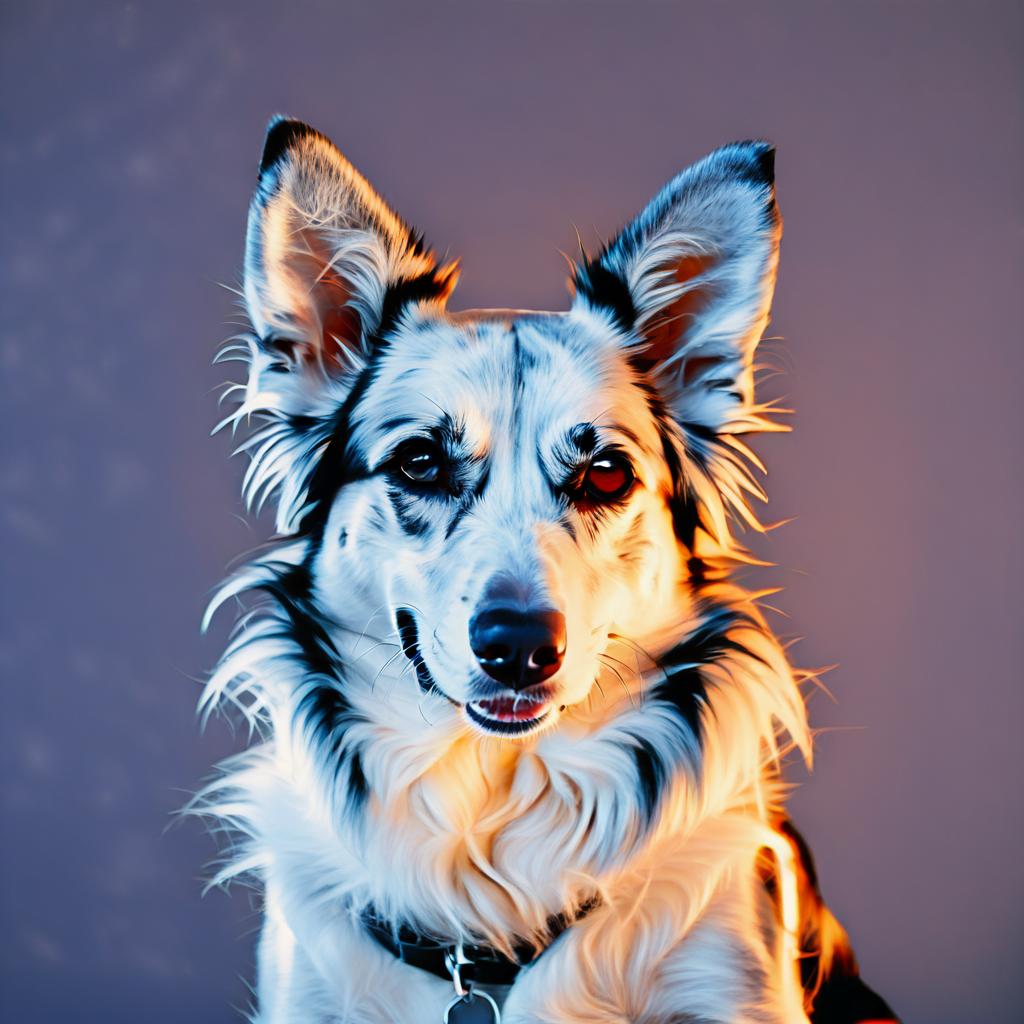} & \\
    \end{tabular}
    }
    \caption{\textbf{Qualitative Comparison on Koala-700m.} \ours generates better images than ZipLoRA.}
    \label{fig:qualitative_comparison_koala700m}
\end{figure*}

\subsection{Results on Lightweight Diffusion Model}
Fig.~\ref{fig:qualitative_comparison_koala700m} shows qualitative results produced with KOALA 700m \cite{lee2024koala}, a lightweight diffusion model, further showing that \ours could be applied to other diffusion model backbones and still outperform ZipLoRA.

\subsection{Generalization to New Concepts}
\begin{figure}[!h]
    \centering
    \resizebox{1\linewidth}{!}{%
    \begin{tabular}[c]{L L L L L L L}
\raisebox{-7mm}{\begin{tikzpicture}
    \draw[->] (0, -1.5) -- (1.5, -1.5) node[below, midway] {Styles};
    \draw[->] (0, -1.5) -- (0, -3) node[right, midway] {Contents};
\end{tikzpicture}}
     &
    \tabfigure{width=0.1\textwidth}{figures/qualitative/3d_rendering.jpg} &
    \tabfigure{width=0.1\textwidth}{figures/qualitative/oil.jpg} &
    \tabfigure{width=0.1\textwidth}{figures/qualitative/watercolor.jpg} & \tabfigure{width=0.1\textwidth}{figures/qualitative/flat.jpg}  & \tabfigure{width=0.1\textwidth}{figures/qualitative/glowing.jpg} &
    \tabfigure{width=0.1\textwidth}{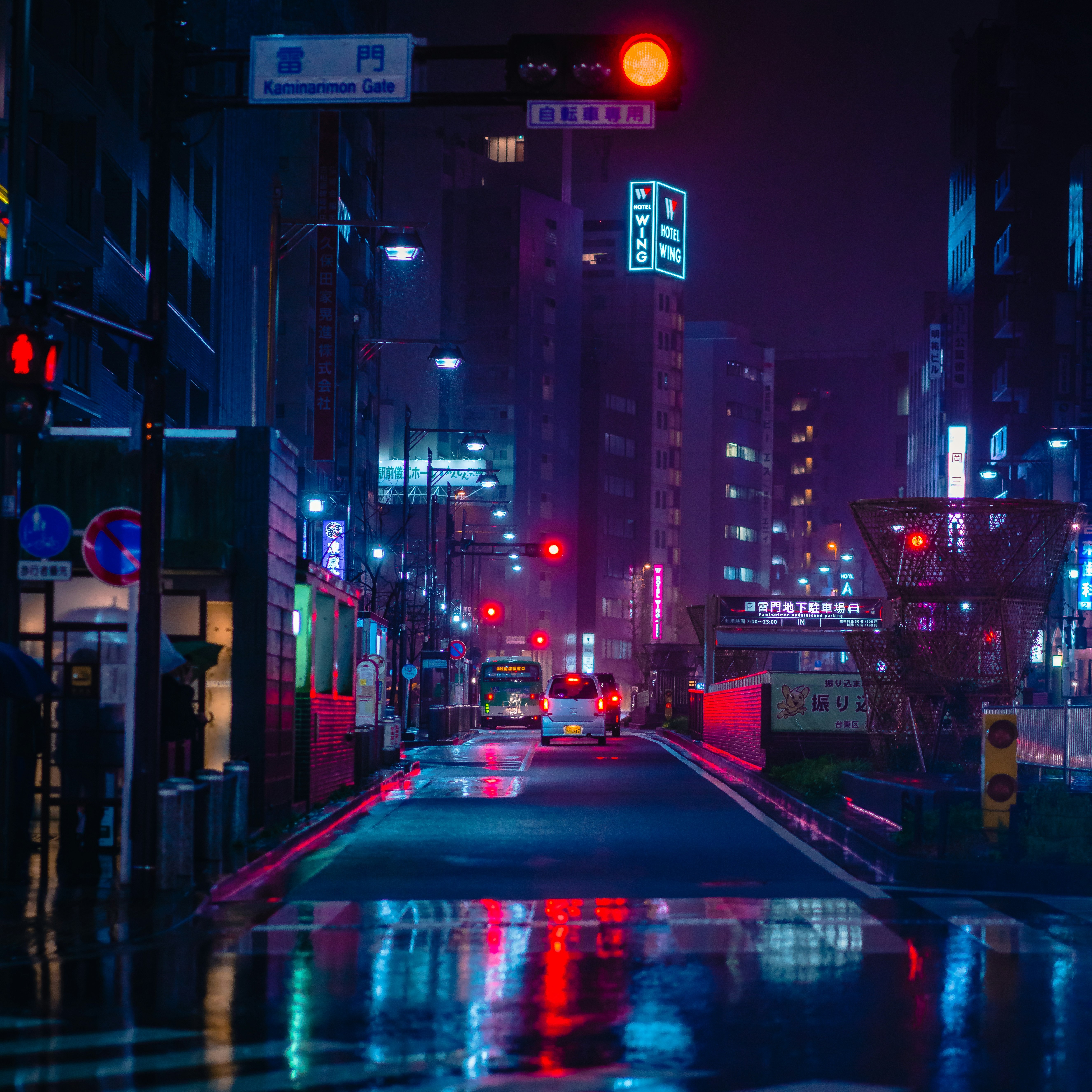} \\ %
    
    \tabfigure{width=0.1\textwidth}{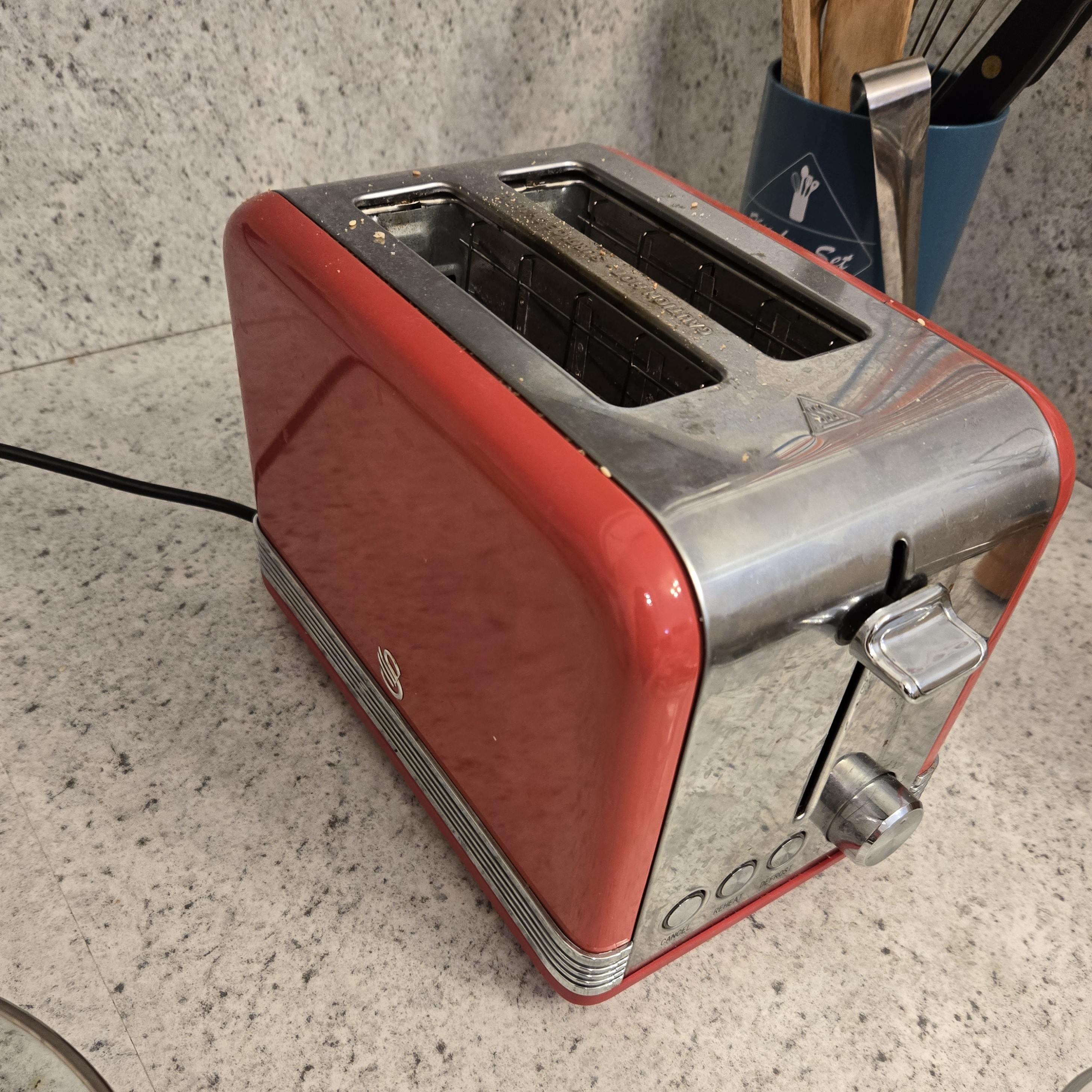} &
    \tabfigure{width=0.1\textwidth}{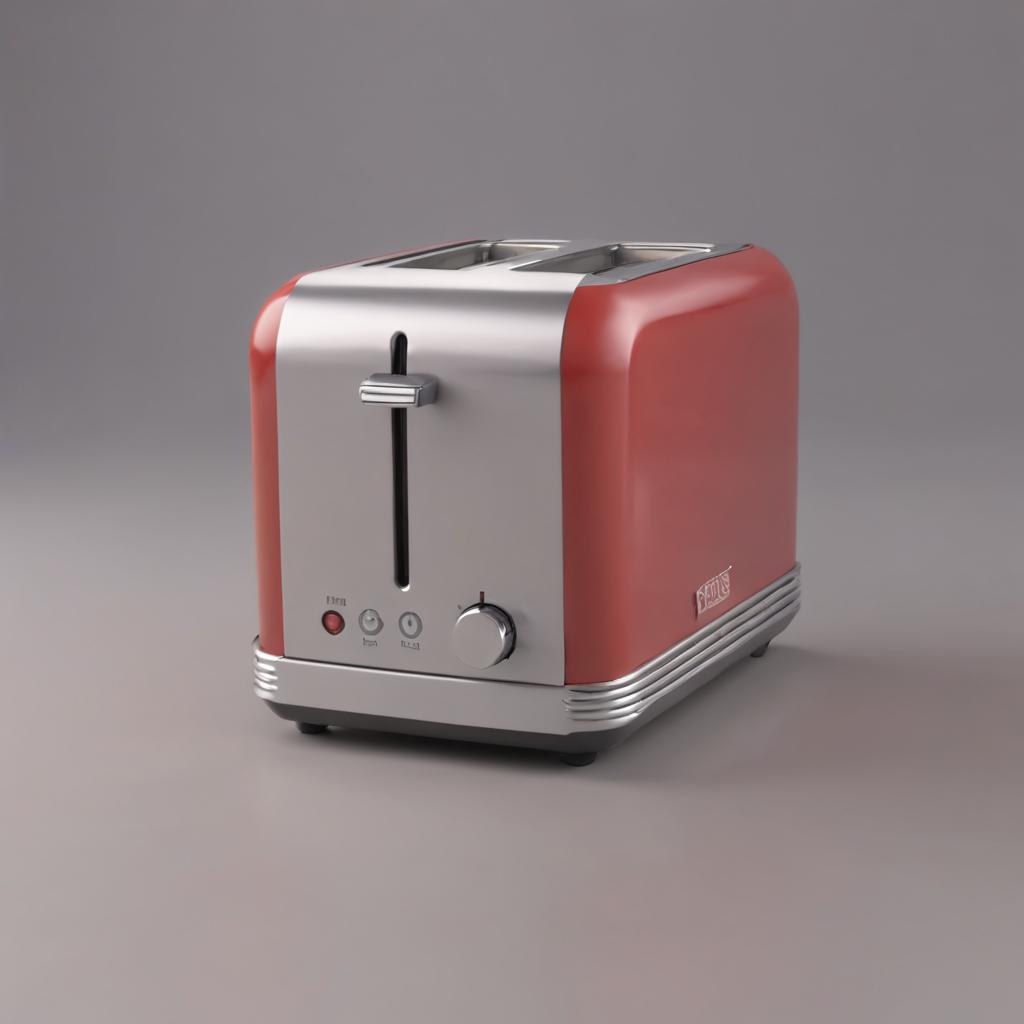} &
    \tabfigure{width=0.1\textwidth}{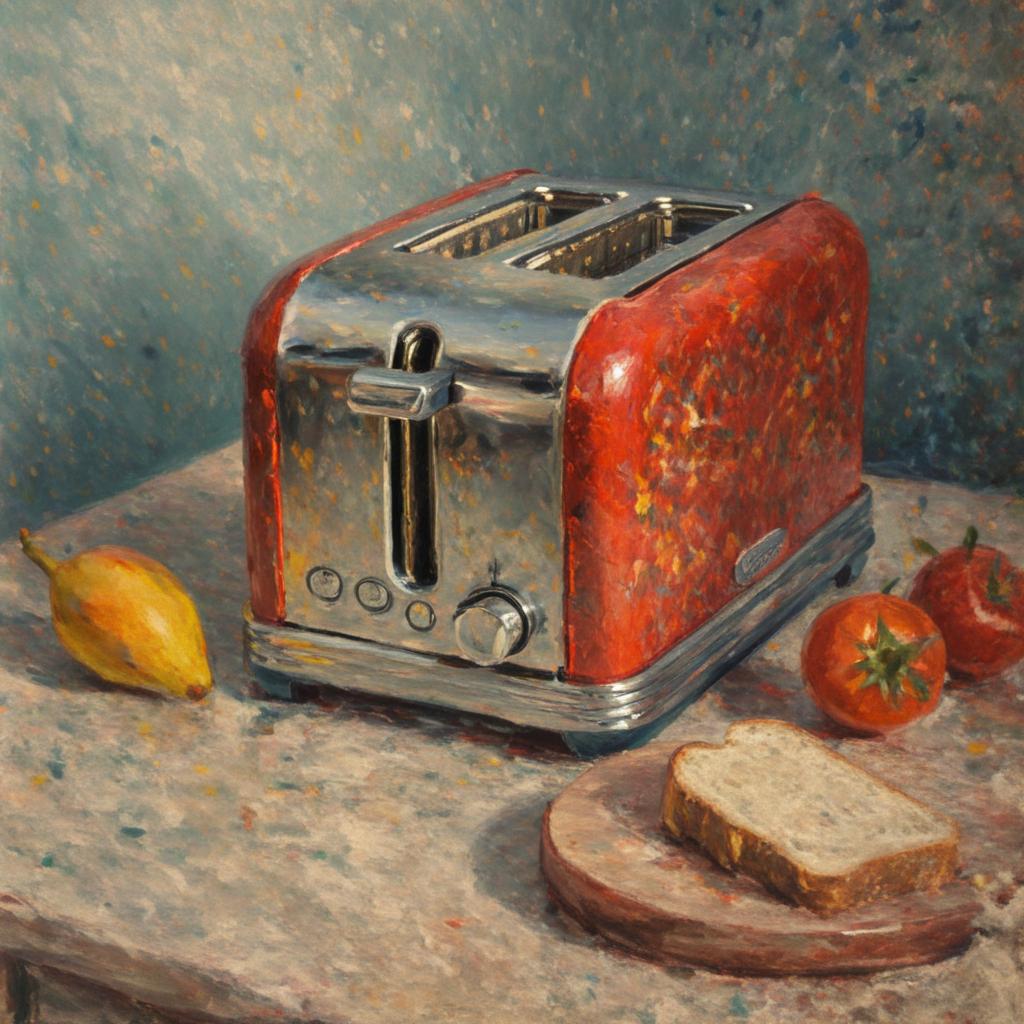} &
    \tabfigure{width=0.1\textwidth}{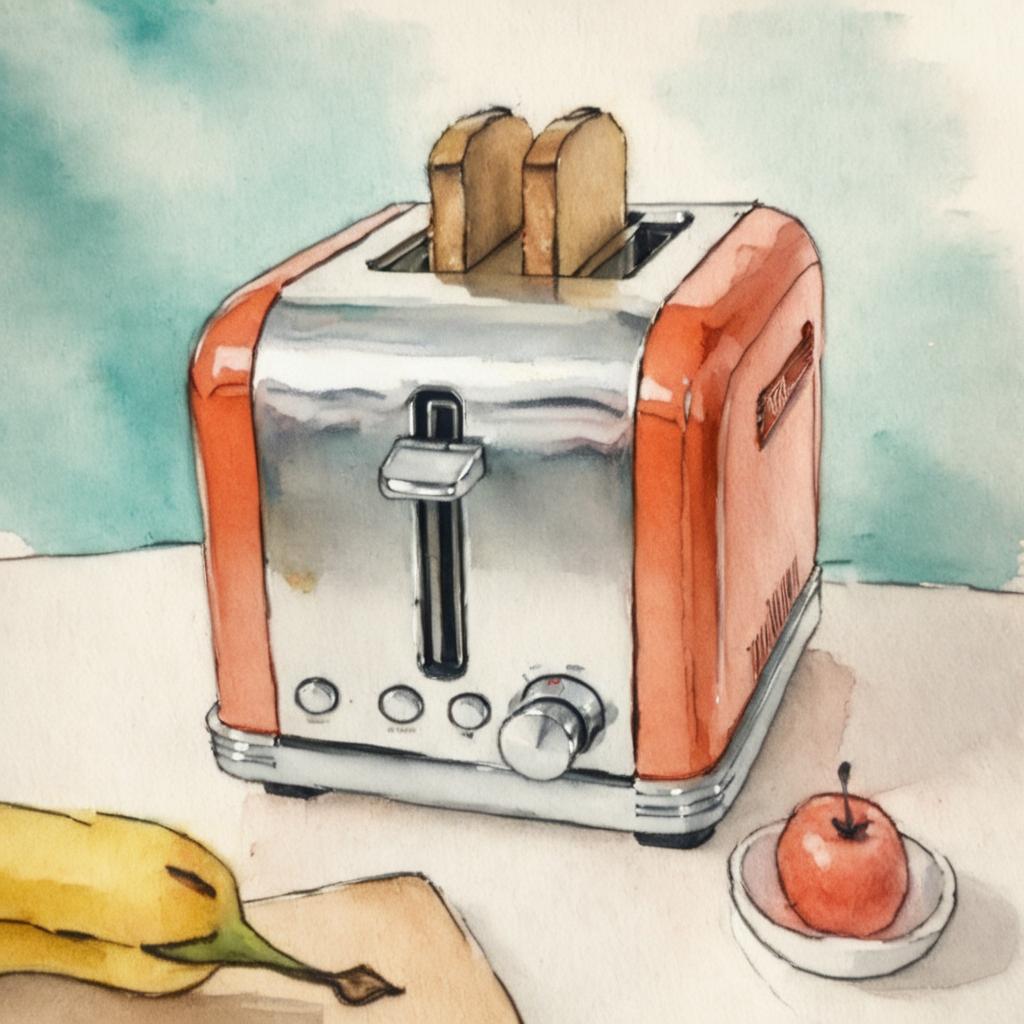} &
    \tabfigure{width=0.1\textwidth}{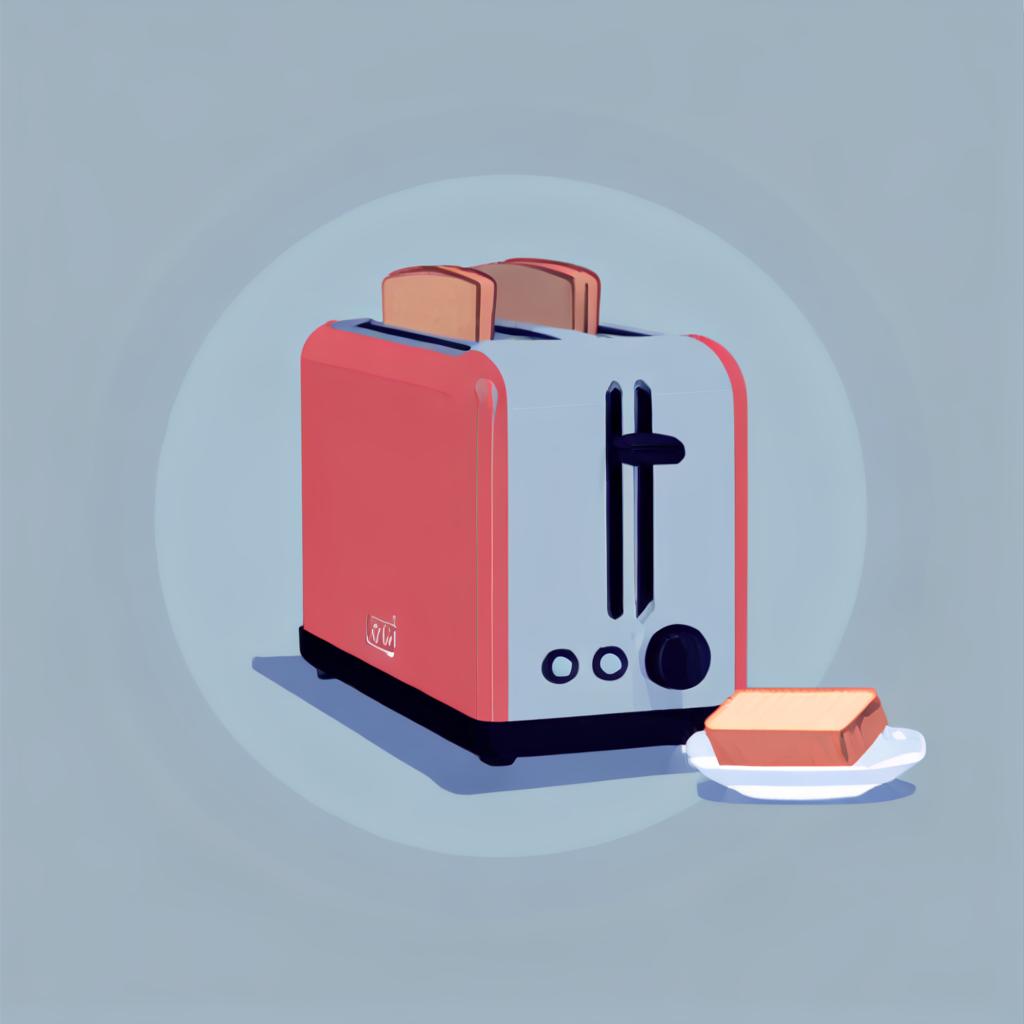} &
    \tabfigure{width=0.1\textwidth}{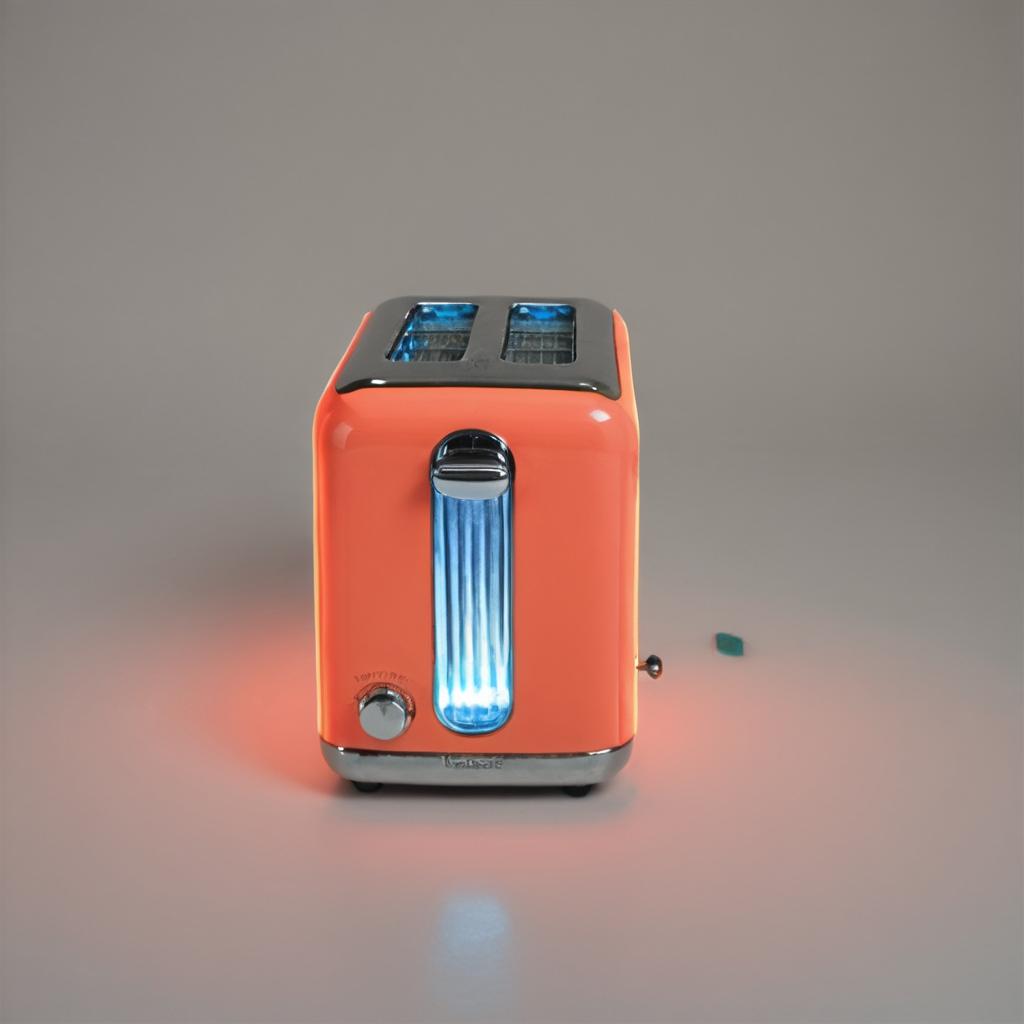} &
    \tabfigure{width=0.1\textwidth}{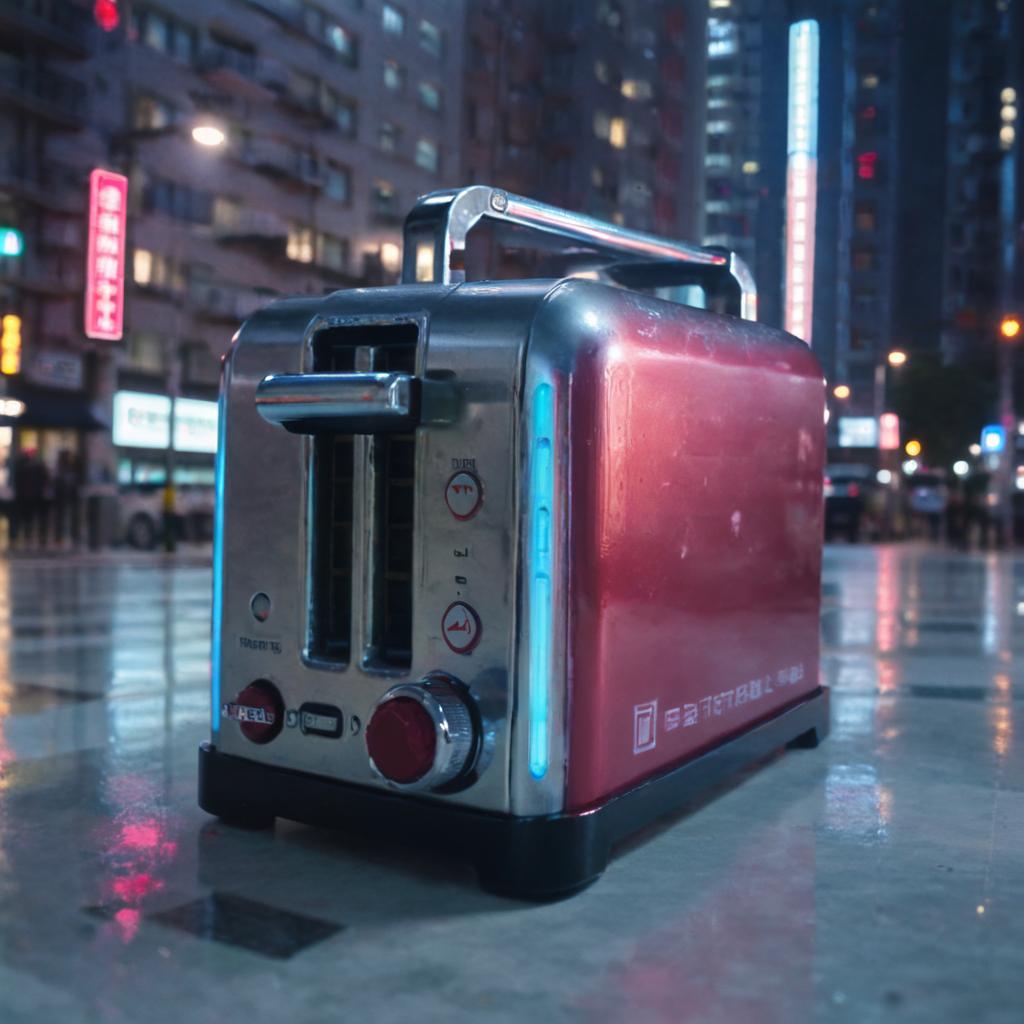} \\ %

    \tabfigure{height=0.1\textwidth}{} &
    \tabfigure{width=0.1\textwidth}{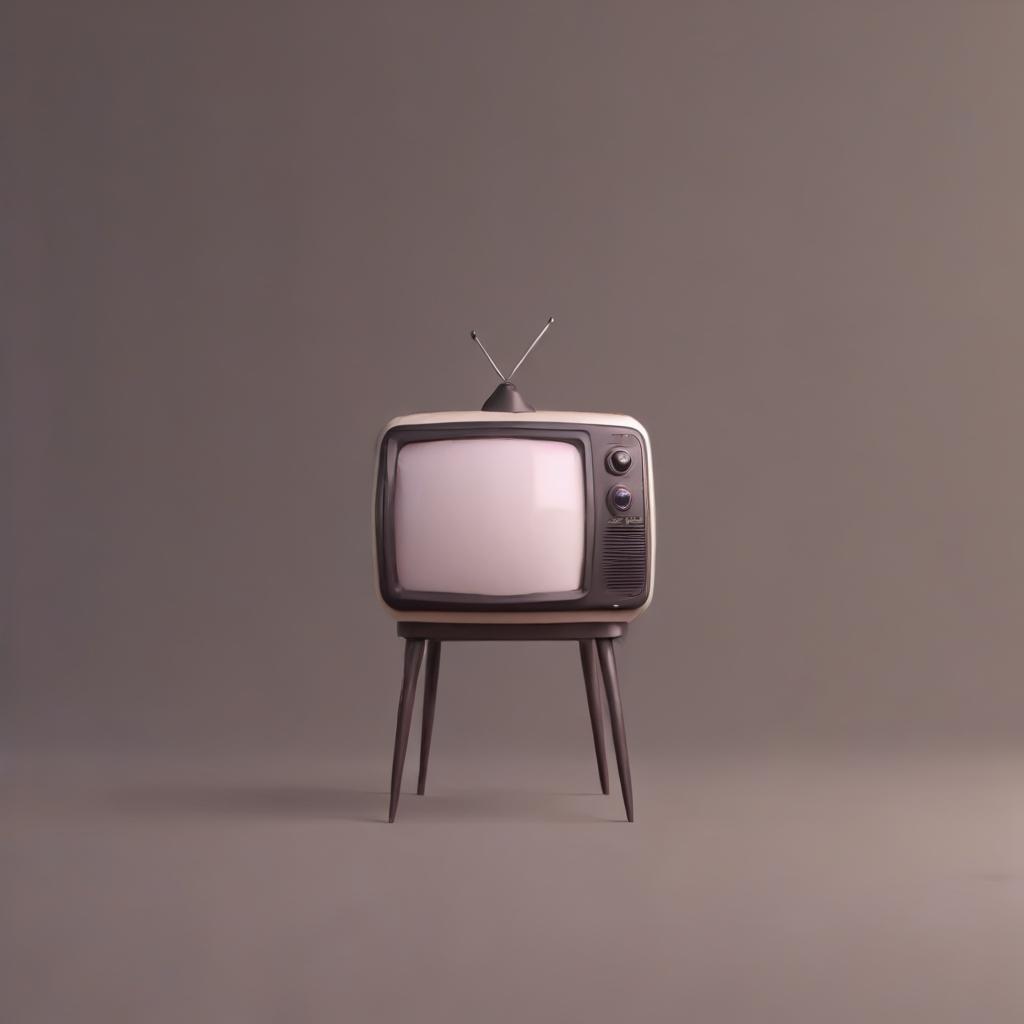} &
    \tabfigure{width=0.1\textwidth}{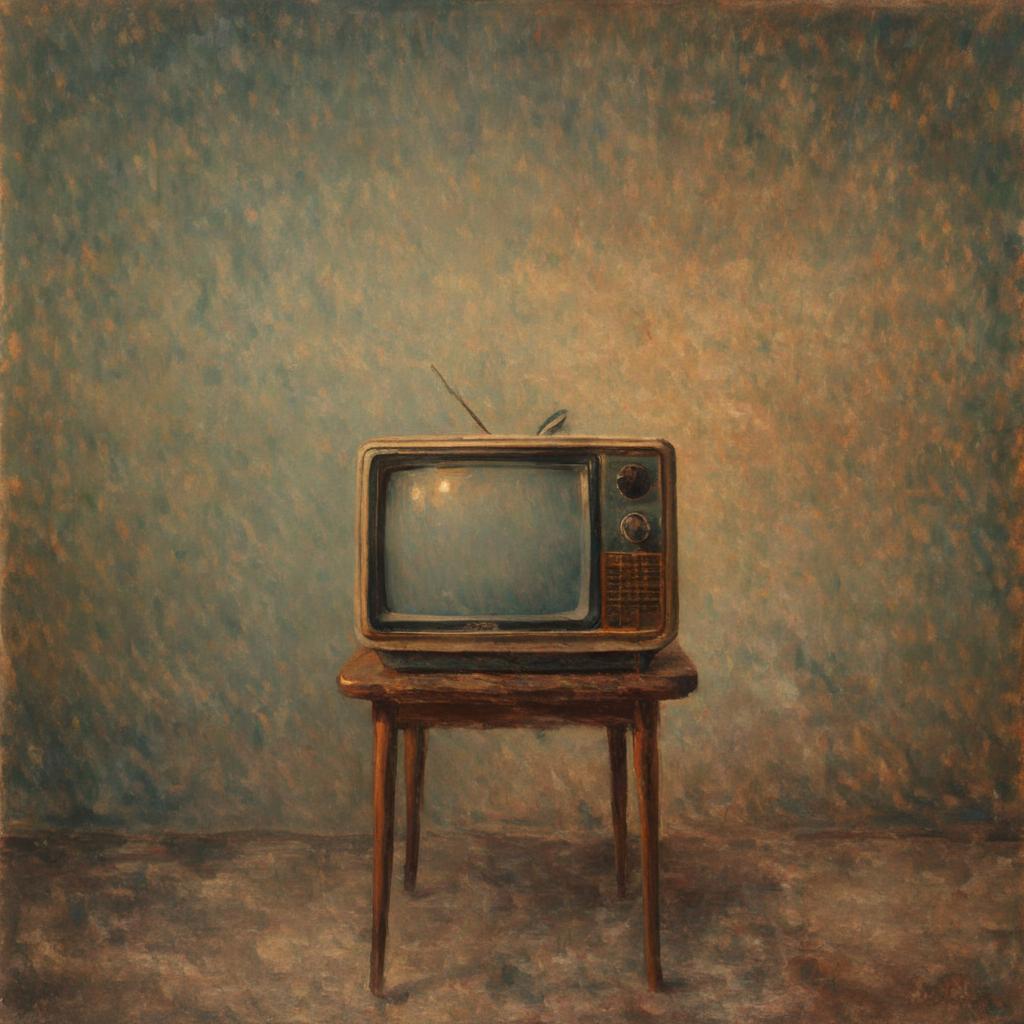} &
    \tabfigure{width=0.1\textwidth}{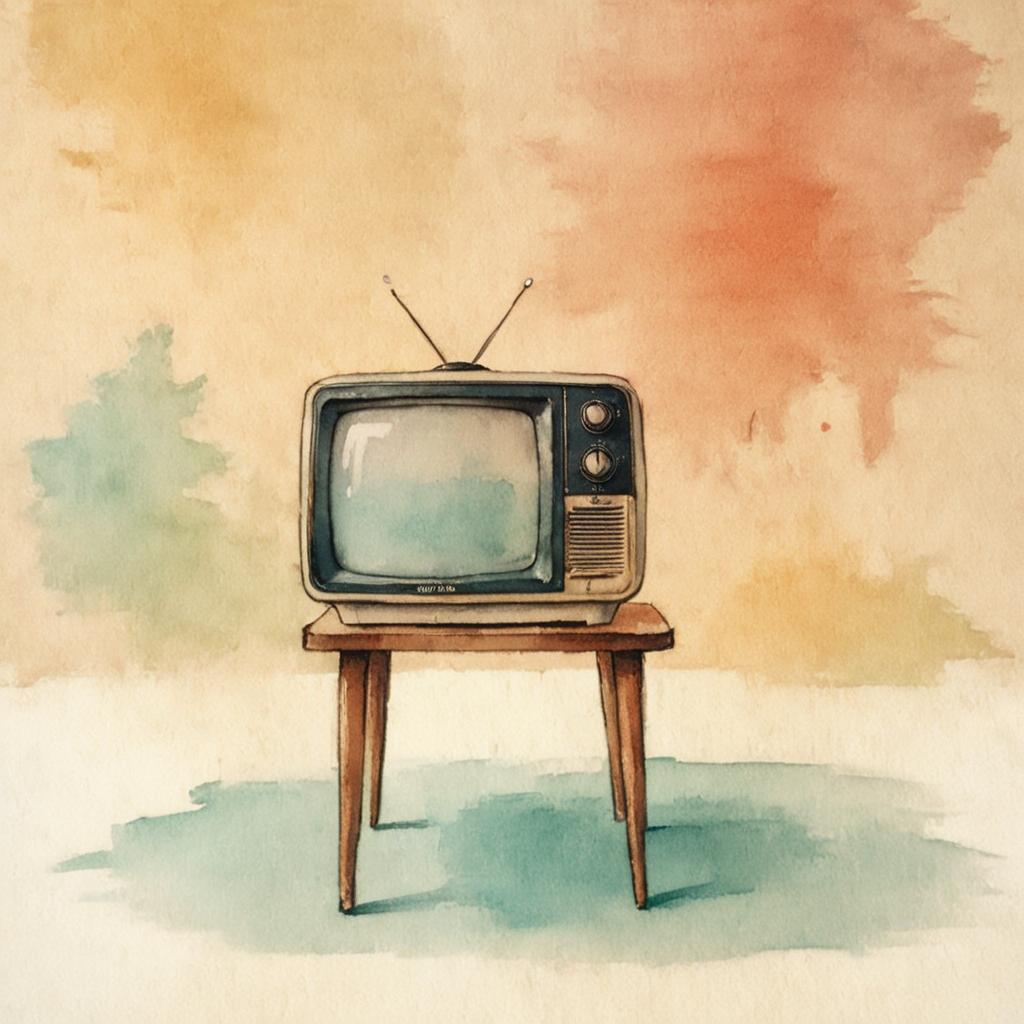} &
    \tabfigure{width=0.1\textwidth}{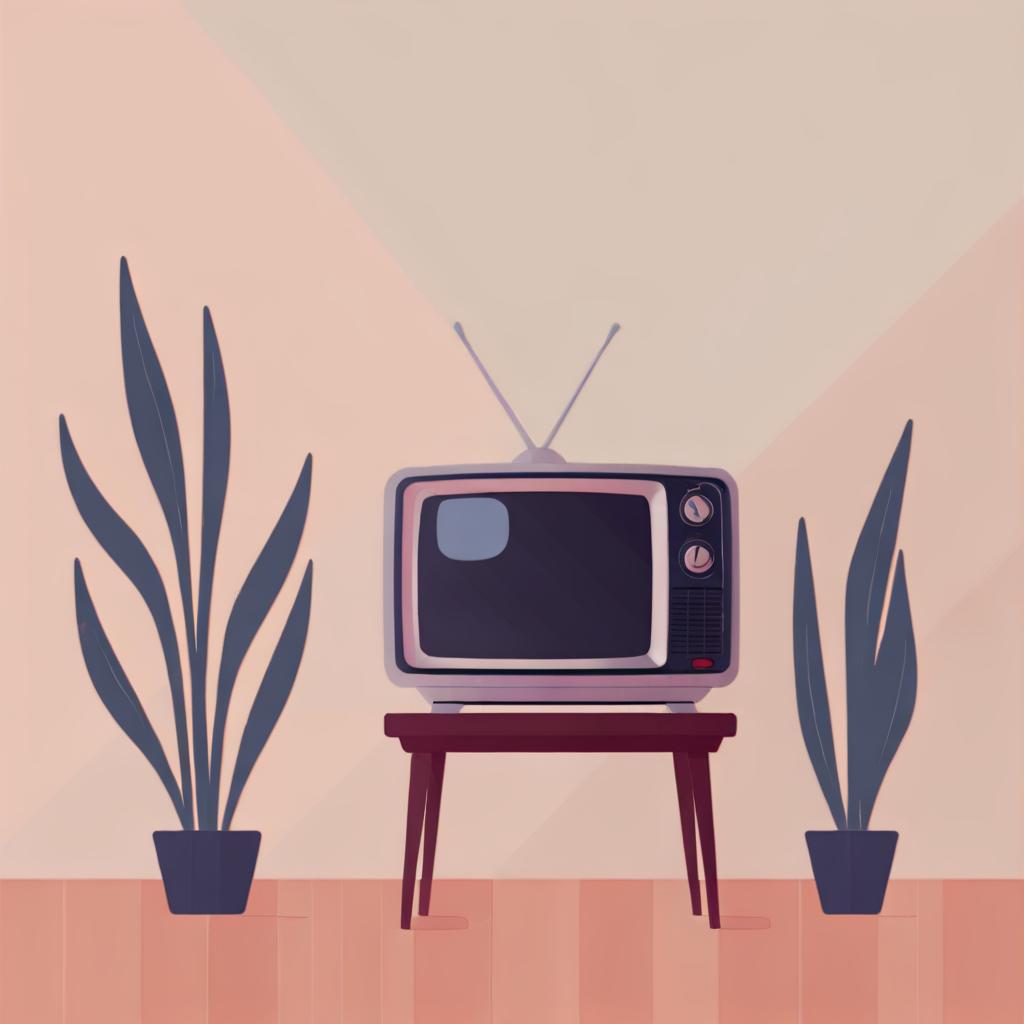} &
    \tabfigure{width=0.1\textwidth}{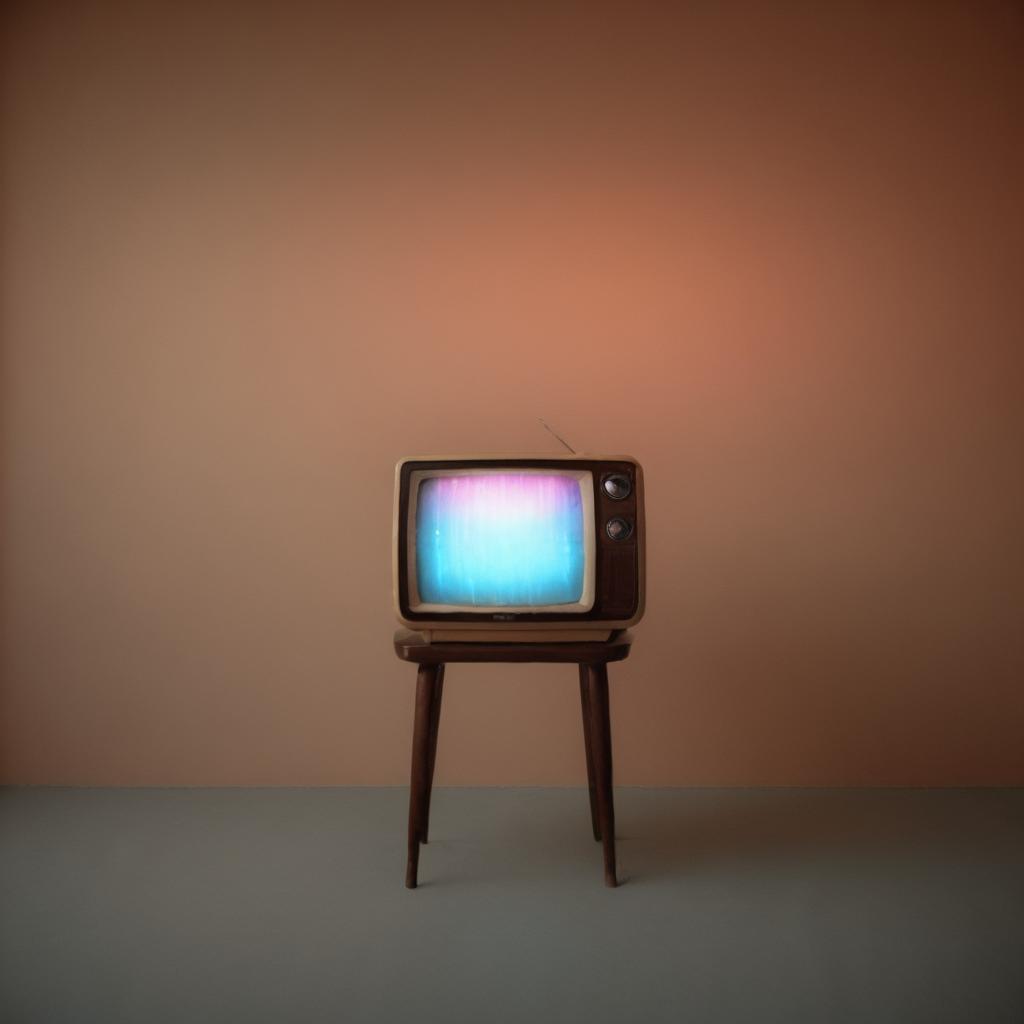}  &
        \tabfigure{width=0.1\textwidth}{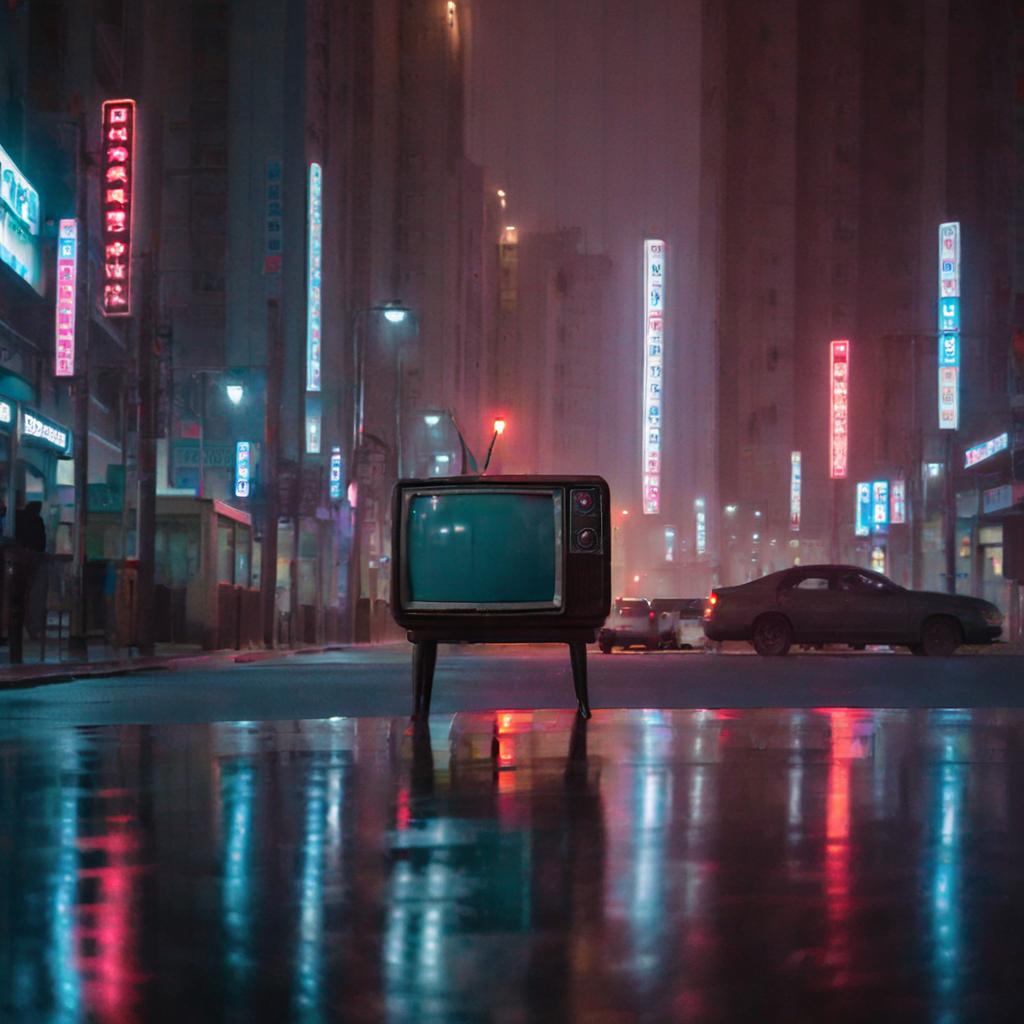} \\ %

    \tabfigure{width=0.1\textwidth}{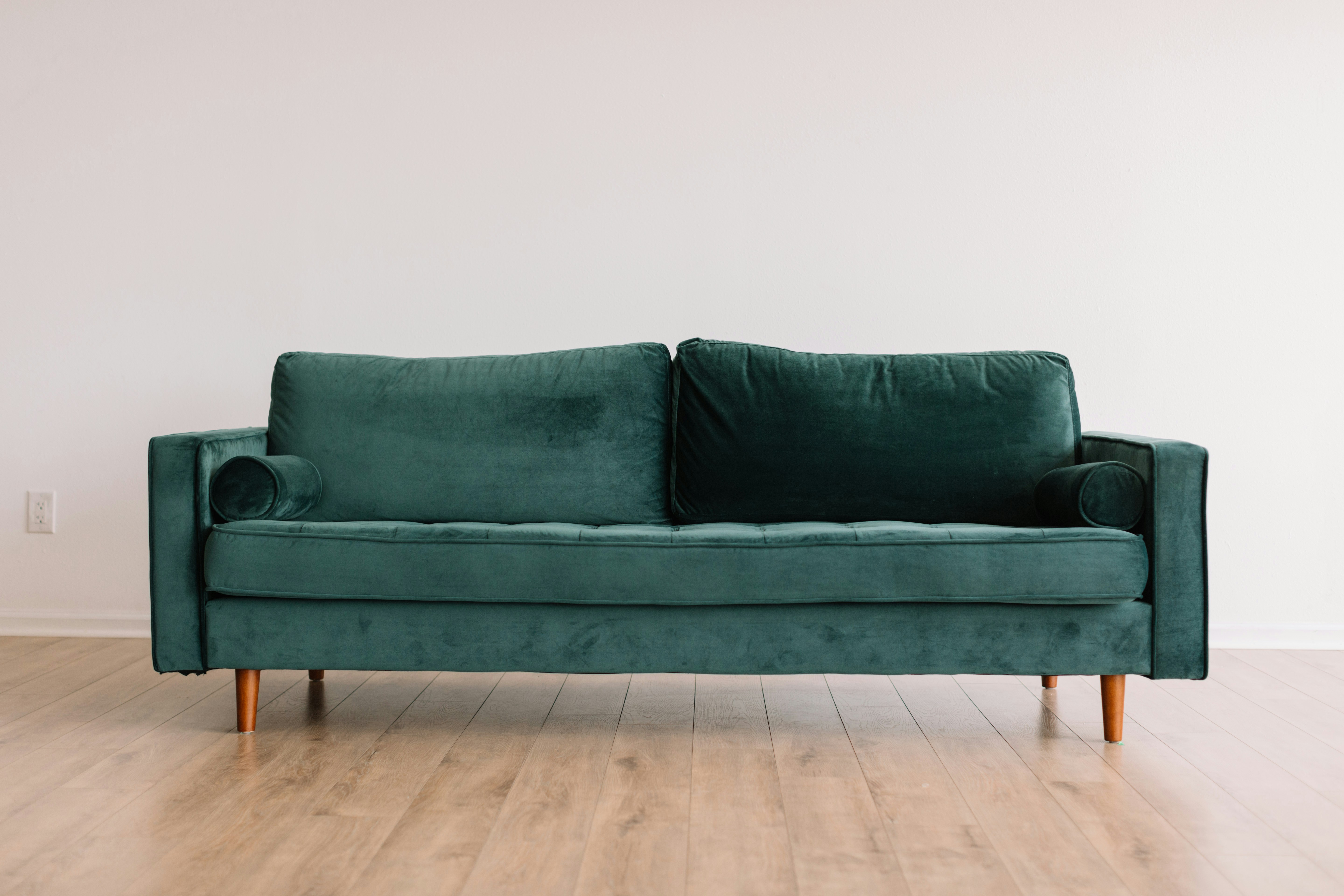} &
    \tabfigure{width=0.1\textwidth}{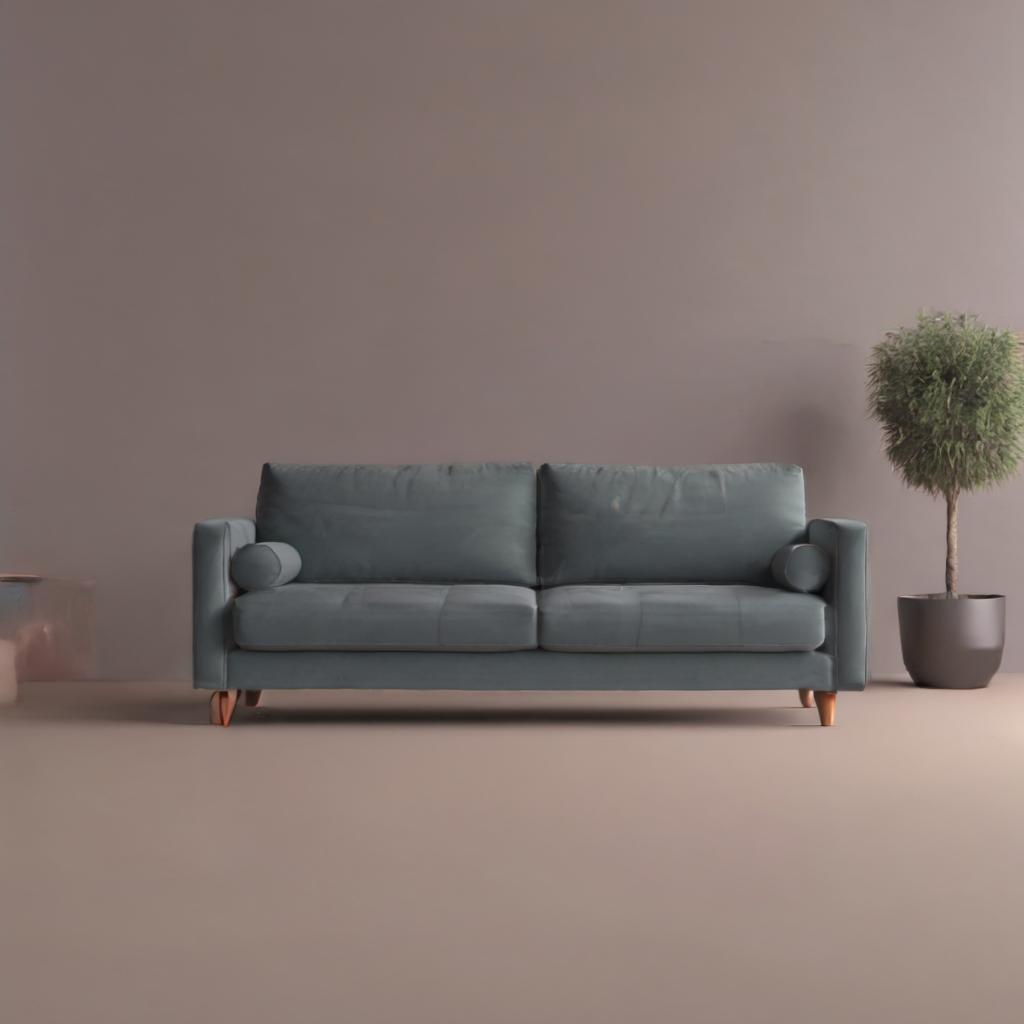} &
    \tabfigure{width=0.1\textwidth}{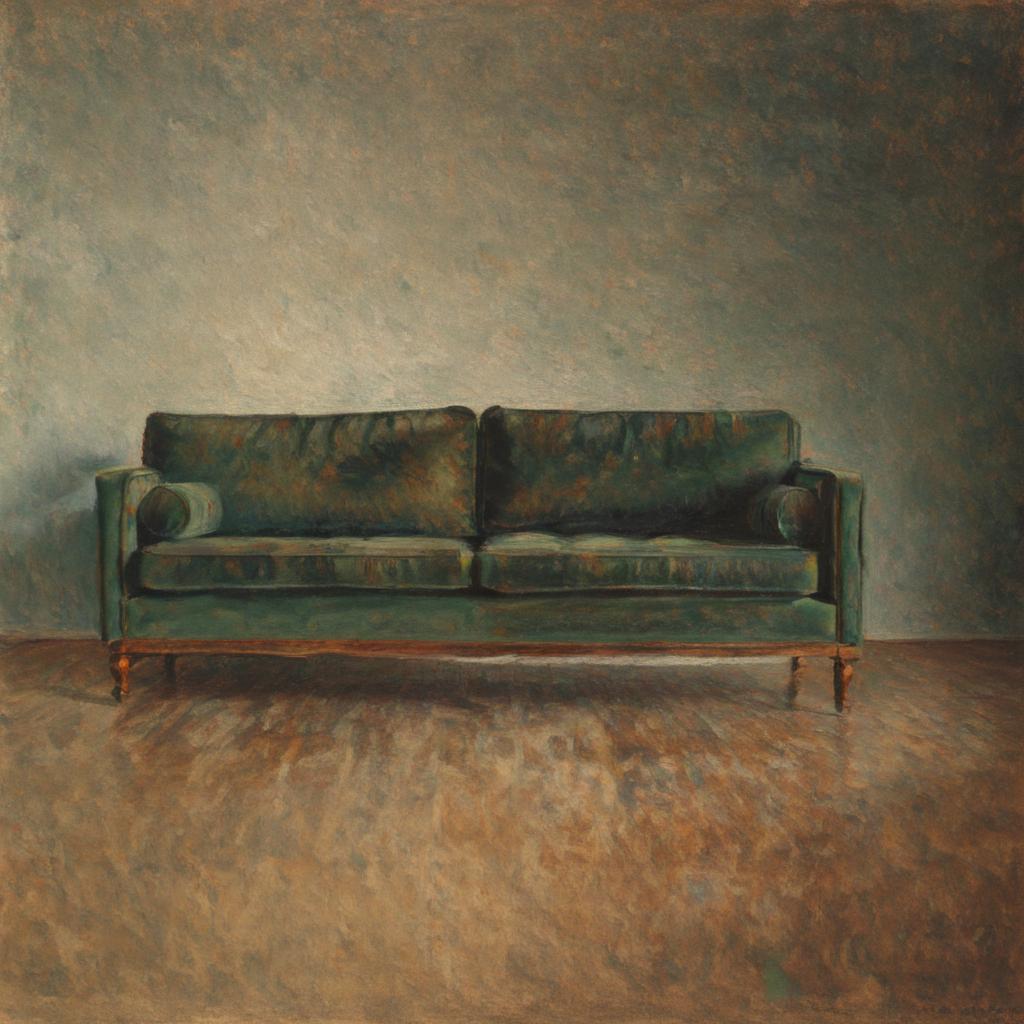} &
    \tabfigure{width=0.1\textwidth}{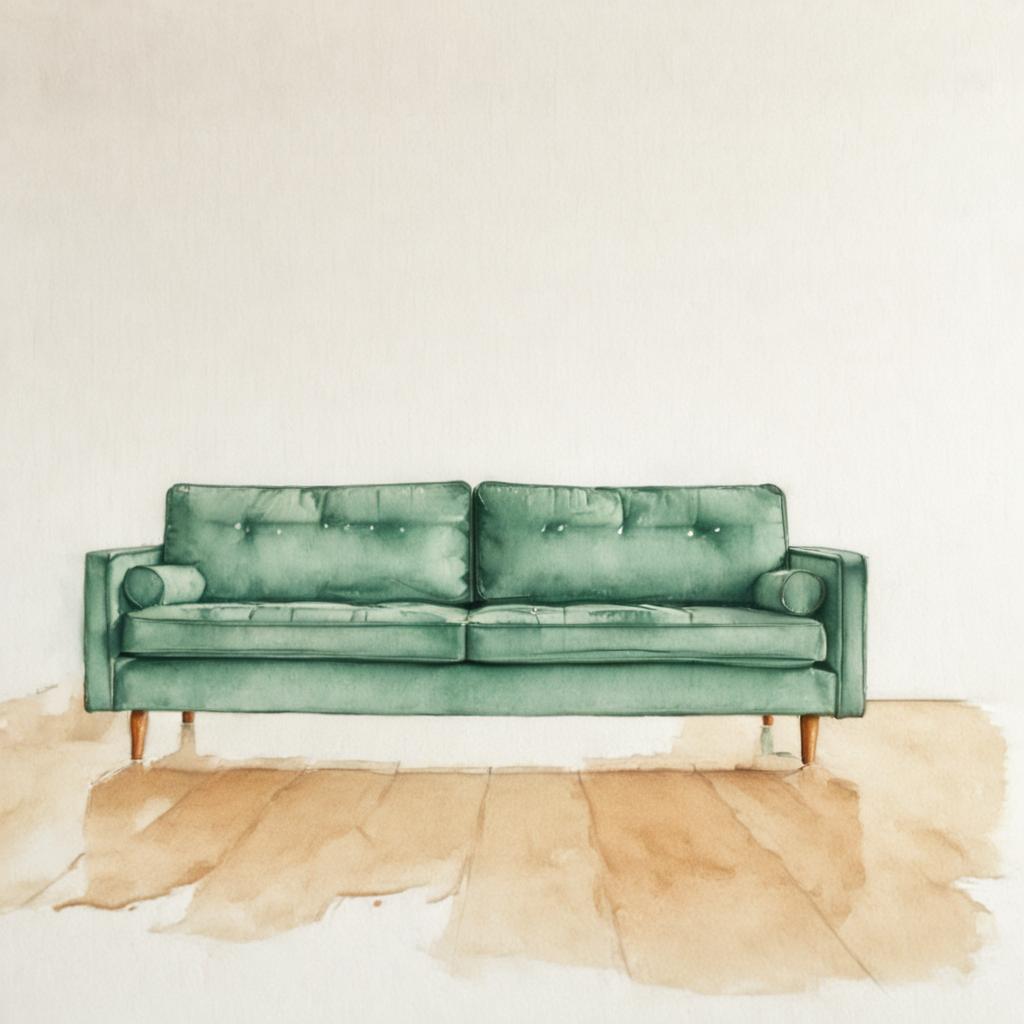} &
    \tabfigure{width=0.1\textwidth}{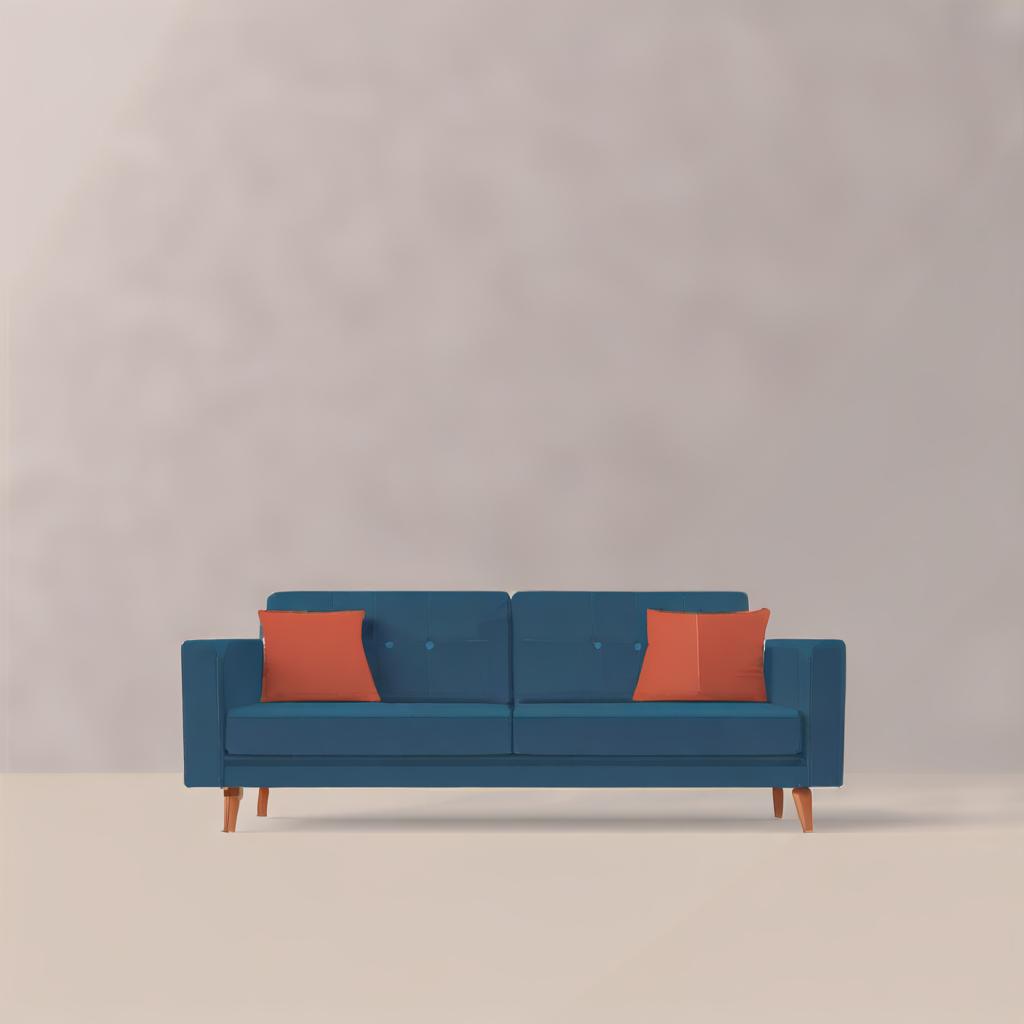} &
    \tabfigure{width=0.1\textwidth}{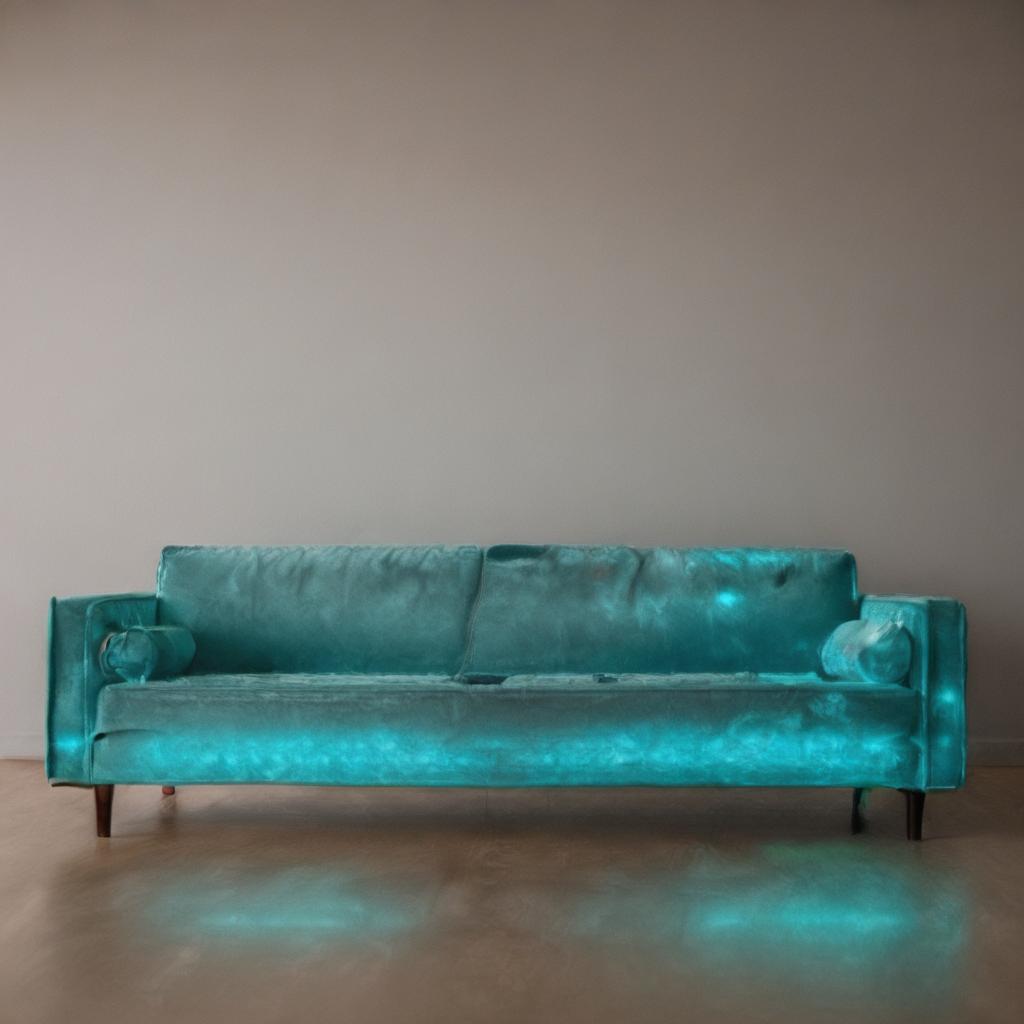}  &
    \tabfigure{width=0.1\textwidth}{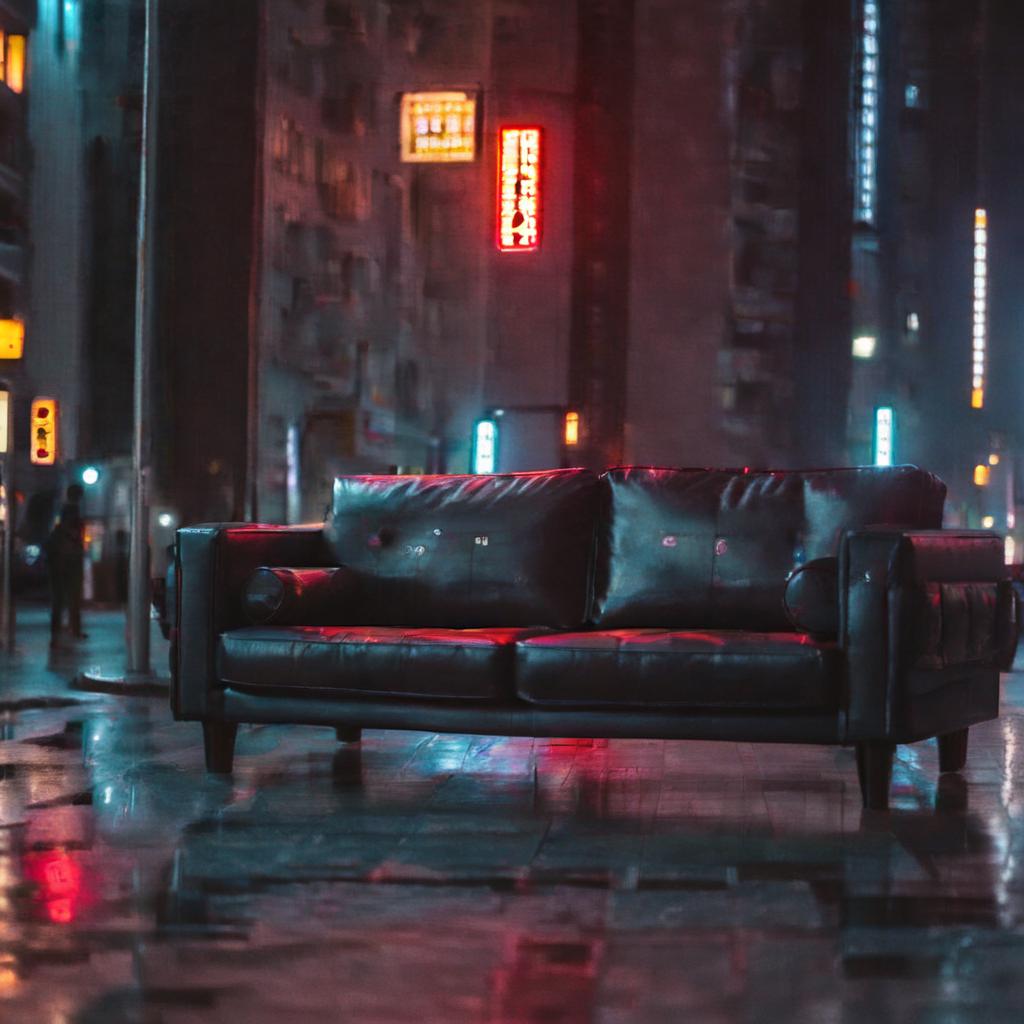} \\ %
    \end{tabular}
    }
    \caption{\textbf{Generalization to New Subjects and Styles.}  \ours performs well also on new objects and styles.}
    \label{fig:generalization}
\end{figure}

As common in the personalized image generation literature, we employed the DreamBooth dataset, which includes a diverse set of objects.
In the main paper, we already tested generalization to new subjects (clock, teapot, and can), different from pre-training categories (see \cref{tab:dataset_split} for details).
We consider three new furniture subjects (\textit{toaster} collected by us; \textit{\href{https://unsplash.com/photos/turned-off-black-television-UBhpOIHnazM}{tv}, \href{https://unsplash.com/photos/green-fabric-sofa-fZuleEfeA1Q}{sofa}} from the web), and a new substantially different style (\textit{\href{https://images.unsplash.com/photo-1601042879364-f3947d3f9c16}{cyberpunk}}). 
The aim of this experiment is twofold: (1) we further prove the generalization of our approach to unseen subjects-style, (2) we demonstrate the simplicity of collecting new LoRAs from single images and merge them for joint subject-style personalized image generation.
Fig.~\ref{fig:generalization} shows that our hypernetwork generalizes well and does not need to be trained every time a new object appears. 
Also in this case, we outperform ZipLoRA in \ourmetric~(0.8 vs.\ 0.6).

\subsection{Generalization to New Splits}
We re-trained the hypernetwork using 2 new splits, with the same hyperparameters as in the other experiments in the paper.
The two splits that we consider are:
\begin{enumerate}
    \item Training subjects: objects (no animals included); \\
Test subjects: stuffed animals; \\
Training styles: 3D renderings; \\
Test styles: cartoon.
    \item Training subjects: animals and stuffed animals; \\
Test subjects: objects; \\
Training styles: watercolor paintings; \\
Test styles:  abstract rainbow, wooden sculpture, melting golden rendering.
\end{enumerate}

\begin{figure}[t]
\vspace{-3.5mm}
    \centering
    \resizebox{0.98\linewidth}{!}{%
    \begin{tabular}[c]{c|L | L L L | L L L|}

\cline{2-8}
 & &\multicolumn{3}{c|}{  \textbf{Ours}}&\multicolumn{3}{c|}{\textbf{ZipLoRA}}\\

&\raisebox{-8mm}
{\begin{tikzpicture}
    \draw[->] (0, -1.5) -- (1.5, -1.5) node[below, midway] {Styles};
    \draw[->] (0, -1.5) -- (0, -3) node[right, midway] {Contents};
\end{tikzpicture}}
 &
         {\tabfigure{width=0.1\textwidth}{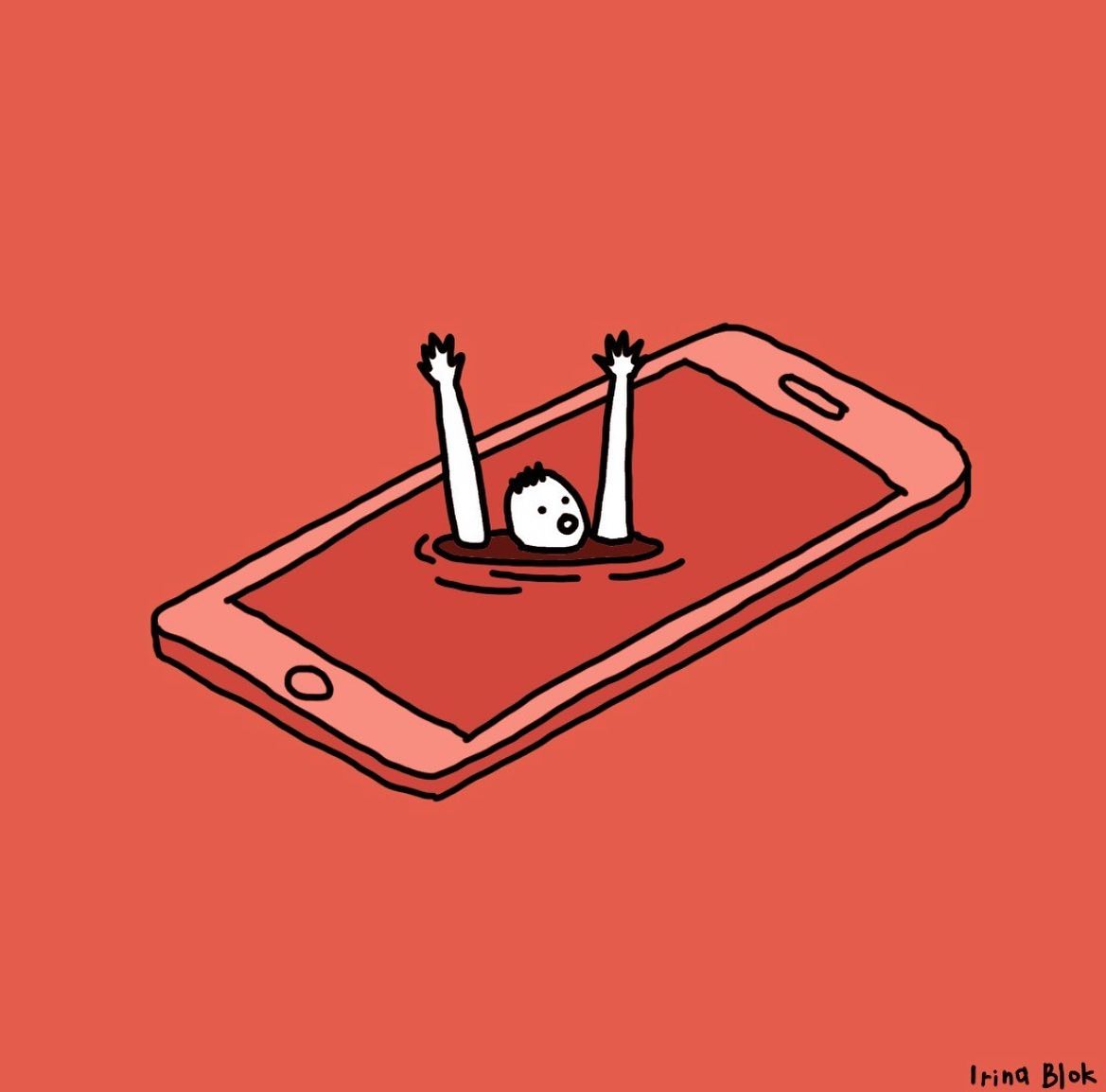}} &
    \tabfigure{width=0.1\textwidth}{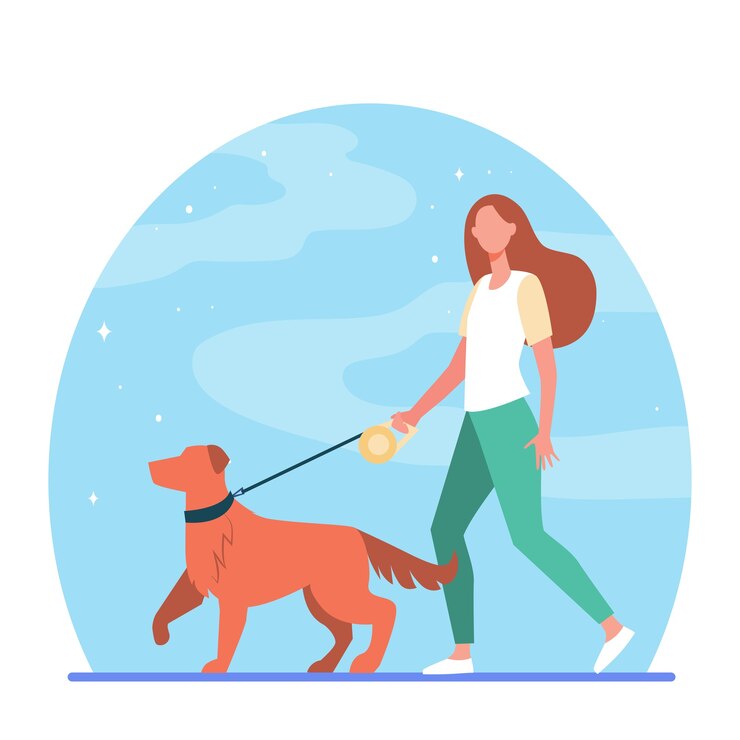} &
    \tabfigure{width=0.1\textwidth}{figures/dataset/test/flat.jpg} &
    \tabfigure{width=0.1\textwidth}{figures/new_splits/cartoon_line.jpg} &
        \tabfigure{width=0.1\textwidth}{figures/new_splits/flat_2.jpg} &
    \tabfigure{width=0.1\textwidth}{figures/dataset/test/flat.jpg}
    \\
    
  &
  \tabfigure{width=0.1\textwidth}{figures/dataset/test/wolf_plushie/00.jpg} &
    \tabfigure{width=0.1\textwidth}{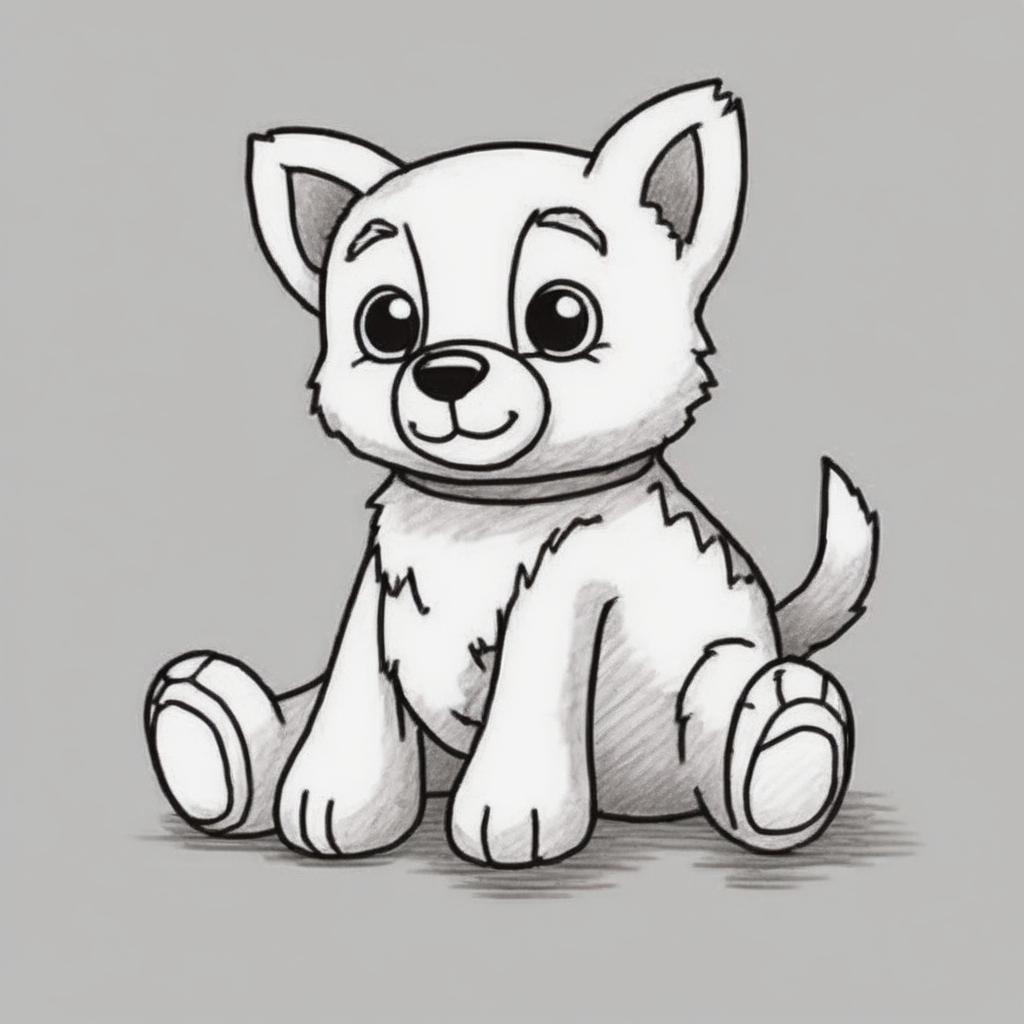} &
    \fcolorbox{blue}{white}{\tabfigure{width=0.096\textwidth}{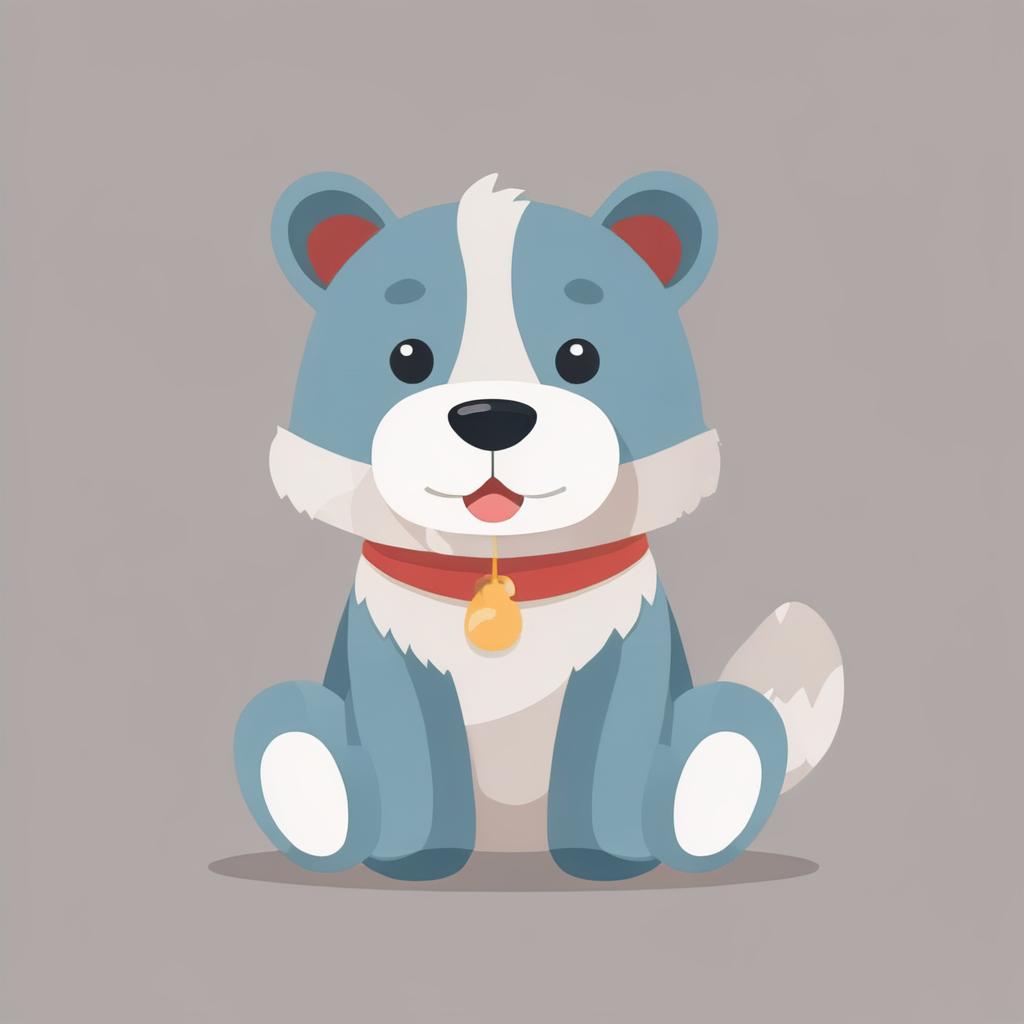}} &
    \tabfigure{width=0.1\textwidth}{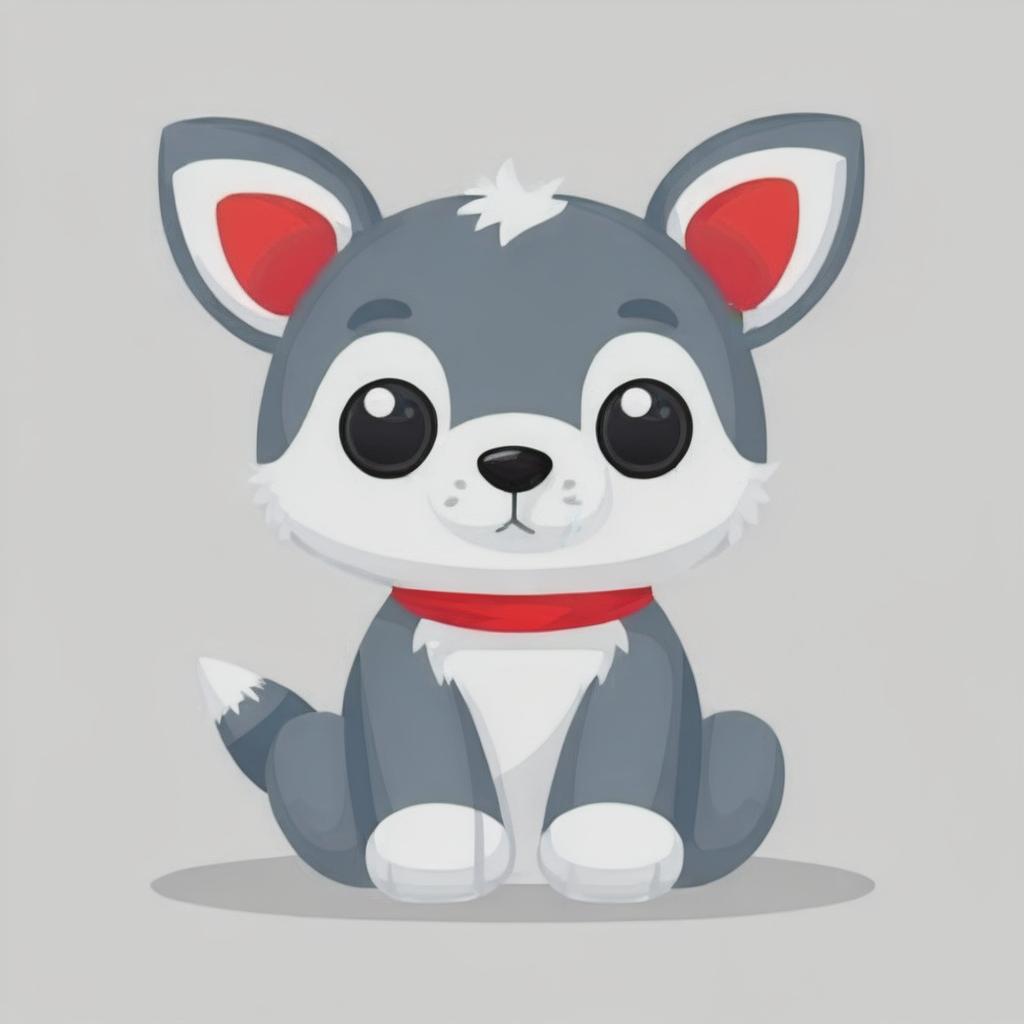} &
    \tabfigure{width=0.1\textwidth}{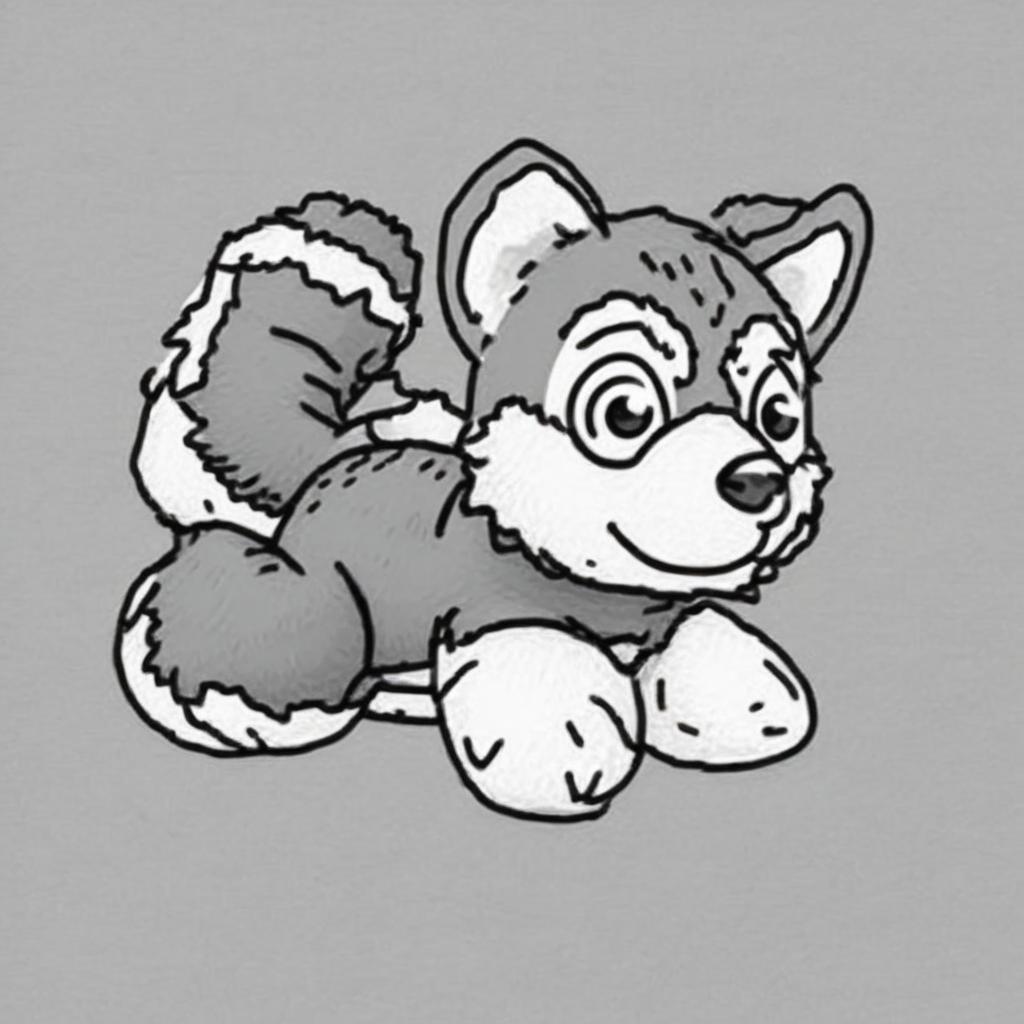} &
    \fcolorbox{blue}{white}{\tabfigure{width=0.096\textwidth}{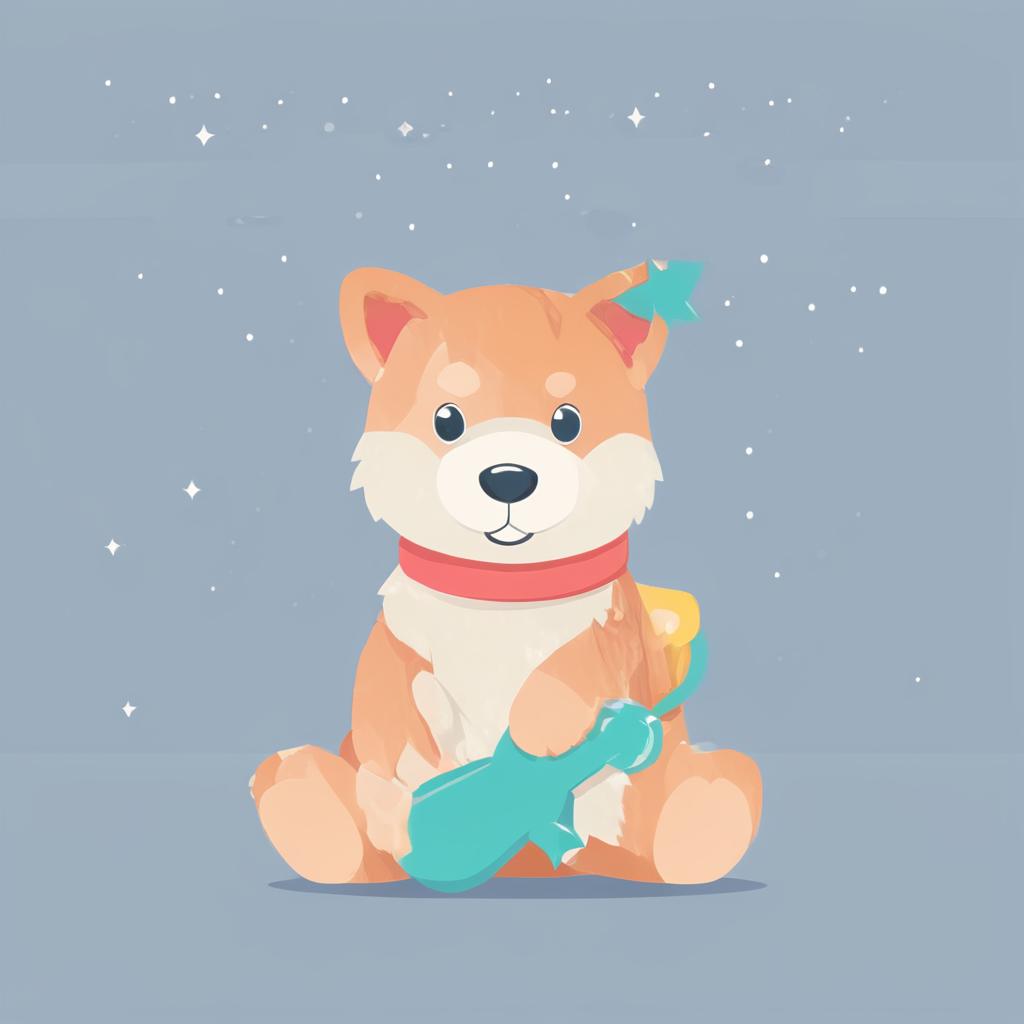}} &
    \tabfigure{width=0.1\textwidth}{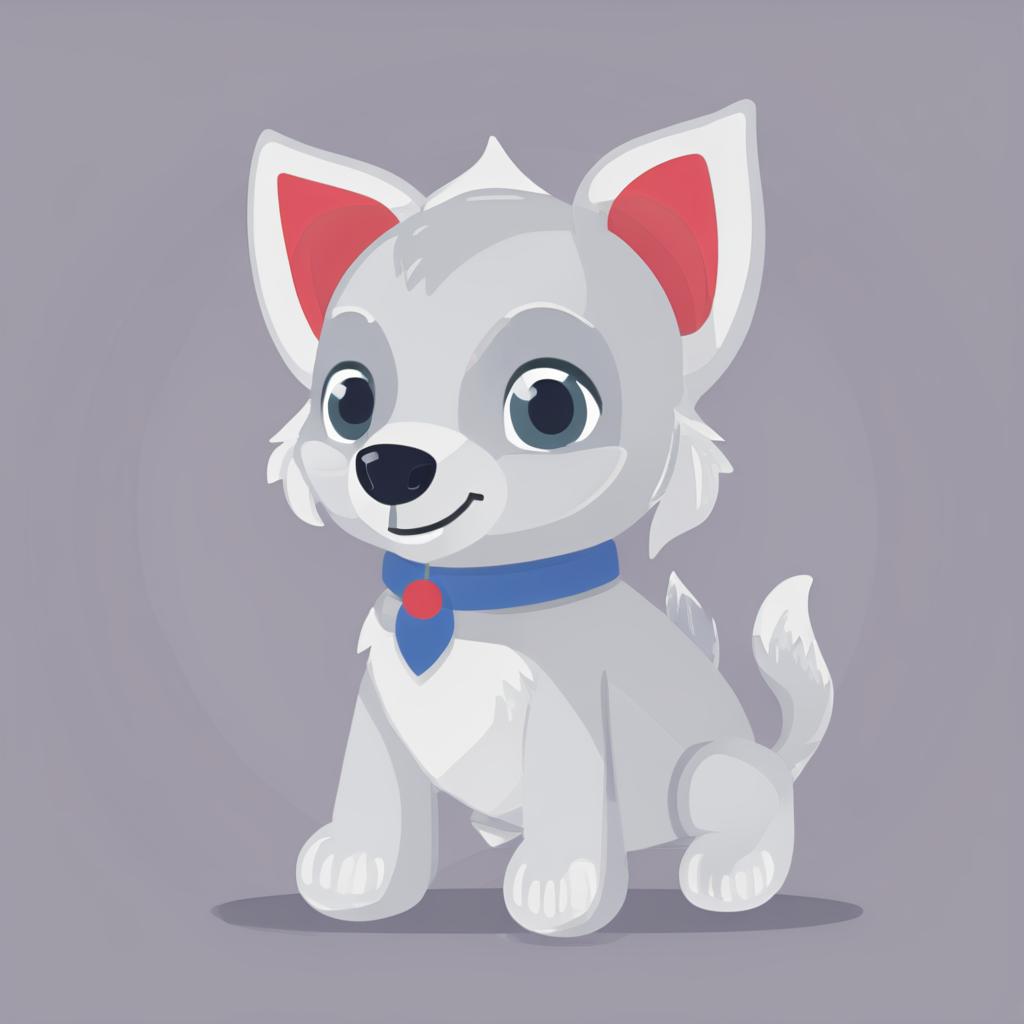} 
    \\    
\textbf{(1)} &
    \tabfigure{width=0.1\textwidth}{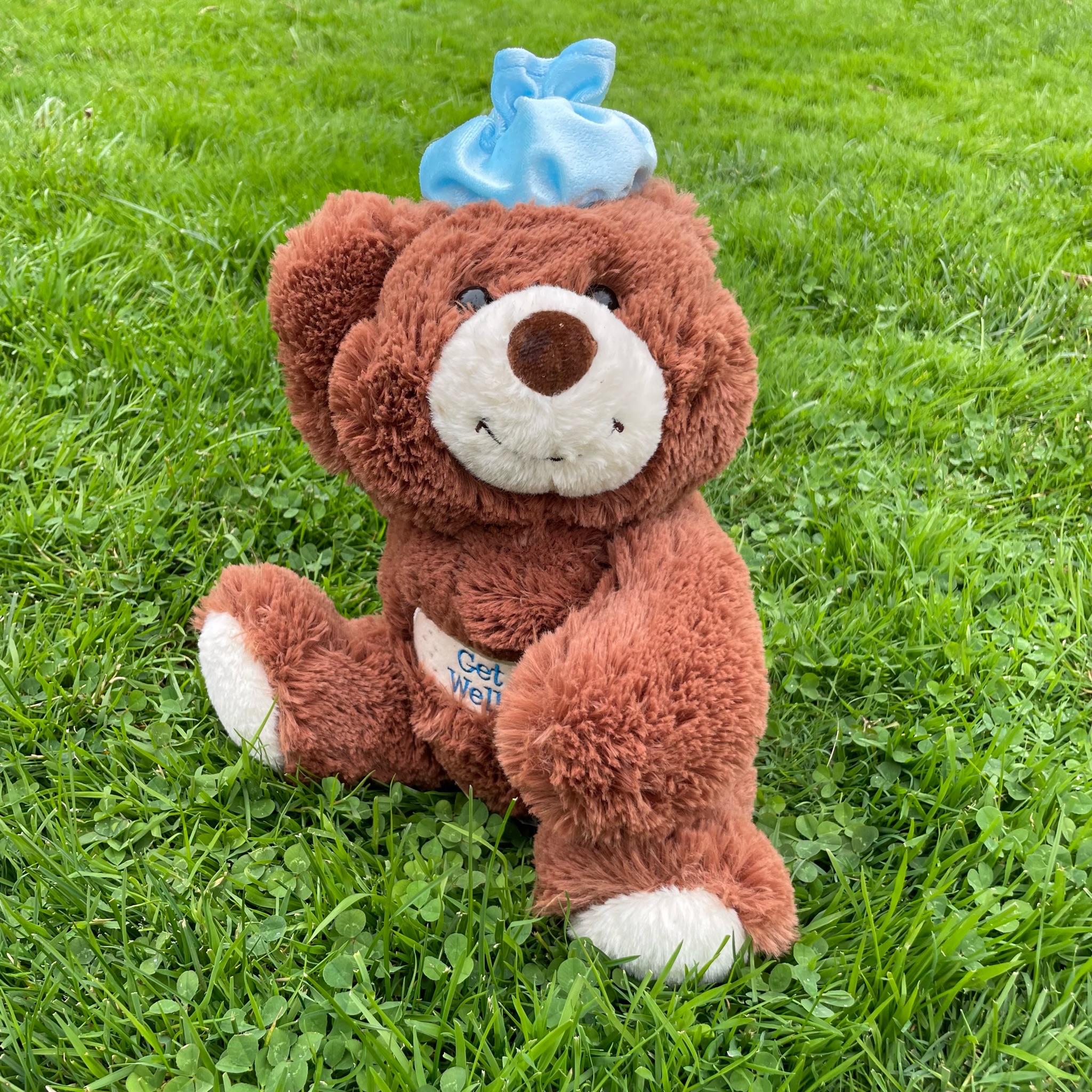} &
        \tabfigure{width=0.1\textwidth}{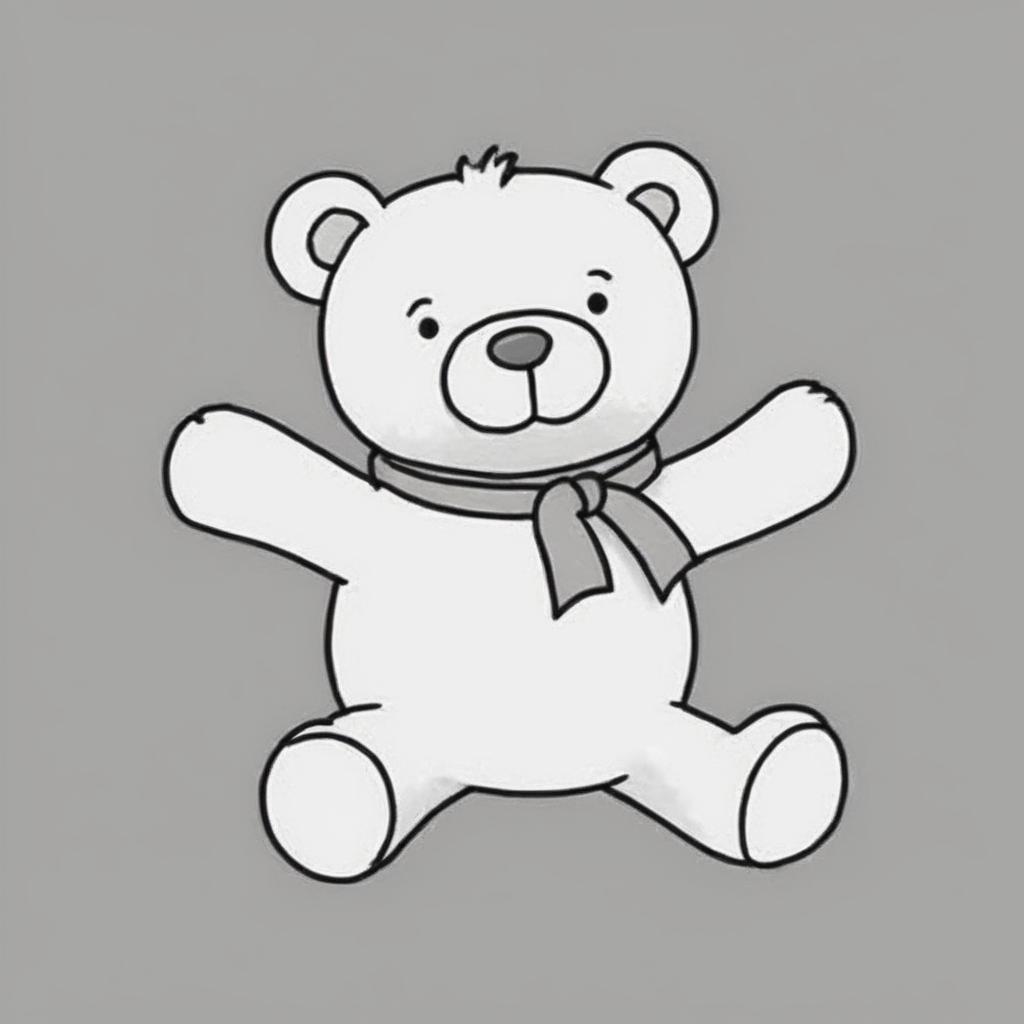} &
    \tabfigure{width=0.1\textwidth}{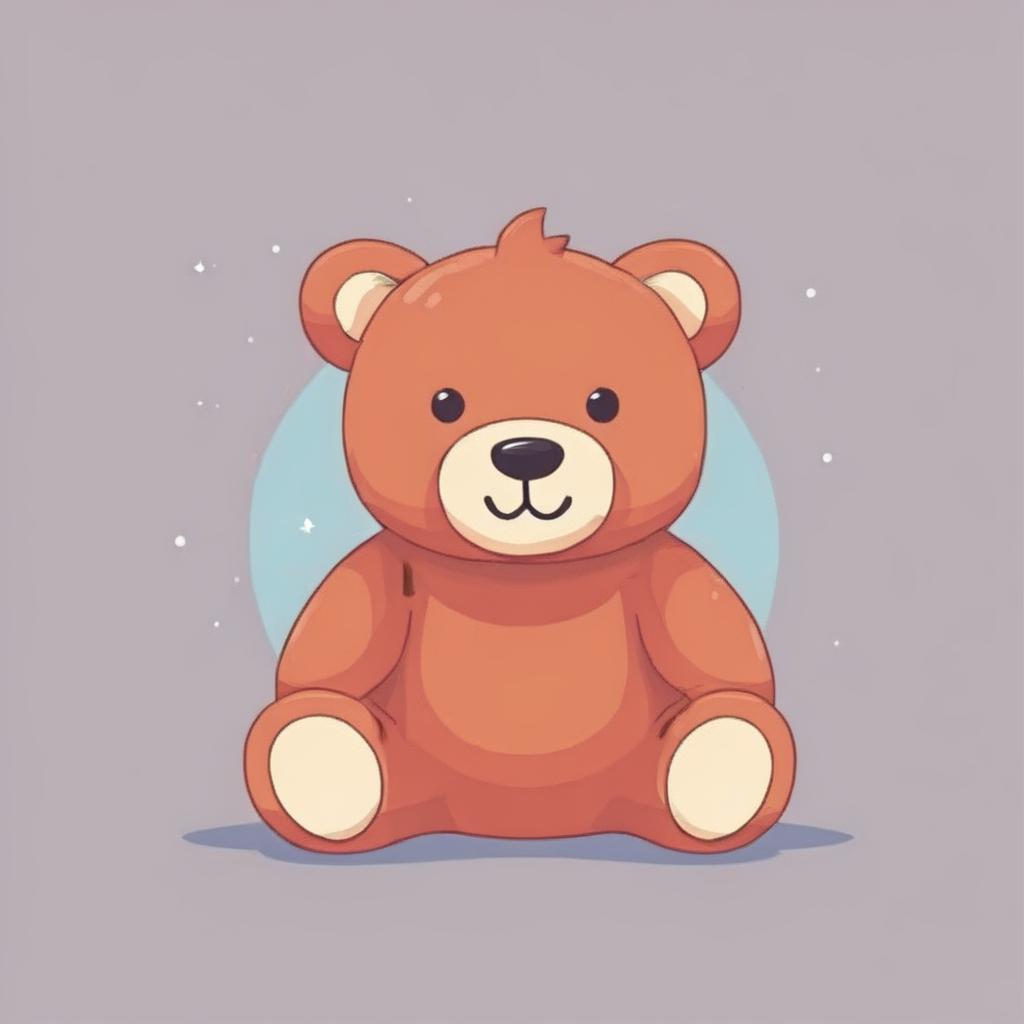} &
    \tabfigure{width=0.1\textwidth}{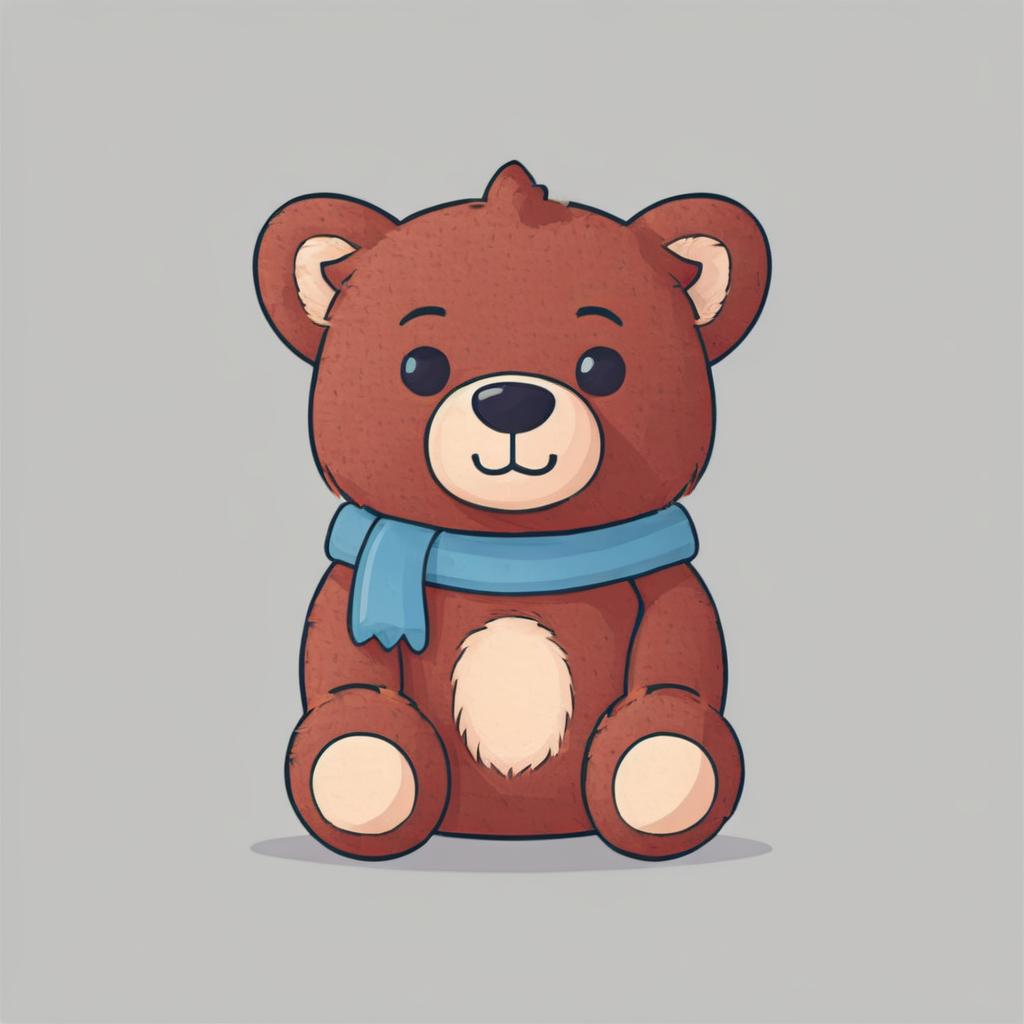} &
        \tabfigure{width=0.1\textwidth}{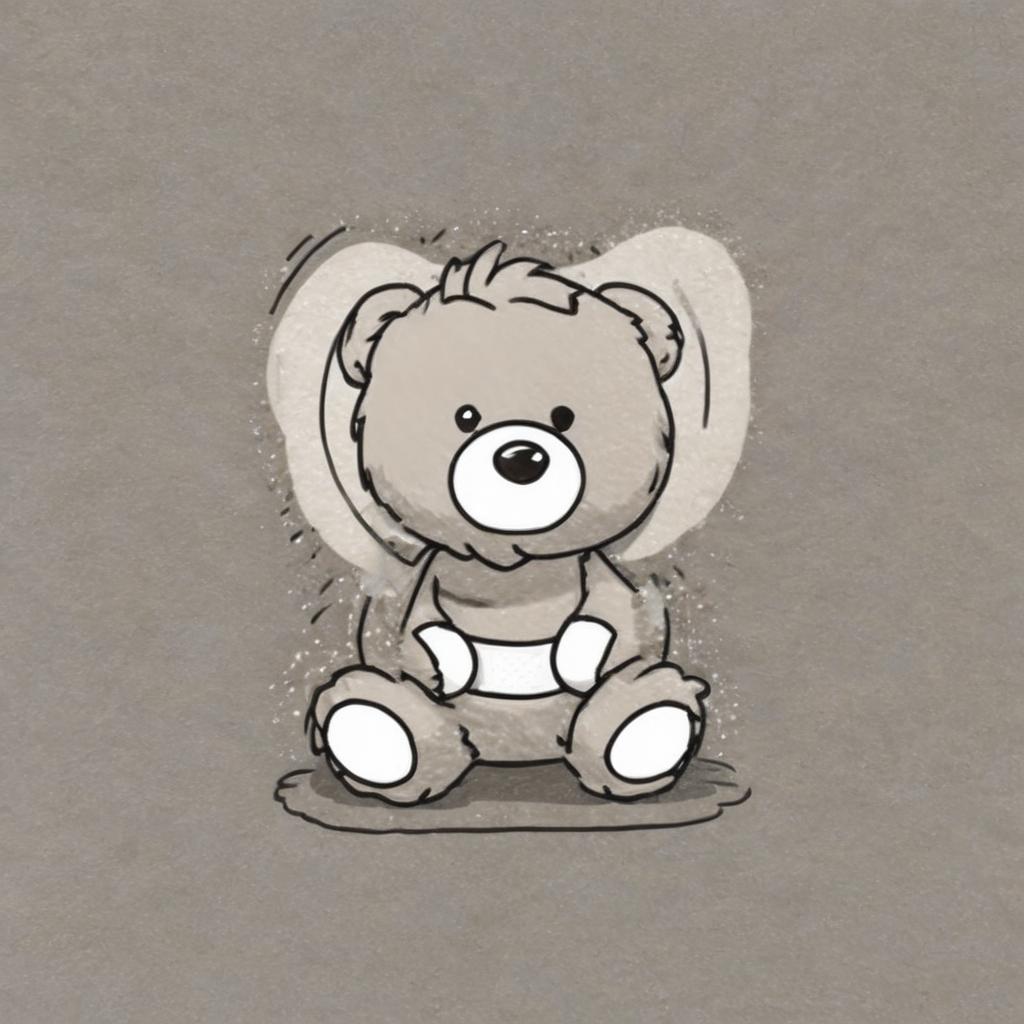} &
        \tabfigure{width=0.1\textwidth}{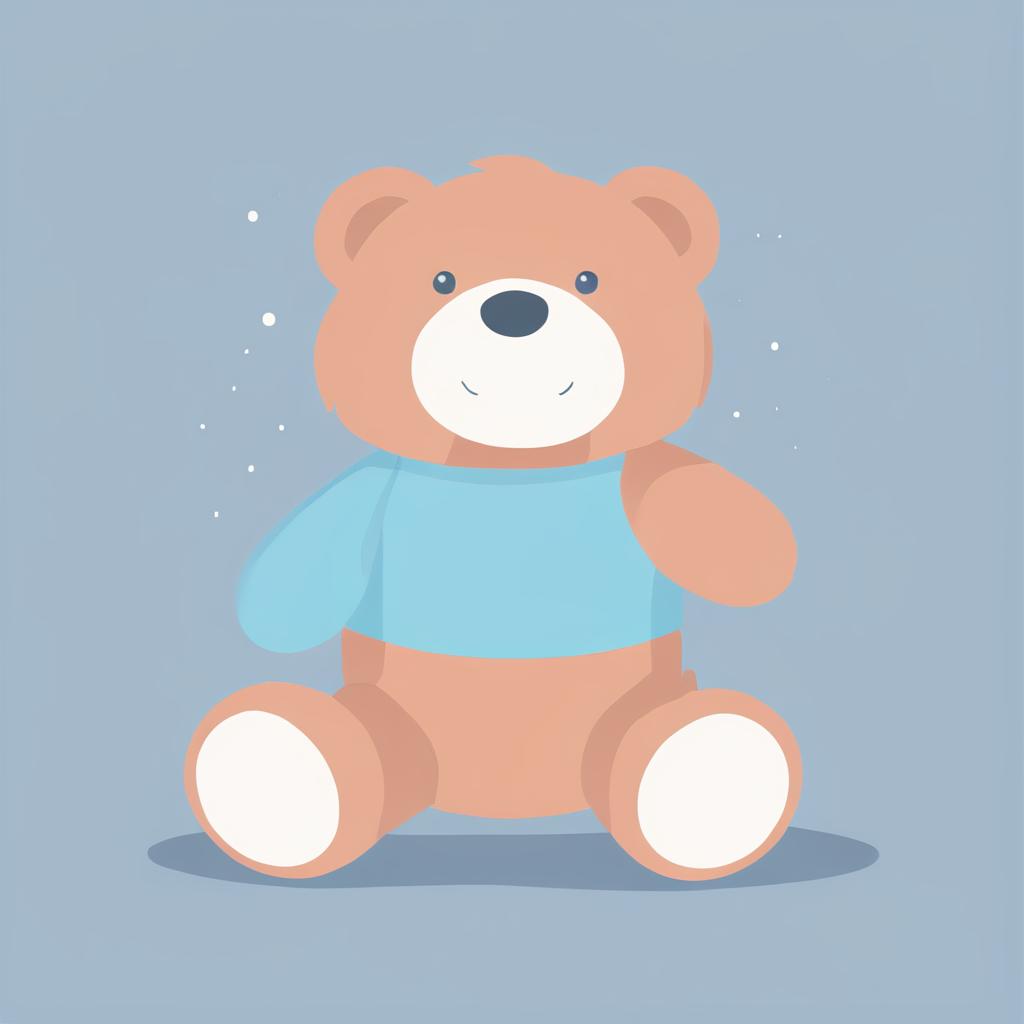} &
    \tabfigure{width=0.1\textwidth}{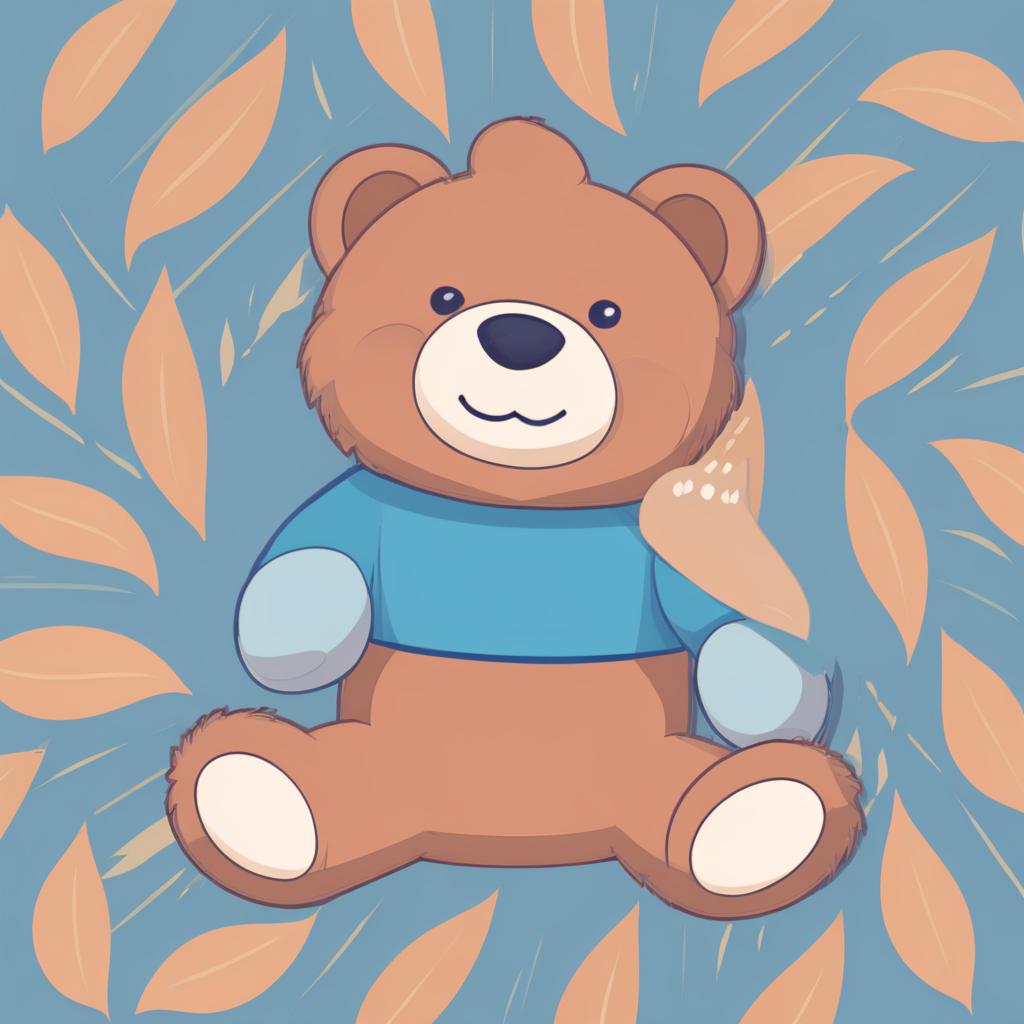} 
    \\
    &
    \tabfigure{width=0.1\textwidth}{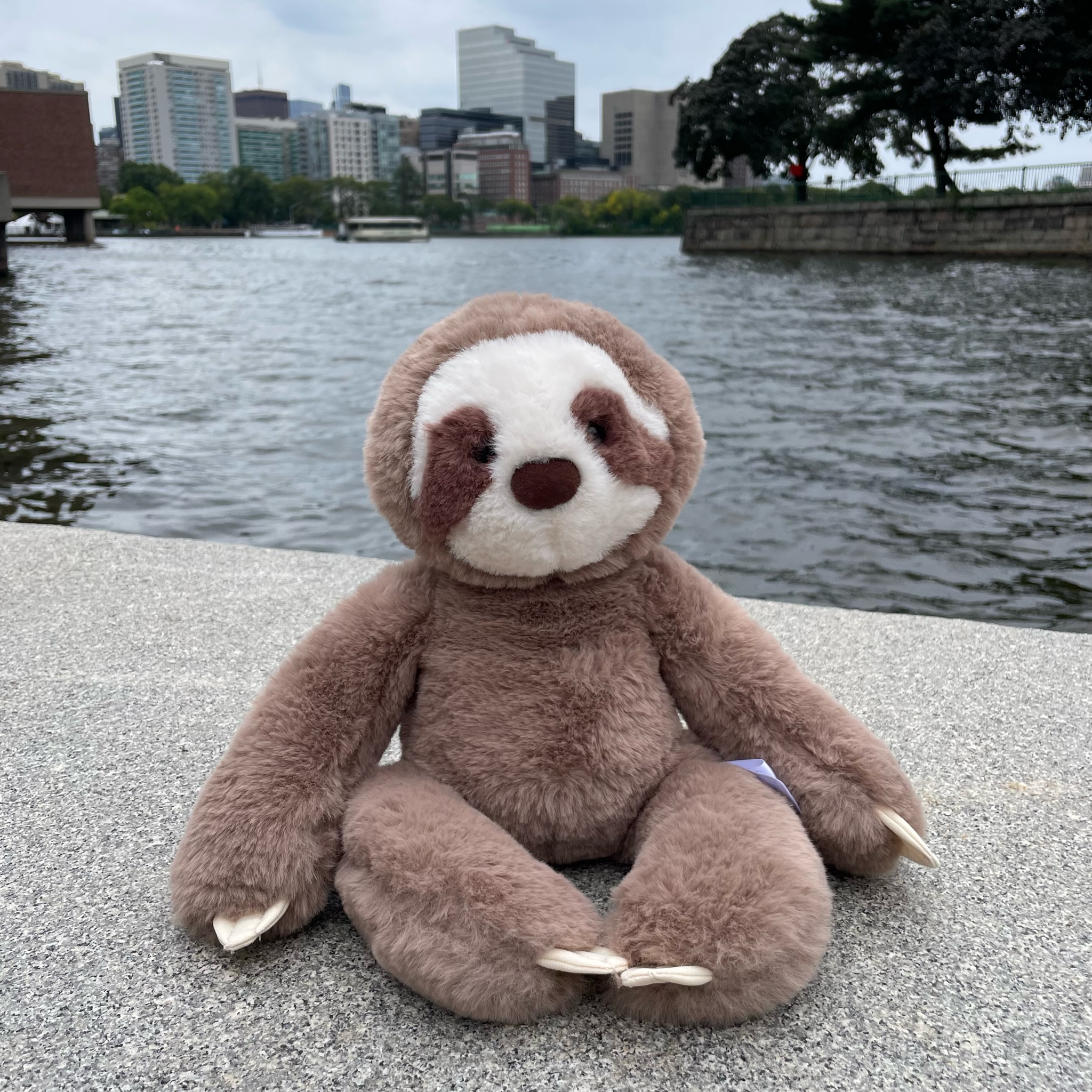} &
    \fcolorbox{red}{white}{\tabfigure{width=0.096\textwidth}{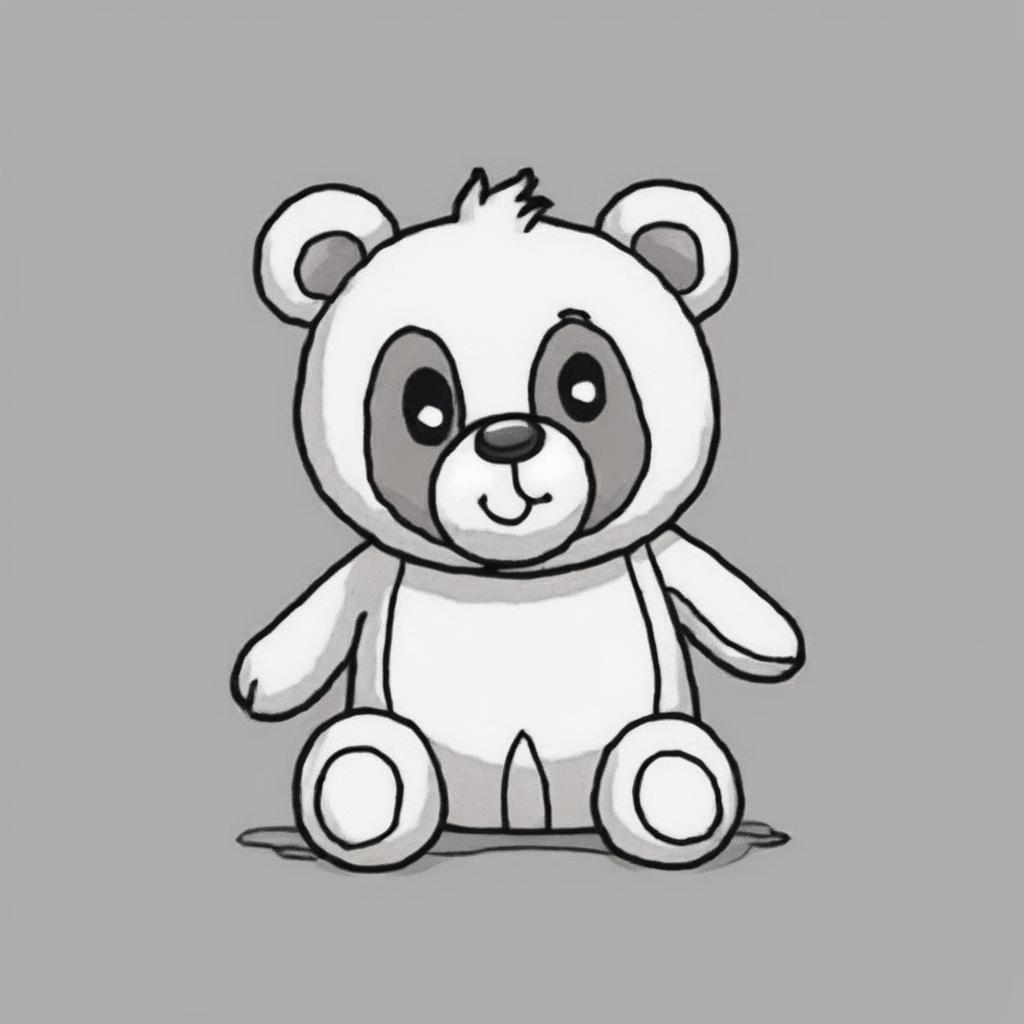}} &
    \tabfigure{width=0.1\textwidth}{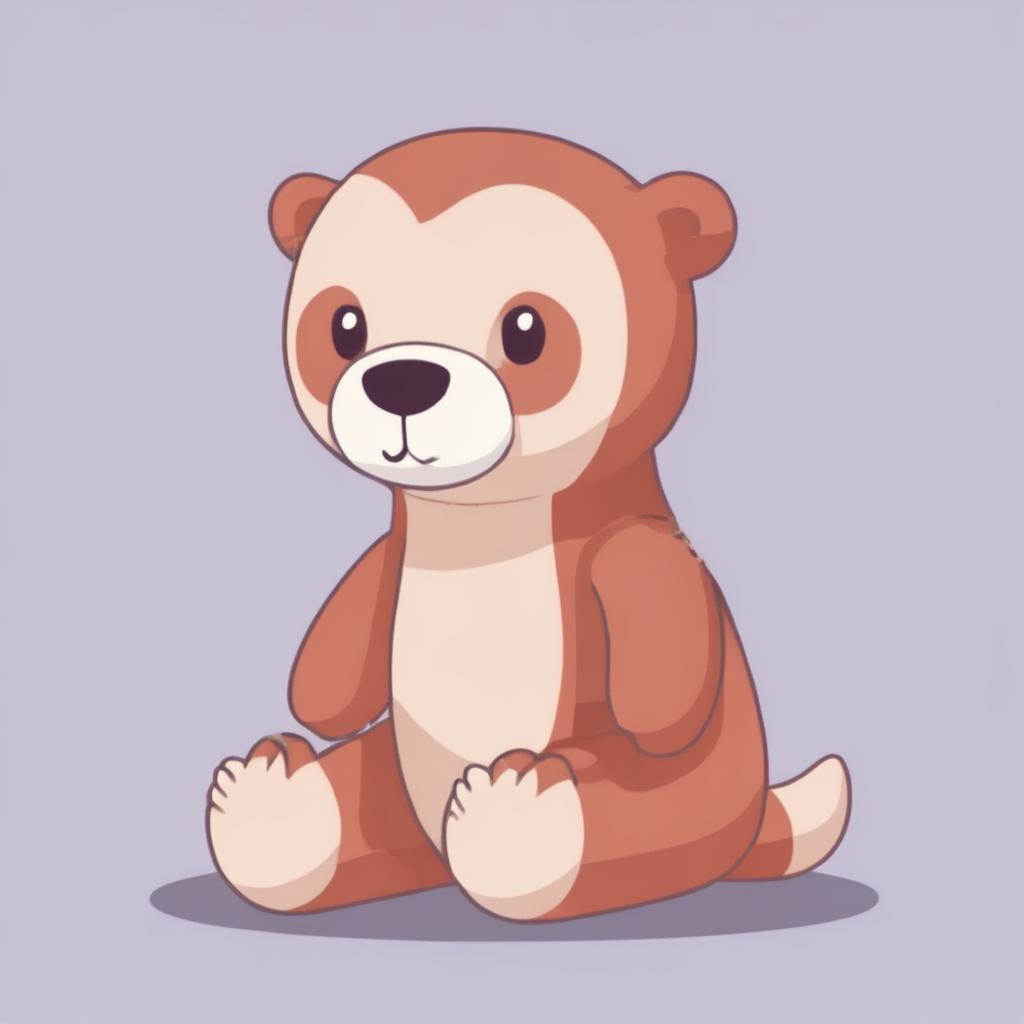} &
    {\tabfigure{width=0.1\textwidth}{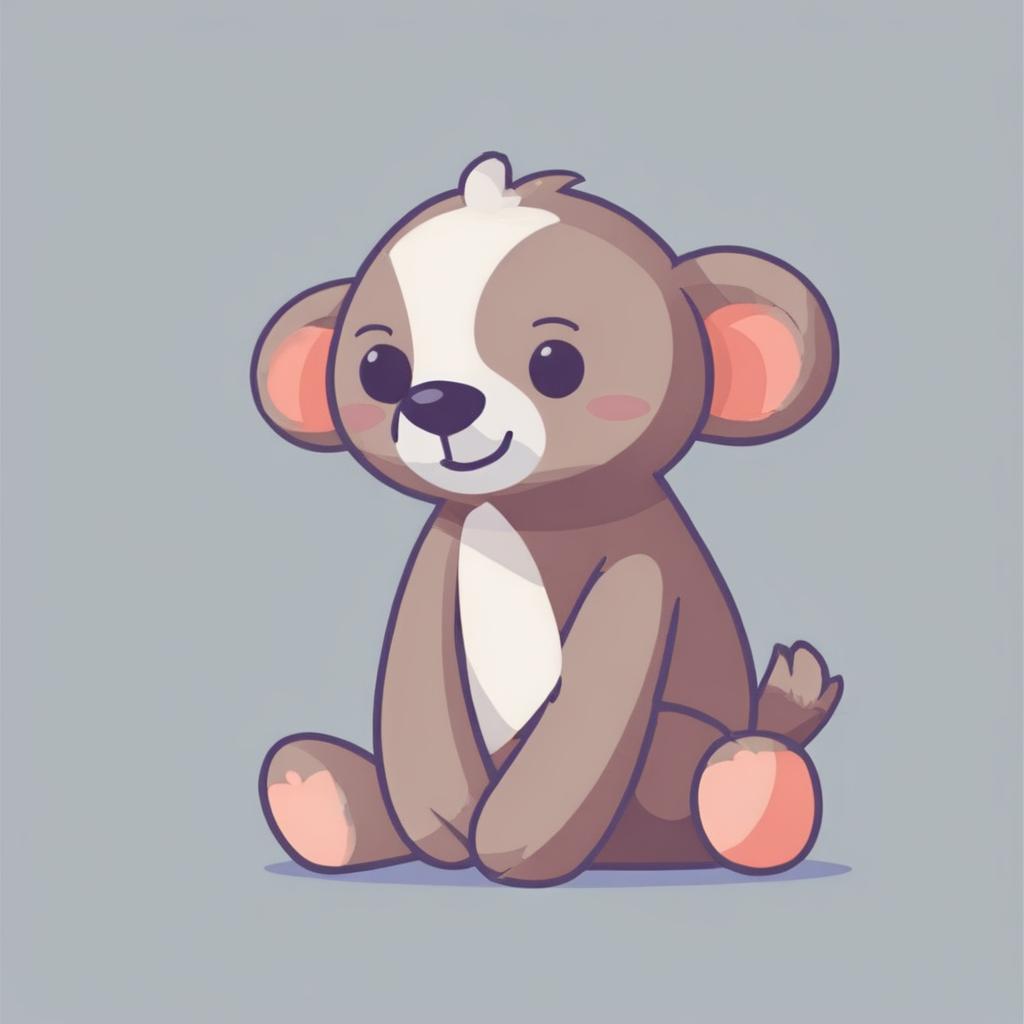}} &
    \fcolorbox{red}{white}{\tabfigure{width=0.096\textwidth}{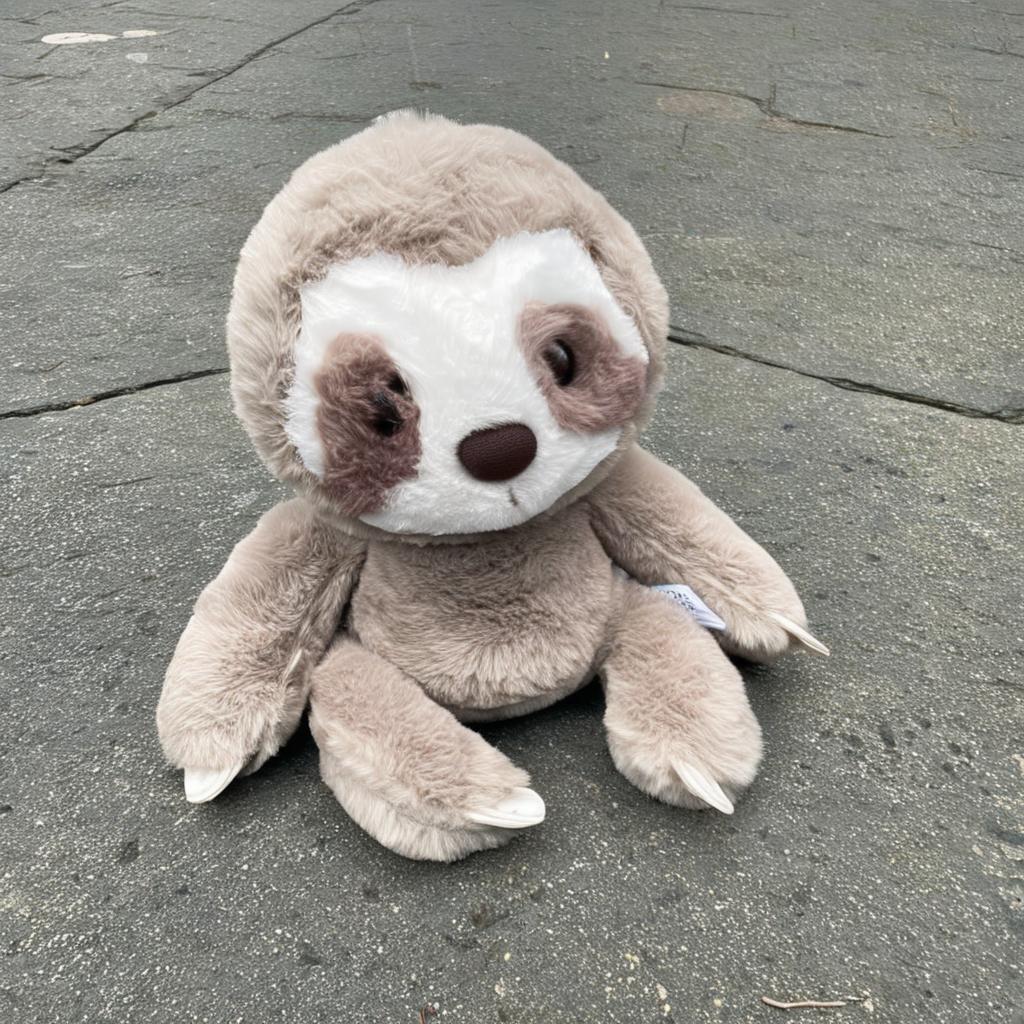}} &
        \tabfigure{width=0.1\textwidth}{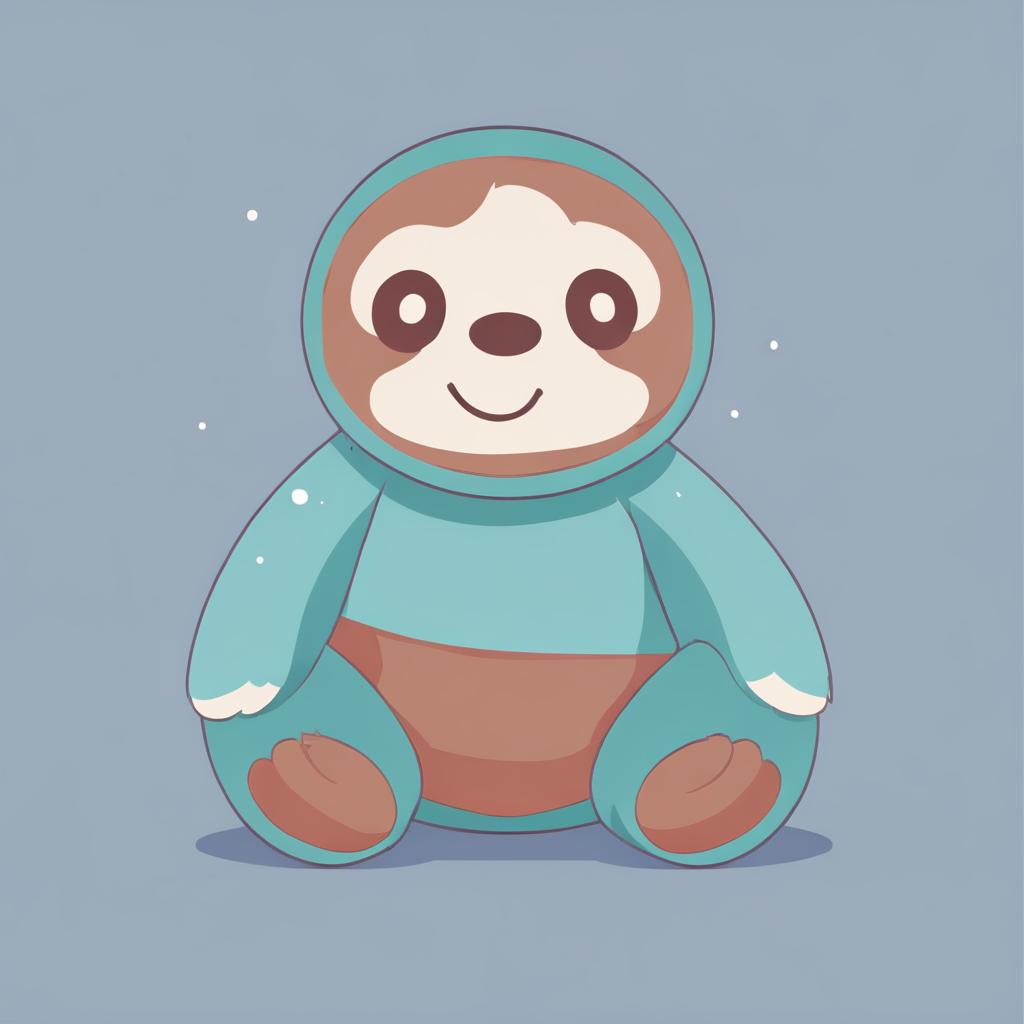} &
    \fcolorbox{ForestGreen}{white}{\tabfigure{width=0.096\textwidth}{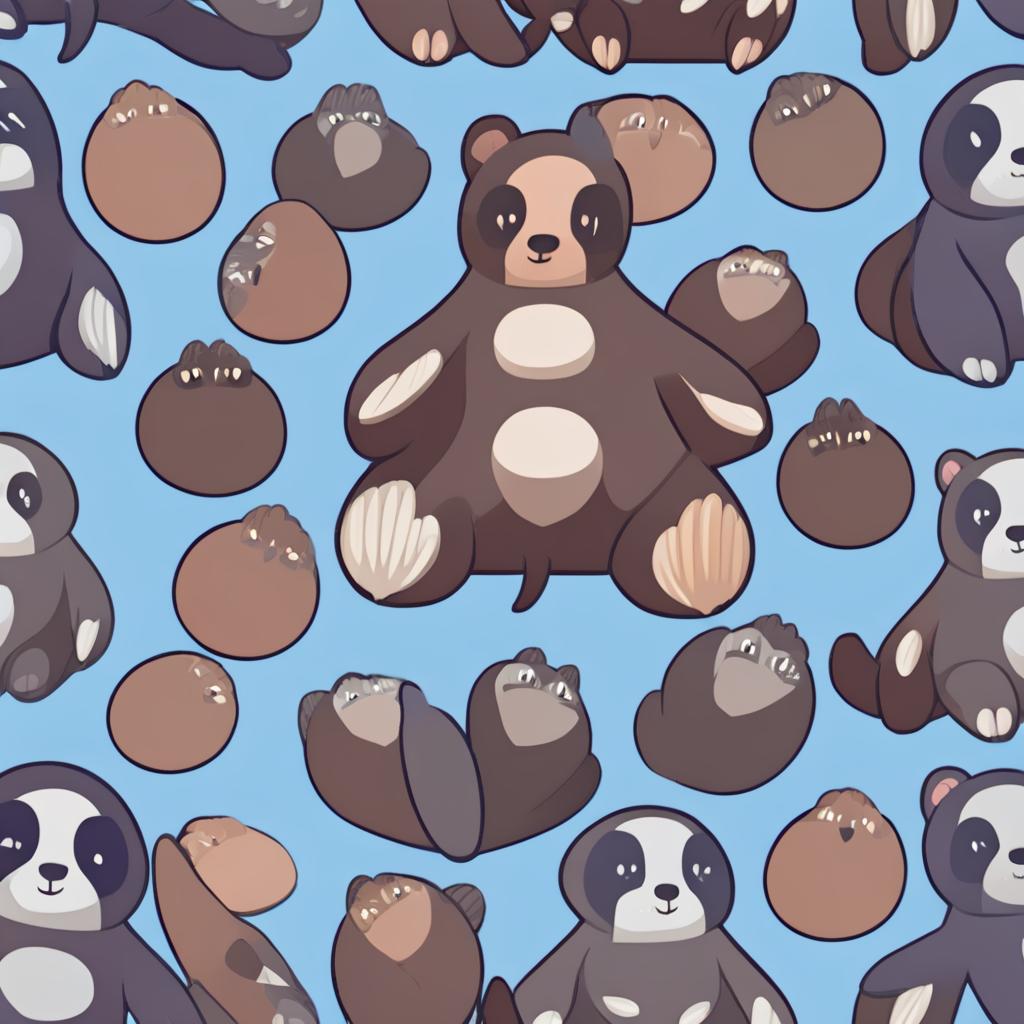}} 
    \\

\cline{2-8}
&
\raisebox{-8mm}
{\begin{tikzpicture}
    \draw[->] (0, -1.5) -- (1.5, -1.5) node[below, midway] {Styles};
    \draw[->] (0, -1.5) -- (0, -3) node[right, midway] {Contents};
\end{tikzpicture}}
     &
    \tabfigure{width=0.1\textwidth}{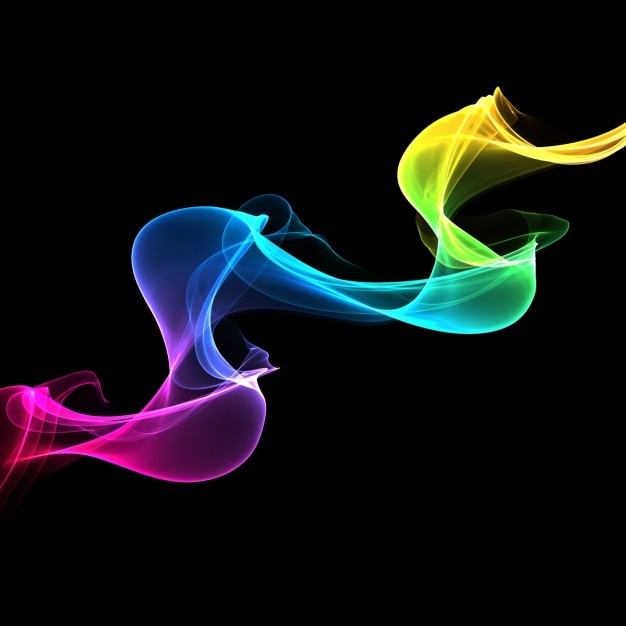} &
    \tabfigure{width=0.1\textwidth}{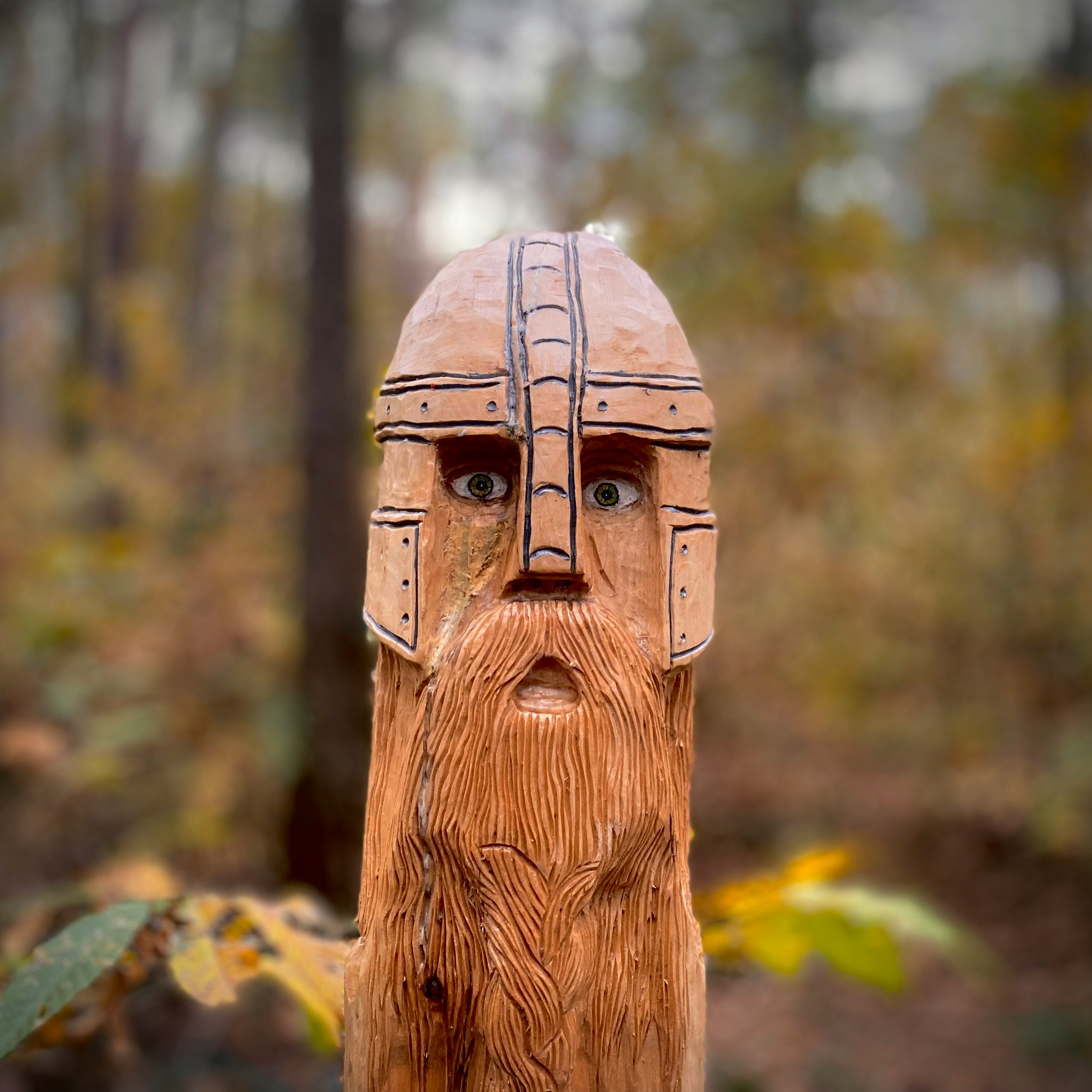} &
    \tabfigure{width=0.1\textwidth}{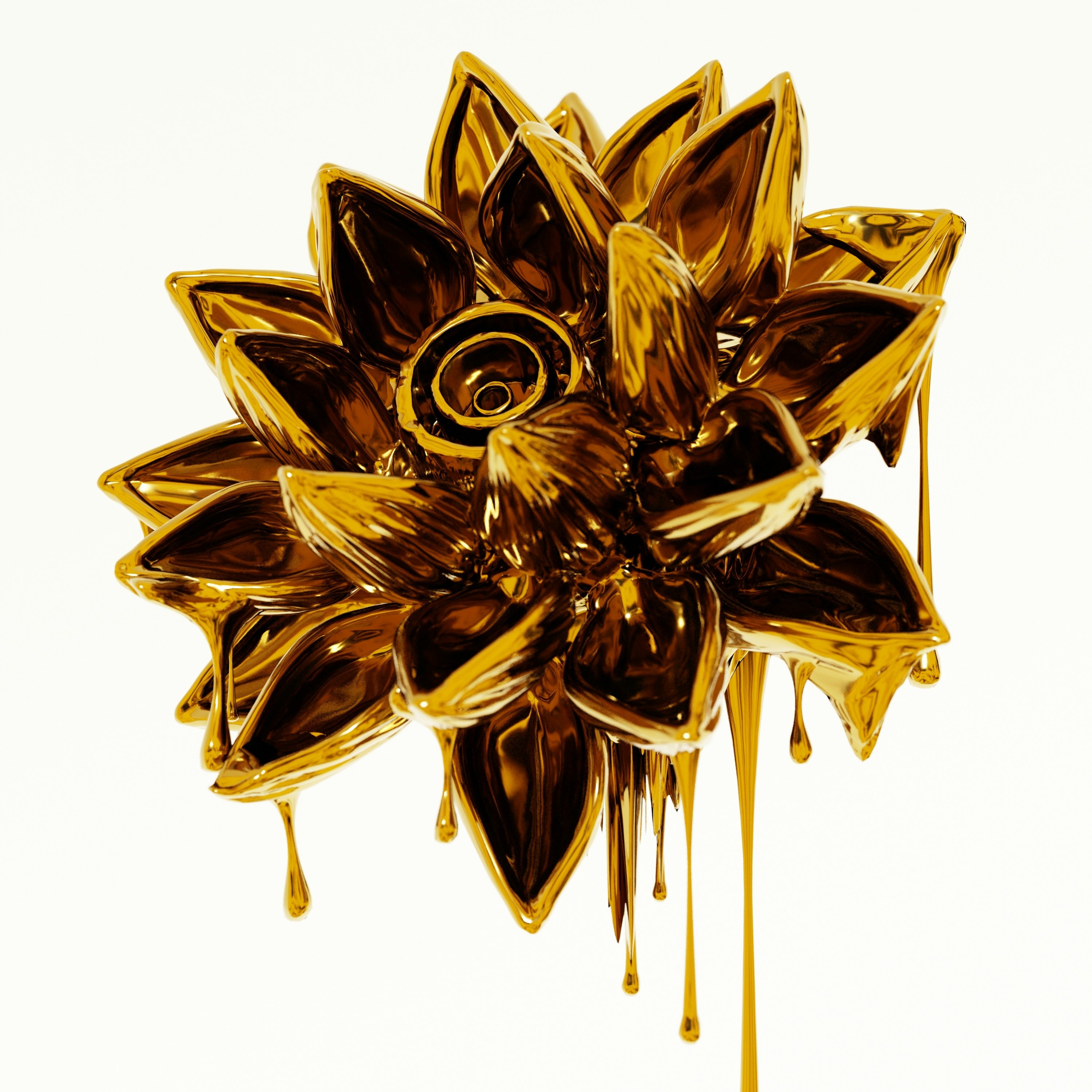}
    & \tabfigure{width=0.1\textwidth}{figures/new_splits/rainbow.jpg} &
    \tabfigure{width=0.1\textwidth}{figures/new_splits/wood.jpg} &
    \tabfigure{width=0.1\textwidth}{figures/new_splits/melting.jpg}
    \\
    
    &
    \tabfigure{width=0.1\textwidth}{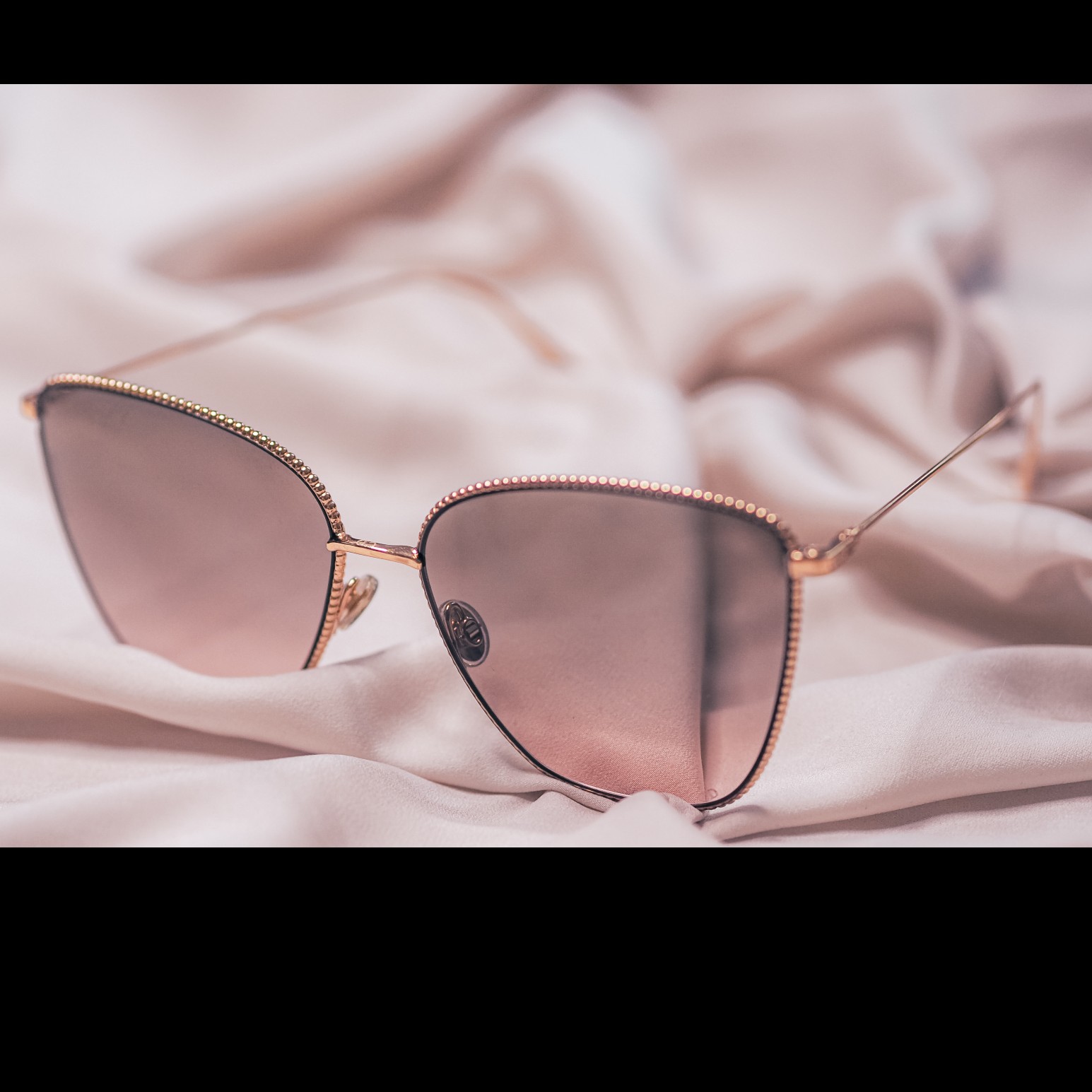} &
    \tabfigure{width=0.1\textwidth}{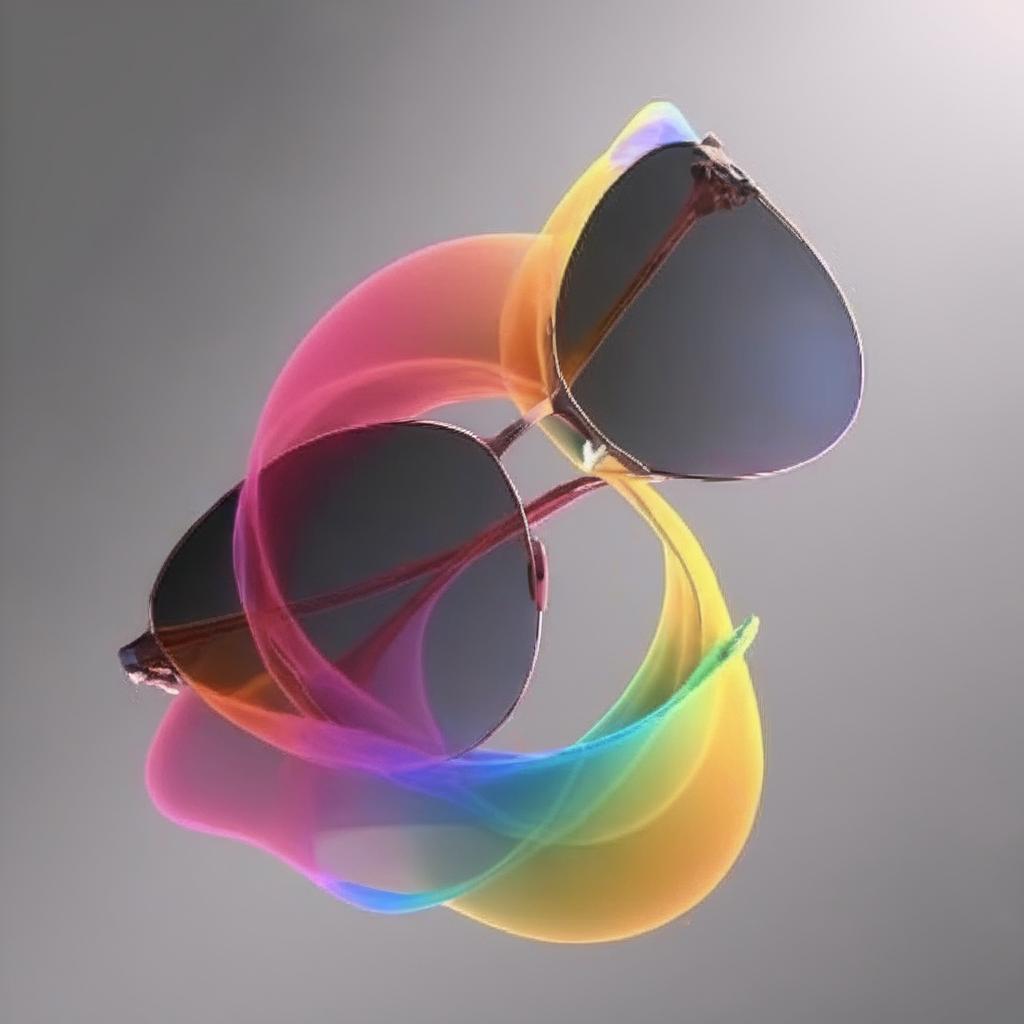} &
    \tabfigure{width=0.1\textwidth}{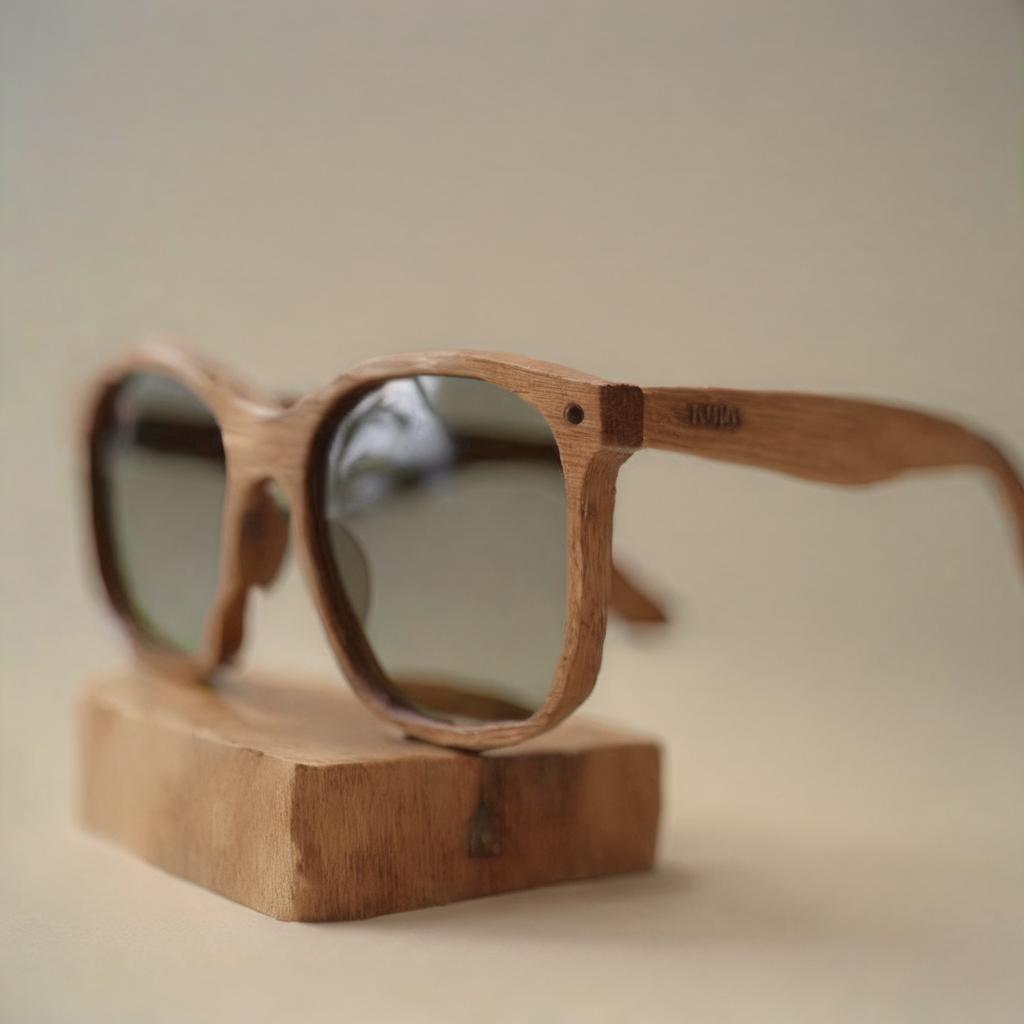} &
    \fcolorbox{red}{white}{\tabfigure{width=0.096\textwidth}{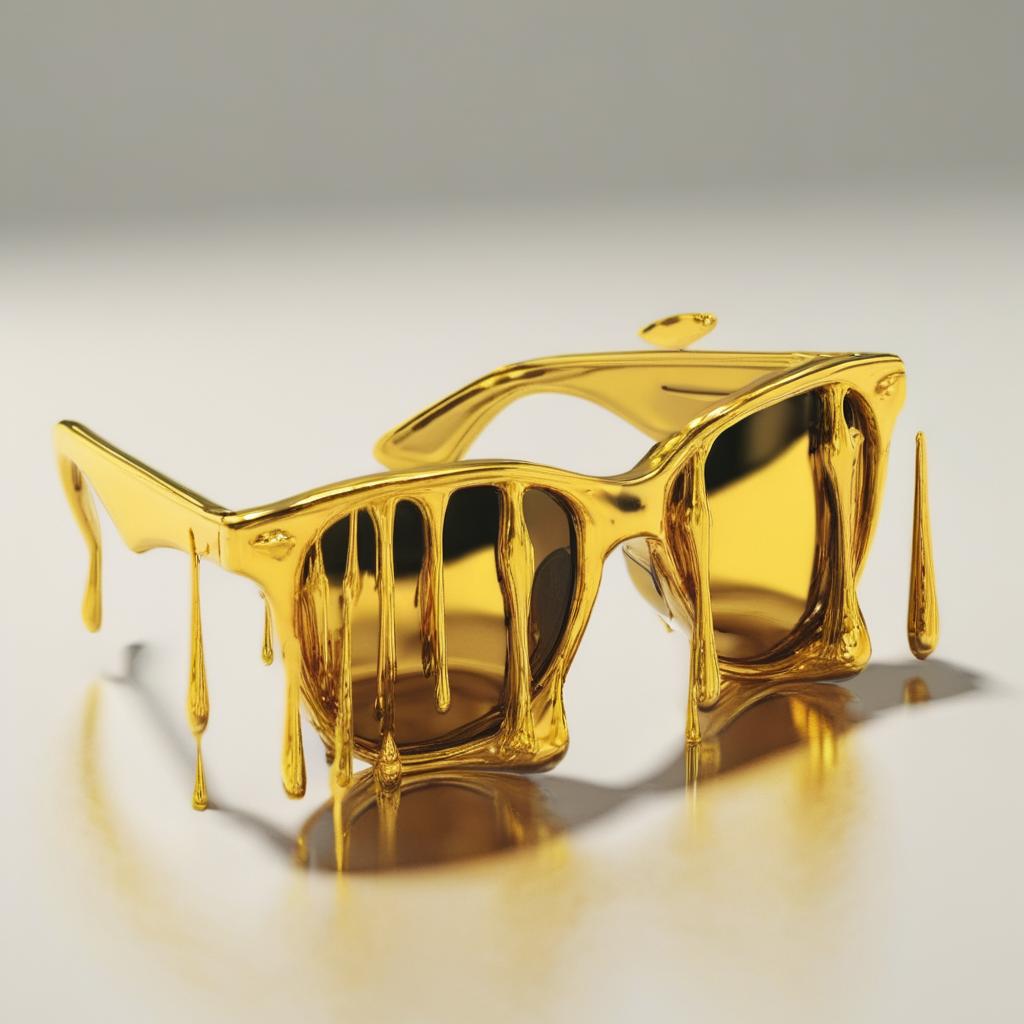}} &
    \tabfigure{width=0.1\textwidth}{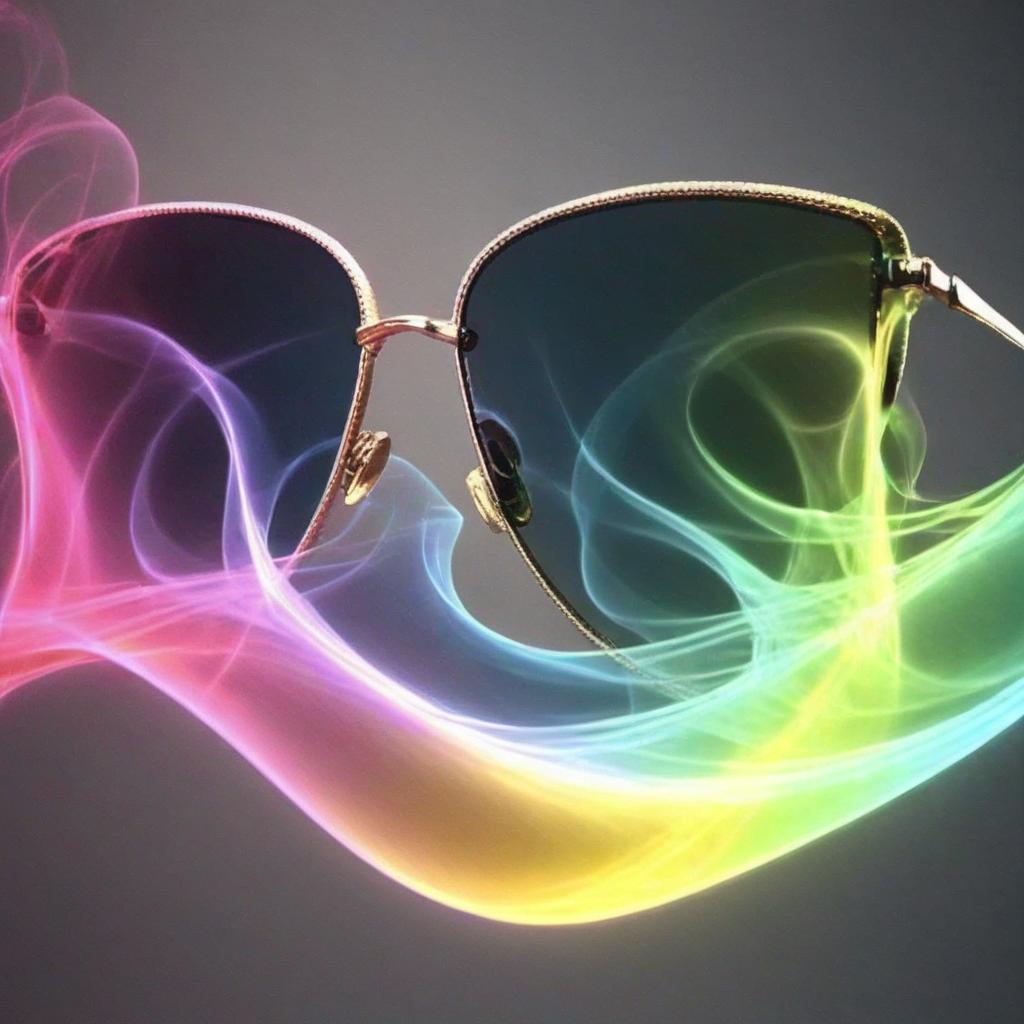} &
    \tabfigure{width=0.1\textwidth}{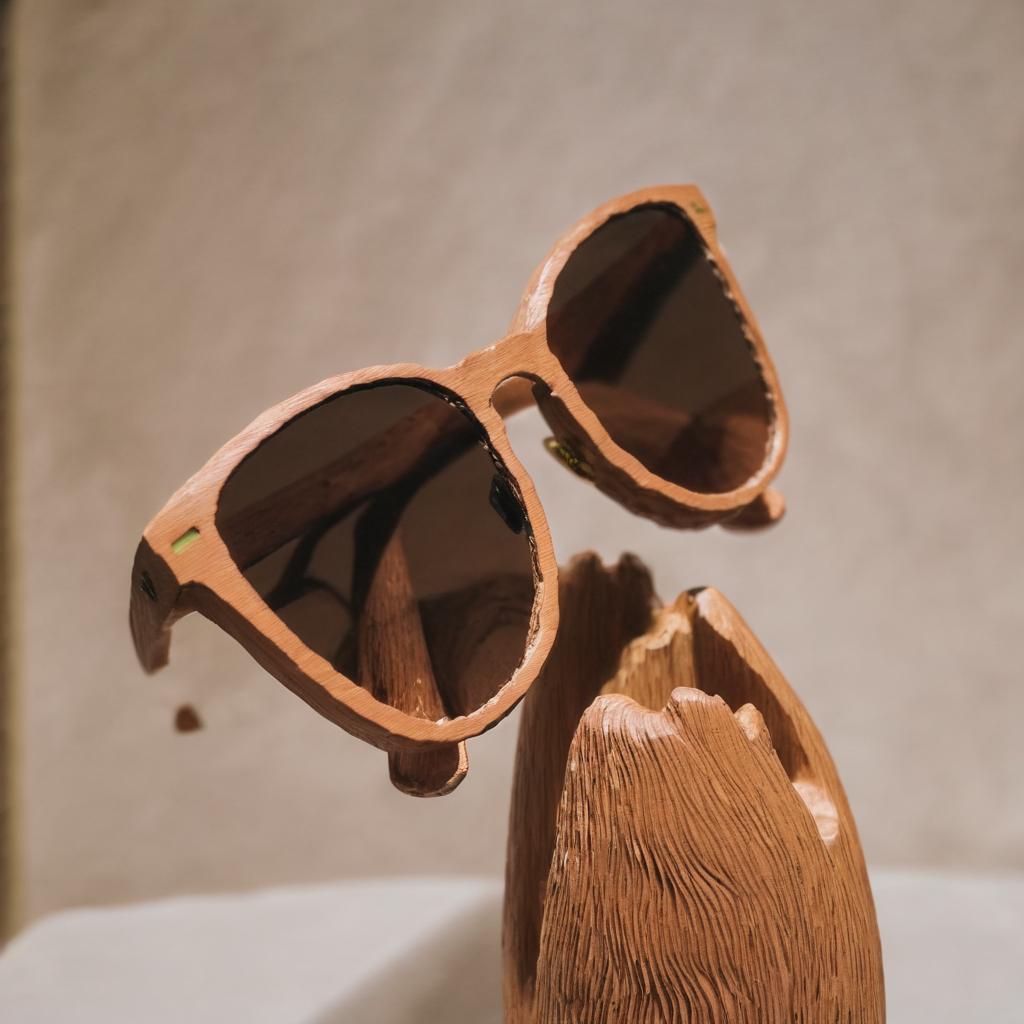} &
    \fcolorbox{red}{white}{\tabfigure{width=0.096\textwidth}{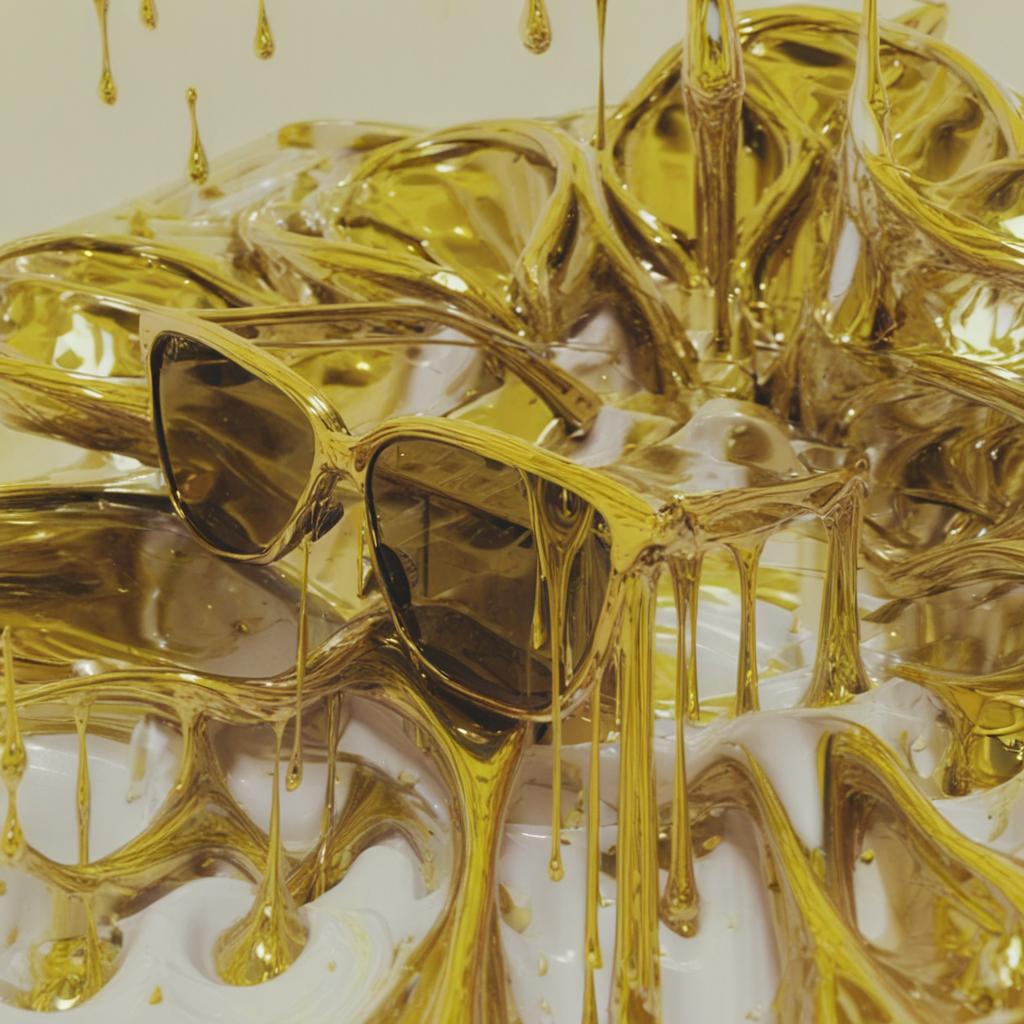}} 
    \\    
\textbf{(2)} &
    \tabfigure{width=0.1\textwidth}{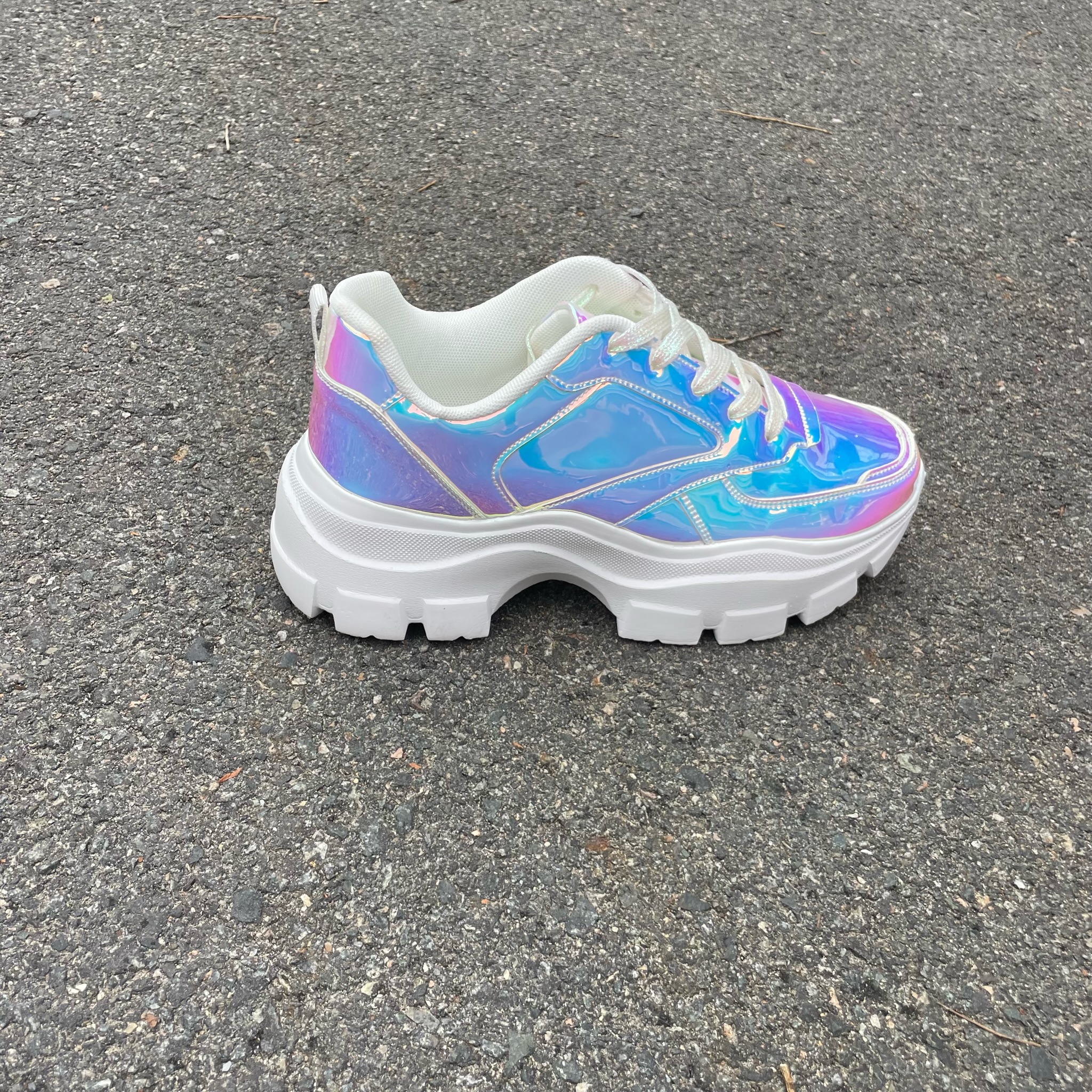} &
    \tabfigure{width=0.1\textwidth}{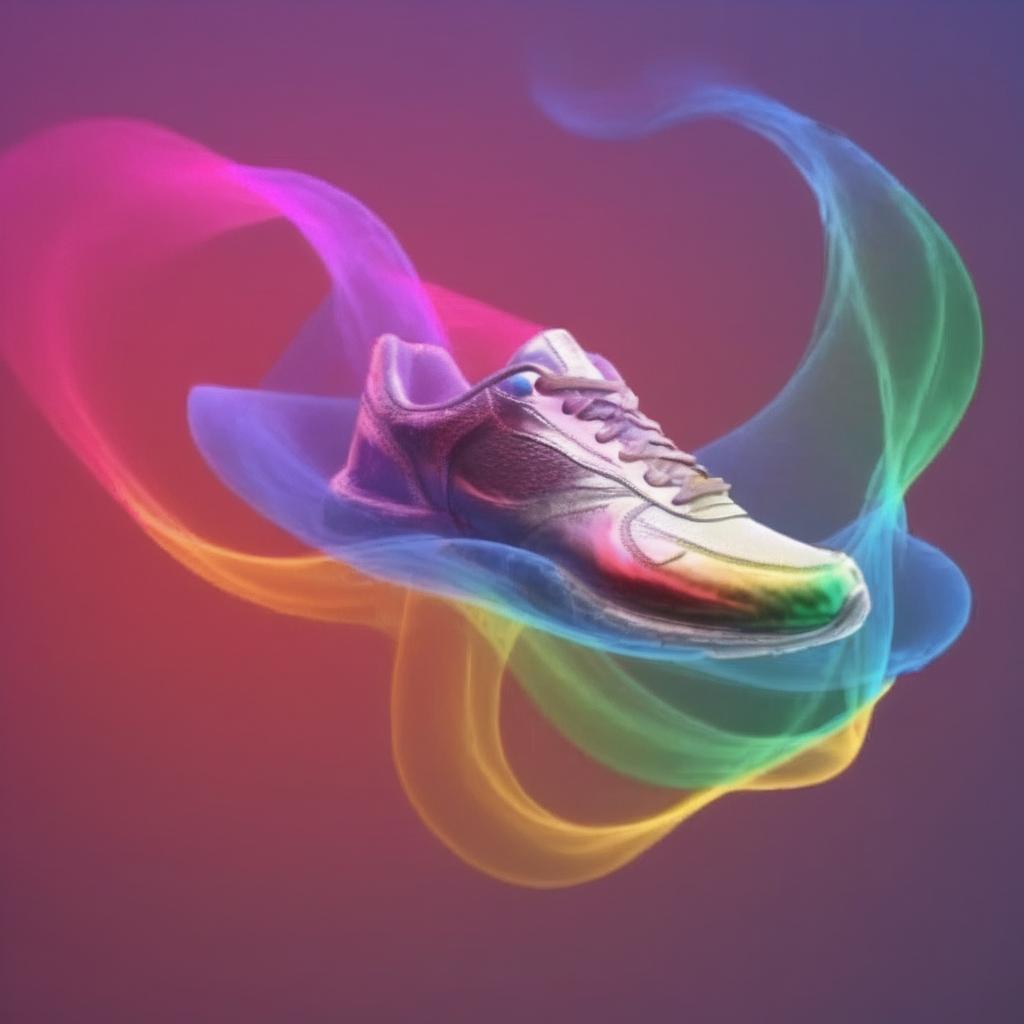} &
    \tabfigure{width=0.1\textwidth}{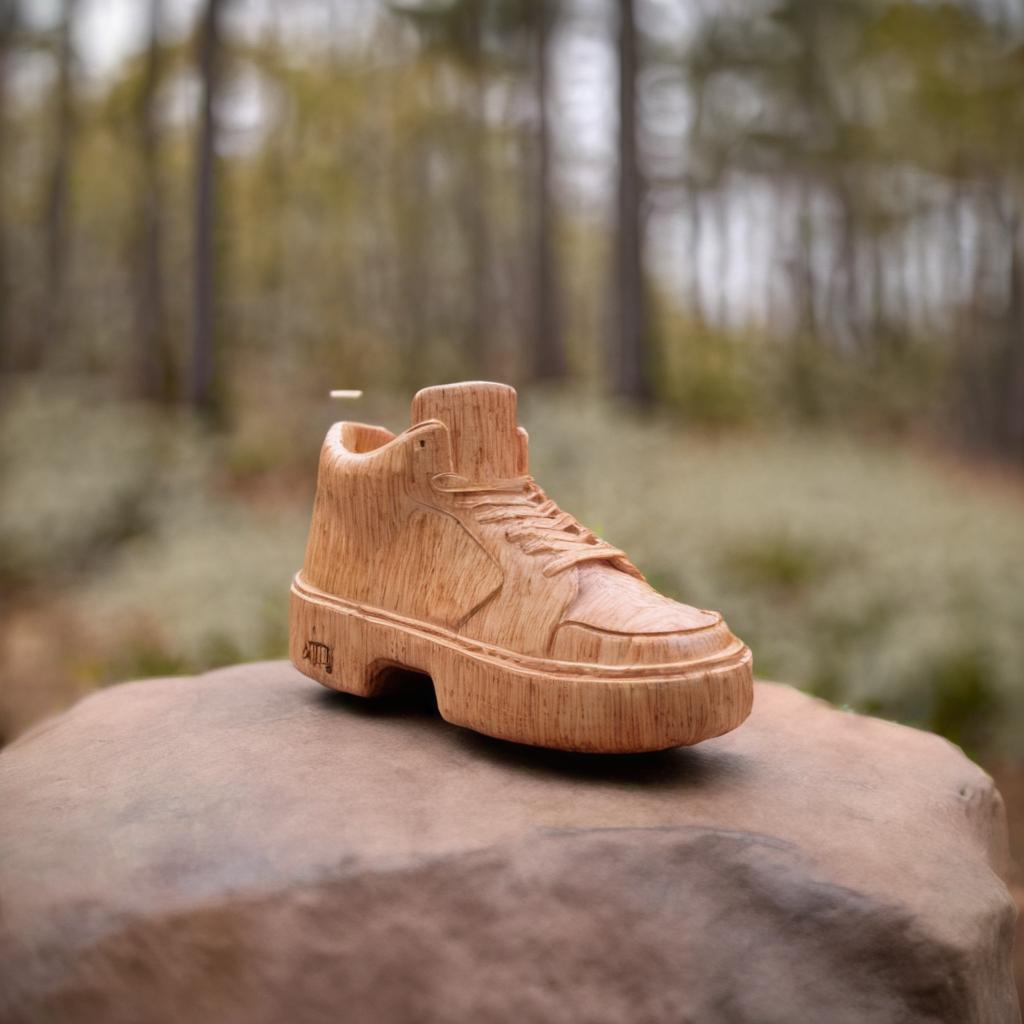} &
    \fcolorbox{red}{white}{\tabfigure{width=0.096\textwidth}{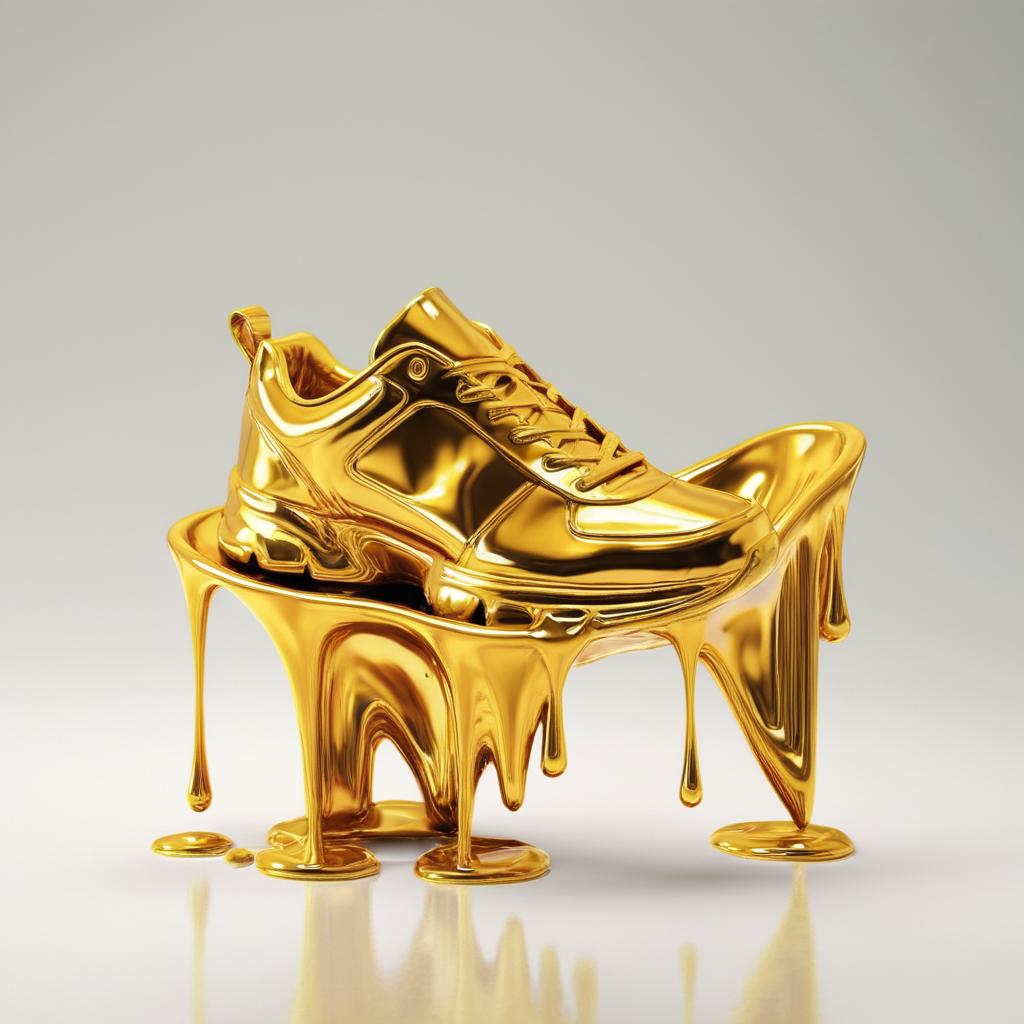}} &
    \fcolorbox{yellow}{white}{\tabfigure{width=0.096\textwidth}{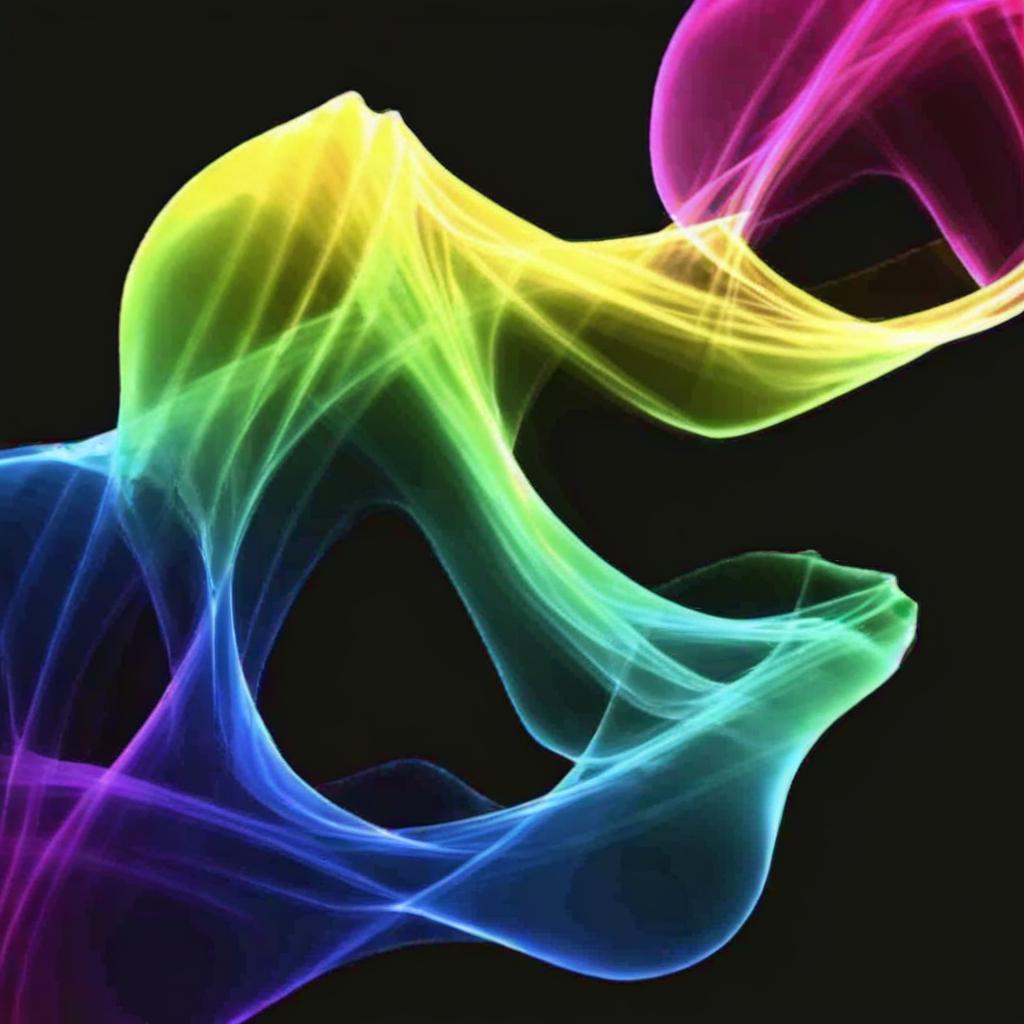}} &
    \tabfigure{width=0.1\textwidth}{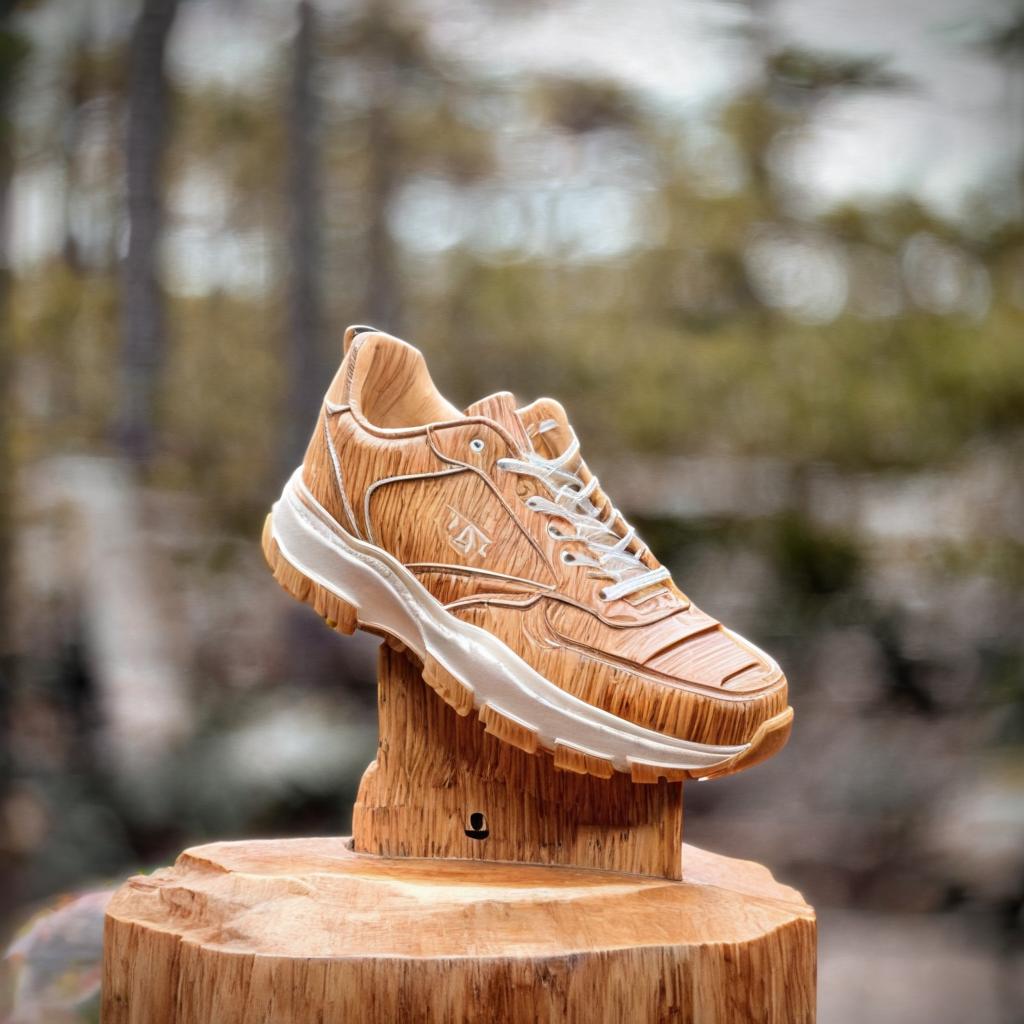} &
    \fcolorbox{red}{white}{\tabfigure{width=0.096\textwidth}{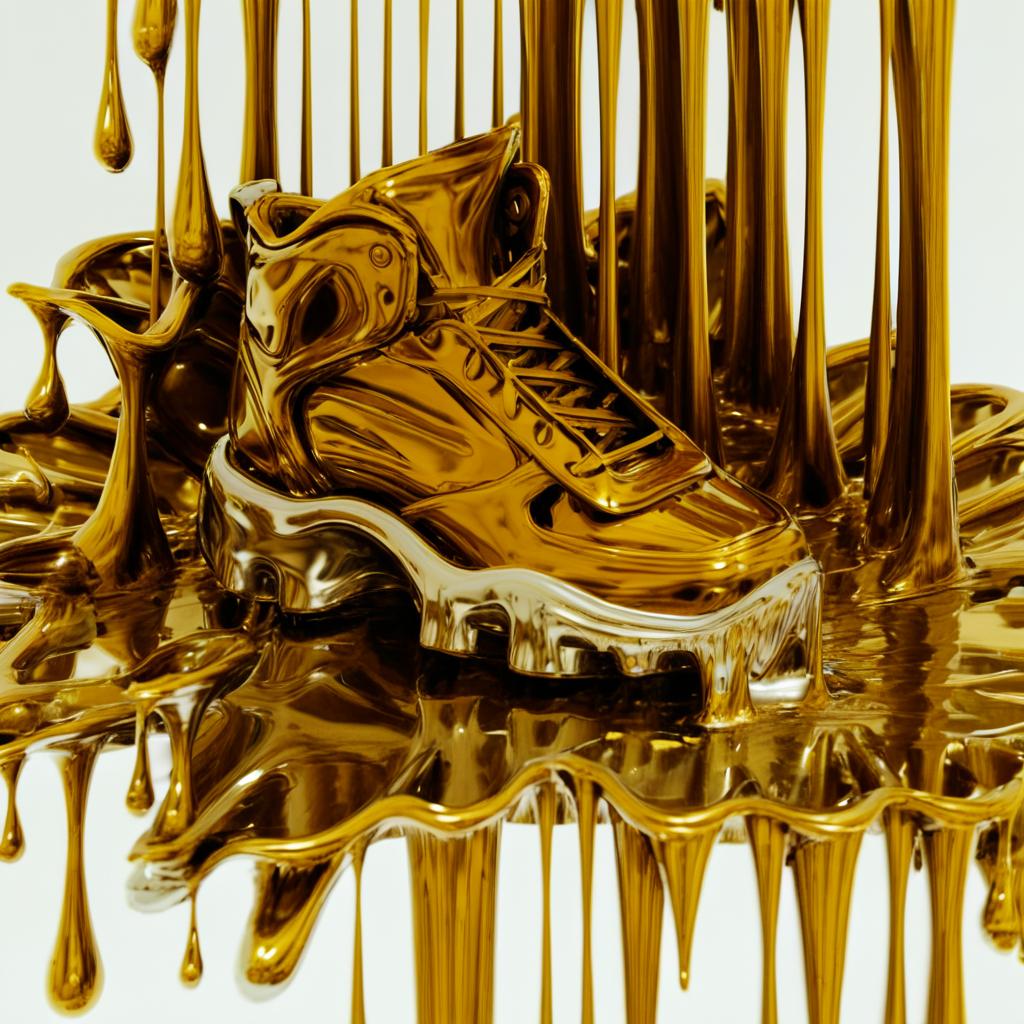}}
    \\
&
    \tabfigure{width=0.1\textwidth}{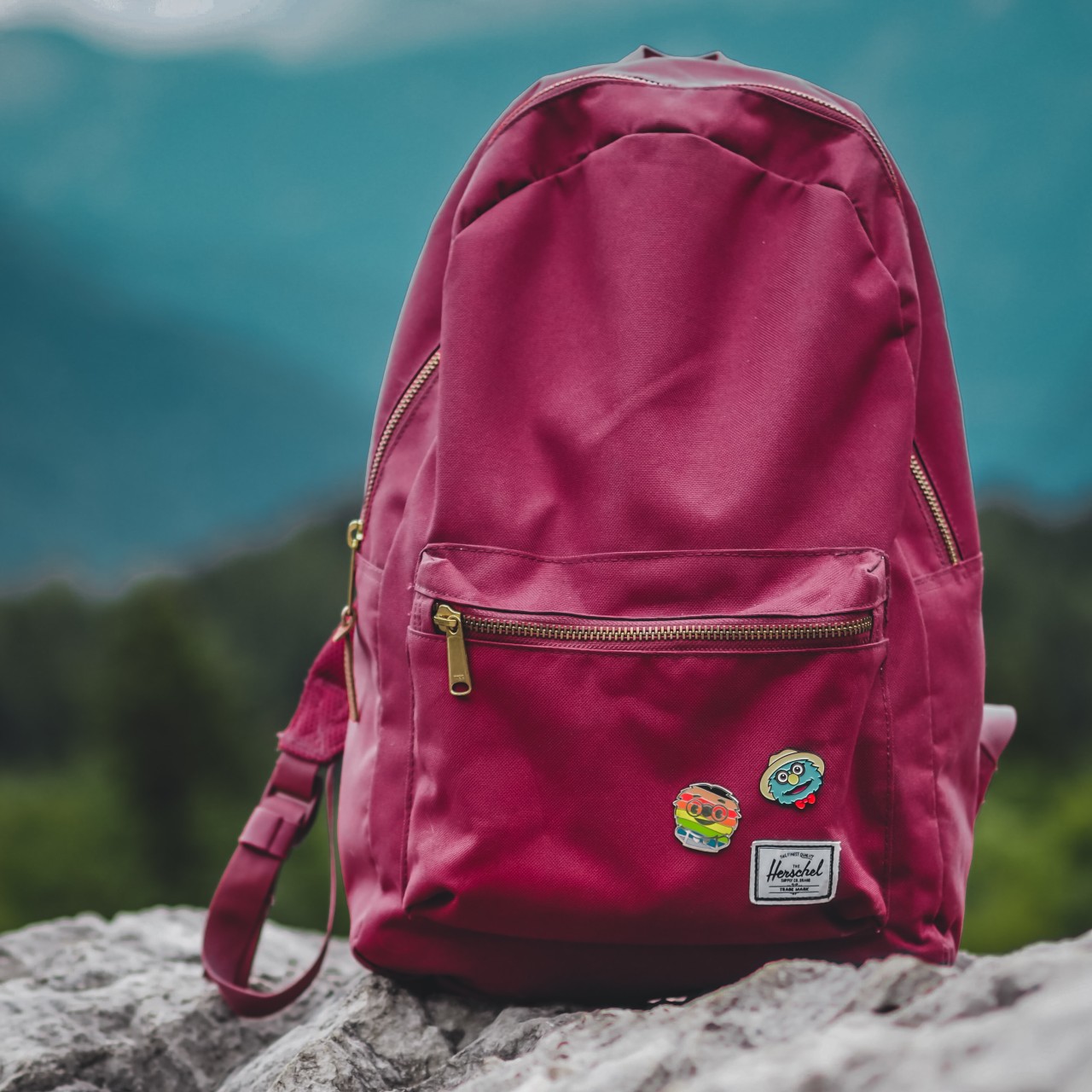} &
    \fcolorbox{red}{white}{\tabfigure{width=0.096\textwidth}{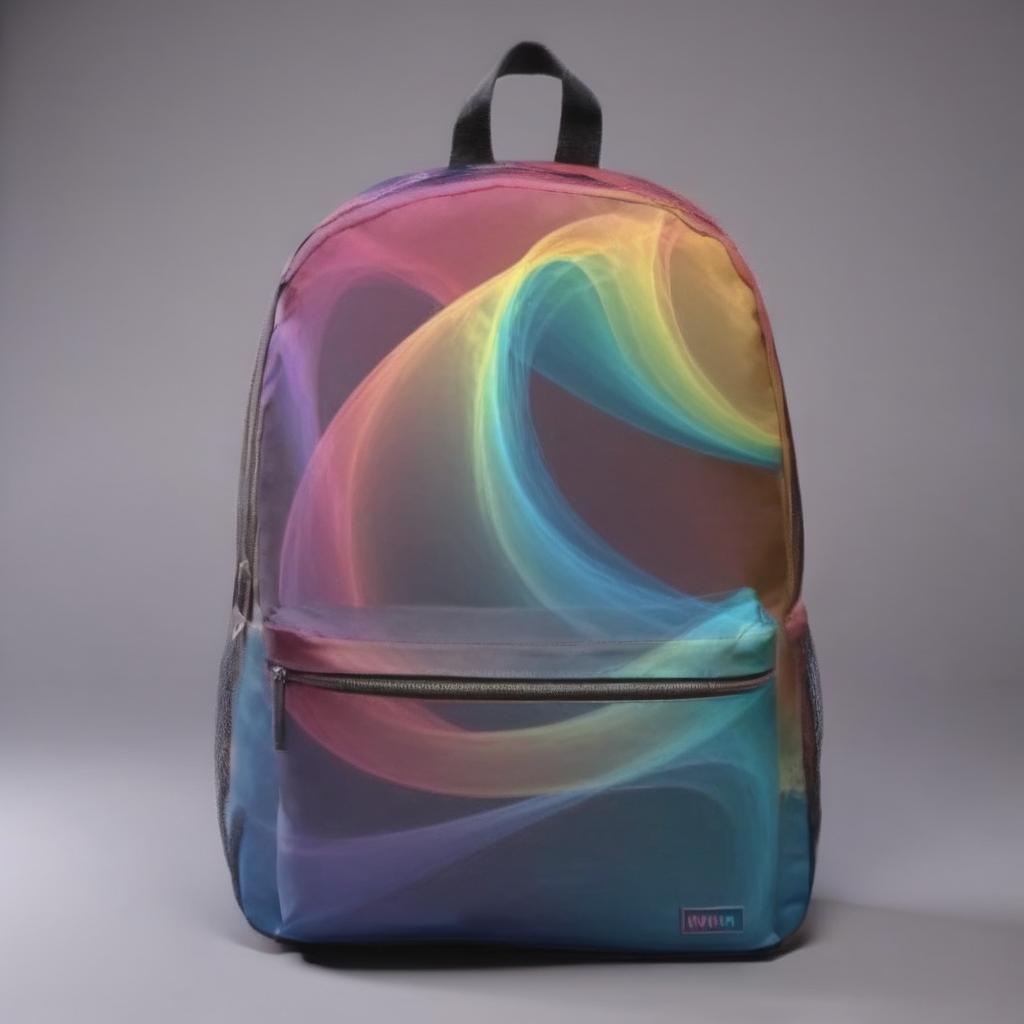}} &
    \tabfigure{width=0.1\textwidth}{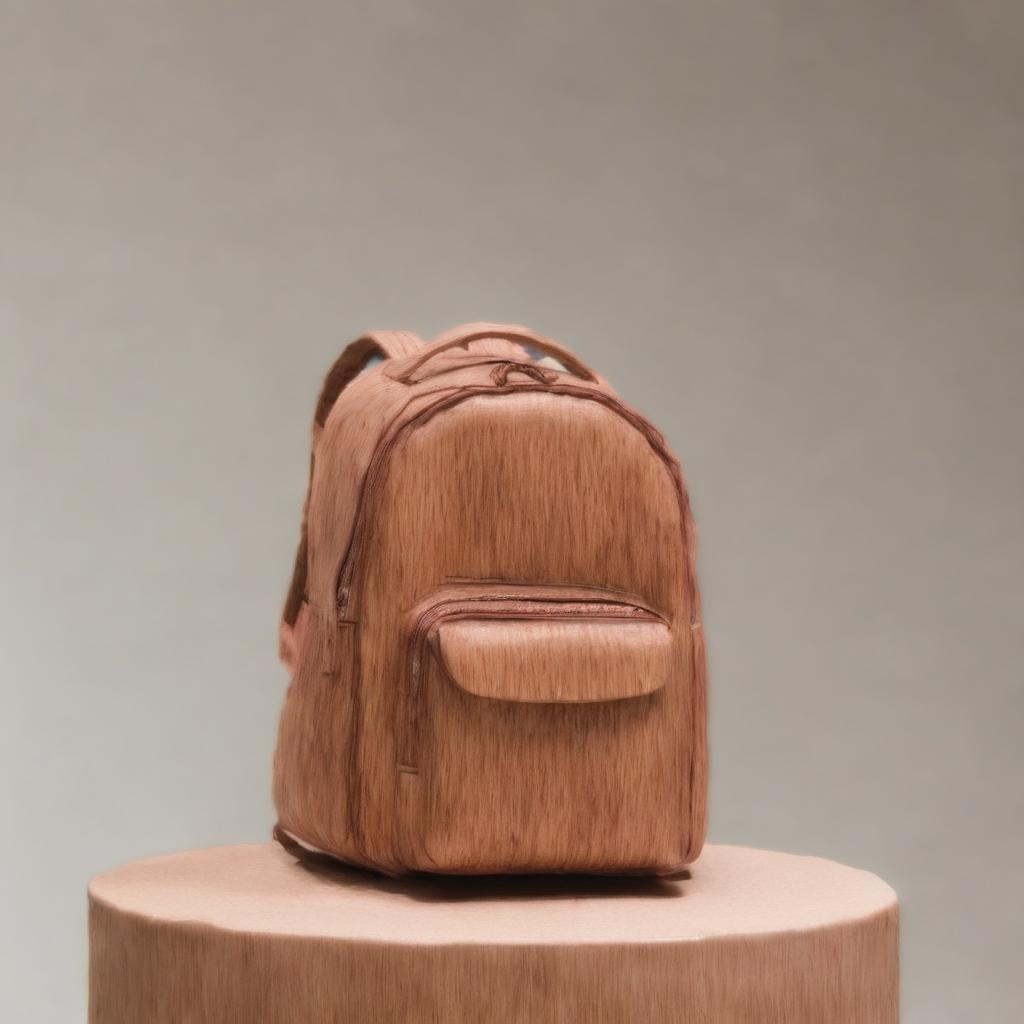} &
    \fcolorbox{red}{white}{\tabfigure{width=0.096\textwidth}{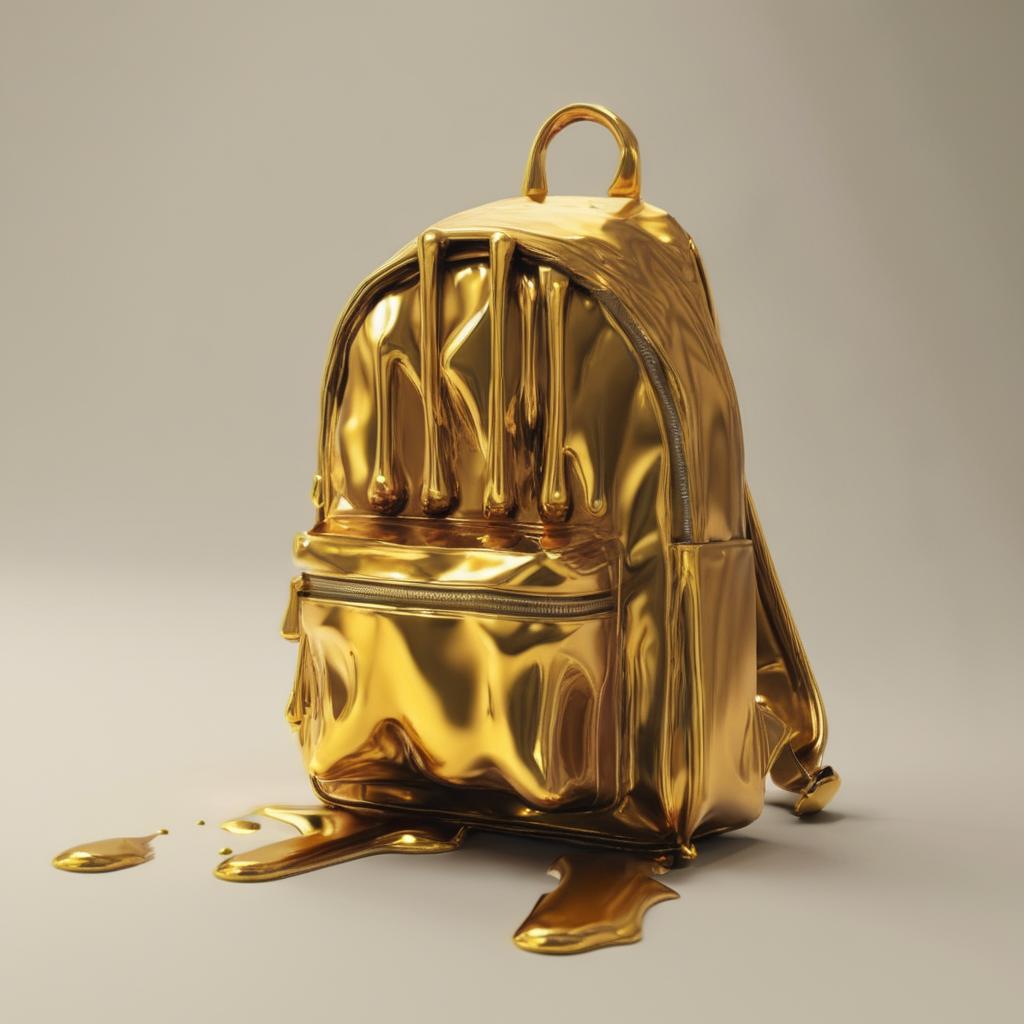}} &
        \fcolorbox{red}{white}{\tabfigure{width=0.096\textwidth}{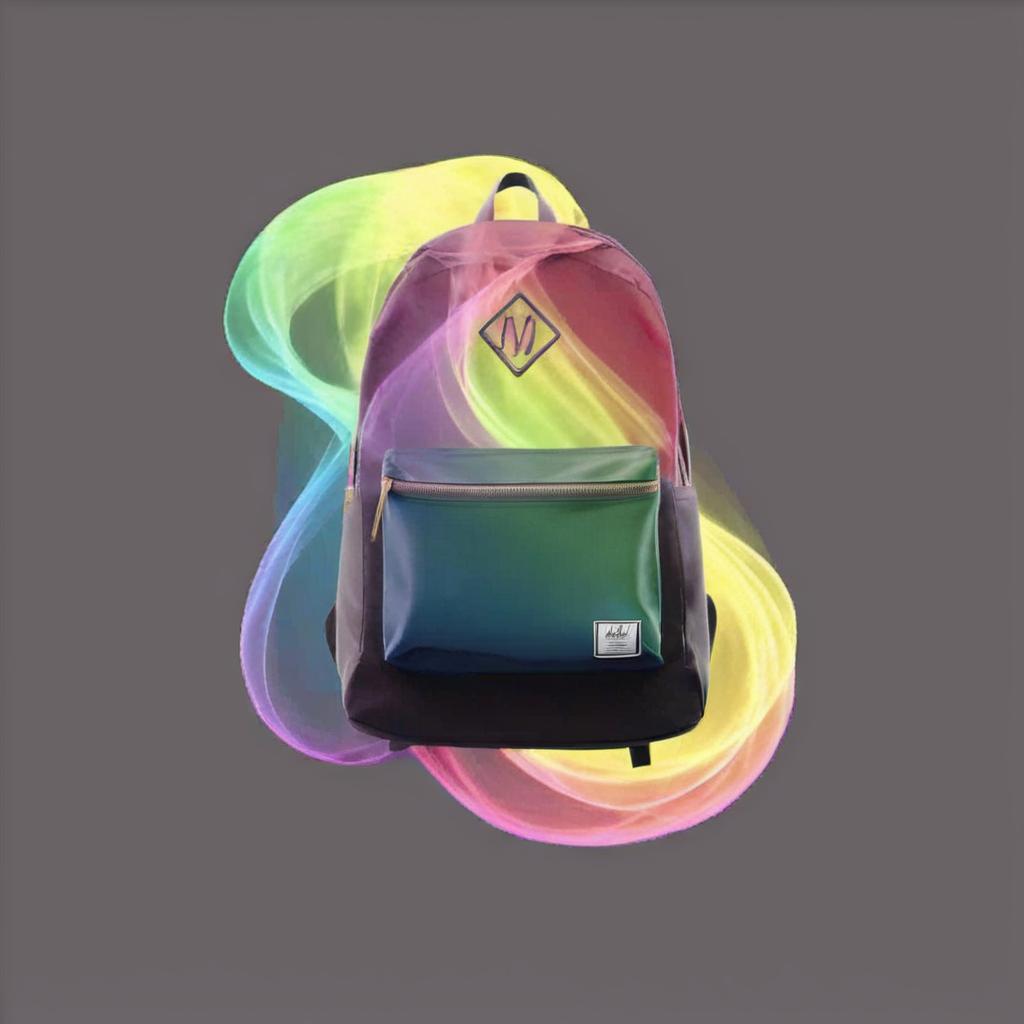}} &
    \tabfigure{width=0.1\textwidth}{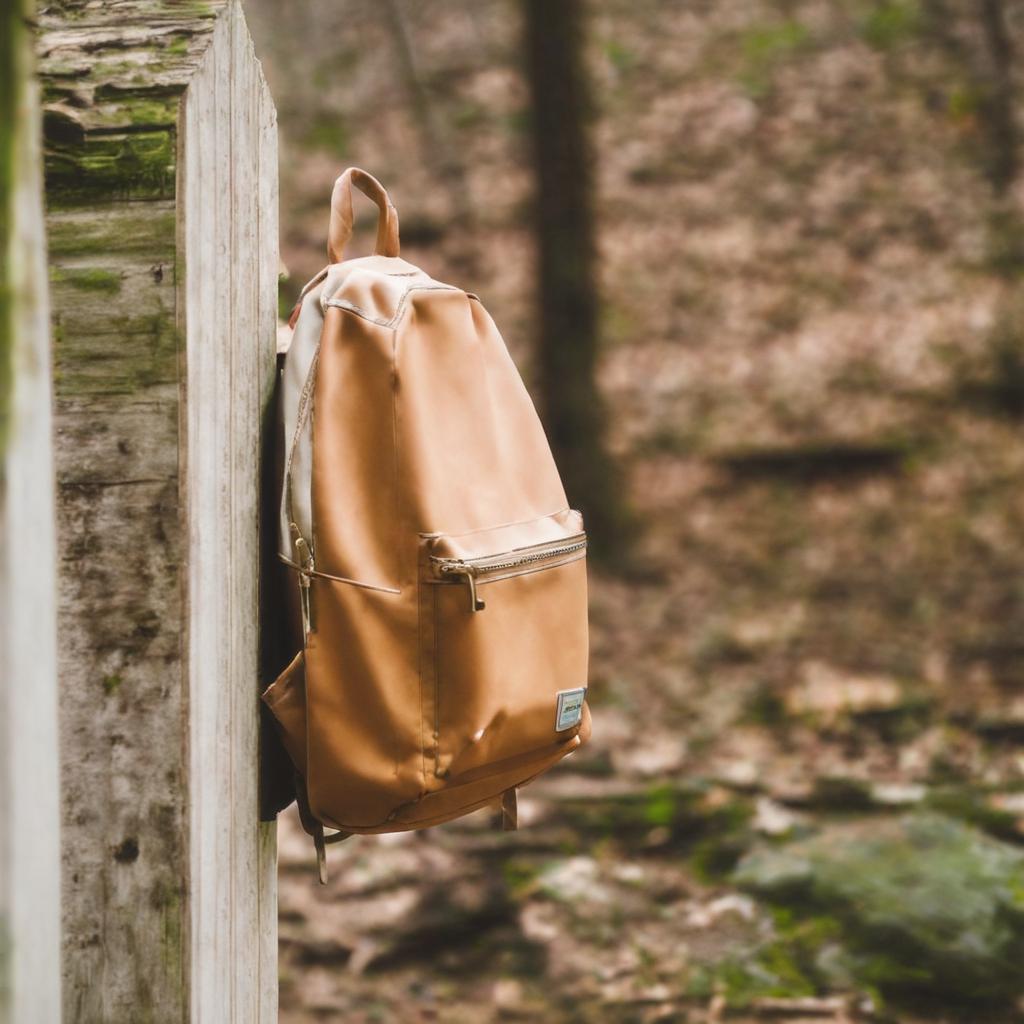} &
    \fcolorbox{red}{white}{\tabfigure{width=0.096\textwidth}{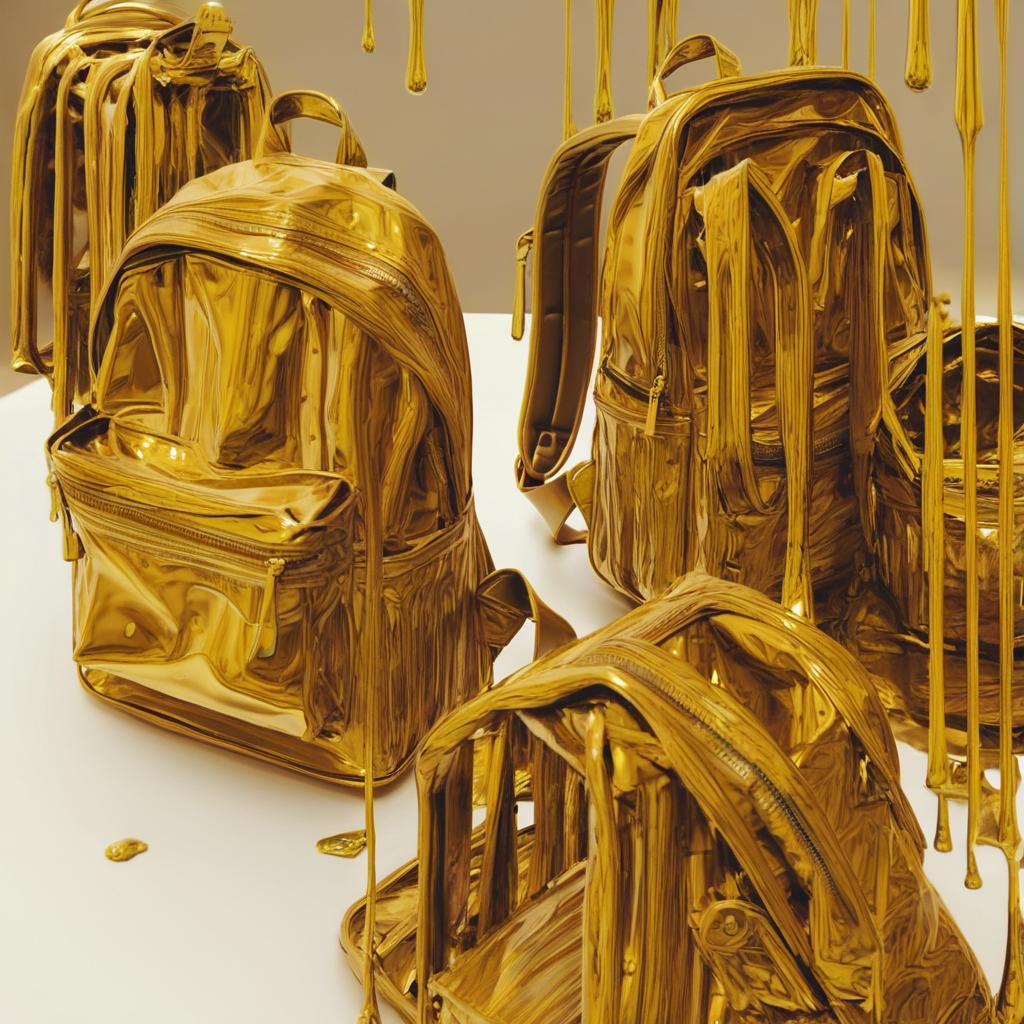}} 
    \\
    \cline{2-8}
    \end{tabular}
    }
        \caption{\textbf{Generalization to New Splits.}  \ours performs well also when trained and tested on more challenging splits.}
    \label{fig:generalization_splits}
\end{figure}

The results are shown in \cref{fig:generalization_splits}. 
Despite the challenging setups with no overlap between training and test set macro-categories, our method still performs well and outperforms ZipLoRA, even if the results are slightly worse than in setups with more diverse training data, as expected.

In both (1) and (2), we observe that  LoRA.rar better preserves the style  (\eg, {\color{red}red}-bordered images) and the subject identity (\eg, {\color{blue}blue} images). At the same time, it reduces hallucinations (\eg, {\color{ForestGreen}green} image, where ZipLoRA unnecessarily repeats the subject), degenerate outputs (\eg, {\color{yellow}yellow} image, where the subject is missing), or unrealistic samples (\eg, in wood style, our samples exhibit a more wooden look and do not float in the air, unlike the first and third outputs of ZipLoRA).

\setlength{\fboxrule}{1.0pt}
\setlength{\fboxsep}{0pt}   %

\subsection{Additional Qualitative Results}
\begin{figure*}[!t]
    \centering
    \resizebox{\textwidth}{!}{%
    \begin{tabular}[c]{L L | L L L L}
     \small A [C] dog \dots & \small \dots in [S] style &  \small \dots playing with a ball & \small \dots catching a frisbie & \small \dots wearing a hat & \small \dots with a crown \\    
    \tabfigure{width=0.1\textwidth}{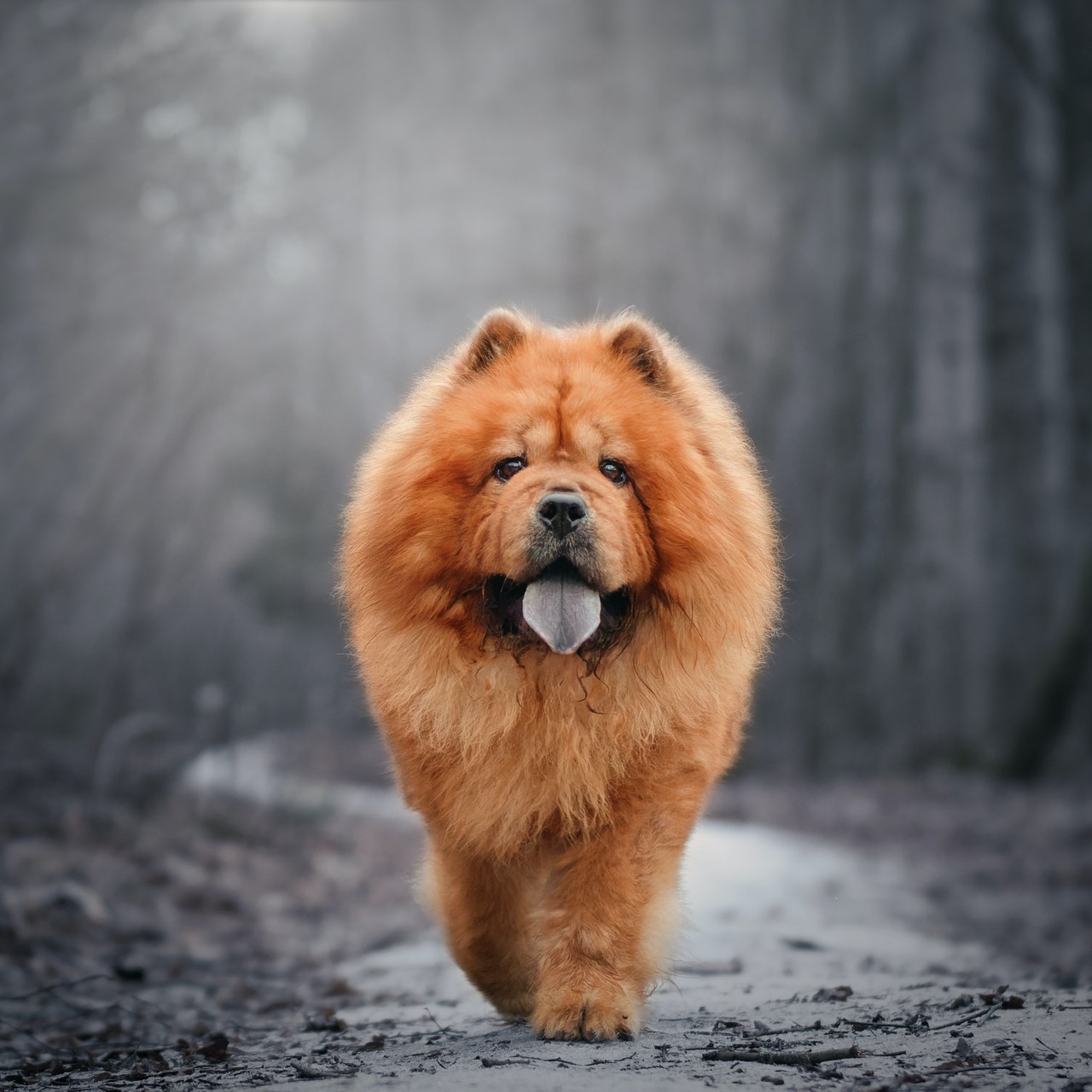} &
    \tabfigure{width=0.1\textwidth}{figures/dataset/test/flat.jpg} &
    \tabfigure{width=0.1\textwidth}{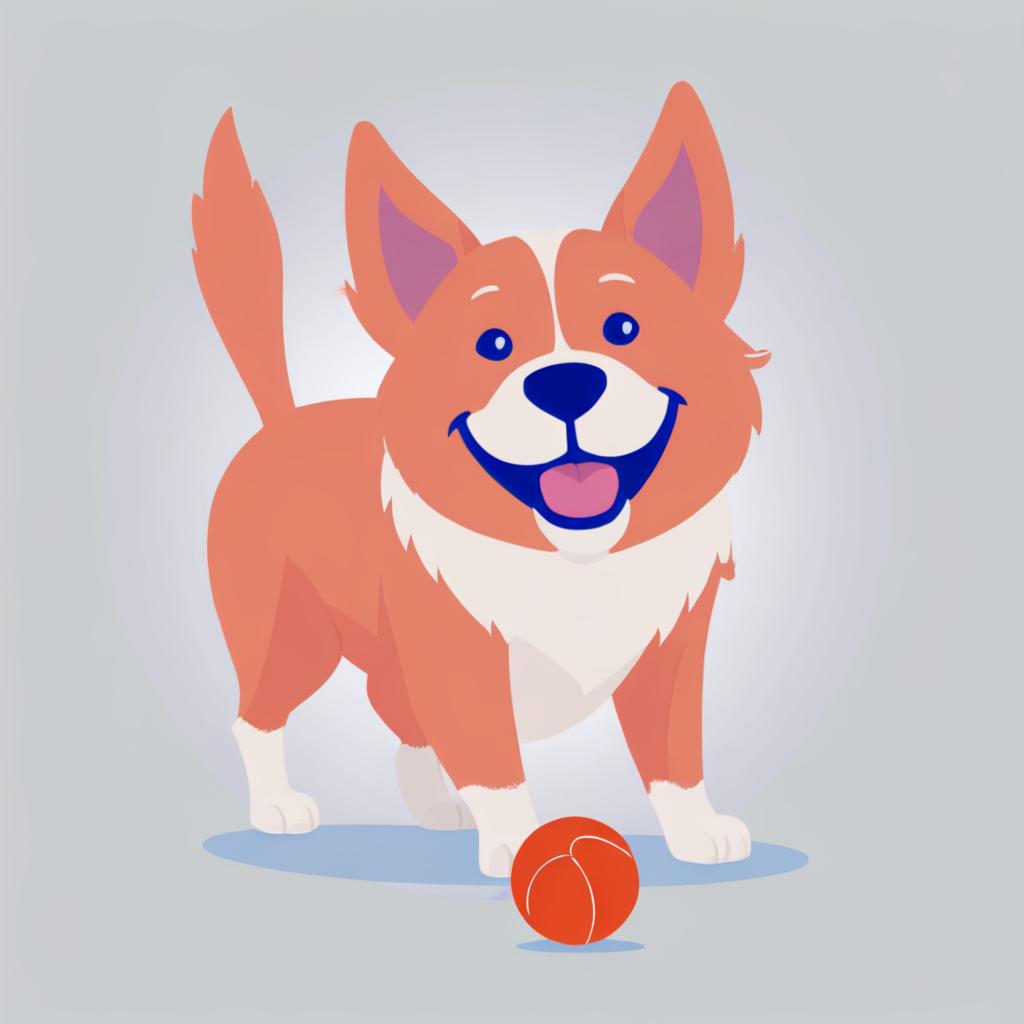} &
    \tabfigure{width=0.1\textwidth}{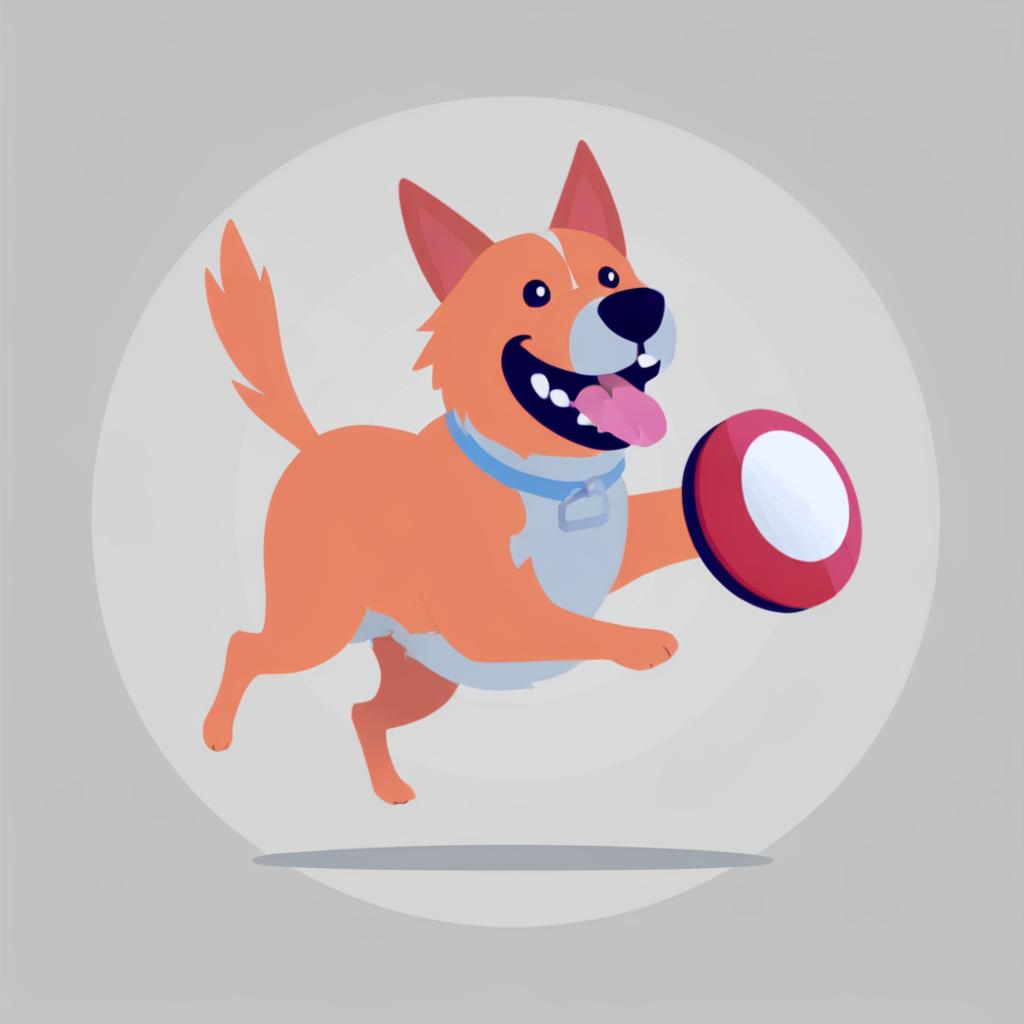} &
    \tabfigure{width=0.1\textwidth}{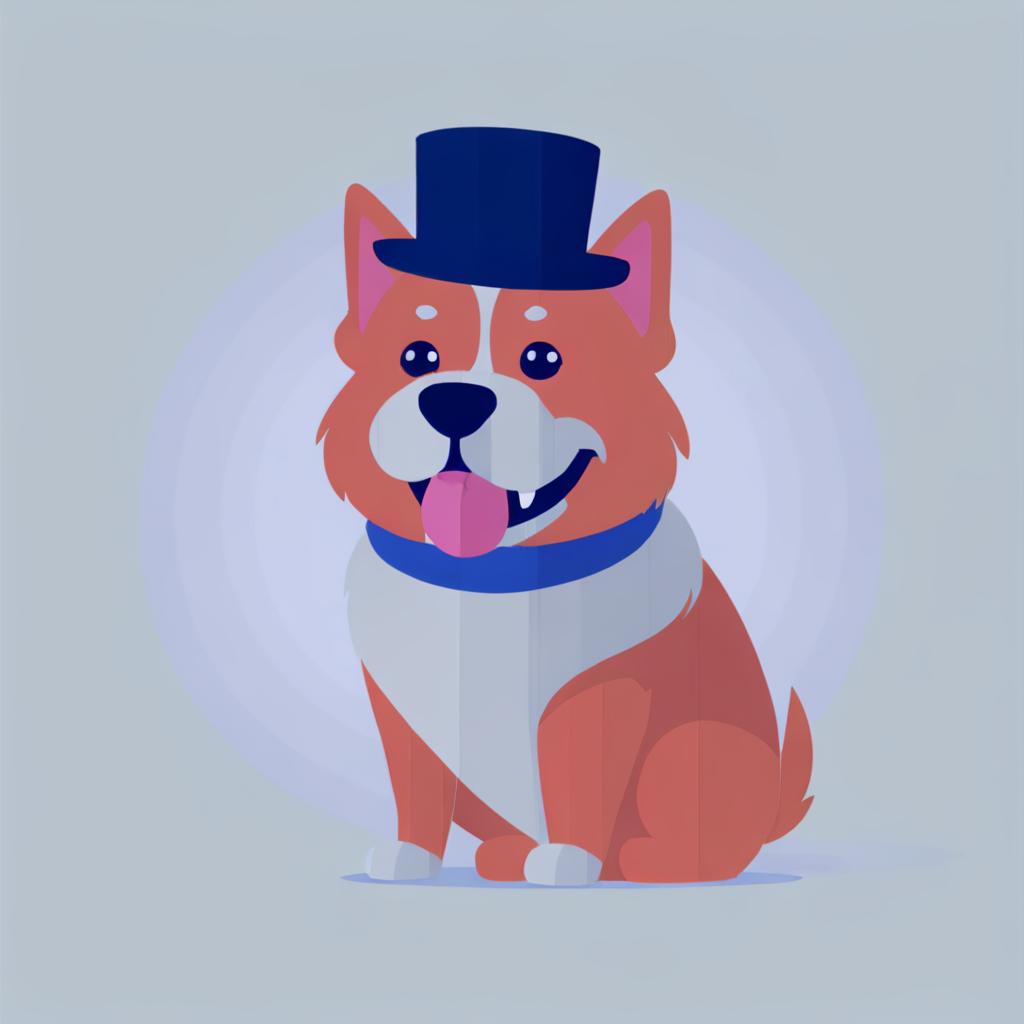} &
    \tabfigure{width=0.1\textwidth}{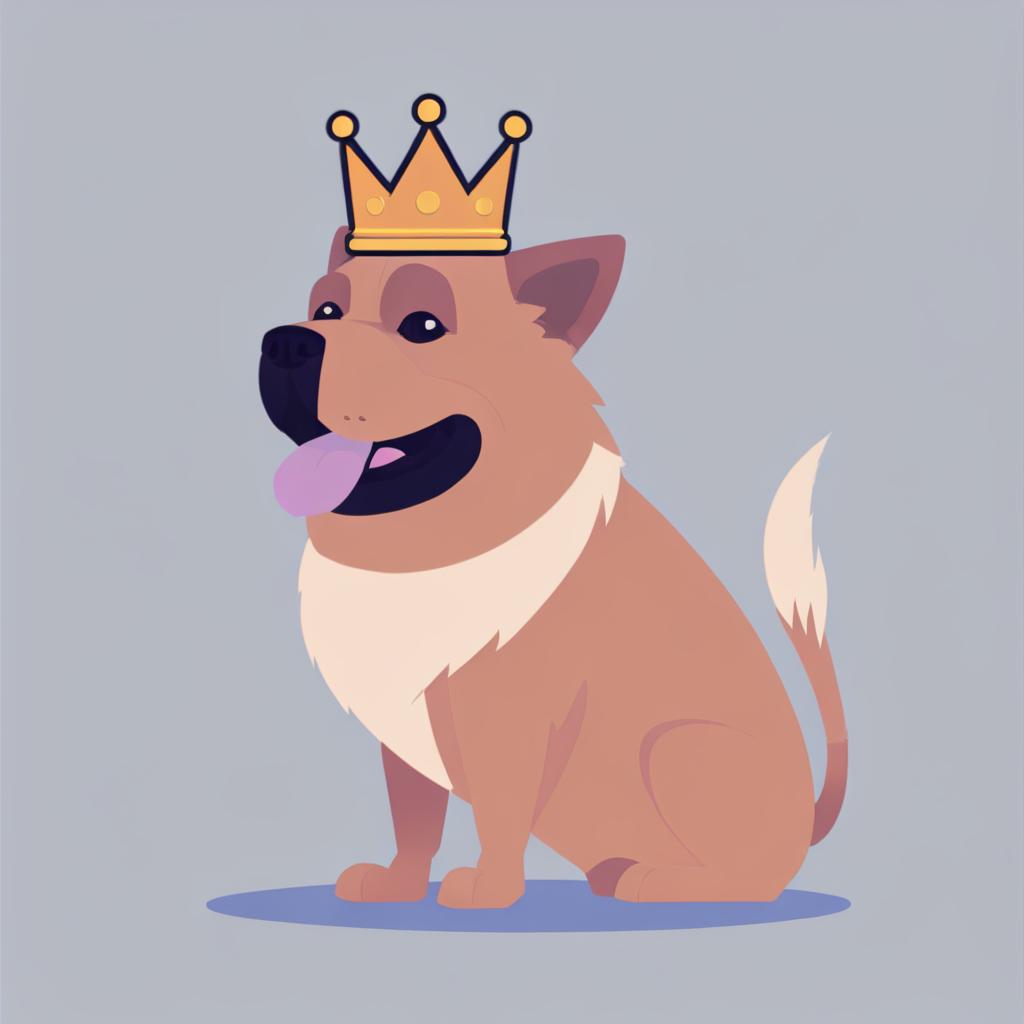} \\
     && \small \dots riding a bicycle & \small \dots sleeping & \small \dots in a boat & \small  \dots driving a car \\
    && \tabfigure{width=0.1\textwidth}{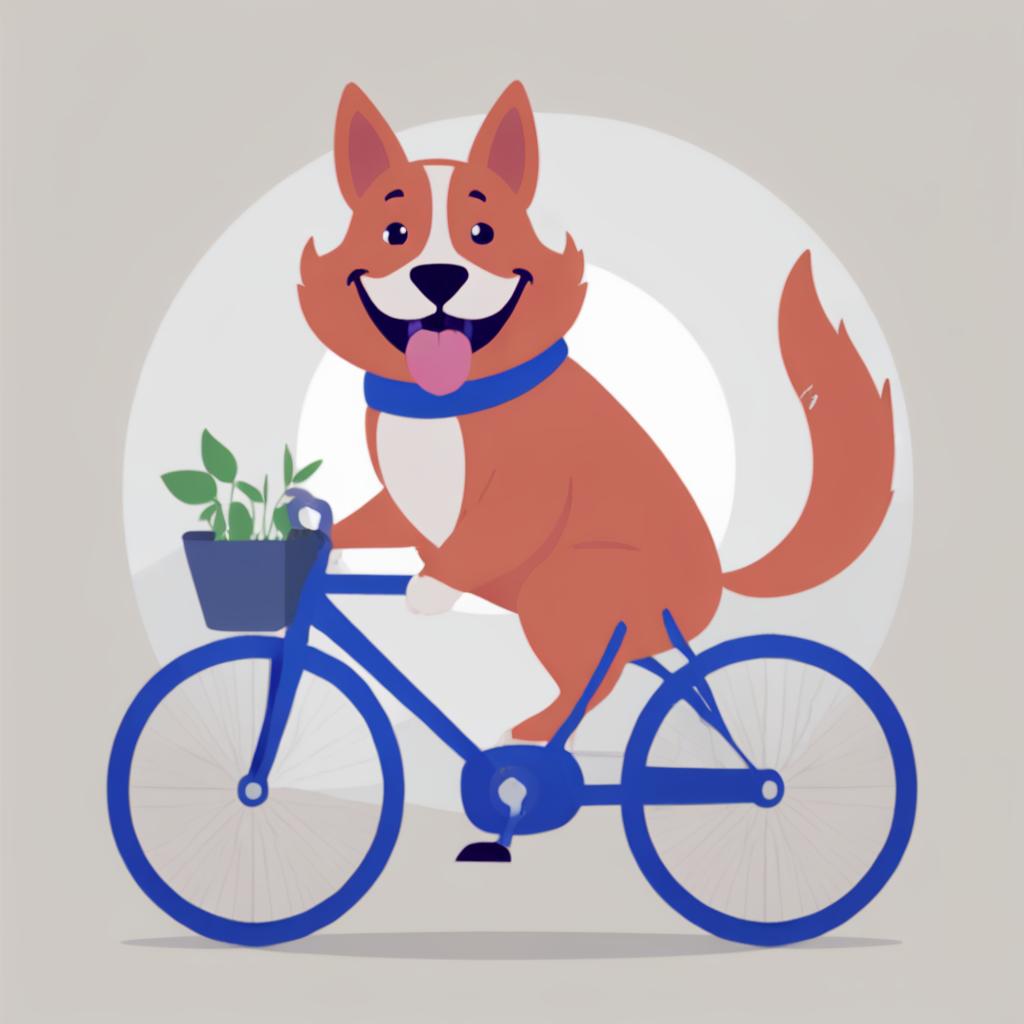} &
    \tabfigure{width=0.1\textwidth}{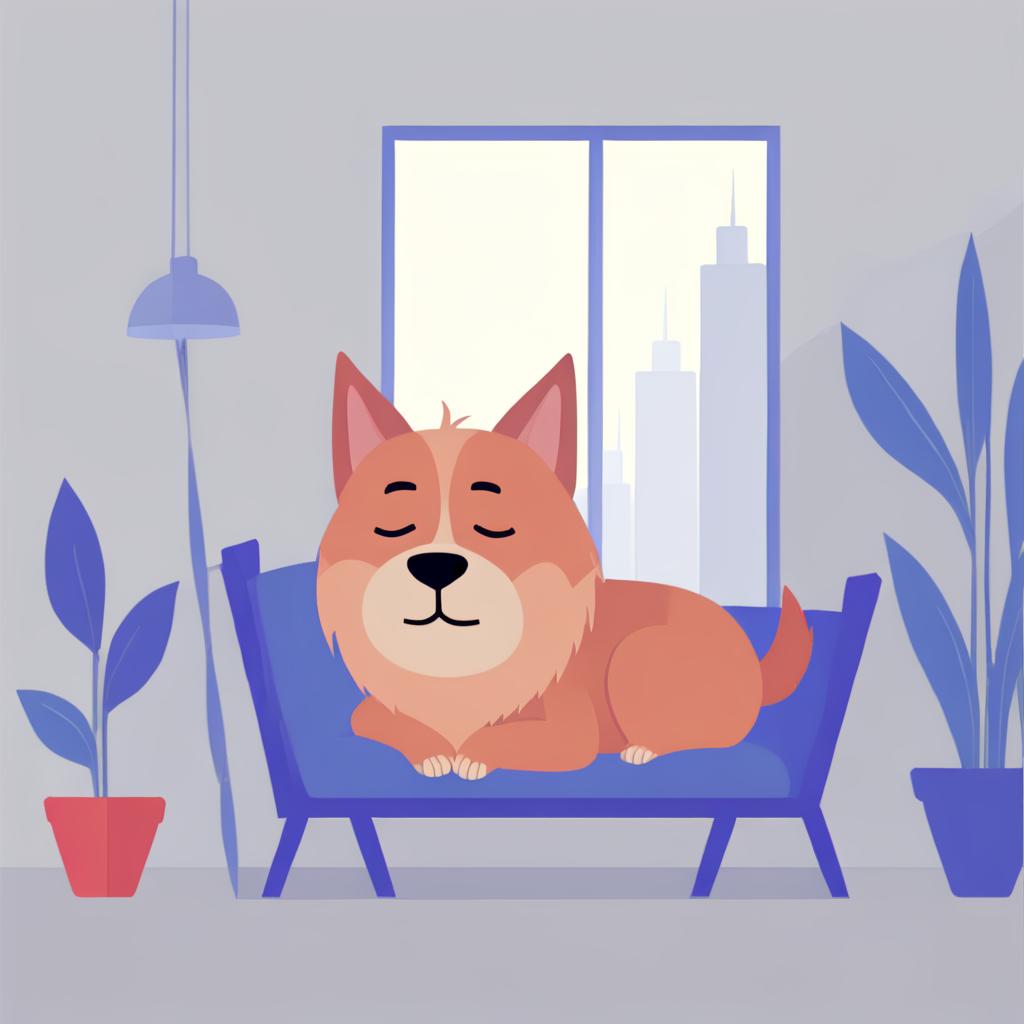} &
    \tabfigure{width=0.1\textwidth}{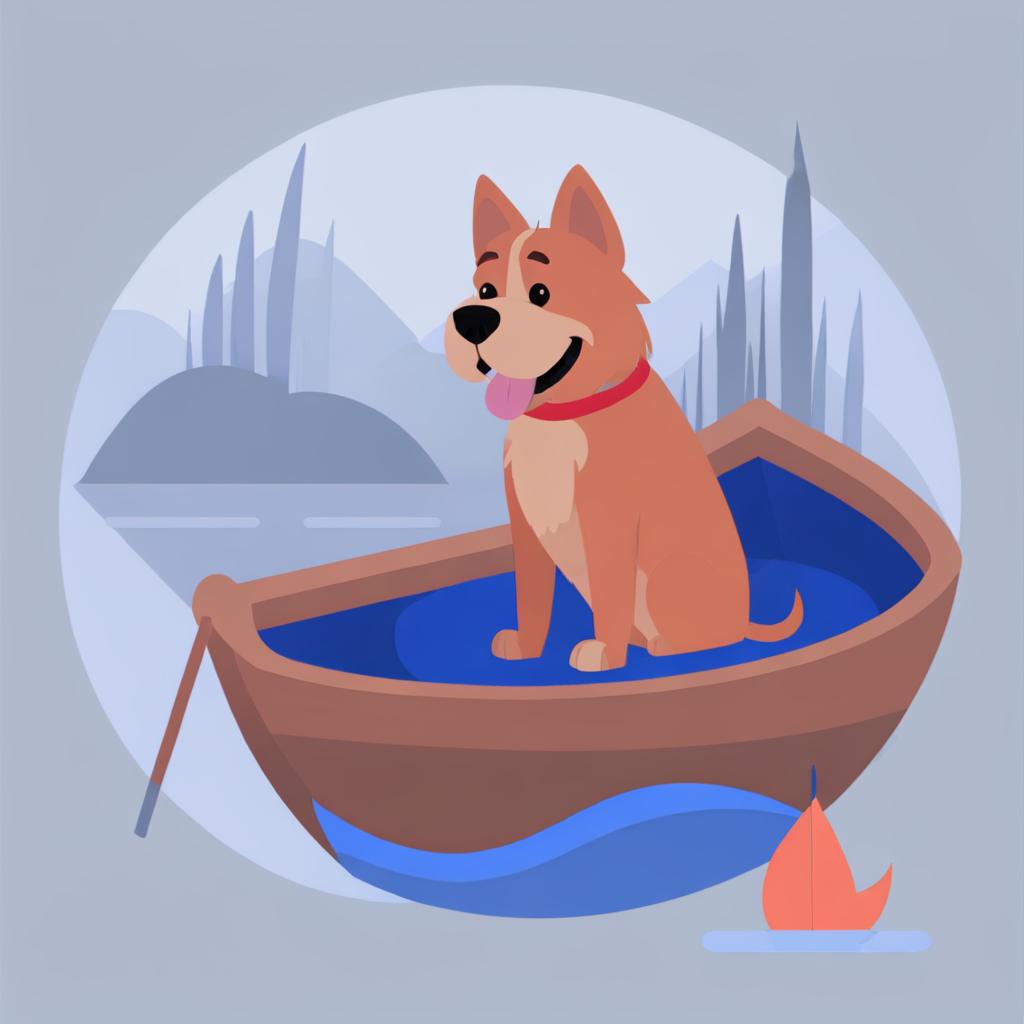} &
    \tabfigure{width=0.1\textwidth}{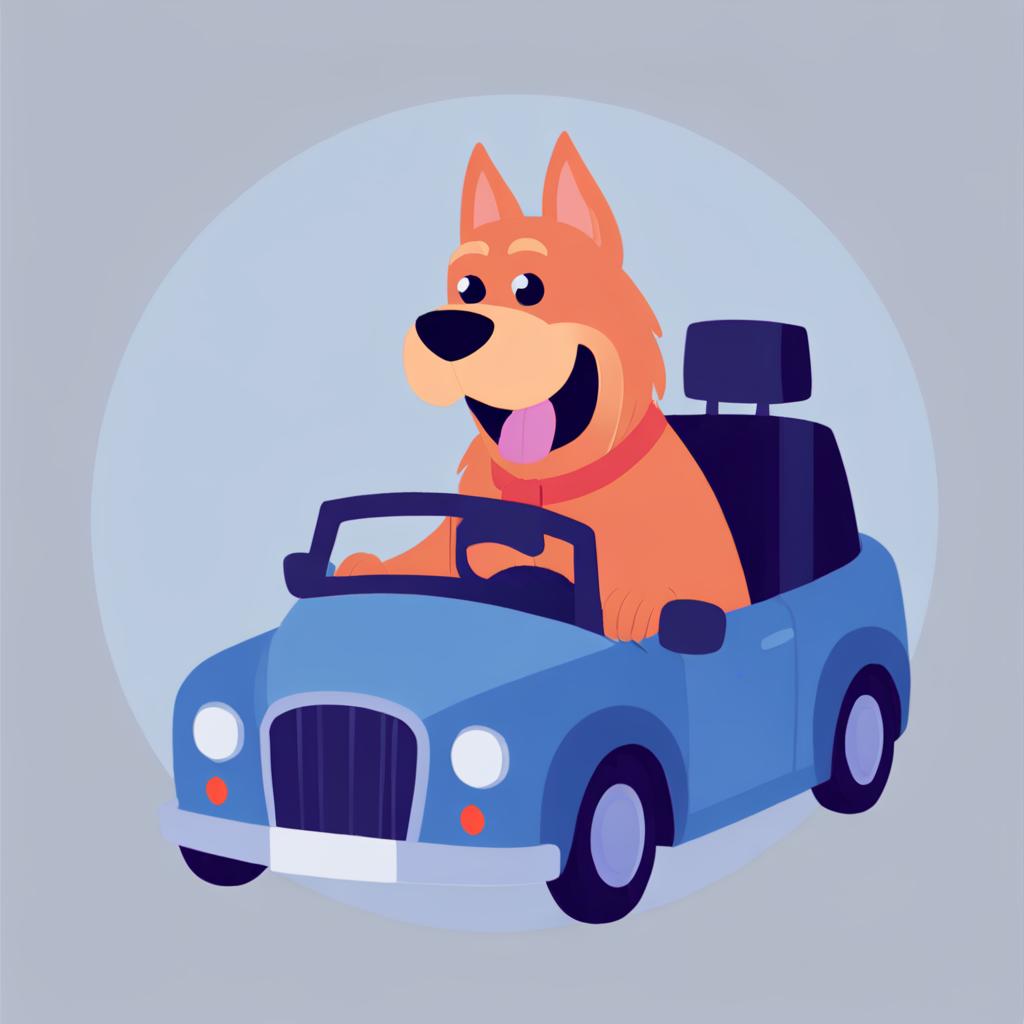} \\ 
    \vspace{4mm} & \vspace{4mm} & \vspace{4mm} & \vspace{4mm} & \vspace{4mm} & \vspace{4mm} \\ 
     \small A [C] stuffed animal \dots & \small \dots in [S] style &  \small \dots playing with a ball & \small \dots catching a frisbie& \small \dots wearing a hat& \small \dots with a crown \\    
    \tabfigure{width=0.1\textwidth}{figures/dataset/test/wolf_plushie/00.jpg} &
    \tabfigure{width=0.1\textwidth}{figures/qualitative/3d_rendering.jpg} &
    \tabfigure{width=0.1\textwidth}{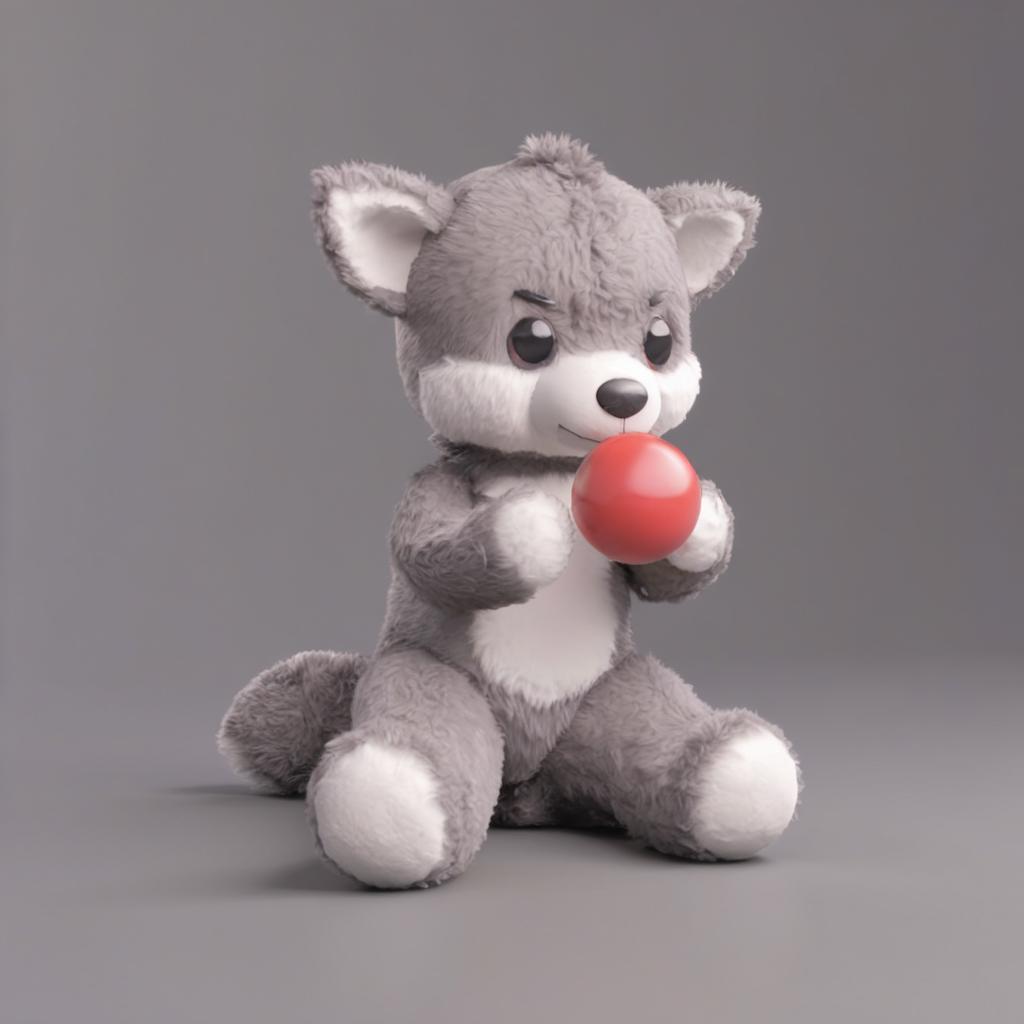} &
    \tabfigure{width=0.1\textwidth}{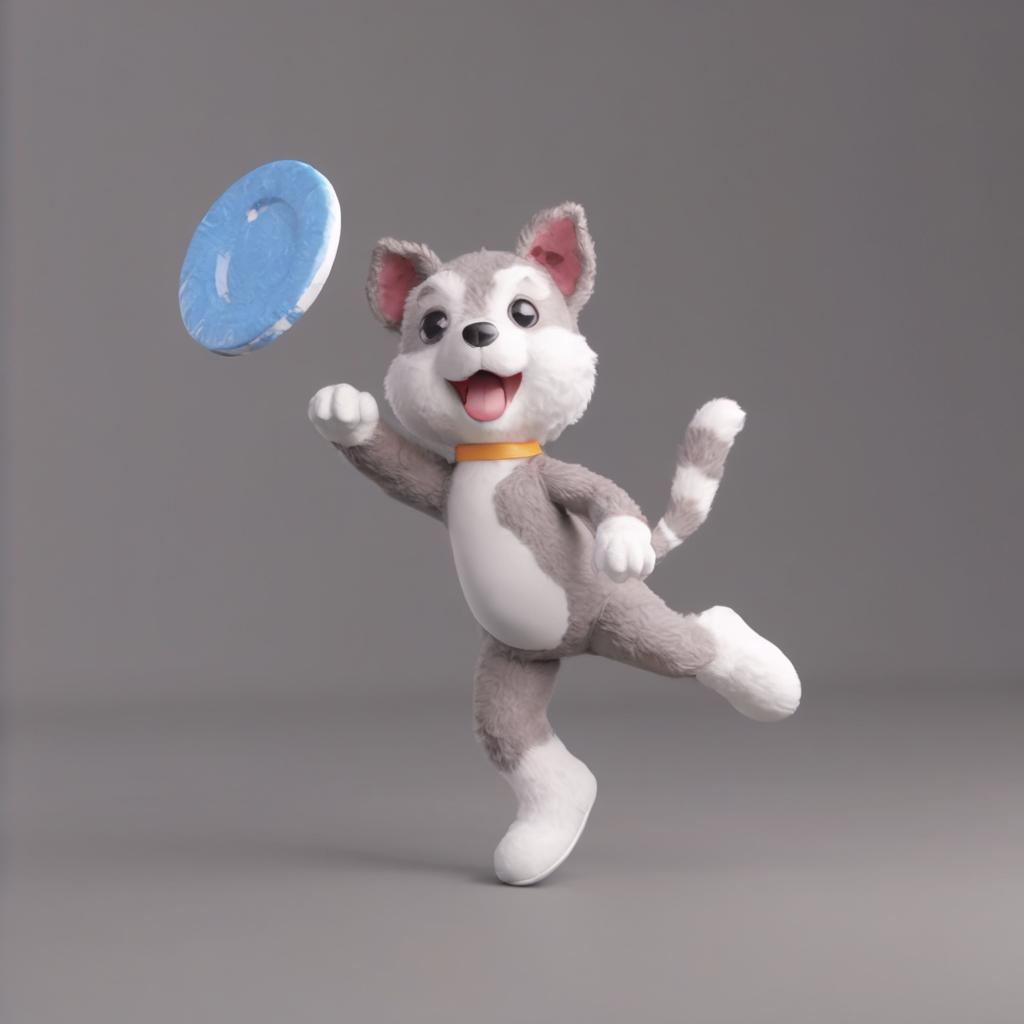} &
    \tabfigure{width=0.1\textwidth}{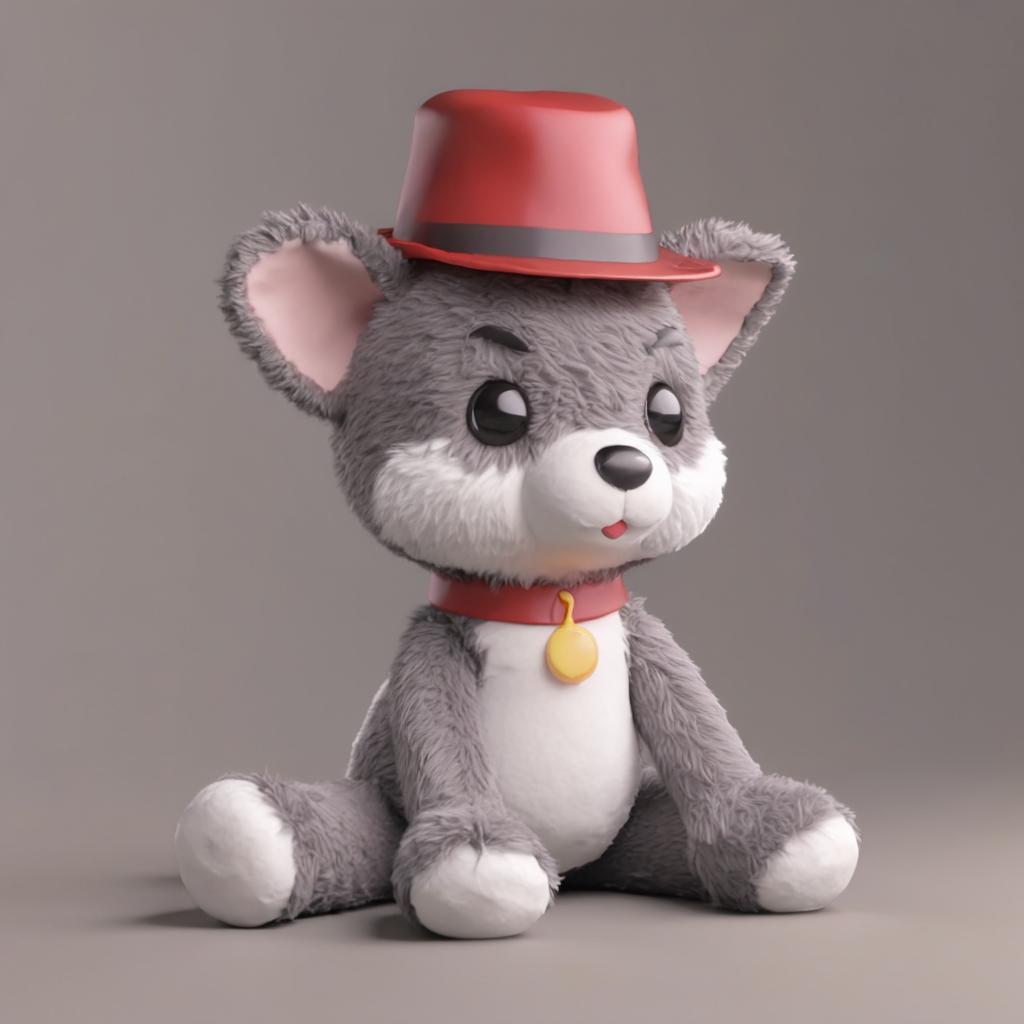} &
    \tabfigure{width=0.1\textwidth}{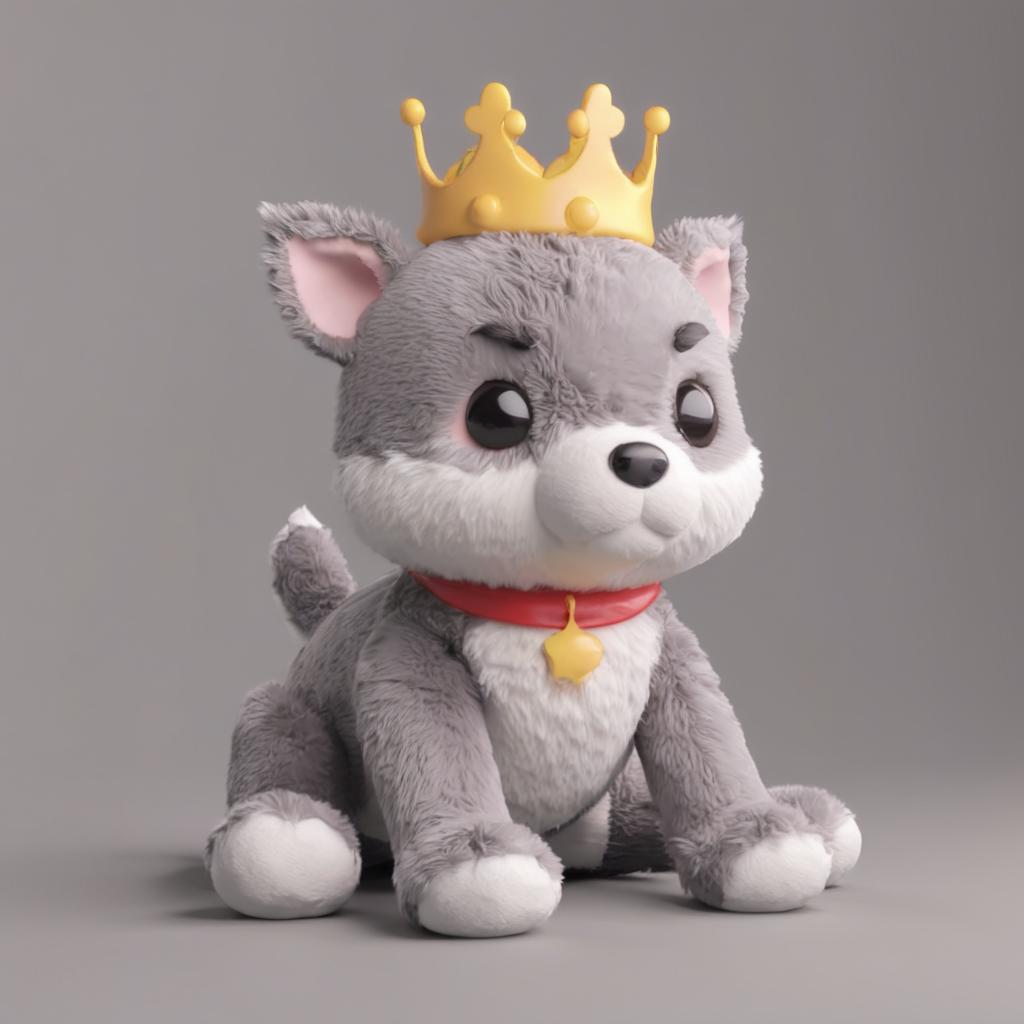} \\
     &&\small \dots riding a bicycle & \small \dots sleeping & \small \dots in a boat & \small  \dots driving a car \\
    & & \tabfigure{width=0.1\textwidth}{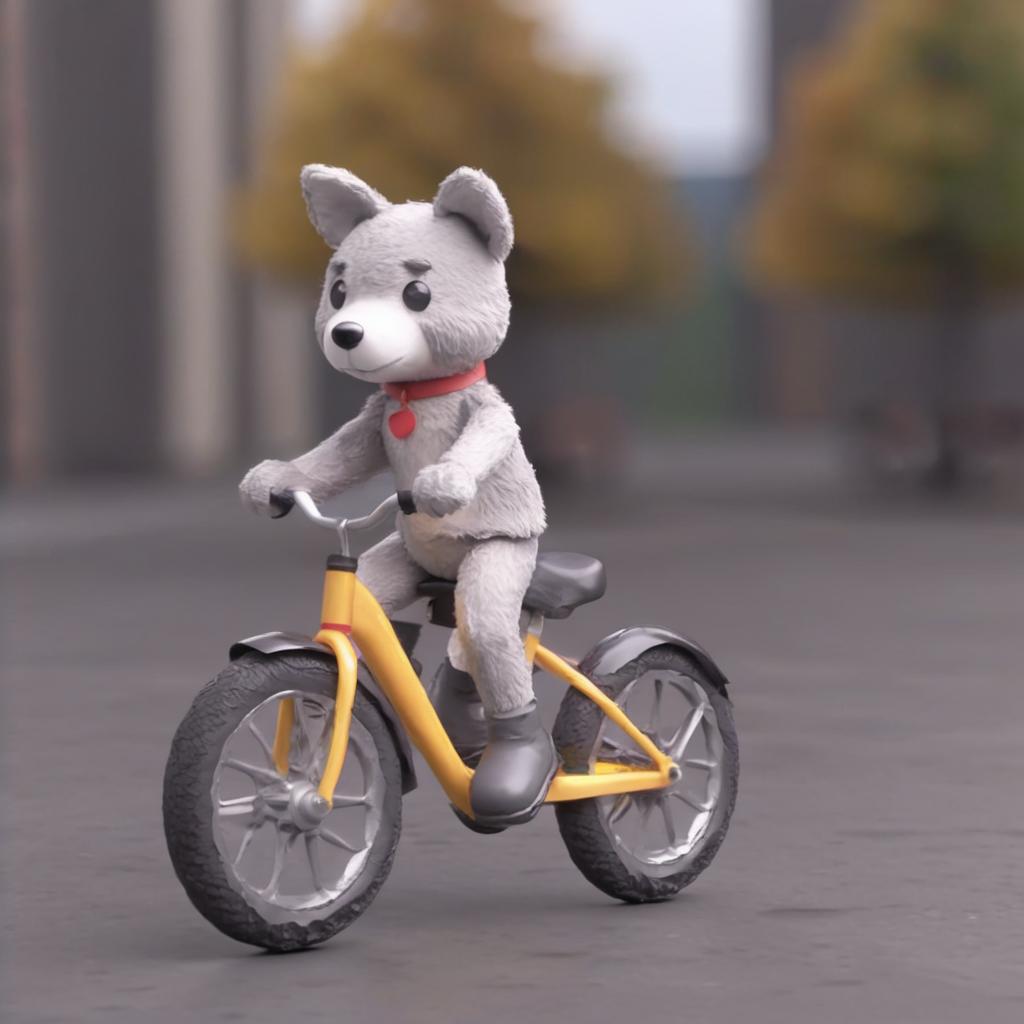} &
    \tabfigure{width=0.1\textwidth}{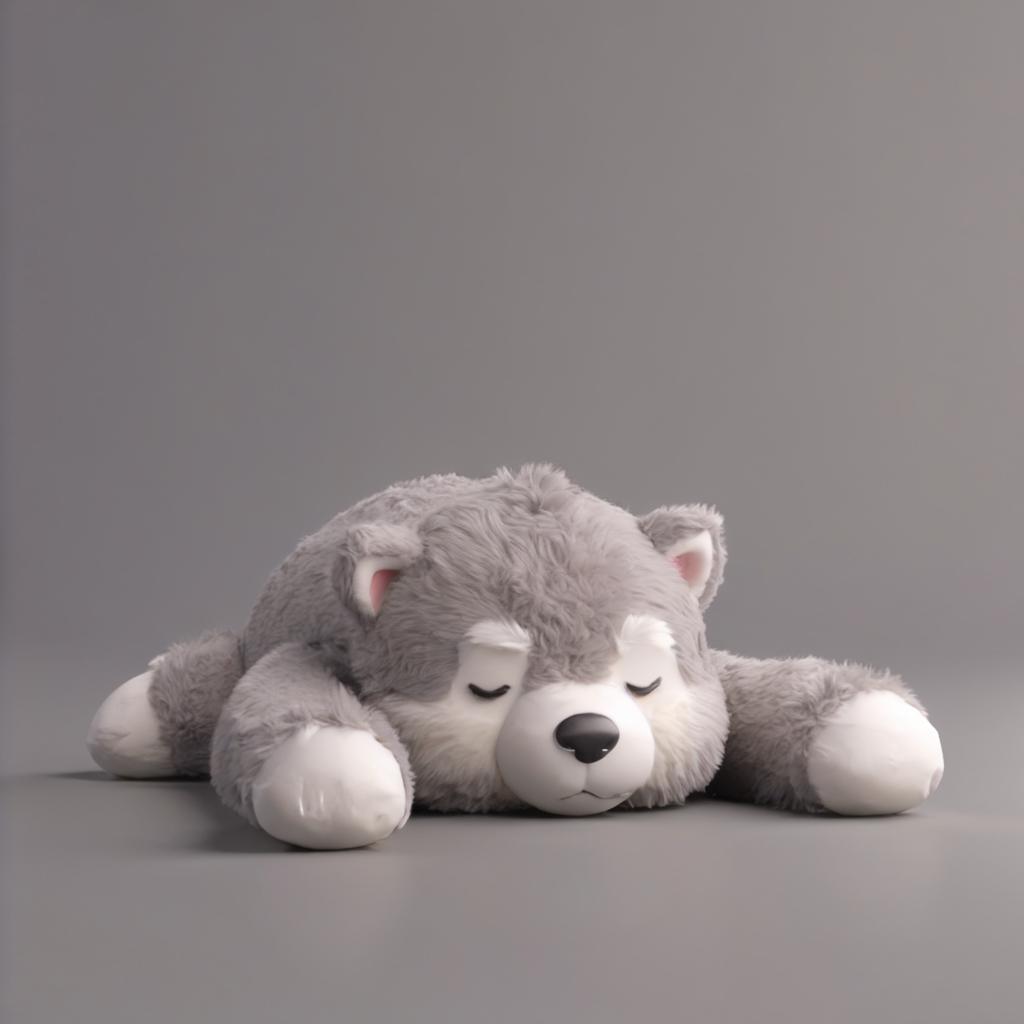} &
     \tabfigure{width=0.1\textwidth}{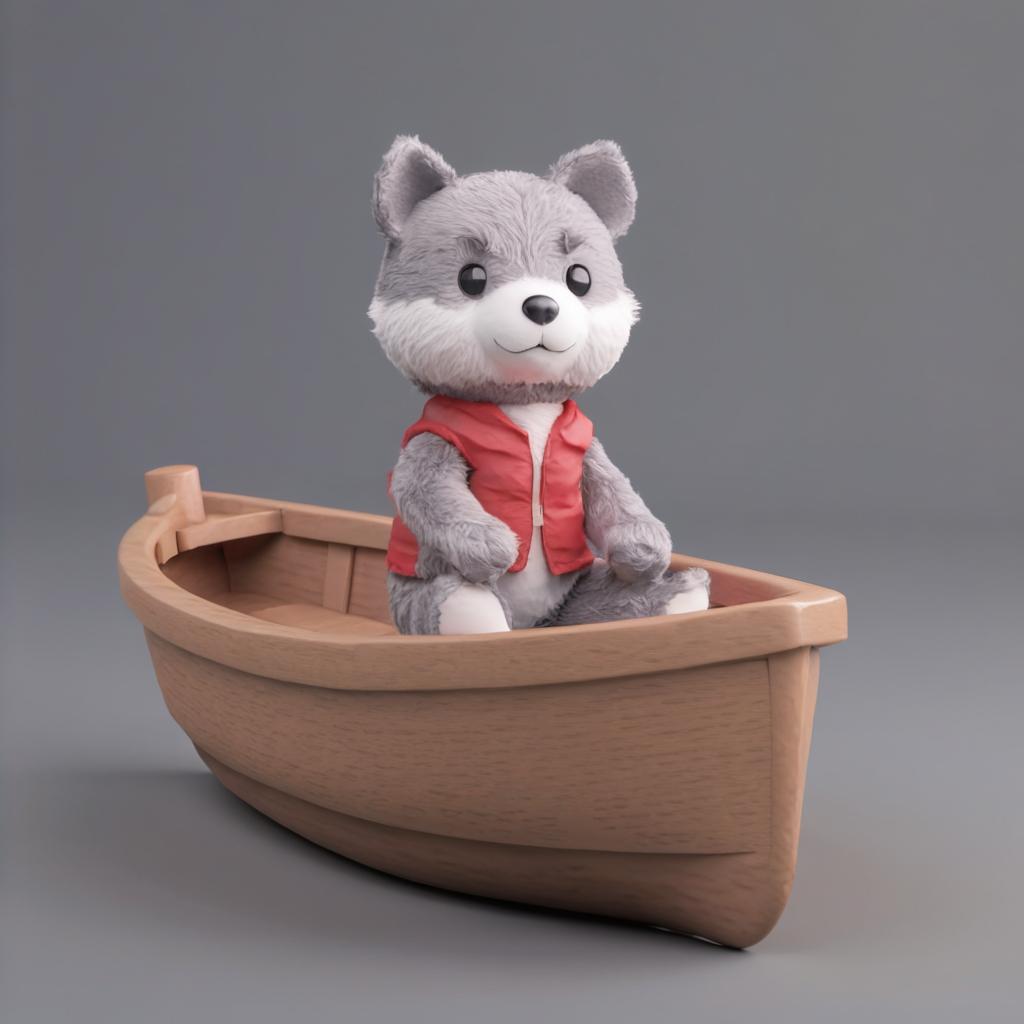} &
    \tabfigure{width=0.1\textwidth}{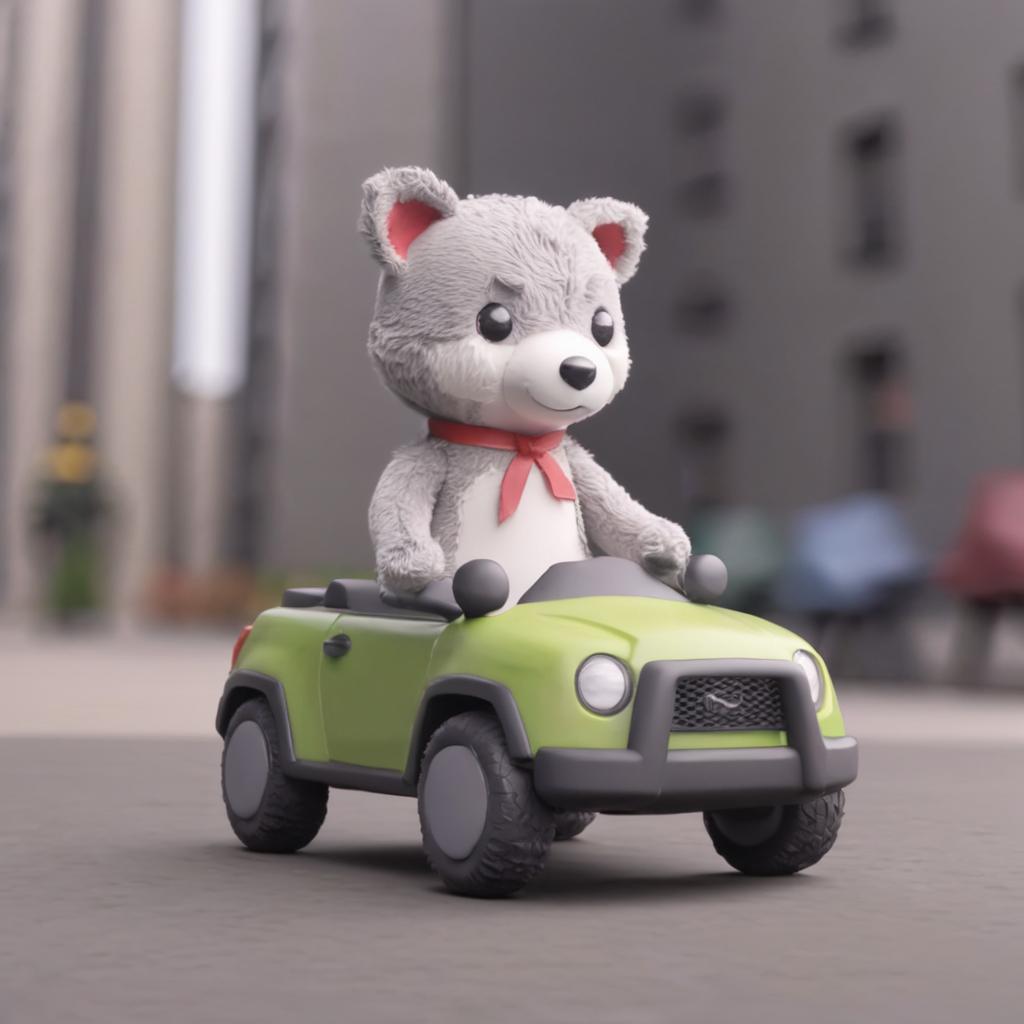}
    \end{tabular}
    }
    \caption{\textbf{Recontextualization Output Generations}. Generated outputs using various prompts for the contents ``dog2" and ``wolf plushie''.}
    \label{fig:recontext_1}
\end{figure*}

\begin{figure*}[!t]
    \centering
    \resizebox{\textwidth}{!}{%
    \begin{tabular}[c]{L L | L L L L}
     \small A [C] dog \dots & \small \dots in [S] style &  \small \dots playing with a ball & \small \dots catching a frisbie& \small \dots wearing a hat& \small \dots with a crown \\    
    \tabfigure{width=0.1\textwidth}{figures/dataset/test/dog8/04.jpg} &
    \tabfigure{width=0.1\textwidth}{figures/dataset/test/glowing.jpg} &
    \tabfigure{width=0.1\textwidth}{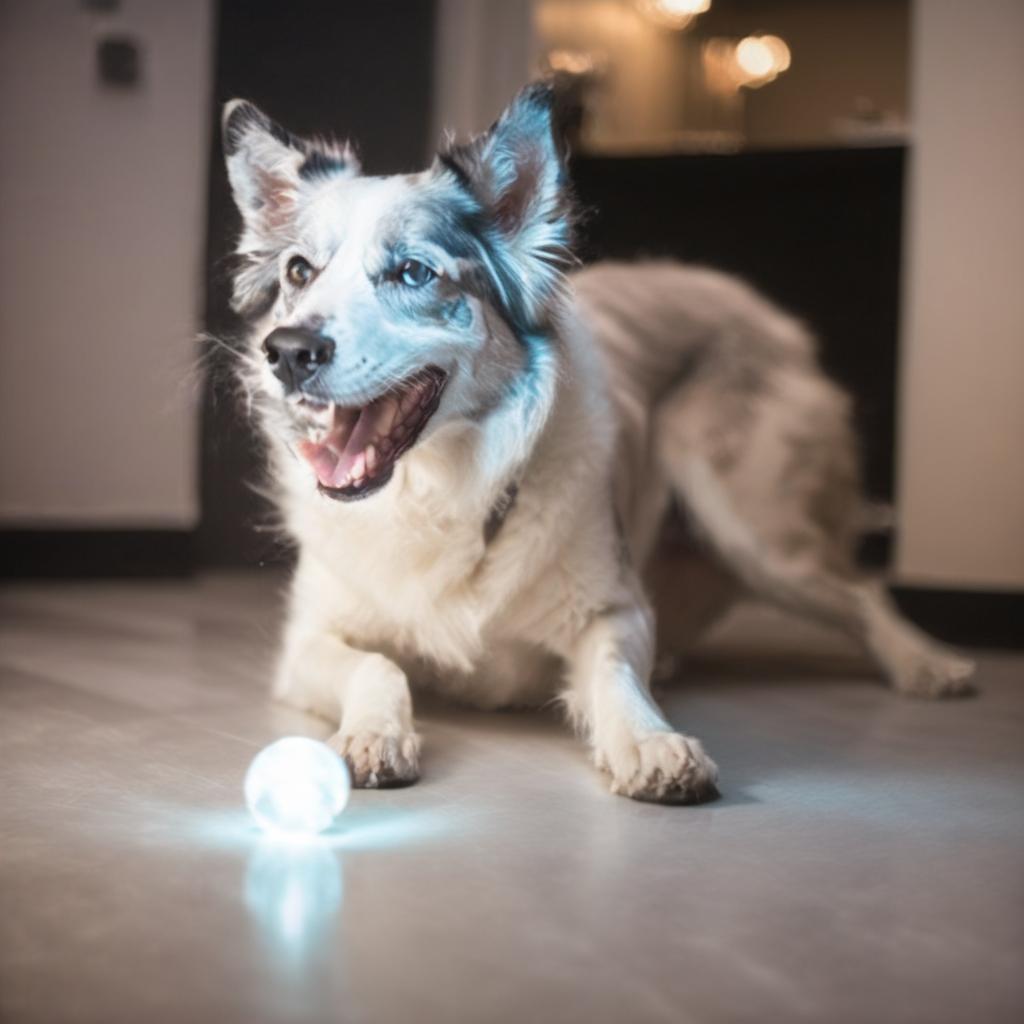} &
    \tabfigure{width=0.1\textwidth}{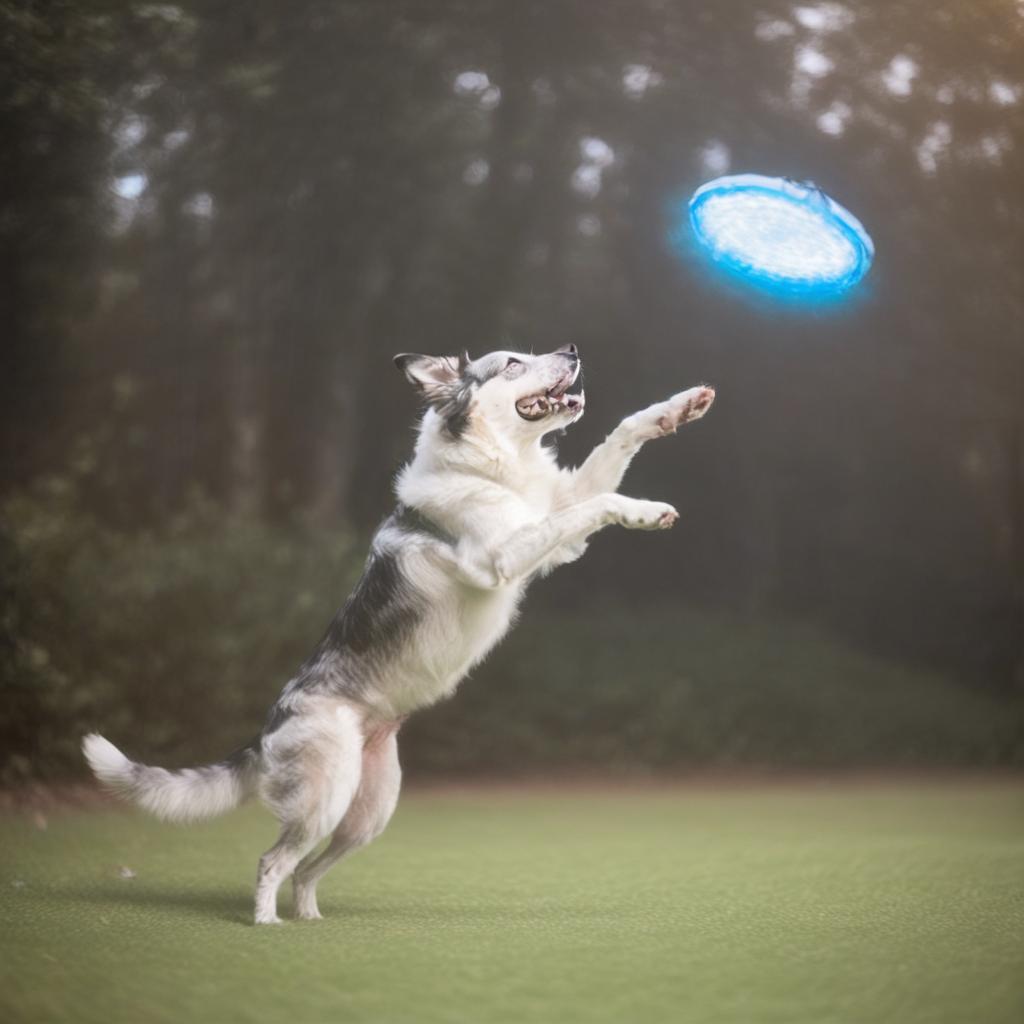} &
    \tabfigure{width=0.1\textwidth}{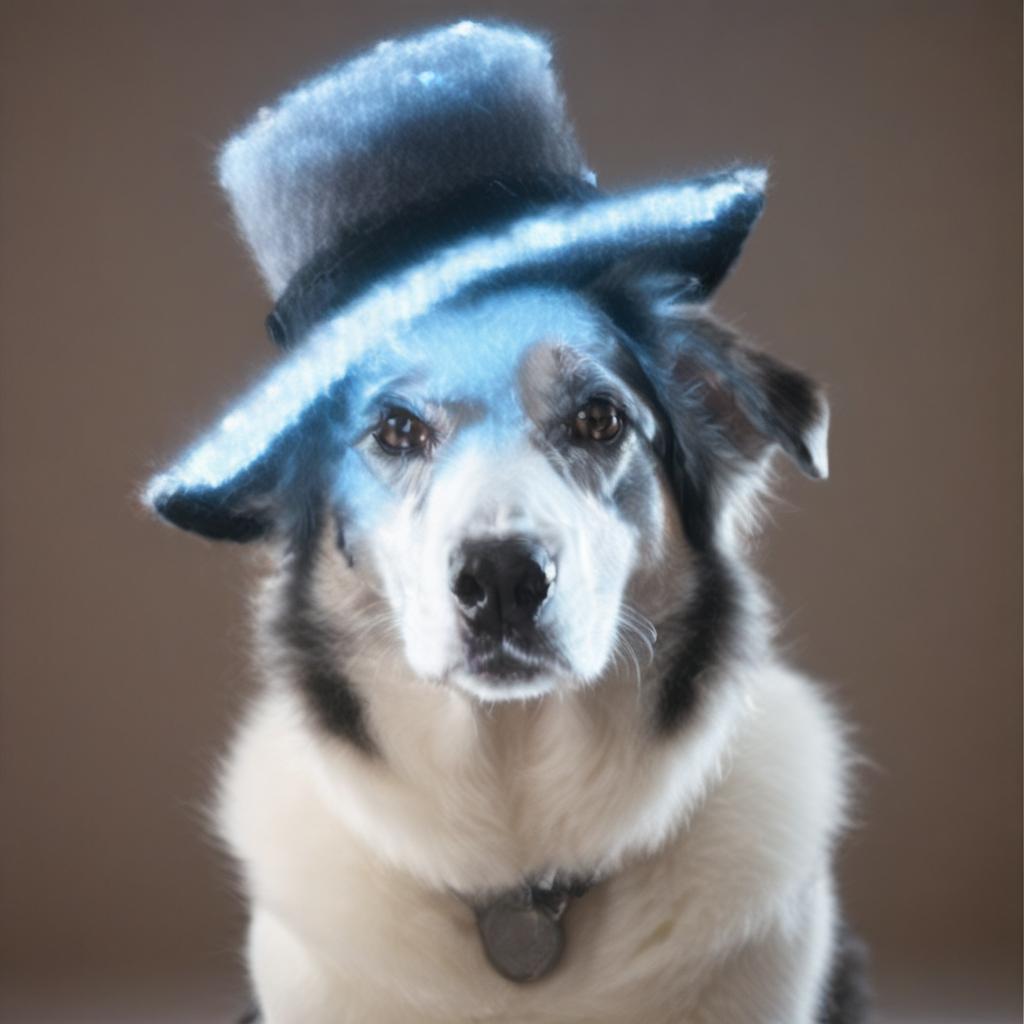} &
    \tabfigure{width=0.1\textwidth}{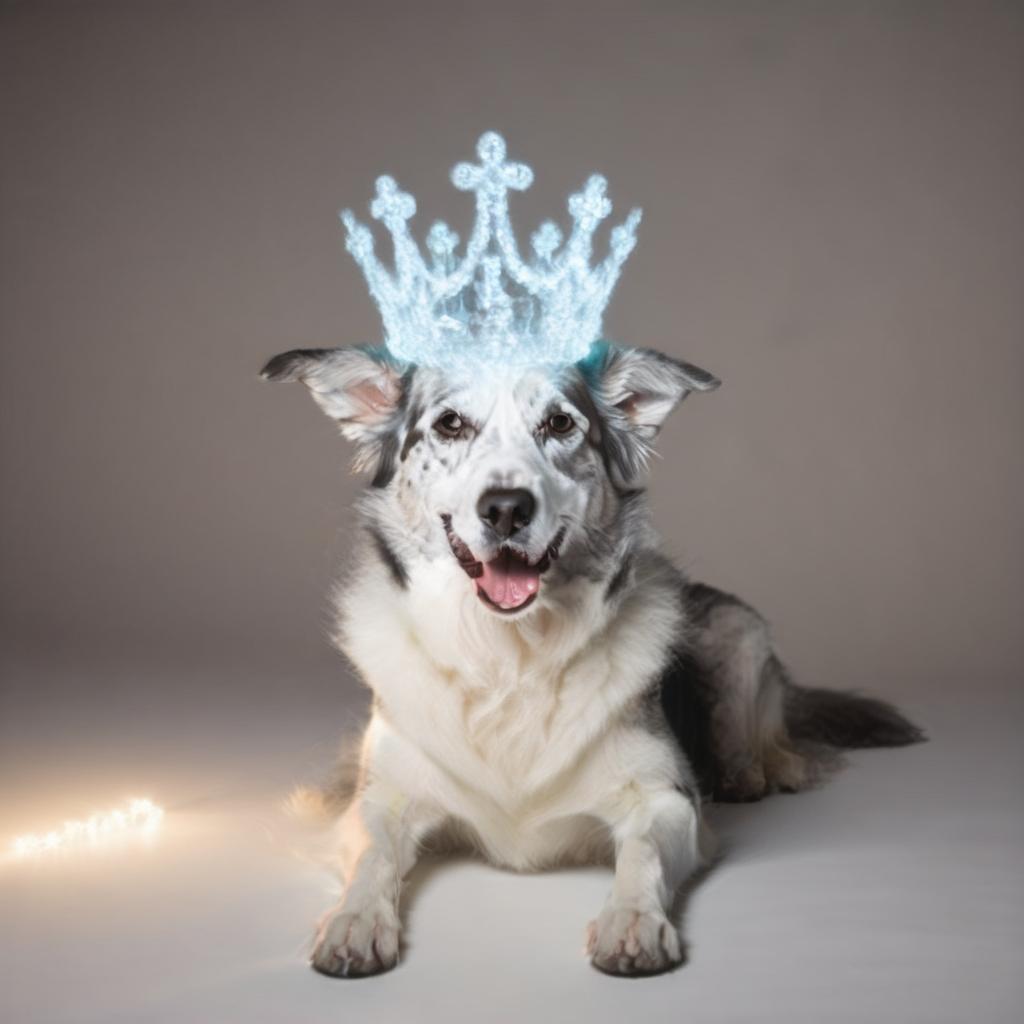} \\
     && \small \dots riding a bicycle & \small \dots sleeping & \small \dots in a boat & \small \dots driving a car \\
    && \tabfigure{width=0.1\textwidth}{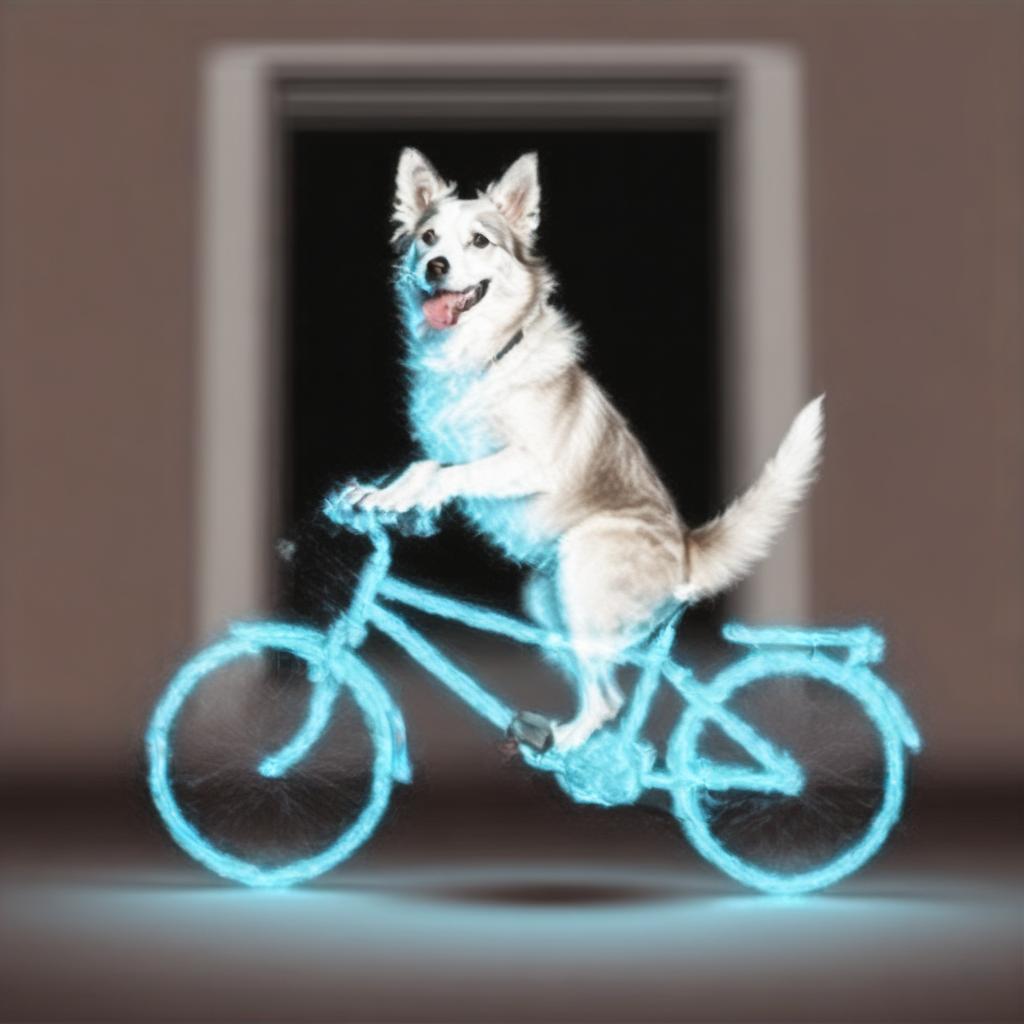} &
    \tabfigure{width=0.1\textwidth}{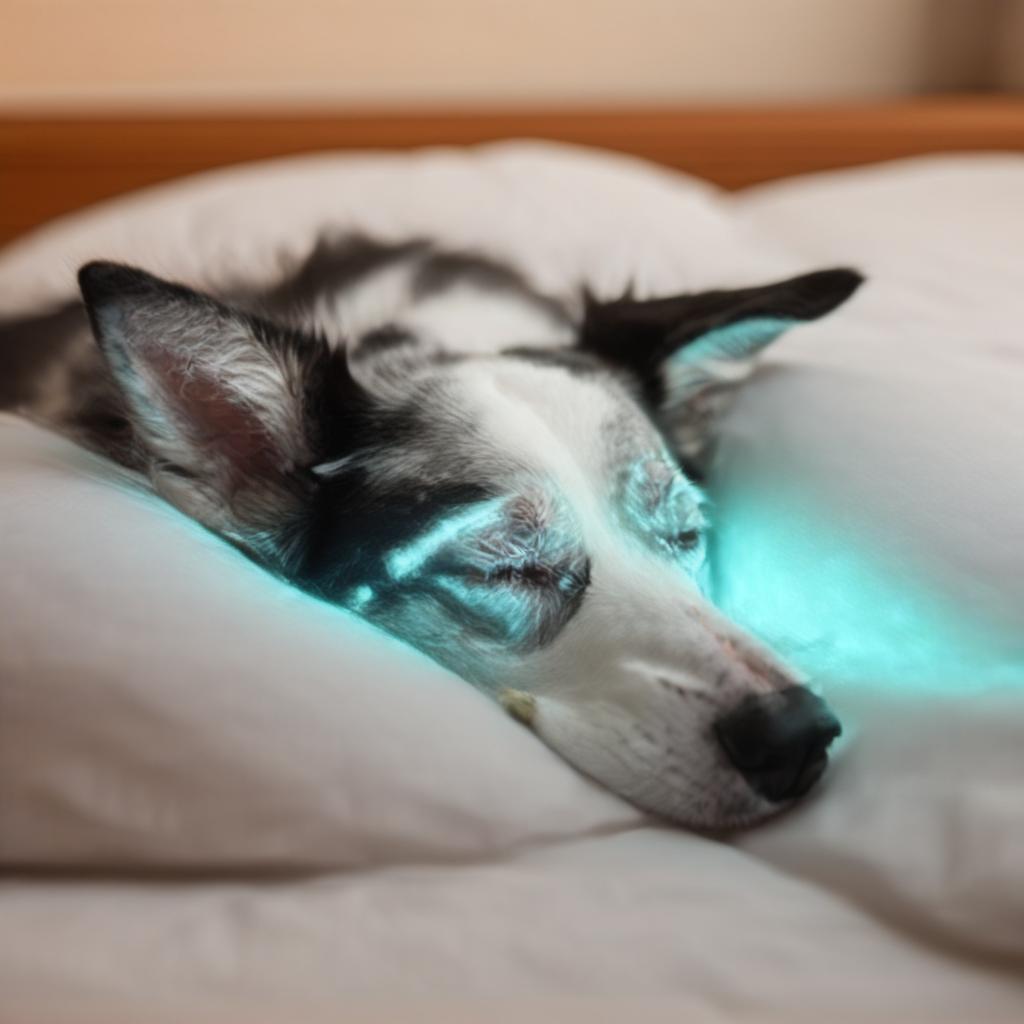} &
    \tabfigure{width=0.1\textwidth}{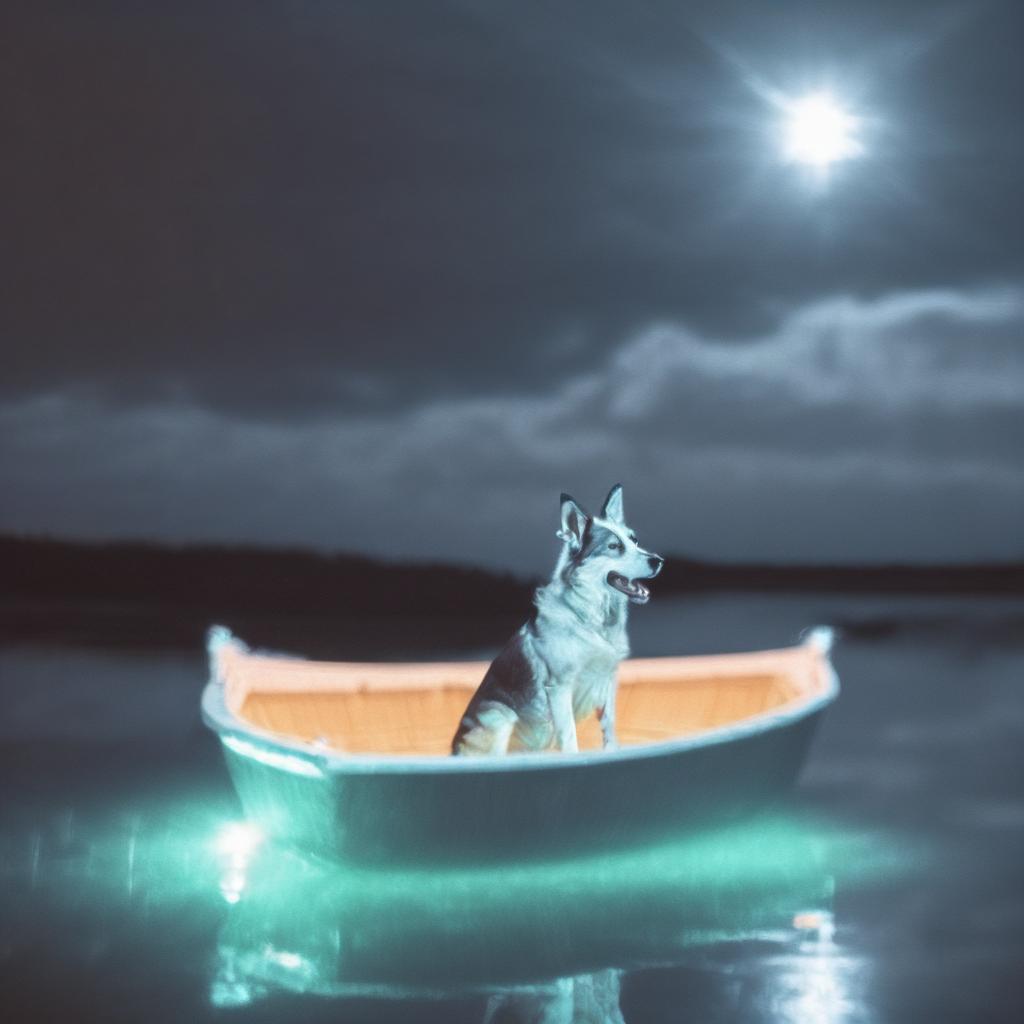} &
    \tabfigure{width=0.1\textwidth}{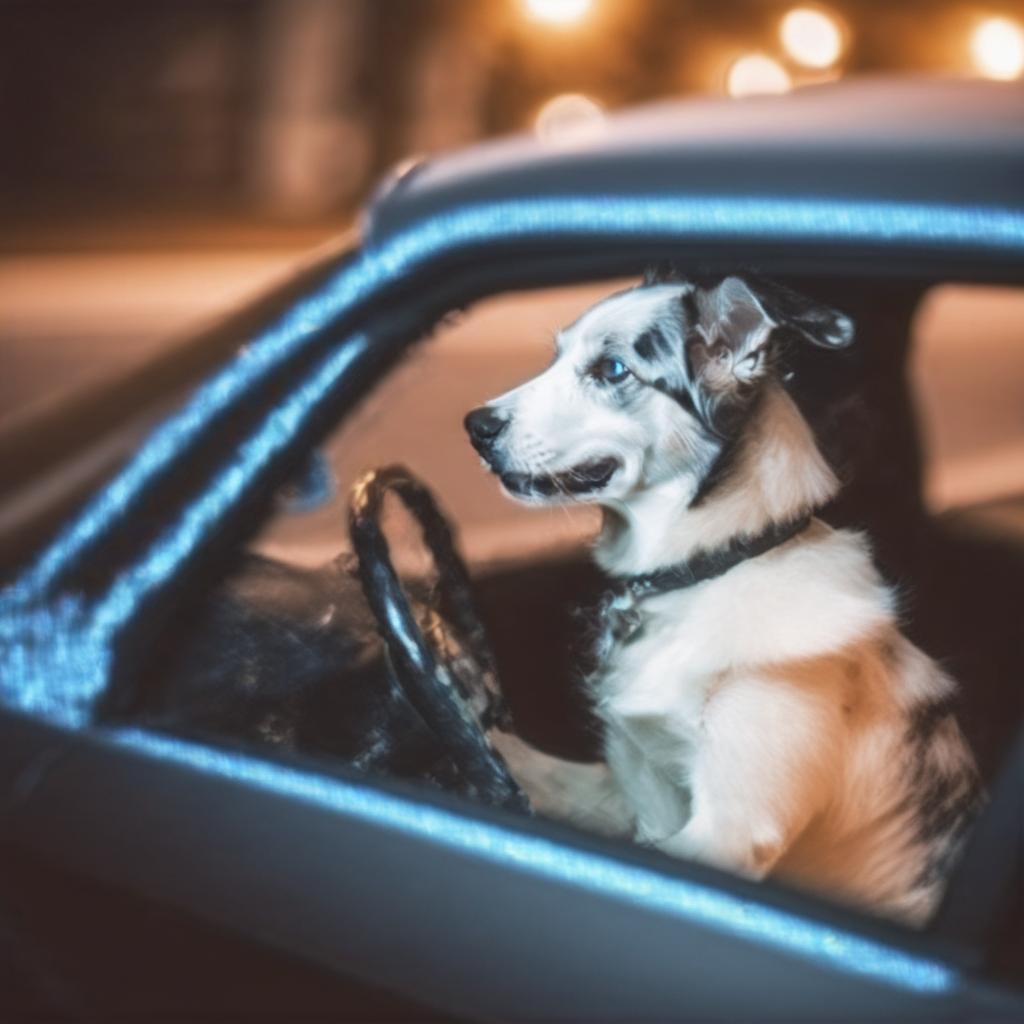} \\ 
    \vspace{4mm} & \vspace{4mm} & \vspace{4mm} & \vspace{4mm} & \vspace{4mm} & \vspace{4mm} \\ 
    
     \small A [C] cat \dots & \small \dots in [S] style &  \small \dots playing with a ball & \small \dots catching a frisbie& \small \dots wearing a hat& \small  \dots with a crown \\    
    \tabfigure{width=0.1\textwidth}{figures/dataset/test/cat2/03.jpg} &
    \tabfigure{width=0.1\textwidth}{figures/qualitative/oil.jpg} &
    \tabfigure{width=0.1\textwidth}{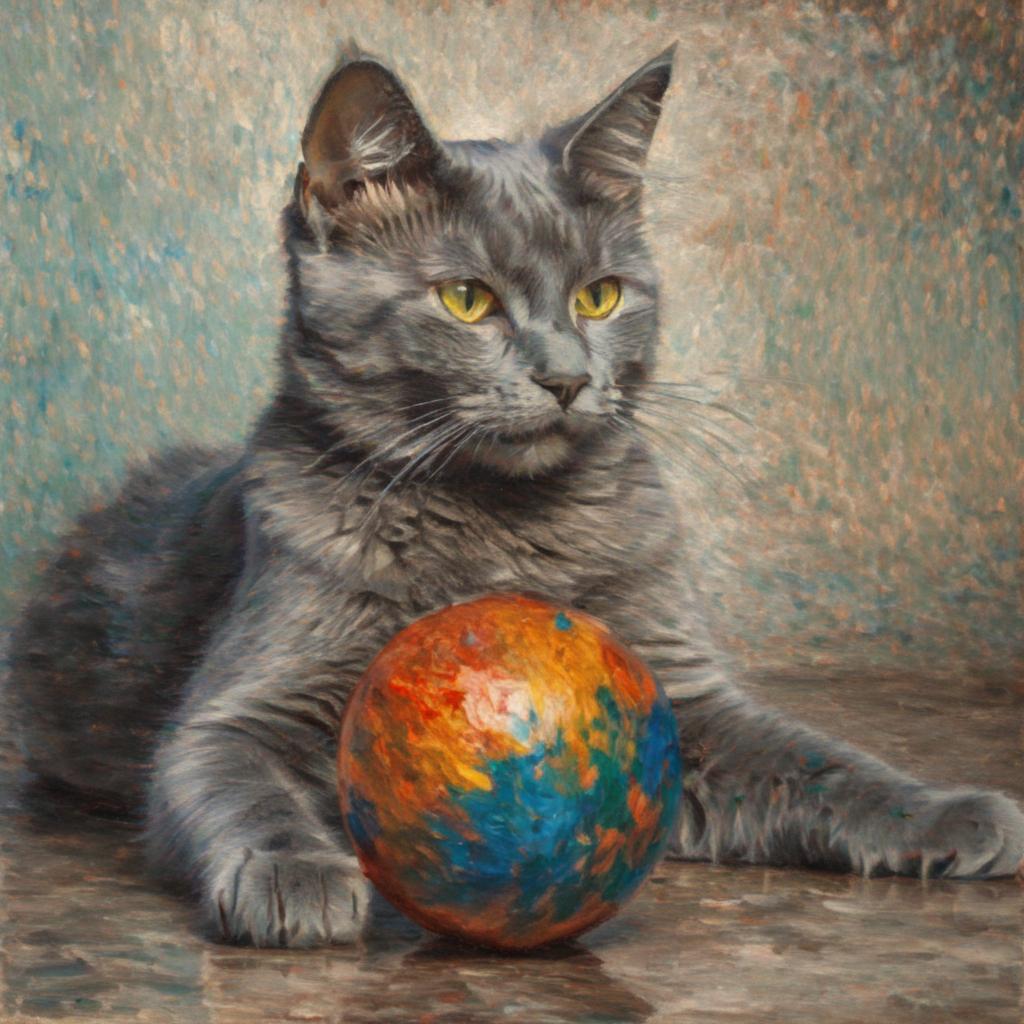} &
    \tabfigure{width=0.1\textwidth}{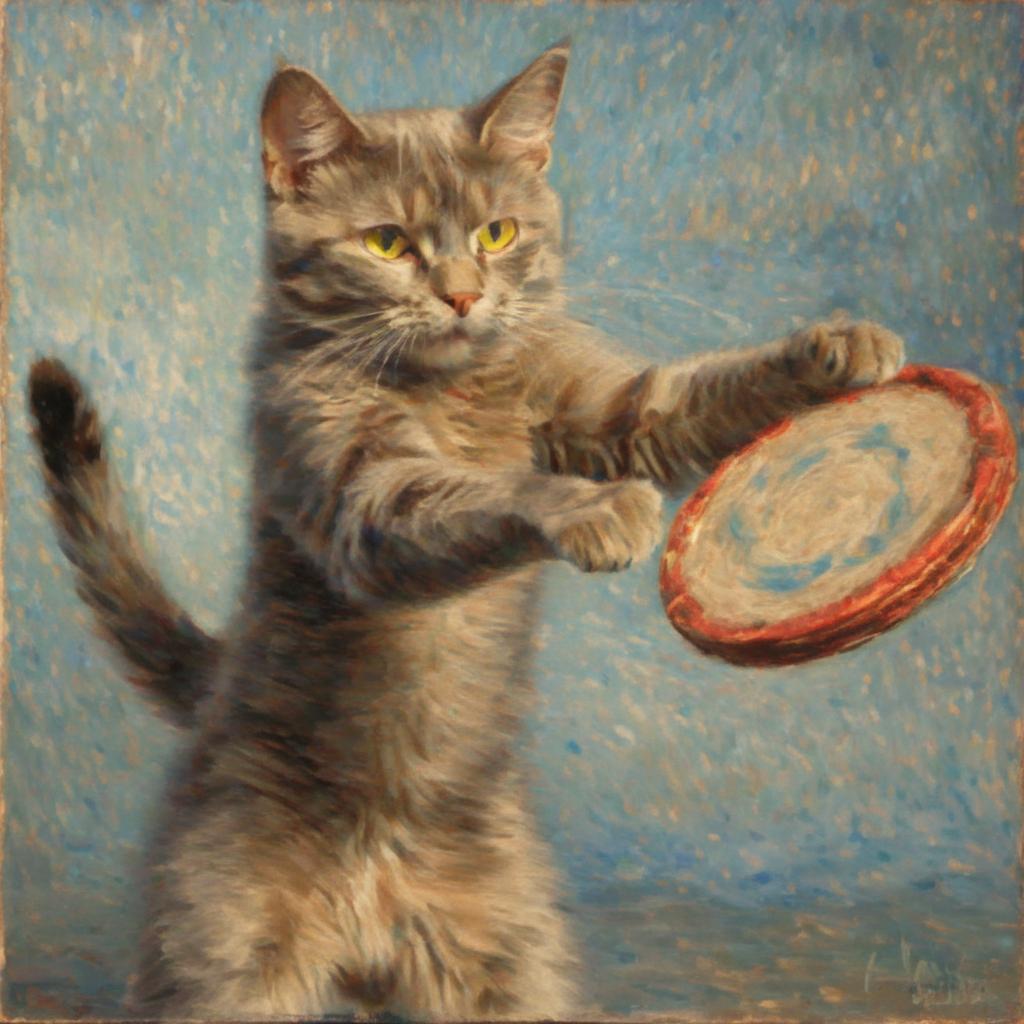} &
    \tabfigure{width=0.1\textwidth}{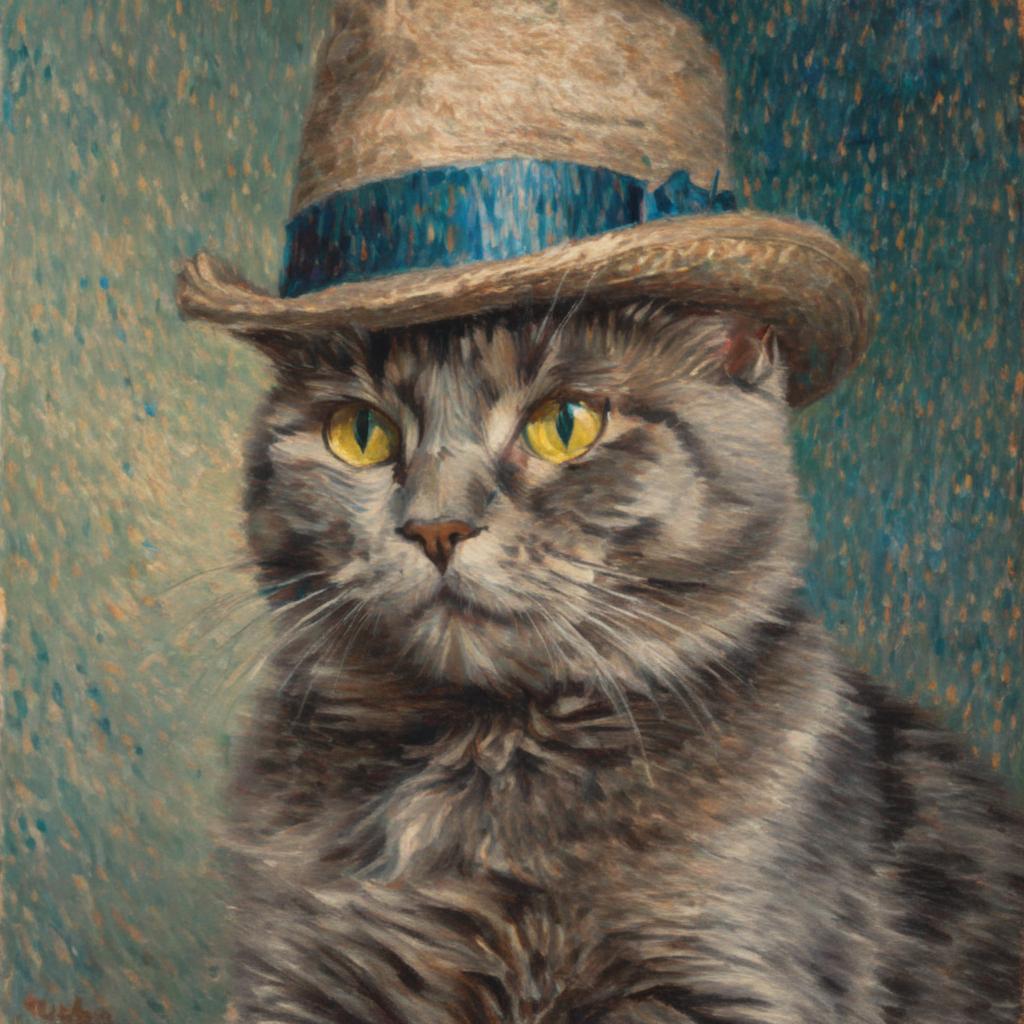} &
    \tabfigure{width=0.1\textwidth}{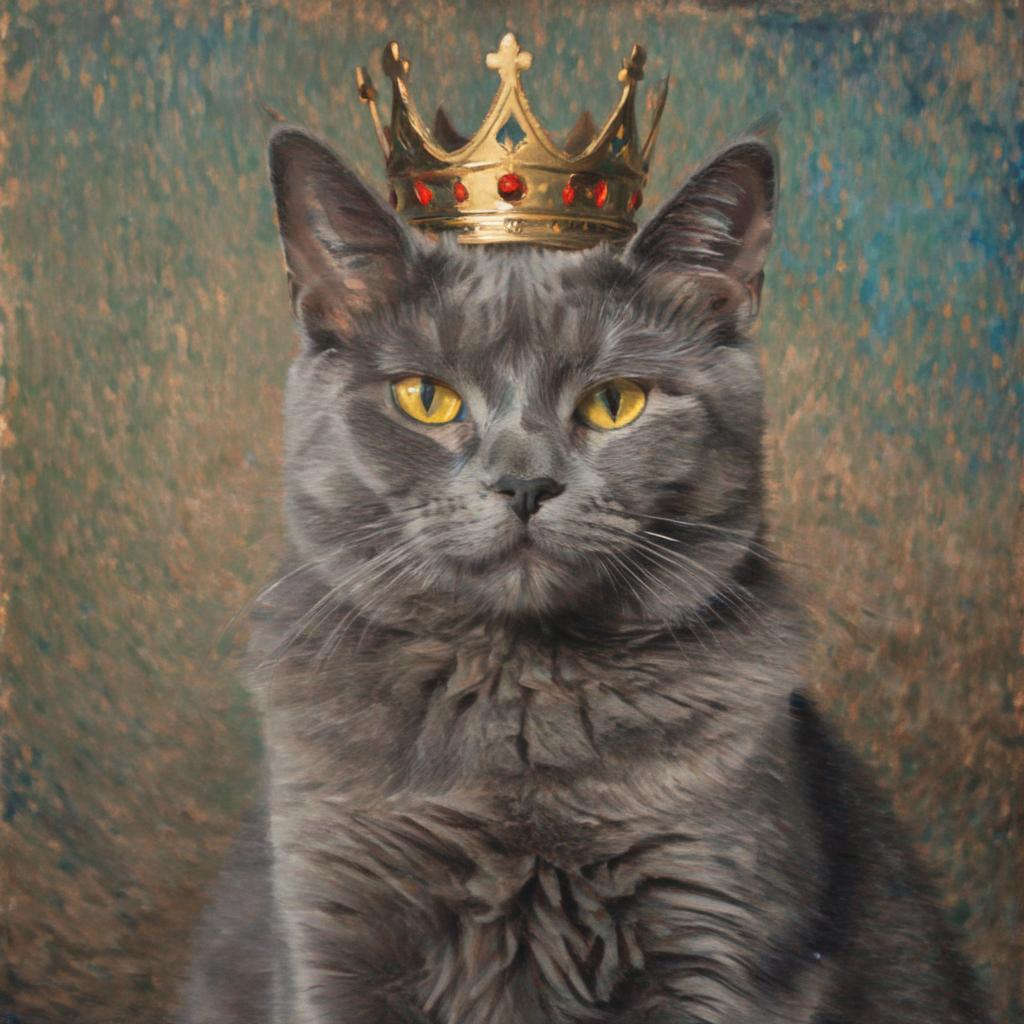} \\
     &&\small \dots riding a bicycle & \small \dots sleeping & \small \dots in a boat & \small \dots driving a car \\
    & & \tabfigure{width=0.1\textwidth}{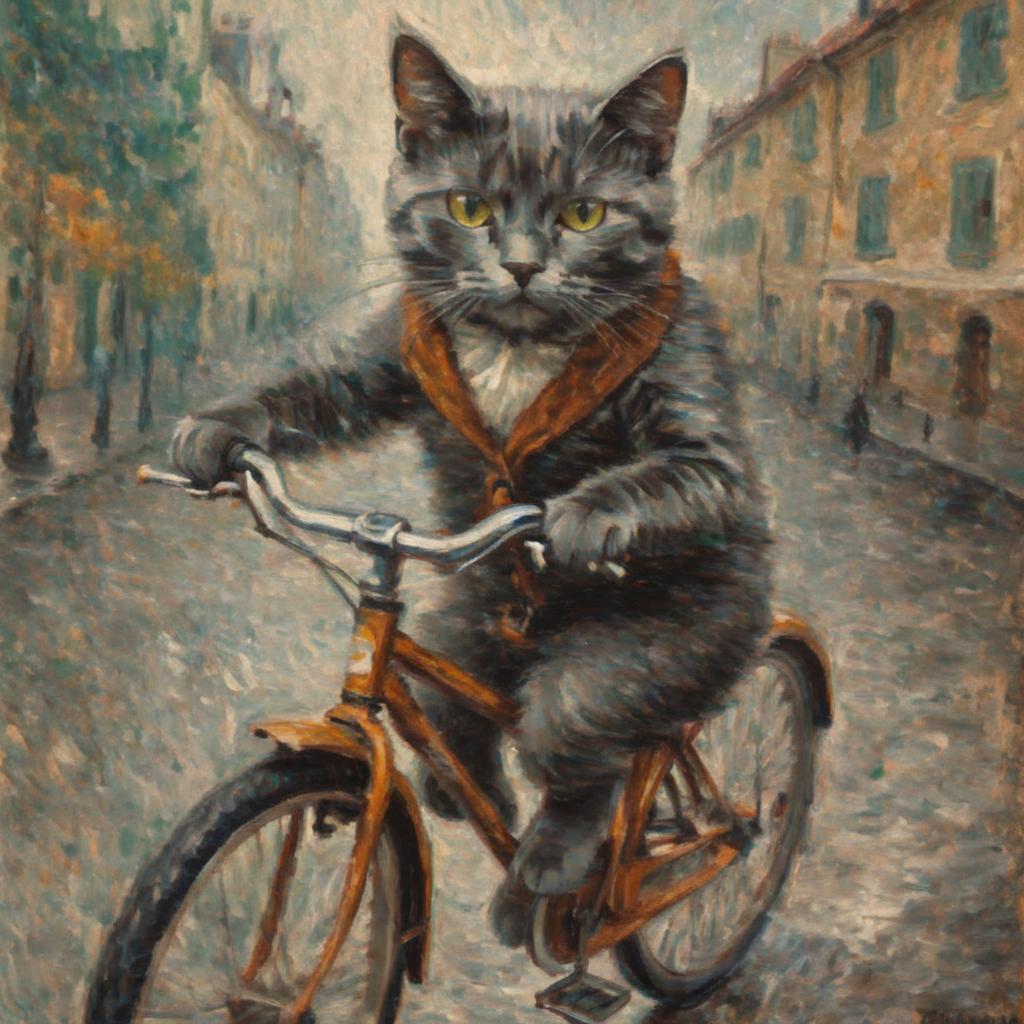} &
    \tabfigure{width=0.1\textwidth}{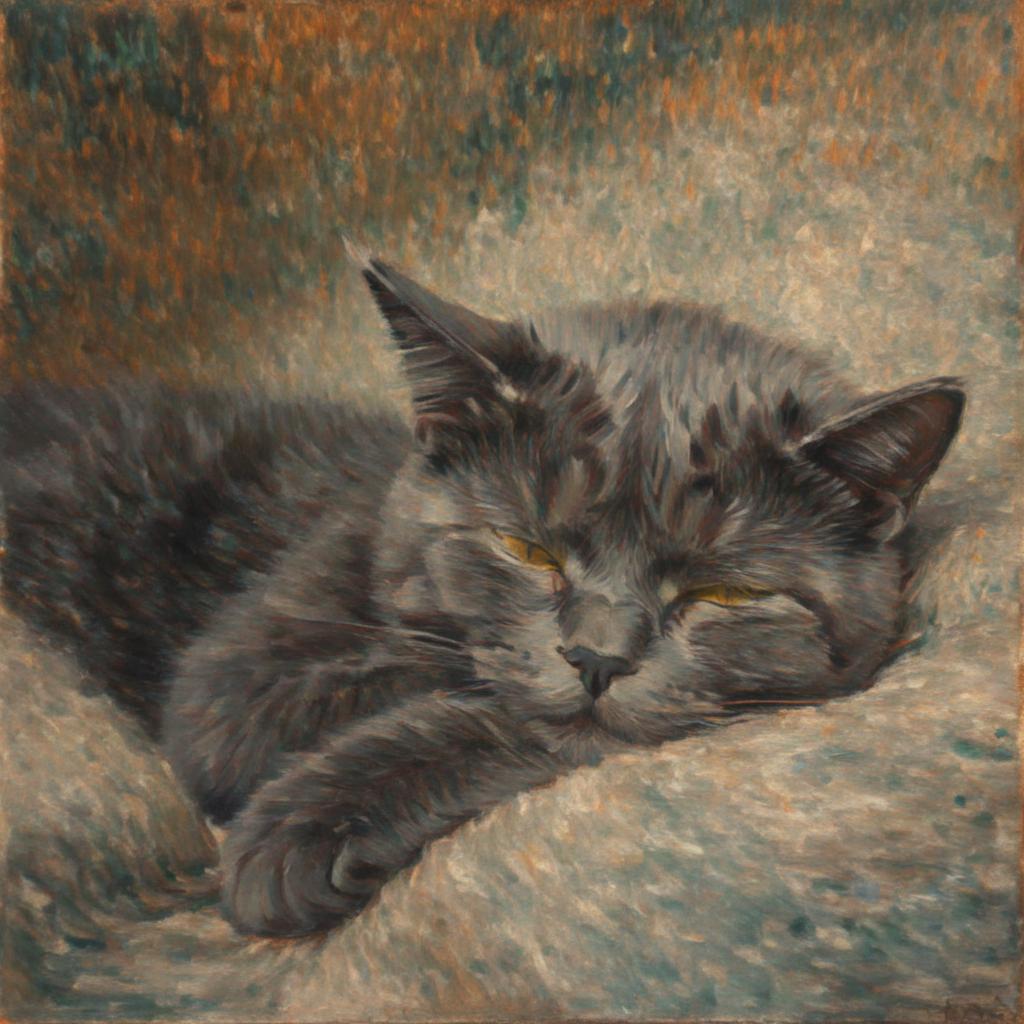} &
     \tabfigure{width=0.1\textwidth}{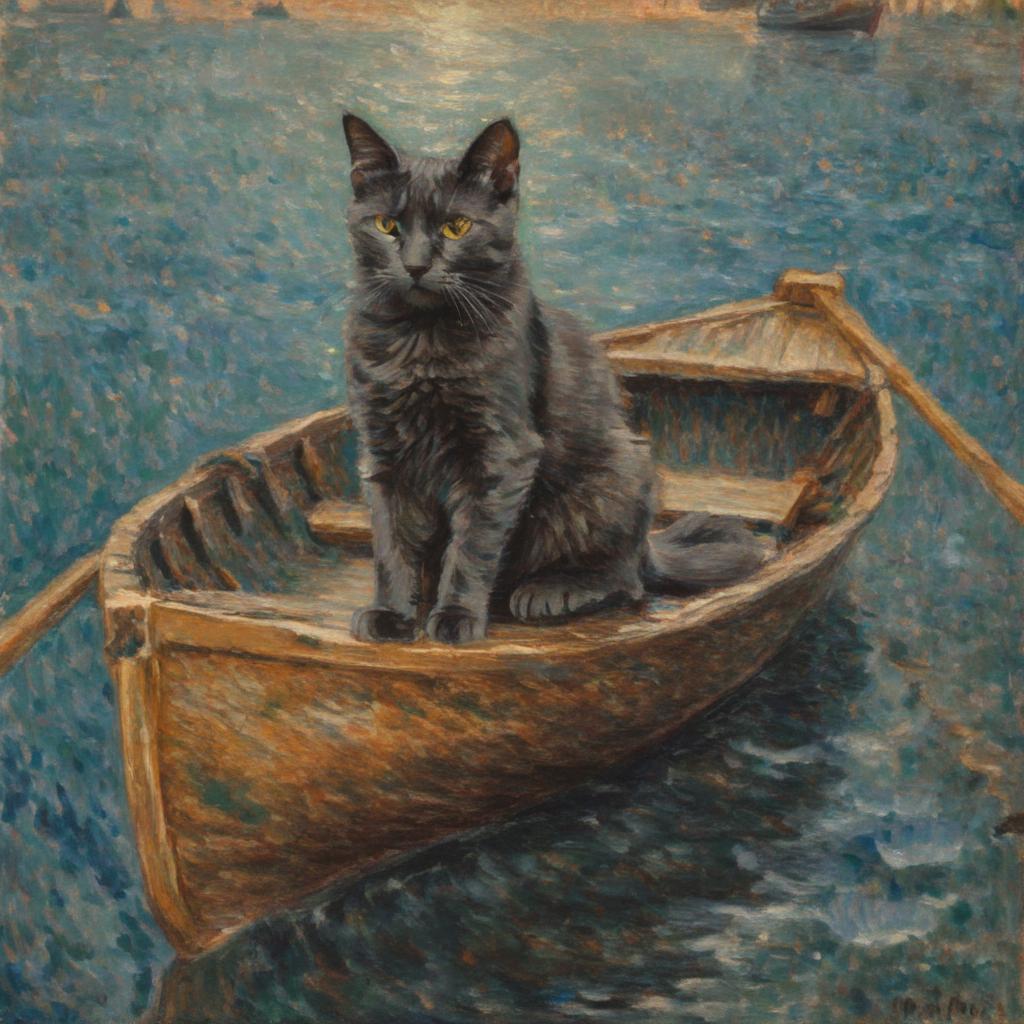} &
    \tabfigure{width=0.1\textwidth}{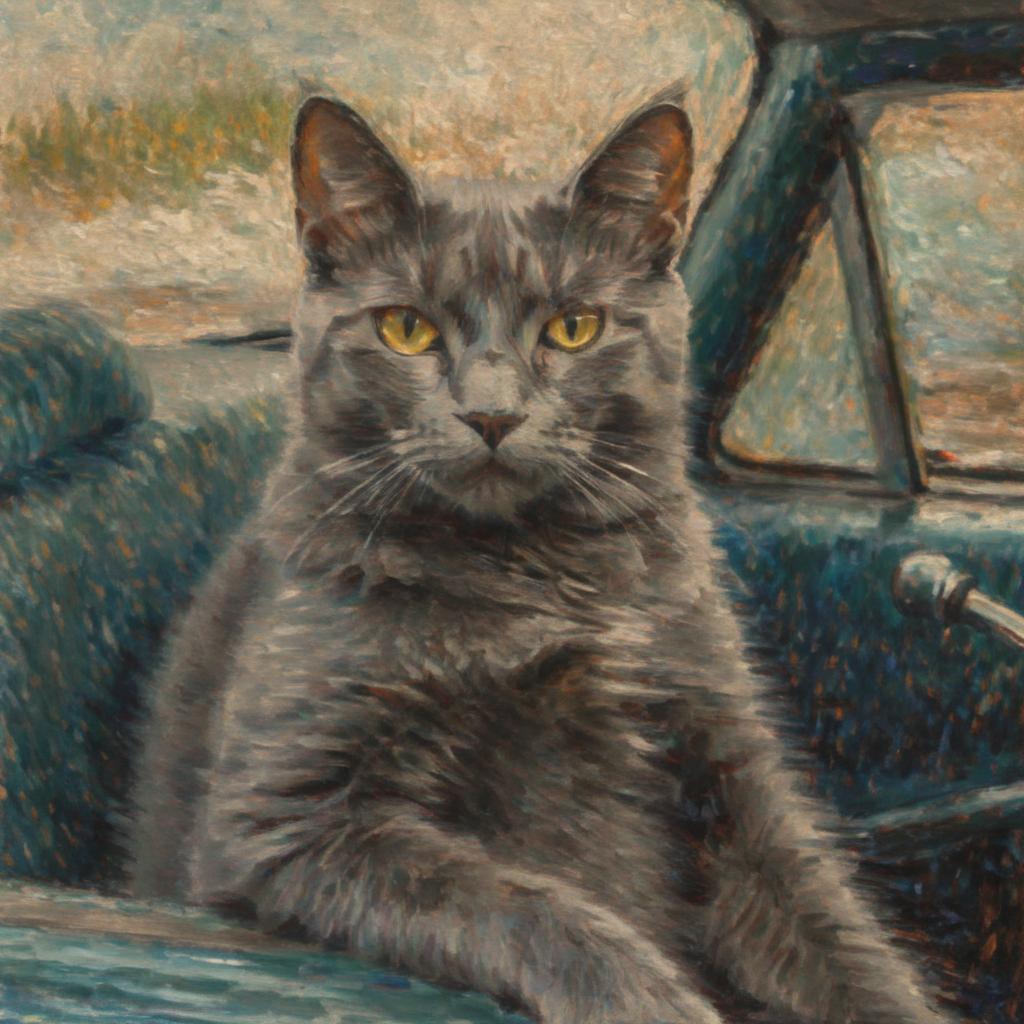}
    \end{tabular}
    }
    \caption{\textbf{Recontextualization Output Generations}. Generated outputs using various prompts for the contents ``dog8" and ``cat2''.}
    \label{fig:recontext_2}
\end{figure*}

In \cref{fig:recontext_1} and \cref{fig:recontext_2} we report a recontextualization analysis for different subjects and styles, demonstrating the effectiveness of our approach.

\begin{figure}[t]
    \centering
    \footnotesize
    \begin{tabular}{c c c}
        Subject & Style & \ours Output\\
        \tabfigure{width=0.12\textwidth}{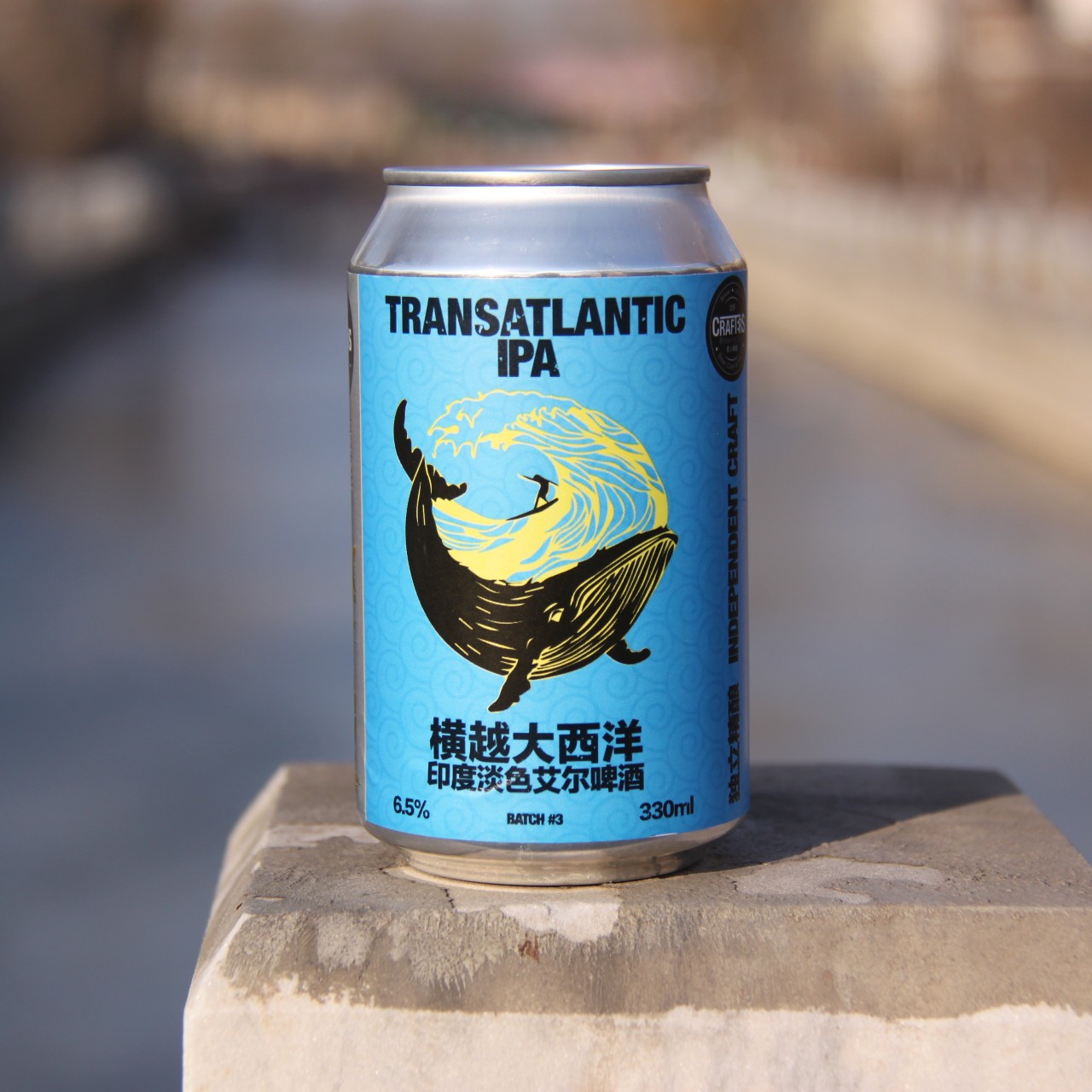} & \tabfigure{width=0.12\textwidth}{figures/qualitative/watercolor.jpg} & \tabfigure{width=0.12\textwidth}{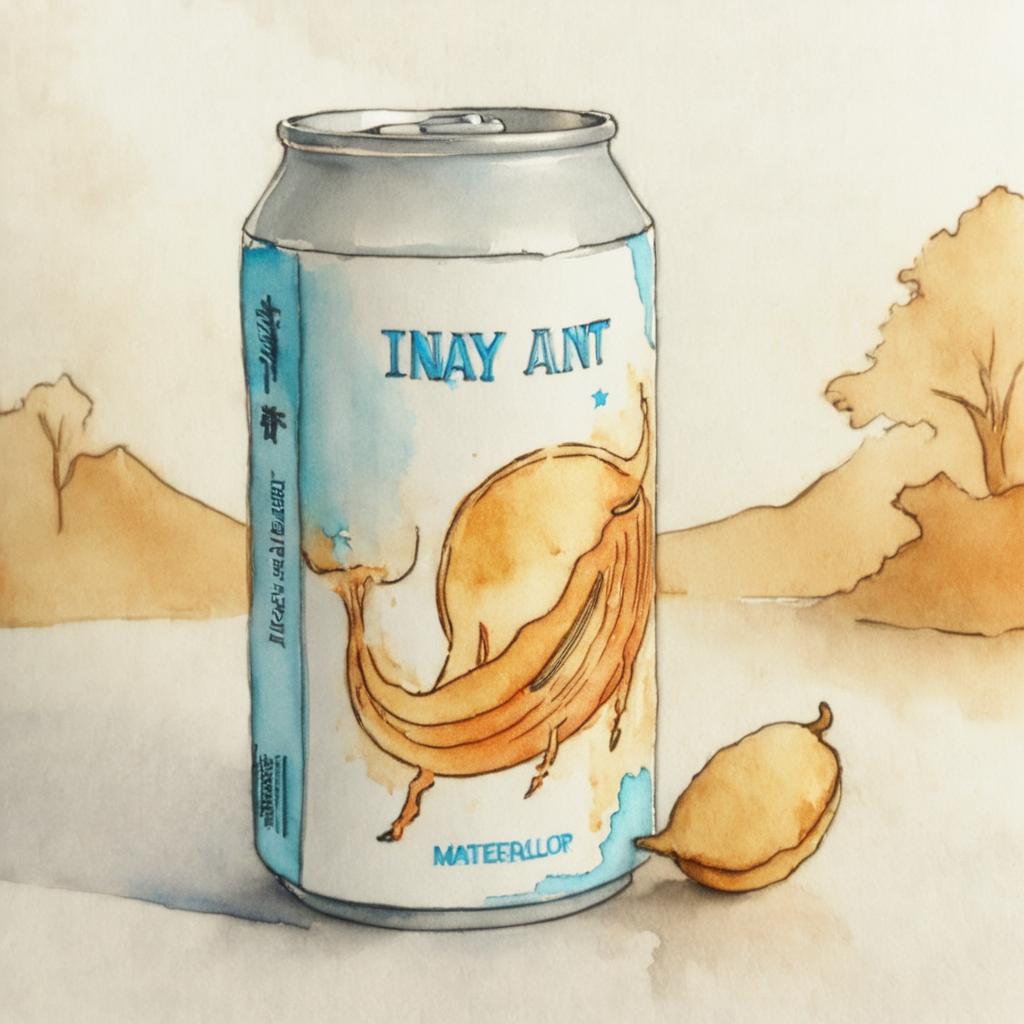}
    \end{tabular}
    \caption{\textbf{Limitation Example.} Example of a challenging generation case, where the generated text and logo are not accurate. }
    \label{fig:can_limitation}
\end{figure}

\section{Discussion}
\label{sec:supp:discussion}
\subsection{Limitations}
Our approach exhibits certain limitations with specific subjects, particularly the \textit{can}.
This limitation is shared by the other tested model merging methods as well. 
The \textit{can} subject is especially challenging because generative models struggle to accurately render text on objects (as we can see in \cref{fig:can_limitation}).

Furthermore, we note that while the MLLM judge is useful for the task of assessing generated images in terms of content and style, it is not perfect and, for example, it may overlook small details specific to the subjects.

\subsection{Societal Impact}

Our work makes it possible to generate personalized images that follow a given style and show a given subject, for example one's pet in watercolor painting style. In particular we make generating personalized images significantly more accessible than before as our solution can be deployed on smartphones, enabling real-time merging of LoRA parameters needed for the personalization. However, this brings risks that are shared with image generative models and image editing methods in general. These solutions can be used for creating deceptive content, and with our method it is even easier than before. Addressing the risks of misuse is an ongoing research priority in generative AI.

\end{document}